\documentclass[a4paper,12pt,twoside,custombib,customfont,print,index]{imartinezthesis}

\input{preamble/preamble}

\title{Diffeomorphic Transformations for Time Series Analysis: An Efficient Approach to Nonlinear Warping}

\subtitle{by}

\author{I\~nigo Martinez Lopez}

\dept{Department of Industrial Management}

\university{University of Navarra}
\crest{
\includegraphics[width=0.32\textwidth]{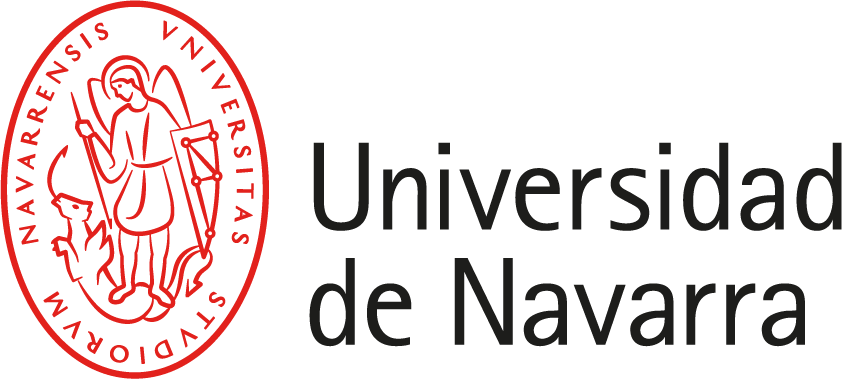}
\hspace{1cm}
\includegraphics[width=0.28\textwidth]{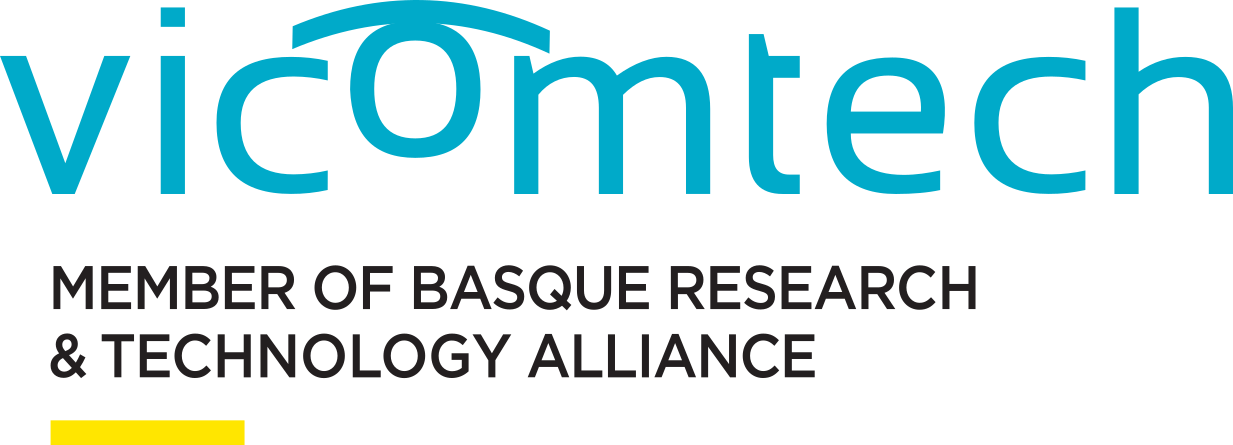}
}

\deptB{Department of Data Intelligence}
\universityB{Vicomtech, Basque Research and Technology Alliance}

\supervisor{Prof. Elisabeth Viles Díez}

\supervisorrole{}

\supervisorlinewidth{0.28\textwidth} 

\advisor{Dr. Igor García Olaizola}
     
\advisorrole{}

\advisorlinewidth{0.28\textwidth} 

\renewcommand{\submissiontext}{A thesis submitted for the degree of}
\degreetitle{Doctor of Philosophy}

\degreedate{July 2023} 

\ifdefineAbstract
\fi

\ifdefineChapter
\fi

\begin{document}

\frontmatter

\maketitle


\begin{dedication}

“ The distinction between past, present, and future is only a stubbornly persistent illusion. ”

\vspace{1em}

Albert Einstein

\end{dedication}


\begin{abstract}
\addcontentsline{toc}{section}{Abstract} 

The proliferation and ubiquity of temporal data across many disciplines has sparked interest  for similarity, classification and clustering methods specifically designed to handle time series data. 
A core issue when dealing with time series is determining their pairwise similarity, i.e., the degree to which a given time series resembles another. 
Traditional distance measures such as the Euclidean are not well-suited due to the time-dependent nature of the data.
Elastic metrics such as dynamic time warping (DTW) offer a promising approach, but are limited by their computational complexity, non-differentiability and sensitivity to noise and outliers. 

This thesis proposes novel elastic alignment methods that use parametric \& diffeomorphic warping transformations as a means of overcoming the shortcomings of DTW-based metrics.
The proposed method is differentiable \& invertible, well-suited for deep learning architectures, robust to noise and outliers, computationally efficient, and is expressive and flexible enough to capture complex patterns.
Furthermore, a closed-form solution was developed for the gradient of these diffeomorphic transformations, which allows an efficient search in the parameter space, leading to better solutions at convergence.

Leveraging the benefits of these closed-form diffeomorphic transformations, this thesis proposes a suite of advancements that include:
(a) an enhanced temporal transformer network for time series alignment and averaging, 
(b) a deep-learning based time series classification model to simultaneously align and classify signals with high accuracy, 
(c) an incremental time series clustering algorithm that is warping-invariant, scalable and can operate under limited computational and time resources, and finally, 
(d) a normalizing flow model that enhances the flexibility of affine transformations in coupling and autoregressive layers. 
Taken together, these advancements demonstrate the versatility and potential of closed-form diffeomorphic transformations for a range of time series applications.

In summary, this thesis aims to enhance time-series tasks such as alignment, averaging, classification and clustering by leveraging the power of fast, efficient, parametric \& diffeomorphic warping methods.

\end{abstract} 

\begingroup
\hypersetup{linkcolor=black}
\dominitoc%

\cleardoublepage
\phantomsection
\addcontentsline{toc}{section}{Table of Contents}
\tableofcontents

\cleardoublepage
\phantomsection
\addcontentsline{toc}{section}{List of Figures}
\listoffigures

\cleardoublepage
\phantomsection
\addcontentsline{toc}{section}{List of Tables}
\listoftables



\chapter*{Originality}
\addcontentsline{toc}{section}{Originality}

\section*{Statement}
\markboth{Statement}{Statement}

The writing of this thesis is my original work. The material in this thesis is either (a) my original work either with or without collaborators, or (b) where relevant prior or concurrent work included for reference, so as to provide a survey of the field.

\section*{Papers}
\markboth{Papers}{Papers}

This thesis contains material from the following papers: 

\newlength{\paperspacing}
\setlength{\paperspacing}{10pt}

\textbf{Closed-form diffeomorphic transformations for time series alignment}\\
I\~{n}igo Martinez, Elisabeth Viles, Igor Garcia Olaizola\\
\textit{International Conference on Machine Learning Data (ICML)}, 2022 \\
\url{https://proceedings.mlr.press/v162/martinez22a.html}

\vspace{\paperspacing}

\textbf{Elastic time series classification of rock drill pressure data}\\
I\~{n}igo Martinez, Elisabeth Viles, Igor Garcia Olaizola\\
\textit{Information Sciences} (In Review), 2023 

\vspace{\paperspacing}

\textbf{Normalizing flows based on diffeomorphic coupling functions}\\
I\~{n}igo Martinez, Elisabeth Viles, Igor Garcia Olaizola\\
\textit{Neurocomputing} (In Review), 2023 

\clearpage
Several other papers were published during the development of this thesis, which contributed to the doctoral student's formation and maturity as a researcher. (in chronological order):

\vspace{\paperspacing}

\textbf{A scalable framework for annotating photovoltaic cell defects in electroluminescence images}\\
Urtzi Otamendi, I\~{n}igo Martinez, Igor Garcia Olaizola, Marco Quartulli\\
\textit{IEEE Transactions on Industrial Informatics}, 2022, JCR 11.648, Q1 \\
\url{https://doi.org/10.1109/TII.2022.3228680}

\vspace{\paperspacing}

\textbf{A novel method for error analysis in radiation thermometry with application to industrial furnaces}\\
I\~{n}igo Martinez, Urtzi Otamendi, Igor Garcia Olaizola, Roger Solsona, Mikel Maiza, Elisabeth Viles, Arturo Fernandez, Ignacio Arzua\\
\textit{Measurement}, 2022, JCR: 5.131, Q1  \\
\url{https://doi.org/10.1016/j.measurement.2021.110646}

\vspace{\paperspacing}

\textbf{ArchABM: An agent-based simulator of human interaction with the built environment. CO\textsubscript{2} and viral load analysis for indoor air quality}\\
I\~{n}igo Martinez, Jan Lukas Bruse, Ane Miren Florez-Tapia, Elisabeth Viles, Igor Garcia Olaizola\\
\textit{Building and Environment}, 2022, JCR: 7.093, Q1 \\
\url{https://doi.org/10.1016/j.buildenv.2021.108495}

\vspace{\paperspacing}

\textbf{On the performance of shared autonomous bicycles: A simulation study}\\
Naroa Coretti Sanchez, I\~{n}igo Martinez, Luis Alonso Pastor, Kent Larson\\
\textit{Communications in Transportation Research}, 2022 \\
\url{https://doi.org/10.1016/j.commtr.2022.100066}

\vspace{\paperspacing}

\textbf{On the simulation of shared autonomous micro-mobility}\\
Naroa Coretti Sanchez, I\~{n}igo Martinez, Luis Alonso Pastor, Kent Larson\\
\textit{Communications in Transportation Research}, 2022 \\
\url{https://doi.org/10.1016/j.commtr.2022.100065}

\vspace{\paperspacing}

\textbf{A survey study of success factors in data science projects}\\
I\~{n}igo Martinez, Elisabeth Viles, Igor Garcia Olaizola\\
\textit{IEEE International Conference on Big Data}, 2021 \\
\url{https://doi.org/10.1109/BigData52589.2021.9671588}

\vspace{\paperspacing}

\textbf{Segmentation of cell-level anomalies in electroluminescence images of photovoltaic modules} \\
Urtzi Otamendi, I\~{n}igo Martinez, Marco Quartulli, Igor Garcia Olaizola, Elisabeth Viles, Werther Cambarau\\
\textit{Solar Energy}, 2021, JCR 7.188, Q2 \\ 
\url{https://doi.org/10.1016/j.solener.2021.03.058}

\vspace{\paperspacing}

\textbf{Data science methodologies: current challenges and future approaches}\\
I\~{n}igo Martinez, Elisabeth Viles, Igor Garcia Olaizola\\
\textit{Big Data Research}, 2021, JCR 3.739, Q1 \\ 
\url{https://doi.org/10.1016/j.bdr.2020.100183}

\vspace{\paperspacing}

\textbf{Labelling drifts in a fault detection system for wind turbine maintenance}\\
I\~{n}igo Martinez, Elisabeth Viles, I\~{n}aki Cabrejas\\
\textit{International Symposium on Intelligent and Distributed Computing}, 2018 \\
\url{https://doi.org/10.1007/978-3-319-99626-4_13}

\section*{Open source software}
\markboth{Open source software}{Open source software}

I have authored or otherwise had a substantial hand in developing:

\vspace{\paperspacing}
\textbf{DIFW - Diffeomorphic Fast Warping}\\
Diffeomorphic transformations based on the closed-form integration of continuous piecewise affine velocity functions. \\
\url{https://github.com/imartinezl/difw}

\vspace{\paperspacing}
\textbf{TSCLUST - Time Series clustering}\\
Incremental clustering of time series with elastic similarity measures\\
\url{https://github.com/imartinezl/tsclust}

\vspace{\paperspacing}
\textbf{PHMChallenge - Elastic time series classification of rock drill data}\\
Elastic time series classification of rock drill data\\
\url{https://github.com/imartinezl/phm-challenge}

\vspace{\paperspacing}
\textbf{DIFW-NF - Diffeomorphic Neural Flows}\\
Normalizing flows based on diffeomorphic coupling functions\\
\url{https://github.com/imartinezl/normalizing-flows}

\vspace{\paperspacing}
\textbf{ArchABM - Architectural Agent Based Modeling}\\
Agent-based simulator for air quality and pandemic risk assessment in architectural spaces\\
\url{https://github.com/Vicomtech/ArchABM}

\vspace{\paperspacing}
\textbf{Micro-mobility ABM}\\
Fleet simulation of MIT autonomous bicycle project\\
\url{https://github.com/CityScope/AutonomousMicroMobility}


\begin{acknowledgements}
\addcontentsline{toc}{section}{Acknowledgements} 

\newlength{\ackspacing}
\setlength{\ackspacing}{6pt}

I would like to express my gratitude to supervisors Elisabeth Viles Díez and Igor García Olaizola for their determined support, guidance, and encouragement throughout my Ph.D. journey. Their expertise, patience, and constructive feedback have been invaluable in shaping the direction and quality of this thesis.

\vspace{\ackspacing}

My sincere appreciation also goes to the staff of Vicomtech and the University of Navarra, for providing me with the resources, funding, and assistance necessary to conduct this research.

\vspace{\ackspacing}

I am also grateful to my department colleagues and peers for their valuable insights, stimulating discussions, and support throughout my doctoral program. You have all fostered a supportive and collaborative environment that has been essential to the success of this thesis.

\vspace{\ackspacing}

Last but not least, I would like to thank my family and friends for their encouragement and unwavering support throughout my academic journey. Their understanding, patience, and constant motivation have been instrumental in helping me overcome the many challenges I faced during my research work.

\vspace{\ackspacing}

To all those who have supported me, in ways big and small, I express my heartfelt gratitude.

\end{acknowledgements}

\endgroup


\mainmatter


\graphicspath{{content/chapter0/}}

\hypertarget{chapter0}{
\chapter{Introduction}
}\label{chapter:0}
\begingroup
\hypersetup{linkcolor=black}
\setstretch{1.0}
\minitoc
\endgroup

\clearpage
\section{State of Artificial Intelligence}\label{sec:state_ai}

Artificial intelligence (AI) technologies are increasingly becoming a part of our daily lives and are having a significant impact on a variety of industries. From self-driving cars and intelligent personal assistants to medical diagnosis and financial forecasting, these technologies are transforming the way we live and work, enabling businesses to automate processes, improve decision-making, and create new products and services. 

Similarly, AI is rapidly changing the way we approach research and development, and it is playing a crucial role in driving breakthroughs in fields such as:
language generation \cite{brown2020language}, 
fusion energy \cite{Degrave2022}, 
protein structure prediction \cite{hassabis_2022, Lin2022.07.20.500902, Nijkamp_2022},
genomic research \cite{doi:10.1126/science.abi6983},
materials science \cite{kirkpatrick2021pushing}, 
plastic recycling \cite{Lu2022},
reinforcement learning \cite{Baker_2022},
methodological errors in machine learning \cite{Kapoor_2022},
code generation \cite{Li_2022},
text-to-image generation \cite{Ramesh_2022,Saharia_2022}
mathematical hypothesis testing \cite{davies2021advancing}, 
representation theory \cite{blundell2022towards, davies2021signature},
matrix multiplication \cite{fawzi2022discovering},
and math problem-solving \cite{polu2022formal}, among others.

\subsection{Definitions}
To provide context for this thesis and prevent any semantic misunderstanding, we begin by reviewing some fundamentals:
\textbf{Artificial intelligence} (AI) is a multidisciplinary field of science and engineering whose goal is to create intelligent machines that simulate human intelligence, including the ability to learn, reason, and make decisions.
\textbf{Machine learning} (ML) is a subset of AI that involves the use of statistical techniques to enable machines to learn from data and improve their performance on a given task without being explicitly programmed. This process is known as \textit{training} a model using a learning algorithm, and it progressively improves model performance on a specific task.
\textbf{Deep learning} (DL) is an area of ML that attempts to mimic the activity in layers of neurons in the brain to learn how to recognize complex patterns in data. The \textit{“deep”} refers to the large number of layers of neurons in contemporary models that help to learn rich representations of data to achieve better performance gains.
\textbf{Big Data} refers to extremely large and complex datasets that cannot be processed and analyzed using traditional data processing tools and techniques. It is commonly defined by the 4 V's (see \cref{fig:big_data}): 
\textbf{Volume} alludes to the sheer size of the datasets, which can be in the terabytes, petabytes, or even exabytes, orders of magnitude larger than those typically used in data analysis and decision-making. 
The vast size of these datasets poses challenges for data storage, management, and analysis.
\textbf{Velocity} relates to the speed at which data is generated, collected, and processed. In today's digital world, data is generated dynamically at an unprecedented rate, with billions of transactions occurring every day and streaming data from sensors or social media feeds. 
\textbf{Variety} refers to the diverse types of data that are generated: structured data from databases, unstructured data from text documents and social media posts, and semi-structured data such as XML and JSON.
\textbf{Veracity} is a measure of the quality and reliability of data. As the volume of data increases, it becomes increasingly difficult to ensure its quality and relevance and organizations must implement effective data governance and quality control processes to ensure the veracity of their data.

\def\rA{3cm}
\def\rB{1.1 * \rA}
\def\s{1mm}

\begin{figure}[t]
  \begin{center}
  \scalebox{0.75}{
\begin{tikzpicture}[
    thin, every path/.style={rounded corners=0.1},
    direction/.style={->,shorten >=2mm,shorten <=2mm},
  ]

  \begin{scope}[shift={(\s,\s)}]
    \draw [fill=OrangeRed!20] (0:\rA) arc(0:90:\rA) |- cycle;
    \draw [fill=OrangeRed!50] (0:\rA) arc(0:60:\rA) -- (0,0) -- cycle;
    \node at (25:0.6*\rA) {\textbf{Veracity}};
    \draw[direction] (0:\rB) arc(0:90:\rB) node[pos=0.05, above right] {high variance} node[pos=0.95, above right] {reference data};
  \end{scope}

  \begin{scope}[shift={(-\s,\s)}]
    \draw [fill=RoyalBlue!20] (90:\rA) arc(90:180:\rA) -| cycle;
    \draw [fill=RoyalBlue!50] (90:\rA) arc(90:120:\rA) -- (0,0) -- cycle;
    \node at (150:0.6*\rA) {\textbf{Volume}};
    \draw[direction] (90:\rB) arc(90:180:\rB) node[pos=0.05, above left] {kilobytes} node[pos=0.95, above left] {terabytes};
  \end{scope}

  \begin{scope}[shift={(-\s,-\s)}]
    \draw [fill=Dandelion!20] (180:\rA) arc(180:270:\rA) |- cycle;
    \draw [fill=Dandelion!50] (180:\rA) arc(180:190:\rA) -- (0,0) -- cycle;
    \node at (230:0.6*\rA) {\textbf{Velocity}};
    \draw[direction] (180:\rB) arc(-180:-90:\rB) node[pos=0.05, below left] {static} node[pos=0.95, below left] {dynamic};
  \end{scope}

  \begin{scope}[shift={(\s,-\s)}]
    \draw [fill=JungleGreen!20] (270:\rA) arc(270:360:\rA) -| cycle;
    \draw [fill=JungleGreen!50] (270:\rA) arc(270:290:\rA) -- (0,0) -- cycle;
    \node at (330:0.6*\rA) {\textbf{Variety}};
    \draw[direction] (270:\rB) arc(-90:0:\rB) node[pos=0.05, below right] {clustered} node[pos=0.95, below right] {heterogeneous};
  \end{scope}

\end{tikzpicture}
}
\caption{Big Data 4V's: Volume, Velocity, Variety and Veracity}
\label{fig:big_data}
\end{center}
\end{figure}
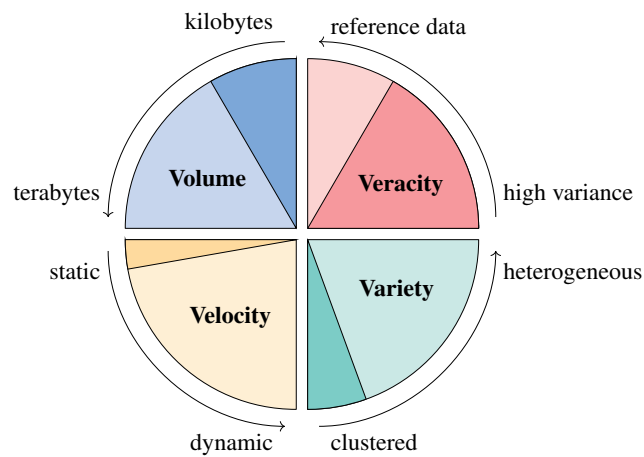

\section{Ubiquity of Time Series Data}\label{sec:ubiquity_time_series}

\begin{figure}[!htb]
  \begin{center}
  \includegraphics[width=0.8\linewidth]{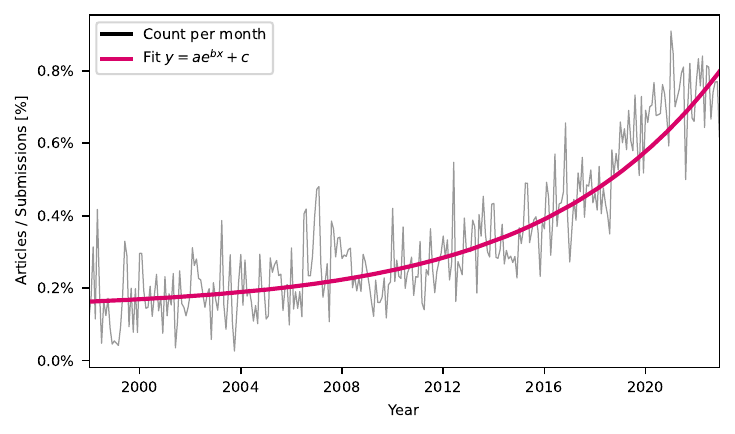}
  \caption{Proportion of article titles containing "time series" on arXiv open-access archive. We use nonlinear least squares to fit an exponential function $y(x)=a * e^{bx} + c$ to monthly data, where $x$ represents the month number starting at 1998 and $a=0.01$, $b=4.23$ and $c=0.14$ are the parameters that minimize the sum of squares of the residuals. The function grows rapidly since $b$ is positive and will be shifted upward since $c$ is positive. Data extracted using arXiv public API \cite{arxiv}.}
  \label{fig:arxiv}
  \end{center}
\end{figure}

Time series data, which is a sequence of data points indexed in time order, are ubiquitous in today's data-driven world. It is often used to forecast future trends and patterns and can be found in almost scientific discipline and business application, from finance and economics to healthcare and technology.
Recent times have seen an explosion in the volume and prevalence of time series data.
Time is a fundamental constituent of everything that is observable so as our world becomes increasingly instrumented, sensors and systems are constantly emitting a relentless stream of time series data.
Whether it be stock market fluctuations, sensor data recording climate change, or activity in the brain, any signal that changes over time can be described as a time series.
This ubiquity has spread to the research field as well, where the number of articles on time series has increased in recent years. In this regard, \cref{fig:arxiv} shows the proportion of article titles containing "time series" on arXiv open-access archive.

\subsection{Applications}\label{sec:applications_time_series}

Time series data has long been a staple in industrial applications, providing valuable insights into trends, patterns, and changes over time. Whether companies design their applications to include time series data from the outset or retrofit time series analysis into established, preexisting applications, the industries and use cases for time series data vary widely.

In \textbf{healthcare}, time series data is used to analyze patients' heartbeat electrocardiograms (ECGs) and brain activity \cite{rathi2022analysis}, as well as to study human DNA sequences \cite{beng2021temporal}. 
Through remote sensing, \textbf{meteorologists} frequently rely on time series data to track weather changes and daily temperatures \cite{zeng2020review}.
\textbf{Geologists} employ time series data to make earthquake predictions \cite{mir2021anomaly}. 
\textbf{Engineers} rely on time series data to analyze the trajectories of moving objects, recognize human activity \cite{qin2020imaging}, and optimize Internet of Things (IoT) systems \cite{cook2019anomaly}. 
Regarding social sciences, \textbf{demographers} exploit time series data to study birth and mortality rates \cite{perla2021time}. 
\textbf{Marketing} professionals also manipulate time series data to analyze weekly sales totals, web usage sequences, and other trends \cite{wang2022applied}. 
\textbf{Financial} service companies utilize time series data to analyze market fluctuations, transaction anomalies, prices of mutual funds, and to perform stock market analysis and economic forecasting \cite{sezer2020financial}. 
\textbf{Streaming services} use time series data to identify and prevent issues before they impact end users \cite{dinaki2021forecasting}, and
\textbf{telecommunications carriers} often use time series data for anomaly detection, network telemetry and capacity planning \cite{jadon2021challenges}.

In the manufacturing sector, for example, ML is helping to drive \textbf{industry 4.0}, i.e., the integration of digital technologies into manufacturing processes \cite{diez2019data}.
This integration is enabling businesses to create smarter, more efficient factories that are able to adapt to changing market conditions and customer needs.
In many manufacturing environments, production processes are complex and often involve numerous variables. By using time series data, manufacturers can develop systems that can analyze the trends and evolution of production data and identify potential areas for improvement \cite{essien2020deep}. For example, an AI system might be able to identify bottlenecks in a production line, or suggest changes to production schedules that could improve efficiency, quality and reduce waste and downtime, improving their overall competitiveness. 

Another use of time series data in industry 4.0 is through \textbf{predictive maintenance} \cite{coelho2022predictive}. In traditional manufacturing environments, equipment is typically inspected and maintained on a regular schedule, regardless of its condition. This approach is not only inefficient, but it can also be costly, as it requires the allocation of resources for maintenance even when they are not needed.
With time series data, however, it is possible to develop predictive maintenance systems that can identify trends and patterns and forecast the likelihood of equipment failures or maintenance needs, allowing manufacturers to focus their resources on only those items that truly need attention.
This allows businesses to proactively schedule maintenance and repair work, reducing downtime and improving overall equipment reliability and performance.

\subsection{Explosion of Time Series Data}

The rapid expansion of time series data use in recent years has been closely linked to several key developments, which are discussed below.

\vspace{-1.0em}
\paragraph{Data collection} 
The proliferation of affordable and reliable interconnected sensors, the Internet of Things and other data-gathering devices (smartphones, smartwatches, and other wearables), has allowed collecting large amounts of data on various phenomena over time. 
As new devices come online, they generate an ever-increasing amount of time series data. Imagine a common metric from a connected car, like speed in kilometers per hour. If we change the time interval for collecting this info to every minute or second, the resulting dataset is orders of magnitude larger. Apply this principle across an entire industry of millions of connected devices and sensors, and the exponential growth of this data becomes obvious.

This growth has also significantly changed the way statistical analysis is conducted. Previously, the high cost of data collection meant that statisticians had to carefully select data in order to answer a specific question. However, the proliferation of these sensors, along with the widespread adoption of the internet, has led to an explosion of machine-generated data. 

\vspace{-1.0em}
\paragraph{Computer performance}
Advances in computing power (both in CPUs and GPUs) and data storage capabilities have made it possible for organizations to process and analyze larger and more complex datasets.
Specially the latest advances in GPUs have pushed forward the expansion of deep learning techniques, very eager of fast matrix operations \cite{shi_benchmarking}.

\vspace{-1.0em}
\paragraph{Distributed and cloud computing} 
Ecosystems such as Hadoop \cite{hou2019time} and Spark \cite{zaharia2016apache} can process data in parallel across multiple machines, significantly increasing the speed and efficiency of data analysis.
In addition, cloud-based solutions, such as Amazon Web Services and Google Cloud Platform, provide organizations with scalable and flexible data storage and processing capabilities without the need for expensive on-premises infrastructure, making it easier and more cost-effective for organizations to access and use big data technologies.

\vspace{-1.0em}
\paragraph{Advanced ML algorithms} 
The increasing popularity of machine learning algorithms has made it easier to analyze time series data. Machine learning algorithms are designed to automatically identify patterns in data and make predictions based on those patterns. This has greatly enhanced the ability of researchers to extract useful insights from time series data, making it a valuable tool for a wide range of applications.

\vspace{-1.0em}
\paragraph{Data-driven decisions}
The growth of time series data has also been fueled by an increasing recognition of its value \cite{martinez2021data}.
Organizations are now using data to inform strategic decisions, optimize operations, and improve customer experiences. 
For example, a retailer might use time series data to identify seasonal trends in customer demand or product usage, or a manufacturer might use time series data to monitor the performance of its production line.
Many researchers and organizations have come to realize the potential of time series data to provide valuable insights into complex phenomena. This has led to a greater demand for time series data and has contributed to its ubiquity.

\vspace{-1.0em}
\paragraph{Open data initiatives}
Governments and organizations around the world have begun to make their time series data publicly available, allowing researchers to access and analyze data that would have previously been difficult or impossible to obtain. This has greatly expanded the scope of research that can be conducted using time series data. 

In conclusion, the growth of time series data has been driven by advances in technology, the use of machine learning algorithms, an increasing recognition of its value and the availability of open data. 
These factors have created a demand for efficient methods of analyzing and exploiting large and complex time series datasets, a demand that is explored in this thesis. 

\section{Thesis Motivation \& Challenges}

The proliferation and ubiquity of temporal data across many disciplines has generated substantial interest in the analysis and mining of time series.
For instance, time series classification and clustering are crucial tasks in many applications (see \cref{sec:applications_time_series}).
In classification, the goal is to train a model that can accurately predict the class of a time series, given a dataset with labeled time sequences. With clustering, we can identify and summarize interesting patterns and correlations in the underlying data.

As such, there is a growing need for the development of similarity, classification and clustering methods specifically designed for the analysis of time series data. This section provides a brief overview of the identified challenges that arise when analyzing time series data and sets the stage for the subsequent chapters.

\subsubsection*{Challenge 1: Differentiable and efficient similarity measures under time warping}\label{sec:challenge_1}

A core issue when dealing with time series is determining their pairwise similarity, i.e., the degree to which a given time series resembles another. 
Time series data recurrently appears misaligned or warped in time despite exhibiting amplitude and shape similarity, which impacts the design of similarity metrics. 
Temporal misalignment, often caused by differences in execution, sampling rates, or the number of measurements, confounds statistical analysis to the extent that even the sample mean of semantically-close time series can obfuscate actual peaks and valleys, generate non-existent features and render the analysis almost meaningless. With time warping, insignificant differences in the time axis may appear as very significant differences in the ordinate axis.

\begin{figure}[!htb]
  \begin{center}
  \includegraphics[width=\linewidth]{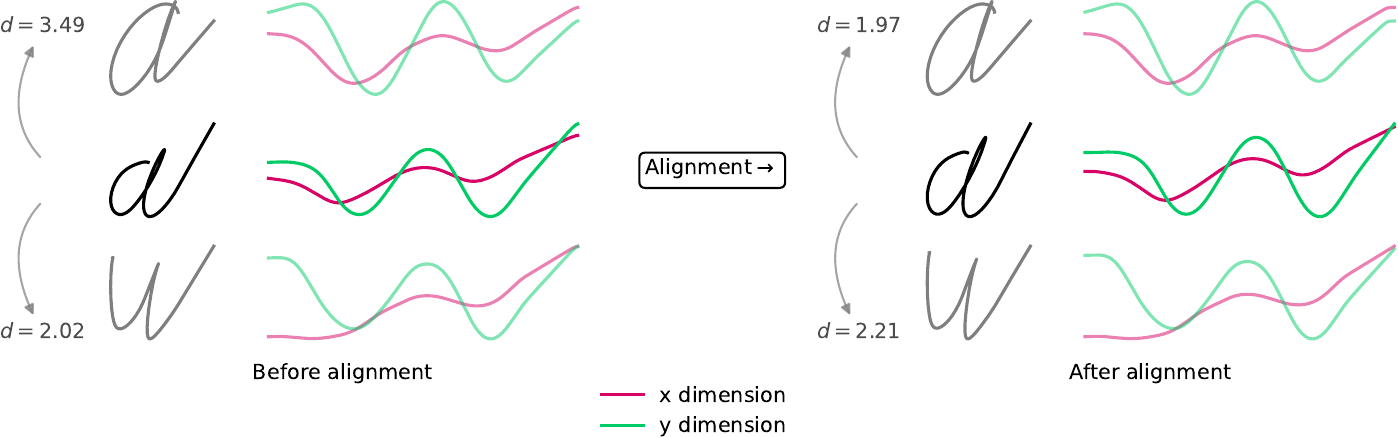}
  \caption{Euclidean vs elastic distance on multivariate (\textcolor{Magenta}{$\mathbf{x}$},\textcolor{Green}{$\mathbf{y}$}) time series from UCR archive \textit{CharacterTrajectories} dataset \cite{dau2019ucr}. Euclidean (\textcolor{Sepia}{elastic}) distance between letters of the same class is larger (\textcolor{Sepia}{smaller}) than between distinct classes. Class $a$ vs $w$.}
  \label{fig:character_trajectories_0}
  \end{center}
\end{figure}

Examples of temporal misalignment are illustrated in \cref{fig:character_trajectories_0}, where we analyze two-dimensional, multivariate time series dataset from UCR archive \cite{dau2019ucr} that represent character trajectories. We compare the letters $a$ vs $w$ before and after elastic alignment. 
Computing the Euclidean distance on the original data (left) shows that the distance between letters of the same class ($d=3.49$) is larger than between distinct classes ($d=2.02$). On the contrary, computing the elastic distance after data alignment (right) fixes this issue, and the distance between letters of the same class ($d=1.97$) is smaller than between distinct classes ($d=2.21$). 
Note that even though the visual differences between before and after alignment are minimal, the impact on the similarity measure is significant.

Therefore, traditional distance measures such as the Euclidean are not well-suited for time series similarity, due to the time-dependent nature of the data.
Elastic metrics such as dynamic time warping (DTW), which are thoroughly reviewed in \textbf{\cref{chapter:1}}, offer a promising approach, but are limited by their computational complexity, non-differentiability and sensitivity to noise and outliers.
To overcome the limitations of DTW-based methods, this thesis proposes \textbf{novel parametric alignment techniques that are well-suited for deep learning} methods. These alignment methods should be differentiable, robust to noise and outliers, computationally efficient, and expressive and flexible enough to capture complex patterns in the data.

\vspace{-2.0em}
\subsubsection*{Challenge 2: Efficient parametric warping functions suitable for deep learning}\label{sec:challenge_2}

In the context of time series alignment, a warping function is a transformation that aligns two series by mapping the time indices of one series to the time indices of another series. A warping function must be strictly monotone with a strictly positive first derivative. 

Non-parametric warping functions, such as the paths derived from DTW, are typically more flexible than parametric warping functions, but are more difficult to work with due to the variable number of parameters and the computational resources required to compute them. 
Parametric warping functions, on the other hand, have a fixed number of parameters that can be learned from the data. Examples of parametric warping functions include linear, polynomial, and spline functions, which are typically easier to work with due to their well-defined form and fixed number of parameters. However, they may not be as flexible as non-parametric functions in capturing complex relationships between the time indices. 

We are seeking warping functions methods that are differentiable, robust to noise and outliers, computationally efficient, and expressive and flexible enough to capture complex patterns in the data. In order to comply with such requirements, in \cref{chapter:2} we propose \textbf{fast and efficient one-dimensional diffeomorphic transformations}, which are functions that are differentiable, invertible and have a differentiable inverse. 

\subsubsection*{Challenge 3: Warping invariant and generalizable time series averaging}\label{sec:challenge_3}

Time series averaging with nonlinear time warping presents a complex optimization problem (\textit{NP}-hard). The time complexity of finding an optimal alignment raises exponentially when the input number of sequences to align increases. Additionally, the number of possible warping functions grows exponentially with the length of the time series. 

In response, researchers have proposed various heuristic algorithms that aim to find a near-optimal solution in a reasonable amount of time. In this regard, we propose to use the power of \textbf{gradient descent methods and deep learning} training (\textit{backpropagation}) to minimize the alignment loss function.

Finding the optimal warping function for pairwise or joint alignment requires solving an optimization problem. 
However, aligning a batch of time series data is insufficient because new optimization problems arise as new data batches arrive. 
In this work we incorporate closed-form diffeomorphic warping functions to Temporal Transformer Networks (TTN) such as \cite{Weber2019,Lohit2019,Nunez2020,Huang2021} that can generalize  \textbf{inferred alignments from the original batch} to the new data without having to solve a new optimization problem each time. 

A related challenge is the sensitivity of the warping function to noise and outliers in the time series data. Since the warping function is designed to align the two time series as closely as possible, any noise or outliers in the data can significantly affect the warping function and the similarity measure. This can lead to poor performance and unreliable results, especially in applications where the data is noisy or has high variability. We address this issue by including  \textbf{regularization to the alignment loss function}.

Furthermore, different similarity measures can result in divergent warping functions and average results, making it difficult to compare the performance of different algorithms or to evaluate the quality of the warping function. 
Given this lack of ground truth for the latent warps in real data, we use a simple classification model (nearest centroid) as a proxy metric for the quality of the joint alignment and the average signal. Extensive experiments are conducted on 84 datasets from the UCR archive \cite{dau2019ucr} to validate the  \textbf{generalization ability of the model to unseen data} for time series joint alignment.

\subsubsection*{Challenge 4: Scalable shape-based time series clustering}\label{sec:challenge_4}

Clustering methods can \textbf{partition data into homogeneous groups}, enabling a more detailed analysis of the data structure, while also aggregating and summarizing the data into a set of small, meaningful, and manageable representatives.
When applied to time series data, these methods must contend with several challenges, including but not limited to: 
(a) large-scale datasets \& each time series' high dimensionality can lead to expensive computational and memory-intensive solutions
(b) pairwise distances between time series need to account for the specificity of the time dimension, 
(c) dynamic and non-stationary processes often generate non-stationary distributions, to which models should adapt. Current incremental clustering algorithms have shown some promise in addressing these challenges, but there is still room for improvement in explicitly capturing the temporal dependencies.

In this work we propose a novel \textbf{incremental clustering algorithm} for time series data that assigns each incoming time series into the nearest cluster using a combination of elastic alignment and incremental clustering techniques. 
The algorithm is warmed-up through an offline process, and in the online phase, new time series instances are added incrementally to existing clusters. The assignment decision is based on the \textbf{diffeomorphic elastic distance} of the new point to the existing clusters. 
When the query time series does not match with any of the existing clusters, it is allocated to a new temporary group that can be updated with more incoming data. 
The performance of the proposed algorithm was evaluated on several datasets and compared it with state-of-the-art clustering methods for time series data. Results show that the proposed algorithm outperforms existing ones in terms of \textbf{clustering quality and scalability}.

\subsubsection*{Challenge 5: Efficient and flexible coupling functions for normalizing flows}\label{sec:challenge_5}

Normalizing flows \cite{rezende2015variational}, also known as flow-based generative models, use a sequence of invertible transformations $f$ to transform a simple distribution (such as a Gaussian) into a more complex target distribution.
The key defining property of flow-based models is that the transformations must be invertible and both $f$ and $f^{-1}$ must be differentiable.
Normalizing flows based on coupling layers require a bijective one-dimensional function $h(z)$ and the derivative of the function with respect the input variable $z$, i.e. $\frac{\partial h}{\partial z}$.
Related \textbf{flows based on coupling layers} such as NICE \cite{dinh2014nice} and RealNVP \cite{dinh2016density} have an analytic one-pass inverse, but are often less flexible than their autoregressive counterparts. 
Based on these limitations, in this work we propose to implement the coupling function $h$ using the \textbf{integration of continuous piecewise-affine (CPA) velocity functions} as a building block. 

The module acts as a \textbf{drop-in replacement for the affine or additive transformations} commonly found in coupling and autoregressive transforms. Unlike the additive and affine transformations, which have limited flexibility, the proposed differentiable monotonic function with sufficiently many intervals can approximate any differentiable monotonic function, yet has a closed-form, tractable Jacobian determinant, and can be inverted analytically. Our parameterization is fully-differentiable, which allows for training by gradient methods.

\section{Thesis Structure}\label{sec:thesis_structure}

\begin{figure}[!htb]
  \begin{center}
    \includegraphics[width=0.95\linewidth]{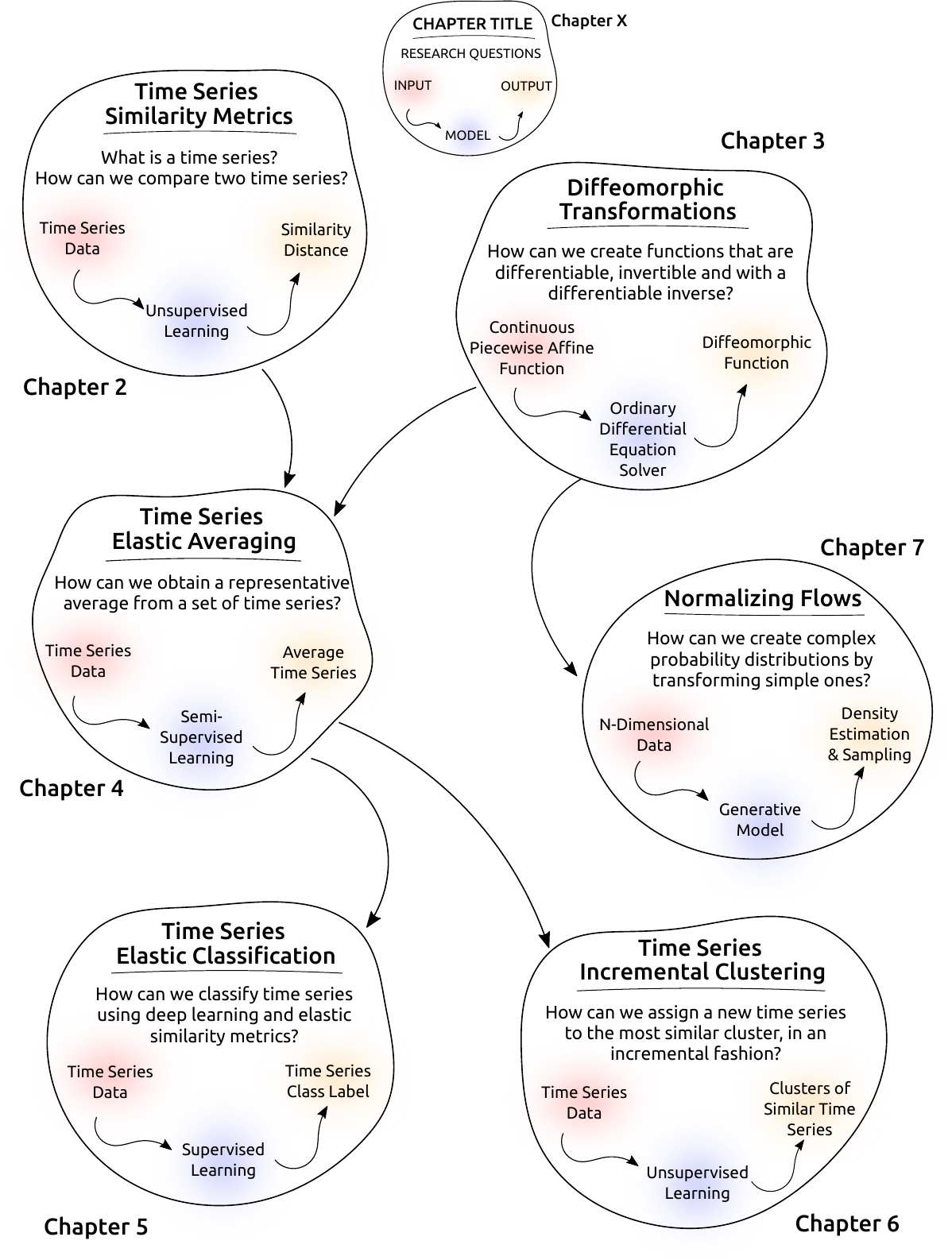}
    \Put(-173,261){ \hyperlink{chapter5}{\tikz \fill [opacity=0.05, rounded corners=11, black] (0,0) rectangle (4.05,0.8) ;}}
    \Put(-377,275){ \hyperlink{chapter4}{\tikz \fill [opacity=0.05, rounded corners=11, black] (0,0) rectangle (3.6,0.8) ;}}
    \Put(-383,637){ \hyperlink{chapter3}{\tikz \fill [opacity=0.05, rounded corners=11, black] (0,0) rectangle (3.1,0.8) ;}}
    \Put(-386,1009){\hyperlink{chapter1}{\tikz \fill [opacity=0.05, rounded corners=11, black] (0,0) rectangle (3.15,0.8) ;}}
    \Put(-194,925){ \hyperlink{chapter2}{\tikz \fill [opacity=0.05, rounded corners=11, black] (0,0) rectangle (3.0,0.8) ;}}
    \Put(-149,577){ \hyperlink{chapter6}{\tikz \fill [opacity=0.05, rounded corners=6, black] (0,0) rectangle (3.2,0.45) ;}}
    \caption{Organization of the thesis, a visual representation. The main research question and the \textcolor{red}{input}/\textcolor{blue}{model}/\textcolor{YellowOrange}{output} of each chapter is also included.}
    \label{fig:main}
    \end{center}
\end{figure}

This thesis aims to extract valuable insights and information from time series data using fast, scalable and efficient warping methods that ultimately improve time series averaging, whole-time-series classification and incremental clustering. 
\cref{fig:main} shows a visual representation of how the chapters of this thesis are related.

\textbf{\cref{chapter:0}} is the introduction chapter and summarizes the framework, aim and scope of this thesis. 
In \textbf{\cref{chapter:1}} we provide a formal definition of what a time series is, and discuss the temporal invariances that frequently arise in time series data, such as scaling, warping, occlusion, etc. Invariances are closely related to similarity measures for time series, which allow two time series to be compared; these are also covered in this chapter.
Among the existing temporal invariances, nonlinear warping is a source of nuisance for time series analysis methods and can seriously compromise their performance. 
As such, this thesis presents novel parametric alignment techniques that are well-suited for deep learning models. 
In \textbf{\cref{chapter:2}} we propose closed-form diffeomorphic
transformations that act as a warping function for time series alignment, efficiently removing the unwanted nonlinear temporal deformations present in the data. 
\cref{chapter:2} also discusses the mathematical properties of these diffeomorphic transformations and their computational efficiency, which is essential for minimizing their impact in subsequent time series applications: alignment, averaging, classification and clustering. Due to the numerous applications of the proposed diffeomorphic transformations, this chapter is presented as a time-series-independent chapter. In fact, this chapter is pivotal for the thesis, since these diffeomorphic transformations serve as a foundation in subsequent chapters, from \cref{chapter:3} all the way to \cref{chapter:6} (see \cref{fig:main}).
In \textbf{\cref{chapter:3}} we incorporate the proposed closed-form diffeomorphic transformations into a temporal transformer network for time series alignment and averaging. 
In \textbf{\cref{chapter:4}} we apply the methodological contributions on time series warping presented in previous chapters to the classification of real-world time series data. Using hydraulic pressure sensor data as input, we developed a deep-learning based fault diagnosis \& classification model for a rock drill application that simultaneously aligns and classifies faulty signals with high accuracy. 
\textbf{\cref{chapter:5}} focuses on time series clustering techniques that are warping-invariant and can operate under limited computational and time resources. 
Clustering can help to understand the underlying dynamics of complex systems and detect anomalies, and in this chapter we specifically propose a novel incremental and scalable clustering algorithm that can efficiently cluster an unlimited number of time series data.
\textbf{\cref{chapter:6}} deals with normalizing flows, a type of generative model that converts a simple probability distribution into a more complex one by applying a sequence of invertible and differentiable mappings. We use the diffeomorphic transformations presented in \cref{chapter:2} to build a novel module that acts as a drop-in replacement for the affine or additive transformations commonly found in coupling and autoregressive flows and that significantly enhances their flexibility. 
Finally, \textbf{\cref{chapter:7}} summarizes the main contributions of this thesis, and also reviews the techniques and rehashes the results obtained.

\paragraph{How to read this thesis}

It is the author's belief that readers will benefit most from reading each chapter as it is presented, reading the dissertation from start to finish.
Some chapters are tightly linked to each other while others are more independent. 
\cref{chapter:1} presents an in-depth review of similarity metrics for time series. It is a unique chapter in that its primary purpose is to lay the necessary background for subsequent chapters. A more in-depth literature review is then conducted in each chapter, in order to better clarify related techniques, as well as to clearly state the research gap and the chapter's objectives.
In this regard, \cref{chapter:6} also presents some particularities. Given that all previous chapters were devoted to time series, an appropriate introduction to the matter is provided in \cref{chapter:6}, with a formal introduction to normalizing flows and an extensive review of the state-of-the-art.

\paragraph{Notation}
It is worth noting that the notation and particular contents of the chapters of this thesis differ from the original journal articles they are based on. This has been done for the sake of clarity and to maintain the integrity of the notation employed herein.
We use bold symbols to indicate vectors (lowercase) and matrices (uppercase), otherwise variables are scalars.

\clearpage
\graphicspath{{content/chapter1/}}

\chapter[Background on Time Series Similarity Metrics]{Background on Time Series \\ Similarity Metrics}\label{chapter:1}
\begingroup
\hypertarget{chapter1}{}
\hypersetup{linkcolor=black}
\setstretch{1.0}
\minitoc
\endgroup

This chapter is meant to give an overview of similarity measures for time series data and is structured as follows.
In \cref{sec:time_series_data} we provide a formal definition of what a time series is and in \cref{sec:invariances} we discuss the temporal invariances that frequently arise in time series data, such as scaling and warping. We thoroughly review shape, feature and model-based similarity metrics in \cref{sec:similarity_metrics}, with special focus on elastic metrics. Final remarks are included in \cref{sec:conclusions_1}.

\section{Time Series Data}\label{sec:time_series_data}

Time series data is a sequence of data points indexed in time order. 
These data points typically consist of successive measurements made from the same source over a time interval and are used to track change over time. Plot the points on a graph, and one of your axes would always be time. 

One of the key (and obvious) characteristics of time series data is that it is ordered in time. This means that the measurements are taken in a specific order, and the order is important in analyzing the data. Past can affect the future, but not vice versa. For example, if we were looking at stock prices over time, the order in which the prices were recorded would be important in identifying trends and patterns in the data. 

Regarding the data storing format, a time series $\mathbf{x}$ consists of two mandatory components: time units and value assigned to the corresponding time step. 
Values of the series need to denote the same meaning and correlate among the nearby values. Restriction is, that at the same time there can be at most one value for each time unit. 
In other words, a time series $\mathbf{x}: \Omega \rightarrow \mathbb{R}^d$ is a sequential list of $n$ time and value tuples $\mathbf{x}=\{(t_{1}, x_{1}), (t_{2}, x_{2}), \cdots, (t_{n}, x_{n}) \}$ that maps the temporal domain $\Omega \subseteq \mathbb{R}$ to the observed $d$-dimensional measurements.
In case there is a regular interval between observations, we can simplify this definition for a time series
$\mathbf{x}=(x_{1}, x_{2}, \cdots, x_{n})$ of length $n$. 

Based on the dimensionality $d$, there exist two kinds of time series: univariate and multivariate. At a given time point, the value of a univariate time series is a single element (typically a real number), while the value of a multivariate time series is a vector (usually also of real numbers). For example, a univariate time series might consist of the daily temperature in a particular city, while a multivariate time series might consist of daily temperature, humidity, and air pressure data.

\begin{definition}[Multivariate time series]
    A $d$-dimensional multivariate time series of length $n$ is defined as
    $\mathbf{x}=(x_{1}, x_{2}, \cdots, x_{n}) \in \mathbb{R}^{n \times d}$ where $x_{i} \in \mathbb{R}^{d}$.
\end{definition}

\begin{definition}[Univariate time series]
    A univariate time series of length $n$ is a one-dimensional ($d=1$) multivariate time series defined as
    $\mathbf{x}=(x_{1}, x_{2}, \cdots, x_{n}) \in \mathbb{R}^{n}$ where $x_{i} \in \mathbb{R}$.
\end{definition}

There are many various time series classifications based on specific criteria. With regard to this thesis, the most significant dependencies are: \textbf{length, frequency and stationarity}.

Time series data can have variable length, meaning that the number of observations in the series may vary. For example, a time series of daily temperature data may have 365 observations in a leap year, but only 364 observations in a non-leap year. On the other hand, time series data with regular length have a fixed number of observations in the series.
Furthermore, time series data can have either regular or irregular frequency. Regular frequency data are recorded at equal intervals, such as daily, monthly, or hourly. Irregular frequency data, on the other hand, are recorded at unequal intervals.
Another important characteristic of time series data is that it is often non-stationary. This means that the underlying trends and patterns in the data may change over time. For example, the stock market may experience a period of growth followed by a period of decline, and these changes can be difficult to predict.

In this thesis, we consider variable-length, non-stationary and regularly sampled time series data. For future work, one could consider extending the proposed methods for time series with irregular frequency.

\section{Invariances in Time Series}\label{sec:invariances}

As stated in \cref{chapter:0}, one of the key challenges in analyzing time series data is identifying and measuring the similarity between time series. Closely related to this issue is the set of invariances that each similarity measure is invariant to.

An \textbf{invariance is a property which remains unchanged} when a certain transformation is applied. 
For example, a function may be scale invariant, meaning that a change in the scale of the input data does not affect the output of the function. 
A classical example of invariance is shift-invariance, arising in computer vision and pattern recognition applications such as image classification. It is often reasonably assumed that the classification result should not be affected by the position of the object in the image, i.e., the classification model must be shift-invariant.
If we however take a closer look at the convolutional layers of CNNs, we will find that they are not shift-invariant but shift-equivariant: in other words, a shift of the input to a convolutional layer produces a shift in the output feature maps by the same amount.

Analogous to how certain tasks in computer vision exhibit spatial invariances, temporal invariances frequently arise in time series data. 
For example, phase invariance relates to a transformation that shifts a signal forward of backward in time, resulting in an alignment in phase. Such transformations can be particularly useful when processing periodic signals e.g., electrocardiogram waveforms.
One-dimensional convolutional neural networks (CNNs) efficiently exploit phase invariance by design. This property, in addition to their computational efficiency achieved by weight sharing, has led to their successful application to a variety of tasks involving sequential data.  

Preprocessing techniques like dynamic time warping are also used to exploit warping invariances and align time series data, facilitating relevant comparisons.
Recognizing that time series data may exhibit other types of invariance beyond phase, in \cref{chapter:3} we propose new methods to explicitly account for amplitude and non-linear warping invariances.

In this section, the temporal axis is modeled by the function $f$, i.e., 
$t_{new}=f(t)$
while the spatial axis is transformed by the function $g$, 
$x_{new}=g(x)$. 
Temporal invariances describe a set of transformations that, when applied to the time series, magnify task-relevant similarities between samples. Tasks involving time series data may exhibit different invariances:


\begin{itemize}
  \item \textbf{Amplitude}: transformation of the amplitude of the time series $g(x)=a \cdot x$, where $a$ is a constant parameter and represents the scale.
  \item \textbf{Offset/Translation}: uniformly increases/decreases the value of a time series. $g(x)=x+b$. Amplitude and offset invariances can be trivially achieved by z-normalizing the data. A classic example of where these invariances are critical is object recognition from video, where zoom and pan/tilt correspond to amplitude and offset, respectively.
  \item \textbf{Uniform Scaling}: globally stretches the duration of the time series.  $f(t)=a \cdot t$. In contrast to the localized scaling that DTW deals with, in many datasets we must also deal with global scaling.
  \item \textbf{Phase/Shift}: shifts the start time of a time series. $f(t)=t+b$. This form of invariance is important when matching periodic time series such as star-light curves or heartbeats. 
  \item \textbf{Warping}: A transformation that locally stretches or warps the duration of the time series. $f(t)$ can be any monotonically increasing function that is sufficiently expressive to capture the time series warping. This invariance is necessary in almost all biological signals, including motion capture, handwriting and ECGs.
  \item \textbf{Occlusion}: A transformation that randomly removes data. This can arise when measurements are irregularly sampled or missing. This form of invariance occurs in domains where a small subsequence of a time series may be missing.
  \item \textbf{Noise}: A transformation that adds or removes noise.
\end{itemize}

For many tasks, some or all of the above invariances are required when we compare time-series sequences. To satisfy the appropriate invariances, we could preprocess the data to eliminate the corresponding distortions before clustering. For example, by z-normalizing \cite{goldin1995similarity} the data, we can achieve the amplitude and offset invariances.

Let's consider a continuous time series signal, a sinusoidal function $x(t)=\sin(t)$ and run some examples. As before, we denote the time axis (domain) as $f(t)$ and the space axis (codomain) as $g(x)$. By default, we choose the identity function on both axis, $f(t)=t$ and $g(x)=x$, and we obtain the original function as visualized in \cref{fig:function_identity}.

\renewcommand\x{0.28}
\begin{figure}[!htb]
    \begin{center}
    \begin{subfigure}[b]{\x\linewidth}
        \centering
        \includegraphics[width=\linewidth, page=1]{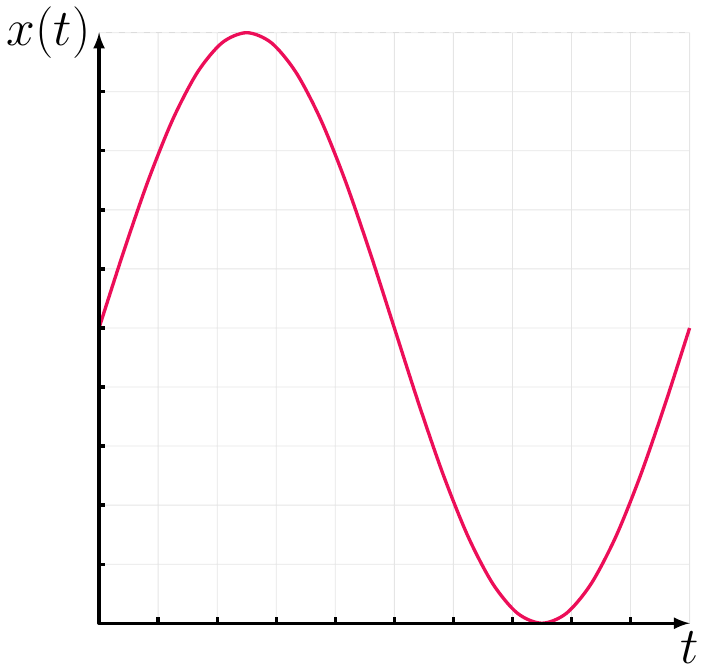}
        \caption{Time series example}
        \label{fig:function_identity_1}
    \end{subfigure}
    \hfill
    \begin{subfigure}[b]{\x\linewidth}
      \centering
      \includegraphics[width=\linewidth, page=3]{figures/warping/warping_identity.pdf}
      \caption{Time axis $f(t)$}
      \label{fig:function_identity_2}
  \end{subfigure}
  \hfill
  \begin{subfigure}[b]{\x\linewidth}
    \centering
    \includegraphics[width=\linewidth, page=4]{figures/warping/warping_identity.pdf}
    \caption{Space axis $g(x)$}
    \label{fig:function_identity_3}
\end{subfigure}
\caption{Example of a time series function $x(t)=\sin(t)$. The time axis is denoted as $f(t)$ and the space axis as $g(x)$, both linear identity functions by default.} 
\label{fig:function_identity}
\end{center}
\vspace{-0.7cm}
\end{figure}

\begin{figure}[!htb]
  \begin{center}
  \begin{subfigure}[b]{\x\linewidth}
      \centering
      \includegraphics[width=\linewidth, page=1]{figures/warping/warping_identity.pdf}
      \caption{Before transformation}
      \label{fig:function_linear_1}
  \end{subfigure}
  \hfill
  \begin{subfigure}[b]{\x\linewidth}
    \centering
    \includegraphics[width=\linewidth, page=3]{figures/warping/warping_linear.pdf}
    \caption{Time axis $f(t)= t/2$}
    \label{fig:function_linear_2}
\end{subfigure}
\hfill
\begin{subfigure}[b]{\x\linewidth}
  \centering
  \includegraphics[width=\linewidth, page=1]{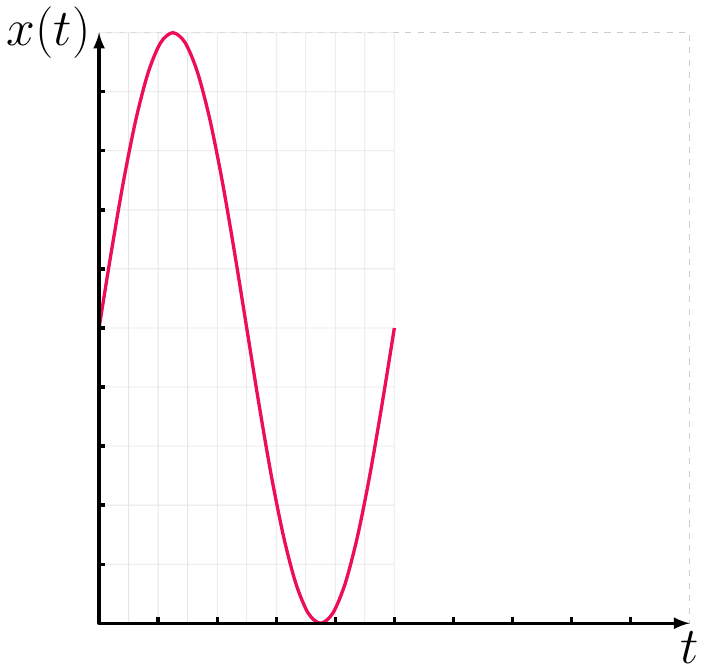}
  \caption{After transformation}
  \label{fig:function_linear_3}
\end{subfigure}
\caption{Temporal transformation $f(t)=t/2$ applied to a time series $x(t)=\sin(t)$.} 
\label{fig:function_linear}
\end{center}
\end{figure}

\paragraph{Example: uniform scaling}
If we apply a linear transformation to the time axis (uniform scaling), for example $f(t) = t/2$, then the temporal axis is linearly scaled ("compressed") to half the duration, as one can see in \cref{fig:function_linear}.

\begin{figure}[!htb]
  \begin{center}
  \includegraphics[width=\linewidth, page=2]{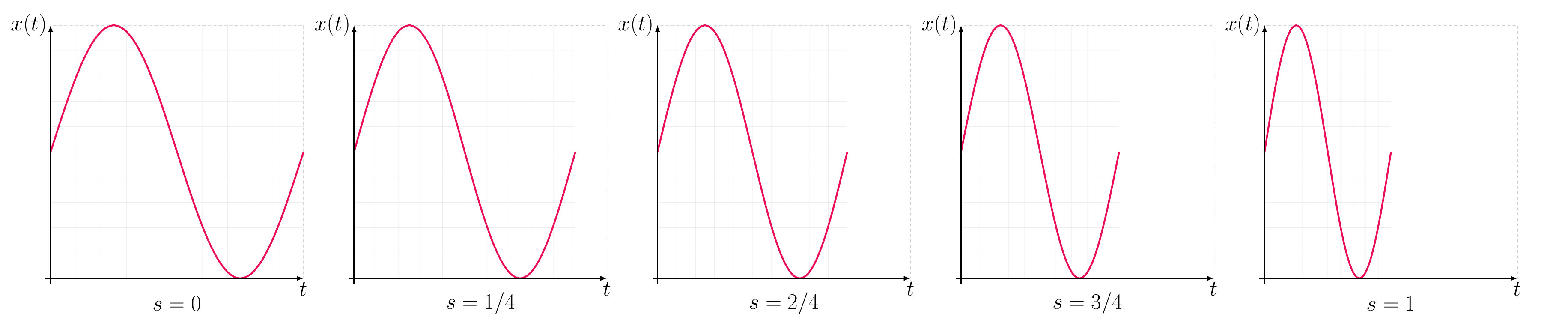}
  \caption{Incremental transformation from the identity function $f(t)= t$ to $f(t)= t/2$} 
  \label{fig:function_linear_evolution}
  \end{center}
  \vspace{-1cm}
\end{figure}

Similarly, we can visualize the transformation evolution as it unfolds in multiple steps by interpolating from the identity function (\cref{fig:function_linear_evolution}). Here $s$ denotes the interpolation parameter, from $s=0$ being the identity function and $s=1$ being the applied transformation.

\clearpage

\begin{figure}[!htb]
  \begin{center}
    \begin{subfigure}[b]{0.3\linewidth}
      \centering
      \includegraphics[width=\linewidth]{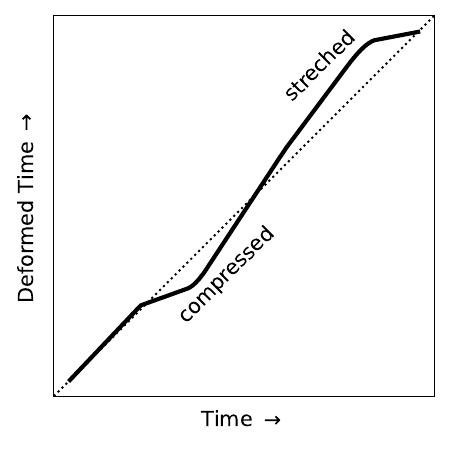}
    \caption{Warping function}
    \label{fig:spring_curve}
  \end{subfigure}
  \begin{subfigure}[b]{0.68\linewidth}
    \centering
    \includegraphics[width=\linewidth]{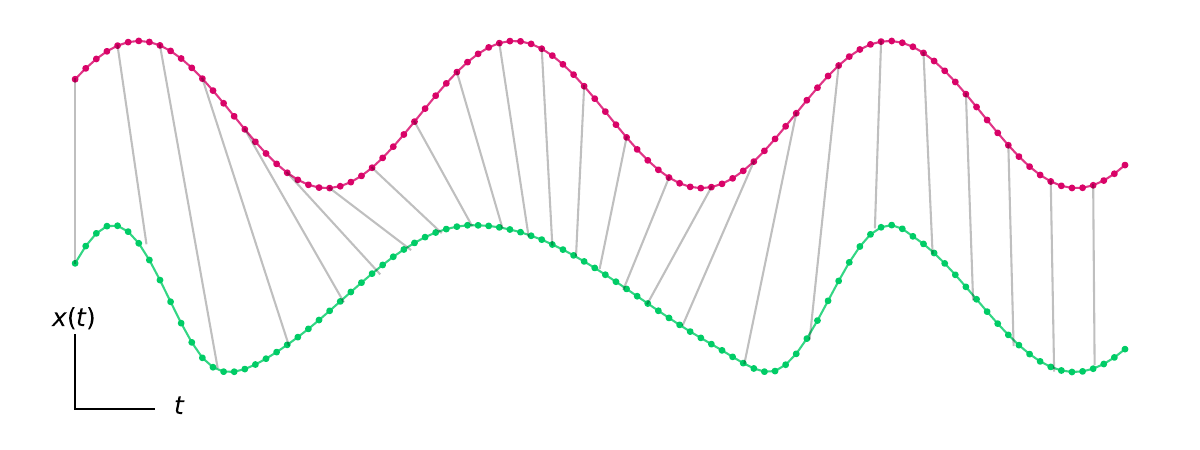}
    \caption{Time series deformed by the warping function on the left.}
    \label{fig:spring_warping}
  \end{subfigure}
  \\
  \vspace{1em}
  \begin{subfigure}[b]{\linewidth}
      \centering
      \includegraphics[width=0.9\linewidth]{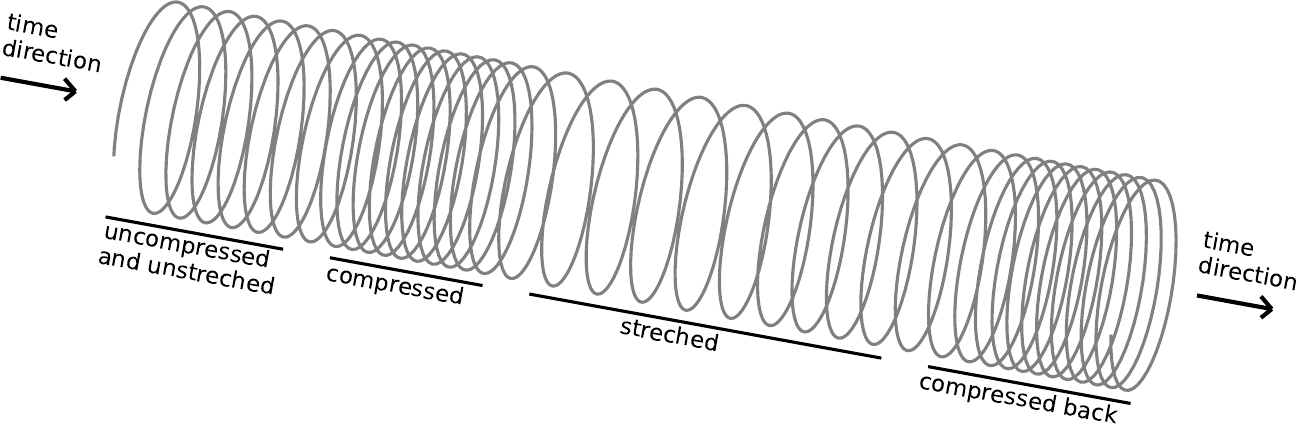}
      \caption{Mechanical spring illustrating a nonlinear space deformation along the longitudinal axis.}
      \label{fig:spring}
  \end{subfigure}
  \caption{Using an analogy, nonlinear time warping is comparable to a mechanical spring that is compressed and streched in various sections.}
  \label{fig:spring_example}
  \end{center}
\end{figure}

\paragraph{Example: warping}
We can extend uniform scaling to more complex, nonlinear transformations $f(t)$. Nonlinear time warping refers to the distortion of the time axis by a nonlinear function (see \cref{fig:spring_example}). 

\begin{figure}[!htb]
  \begin{center}
  \begin{subfigure}[t]{\x\linewidth}
    \centering
    \includegraphics[width=\linewidth, page=1]{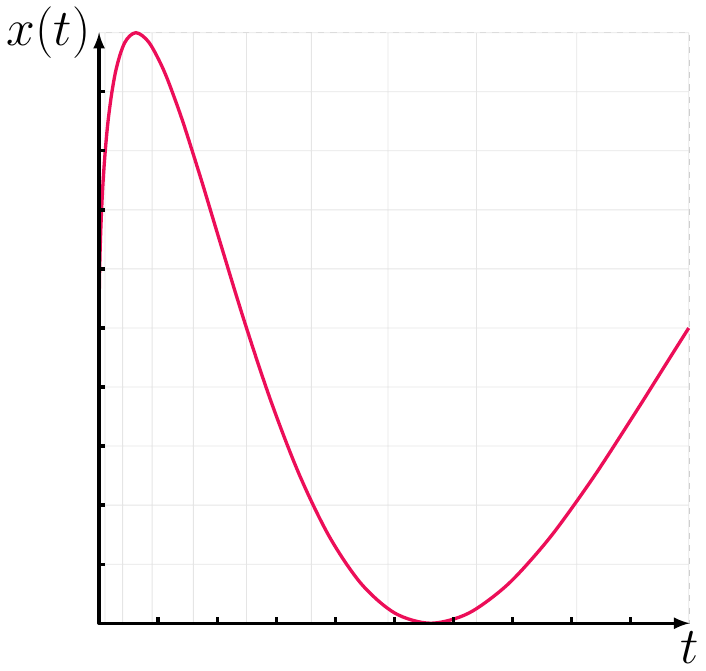}
    \includegraphics[width=\linewidth, page=3]{figures/warping/warping_quadratic.pdf} \\
    \captionsetup{justification=centering}
    \caption{Quadratic function \\ $f(t)= t^{2}$}
    \label{fig:function_quadratic_vertical}
  \end{subfigure}
  \hfill
  \begin{subfigure}[t]{\x\linewidth}
      \centering
      \includegraphics[width=\linewidth, page=1]{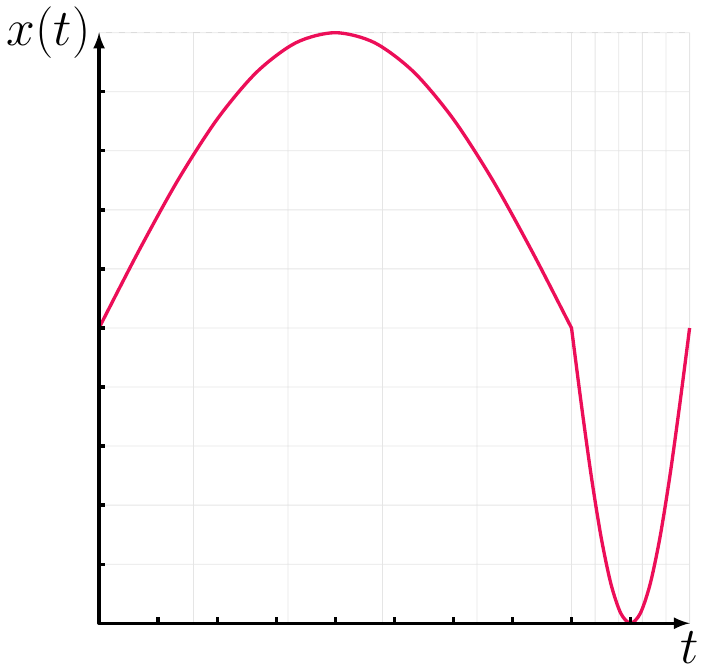}
      \includegraphics[width=\linewidth, page=3]{figures/warping/warping_piecewise.pdf} \\
      \captionsetup{justification=centering, width=1.2\linewidth}
      \caption{Piecewise affine function \\ $f(t) = \left\{\begin{matrix*}[l]
        1.6t & \; t \leq \frac{1}{2} \\
        0.6+0.4t & \; t \geq \frac{1}{2} \\
      \end{matrix*}\right.$}
      \label{fig:function_piecewise_vertical}
  \end{subfigure}
  \hfill
  \begin{subfigure}[t]{\x\linewidth}
    \centering
    \includegraphics[width=\linewidth, page=1]{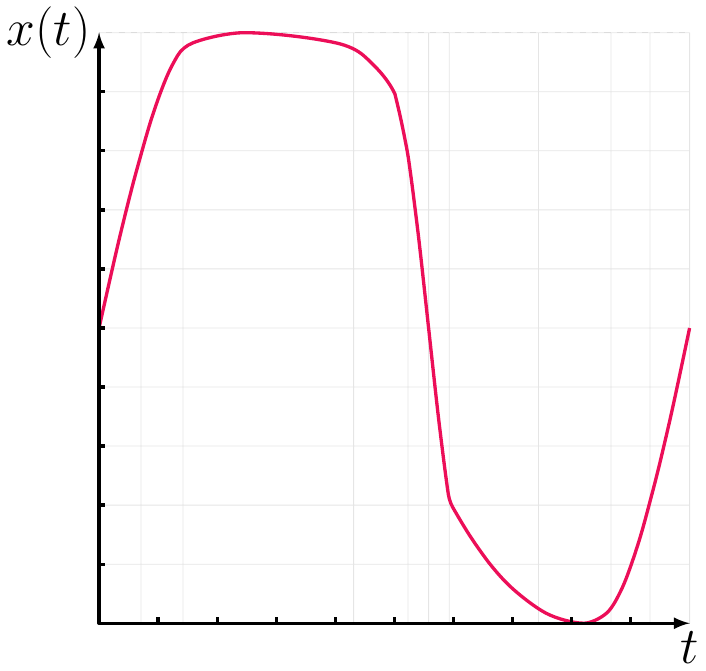}
    \includegraphics[width=\linewidth, page=3]{figures/warping/warping_cpab.pdf} \\
    \captionsetup{justification=centering}
    \caption{Function from ODE \\ integration  $f'(t)=v(f(t))$}
    \label{fig:function_cpab_vertical}
  \end{subfigure}
\caption{Examples of temporal transformations $f(t)$ applied to $x(t)=\sin(t)$.} 
\label{fig:temporal_transformations}
\end{center}
\end{figure}

\begin{figure}[!htb]
  \begin{center}
\begin{subfigure}[b]{\linewidth}
  \centering
  \includegraphics[width=\linewidth, page=2]{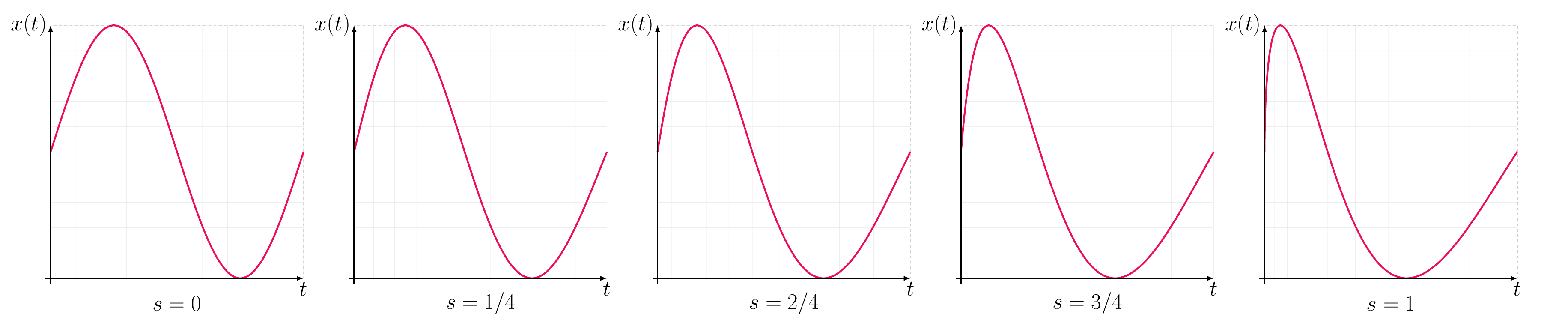}
  \caption{Quadratic function $f(t)= t^{2}$}
  \label{fig:function_quadratic_evolution}
\end{subfigure}
\\
\begin{subfigure}[b]{\linewidth}
  \centering
  \includegraphics[width=\linewidth, page=2]{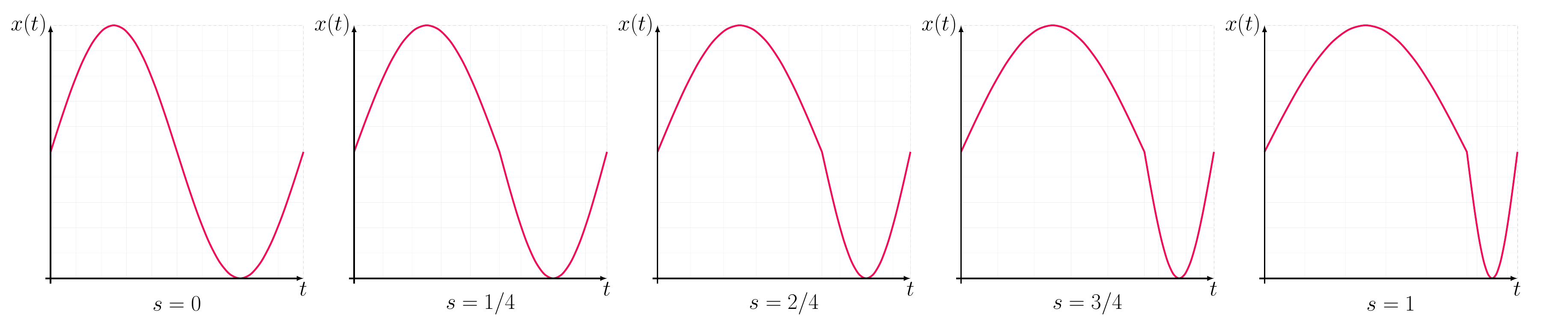}
  \caption{Piecewise affine function $f(t) = \left\{\begin{matrix*}[l]
    1.6t & \; t \leq \frac{1}{2}; \quad
    0.6+0.4t & \; t \geq \frac{1}{2} 
  \end{matrix*}\right\}$}
  \label{fig:function_piecewise_evolution}
\end{subfigure}
\\
\begin{subfigure}[b]{\linewidth}
  \centering
  \includegraphics[width=\linewidth, page=2]{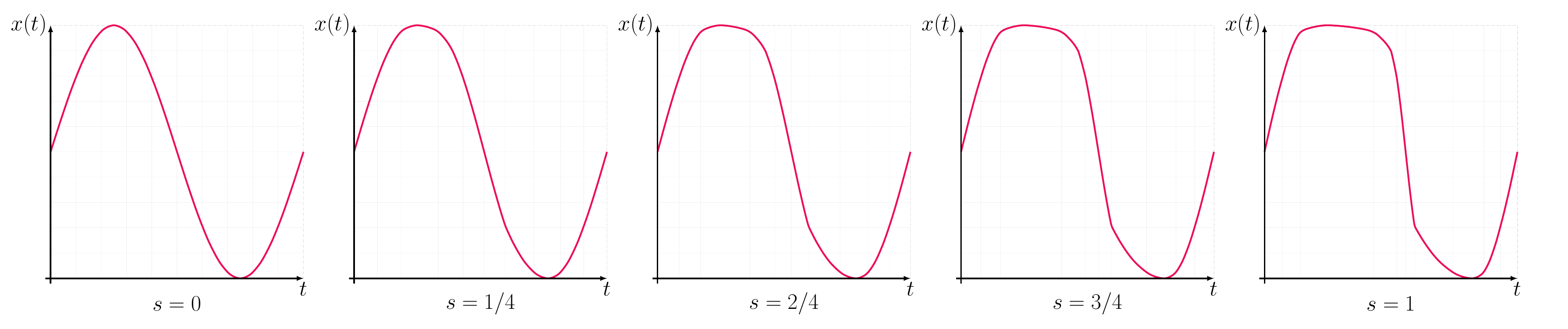}
  \caption{Function from ODE integration $f'(t)=v(f(t))$}
  \label{fig:function_cpab_evolution}
\end{subfigure}
\caption{Incremental transformation from the identity function $f(t)= t$ to different temporal transformations $f(t)$, applied to a time series function $x(t)=\sin(t)$.}
\label{fig:function_evolution}
\end{center}
\end{figure}

Some examples of nonlinear transformations are provided below and visualized in \cref{fig:temporal_transformations,fig:function_evolution}:

\begin{itemize}
  \item Quadratic function $f(t)= t^{2}$
  \item Piecewise affine function $f(t) = \left\{\begin{matrix*}[l]
    1.6t & \; t \leq \frac{1}{2}; \\
    0.6+0.4t & \; t \geq \frac{1}{2} \\ 
  \end{matrix*}\right.$
  \item Function from ODE integration $f'(t)=v(f(t))$, which yields diffeomorphic transformations (more on this in \cref{chapter:2}).
\end{itemize}

\clearpage

\paragraph{Dealing with unwanted invariances}
In other domains, to exploit invariances researchers either augment their training data by applying a variety of transformations or modify the neural network architecture. The first approach is most popular in domains where it is straightforward to generate realistic training examples (e.g., images). Common image invariances include rotation, scale, translation and warping. Such transformations are easily applied to existing images to create additional, realistic training examples.

Recognizing the difficulty in identifying potential invariances or transformation a priori in the time series domain, \textbf{we focus on learning the invariances directly from the data}. Specifically, our proposed method tackles amplitude and offset invariances, phase invariance, and uniform scale invariance, and learns input-specific transformation parameters directly from the data. More on this in \cref{chapter:3}.

\clearpage
\section{Similarity Metrics}\label{sec:similarity_metrics}

The definition of appropriate metrics for comparing objects is at the core of many machine learning techniques.
When complex objects are involved, such metrics have to be carefully designed in order to leverage on desired notions of similarity.

Similarity metrics in time series, also called similarity distances\footnote{the terms "metric" and "distance" are often used interchangeably}, are measures of how closely two time series datasets match each other. There are many types of similarity metrics, each of which has its own strengths and weaknesses. 
By calculating the similarity distance between two time series, we can determine how closely they match and make inferences about the underlying processes that generated the data. 

Let $\mathbf{x}=(x_{1}, \cdots, x_{n}) \in \mathbb{R}^{n}$ and $\mathbf{y}=(y_{1}, \cdots, y_{m}) \in \mathbb{R}^{m}$ be two time series of length $n$ and $m$ respectively. Time series $\mathbf{x}$ and $\mathbf{y}$ can be seen as numerical vectors with dimensions $n$ and $m$, which can differ in value. 

Formally, a metric $D: S \times S \rightarrow \mathbb{R}$ is a function that assigns a non-negative real number to any pair of elements in a set $S$, which satisfies the properties of non-negativity $D(\mathbf{x},\mathbf{y})>0$, symmetry $D(\mathbf{x},\mathbf{y})=D(\mathbf{y},\mathbf{z})$, and the triangle inequality $D(\mathbf{x},\mathbf{z}) \leq D(\mathbf{x},\mathbf{y}) + D(\mathbf{y},\mathbf{z})$ for all points $\mathbf{x}, \mathbf{y}, \mathbf{z} \in S$. 
Hence, we denote $D(\mathbf{x},\mathbf{y})$ as the distance between time series $\mathbf{x}$ and $\mathbf{y}$. 
The lower the value of $D(\mathbf{x},\mathbf{y})$, the closer the two observations are according to the chosen distance measure. 

Computational complexity is a major concern when it comes to similarity metrics. 
Some clustering methods require computing the distance or dissimilarity matrix, which contains the distance between each pair of observations, i.e. all pairwise distances. 
When determining the distance matrix for $N$ different observations, a total of $\frac{N(N-1)}{2}$ distances have to be calculated. 
The required calculation time for determining a distance matrix thus grows with $\mathcal{O}(N^2)$ and this is without taking into account the time complexity of the individual distance measures. Due to the quadratic time complexity, determining the distance matrix for large sets of observations ($\gg 1000$) can be a time-consuming operation. On the other hand, pairwise distance calculations are easy to parallelize, and finding efficient approximate methods is an active field of research.

This section covers similarity metrics for time series data. \cref{tab:similarity_metrics} shows the methods reviewed in this section, and their time complexity, a brief description and the respective references.
We present several representative examples of different families of time series similarity measures: lock-step measures (e.g. Euclidean distance), elastic measures (e.g. dynamic time warping), feature-based measures (e.g. Fourier coefficients), and model-based measures (e.g. autoregressive).

\begin{landscape}
  \begin{table}[!ht]
    \caption{Similarity metrics}
    \label{tab:similarity_metrics}
    \vspace{-20pt}
    \begin{center}
    \resizebox*{!}{0.98\textheight}{%
    \begin{tabular}{lllll}
    \toprule
      \textbf{Type} &
      $\quad$ \textbf{Method} &
      \textbf{Time Complexity} &
      \textbf{Description} &
      \textbf{Reference} \\ 
      
    \midrule \multirow{7}{*}{\rotatebox[origin=c]{90}{\begin{tabular}[t]{c}\textbf{Shape-based} \\ \textbf{Lock-step} \end{tabular}}}
    & $\quad$ Minkowski & $\mathcal{O}(n)$ & Generalization of Euclidean, Manhattan and Chebyschev & \\
    & $\quad$ Pearson           & $\mathcal{O}(n)$         & Invariant to scale and location         &  \cite{pearson1895vii} \\
    & $\quad$ Spearman          & $\mathcal{O}(n \log n)$  & Less sensitive to noise                 &  \cite{spearman1961proof} \\
    & $\quad$ Kendall           & $\mathcal{O}(n^2)$       & Less sensitive to noise                 &  \cite{kendall1938new} \\
    & $\quad$ Cross-correlation & $\mathcal{O}(n \log n)$  & Accounts for lags in time               &  \\
    & $\quad$ SBD               & $\mathcal{O}(n)$         & Normalized cross-correlation            &  \cite{paparrizos2017fast} \\
    \midrule \multirow{12}{*}{\rotatebox[origin=c]{90}{\begin{tabular}[t]{c}\textbf{Shape-based} \\ \textbf{Elastic} \end{tabular}}}
    & $\quad$ DTW & $\mathcal{O}(nm)$ & Warping invariant, robust to noise, shifts and scaling & \cite{sakoe1978dynamic} \\
    & $\quad$ EDIT              & $\mathcal{O}(nm)$        & Similarity between characters           &  \cite{marzal1993computation} \\
    & $\quad$ EDR               & $\mathcal{O}(nm)$        & EDIT + real values                      &  \cite{chen2005robust} \\
    & $\quad$ ERP               & $\mathcal{O}(nm)$        & EDR + penalty for non-matching elements &  \cite{chen2004marriage} \\
    & $\quad$ TWED & $\mathcal{O}(nm)$ & DTW + time distance penalty, allows different sampling rates &  \cite{marteau2008time} \\
    & $\quad$ LCSS & $\mathcal{O}(n \delta)$ & Longest common subsequence, noise robustness & \cite{vlachos2002discovering} \\
    & $\quad$ FastDTW           & $\mathcal{O}(\max(n,m))$ & DTW + reduced resolution scale          &  \cite{salvador2007toward} \\
    & $\quad$ DDTW              & $\mathcal{O}(nm)$        & DTW + first order differences           &  \cite{keogh2001derivative} \\
    & $\quad$ WDTW              & $\mathcal{O}(nm)$        & DTW + multiplicative weight penalty     &  \cite{jeong2011weighted} \\
    & $\quad$ ShapeDTW          & $\mathcal{O}(nm)$        & Locally sensible matchings              &  \cite{zhao2018shapedtw} \\
    & $\quad$ GDTW              & $\mathcal{O}(n)$         & DTW + multiple point-to-point distances &  \cite{neamtu2018generalized} \\
    & $\quad$ LDTW              & $\mathcal{O}(nm)$        & DTW + limited warping path length       &  \cite{zhang2017dynamic} \\
    \midrule \multirow{3}{*}{\rotatebox[origin=c]{90}{\begin{tabular}[t]{c}\textbf{Kernel} \\ \textbf{based} \end{tabular}}}
    & $\quad$ GAK               & $\mathcal{O}(nm)$        & Positive-definite kernel                &  \cite{cuturi2011fast} \\
    & $\quad$ SoftDTW           & $\mathcal{O}(nm)$        & Differentiable distance by smoothing    &  \cite{cuturi2017soft} \\
    & $\quad$ DTWNet & $\mathcal{O}(nm) + \mathcal{O}(n+m)$ & Differentiable neural network layer, convolution kernel & \cite{cai2019dtwnet} \\
    \midrule \multirow{7}{*}{\rotatebox[origin=c]{90}{\begin{tabular}[t]{c}\textbf{Feature} \\ \textbf{based} \end{tabular}}}
    & $\quad$ SAX               &                          & Symbolic representation                 &  \cite{lin2003symbolic}\\
    & $\quad$ BOP               &                          & Discretized histogram                   &  \cite{lin2012rotation}\\
    & $\quad$ SAX-VSM           &                          & SAX + TF-IDF matrix                     & \cite{senin2013sax}\\
    & $\quad$ DFT               & $O(n \log n)$            & Spectral representation                 &  \\
    & $\quad$ DWT               &                          & Wavelet decomposition                   &  \cite{haar1909theorie} \\
    & $\quad$ SFA               &                          & DFT + SAX                               &  \cite{schafer2012sfa} \\
    & $\quad$ BOSS              &                          & SFA + histogram                         &  \cite{schafer2015boss} \\
    \midrule \multirow{3}{*}{\rotatebox[origin=c]{90}{\begin{tabular}[t]{c}\textbf{Model} \\ \textbf{based} \end{tabular}}}
    & $\quad$ Shapelets      & $O(n^2)$  & Finds short repeated subsequences       &  \cite{cheng2020time2graph} \\ 
    & $\quad$ Autoregressive & & Model captures a linear combination of past values & \\
    & $\quad$ Siamese network      & & Identical sub-networks &  \cite{hou2019time} \\ 
    \bottomrule
    \end{tabular}%
    }
    \end{center}
  \end{table}
\end{landscape}

Shape-based distances (lock-step or elastic) compare the overall shape of time series based on the actual (scaled) values of the time series. Unlike feature-based approaches, which rely on specific features or characteristics of the data, shape-based distances focus on the overall shape of the time series. This approach is particularly useful in cases where a suitable distance can be found, but a set of features cannot be identified.

Though both shape- and feature-based methods are very powerful, they can be outperformed by data-driven methods when sufficient data is available. Data-driven approaches, such as Siamese neural network (SNN), can be viewed as automatically constructing the set of features or distance metric to use based on the available data, as opposed to relying on domain knowledge introduced by an expert.

\paragraph{Software Libraries}
Regarding available software libraries, \cref{tab:software} summarizes some of the most popular and well-developed packages for R and Python.

\begin{table}[!htb]
  \caption{Software libraries to compute similarity metrics between time series.}
  \label{tab:software}
  \begin{center}
  \resizebox{0.8\textwidth}{!}{%
  \begin{tabular}{llccccccr}
  \toprule
  \rotatebox{0}{\textbf{Language} } & 
  \rotatebox{0}{\textbf{Library} } & 
  \rotatebox{90}{\textbf{Lock-step} } & 
  \rotatebox{90}{\textbf{Elastic metrics} } & 
  \rotatebox{90}{\textbf{Kernel based} } & 
  \rotatebox{90}{\textbf{Feature based} } & 
  \rotatebox{90}{\textbf{Model based} } & 
  \rotatebox{90}{\textbf{Shapelet [REVIEW]} } &
  \rotatebox{0}{\textbf{Reference} } 
  \\ \midrule
  \multirow{4}{*}{R}      & dtwclust     & \textcolor{ForestGreen}{\cmark}   & \textcolor{ForestGreen}{\cmark}  & \textcolor{ForestGreen}{\cmark}  &         &         &        &  \cite{dtwclust}         \\
                          & tsclust      & \textcolor{ForestGreen}{\cmark}   & \textcolor{ForestGreen}{\cmark}  &         & \textcolor{ForestGreen}{\cmark}  & \textcolor{ForestGreen}{\cmark}  &        &  \cite{tsclust}          \\
                          & tsdist       & \textcolor{ForestGreen}{\cmark}   & \textcolor{ForestGreen}{\cmark}  & \textcolor{ForestGreen}{\cmark}  & \textcolor{ForestGreen}{\cmark}  & \textcolor{ForestGreen}{\cmark}  &        &  \cite{tsdist}           \\
                          & tsrepr       &          &         &         & \textcolor{ForestGreen}{\cmark}  & \textcolor{ForestGreen}{\cmark}  &        &  \cite{tsrepr}           \\ \midrule
  \multirow{8}{*}{Python} & tslearn      &          & \textcolor{ForestGreen}{\cmark}  & \textcolor{ForestGreen}{\cmark}  &         & \textcolor{ForestGreen}{\cmark}  & \textcolor{ForestGreen}{\cmark} &  \cite{JMLR:v21:20-091}  \\
                          & seglearn     & \textcolor{ForestGreen}{\cmark}   &         &         & \textcolor{ForestGreen}{\cmark}  &         &        &  \cite{arXiv:1803.08118} \\
                          & tsfresh      & \textcolor{ForestGreen}{\cmark}   &         &         & \textcolor{ForestGreen}{\cmark}  & \textcolor{ForestGreen}{\cmark}  &        &  \cite{christ2018time}   \\
                          & pyts         &          & \textcolor{ForestGreen}{\cmark}  &         & \textcolor{ForestGreen}{\cmark}  & \textcolor{ForestGreen}{\cmark}  & \textcolor{ForestGreen}{\cmark} &  \cite{JMLR:v21:19-763}  \\
                          & sktime       & \textcolor{ForestGreen}{\cmark}   & \textcolor{ForestGreen}{\cmark}  &         &         &         &        &  \cite{loning2019sktime} \\
                          & stumpy       & \textcolor{ForestGreen}{\cmark}   &         &         & \textcolor{ForestGreen}{\cmark}  &         &        &  \cite{law2019stumpy}    \\
                          & pyflux       &          &         &         &         & \textcolor{ForestGreen}{\cmark}  &        &  \cite{taylor2016pyflux} \\
                          & dtaidistance & \textcolor{ForestGreen}{\cmark}   & \textcolor{ForestGreen}{\cmark}  &         &         &         &        &  \cite{meert2020wannesm} \\ \bottomrule
  \end{tabular}%
  }
  \end{center}
\end{table}

\clearpage

\subsection{Shape-based: Lock-Step Distances}\label{sec:shape_lock_step}

Let $\mathbf{x}=(x_{1}, \cdots, x_{n}) \in \mathbb{R}^{n}$ and $\mathbf{y}=(y_{1}, \cdots, y_{n}) \in \mathbb{R}^{n}$ be two time series of the same length $n$.
Lock-step distances require both time series to be of equal length ($n = m$) and compare time point $i$ of time series $\mathbf{x}$ with the same temporal location $i$ of time series $\mathbf{y}$.

\subsubsection{$\mathbf{L_p}$-norm Distances}
The Minkowski distance is defined as the $L_{p}$-norm of the difference between two vectors of equal length ($n = m$) and is the generalization of the commonly used Euclidean ($p=2$), Manhattan ($p=1$) and Chebyshev ($p= \infty$) distances. The time complexity for the Minkowsky distance is $\mathcal{O}(n)$ and thus determining the distance matrix for $N$ time series with this measure takes $\mathcal{O}(n\times N ^2)$ time. 

\begin{itemize}
  \item Minkowski: $D(\mathbf{x},\mathbf{y}) = \big(\sum_{i=1}^{n} |x_{i} - y_{i}|^p \big)^{1/p}$
  \item Euclidean: $D(\mathbf{x},\mathbf{y}) = \sqrt{\sum_{i=1}^{n} (x_i - y_i)^2}$
  \item Manhattan: $D(\mathbf{x},\mathbf{y}) = \sum_{i=1}^{n} |x_i - y_i|$
  \item Chebyshev: $D(\mathbf{x},\mathbf{y}) = \max_{i=1,\cdots,n} |x_i - y_i|$
\end{itemize}

The Euclidean distance is one of the most used time series dissimilarity measures, favored by its speed, computational simplicity, and indexing capabilities. It is used as benchmark in many works, because it is parameter free. However, this measure is very weak and sensitive to small shifts across the time axis, as illustrated in \cref{sec:challenge_4}.


\subsubsection{Pearson Correlation Distance}
The Pearson correlation coefficient takes into account the linear association between two time series and is defined as: 
\begin{equation}  
  \rho(\mathbf{x},\mathbf{y}) = \frac{\text{Cov}(\mathbf{x},\mathbf{y})}{\sigma_x\sigma_y} = \frac{\mathbb{E}[(\mathbf{x}-\mu_{\mathbf{x}})(y-\mu_{\mathbf{y}})]}{\sigma_{\mathbf{x}}\sigma_\mathbf{y}} = \frac{\sum_{i=1}^{n}(x_i - \bar{\mathbf{x}})(y_i - \bar{\mathbf{y}})}{\sqrt{\sum_{i=1}^{n}(x_i - \bar{\mathbf{x}})^2}\sqrt{\sum_{i=1}^{n}(y_i - \bar{\mathbf{y}})^2}}
\end{equation}

where $\mu_{\mathbf{x}}$ and $\mu_{\mathbf{y}}$ are the means of $\mathbf{x}$ and $\mathbf{y}$ and $\sigma_{\mathbf{x}}$ and $\sigma_{\mathbf{y}}$ are the standard deviations of $\mathbf{x}$ and $\mathbf{y}$, respectively. The values of $\rho$ lie within the range [-1,1], where $\rho = 1$ indicates a perfect positive relationship between $\mathbf{x}$ and $\mathbf{y}$, $\rho = -1$ indicates a perfect negative relationship between $\mathbf{x}$ and $\mathbf{y}$, and $\rho = 0$ indicates no relationship between the two variables. The Pearson correlation distance is defined as:
$D(\mathbf{x}, \mathbf{y}) = 1-\rho(\mathbf{x}, \mathbf{y})$
This distance measure can take values in the range [0,2]. The time complexity again is $\mathcal{O}(n)$ and thus determining the distance matrix for $N$ time series with this measure takes $\mathcal{O}(n\times N^2)$ time.

Alternative correlation measures include Spearman's Rank and Kendall's Tau correlation coefficients. 
These coefficients indicate correlation based on rank, whereas the Pearson coefficient is based on a linear relationship between two vectors. This makes Spearman's Rank and Kendall's Tau less sensitive to noise and outliers. However, this comes with an increase in time complexity: $\mathcal{O}(n\log(n))$ for Spearman's Rank and $\mathcal{O}(n^2)$ for Kendall's Tau.

\subsubsection{Spearman Correlation Distance}
Spearman correlation indicates the direction of association between $\mathbf{x}$ (the independent variable) and $\mathbf{y}$ (the dependent variable). If $\mathbf{y}$ tends to increase when $\mathbf{x}$ increases, the Spearman correlation coefficient is positive. 
\begin{equation}
  \rho_{spearman} = \frac{\text{Cov}(rg_{\mathbf{x}},rg_{\mathbf{y}})}{\sigma_{rg_{\mathbf{x}}}\sigma_{rg_\mathbf{y}}}
\end{equation}
where $rg_{\mathbf{x}}$ is the rank of the vector $\mathbf{x}$, which requires sorting the vector.

\subsubsection{Kendall Correlation Distance}
In the same way, Kendall correlation between two variables will be high when observations have a similar rank (i.e. relative position label of the observations within the variable) between the two variables.
\begin{equation}
\tau = \frac{\text{concordant}_{\text{pairs}} - \text{discordant}_{\text{pairs}}}{\binom{N}{2}}
\end{equation}
Any pair of observations $(x_i, y_i)$ and $(x_j, y_j)$ are said to be concordant if the ranks for both elements agree. In order to count the number of concordant and discordant pairs, it is necessary to compare all pairs of observation, thus the time complexity $\mathcal{O}(n^2)$. Another alternative is a distance based on autocorrelation, which accounts for lags in time and has a time complexity of $\mathcal{O}(n\log(n))$.

\clearpage
\subsubsection{Cross-Correlation}

Cross-correlation can determine the similarity of two sequences even if they are not properly aligned. To achieve shift-invariance, cross-correlation keeps the time series $\mathbf{y}$ static and slides $\mathbf{x}$ the over it to compute their inner product for each shift $s$ of $\mathbf{x}$. We denote a shift of a sequence as follows:
\begin{equation}
\mathbf{x}_{s}(t) = \left\{\begin{matrix}
  (\overbrace{0,\cdots, 0}^{|s|}, x_{1}, \cdots, x_{n-s}), \quad \text{if} \quad s \geq 0 \\
  (x_{1-s}, \cdots, x_{n-s},\underbrace{0,\cdots, 0}_{|s|}), \quad \text{if} s \le 0
\end{matrix}\right.
\end{equation}
Shape-based distance (SBD) \cite{paparrizos2017fast} is a normalized version of the cross-correlation measure, that considers the shapes of time series while comparing them. It is an efficient and parameter-free distance measure that is scale and shift invariant, and can be computed efficiently by exploiting intrinsic characteristics of Fourier transform algorithms.

\subsubsection{Limitations}
$L_p$-norm distances  and lock-steps distances in general are not well suited to compare time series because they: 
\begin{enumerate}[label=(\roman*)]
  \item are only defined for two vectors with the same length, whereas the time series of a given dataset often have different lengths.
  \item compare the values of both time series at each time point independently, whereas the values of time series may be correlated. In this sense, considering the minimum of the Euclidean distances between the smaller time series and the subsequences of the same length from the larger time series may not be optimal.
  \item are not invariant to noise, amplitude, offset, phase shifts (see \cref{sec:invariances}). 
\end{enumerate}

\subsection{Shape-based: Elastic Distances}\label{sec:shape_elastic}

Elastic similarity measures (ESMs) such as dynamic time warping (DTW) address the limitations of $L_p$-norm distances (\cref{sec:shape_lock_step}). 
ESMs are particularly well-suited for natural signals, such as those arising from social, economic, or consumer-behavior processes, where the “shape” of the signal is more meaningful than the exact temporal structure.
Elastic measures allow one-to-many and one-to-none matching, and are able to warp the time axis in order to achieve the best alignment and thus are more robust in handling outliers. 

Lock-step distance measures are often outperformed by elastic ones, due to their sensitivity to noise, scale and time shifts. The main disadvantage, however, is that elastic distance measures generally come with an increase in time complexity. ESMs warp two sequences $\mathbf{x}$ and $\mathbf{y}$ non-linearly in time in order to cope with time deformations and varying speeds in time dependent data.

Let $\mathbf{x}=(x_{1}, \cdots, x_{n}) \in \mathbb{R}^{n \times d}$ and $\mathbf{y}=(y_{1}, \cdots, y_{m}) \in \mathbb{R}^{m \times d}$ be two time series of length $n$ and $m$ respectively. Here, all elements  $x_i$  and  $y_j$ are assumed to lie in the same $d$-dimensional space and the exact timestamps at which observations occur are disregarded: only their ordering matters.

The cost matrix, denoted by $C$, is a $n \times m$ matrix consisting of the cost between each pair of values in both time series:
\begin{equation}
  C_{ij} = f(x_{i}, y_{j}) \quad \quad \forall i, j \in \{1, \cdots, n\} \times \{1,\cdots,m\}
\end{equation}

where $f$, often called local divergence or ground metric, is a function evaluating the cost between any pair of real numbers; $f$ is usually the squared Euclidean distance function:
\begin{equation}
f(x,y)=(x-y)^{2} \quad \quad \forall x, y \in \mathbb{R}
\end{equation}

A warping path of order $n \times m$ is a sequence $p=(p_{1}, \cdots, p_{L})$ of $L$ points $p_{l}=(i_{l}, j_{l}) \in [n] \times [m]$ where we denote $[n] = \{1, \cdots, n\}$ such that:

\begin{enumerate}
  \item \textbf{Boundary conditions}: path must start and end in the diagonal corners; $p_{1}=(1,1)$ and $p_{L}=(n,m)$
  \item \textbf{Continuity and step-size constraint}: only adjacent elements in the matrix are allowed for steps in the path; for example $p_{l+1}-p_{l} \in \{(0,1),(1,0),(1,1)\} \quad \forall l\in \{1,\cdots,L-1\}$. (see \cref{sec:step_pattern} for alternative step patterns.)
  \item \textbf{Monotonicity constraint}: subsequent steps in the path must be monotonically spaced in time,  i.e. every index from one time series must be matched with at least one index from the other time series; $p_{l}=(i_{l},j_{l}) \in \{1,\cdots,n\} \times \{1,\cdots,m\}; \quad \forall l\in \{1,\cdots,L\}$
\end{enumerate}

A warping path of order $n \times m$ can be thought as a path in a $[n] \times [m]$ grid where the boundary condition demands that the path starts in the lower left corner and ends in the upper right corner of the grid. The step condition demands that a transition from point to the next point moves a unit in exactly one of the following directions: down, diagonal, and right. Every point $p_{l}=(i_{l}, j_{l})$ of warping path $p$ aligns element $x_{i_{l}}$ to element $y_{j_{l}}$. 

\begin{figure}[!htb]
  \begin{center}
    \begin{subfigure}[t]{0.48\linewidth}
      \centering
      \includegraphics[width=\linewidth,trim=0 -60 0 0, clip]{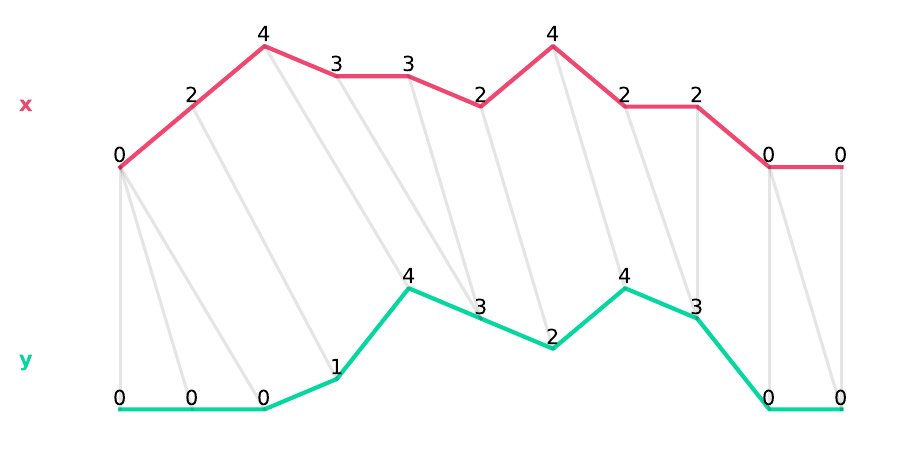}
      \caption{Time series $\mathbf{x}$ and $\mathbf{y}$ of length 10. Gray lines illustrate the one-to-many warping assignments.}
      \label{fig:dtw_time_series_2}
    \end{subfigure}
    \hfill
    \begin{subfigure}[t]{0.48\linewidth}
      \centering
      \includegraphics[width=0.9\linewidth]{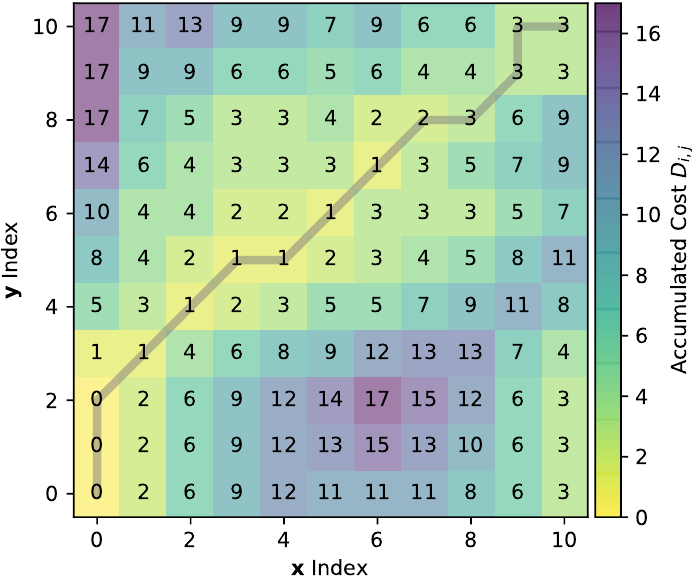}
      \caption{Cost matrix with cumulative distances $D_{i,j}$. Optimal alignment path $p^*$ is represented by the black line.}
      \label{fig:dtw_cost_matrix_2}
    \end{subfigure}
    \caption{Elastic similarity measure between two generic time series $\mathbf{x}$ and $\mathbf{y}$, with $n=m=10$. The cost function corresponds to the Euclidean distance, $f(x,y)=(x-y)^{2}$. }
    \label{fig:dtw_2}
  \end{center}
\end{figure}

The cost associated with a warping path $p$, denoted by $C_{p}$, is the sum of the elements of the cost matrix that belong to the warping path:
\begin{equation}
  C_{p}(\mathbf{x},\mathbf{y})=\sum_{l=1}^{L} f(i_{l},j_{l})
\end{equation}
where $f(x,y)$ is the local cost function (usually the Euclidean norm on $\mathbb{R}^{d}$).

The elastic distance score is defined as the minimum cost among all the warping paths:
\begin{equation}
  \text{D}(\mathbf{x},\mathbf{y}) = \min_{p \in \mathcal{P}} C_{p}(\mathbf{x},\mathbf{y})
\end{equation}
where $\mathcal{P}$ is the set of all admissible warping paths. 
The path $p^{*} \in \mathcal{P}$ that satisfies this condition will be referred as optimal path. 

\begin{figure}[!htb]
  \begin{center}
    \begin{subfigure}[t]{0.48\linewidth}
      \centering
      \includegraphics[width=\linewidth,trim=0 -60 0 0, clip]{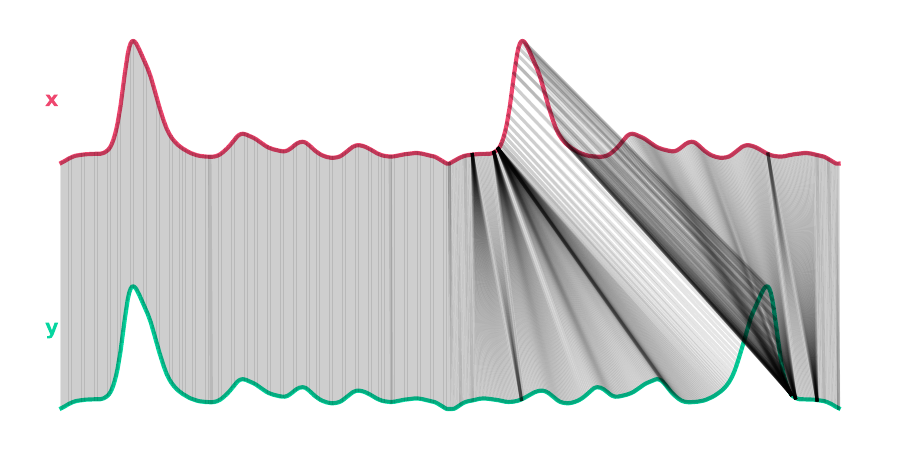}
      \caption{Time series $\mathbf{x}$ and $\mathbf{y}$ of length 10. Gray lines illustrate the one-to-many warping assignments.}
      \label{fig:dtw_time_series_1}
    \end{subfigure}
    \hfill
    \begin{subfigure}[t]{0.48\linewidth}
      \centering
      \includegraphics[width=0.9\linewidth]{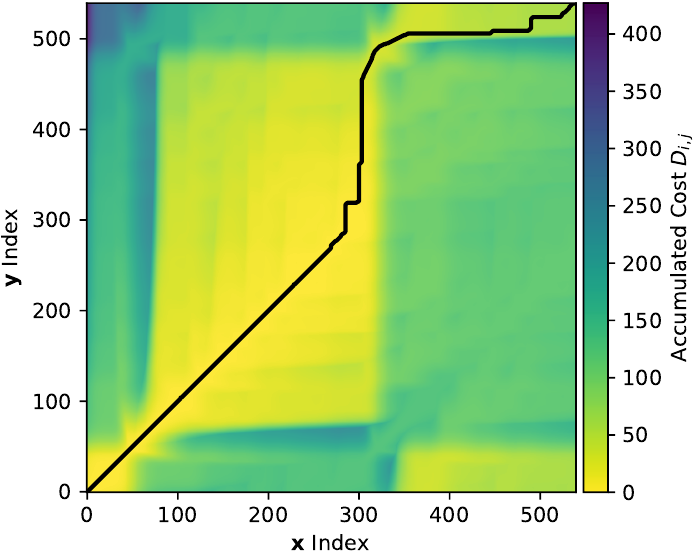}
      \caption{Cost matrix with cumulative distances $D_{i,j}$. Optimal alignment path $p^*$ is represented by the black line.}
      \label{fig:dtw_cost_matrix_1}
    \end{subfigure}
    \caption{Elastic similarity measure between two generic time series $\mathbf{x}$ and $\mathbf{y}$, with $n=m=540$. The cost function corresponds to the Euclidean distance, $f(x,y)=(x-y)^{2}$.}
    \label{fig:dtw_1}
  \end{center}
\end{figure}

Because the number of allowed paths in $\mathcal{P}$ grows exponentially with the size of the time series ($m$ and $n$), brute force is not the best option to consider when solving this equation.
Instead of computing the costs for all the warping paths, a more efficient computation in quadratic time is possible using dynamic programming.
The ESM is computed using the recursion:
\begin{equation}
D_{i,j} = C_{i,j} + \min\{D_{i-1,j}, D_{i-1,j-1}, D_{i,j-1}\}
\end{equation}
where $D_{0,0} = 0$ and $D_{0,j} = D_{i,0} = \infty$ for $i \in \{1,2,\cdots,n\}$ and $j\in \{1,2,\cdots,m\}$. \cref{fig:dtw_2} and \cref{fig:dtw_1} illustrate two examples of elastic similarity measure with different time series lengths. 

\subsubsection{Dynamic Time Warping (DTW) \cite{sakoe1978dynamic}}
For the well-known dynamic time warping, the cost function $f$ corresponds to the Euclidean Distance. Alternatively the squared difference can be used and define the DTW as the square root of $D_{n,m}$, so that the distance score has the same “unit” as the time series, as it is done for the Euclidean distance and more generally for $L_{p}$ norms.
DTW is equal to the Euclidean distance for the case where $n=m$ and only the diagonal of the cost matrix is traversed. 
\begin{equation}
f(x_{i}, y_{j}) = || x_{i} - y_{j} ||_{2}
\end{equation}

DTW is invariant to any monotonically increasing temporal map that would align starting and end times, and holds the following properties: $\forall x, y, \text{DTW}(x,y) \geq 0$ and $\forall x, \text{DTW}(x,x) = 0$. Mathematically speaking, DTW does not satisfy the triangle inequality, even when the local distance measure is a metric. Thus, DTW is not a valid metric since it satisfies neither the triangular inequality nor the identity of indiscernibles \cite{Jain2018b}. 
Extensions of DTW to multivariate time series were proposed in \cite{wollmer2009multidimensional, ten2007multi}.

\subsubsection{Edit Distance \cite{marzal1993computation}}
The Edit or Levensthein distance \cite{levenshtein1966binary} quantifies the similarity between two character sequences by counting the number of operations needed to convert one string into another. Based on the minimum number of operations that are required to transform one series into the other. DTW was initially devised for the specific task of spoken word recognition, and edit distances were introduced for measuring the dissimilarity between two strings.

\begin{equation}
  f(x_{i}, y_{j}) = \left\{\begin{matrix*}[l]
    0 & \text{if} \quad x_{i} == y_{j} \\
    1 & \text{otherwise} \\
  \end{matrix*}\right.
\end{equation}

\subsubsection{Edit Distance for Real Sequences (EDR) \cite{chen2005robust}}
Given two numerical time series, EDR and ERP (next section) also count the number of operations needed to convert two sequences. 
EDR corresponds to the extension of the original edit or Levensthein distance to real-valued time series.
It uses a distance threshold to define when two elements of a series match and the cost of each operation is $1$.

\begin{equation}
f(x_{i}, y_{j}, \delta) = \left\{\begin{matrix*}[l]
    0 & \text{if}  \quad |x_{i} - y_{j}| \leq \delta \\
    1 & \text{otherwise} \\
\end{matrix*}\right.
\end{equation}

\subsubsection{Edit Distance with Real Penalty (ERP) \cite{chen2004marriage}}
It uses a distance threshold to define when two elements of a series match, but also includes a constant penalty that is applied for non-matching elements and where gaps are inserted to create optimal alignments. However, EDR does not satisfy triangular inequality, as equality is relaxed by assuming elements are equal when the distance between them is less than or equal to the threshold $\delta$.

\begin{equation}
f(x_{i}, y_{j}, \delta) = \left\{\begin{matrix*}[l]
    0 & \text{if}  \quad |x_{i} - y_{j}| \leq \delta \\
    |x_{i} - y_{j}| & \text{otherwise} \\
\end{matrix*}\right.
\end{equation}

\clearpage
\subsubsection{Time Warp Edit Distance (TWED) \cite{marteau2008time}}
A combination of DTW and EDR, TWED comprises a mismatch penalty $\lambda$ and a stiffness parameter $\nu$ controlling its elasticity. TWED is a metric and thus one can exploit the triangular inequality to speed up the search in the metric space. It also takes time stamp differences into account and as a result it is able to cope with time series of different sampling rates, including down-sampled time series.
The formulation of TWED corresponds to:
\begin{equation}
D_{i,j} = \min\{D_{i-1,j} + \Tau_{x}, D_{i-1,j-1} + \Tau_{xy}, D_{i,j-1} + \Tau_{y}\}
\end{equation}
for $i = \{1,\cdots,n\}$ and $j = \{1,\cdots,m\}$, with
\begin{equation}
\begin{split}
  \Tau_{x} &= f(x_{i},x_{i-1}) + \nu + \lambda \\
  \Tau_{y} &= f(y_{i},y_{i-1}) + \nu + \lambda \\
  \Tau_{xy} &= f(x_{i},y_{i}) + f(x_{i-1},y_{i-1}) + 2*\nu|i-j| \\
\end{split}
\end{equation}
where $f(x,y)$ can be any $L_{p}$ metric.

\subsubsection{Longest Common Subsequence (LCSS) \cite{vlachos2002discovering}}
LCSS finds the optimal alignment between two series by inserting gaps to find the greatest number of matching pairs. In other words, it finds the longest subsequence that is common to two or more sequences, where a subsequence is defined as a sequence that appears in the same relative order, but where the individual elements are not necessarily contiguous. Given the difficulty to estimate the threshold parameter $\epsilon$, \cite{soleimani2020dlcss} proposed a new similarity method based on LCSS, called DLCSS.
\begin{equation}  
  D(i,j)
  \begin{cases}
    0 & \text{if } i=0 \text{ or } j=0 \\
    1 + D(i-1, j-1) & \text{for } |x_i-y_j|<\epsilon \\
    \max\{D(i-1,j),L(i,j-1)\} & \text{otherwise}
  \end{cases}
\end{equation}

\subsubsection{Limitations}
Despite its advantages over the Euclidean distance to compare time series, DTW has several important limitations:

\begin{enumerate}[label=(\roman*)]
  \item Quadratic algorithmic complexity: $\mathcal{O}(n \times m)$, where $n$ and $m$ are the lengths of both time series. 
  \item Not a metric: it does not satisfy the separation property (the DTW score between two different time series can be zero) and the triangle inequality, meaning that efficient nearest-neighbor search algorithms such as the K-dimensional tree \cite{bentley1975multidimensional} and the ball tree \cite{omohundro1989five} structures cannot be used. Both limitations make nearest-neighbor classification with DTW a computationally intensive algorithm. 
  \item Allows for very large time warps, which may be undesired as it may align unrelated segments of the data. 
  \item Not differentiable, making it difficult to use with machine learning algorithms that rely on minimizing an objective function with gradient descent or a variant.
\end{enumerate}

EDR seems to be very competitive and has an efficacy comparable to DTW, but DTW currently stands as the main benchmark against which new similarity measures need to be compared: \cite{wang2013experimental} performed an extensive comparison of classification accuracies for 9 measures (plus 4 variants) across 38 datasets coming from various scientific domains. One of the main conclusions of the study is that, even though the newly proposed measures can be theoretically attractive, the efficacy of some common and well-established measures is, in the vast majority of cases, very difficult to beat. Specifically, DTW is found to be consistently superior to the other studied measures (or, at worst, for a few datasets, equivalent). In addition, the authors emphasize that the Euclidean distance remains a quite accurate, robust, simple, and efficient way of measuring the similarity between two time series.

\subsection{DTW Variants and Extensions}
Several variants of DTW have been proposed to address one or several limitations of its original version. DTW's quadratic time complexity makes it a time-consuming process, hence different methods are proposed in the literature to speed up the distance measure, with the drawback of deviating from the true DTW distance (efficient lower bound approximations). These methods fall into three categories: adding constraints, abstracting the data and indexing.

\begin{enumerate}[label=(\alph*)]
  \item Constraints: limiting the number of elements in the cost matrix that are evaluated
  \item Abstracting the data: the DTW algorithm is only run on a reduced representation of the data.
  \item Indexing: uses lower bounds to reduce the number of times the DTW algorithm has to be executed.
\end{enumerate}

These methods increase the speed by a constant factor, thus remaining on $\mathcal{O}(n \times m)$.

\subsubsection{Windowing and Global Constraints}

A global constraint, or window, explicitly forbids warping curves to enter some region of the cost matrix. A global constraint translates an a priori knowledge about the fact that the time distortion is limited. These global constraints discard the subset of paths of higher length and (on average) with higher weights; and do not depend on the values of the time series, but only on their lengths. This approach also decreases the computational complexity of the cost and accumulated cost matrices since only the entries belonging to the constraint region have to be computed, but adds the computational cost of the constraint region.

\begin{itemize}
  \item \textbf{Sakoe-Chiba band} \cite{sakoe1978dynamic}: limits the time warps to be no greater than half the bandwidth parameter $w$. Intuitively, the constraint creates an allowed band of fixed width about the main diagonal of the alignment plane. 
  \item \textbf{Slanted window}: creates a band centered around the jagged line segment which joins element (1, 1) to element ($n$, $m$) and is $w$ (window size parameter) elements wide along the first axis. In other words, the “diagonal” goes from one corner to the other of the possibly rectangular cost matrix, therefore having a slope of $m/n$, not 1.
  \item \textbf{Itakura band} \cite{itakura1975minimum}: constraints the region to a parallelogram defined by the window size parameter $w$. Contrary to the Sakoe-Chiba band, whose width is constant, the Itakura parallelogram has a varying width, allowing for larger time shifts in the middle than at the first and last time points.
\end{itemize}

The flexibility of DTW is, for some applications, a double-edged sword so a maximum distortion constraint of 10\%, following Sakoe and Chiba, is imposed.

\begin{figure}[!htb]
  \begin{center}
    \begin{subfigure}[t]{0.32\linewidth}
      \centering
      \includegraphics[width=\linewidth]{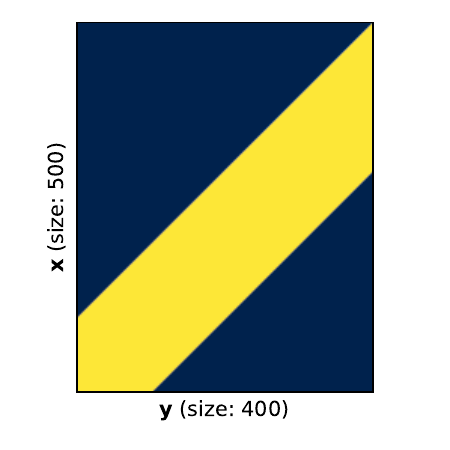}
      \caption{Sakoe-Chiba band}
      \label{fig:dtw_window:sakoechiba}
    \end{subfigure}
    \hfill
    \begin{subfigure}[t]{0.32\linewidth}
      \centering
      \includegraphics[width=\linewidth]{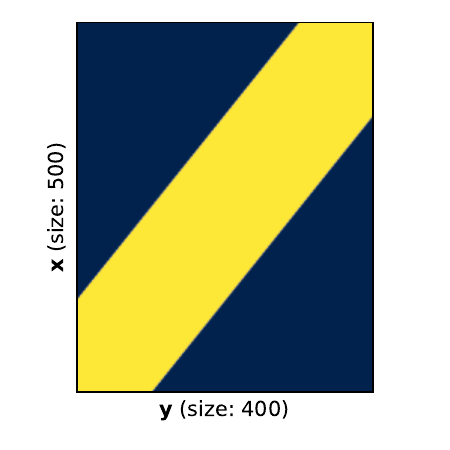}
      \caption{Slanted band}
      \label{fig:dtw_window:slanted}
    \end{subfigure}
    \hfill
    \begin{subfigure}[t]{0.32\linewidth}
      \centering
      \includegraphics[width=\linewidth]{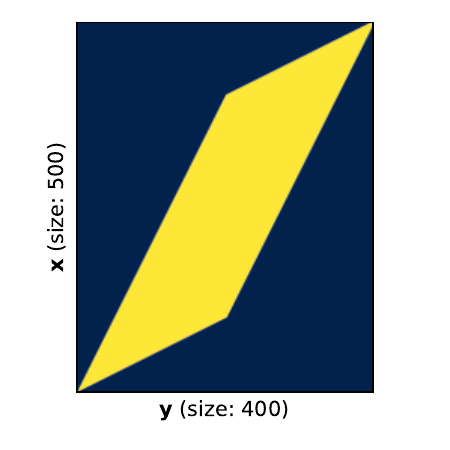}
      \caption{Itakura band}
      \label{fig:dtw_window:itakura}
    \end{subfigure}
    \caption{Windowing functions: example with $n=500$, $m=400$ and $w=100$}
    \label{fig:dtw_window}
  \end{center}
\end{figure}

\subsubsection{Step Patterns and Local Slope Constraints}\label{sec:step_pattern}
The elegant formulation of the DTW alignment problem lends itself to a variety of extensions. 
For example, one usually wants to limit the number of consecutive elements which are skipped in either time series, i.e., are left unmatched. How many repeated elements can be matched consecutively, or how many can be skipped, put limits on the local slope of the warping curve. This property can be controlled by a very flexible scheme called step patterns. 

Step patterns impose a lower and/or an upper bound to the local slope of the alignment. In other words, they limit the maximum amount of time stretch and compression allowed at any point of the alignment.
A variety of classification schemes have been proposed for step patterns (see \cref{fig:step_patterns}), including Sakoe-Chiba \cite{sakoe1978dynamic}; Rabiner-Juang \cite{rabiner1993fundamentals}; and Rabiner-Myers \cite{myers1981comparative}.

\begin{itemize}

\item \textbf{Well-known step patterns}: 
Symmetric recursion allows an unlimited number of elements of the query to be matched to a single element of the reference, and vice-versa; in other words, there is no limit in the amount of time expansion or compression allowed at any point.
\textit{symmetric2} is the normalizable, symmetric, with no local slope constraints. Since one diagonal step costs as much as the two equivalent steps along the sides, it can be normalized dividing by $n+m$ ($\mathbf{x}$+$\mathbf{y}$ length). 
\textit{asymmetric} is asymmetric with slope constrained between 0 and 2, normalized by $n$. Matches each element of the query time series exactly once, so the warping path is guaranteed to be single-valued. 
\textit{symmetric1} (or White-Neely) is quasi-symmetric, no local constraint, non-normalizable. It is biased in favor of oblique steps.

\item \textbf{Slope-constrained step patterns from Sakoe-Chiba} \cite{sakoe1978dynamic}: 
They are called ``symmetricP<s>'' and ``asymmetricP<s>'', where <s> corresponds to Sakoe's integer slope parameter. Values available are accordingly: 0 (no constraint), 1, 05 (one half) and 2. See \cite{sakoe1978dynamic} for details.

\item \textbf{The Rabiner-Myers set} \cite{myers1981comparative}: 
The ``type <g>-<h>'' step patterns follow the older Rabiner-Myers classification proposed in \cite{myers1981comparative} and \cite{myers1980performance}. Note that this is a subset of the Rabiner-Juang set \cite{rabiner1993fundamentals}. ``<g>'' is a Roman numeral specifying the shape of the transitions; ``<h>'' is a letter in the range ``A-D'' specifying the weighting used per step. ``type II<h>'' patterns also have a version ending in 's', meaning the smoothing is used, which does not permit skipping points. The ``type I-D'', ``type II-D'' and ``type II-Ds'' are unbiased and symmetric.

\end{itemize}

\renewcommand\x{0.19}
\begin{figure}[!htb]
  \begin{center}
    \begin{subfigure}[b]{\x\linewidth}
      \centering
      \includegraphics[width=\linewidth]{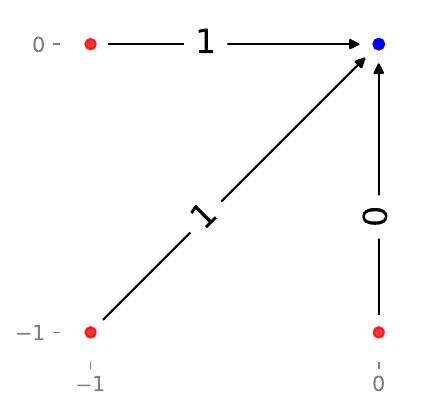}
      \caption{Asymmetric P0}
      \label{fig:step_patterns:asymmetricP0}
    \end{subfigure}
    \hfill
    \begin{subfigure}[b]{\x\linewidth}
      \centering
      \includegraphics[width=\linewidth]{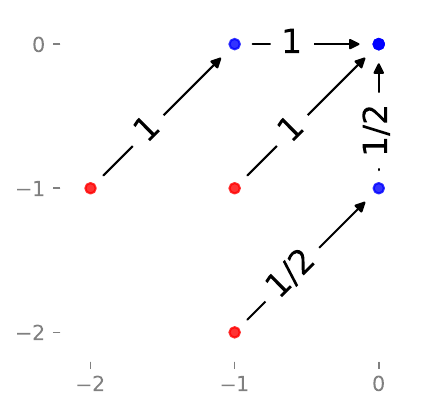}
      \caption{Asymmetric P1}
      \label{fig:step_patterns:asymmetricP1}
    \end{subfigure}
    \hfill
    \begin{subfigure}[b]{\x\linewidth}
      \centering
      \includegraphics[width=\linewidth]{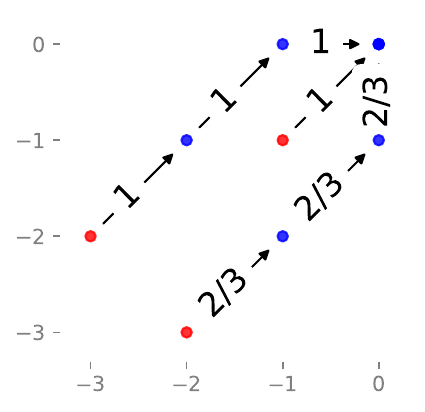}
      \caption{Asymmetric P2}
      \label{fig:step_patterns:asymmetricP2}
    \end{subfigure}
    \hfill
    \begin{subfigure}[b]{\x\linewidth}
      \centering
      \includegraphics[width=\linewidth]{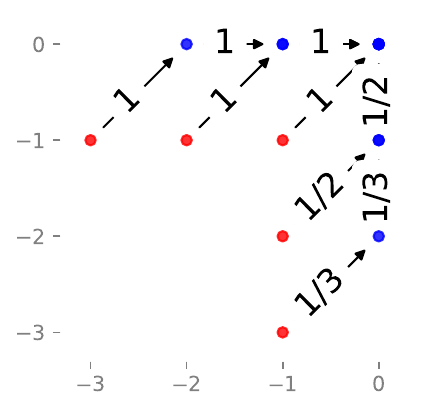}
      \captionsetup{width=1.2\linewidth}
      \caption{Asymmetric P05}
      \label{fig:step_patterns:asymmetricP05}
    \end{subfigure}
    \hfill
    \begin{subfigure}[b]{\x\linewidth}
      \centering
      \includegraphics[width=\linewidth]{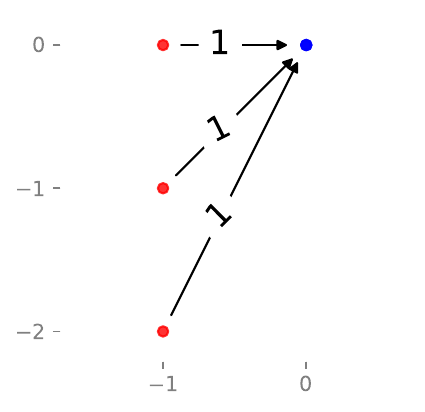}
      \caption{Asymmetric}
      \label{fig:step_patterns:asymmetric}
    \end{subfigure}
    \hfill
    \begin{subfigure}[t]{\x\linewidth}
      \centering
      \includegraphics[width=\linewidth]{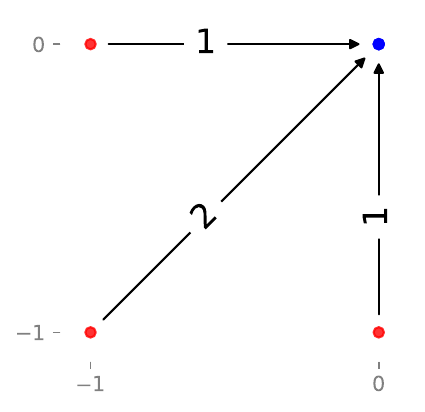}
      \caption{Symmetric P0}
      \label{fig:step_patterns:symmetricP0}
    \end{subfigure}
    \hfill
    \begin{subfigure}[t]{\x\linewidth}
      \centering
      \includegraphics[width=\linewidth]{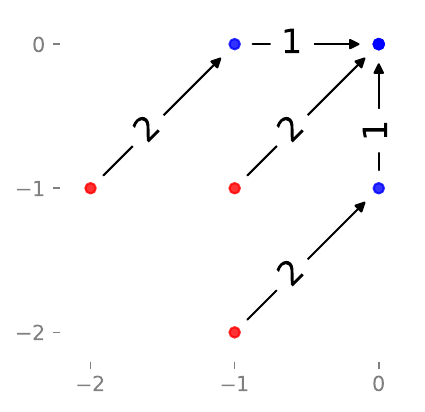}
      \caption{Symmetric P1}
      \label{fig:step_patterns:symmetricP1}
    \end{subfigure}
    \hfill
    \begin{subfigure}[t]{\x\linewidth}
      \centering
      \includegraphics[width=\linewidth]{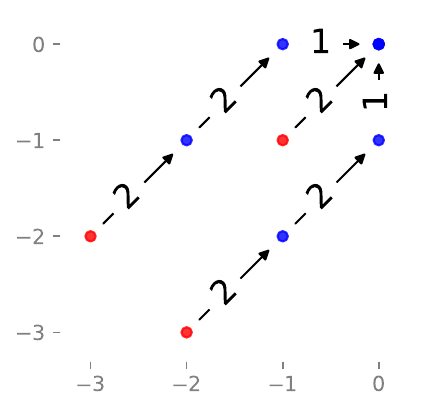}
      \caption{Symmetric P2}
      \label{fig:step_patterns:symmetricP2}
    \end{subfigure}
    \hfill
    \begin{subfigure}[t]{\x\linewidth}
      \centering
      \includegraphics[width=\linewidth]{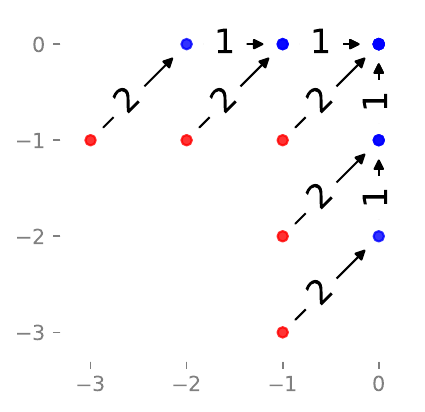}
      \caption{Symmetric P05}
      \label{fig:step_patterns:symmetricP05}
    \end{subfigure}
    \hfill
    \begin{subfigure}[t]{\x\linewidth}
      \centering
      \includegraphics[width=\linewidth]{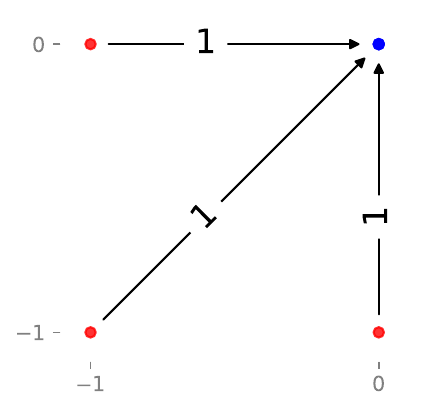}
      \caption{Symmetric 1}
      \label{fig:step_patterns:symmetric1}
    \end{subfigure}
    \hfill
    \begin{subfigure}[t]{\x\linewidth}
      \centering
      \includegraphics[width=\linewidth]{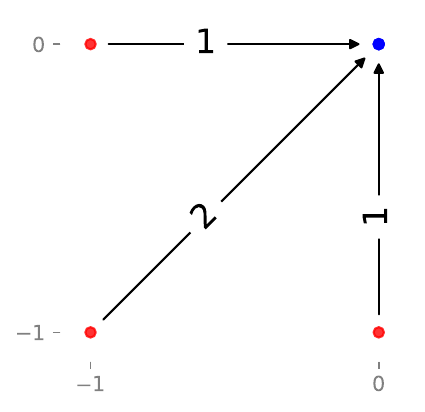}
      \caption{Symmetric 2}
      \label{fig:step_patterns:symmetric2}
    \end{subfigure}
    \hfill
    \begin{subfigure}[t]{\x\linewidth}
      \centering
      \includegraphics[width=\linewidth]{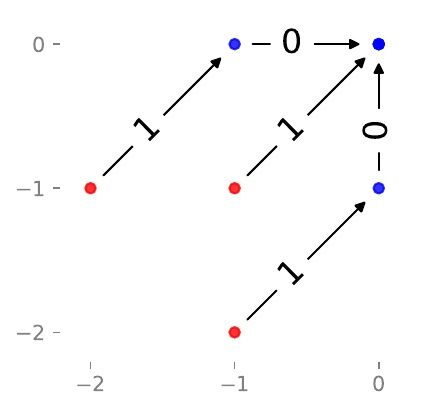}
      \caption{Type I-A}
      \label{fig:step_patterns:typeIa}
    \end{subfigure}
    \hfill
    \begin{subfigure}[t]{\x\linewidth}
      \centering
      \includegraphics[width=\linewidth]{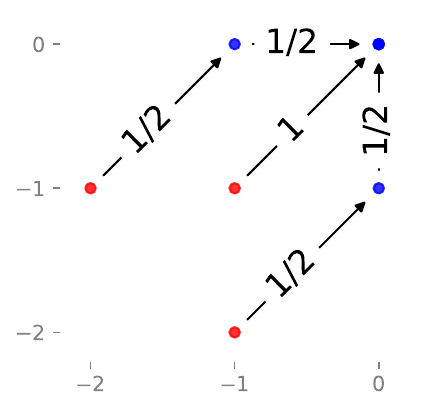}
      \caption{Type I-As}
      \label{fig:step_patterns:typeIas}
    \end{subfigure}
    \hfill
    \begin{subfigure}[t]{\x\linewidth}
      \centering
      \includegraphics[width=\linewidth]{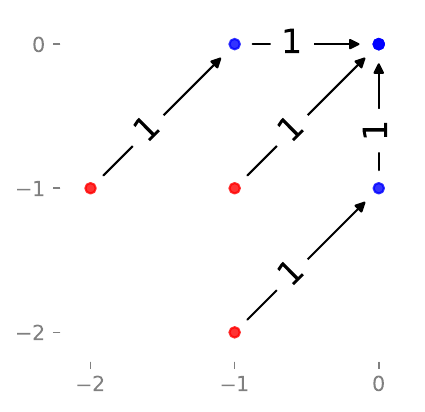}
      \caption{Type I-B}
      \label{fig:step_patterns:typeIb}
    \end{subfigure}
    \hfill
    \begin{subfigure}[t]{\x\linewidth}
      \centering
      \includegraphics[width=\linewidth]{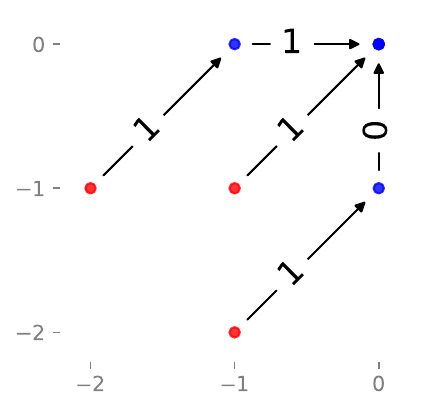}
      \caption{Type I-C}
      \label{fig:step_patterns:typeIc}
    \end{subfigure}
    \hfill
    \begin{subfigure}[t]{\x\linewidth}
      \centering
      \includegraphics[width=\linewidth]{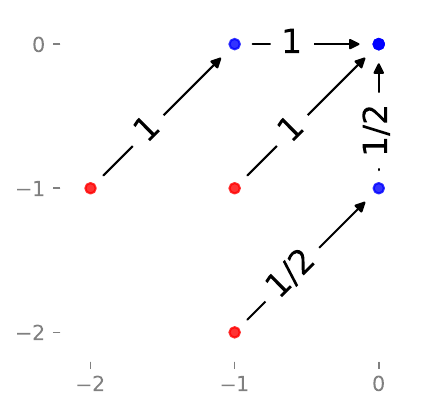}
      \caption{Type I-Cs}
      \label{fig:step_patterns:typeIcs}
    \end{subfigure}
    \hfill
    \begin{subfigure}[t]{\x\linewidth}
      \centering
      \includegraphics[width=\linewidth]{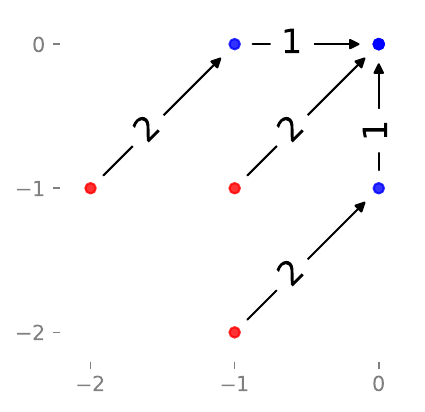}
      \caption{Type I-D}
      \label{fig:step_patterns:typeId}
    \end{subfigure}
    \hfill
    \begin{subfigure}[t]{\x\linewidth}
      \centering
      \includegraphics[width=\linewidth]{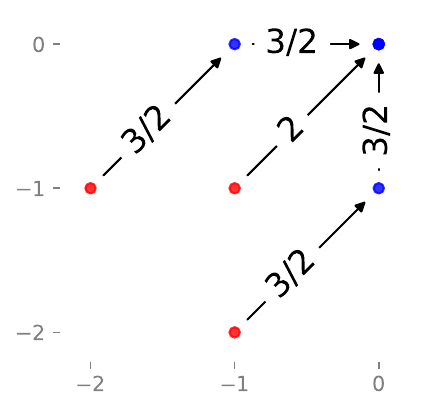}
      \caption{Type I-Ds}
      \label{fig:step_patterns:typeIds}
    \end{subfigure}
    \hfill
    \begin{subfigure}[t]{\x\linewidth}
      \centering
      \includegraphics[width=\linewidth]{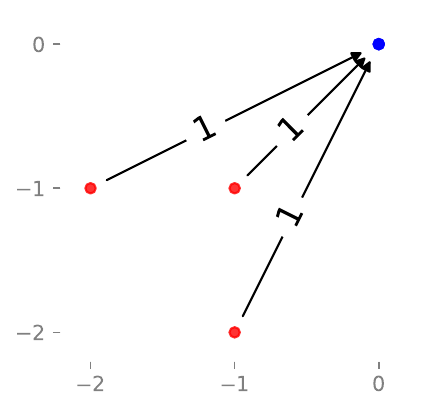}
      \caption{Type II-A}
      \label{fig:step_patterns:typeIIa}
    \end{subfigure}
    \hfill
    \begin{subfigure}[t]{\x\linewidth}
      \centering
      \includegraphics[width=\linewidth]{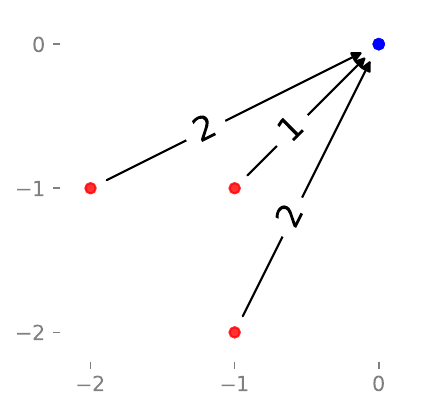}
      \caption{Type II-B}
      \label{fig:step_patterns:typeIIb}
    \end{subfigure}
    \hfill
    \begin{subfigure}[t]{\x\linewidth}
      \centering
      \includegraphics[width=\linewidth]{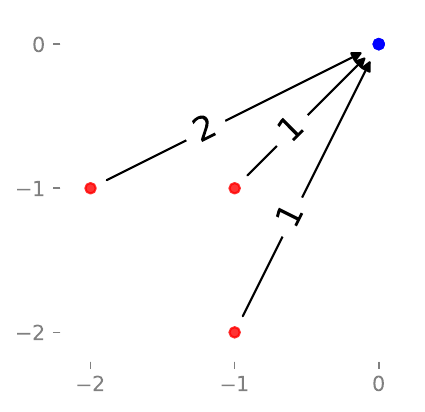}
      \caption{Type II-C}
      \label{fig:step_patterns:typeIIc}
    \end{subfigure}
    \hfill
    \begin{subfigure}[t]{\x\linewidth}
      \centering
      \includegraphics[width=\linewidth]{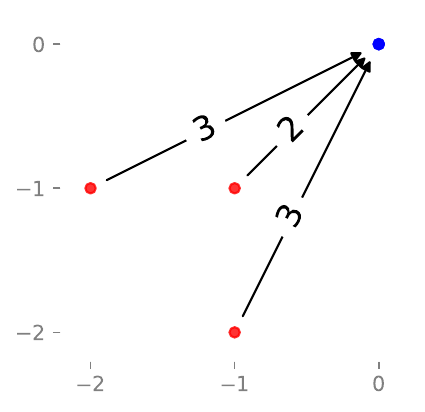}
      \caption{Type II-D}
      \label{fig:step_patterns:typeIId}
    \end{subfigure}
    \hfill
    \begin{subfigure}[t]{\x\linewidth}
      \centering
      \includegraphics[width=\linewidth]{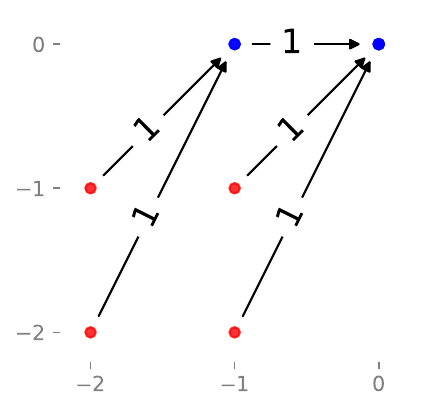}
      \caption{Type III-C}
      \label{fig:step_patterns:typeIIIc}
    \end{subfigure}
    \hfill
    \begin{subfigure}[t]{\x\linewidth}
      \centering
      \includegraphics[width=\linewidth]{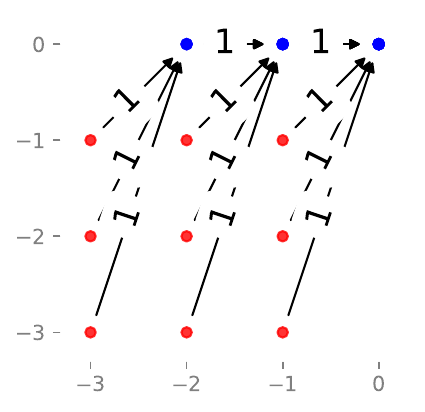}
      \caption{Type IV-C}
      \label{fig:step_patterns:typeIVc}
    \end{subfigure}
    \hfill
    \begin{subfigure}[t]{\x\linewidth}
      \centering
      \includegraphics[width=\linewidth]{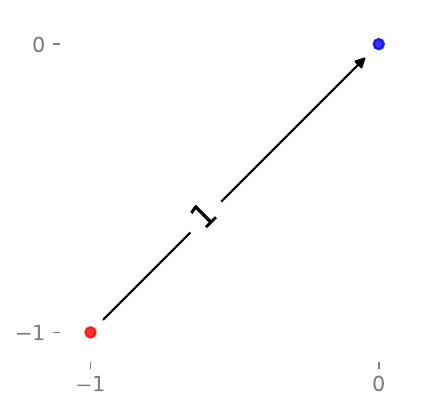}
      \caption{Unitary}
      \label{fig:step_patterns:unitary}
    \end{subfigure}
    \caption{Step patterns}
    \label{fig:step_patterns}
  \end{center}
\end{figure}

\clearpage
\subsubsection{FastDTW \cite{salvador2007toward}}
It improves upon traditional DTW by using a constrained search strategy to find the optimal path between the two series, allowing for faster computation times. 
FastDTW is an approximation algorithm that has a time complexity of $\mathcal{O}(\max(n,m))$. FastDTW reduces the dimensions of the time series by averaging adjacent pairs of points in time, then finds the minimum warping path through local adjustments by considering neighbouring cells in the original local cost matrix. Even though FastDTW has a linear time complexity, accurate approximations come with large constant factors in time complexity. 

\begin{figure}[!htb]
  \begin{center}
    \begin{subfigure}[t]{0.22\linewidth}
      \centering
      \includegraphics[width=\linewidth]{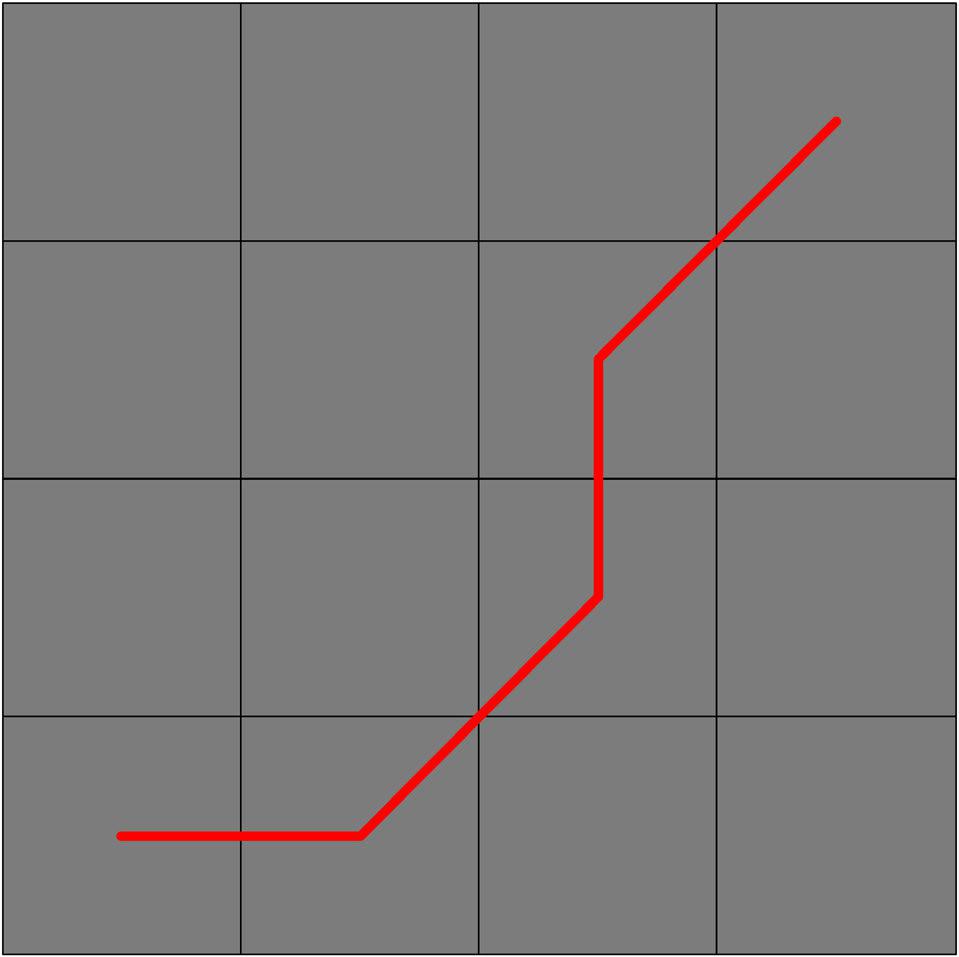}
      \caption{Resolution 1/8}
      \label{fig:fastdtw_1}
    \end{subfigure}
    \hfill
    \begin{subfigure}[t]{0.22\linewidth}
      \centering
      \includegraphics[width=\linewidth]{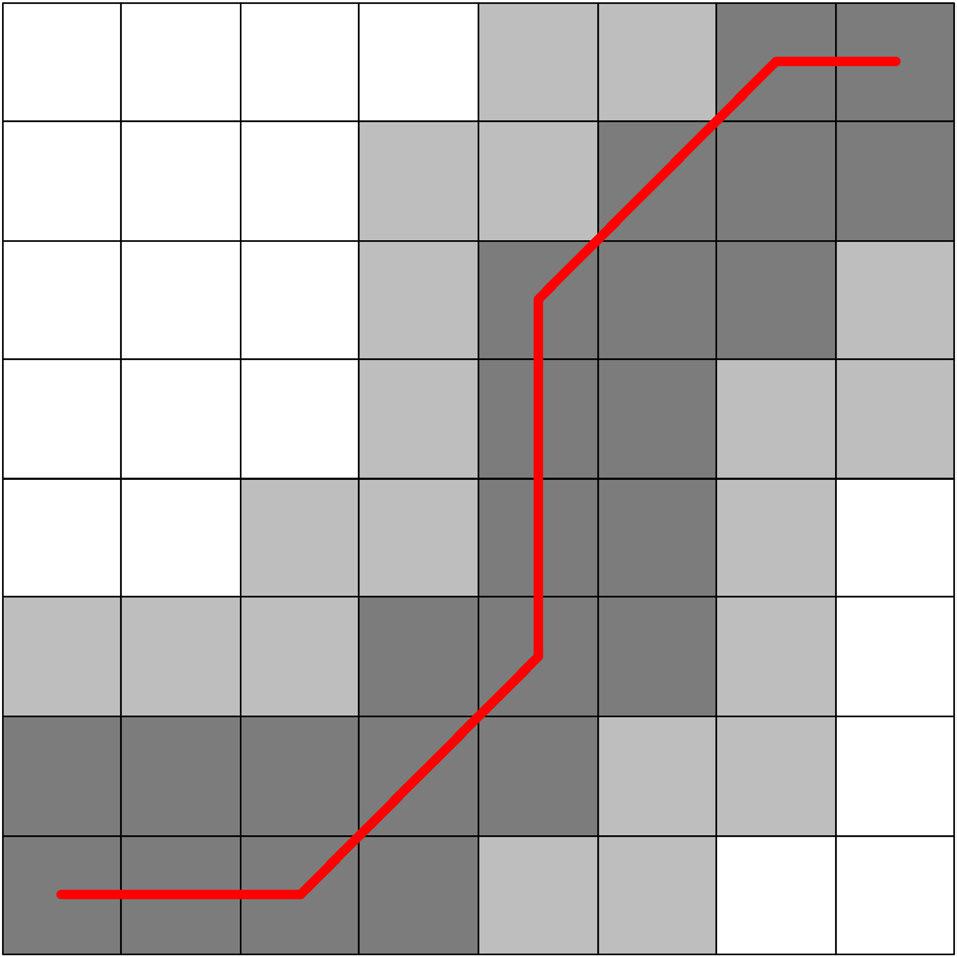}
      \caption{Resolution 1/4}
      \label{fig:fastdtw_2}
    \end{subfigure}
    \hfill
    \begin{subfigure}[t]{0.22\linewidth}
      \centering
      \includegraphics[width=\linewidth]{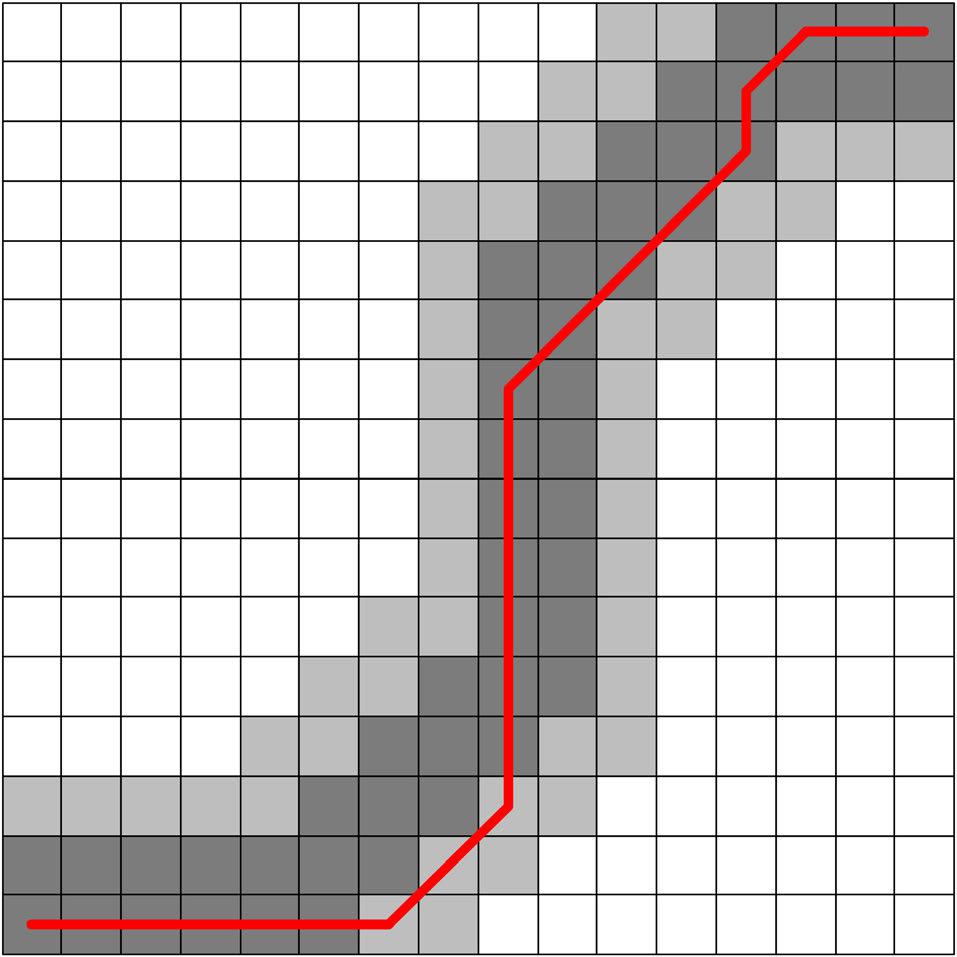}
      \caption{Resolution 1/2}
      \label{fig:fastdtw_3}
    \end{subfigure}
    \hfill
    \begin{subfigure}[t]{0.22\linewidth}
      \centering
      \includegraphics[width=\linewidth]{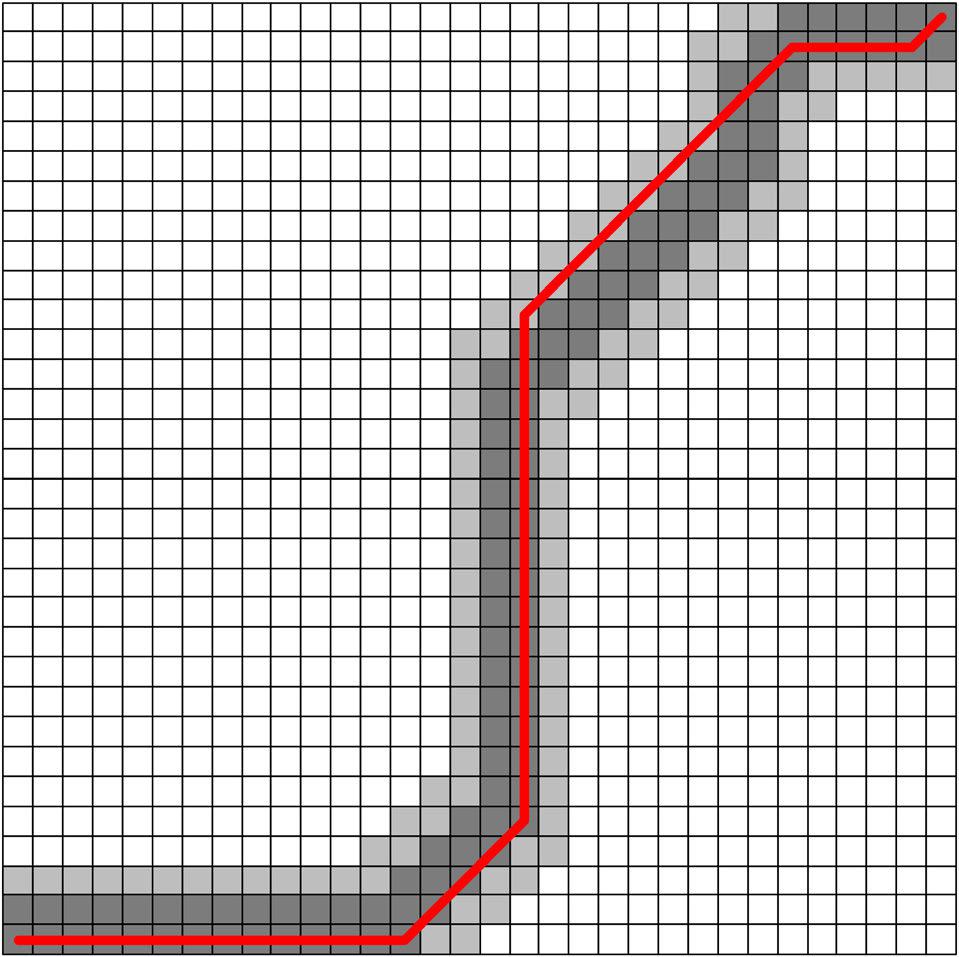}
      \caption{Original resolution}
      \label{fig:fastdtw_4}
    \end{subfigure}
    \caption{Accumulated cost matrix of four resolutions for the FastDTW algorithm}
    \label{fig:fastdtw}
  \end{center}
\end{figure}

\subsubsection{Derivative Dynamic Time Warping (DDTW) \cite{keogh2001derivative}}
It first transforms the series into a series of first order differences. The motivation for DDTW was to introduce a measure that avoids singularities, where a single point on one series may map onto a large subsection of another time series and create pathological results. This avoids having a single point of one series to map to many point in another series, which can negatively impact the DTW distance. DDTW is designed to mitigate against noise in the series that can adversely affect DTW.  Given a time series $\mathbf{x}=\{x_{1},x_{2},\cdots,x_{n}\}$, the difference series $z_i$ is defined as the average of slopes between $x_{i-1}$ and $x_{i}$ and $x_{i}$ and $x_{i+1}$.

\subsubsection{Weighted Dynamic Time Warping (WDTW) \cite{jeong2011weighted}}
It adds a multiplicative weight penalty based on the warping distance between points in the warping path. Specifically, WDTW uses a logistic weight function to add penalties to the local cost matrix to favour reduced warping, which brings more balance between shape matching and alignment in time. It favours reduced warping, and is a smooth alternative to the cutoff point approach of using a warping window. 

\subsubsection{ShapeDTW \cite{zhao2018shapedtw}}
ShapeDTW enhances DTW by taking point-wise local structural information into consideration. It attempts to pair locally similar structures and to avoid matching points with distinct neighborhood structures. 
It first represents each temporal point by some shape descriptor, which encodes structural information of local subsequences around that point; in this way, the original time series is converted into a sequence of descriptors. Then, it uses DTW to align two sequences of descriptors. Since the first step takes linear time while the second step is a typical DTW, which takes quadratic time, the total time complexity is quadratic, $\mathcal{O}(mn)$.

\subsubsection{Generalized Dynamic Time Warping (GDTW) \cite{neamtu2018generalized}}

Generalized Dynamic Time Warping (GDTW) generalizes DTW and extends its warping capabilities to a rich diversity of point-to-point distances, and supports alignment of a large array of domain-specific distances in a uniform manner. 

\subsubsection{DTW with Limited Length (LDTW) \cite{zhang2017dynamic}}

LDTW limits the total number of links during the optimization process of DTW, effectively reducing the pathological alignment problem while being flexible enough to measure similarities. LDTW sets a global and softer constraint instead of other local and rigid ones
used by existing variants of DTW, because it lets the optimization process of DTW decide how many links to allocate to each data point and where to put these links.

\subsection{Kernel methods}

Kernel methods rely on a kernel function measuring similarity between any pair of inputs. A key necessary assumption of kernel methods is that the kernel is positive-definite. As mentioned in the previous section, DTW is not a distance because it does not satisfy the triangle inequality, implying that DTW cannot be used to define a positive-definite kernel. Although DTW has been used with kernel methods in several publications with some tricks, the fact that the theoretical assumptions are not satisfied is an important limitation.

\clearpage
\subsubsection{Global Alignment Kernel (GAK) \cite{cuturi2011fast}}
Denoted  by $k_{GA}^\gamma$, the global alignment kernel is a true positive-definite kernel proposed for time series, and is defined as the sum of all the negatively exponentiated costs over all the possible warping paths:
$k_{GA}^{\gamma}=\sum_{p\in \mathcal{P}} \exp \left( -{C_{p}(\mathbf{x},\mathbf{y})}/{\gamma} \right)$, where $C_{p}(\mathbf{x},\mathbf{y})$ is the cost associated with the warping path $p$, $\mathcal{P}$ is the set of all the warping paths, and $\gamma > 0$ is a smoothing parameter.

The global alignment kernel has the same computational complexity as DTW, that is $\mathcal{O}(n \times m)$, because the score between two time series can be computed using a recurrence relation. Constraint regions such as the Sakoe-Chiba band \cite{sakoe1978dynamic} and the Itakura parallelogram \cite{itakura1975minimum} can also be used with global alignment kernels. Support vector machines with the global alignment kernel have been shown to yield better predictive performances than with other pseudo kernels based on DTW for several multivariate time series classification tasks \cite{cuturi2011fast}.

\subsubsection{Soft-DTW \cite{cuturi2017soft}}
Soft-DTW is DTW's smoothed counterpart and replaces the mininum function in DTW, which is not differentiable, with a smooth minimum function, namely the LogSumExp function, which is differentiable. The soft-DTW function, being differentiable, can be used as a loss function for machine learning algorithms, in particular artificial neural networks. 
The smoothness of the LogSumExp function is controlled by a parameter $\gamma > 0$. In the limit case $\gamma = 0$, softDTW reduces to a hard min operator and is defined as equivalent to the DTW algorithm \cite{vayer2020time}. 
Similarly to DTW, soft-DTW can be computed in quadratic time using dynamic programming:
$\text{soft-min}_{\gamma}(a_{1}, \cdots, a_{n}) = -\gamma \sum_{i} e^{-a_{i}/\gamma}$

\begin{figure}[!htb]
  \begin{center}
    \begin{subfigure}[t]{0.28\linewidth}
      \centering
      \includegraphics[width=\linewidth]{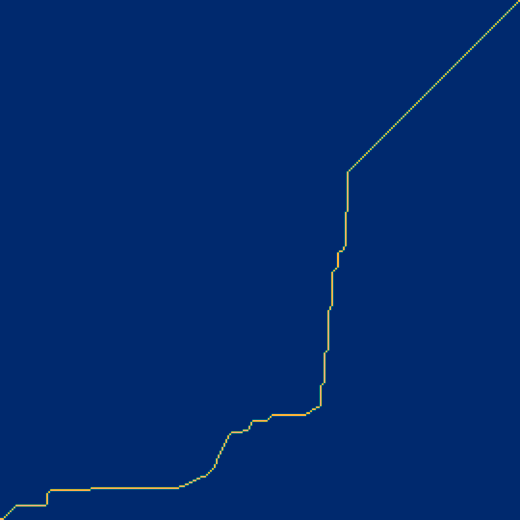}
      \caption{$\gamma = 0.0$}
      \label{fig:softdtw:00}
    \end{subfigure}
    \hfill
    \begin{subfigure}[t]{0.28\linewidth}
      \centering
      \includegraphics[width=\linewidth]{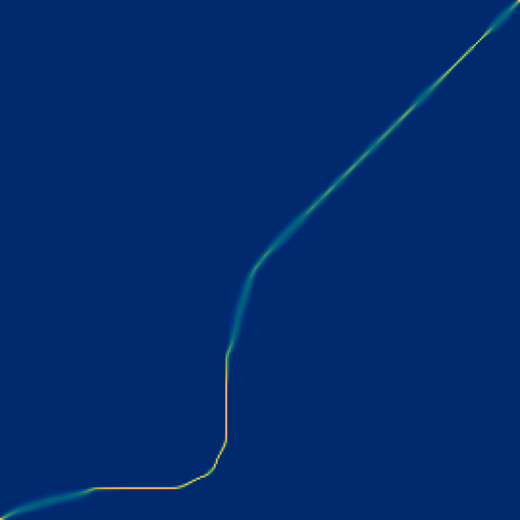}
      \caption{$\gamma = 0.1$}
      \label{fig:softdtw:01}
    \end{subfigure}
    \hfill
    \begin{subfigure}[t]{0.28\linewidth}
      \centering
      \includegraphics[width=\linewidth]{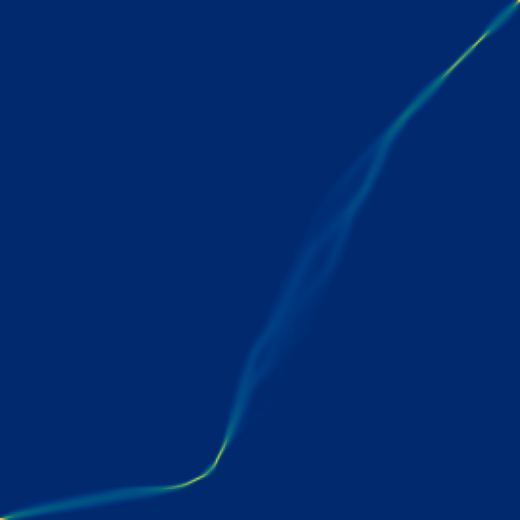}
      \caption{$\gamma = 1.0$}
      \label{fig:softdtw:10}
    \end{subfigure}
    \caption{Optimal soft alignment matrices between two time series. Adapted with permission from \cite{JMLR:v21:20-091}.}
    \label{fig:softdtw}
  \end{center}
\end{figure}

\subsubsection{DTWNet \cite{cai2019dtwnet}}

DTWNet is a neural network with learnable DTW kernels. This is not explicitly a similarity measure, but one can use this architecture to obtain better feature extraction and consequently improve other task (classification, clustering, etc.). To learn the DTW kernel, a stochastic backpropagation method based on the warping path is proposed, that is able to compute the gradient of the dynamic programming process in DTW.

\subsection{Bag-of-words (dictionary-based, feature-based)}\label{sec:bag_words}

Bag-of-words methods represent each time series as a collection of individual words, typically derived from the time series data itself. The similarity between two time series is then calculated based on the number of common words shared between the two series. This approach allows for efficient computation and is effective in capturing the important characteristics of time series data.

These methods first extract features from the time series and then measure the distance between these features.
A simple extension of the Euclidean distance is not to compute it directly using the raw time series, but using features extracted from it.
Feature-based distances are often applied to obtain a reduction in both dimensionality and noise.
Thus, bag-of-words methods consist in discretizing time series into sequences of symbols, then extracting words from these sequences with a sliding window, and finally counting the number of words for all the words in the dictionary. These type of classifiers discriminate time series by the frequency of repetition of some subseries.

Finding a trade-off between the accuracy and speed is a controversial and non-trivial task in representation methods: as more dimensionally reduction occurs, more data is lost and becomes inaccurate, and consequently less time execution. 

\paragraph{Symbolic Aggregation approXimation (SAX) \cite{lin2003symbolic}}
It makes a symbolic representation of the time series, mapping each real-valued observation of the time series to its corresponding bin and consists of two steps: first, reducing the time series using Piecewise Aggregate Approximation (PAA) (divides the time series horizontally) and second, assigning a letter or discrete label to every PAA element.
Several strategies to compute the bin edges are possible: quantiles of the standard normal distribution if the time series was standardized (zero mean, unit variance), uniform bins based on the extreme values of the time series, or quantiles of the time series.

\paragraph{Bag-of-patterns (BOP) \cite{lin2012rotation}}
It uses a sliding window to extract words from the discretized time series and computes the corresponding histogram.

\paragraph{Symbolic Aggregation approXimation in Vector Space Model (SAX-VSM) \cite{senin2013sax}}
A sliding window is applied to the raw time series to extract subsequences. Classification is based on a simple numerical statistic used in natural language processing: the term frequency—inverse document frequency (TF-IDF) matrix \cite{ramos2003using}.

\paragraph{Discrete Fourier Transform (DFT)} 
It transforms the time series from the \textit{time-domain}" $x(t)$ to a \textit{frequency-domain} representation $X(f)$. The Fourier transform decomposes a time series into all different cycles (amplitude, offset and rotation speed).
Spectral representations have the advantage of limiting the temporal dependence and integrating well with established signal processing techniques \cite{weylandt2021automatic}. 
DFT is calculated by the inner product of the time series and a sine wave:

\begin{equation}
  X(f) = \sum_{t=0}^{n-1} x_t \, e^{-i \frac{2\pi f}{n}t}
\end{equation}

The resulting vector $X(f)$ if a vector of $n$ complex numbers. The inverse DTF transform a collection of frequencies $X(f)$ back to the time-domain:

\begin{equation}
  x(t) =\frac{1}{n}\sum_{f=0}^{n-1} X(f) \, e^{i \frac{2\pi f}{n}t}
\end{equation}

Following Parseval's theorem \cite{parseval1806memoire}, the DTF preserves the Euclidean distance between two time series: When all the frequencies are use, the Euclidean distance between two Fourier transforms is equal to the Euclidean distance between the raw/original time series. This is caused by the fact that DFT is a linear transformation.
However, most of the energy in real-world signals is concentrated in the low frequencies.  This is where an advantage of DFT arises: only a limited number of low frequencies $q$ are required to obtain a good approximation of the original time series. One usually takes the opportunity to filter out high-frequency coefficients. 
According to the Nyquist-Shannon sampling theorem, only the first $\frac{n}{2}$ of fewer frequencies should be used (usually the upper bound $q=\frac{n}{2}$ is used). Besides the advantage of dimensionality reduction, disregarding the higher frequencies also denoises the time series in the approximation.
This has the effect of removing rapidly-fluctuating components of the signal. Hence, if high frequencies are not relevant for the intended analysis, or we have some high-frequency noise, this operation will usually carry some increase in accuracy. Furthermore, similarity computations can be substantially accelerated.

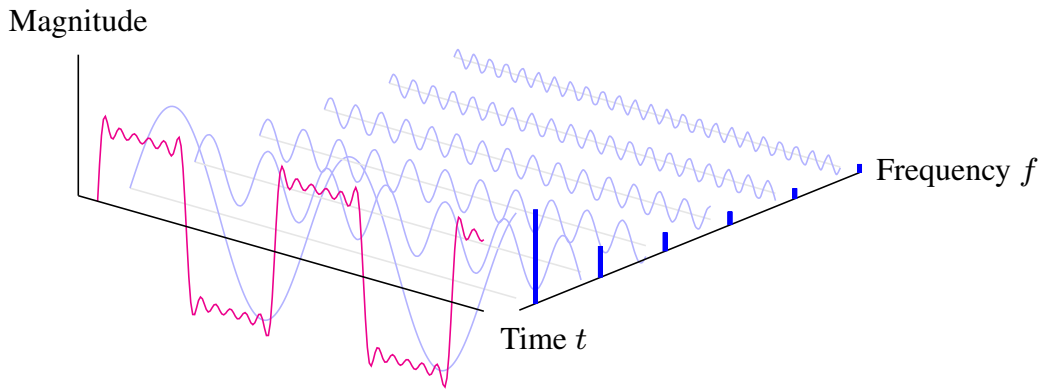
\begin{figure}[!t]
    \begin{center}
    \scalebox{1.5}{
\begin{tikzpicture}
    
    \begin{axis}[
        set layers=standard,
        domain=0:10,
        samples y=1,
        view={40}{20},
        hide axis,
        unit vector ratio*=1 2 2,
        xtick=\empty, ytick=\empty, ztick=\empty,
        clip=false
    ]
    \def\sumcurve{0}
    \pgfplotsinvokeforeach{0.5,1.5,...,5.5}{
        \draw [on layer=background, gray!20] (axis cs:0,#1,0) -- (axis cs:10,#1,0);
        \addplot3 [on layer=main, blue!30, smooth, samples=101] (x,#1,{sin(#1*x*(157))/(#1*2)});
        
        \addplot3 [on layer=axis foreground, very thick, blue,ycomb, samples=2] (10.5,#1,{1/(#1*2)});
        \xdef\sumcurve{\sumcurve + sin(#1*x*(157))/(#1*2)}
    }
    \addplot3 [magenta, samples=200] (x,0,{\sumcurve});
        
    \draw [on layer=axis foreground]  (axis cs:-0.5,0,0) -- (axis cs:10,0,0) node[below right] {\scriptsize Time $t$};
    \draw (axis cs:10.5,0.25,0) -- (axis cs:10.5,5.5,0) node[right] {\scriptsize Frequency $f$};
    \draw (axis cs:-0.5,0,0) -- (axis cs:-0.5,0,1.5) node[above] {\scriptsize Magnitude};
    \end{axis}
\end{tikzpicture}
}
\caption{Fourier transform: illustration of time-frequency correspondence}
\label{fig:}
\end{center}
\end{figure}

DFT has a time complexity of $\mathcal{O}(n^2)$ due to the matrix-vector multiplication. This can be improved to $\mathcal{O}(n\log(n))$ by using the Fast Fourier Transform (FFT) algorithms, which compute the same results by factorizing the DFT matrix into a product of sparse factors to avoid redundant calculations. Calculating the distance, euclidean for example, between two time series based on the Fourier coefficients has time complexity $\mathcal{O}(q)$, and along with the $\mathcal{O}(n\log(n))$ time required for the DFT, the distance matrix for $N$ time series takes $\mathcal{O}(Nn\log(n) + qN^2)$.

\paragraph{Discrete Wavelet Transform (DWT) \cite{haar1909theorie}} 
DWT is a dimensionality reduction method that also reduces noise. It decomposes a time series into a set of bases functions that are called wavelets. A wavelet is a rapid decaying, wave-like oscillation with mean zero, and have finite duration. Wavelets are defined by two functions: the wavelet function (mother) $\psi$ and the scaling function (father) $\varphi$. The mother defines the basic shape, and the father the scale (frequency).
The DWT is obtained by successively passing a time series through high-pass and low-pass filters, which produces detail and approximation coefficients for different levels of decomposition. The main advantage over DFT is that it captures frequency and location in time.

\paragraph{Symbolic Fourier Approximation (SFA) \cite{schafer2012sfa}}
First the discrete Fourier transform of the time series is computed and a subset of the Fourier coefficients is kept. Second, these Fourier coefficients are discretized as in SAX.

\paragraph{Bag-of-SFA-Symbols (BOSS) \cite{schafer2015boss}} First, subsequences of a time series are extracted with a sliding window and the SFA transformation is applied to each subsequence, resulting in an ordered sequence of words. The frequency of each word is computed to obtain the word histogram of the time series.
Several extensions of the BOSS algorithm have been proposed: BOSS VS \cite{schafer2015bag}, WEASEL \cite{schafer2017fast}, SCALE-BOSS \cite{glenis2022scale}.

\subsection{Model-based Distances}
Model-based similarity measures, such as those based on autoregressive (AR) coefficients, are useful tools for comparing the similarity of different time series. By estimating the AR coefficients of a time series and comparing them to those of other time series, it is possible to make inferences about the underlying patterns and dynamics of the data.
In general, model-based measures first need to learn a model of the two time series and then use the obtained model parameters for computing a similarity value. Here we review one of the most popular model, the autoregressive model.

\paragraph{Autoregressive (AR) model coefficients}
An AR model is a statistical model that describes the time-varying behavior of a time series as a linear combination of its past values, as well as any exogenous variables that may be present. The coefficients of this model can be used as a measure of similarity between time series. Specifically, two time series with similar AR coefficients can be considered more similar to each other than two time series with dissimilar AR coefficients.

Let $\mathbf{x}=(x_{1}, \cdots, x_{n})$ be a time series of $n$ real-valued observations, an autoregressive (AR) model of the form: $x_{i} = a_{0} + \sum_{j=1}^{p} a_{j} x_{i-j}$ where $a_j$ denotes the $j$th regression coefficient and $p$ is the order of the model. There are several methods for estimating the AR coefficients of a time series: least squares method, maximum likelihood method and the Yule-Walker equations. Then, the dissimilarity between two time series can be calculated, for instance, using the Euclidean distance between their estimated coefficients. The number of AR coefficients is controlled by the parameter $p$ which, similarly with Fourier coefficients directly affects the final speed of similarity calculations. 

\paragraph{Shapelets}
Shapelets are defined as subsequences that are in some sense maximally representative of a class \cite{ye2009time}.
This family of algorithms focuses on finding relatively short repeated subsequences or time-independent patterns to identify a certain class. 
Ideally, a good shapelet candidate should be a sub-sequence similar to time series from the same class, and dissimilar to time series from other classes.

Let $\mathbf{x}=(x_{1}, \cdots, x_{n})$ be a time series of $n$ real-valued observations and $\mathbf{s}=(s_{1}, \cdots, s_{l})$ be a shapelet of $l$ real numbers, with $l \leq n$. The distance $D(\mathbf{s},\mathbf{x})$ between the shapelet $\mathbf{s}$ and the time series $\mathbf{x}$ is defined as the minimum of the squared Euclidean distances between $\mathbf{s}$ and all the shapelets of length $l$ from $\mathbf{x}$:
\begin{equation}
  D(\mathbf{s},\mathbf{x})=\min_{j \in \{0,\cdots,n-l\}} || \mathbf{s} - \mathbf{x}_{j:j+l}||_{2}^{2} = \min_{j \in \{0,\cdots,n-l\}} \sum_{i=1}^{l} (s_{i} - x_{j+i})^2
\end{equation}

The algorithm extracts all the shapelets whose length belongs to a range, which is a hyperparameter of the algorithm, and selects the $k$ best shapelets, $k$ being another hyperparameter of the algorithm. When the $k$ best shapelets have been identified and the corresponding features have been generated, any standard machine learning classifier can be applied to this new dataset \cite{cheng2020time2graph}. Informally, if we assume a binary classification setting, a shapelet is discriminant if it is present in most series of one class and absent from series of the other class \cite{JMLR:v21:20-091}. 
One limitation of this algorithm is its computational complexity. For a dataset of $N$ time series of length $n$, there are $N \times (n - l + 1)$ shapelets of length $l$. Since many values of $l \in \{1,\cdots,n\}$ are investigated, the maximal computational complexity is $\mathcal{O}(N \times n^2)$.

\paragraph{Siamese Neural Network (SNN) \cite{hou2019time}}
A Siamese neural network is a class of neural network architecture that contains two or more identical sub-networks that share the same weights and configuration. Parameter updating is mirrored across both sub-networks and it is used to find similarities between inputs by comparing its feature vectors. To conveniently train such model, the authors use the label information of datasets and construct a new binary dataset which each example contains two original time series and a binary label. SNNs require a large amount of labeled data in order to be trained effectively, and may not perform well on tasks that require more complex comparisons, such as determining the similarity of long documents or sequences.

\clearpage

\section{Conclusions}
\label{sec:conclusions_1}

Computing the \textbf{similarity between two time series} is a complex and multifaceted problem. Traditional distance measures may not well-suited to this task, due to the time-dependent nature of time series data and the presence of noise and other sources of variability. Feature-based distances are more convenient when domain knowledge is available and the set of extracted features can summarize the most relevant information of the time series.  

Elastic similarity metrics (ESM) have been successful at measuring the distance between two time series by aligning their corresponding elements and computing an adjusted metric between them. However, they suffer from several limitations that hinder their effectiveness: \textbf{computational complexity, non-differentiability and sensitivity to noise and outliers}.
ESMs are highly susceptible to the presence of noise and outliers in the input data, which can lead to incorrect distance measurements, inaccurate alignment and as a consequence, and a degraded performance in time series models that depend on these metrics for pattern recognition tasks.
ESMs are computationally intensive algorithms, and their quadratic time complexity with respect to the length of the time series makes them impractical for large-scale datasets and real-time applications. This issue can complicate to use these metrics in deep learning applications where large amounts of data are often required for training and evaluation.
Soft-DTW proposes a differentiable solution for elastic alignment, but has also some limitations: it is susceptible to the soft minimum parameter $\gamma$, can be negative, which is a nuisance when used as a loss, and more problematically, it is never minimized when the two time series are equal and the squared Euclidean cost is used.

To overcome these limitations, this thesis proposes novel parametric alignment methods that are well-suited for deep learning methods. These alignment methods should be differentiable, robust to noise and outliers, computationally efficient, and expressive and flexible enough to capture complex patterns in the data. 
To comply with such requirements, in \textbf{\cref{chapter:2}} we propose fast and efficient one-dimensional diffeomorphic transformations, which are functions that are differentiable, invertible and have a differentiable inverse. 

\graphicspath{{content/chapter2/}}

\chapter{Closed-Form Diffeomorphic Transformations}\label{chapter:2}
\begingroup
\hypertarget{chapter2}{}
\hypersetup{linkcolor=black}
\setstretch{1.0}
\minitoc
\endgroup

\section{Introduction}\label{sec:introduction_2}

Diffeomorphic transformations, also known as diffeomorphisms, are mathematical transformations that preserve the structure and smoothness of a given space or object.
Diffeomorphisms are invertible and continuous, meaning that they can be reversed and smoothly deformed without breaking or tearing. 
They play a crucial role in many areas of mathematics, physics, engineering and computer graphics as they provide a way to compare and analyze objects that may appear to be different, but are actually equivalent under such transformations. This makes them a powerful tool for studying the geometric and topological structures of shapes and surfaces.

Formally, a $C^1$ \textbf{diffeomorphism}
\footnote{A function $f$ is a $C^k$-diffeomorphism if and only if $f$ is a bijection of class $C^k$ with an inverse of class $C^k$.}
is a differentiable invertible map with a differentiable inverse \cite{duistermaat2000lie}. 
Diffeomorphisms belong to the group of homeomorphisms that preserve topology by design, and thereby are a natural choice in the context of nonlinear time warping, where continuous, differentiable and invertible functions are preferred \cite{mumford2010pattern, bertrand2016}. 
Diffeomorphic warping functions $\boldsymbol{\phi}(x, t)$ can be generated via integration of regular stationary or time-dependent velocity fields $v(x,t)$ specified by the ordinary differential equation (ODE) in \cref{eq:ODE}
\cite{Freifeld2017, Detlefsen2018, Ouderaa2021}. 
\begin{equation}\label{eq:ODE}
\cfrac{\partial\boldsymbol{\phi}(x, t)}{\partial t} = v(\boldsymbol{\phi}, t)
\end{equation}
where $x$ represents the transformable spatial dimension, and $t$ represents the integration time.

By parametrizing the transformation via integration of regular stationary or time-dependent velocity fields and imposing sufficient regularization, diffeomorphic transformations can be assured.
A key advantage to defining diffeomorphisms via velocity fields is that while a diffeomorphism family is always in the nonlinear space, its corresponding velocity-field family is usually a linear space \cite{vaillant2004statistics}.

Under such gradient-based optimization frameworks, neural networks that include diffeomorphic transformations require calculating derivatives to the differential equation's solution with respect to the model parameters, i.e. sensitivity analysis. 
Unfortunately, deep learning frameworks typically lack automatic-differentiation-compatible sensitivity analysis methods; and implicit functions, such as the solution of ODE, require particular care. 
Current solutions appeal to adjoint sensitivity methods \cite{chen2019neural}, ResNet's Eulerian discretization  \cite{Huang2021} or ad-hoc numerical solvers and automatic differentiation \cite{Freifeld2017}. More on this in \cref{sec:related_work_2}.

In contrast, our proposal is to formulate the \textbf{closed-form}\footnote{
A closed-form expression may use constants, variables and 
a finite number of standard operations 
($+,-,\times,\div$), and functions 
(e.g., $\sqrt[n]{\;}$, $\exp$, $\log$, $\sin$, $\sinh$), but no limit, differentiation, or integration.
\label{fn:closedform}
}
\textbf{expression for the ODE's diffeomorphic solution and its gradient} under continuous piecewise-affine (CPA) velocity functions. These closed-form diffeomorphic transformations are presented in the package Diffeomorphic Fast Warping (DIFW)\footnote{\url{https://github.com/imartinezl/difw}}, an open-source library and highly optimized implementation of 1D diffeomorphic transformations on multiple backends for CPU (NumPy and PyTorch with \textit{C++}) and GPU (PyTorch with \textit{CUDA}).

CPA velocity functions yield well-behaved parametrized diffeomorphic transformations, which are efficient to compute, accurate and highly expressive \cite{Freifeld2015}. These finite-dimensional transformations can handle optional constraints (zero-velocity at the boundary) and support convenient modeling choices such as smoothing priors and coarse-to-fine analysis. The term “piecewise” refers to a partition of the temporal domain into sub-intervals. The fineness of the partition controls the trade-off between expressiveness and computational complexity.
Unlike other spaces of highly-expressive velocity fields, integration of CPA velocity fields is given in either closed form (in 1D) or “essentially” closed form for higher dimensions \cite{Freifeld2017}. However, the gradient of CPA-based transformations is only available via the solution of a system of coupled integral equations \cite{freifeld2018deriving}.
Following the first principle of automatic differentiation --- \emph{if the analytical solution to the derivative is known, then replace the function with its derivative}---, in this work we present a closed-form solution for the ODE gradient in 1D that is not available in the current literature. 

\textbf{A closed-form solution provides efficiency, speed and precision}. Fast computation is essential if the transformation must be repeated several times, as is the case with iterative gradient descent methods used in deep learning training. Furthermore, a precise (exact) gradient of the transformation translates to efficient search in the parameter space, which leads to better solutions at convergence. 
In addition, closed-form solutions require less computational power than approximate methods such as numerical methods, and allow for the solution to be expressed in a compact form, which makes it easier to interpret and understand.
Encapsulating the ODE solution (forward operation) and its gradient (backward operation) under a closed-form transformation block shortens the chain of operations and decreases the overhead of managing a long tape with lots of scalar arithmetic operations. Indeed, explicitly implementing the backward operation is more efficient than letting the automatic differentiation (AD) system naively differentiate the closed-form forward function.

This chapter is structured as follows.
We discuss related work in \cref{sec:related_work_2} and outline the mathematical formalism relevant to our method in \cref{sec:method_2}.
The experiments with extensive results are described in \cref{sec:results_2} and final remarks are included in \cref{sec:conclusions_2}.

\section{Related Work}\label{sec:related_work_2}

In this section we review available solutions to compute derivatives to the ODE's solution:

\textbf{Numerical differentiation} is a method of approximating the derivative of a mathematical function by using a finite difference to compute the slope of the function at a given point. A mathematical formula is used to calculate the difference between two nearby points on the function and then divide this difference by the distance between the two points. This allows the derivative of the function to be calculated without having to solve for the exact derivative using calculus. Numerical differentiation is often used in scientific and engineering applications when the exact derivative of a function is difficult or impossible to calculate, or when the function is given to us only as a set of discrete data points rather than as an analytical expression.

Numerical differentiation should be avoided because it requires two numerical ODE solutions for each parameter (very inefficient) and is prone to numerical error: if the step size is too small, it may exhibit floating point cancellation; if the step size is chosen too large, then the error term of the approximation is large. 

\textbf{Forward-mode continuous sensitivity analysis} extends the ODE system and studies the model response when each parameter is varied while holding the rest at constant values. Thus, it involves taking the derivative of the model with respect to each of the input variables. This type of sensitivity analysis is useful for understanding how a system will respond to changes in its inputs, and for identifying which input variables are the most important for determining the output of the model. However, since the number of ODEs in the system scales proportionally with the number of parameters, this method is impractical for systems with a large number of parameters.

\begin{table}[!htb]
    \caption{Available solutions to compute derivatives to the ODE's solution}
    \label{tab:sota_derivatives}
    \vspace{-0.5cm}
    \begin{center}
      \resizebox{0.9\linewidth}{!}{%
      \begin{tabular}{@{}lllll@{}}
        \toprule
        \textbf{Method}                                                                      & \textbf{Forward$\rightarrow$}                                        & \textbf{Backward $\leftarrow$}                                 & \textbf{Advantages \textcolor{ForestGreen}{\cmark}}                                                                                        & \textbf{Disadvantages \textcolor{red}{\xmark}}                                                                             \\ \midrule \\[-0.9em]
        \begin{tabular}[c]{@{}l@{}}Numerical\\ differentiation\end{tabular}            & \begin{tabular}[c]{@{}l@{}}Generic\\ solver\end{tabular}             & \begin{tabular}[c]{@{}l@{}}Finite\\ differences\end{tabular}   & Easy to implement                                                                                    & \begin{tabular}[c]{@{}l@{}}Sensitive to round-off\\ and truncation errors\end{tabular}          \\ \\[-0.8em] \rowcolor[rgb]{0.92,0.92,0.92} & & & & \\[-0.8em]
        \rowcolor[rgb]{0.92,0.92,0.92} 
        \begin{tabular}[c]{@{}l@{}}Forward-mode\\ sensitivity \\ analysis\end{tabular} & \begin{tabular}[c]{@{}l@{}}Generic\\ solver\end{tabular}             & \begin{tabular}[c]{@{}l@{}}Extended\\ ODE system\end{tabular}  & \begin{tabular}[c]{@{}l@{}}Requires a single \\ numerical ODE \\ solver call\end{tabular}            & \begin{tabular}[c]{@{}l@{}}Impractical for large\\ number of parameters\end{tabular}            \\[1.5em] \\[-0.8em]
        \begin{tabular}[c]{@{}l@{}}Residual\\ networks\end{tabular}                    & \begin{tabular}[c]{@{}l@{}}Layers of\\ residual\\ units\end{tabular} & Backpropagation                                                & \begin{tabular}[c]{@{}l@{}}Non-stationary \\ velocity functions, \\ higher expressivity\end{tabular} & \begin{tabular}[c]{@{}l@{}}Complex neural\\ architecture, high \\ number of layers\end{tabular} \\ \\[-0.8em] \rowcolor[rgb]{0.92,0.92,0.92} & & & & \\[-0.8em]
        \rowcolor[rgb]{0.92,0.92,0.92} 
        \begin{tabular}[c]{@{}l@{}}Adjoint \\ sensitivity \\ analysis\end{tabular}     & \begin{tabular}[c]{@{}l@{}}Generic\\ solver\end{tabular}             & \begin{tabular}[c]{@{}l@{}}Adjoint\\ method\end{tabular}       & \begin{tabular}[c]{@{}l@{}}Tractable change\\ of variables \& \\ continuous time\end{tabular}        & \begin{tabular}[c]{@{}l@{}}Slow \& requires \\ extra hyperparameters\end{tabular}               \\[1.5em] \\[-0.8em]
        \begin{tabular}[c]{@{}l@{}}Discrete \\ sensitivity \\ analysis\end{tabular}    & \begin{tabular}[c]{@{}l@{}}Generic\\ solver\end{tabular}             & Backpropagation                                                & \begin{tabular}[c]{@{}l@{}}Low overhead \& \\ efficiency gains\end{tabular}                          & \begin{tabular}[c]{@{}l@{}}Requires specialized\\ ODE solver\\ implementation\end{tabular}      \\ \\[-0.8em] \rowcolor[rgb]{0.92,0.92,0.92} & & & & \\[-0.8em]
        \rowcolor[rgb]{0.92,0.92,0.92} 
        \begin{tabular}[c]{@{}l@{}}Closed-form\\ solution\end{tabular}                 & \begin{tabular}[c]{@{}l@{}}Closed-form\\ integration\end{tabular}    & \begin{tabular}[c]{@{}l@{}}Closed-form\\ gradient\end{tabular} & \begin{tabular}[c]{@{}l@{}}Efficient \& \\ exact computation\end{tabular}                            & \begin{tabular}[c]{@{}l@{}}Requires specialized\\ ad-hoc solution\end{tabular}                  \\ \bottomrule
        \end{tabular}%
    }
\end{center}
\vspace{-0.5cm}
\end{table}

\textbf{Residual networks}, also known as ResNets, are a type of deep learning neural network that is designed to facilitate the training of very deep neural networks. This is achieved by using a shortcut connection, or a skip connection, that allows the gradient to be directly backpropagated to earlier layers. This helps to alleviate the vanishing gradient problem, which is a common issue in deep neural networks where the gradient becomes increasingly small as it is backpropagated through the network, making it difficult for the network to learn. By using residual networks, it is possible to train much deeper networks without experiencing the same issues with vanishing gradients.

Residual networks can be interpreted as discrete numerical integrators of continuous dynamical systems, with each residual unit acting as one step of Euler's forward method. ResNet-TW \cite{Huang2021} uses this Eulerian discretization schema of the ODE to generate smooth and invertible transformations. The gradient is computed using reverse-mode automatic differentiation back pass, also known as backpropagation\footnote{Automatic differentiation have become the pervasive backbone behind all machine learning libraries. The adjoint technique, backpropagation, and reverse-mode automatic differentiation are in some sense all equivalent phrases given to this method from different disciplines}.  
However, ResNets are difficult to compress without significantly decreasing model accuracy, and do not learn to represent continuous dynamical systems in any meaningful sense \cite{queiruga2020continuousindepth}. Even though highly expressive diffeomorphic transformations can be generated via non-stationary velocity functions, the integration error is proportional to the number of residual units. Computing precise ODE solutions with ResNets require increasing the number of layers and as a result, the memory use of the model.

\textbf{Neural ODEs} \cite{chen2019neural} are a type of neural network that use ordinary differential equations (ODEs) to model the dynamics of a system. In this case, the ODE is used to define the way that the network's internal states evolve over time, allowing the network to model complex, non-linear relationships between input and output. 
Neural ODE's can be interpreted as the continuous version of ResNets, and compute gradients via \textbf{continuous adjoint sensitivity analysis}, i.e. solving a second augmented ODE backwards in time. In this case, the sensitivity of the system with respect to a parameter is calculated by solving a set of adjoint equations, which are derived from the original system of equations. One of the key advantages of continuous adjoint sensitivity analysis is that it allows for the efficient calculation of sensitivities for systems with a large number of parameters. This can be particularly useful in complex optimization problems, where the number of parameters may be very large and traditional sensitivity analysis methods may be computationally impractical. 

The efficiency issue with adjoint sensitivity analysis methods is that they require multiple forward ODE solutions, which can become prohibitively costly in large models. Neural ODEs reduce the computational complexity to a single solve, while retaining low memory cost by solving the adjoint gradients jointly backward-in-time alongside the ODE solution. 
However, this method implicitly makes the assumption that the ODE integrator is time-reversible and sadly there are no reversible adaptive integrators for first-order ODEs solvers, so this method diverges on some systems \cite{rackauckas2019diffeqfluxjl}. Other notable approaches for solving the adjoint equations with different memory-compute trade-offs are
interpolation schemes \cite{daulbaev2020interpolation},
symplectic integration \cite{zhuang2021mali},
storing intermediate quantities \cite{zhang2014fatode} and
checkpointing \cite{zhuang2020adaptive}. 

\textbf{Discrete sensitivity analysis} calculates model sensitivities by directly differentiating the numerical method's steps. Yet, this approach requires specialized implementation of the first-order ODE solvers to propagate said derivatives. Automatic differentiation (AD) can be used on a solver implemented fully in a language with AD (a differential programming approach) \cite{ma2021comparison}. However, pure tape-based reverse-mode AD software libraries (such as PyTorch \cite{paszke2019pytorch}, \textit{ReverseDiff.jl} \cite{reverseDiff} and \textit{Tensorflow Eager} \cite{agrawal2019tensorflow}), 
have been generally optimized for large linear algebra and array operations, which decrease the size of the tape in relation to the amount of work performed. Given that ODEs are typically defined by nonlinear functions with scalar operations, the tape handling to work ratio gets cut down and is no longer competitive with other forms of derivative calculations \cite{ma2021comparison}.
Other frameworks, such as \textit{JAX} \cite{jax2018github} cannot JIT optimize the non-static computation graphs of a full ODE solver. These issues may be addressed in the next generation reverse-mode source-to-source AD packages like \textit{Zygote.jl} \cite{Zygote.jl-2018} or \textit{Enzyme.jl} \cite{NEURIPS2020_9332c513} by not relying on tape generation.

\section{Closed-Form Diffeomorphic Transformations}\label{sec:method_2}

To overcome the mentioned limitations of current elastic similarity metrics in \cref{chapter:2}, this thesis proposes novel parametric alignment methods that are well-suited for deep learning methods. In particular, we propose a diffeomorphic warping function that is based on fast and exact integration of continuous piecewise-affine (CPA) velocity fields.
For this section, we inherit the notation used by \cite{Freifeld2015,Freifeld2017}, who originally proposed CPA-based diffeomorphic transformations.

\subsection{Continuous Piecewise-Affine (CPA) Velocity Functions}\label{sec:cpa_velocity_functions}

A piecewise function is a mathematical function that is defined by multiple sub-functions, each of which applies to a different range of the function's input values. In other words, a piecewise function is a combination of several functions that have been pieced together to form a single, cohesive function. This type of function is typically used in a variety of applications, including computer graphics, signal processing, and numerical analysis, as it allows for the representation of complex or discontinuous phenomena in a more manageable form.  In order to determine the value of a piecewise function at a given point, one must first determine which sub-function the point belongs to. This can be done by evaluating the conditions associated with each sub-function and identifying which interval the point lies within. Once the appropriate sub-function has been identified, the value of the function at that point can be determined by evaluating the sub-function at the given point. For example, consider the function:
\begin{equation}
f(x) = 
\begin{cases}
    x      & \text{if $x \leq 1$} \\
    2x-1   & \text{if $1 \le x \le 2$} \\
    5-x    & \text{if $x \geq 2$} \\
\end{cases}
\end{equation}

This piecewise function is defined on the entire real line $\mathbb{R}$ (interval $[-\infty, \infty]$), and is composed of three sub-functions: $f_{1}(x) = x$, $f_{2}(x) = 2x-1$, and $f_{3}(x) = 5-x$. These sub-functions are connected at the breakpoints $x = 1$ and $x = 2$, resulting in a continuous piecewise function that transitions from one sub-function to the next.

Continuous piecewise functions have several key properties that make them useful in mathematical modeling. First, they are continuous, meaning that they have no breaks or gaps in their graphs. This allows for the representation of smoothly varying phenomena, such as the changing temperature of a liquid over time. Second, being piecewise allows for the representation of phenomena that exhibit distinct behavior in different regions, such as the movement of a particle in a potential well.

The sub-functions can be of any type and complexity, but linear affine functions are usually chosen as they are relatively easy to work with mathematically. An affine function maps a linear combination of input variables to a scalar output and have the form $f(x) = ax + b$, where $a$ and $b$ are constants. 
When a piecewise function is defined by affine functions in each interval, it is called a continuous piecewise affine function. This means that the function is linear within each interval, and the function's graph is composed of multiple straight line segments, each corresponding to one of the intervals.

We now formally introduce the necessary constructs to mathematically define a continuous piecewise function.
Let $\Omega \subseteq \mathbb{R}$ be the Cartesian product of 1D compact intervals that represents the temporal domain. 
\begin{definition}[Tessellation]\label{def:tessellation}
A finite tessellation $\mathcal{P} = \{U_{c}\}_{c=1}^{N_\mathcal{P}}$ is a set of $N_\mathcal{P}$ closed subsets of $\Omega$, also called cells $U_c$, such that their union is $\Omega$ and the intersection of any pair of adjacent cells is their shared border. 
Notation: $N_\mathcal{P}$ is the number of cells, $N_v = N_p + 1$ number of vertices and $N_e = N_v - 2$ number of shared vertices (all three quantities positive integers). 
\end{definition}

\begin{definition}[Membership function]\label{def:membership}
Fix a tessellation $\mathcal{P}$, let $x \in \Omega$ and define the membership function $\gamma: \Omega \rightarrow \{1,...,N_\mathcal{P}\}$, $\gamma: x \rightarrow \text{min}\{c: x \in U_c\}$. If $x$ is not on an inter-cell border, then $\gamma(x) = c \leftrightarrow x \in U_c$.
\end{definition}

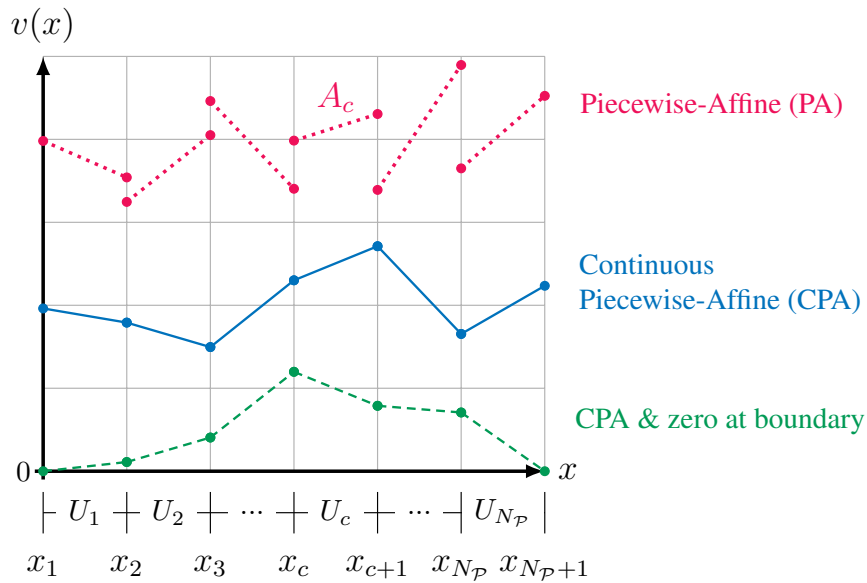
\begin{figure}[!htb]
    \begin{center}
    \scalebox{1.1}{
    \begin{tikzpicture}
        \pgfmathsetmacro{\N}{6};
        \pgfmathsetmacro{\M}{5};
        \pgfmathsetmacro{\P}{\N-1};
        \pgfmathsetmacro{\h}{-0.5};
        \draw[stepx=1, stepy=1, black!30, thin] (0,0) grid (\N, \M);
        \draw[-latex, very thick] (-0.1,0) -- (\N,0) node[right] {$x$};
        \draw[-latex, very thick] (0,-0.1) -- (0,\M) node[above] {$v(x)$};
        \draw (0,0.25+\h) -- (0,-0.25+\h)  node[anchor=north, pos=1.25] {$x_{1}$};
        \draw (1,0.25+\h) -- (1,-0.25+\h)  node[anchor=north, pos=1.25] {$x_{2}$};
        \draw (2,0.25+\h) -- (2,-0.25+\h)  node[anchor=north, pos=1.25] {$x_{3}$};
        \draw (3,0.25+\h) -- (3,-0.25+\h)  node[anchor=north, pos=1.25] {$x_{c}$};
        \draw (4,0.25+\h) -- (4,-0.25+\h)  node[anchor=north, pos=1.25] {$x_{c+1}$};
        \draw (5,0.25+\h) -- (5,-0.25+\h)  node[anchor=north, pos=1.25] {$x_{N_\mathcal{P}}$};
        \draw (6,0.25+\h) -- (6,-0.25+\h)  node[anchor=north, pos=1.25] {$x_{N_\mathcal{P}+1}$};
    
        \draw[-, thin] (0,\h) -- (1,\h) node[midway,fill=white, font=\footnotesize] {$U_{1}$};
        \draw[-, thin] (1,\h) -- (2,\h) node[midway,fill=white, font=\footnotesize] {$U_{2}$};
        \draw[-, thin] (2,\h) -- (3,\h) node[midway,fill=white, font=\footnotesize] {$...$};
        \draw[-, thin] (3,\h) -- (4,\h) node[midway,fill=white, font=\footnotesize] {$U_{c}$};
        \draw[-, thin] (4,\h) -- (5,\h) node[midway,fill=white, font=\footnotesize] {$...$};
        \draw[-, thin] (5,\h) -- (6,\h) node[midway,fill=white, font=\footnotesize] {$U_{N_\mathcal{P}}$};
    
        \node[black, font=\footnotesize, left] at (0,0) {$0$};
        \node[OrangeRed, font=\normalsize] at (3.5,4.5) {$A_{c}$};
        \node[OrangeRed,   align=left, font=\footnotesize] at (8.0,4.4) {Piecewise-Affine (PA)};
        \node[RoyalBlue,  align=left, font=\footnotesize] at (8.1,2.25) {Continuous\\Piecewise-Affine (CPA)};
        \node[ForestGreen, align=left, font=\footnotesize] at (8.1,0.6) {CPA \& zero at boundary};
    
        \pgfmathsetseed{2}
        \foreach \i in {0,...,\P}{
            \pgfmathparse{3.0 + \M * random() / 2.0};
            \coordinate (A) at (\i,\pgfmathresult);
            \pgfmathparse{3.0 + \M * random() / 2.0};
            \coordinate (B) at (\i+1,\pgfmathresult);
            \filldraw[OrangeRed] (A) circle (1.5pt);
            \draw[dotted, OrangeRed, very thick] (A) -- (B);
            \filldraw[OrangeRed] (B) circle (1.5pt);
        }
        \pgfmathsetseed{2}
        \pgfmathparse{\M * random()}\let\var=\pgfmathresult;
        \foreach \i in {0,...,\P}{
            \coordinate (A) at (\i,\var);
            \pgfmathparse{1.25 + \M * random() / 2.0};
            \coordinate (B) at (\i+1,\pgfmathresult);
            \filldraw[RoyalBlue] (A) circle (1.5pt);
            \draw[-, solid, RoyalBlue, thick] (A) -- (B);
            \filldraw[RoyalBlue] (B) circle (1.5pt);
            \global\let\var=\pgfmathresult
        }
        \pgfmathsetseed{3}
        \pgfmathsetmacro{\var}{0}
        \foreach \i in {0,...,\P}{
            \coordinate (A) at (\i,\var);
            \pgfmathparse{\M * random() / 4.0};
            \coordinate (B) at (\i+1,\pgfmathresult);
            \ifnum\i=\P
                \coordinate (B) at (\i+1,0);
            \fi
            \filldraw[ForestGreen] (A) circle (1.5pt);
            \draw[-, densely dashed, ForestGreen, thick] (A) -- (B);
            \filldraw[ForestGreen] (B) circle (1.5pt);
            \global\let\var=\pgfmathresult
        }
    \end{tikzpicture}
    }
    \caption{Velocity functions $v(x)$. Each cell $U_c$ in the tessellation $\mathcal{P}$ defines an affine transformation $\mathbf{A}_{c}=[a_c \;\; b_c] \in \mathbb{R}^{1 \times 2}$.}
    \label{fig:velocity_functions}
    \end{center}
\end{figure}

\begin{definition}[Piecewise affine]\label{def:piecewise_affine}
A map $f: \Omega \rightarrow \mathbb{R}$ is called piecewise affine (PA) w.r.t tessellation $\mathcal{P}$ if $\{f|_{U_c}\}_{c=1}^{N_\mathcal{P}}$ are affine, i.e., 

\begin{center}
    $f(x) = \mathbf{A}_{\gamma(x)} \tilde{x}\quad$ 
    where $\quad\tilde{x} \stackrel{\Delta}{=} \begin{bmatrix} x & 1 \end{bmatrix}^{T} \in \mathbb{R}^{2 \times 1} ,\quad \mathbf{A}_c \in \mathbb{R}^{1 \times 2} \quad \forall c \in \{1,...,N_\mathcal{P}\}$.
\end{center}
    
Let's define $\mathbf{A}_c = \begin{bmatrix} a_c & b_c \end{bmatrix} \in \mathbb{R}^{1 \times 2}$, then $f(x) = a_{\gamma(x)}x + b_{\gamma(x)} = a_c x + b_c$. 
\end{definition}

\begin{definition}[Continuous piecewise affine]\label{def:continuous_piecewise_affine}
$f$ is called CPA if it is continuous and piecewise affine.
\end{definition}

Let $\mathcal{V}_{\Omega, \mathcal{P}}$ be the spaces of CPA velocity fields on $\Omega$ w.r.t $\mathcal{P}$ and $d = \text{dim}(\mathcal{V}) = N_v$ its dimensionality.
A generic element of $\mathcal{V}_{\Omega, \mathcal{P}}$ is denoted by $v_\mathbf{A}$ where 

\begin{equation}
\mathbf{A}=(\mathbf{A}_1, ...\,, \mathbf{A}_{N_\mathcal{P}}) = \begin{bmatrix}
a_1 \,\cdots\, a_c\, \cdots \, a_{N_\mathcal{P}} \\
b_1 \,\cdots\, b_c\, \cdots \, b_{N_\mathcal{P}} 
\end{bmatrix}^T \in \mathbb{R}^{N_\mathcal{P} \times 2}
\end{equation}

The vectorize operation $vec(\mathbf{A})$ computes the row by row flattening to a column vector. If $\mathbf{A}_c = \begin{bmatrix} a_c & b_c \end{bmatrix} \in \mathbb{R}^{1 \times 2}$ then $vec(\mathbf{A}_c)=\mathbf{A}_{c}^{T} = \begin{bmatrix} a_c & b_c \end{bmatrix}^T \in \mathbb{R}^{2 \times 1}$. Also, if $\mathbf{A} \in \mathbb{R}^{{N_\mathcal{P}} \times 2}$ then $vec(\mathbf{A}) \in \mathbb{R}^{2{N_\mathcal{P}} \times 1}$, as follows:
\begin{equation*}\label{eq:vectorize}
\begin{split}
vec(\mathbf{A}) &= 
\begin{bmatrix} vec(\mathbf{A}_1)^T \, \cdots \, vec(\mathbf{A}_c)^T  \, \cdots & vec(\mathbf{A}_{N_\mathcal{P}})^T \end{bmatrix}^T \\ &= 
\begin{bmatrix} \mathbf{A}_1 \, \cdots \, \mathbf{A}_c \, \cdots \, \mathbf{A}_{N_\mathcal{P}} \end{bmatrix}^T \\ &=
\begin{bmatrix} a_1 \;\; b_1 \, \cdots \, a_c \;\; b_c \, \cdots \, a_{N_\mathcal{P}} \;\; b_{N_\mathcal{P}}
\end{bmatrix}^T
\end{split}
\end{equation*}

\subsection{Velocity Continuity Constraints}\label{sec:velocity_continuity_constraints}

Let's consider three adjacent cells $U_i, U_j, U_k$ with affine transformations

\begin{center}
    $\mathbf{A}_i = \begin{bmatrix} a_i & b_i \end{bmatrix}$,
    $\mathbf{A}_j = \begin{bmatrix} a_j & b_j \end{bmatrix}$ and
    $\mathbf{A}_k = \begin{bmatrix} a_k & b_k \end{bmatrix}$. 
\end{center}

The velocity field $v(x)$ must be continuous. The velocity field is denoted as $v_\mathbf{A}(x)$, since it depends on the affine transformation $\mathbf{A}$. $v_\mathbf{A}$ is continuous on every cell, because it is a linear function, but it is discontinuous on cell boundaries.
Continuity of $v_\mathbf{A}$ at $x_j$ implies one linear constraint on $\mathbf{A}_i$ and $\mathbf{A}_j$. In the same way, continuity of $v_\mathbf{A}$ at $x_k$ implies another linear constraint on $\mathbf{A}_j$ and $\mathbf{A}_k$.

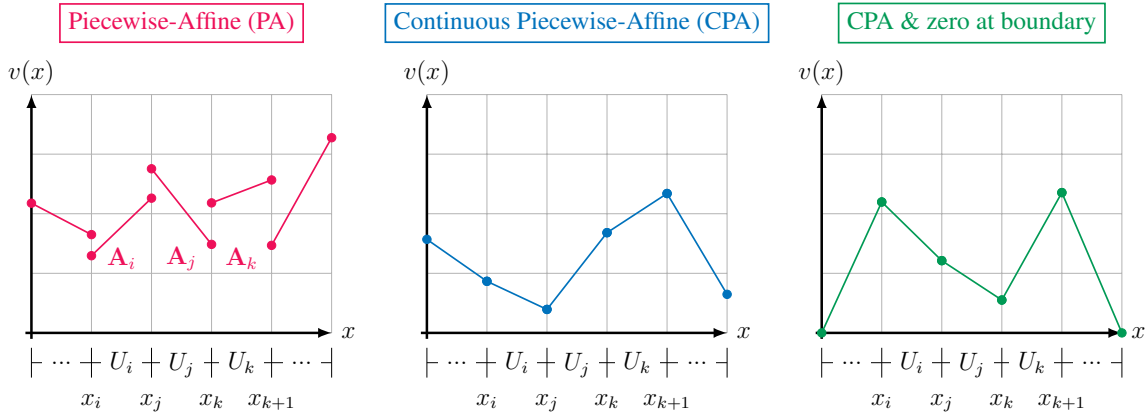
\begin{figure}[!ht]
    \begin{center}
    \scalebox{0.79}{
\begin{tikzpicture}[scale=1]
    \pgfmathsetmacro{\N}{5};
    \pgfmathsetmacro{\M}{4};
    \pgfmathsetmacro{\P}{\N-1};
    \pgfmathsetmacro{\h}{-0.5};
    \draw[step=1, black!30, thin] (0,0) grid (\N, \M);
    \draw[-latex, very thick] (-0.1,0) -- (\N,0) node[right] {$x$};
    \draw[-latex, very thick] (0,-0.1) -- (0,\M) node[above] {$v(x)$};

    \draw (0,0.25+\h) -- (0,-0.25+\h)  ;
    \draw (1,0.25+\h) -- (1,-0.25+\h)  node[anchor=north,  pos=1.25] {$x_{i}$};
    \draw (2,0.25+\h) -- (2,-0.25+\h)  node[anchor=north,  pos=1.25] {$x_{j}$};
    \draw (3,0.25+\h) -- (3,-0.25+\h)  node[anchor=north,  pos=1.25] {$x_{k}$};
    \draw (4,0.25+\h) -- (4,-0.25+\h)  node[anchor=north,  pos=1.25] {$x_{k+1}$};
    \draw (5,0.25+\h) -- (5,-0.25+\h);

    \draw[-, thin] (0,\h) -- (1,\h) node[midway,fill=white] {$...$};
    \draw[-, thin] (1,\h) -- (2,\h) node[midway,fill=white] {$U_{i}$};
    \draw[-, thin] (2,\h) -- (3,\h) node[midway,fill=white,text height=0.36cm] {$U_{j}$};
    \draw[-, thin] (3,\h) -- (4,\h) node[midway,fill=white] {$U_{k}$};
    \draw[-, thin] (4,\h) -- (5,\h) node[midway,fill=white] {$...$};

    \node[thin, OrangeRed] at (1.5,1.25) {$\mathbf{A}_{i}$};
    \node[thin, OrangeRed] at (2.5,1.25) {$\mathbf{A}_{j}$};
    \node[thin, OrangeRed] at (3.5,1.25) {$\mathbf{A}_{k}$};

    \pgfmathsetseed{2}
    \foreach \i in {0,...,4}{
        \pgfmathparse{3 * random()};
        \coordinate (A) at (\i,1+\pgfmathresult);
        \pgfmathparse{3 * random()};
        \coordinate (B) at (\i+1,1+\pgfmathresult);
        \filldraw[OrangeRed] (A) circle (2pt);
        \draw[OrangeRed, thick] (A) -- (B);
        \filldraw[OrangeRed] (B) circle (2pt);
    }

    \node[OrangeRed, draw=OrangeRed, align=center, font=\normalsize, rectangle, thick] at (2.5,5.2) {Piecewise-Affine (PA)};

\end{tikzpicture}
    }
    \scalebox{0.79}{
\begin{tikzpicture}[scale=1]
    \pgfmathsetmacro{\N}{5};
    \pgfmathsetmacro{\M}{4};
    \pgfmathsetmacro{\P}{\N-1};
    \pgfmathsetmacro{\h}{-0.5};
    \draw[step=1, black!30, thin] (0,0) grid (\N, \M);
    \draw[-latex, very thick] (-0.1,0) -- (\N,0) node[right] {$x$};
    \draw[-latex, very thick] (0,-0.1) -- (0,\M) node[above] {$v(x)$};

    \draw (0,0.25+\h) -- (0,-0.25+\h)  ;
    \draw (1,0.25+\h) -- (1,-0.25+\h)  node[anchor=north,  pos=1.25] {$x_{i}$};
    \draw (2,0.25+\h) -- (2,-0.25+\h)  node[anchor=north,  pos=1.25] {$x_{j}$};
    \draw (3,0.25+\h) -- (3,-0.25+\h)  node[anchor=north,  pos=1.25] {$x_{k}$};
    \draw (4,0.25+\h) -- (4,-0.25+\h)  node[anchor=north,  pos=1.25] {$x_{k+1}$};
    \draw (5,0.25+\h) -- (5,-0.25+\h);

    \draw[-, thin] (0,\h) -- (1,\h) node[midway,fill=white] {$...$};
    \draw[-, thin] (1,\h) -- (2,\h) node[midway,fill=white] {$U_{i}$};
    \draw[-, thin] (2,\h) -- (3,\h) node[midway,fill=white,text height=0.36cm] {$U_{j}$};
    \draw[-, thin] (3,\h) -- (4,\h) node[midway,fill=white] {$U_{k}$};
    \draw[-, thin] (4,\h) -- (5,\h) node[midway,fill=white] {$...$};

    \pgfmathsetseed{2}
    \pgfmathparse{\M * random()}\let\var=\pgfmathresult;
    \foreach \i in {0,...,\P}{
        \coordinate (A) at (\i,\var);
        \pgfmathparse{\M * random()};
        \coordinate (B) at (\i+1,\pgfmathresult);
        \filldraw[RoyalBlue] (A) circle (2pt);
        \draw[-, solid, RoyalBlue, thick] (A) -- (B);
        \filldraw[RoyalBlue] (B) circle (2pt);
        \global\let\var=\pgfmathresult
    }
    
    \node[RoyalBlue, draw=RoyalBlue, align=center, font=\normalsize, rectangle, thick] at (2.5,5.2) {Continuous Piecewise-Affine (CPA)};

\end{tikzpicture}
    }
    \scalebox{0.79}{
\begin{tikzpicture}[scale=1]
    \pgfmathsetmacro{\N}{5};
    \pgfmathsetmacro{\M}{4};
    \pgfmathsetmacro{\P}{\N-1};
    \pgfmathsetmacro{\h}{-0.5};
    \draw[step=1, black!30, thin] (0,0) grid (\N, \M);
    \draw[-latex, very thick] (-0.1,0) -- (\N,0) node[right] {$x$};
    \draw[-latex, very thick] (0,-0.1) -- (0,\M) node[above] {$v(x)$};

    \draw (0,0.25+\h) -- (0,-0.25+\h)  ;
    \draw (1,0.25+\h) -- (1,-0.25+\h)  node[anchor=north,  pos=1.25] {$x_{i}$};
    \draw (2,0.25+\h) -- (2,-0.25+\h)  node[anchor=north,  pos=1.25] {$x_{j}$};
    \draw (3,0.25+\h) -- (3,-0.25+\h)  node[anchor=north,  pos=1.25] {$x_{k}$};
    \draw (4,0.25+\h) -- (4,-0.25+\h)  node[anchor=north,  pos=1.25] {$x_{k+1}$};
    \draw (5,0.25+\h) -- (5,-0.25+\h);

    \draw[-, thin] (0,\h) -- (1,\h) node[midway,fill=white] {$...$};
    \draw[-, thin] (1,\h) -- (2,\h) node[midway,fill=white] {$U_{i}$};
    \draw[-, thin] (2,\h) -- (3,\h) node[midway,fill=white,text height=0.36cm] {$U_{j}$};
    \draw[-, thin] (3,\h) -- (4,\h) node[midway,fill=white] {$U_{k}$};
    \draw[-, thin] (4,\h) -- (5,\h) node[midway,fill=white] {$...$};

    \pgfmathsetseed{2}
    \pgfmathsetmacro{\var}{0}
    \foreach \i in {0,...,\P}{
        \coordinate (A) at (\i,\var);
        \pgfmathparse{1.4*\M * random()};
        \coordinate (B) at (\i+1,\pgfmathresult);
        \ifnum\i=\P
            \coordinate (B) at (\i+1,0);
        \fi
        \filldraw[ForestGreen] (A) circle (2pt);
        \draw[-, ForestGreen, thick] (A) -- (B);
        \filldraw[ForestGreen] (B) circle (2pt);
        \global\let\var=\pgfmathresult
    }
    \node[ForestGreen, draw=ForestGreen, align=center, font=\normalsize, rectangle, thick] at (2.5,5.2) {CPA \& zero at boundary};

\end{tikzpicture}
    }
    \caption{Piecewise-affine velocity function. Here three adjacent cells $U_i, U_j, U_k$ are represented. Continuity conditions at the boundary are necessary to comprise a continuous piecewise-affine velocity function}
    \label{fig:velocity_continuity_constraints}
    \end{center}
\end{figure}

Note that $\forall x \in \{U_i, U_j, U_k\}$, $v_\mathbf{A}$ is continuous. In order to be continuous on points $x_j$ and $x_k$, two linear constraints must be satisfied:

\begin{equation}
\left\{\begin{matrix}
    \mathbf{A}_i \cdot \tilde{x_j} = \mathbf{A}_j \cdot \tilde{x_j} \\
    \mathbf{A}_j \cdot \tilde{x_k} = \mathbf{A}_k \cdot \tilde{x_k}
\end{matrix}\right.
\Rightarrow 
\left\{\begin{matrix}
    a_i \cdot x_j + b_i = a_j \cdot x_j + b_j \\
    a_j \cdot x_k + b_j = a_k \cdot x_k + b_k
\end{matrix}\right.
\Rightarrow 
\left\{\begin{matrix}
    a_i \cdot x_j + b_i - a_j \cdot x_j - b_j = 0\\
    a_j \cdot x_k + b_j - a_k \cdot x_k - b_k = 0
\end{matrix}\right.
\end{equation}

To place the linear constraints in matrix form, let's recall the vectorize operation $vec$ for this case:
\begin{equation}
\begin{split}
vec(\mathbf{A}) &= 
\begin{bmatrix} vec(\mathbf{A}_i)^T & vec(\mathbf{A}_j)^T & vec(\mathbf{A}_k)^T  \end{bmatrix}^T \\ &= 
\begin{bmatrix} \mathbf{A}_i & \mathbf{A}_j & \mathbf{A}_k  \end{bmatrix}^T \\ &= 
\begin{bmatrix} a_i & b_i & a_j & b_j & a_k & b_k \end{bmatrix}^T 
\end{split}
\end{equation}

Therefore,
\begin{equation}
\begin{bmatrix} 
    x_j & 1 & -x_j & -1 & 0 & 0 \\
    0 & 0 & x_k & 1 & -x_k & -1
\end{bmatrix} 
\begin{bmatrix} a_i \\ b_i \\ a_j \\ b_j \\ a_k \\ b_k \end{bmatrix} =
\begin{bmatrix} 0 \\ 0 \end{bmatrix}
\Longrightarrow 
\boxed{\mathbf{L} \cdot vec(\mathbf{A}) = \vec{\mathbf{0}}}
\end{equation}

Extending the continuity constraints to a tessellation with $N_\mathcal{P}$ cells, the constraint matrix $\mathbf{L}$ has dimensions $N_{e} \times 2N_\mathcal{P}$, $vec(\mathbf{A})$ is $2N_\mathcal{P} \times 1$ and the null vector $\vec{0}$ is $N_{e} \times 1$. The number of shared vertices is $N_{e} = N_{v}-2 = N_\mathcal{P}-1$.

Any matrix $\mathbf{A}$ that satisfies $\mathbf{L} \cdot vec(\mathbf{A}) = \vec{\mathbf{0}}$ will be continuous everywhere. 
The null space of $\mathbf{L}$ coincides with the CPA vector-field space.
In this work, we implement four different methods to obtain the null space of $\mathbf{L}$: SVD decomposition, QR decomposition, Reduced Row Echelon Form (RREF) and Sparse Form. We refer the reader to \cref{sec:null_space} for a comparison of these four spaces about sparsity and computation times.

Let $\mathbf{B} = \begin{bmatrix} \mathbf{B}_1 & \mathbf{B}_2 & \cdots & \mathbf{B}_d \end{bmatrix} \in \mathbb{R}^{2N_\mathcal{P} \times N_v}$ be the orthonormal basis of the null space of $\mathbf{L}$.
Under this setting, $\boldsymbol{\theta} = \begin{bmatrix} \theta_1 & \theta_2 & \cdots & \theta_d \end{bmatrix} \in \mathbb{R}^{d} = \mathbb{R}^{N_v}$ are the coefficients (parameters) of each basis vector, and we can compute the matrix $\mathbf{A}$ as follows:
\begin{equation}\label{eq:linear_matrix}
vec(\mathbf{A}) = \mathbf{B} \cdot \boldsymbol{\theta} = \theta_1 \cdot \mathbf{B}_1 + \theta_2 \cdot \mathbf{B}_2 + \cdots + \theta_d \cdot \mathbf{B}_d = \sum_{j=1}^{d} \theta_j \cdot \mathbf{B}_j    
\end{equation}
If the velocity field is built using the orthonormal basis $\mathbf{B}$ such that $vec(\mathbf{A}) = \mathbf{B} \cdot \boldsymbol{\theta}$, then $v_{\mathbf{A}}$ is CPA.

\paragraph{Additional Constraints}
To allocate additional constraints, we must extend the constraint matrix $\mathbf{L}$ to have more rows. The null space of the extended $L$ is a linear subspace of the null space of the original $\mathbf{L}$. For instance, constraining the velocity field to be zero at the border of $\Omega$ : $v(\delta \Omega) = 0$ (zero-boundary constraint) adds two additional equations (one at each limit of $\Omega$), thus the number of constraints $d'=d-2$ and basis $\mathbf{B} = \begin{bmatrix} \mathbf{B}_1 & \mathbf{B}_2 & \cdots & \mathbf{B}_d' \end{bmatrix} \in \mathbb{R}^{2N_\mathcal{P} \times d'}$

\paragraph{Smoothness Priors}
We include smoothness priors on CPA velocity functions as done by \cite{Freifeld2017} to handle scarce-data regions: First sampling a zero-mean Gaussian with $D \times D$ covariance matrix $\boldsymbol{\Sigma}_{PA}$ whose correlations decay with inter-cell distances $vec(\textbf{A}) \sim \mathcal{N}(0_{D \times 1}, \boldsymbol{\Sigma}_{PA})$; and then projecting it into the CPA space: $\boldsymbol{\theta} = \textbf{B}^T \cdot vec(\textbf{A})$. With this procedure we can sample transformation parameters $\boldsymbol{\theta}$ from a prior distribution:
$p(\boldsymbol{\theta}) = \mathcal{N}(0_{d \times 1}, \boldsymbol{\Sigma}_{CPA})$, where $\boldsymbol{\Sigma}_{CPA} = \textbf{B}^T \cdot \boldsymbol{\Sigma}_{PA} \cdot \textbf{B}$ uses the squared exponential kernel and has two parameters: $\lambda_{\sigma}$ which controls the overall variance and $\lambda_{s}$ which controls the kernel's length-scale. Small $\lambda_{\sigma}$ generate close to the identity warps and vice versa. Large $\lambda_{s}$ favors purely affine velocity fields.

\subsection{Null Space of the Constraint Matrix $\mathbf{L}$}\label{sec:null_space}

In order to properly define the null space of a matrix, let's first review some auxiliary definitions of linear algebra, as introduced by \cite{poole2014linear}.

\subsubsection{Linear Algebra Recap}
Given an $m \times n$ matrix $\mathbf{L}$, let us define: 

    \begin{definition_plain}[Subspace]
    A subspace of $\mathbb{R}^n$ is any collection $S$ of vectors in $\mathbb{R}^n$ such that:
    \begin{itemize}
        \item The zero vector $\vec{\mathbf{0}}$ is in $S$.
        \item If $\mathbf{u}$ and $\mathbf{v}$ are in $S$, then $\mathbf{u} + \mathbf{v}$ is in $S$. ($S$ is closed under addition)
        \item If $\mathbf{u}$ is in $S$ and $c$ is a scalar, then $c\cdot\mathbf{u}$ is in $S$. ($S$ is closed under scalar multiplication)
    \end{itemize}
    \end{definition_plain}

    \begin{definition_plain}[Basis]
    A basis for a subspace $S$ of $\mathbb{R}^n$ is a set of vectors in $S$ that spans $S$ and is linearly independent.
    \end{definition_plain}

    \begin{definition_plain}[Dimension]
    If $S$ is a subspace of $\mathbb{R}^n$, then the number of vectors in a basis for $S$ is called the dimension of $S$, denoted as $\operatorname{dim}(S)$.
    \end{definition_plain}
    
    \begin{definition_plain}[Column space]
    The \textit{column space} of $\mathbf{L}$ the subspace of $\mathbb{R}^m$ spanned by the columns of $\mathbf{L}$. Its dimension is called the $\text{rank}(\mathbf{L})$. We denote this subspace $\operatorname{col}(\mathbf{L})$.
    \end{definition_plain}

    \begin{definition_plain}[Row space]
        The \textit{row space} of $\mathbf{L}$ the subspace of $\mathbb{R}^n$ spanned by its rows. We denote this subspace $\operatorname{row}(\mathbf{L})$.
    \end{definition_plain}

Therefore, based on these definitions, the null space can be formalized as:

\begin{definition}[Null space]
    The \textit{null space}, (also called the \textbf{kernel}) of $\mathbf{L}$ is the set of all solutions to the homogeneous equation represented by $\mathbf{L}$: 
    \begin{equation}
        \ker(\mathbf{L}) = {\, \mathbf{x} \in \mathbb{R}^n \mid \mathbf{L} \mathbf{x} = \mathbf{0} \,}
    \end{equation}
    In other words, the null space is a subspace of the vector space spanned by the columns of the matrix, that is, a subspace of $\mathbb{R}^n$. Its dimension is called the $\operatorname{nullity}(\mathbf{L})$. The null space of a matrix can be found using a number of different techniques, including row reduction, Gaussian elimination, and the computation of the matrix's nullity. It is important to note that the null space of a matrix is not unique, as it depends on the particular representation of the matrix.
\end{definition}

\clearpage
Four different methods have been implemented to obtain the null space of $\mathbf{L}$. Each method has a different computational cost and yields a base with different mathematical properties that can suite specific applications as well. 

\begin{itemize}
    \item Singular Value Decomposition (SVD): \cref{sec:svd}.
    \item QR decomposition: \cref{sec:qr}
    \item Reduced Row Echelon Form (RREF): \cref{sec:rref}
    \item Sparse form (SPARSE): \cref{sec:sparse}
\end{itemize}

Let's now analyze each of these methods for a simple example with 5 cells ($N_\mathcal{P}=5$), defined over the $[0,1]$ interval and with the additional constraint of zero velocity at the boundary. These 5 cells over the unit interval define the shared vertices $\{0.2, 0.4, 0.6, 0.8\}$, and the constraint matrix has 6 rows ($N_{e}=4$ + 2 additional constraints at the boundary) and 10 columns corresponding to 5 tuples $(a_{c}, b_{c})$:

\begin{equation}
    \mathbf{L}=
    \begin{blockarray}{rrrrrrrrrr}
        a_1 & b_1 & a_2 & b_2 & a_3 & b_3 & a_4 & b_4 & a_5 & b_5 \\
        \begin{block}{[rrrrrrrrrr]}
            \mathbf{0.2} & \mathbf{1} & \mathbf{-0.2} & \mathbf{-1} & 0 & 0 & 0 & 0 & 0 & 0 \\
            0  & 0 & \mathbf{0.4} & \mathbf{1} & \mathbf{-0.4} & \mathbf{-1} & 0 & 0 & 0 & 0 \\
            0  & 0 & 0  & 0 & \mathbf{0.6} & \mathbf{1} & \mathbf{-0.6} & \mathbf{-1} & 0 & 0 \\
            0  & 0 & 0  & 0 & 0 & 0 & \mathbf{0.8} & \mathbf{1} & \mathbf{-0.8} & \mathbf{-1}\\ 
            \mathbf{0}  & \mathbf{-1} & 0  & 0 & 0 & 0 & 0 & 0 & 0 & 0   \\
            0  & 0 & 0  & 0 & 0 & 0 & 0 & 0 & \mathbf{-1} & \mathbf{-1} \\
        \end{block}
    \end{blockarray}
\end{equation}

Note that the first four rows in $\mathbf{L}$ represent the continuity constraints at each shared vertex, while the last two rows are the additional constraints for zero velocity at the boundary ($b_1 = 0$ and $a_5 = b_5$). 

\subsubsection{SVD decomposition}\label{sec:svd}

It's worth noting that the null space of a matrix depends on the choice of basis, so the null space we find using SVD may not be the same as the null space we would find using a different method like QR. 

\begin{definition}[SVD decomposition]
    The singular value decomposition (SVD) is a matrix factorization technique that decomposes a matrix into its singular values and singular vectors. If $\mathbf{L}$ is an $m \times n$ matrix, then we may write $\mathbf{L}$ as a product of three factors: $\mathbf{L} = \mathbf{U} \boldsymbol{\Sigma} \mathbf{V}^T$, 
    where $\mathbf{U}$ is an orthogonal $m \times m$ matrix, $\mathbf{V}$ is an orthogonal $n \times n$ matrix, $\mathbf{V}^T$ is the transpose of $\mathbf{V}$, and $\boldsymbol{\Sigma}$ is an $m \times n$ matrix that has all zeros except for its diagonal entries, which are non-negative real numbers.  If $\sigma_{ij}$ is the $i,j$ entry of $\boldsymbol{\Sigma}$, then $\sigma_{ij} = 0$ unless $i=j$ and $\sigma_{ii} = \sigma_i \geq 0$.  The $\sigma_i$ are the \textit{singular values} and the columns of $u$ and $v$ are respectively the right and left singular vectors. For a matrix of size $m \times n$, the time complexity of the SVD is $\mathcal{O}(m^2n + n^3)$ for full SVD and $\mathcal{O}(mn^2)$ for the truncated SVD algorithm.
\end{definition}

To construct an orthonormal basis for the null space of $\mathbf{L}$ using SVD, we need to:
\begin{enumerate}
    \item Compute the SVD of the matrix $\mathbf{L}$: $\mathbf{L} = \mathbf{U} \boldsymbol{\Sigma} \mathbf{V}^T$
    \item Identify the singular values in $\boldsymbol{\Sigma}$ that are close to zero. These correspond to the dimensions of the null space of $\mathbf{L}$. By default, the tolerance is determined by the machine limit for floating point operations ($\varepsilon$) multiplied by $\max(m,n)$.
    \item For each singular value that is close to zero, the corresponding column in $\mathbf{V}$ is a basis vector for the null space of $\mathbf{L}$. 
\end{enumerate}

For the example with $N_\mathcal{P}=5$, this procedure using SVD yields the following null space:
\begin{equation}
    \mathbf{B}_{SVD}=
    \begin{blockarray}{rrrr}
        \mathbf{B}_1 & \mathbf{B}_2 & \mathbf{B}_3 & \mathbf{B}_4 \\
        \begin{block}{[rrrr]}
            0.121  & 0.135  & -0.549 & -0.589 \\
            0      & 0      & 0      & 0     \\
            0.189  & 0.658  & 0.206  & 0.412  \\
            -0.014 & -0.105 & -0.151 & -0.2   \\
            -0.655 & -0.227 & -0.203 & 0.316  \\
            0.324  & 0.249  & 0.013  & -0.162 \\
            0.491  & -0.495 & 0.008  & 0.22   \\
            -0.363 & 0.411  & -0.114 & -0.104 \\
            -0.146 & -0.071 & 0.538  & -0.359 \\
            0.146  & 0.071  & -0.538 & 0.359  \\
        \end{block}
    \end{blockarray}
\end{equation}

\subsubsection{QR decomposition}\label{sec:qr}

\begin{definition}[QR decomposition]
Let $\mathbf{L}$ be an $m \times n$ matrix with linearly independent columns. Then $\mathbf{L}$ can be factored as $\mathbf{L} = \mathbf{Q}\mathbf{R}$, where $\mathbf{Q}$ is an $m \times n$ matrix with orthonormal columns and $\mathbf{R}$ is an invertible upper triangular matrix with positive entries on its diagonal. For a real matrix of size $m \times m$, the time complexity of the QR is $\mathcal{O}(mn^2)$ using the Schwarz-Rutishauser algorithm.
\end{definition}

To obtain the null space of the constraint matrix $\mathbf{L}$ using QR decomposition, we need to follow these steps:
\begin{enumerate}
    \item Perform QR decomposition on the matrix $\mathbf{L}$. This involves finding an orthogonal matrix $\mathbf{Q}$ and an upper-triangular matrix $\mathbf{R}$ such that $\mathbf{L} = \mathbf{Q}\mathbf{R}$.
    \item Determine the rank of $\mathbf{L}$ by calculating the number of entries on the main diagonal of $\mathbf{R}$ whose magnitude exceeds a tolerance. Similarly to SVD, by default the tolerance is estimated using the machine limit for floating point operations ($\varepsilon$) multiplied by $\max(m,n)$.
    \item Once the rank $r$ is computed, the last $n-r$ columns of $\mathbf{Q}$ make up the null space of $\mathbf{L}$.
\end{enumerate}

For the example with $N_\mathcal{P}=5$, this procedure using QR yields the following null space:
\begin{equation}
    \mathbf{B}_{QR}=
    \begin{blockarray}{rrrr}
        \mathbf{B}_1 & \mathbf{B}_2 & \mathbf{B}_3 & \mathbf{B}_4 \\
        \begin{block}{[rrrr]}
            -0.259 & -0.018 & 0.416   & 0.664 \\
            0     & 0     & 0      & 0   \\
            -0.392 & -0.547 & -0.369  & -0.305\\
            0.027  & 0.106  & 0.157   & 0.194 \\
            -0.036 & 0.658  & -0.433  & -0.003\\
            -0.116 & -0.376 & 0.183   & 0.073 \\
            0.682  & -0.231 & -0.082  & 0.098 \\
            -0.547 & 0.157  & -0.028  & 0.012 \\
            0.005  & 0.138  & 0.468   & -0.455\\
            -0.005 & -0.138 & -0.468  & 0.455 \\
        \end{block}
    \end{blockarray}
\end{equation}

\subsubsection{Reduced Row Echelon Form (RREF)}\label{sec:rref}

We now describe the procedure by which any matrix can be reduced to a matrix in row echelon form. The allowable operations, called elementary row operations, correspond to the operations that can be performed on a system of linear equations to transform it into an equivalent system.
    The following elementary row operations can be performed on a matrix:
    \begin{itemize}
        \item Interchange two rows.
        \item Multiply a row by a nonzero constant.
        \item Add a multiple of a row to another row
    \end{itemize}

\begin{definition}[Gaussian Elimination]
When row reduction is applied to the augmented matrix of a system of linear equations, we create an equivalent system that can be solved by back substitution. The entire process is known as Gaussian elimination:
\begin{enumerate}
    \item Write the augmented matrix of the system of linear equations.
    \item Use elementary row operations to reduce the augmented matrix to row echelon form.
    \item Using back substitution, solve the equivalent system that corresponds to the row-reduced matrix.
\end{enumerate}
\end{definition}

\begin{definition}[Row Echelon Form]
    A matrix is in row echelon form if it satisfies the following properties:
    \begin{itemize}
        \item Any rows consisting entirely of zeros are at the bottom.
        \item In each nonzero row, the first nonzero entry (called the leading entry) is in a column to the left of any leading entries below it.
    \end{itemize}
\end{definition}

A modification of Gaussian elimination greatly simplifies the back substitution phase and is particularly helpful when calculations are being done by hand on a system with infinitely many solutions. This variant, known as Gauss-Jordan elimination, relies on reducing the augmented matrix even further.

\begin{definition}[Reduced Row Echelon Form]
A matrix is in reduced row echelon form if it satisfies the following properties:
\begin{itemize}
    \item It is in row echelon form.
    \item The leading entry in each nonzero row is a 1 (called a leading 1).
    \item Each column containing a leading 1 has zeros everywhere else.
\end{itemize}
\end{definition}

Using the Reduced Row Echelon Form (RREF) we can obtain the null space of the constraint matrix $\mathbf{L}$ with the additional constraints of zero velocity at the boundary: 
\begin{enumerate}
    \item Initialize the basis matrix $\mathbf{B}$ with $2 N_\mathcal{P}$ rows and $N_\mathcal{P}-1$ columns. 
    \item Let $x_0$ be the left boundary point in $\Omega$ and $s$ the cell size.
    \item Fill the $\mathbf{B}$ matrix with the following rules\footnote{The slice notation \textit{start : step : stop} is used to extract a portion of a sequence. The slice starts at the index specified by \textit{start}, and continues up to but not including the index specified by \textit{stop}. The \textit{step} argument specifies the number of indices to skip between successive elements in the slice.}, applied $\forall k \in [1, N_\mathcal{P}-1]$: 
    \begin{itemize}
        \renewcommand{\labelitemi}{\scriptsize\textcolor{Magenta}{$\blacksquare$}}
        \item $B[1,k]= x_0 + s $
        \renewcommand{\labelitemi}{\scriptsize\textcolor{Violet}{$\blacksquare$}}
        \item $B[2,k]= -x_0 * (x_0 + s) $
        \renewcommand{\labelitemi}{\scriptsize\textcolor{ForestGreen}{$\blacksquare$}}
        \item $B[3:2:2k+1, k]= s$\
        \renewcommand{\labelitemi}{\scriptsize\textcolor{Cyan}{$\blacksquare$}}
        \item $B[2k+1, k]= -(x_0 + k * s) $
        \renewcommand{\labelitemi}{\scriptsize\textcolor{Orange}{$\blacksquare$}}
        \item $B[2k+2, k]= (x_0 + k * s) * (x_0 + (k + 1) * s) $
    \end{itemize}
    \item An optional final step is to normalize each column in $\mathbf{B}$ by the column norm.
\end{enumerate}

For the example with $N_\mathcal{P}=5$, this procedure using RREF yields the following null space. The matrix has been colored to illustrate where each of the five rules apply:

\begin{equation}
    \mathbf{B}_{RREF}=
    \begin{blockarray}{rrrr}
        \mathbf{B}_1 & \mathbf{B}_2 & \mathbf{B}_3 & \mathbf{B}_4 \\
        \begin{block}{[rrrr]}
            \textcolor{Magenta}{\mathbf{0.2}} & \textcolor{Magenta}{\mathbf{0.2}} & \textcolor{Magenta}{\mathbf{0.2}} & \textcolor{Magenta}{\mathbf{0.2}} \\
            \textcolor{Violet}{\mathbf{0.}} & \textcolor{Violet}{\mathbf{0.}} & \textcolor{Violet}{\mathbf{0.}} & \textcolor{Violet}{\mathbf{0.}} \\
            \textcolor{Cyan}{\mathbf{-0.2}} & \textcolor{ForestGreen}{\mathbf{0.2}}  & \textcolor{ForestGreen}{\mathbf{0.2}}   & \textcolor{ForestGreen}{\mathbf{0.2}}  \\
            \textcolor{Orange}{\mathbf{0.08}} & 0  & 0  & 0   \\
            0   & \textcolor{Cyan}{\mathbf{-0.4}} & \textcolor{ForestGreen}{\mathbf{0.2}}   & \textcolor{ForestGreen}{\mathbf{0.2}}  \\
            0   & \textcolor{Orange}{\mathbf{0.24}} & 0 & 0   \\
            0   & 0   & \textcolor{Cyan}{\mathbf{-0.6}} & \textcolor{ForestGreen}{\mathbf{0.2}}  \\
            0   & 0   & \textcolor{Orange}{\mathbf{0.48}} & 0   \\
            0   & 0   & 0  & \textcolor{Cyan}{\mathbf{-0.8}} \\
            0   & 0   & 0  &\textcolor{Orange}{\mathbf{0.8}}  \\
        \end{block}
    \end{blockarray}
\end{equation}

\subsubsection{Sparse form (SPARSE)}\label{sec:sparse}

An alternative to the RREF method is to compute a Gaussian elimination process to yield a highly sparse null space. This process is ad-hoc for each case, and does not use any decomposition or factorization. The obtained base is not orthogonal, but presents the highest sparsity among the four proposed methods, as can be seen on \cref{fig:basis_spy}.

For the example with $N_\mathcal{P}=5$, this procedure using SPARSE yields the following null space:
\begin{equation}
    \mathbf{B}_{SPARSE}=
    \begin{blockarray}{rrrr}
        \mathbf{B}_1 & \mathbf{B}_2 & \mathbf{B}_3 & \mathbf{B}_4 \\
        \begin{block}{[rrrr]}
            \mathbf{1} & 0 & 0 & 0\\
            \mathbf{0} & 0 & 0 & 0\\
            \mathbf{-1} & \mathbf{1} & 0 & 0\\
            \mathbf{0.4} & \mathbf{-0.2} & 0 & 0\\
            0 & \mathbf{-1} & \mathbf{1} & 0\\
            0 & \mathbf{0.6} & \mathbf{-0.4} & 0\\
            0 & 0 & \mathbf{-1} & \mathbf{1}\\
            0 & 0 & \mathbf{0.8} & \mathbf{-0.6}\\
            0 & 0 & 0 & \mathbf{-1}\\
            0 & 0 & 0 & \mathbf{1}\\
        \end{block}
    \end{blockarray}
\end{equation}

\renewcommand\x{0.48}
\begin{figure}[!htb]
    \begin{center}
    \begin{subfigure}{\x\linewidth}
        \centering
        \includegraphics[width=\linewidth]{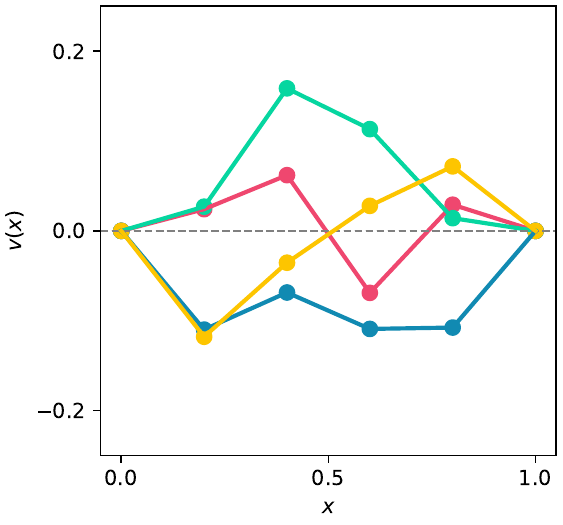}
        \caption{SVD velocity basis}
        \label{fig:basis_svd}
    \end{subfigure}
    \hfill
    \begin{subfigure}{\x\linewidth}
        \centering
        \includegraphics[width=\linewidth]{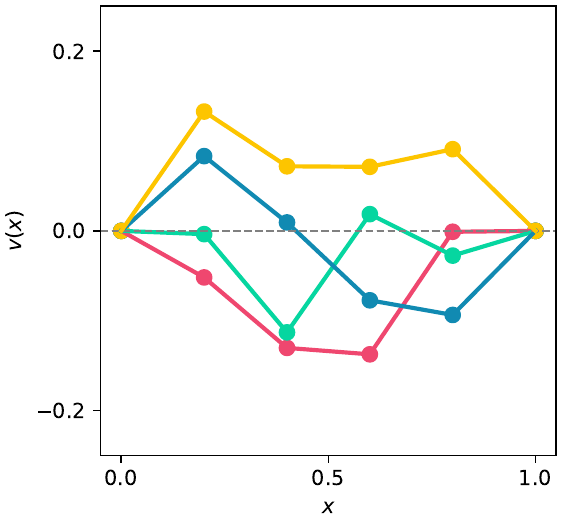}
        \caption{QR velocity basis}
        \label{fig:basis_qr}
    \end{subfigure}
    \\
    \begin{subfigure}{\x\linewidth}
        \centering
        \includegraphics[width=\linewidth]{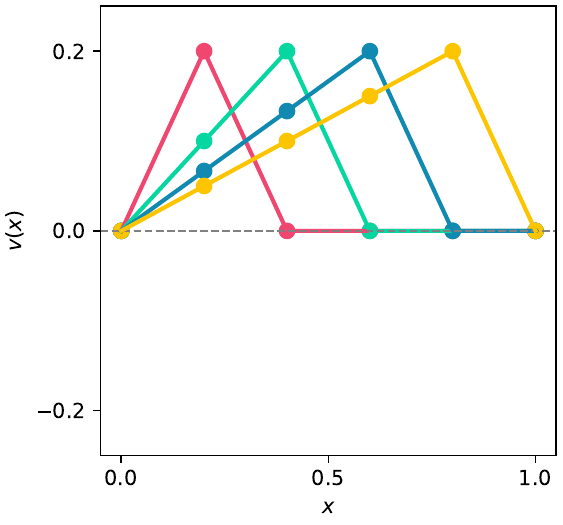}
        \caption{RREF velocity basis}
        \label{fig:basis_rref}
    \end{subfigure}
    \hfill
    \begin{subfigure}{\x\linewidth}
        \centering
        \includegraphics[width=\linewidth]{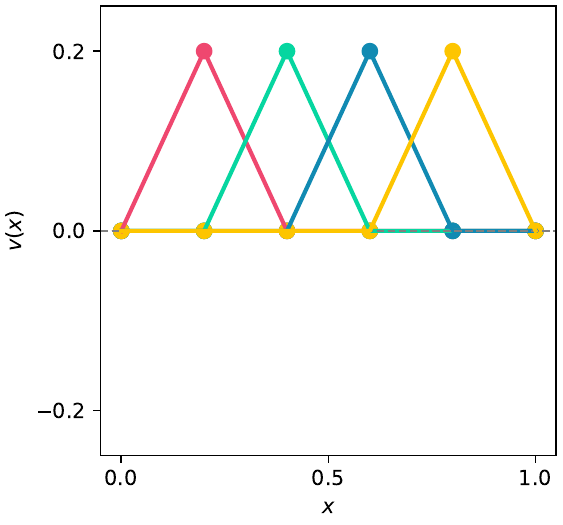}
        \caption{SPARSE velocity basis}
        \label{fig:basis_sparse}
    \end{subfigure}
    \caption{Visual comparison of 4 different velocity field basis (SVD, QR, RREF, SPARSE) with 5 cells ($N_\mathcal{P}=5$) and 4 degrees of freedom ($d=4$).}
    \label{fig:basis_velocity}
    \end{center}
\end{figure}

\clearpage
\subsubsection{Properties Comparison}

Here we analyze various properties for each of the proposed velocity field basis $\mathbf{B}$: 

\begin{itemize}
    \item The matrix \textbf{norm} is a function that assigns a non-negative value to each matrix, with certain properties. Here we use the Euclidean norm, also known as the Frobenius norm, and is defined as the square root of the sum of the squares of the elements of the matrix: $\Vert B \Vert =\sqrt{\sum_{i,j}^{m,n} B_{i,j}^{2}}$
    \item The \textbf{sparsity} counts the number of zero-valued elements divided by the total number of elements (e.g., $m \times n$ for a matrix of $m$ rows and $n$ columns), which is equal to 1 minus the density of the matrix. 
    \item The \textbf{condition number} is defined as the norm of the matrix $\mathbf{B}$ times the norm of the matrix inverse $\mathbf{B}^{-1}$, that is: $\Vert \mathbf{B} \mathbf{B}^{-1} \Vert$. It measures the stability or sensitivity of a matrix (or the linear system it represents) to numerical operations, that is, how sensitive the output is to perturbations in the input.
    \item The \textbf{orthogonality} $\Vert \mathbf{B}^{T} \mathbf{B} - I \Vert$ measures how close the matrix $\mathbf{B}$ is to an orthogonal matrix. A real square matrix is orthogonal if and only if its columns form an orthonormal basis of the Euclidean space. A matrix with orthogonal (not orthonormal) columns might be tempted to be referred to as an orthogonal matrix, but such matrices have no special significance and no special name; they only satisfy $\mathbf{B}^{T} \mathbf{B}=\mathbf{D}$, with D a diagonal matrix. An important property of orthonormal matrices is that their inverse is equal to their transpose $\mathbf{B}^{-1} = \mathbf{B}^{T}$.

\end{itemize}

\begin{figure}[!htb]
    \begin{center}
        \begin{subfigure}{0.8\linewidth}
            \centering
            \includegraphics[width=\linewidth]{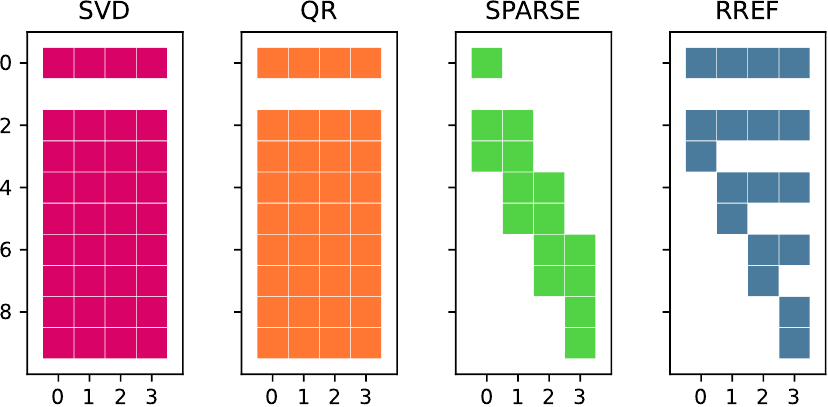}
            \caption{Basis with 5 cells ($N_\mathcal{P}=5$) and 4 degrees of freedom ($d=4$).}
            \label{fig:basis_spy_5}
        \end{subfigure}
        \vspace{0.5em}
        \\
        \begin{subfigure}{0.8\linewidth}
            \centering
            \includegraphics[width=\linewidth]{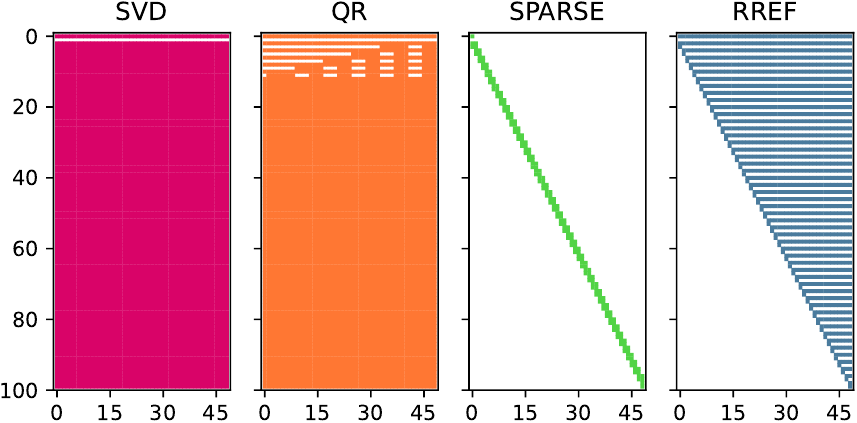}
            \caption{Basis with 50 cells ($N_\mathcal{P}=50$) and 49 degrees of freedom ($d=49$).}
            \label{fig:basis_spy_50}
        \end{subfigure}
        \vspace{-0.5em}
    \caption{Sparsity pattern of 4 different velocity field basis (SVD, QR, SPARSE, RREF) with 5 (top) and 50 (bottom) cells ($N_\mathcal{P}$) and 4 degrees of freedom ($d=4$). Columns indicate each of the basis vectors and colored cells indicate the non-zero values of the basis.}
    \label{fig:basis_spy}
    \end{center}
\end{figure}

For example, for a basis with 50 cells ($N_\mathcal{P}=5$) and 49 degrees of freedom ($d=49$), \cref{tab:basis} summarizes the properties presented above. Also related, \cref{fig:basis_spy} visualizes the sparsity pattern of each basis. Regarding SVD and QR, the norm is equal the norm of the inverse matrix, has a unit condition number and is orthogonal. However, these bases are dense matrices with almost zero sparsity. On the contrary, RREF is very sparse (0.73), non-orthogonal and with a condition number slightly greater than the unit (1.32). Finally, as was expected, the SPARSE basis is super sparse (0.96), also non-orthogonal but very ill-conditioned (34.4). A performance analysis of these different methods is included in \cref{sec:results:null_space}.

\begin{table}[!htb]
    \captionsetup{width=0.7\textwidth}
    \caption{Properties comparison of 4 different velocity field basis (SVD, QR, RREF, SPARSE) with 50 cells ($N_\mathcal{P}=5$) and 49 degrees of freedom ($d=49$).}
    \label{tab:basis}
    \vspace{-1em}   
    \begin{center}
    \begin{tabular}{lccccc}
        \toprule
        \textbf{Basis} $\mathbf{B}$ & $\Vert \mathbf{B} \Vert$ & $\Vert \mathbf{B}^{-1} \Vert$ & \begin{tabular}[t]{@{}c@{}}Condition\\ Number\end{tabular} &  Orthogonality & Sparsity  \\ \midrule
        \textbf{SVD}     & 7.00  & 7.00  & 1.00 & \textcolor{ForestGreen}{\cmark} (0.00) & 0.01\\
        \textbf{QR}      & 7.00  & 7.00  & 1.00 & \textcolor{ForestGreen}{\cmark} (0.00) & 0.04\\
        \textbf{RREF}    & 8.40  & 5.92  & 1.32 & \textcolor{red}{\xmark}         (3.63) & 0.73\\
        \textbf{SPARSE}  & 11.4  & 17.0  & 34.4 & \textcolor{red}{\xmark}         (18.1) & 0.96\\
        \bottomrule
    \end{tabular}
    \end{center}
\end{table}

\clearpage
\subsection{CPA Diffeomorphic Transformations}\label{sec:cpa_diffeomorphic_transformations}

\begin{definition}\label{def:diffeomorphism}
A map $T: \Omega \rightarrow \Omega$ is called a ($\mathcal{C}^1$) diffeomorphism on $\Omega$ if $T^{-1}$ exists and both  $T$ and $T^{-1}$ are differentiable. A diffeomorphism can be obtained, via integration, from uniformly continuous stationary velocity fields $T^{\theta}(x) = \phi^{\theta}(x,1)$ 
where $\phi^{\theta}(x,t) = x + \int_0^t v^{\theta}(\phi^{\theta}(x,\tau)) d\tau$ for uniformly continuous $v: \Omega \rightarrow \mathbb{R}$ and integration time $t$.
\end{definition}
Any continuous velocity field, whether piecewise-affine or not, defines differentiable $\mathbb{R} \rightarrow \Omega$ trajectories. 
Let $x \in \Omega$ and $v^\theta \in \mathcal{V}_{\Omega, \mathcal{P}}$ define a function: $t \rightarrow \phi^\theta(x,t)$ 
such that $\phi^\theta(x,0) = x$ and $\phi^\theta(x,t)$ solves the integral equation:
\begin{equation}\label{eq:integral}
\phi^\theta(x,t) = x + \int_0^t v^\theta(\phi^\theta(x,\tau)) d\tau
\end{equation}
or, the equivalent ordinary differential equation (ODE):
\begin{equation}\label{eq:ode}
\frac{\partial\phi^\theta(x,t)}{\partial t} = v(\phi^\theta(x,t))
\end{equation}
This integral equation should not be confused with the piecewise-quadratic $\mathbb{R} \rightarrow \Omega$ map, $y \rightarrow \int_0^y v^\theta(x) dx$. Both $x \rightarrow \phi^\theta(x,t)$ and $t \rightarrow \phi^\theta(x,t)$ are not piecewise quadratic.

Letting $x$ vary and fixing $t$, $x \rightarrow \phi^\theta(x,t)$ is an $\Omega \rightarrow \Omega$ transformation. Without loss of generality, we may set $t=1$ and define $T^{\theta}(x) = \phi^\theta(x,1)$ where $\theta \in \mathbb{R}^d$.
The solution $\phi(x,t)$ to this ODE is the composition of a finite number of solutions $\psi$: 
\begin{equation}\label{eq:ode_solution}
\phi^\theta(x,t) = \big(\psi_{\theta,c_m}^{t_m} \circ \psi_{\theta,c_{m-1}}^{t_{m-1}} \circ \cdots \circ \psi_{\theta,c_2}^{t_2} \circ \psi_{\theta,c_1}^{t_1} \big)(x)
\end{equation}
where $m$ is the number of cells visited and $\psi_{\theta,c}^{t}$ is the solution of a basic ODE $\cfrac{\partial\psi(x,t)}{\partial t}=v^\theta(\psi(x,t))$
with an $\mathbb{R} \rightarrow \mathbb{R}$ affine velocity field: 
$v^\theta(\psi) = a^\theta \psi + b^\theta$ and an initial condition $\psi(x,0) = x$.
The integration details for this ODE are included in \cref{sec:integration_details}.
\begin{equation*}\label{eq:integral_simple}
\frac{\partial\psi(x,t)}{\partial t}=v^\theta(\psi(x,t))=a^\theta \psi(x,t) + b^\theta \longrightarrow 
\psi(x,t) = x e^{t a^\theta} + \Big(e^{t a^\theta}-1\Big) \frac{b^\theta}{a^\theta}
\end{equation*}

CPA-based transformations are easy to implement (including in GPU) and can be fast evaluated with very high accuracy using ad-hoc numerical methods or even exactly using the closed-form solution \cite{Freifeld2015}. In addition, these finite dimensional transformations have continuously-defined velocity fields (without discretization) and allow optional constraints such as zero-velocity at the boundary.

\subsection{Integration of Affine ODE}\label{sec:integration_details}

Given the initial value problem (IVP) defined by an ordinary differential equation (ODE) together with an initial condition\footnote{we use $\psi$ to refer to the function $\psi(x,t)$}:
\begin{equation}
\frac{\partial\psi(x,t)}{\partial t}=v^\theta(\psi(x,t))=a^\theta \psi + b^\theta \quad; \quad \psi(x,0) = x
\end{equation}

\subsubsection{Analytical Solution}\label{sec:analytical_solution}

To obtain the analytical solution we follow the separation of variables method, in which algebra allows one to rewrite the equation so that each of two variables occurs on a different side of the equation. We rearrange the ODE and integrate both sides:
\begin{equation}
\frac{\partial\psi}{\partial t}=a^\theta \psi + b^\theta \rightarrow 
\frac{\partial\psi}{a^\theta \psi + b^\theta}=\partial t \rightarrow
\frac{1}{a^\theta}\int\frac{a^\theta}{a^\theta \psi + b^\theta}\partial\psi=\int \partial t 
\end{equation}

\begin{gather}
\frac{1}{a^\theta}\log (a^\theta \psi + b^\theta)=t + C \\
a^\theta \psi + b^\theta = \exp (t a^\theta + C a^\theta ) \\
\psi = \frac{1}{a^\theta} \Big( \exp (t a^\theta + C a^\theta ) -  b^\theta \Big)
\end{gather}

The initial condition is then used to solve the unknown integration constant $C$:
\begin{equation}
\psi(x,0) = x \rightarrow C = \frac{1}{a^\theta} \log (a^\theta x + b^\theta)
\end{equation}

With that, we can obtain the closed-form expression for the IVP:
\begin{equation}
\psi = \frac{1}{a^\theta} \Big( \exp \big(t a^\theta + \frac{a^\theta}{a^\theta} \log (a^\theta x + b^\theta)\big) -  b^\theta \Big)  = 
\frac{1}{a^\theta} \Big( (a^\theta x + b^\theta) e^{t a^\theta} -  b^\theta \Big)
\end{equation}

\begin{equation}\label{eq:analytical_solution}
\boxed{\psi(x,t) = x e^{t a^\theta} + \Big(e^{t a^\theta}-1\Big) \frac{b^\theta}{a^\theta}}
\end{equation}

\subsubsection{Matrix Form Solution}\label{sec:matrix_form_solution}

As an alternative, we can compute the IVP solution by transforming the system into matrix form and then using the matrix exponential operation. First, let's rearrange the ODE into a $2\times2$ matrix:

\begin{equation}
\frac{\partial \psi}{\partial t}=a^\theta \psi + b^\theta \rightarrow 
\begin{bmatrix} \frac{\partial\psi}{\partial t} \\ 0 \end{bmatrix} = 
\begin{bmatrix} a^\theta & b^\theta \\ 0 & 0 \end{bmatrix}\begin{bmatrix} \psi \\ 1 \end{bmatrix} \rightarrow 
\frac{\partial \tilde{\psi}}{\partial t} = 
\tilde{A} \tilde{\psi} 
\end{equation}

where $A^\theta = \begin{bmatrix} a^\theta & b^\theta \end{bmatrix}$ is the affine vector, $\tilde{A} = 
\begin{bmatrix} A \\ \vec{0} \end{bmatrix} = 
\begin{bmatrix} a^\theta & b^\theta \\ 0 & 0 \end{bmatrix} $ is the augmented affine matrix, and $\tilde{x} = \begin{bmatrix} x \\ 1 \end{bmatrix}$ and $\tilde{\psi} = \begin{bmatrix} \psi \\ 1 \end{bmatrix}$ are augmented vectors for $x$ and $\psi$ respectively.

The solution to the augmented ODE $\dot{\tilde{\psi}} = \tilde{A} \tilde{\psi}$ is given via the exponential matrix action: $\tilde{\psi} = \tilde{x} \cdot e^{t\tilde{A}}$.
The matrix exponential of a matrix $M$ can be mathematically defined by the Taylor series expansion, likewise to the exponential function $e^{x}$:
\begin{equation}
e^{x} = \sum_{k=0}^{\infty} \frac{x^k}{k!} \quad \rightarrow \quad e^{M} = \sum_{k=0}^{\infty} \frac{M^k}{k!}
\end{equation}
In this case, $M = t \tilde{A} = \begin{bmatrix} ta^\theta & tb^\theta \\ 0 & 0 \end{bmatrix}$.
To calculate the matrix exponential of $M$, we need to derive an expression for the infinite powers of $M$. This can be obtained by generalizing the sequence of matrix products, or by using the Cayley Hamilton theorem.

(a) If we try to generalize the sequence of matrix product: 
\begin{quote}
\begin{equation}
M^0 = \mathbb{I} = \begin{bmatrix} 1 & 0 \\ 0 & 1 \end{bmatrix}
\end{equation}

\begin{equation}
M^1 = M = \begin{bmatrix} ta^\theta & tb^\theta \\ 0 & 0 \end{bmatrix}
\end{equation}

\begin{equation}
M^2 = M \cdot M = \begin{bmatrix} ta^\theta & tb^\theta \\ 0 & 0 \end{bmatrix}\begin{bmatrix} ta^\theta & tb^\theta \\ 0 & 0 \end{bmatrix}=\begin{bmatrix} t^2(a^\theta)^2 & a^\theta b^\theta t^2 \\ 0 & 0 \end{bmatrix}
\end{equation}

\begin{equation}
\begin{aligned}
M^3 = M^2 \cdot M &= \begin{bmatrix} t^2(a^\theta)^2 & a^\theta b^\theta t^2 \\ 0 & 0 \end{bmatrix}\begin{bmatrix} ta^\theta & tb^\theta \\ 0 & 0 \end{bmatrix} =\\ &=
\begin{bmatrix} t^3(a^\theta)^3 & (a^\theta)^2 b^\theta t^3 \\ 0 & 0 \end{bmatrix}
\end{aligned}
\end{equation}
\begin{center}
    $\vdots$
\end{center}
\begin{equation}
M^k = M^{k-1} \cdot M = 
\begin{bmatrix} t^k(a^\theta)^k & (a^\theta)^{k-1} b^\theta t^k \\ 0 & 0 \end{bmatrix} = 
t^k(a^\theta)^k \begin{bmatrix} 1 & \frac{b^\theta}{a^\theta} \\ 0 & 0 \end{bmatrix}
\end{equation}

Then, 
\begin{equation}
\begin{aligned}\label{eq:matrix_solution_1}
e^{M} &= \sum_{k=0}^{\infty} \frac{M^k}{k!} = \frac{M^0}{0!} + \sum_{k=1}^{\infty} \frac{M^k}{k!} = \mathbb{I} + \sum_{k=1}^{\infty} \frac{M^k}{k!} = \\ &= 
\begin{bmatrix} 1 & 0 \\ 0 & 1 \end{bmatrix} + 
\sum_{k=1}^{\infty} \frac{t^k(a^\theta)^k}{k!} \begin{bmatrix} 1 & \frac{b^\theta}{a^\theta} \\ 0 & 0 \end{bmatrix} = 
\begin{bmatrix} 1 & 0 \\ 0 & 1 \end{bmatrix} + 
(e^{ta^\theta}-1) \begin{bmatrix} 1 & \frac{b^\theta}{a^\theta} \\ 0 & 0 \end{bmatrix} = \\ &=
\begin{bmatrix} e^{ta^\theta} & (e^{ta^\theta}-1)\frac{b^\theta}{a^\theta} \\ 0 & 1 \end{bmatrix}
\end{aligned}
\end{equation}

\end{quote}

(b) Rather, using the Cayley Hamilton theorem: 
\begin{quote}
The characteristic polynomial of the matrix $M$ is given by:
\begin{equation}
    p(\lambda) = \text{det}(\lambda I_2 - A) = 
    \begin{vmatrix}
        \lambda - ta^\theta & tb^\theta\\ 
        0 & \lambda 
    \end{vmatrix} = 
    \lambda^2 - ta^\theta \lambda   
\end{equation}

The Cayley-Hamilton theorem claims that, if we define $p(X) = X^2 - ta^\theta X$ then $p(M) = M^2 - ta^\theta M = 0_{2,2}$. Hence, $M^2 =  ta^\theta M$ and as a consequence, $M^k = ta^\theta M^{k-1} = (ta^\theta)^{k-1} M$.
Now, using the exponential matrix expression we arrive at the same outcome as before: 
\begin{equation}\label{eq:matrix_solution_2}
\begin{aligned}
e^{M} &= 
\mathbb{I} + \sum_{k=1}^{\infty} \frac{M^k}{k!} = 
\mathbb{I} + \sum_{k=1}^{\infty} \frac{(ta^\theta)^{k-1}}{k!} M = \\ &= 
\mathbb{I} + \frac{1}{ta^\theta} \sum_{k=1}^{\infty} \frac{(ta^\theta)^{k}}{k!} M = 
\mathbb{I} + \frac{1}{ta^\theta} (e^{ta^\theta}-1) M = \\ &= 
\begin{bmatrix} 1 & 0 \\ 0 & 1 \end{bmatrix} + 
\frac{e^{ta^\theta}-1}{ta^\theta} 
\begin{bmatrix} ta^\theta & tb^\theta \\ 0 & 0 \end{bmatrix} = 
\begin{bmatrix} e^{ta^\theta} & (e^{ta^\theta}-1)\frac{b^\theta}{a^\theta} \\ 0 & 1 \end{bmatrix}
\end{aligned}
\end{equation}

\end{quote}

Finally, from any of these two identical expressions (\cref{eq:matrix_solution_1,eq:matrix_solution_2}), we can extract the same result obtained in the analytical solution (\cref{eq:analytical_solution}):
\begin{equation}
e^{M} = e^{t \tilde{A}}=\begin{bmatrix} e^{ta^\theta} & (e^{ta^\theta}-1)\frac{b^\theta}{a^\theta} \\ 0 & 1 \end{bmatrix}
\end{equation}
\begin{equation}
\tilde{\psi} = \tilde{x} \cdot e^{t\tilde{A}} \quad \rightarrow \quad 
\begin{bmatrix} \psi \\ 1 \end{bmatrix} = 
\begin{bmatrix} x \\ 1 \end{bmatrix}
\begin{bmatrix} e^{ta^\theta} & (e^{ta^\theta}-1)\frac{b^\theta}{a^\theta} \\ 0 & 1 \end{bmatrix} \quad \rightarrow \quad 
\boxed{\psi = x e^{t a^\theta} + \Big(e^{t a^\theta}-1\Big) \frac{b^\theta}{a^\theta}}
\end{equation}

\subsection{Closed-Form Integration of Continuous Piecewise-Affine ODE}\label{sec:closed_form_integration}

\cite{Freifeld2017} proposed the algorithm to compute the closed-form solution for the transformation $x \rightarrow T^\theta(x)$, that can be summarized as follows:

\begin{algorithm}
    \caption{Closed-Form Integration of Continuous Piecewise-Affine ODE}\label{alg:cap}
    \textbf{Inputs} \hspace{0.6em}
    \begin{tabular}[t]{ll}
        $x$ & \textit{input state}  \\
        $t$ & \textit{integration time} \\
        $\gamma(x)$ & \textit{membership function} \\
        $\textbf{L}$ & \textit{constraint matrix} \\
        $\boldsymbol{\theta}$ & \textit{transformation parameters} 
    \end{tabular}

    \textbf{Output} \hspace{0.4em}
    \begin{tabular}{ll}
        $\phi(x,t)$ & \textit{transformed state}
    \end{tabular}\\

    \begin{algorithmic}[1]
        \State $\textbf{B} \gets NullSpace(\textbf{L})$  \Comment{use \cref{sec:null_space}}
        \State $vec(\textbf{A}) = \textbf{B} \cdot \boldsymbol{\theta} = \sum_{j=1}^{d} \theta_j \cdot \textbf{B}_j = [\; \cdots \; \sum_{j=1}^{d} \theta_j a_c^{(j)} \; \sum_{j=1}^{d} \theta_j b_c^{(j)} \; \cdots \;]^T$
        \State $c \gets \gamma(x)$ \Comment{cell index}
        
        \Loop
        \State $a_c^\theta = \sum_{j=1}^{d} \theta_j a_c^{(j)}$ \Comment{slope coefficient}
        \State $b_c^\theta = \sum_{j=1}^{d} \theta_j b_c^{(j)}$ \Comment{intercept coefficient}
        \If{$v(x) \geq 0$} 
            \State $x_c \gets x_{c}^{max}$ \Comment{right boundary}
        \Else
            \State $x_c \gets x_{c}^{min}$ \Comment{left boundary}
        \EndIf

        \State $t_{hit}^\theta \gets \cfrac{1}{a_c^\theta} \log \bigg( \cfrac{a_c^\theta x_c + b_c^\theta}{a_c^\theta x + b_c^\theta} \bigg)$ \Comment{cell boundary hitting time}

        \If{$t_{hit}^\theta \ge t$}
            \State \Return $\phi^\theta(x,t) = \psi_c(x,t)$ \Comment{cell solution}
        \Else
            \State $t \gets t - t_{hit}^\theta$ \Comment{subtract traversed time}
            \State $x \gets x_c$  \Comment{move state to boundary}
            \If{$v(x) \geq 0$}
                \State $c \gets c+1$  \Comment{new cell at right}
            \Else
                \State $c \gets c-1$  \Comment{new cell at left}
            \EndIf
        \EndIf
        \EndLoop
        
    \end{algorithmic}
\end{algorithm}

Without loss of generality, let $t > 0$. For input state $x$, let $t$ be the integration time $t$, $\gamma(x)$ the membership function, $\textbf{L}$ the constraint matrix and $\boldsymbol{\theta}$ the transformation parameters. 

\begin{itemize}
    
    \item \textbf{Step 1:} Compute the null space of the constraint matrix $\textbf{L}$ using any of the four methods presented above (\cref{sec:null_space}): SVD decomposition, QR decomposition, Reduced Row Echelon Form (RREF) and Sparse Form. 
    
    \item \textbf{Step 2:} Use the obtained orthogonal basis $\textbf{B}$ to compute the linear transformation matrix $\textbf{A}$, following \cref{eq:linear_matrix}.
    
    \item \textbf{Step 3:} Compute the cell index $c = \gamma(x)$ using the membership function $\gamma(x)$
    
    \item \textbf{Step 4:} Obtain the slope and intercept coefficients for the current cell using parameters $\theta$ and the linear transformation matrix $\textbf{A}$:  
    
    $a_c^\theta = \sum_{j=1}^{d} \theta_j a_c^{(j)} \quad\quad b_c^\theta = \sum_{j=1}^{d} \theta_j b_c^{(j)}$

    \item \textbf{Step 5:} Compute the cell boundary points, based on the sign of the velocity function:  
    
    $\left\{\begin{matrix}
        x_c = x_{c}^{max}, \text{ if }  v(x) \geq 0 \\
        x_c = x_{c}^{min}, \text{ if }  v(x) < 0
    \end{matrix}\right.$

    \item \textbf{Step 6:}
    Calculate the \textit{hitting time} $t_{hit}$ at the boundary $x_c$:
    $$
    \psi_c^\theta(x,t_{hit}) = x_c 
    \longrightarrow
    t_{hit}^\theta = \frac{1}{a_c^\theta} \log \bigg( \frac{a_c^\theta x_c + b_c^\theta}{a_c^\theta x + b_c^\theta} \bigg)
    $$

    \item \textbf{Step 7:} 
    If $t_{hit}^\theta > t$ then $\phi^\theta(x,t) = \psi_c(x,t)$, otherwise repeat from step 4, with new values for
    $t = t - t_{hit}^\theta$, $x = x_c$ and
    $\left\{\begin{matrix}
        c = c+1, \text{ if }  v(x) \geq 0 \\
        c = c-1, \text{ if }  v(x) < 0
    \end{matrix}\right.$
\end{itemize}

This is an iterative process until convergence. The upper bound for $m$ is $\max(c_{1}, N_{\mathcal{P}}-c_{1} + 1)$, where $c_1$ refers to the first visited cell index.

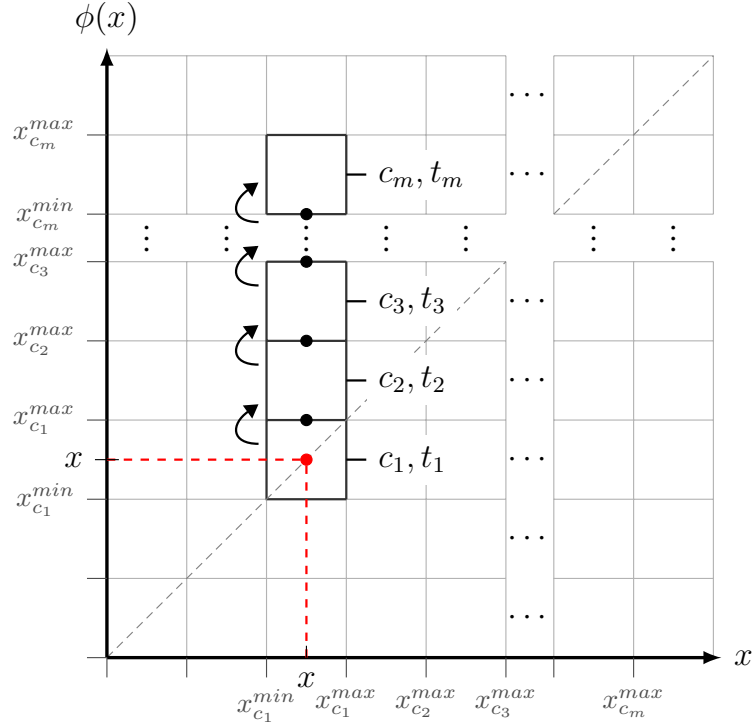
\begin{figure}[!ht]
    \begin{center}
    \scalebox{1.05}{
    \begin{tikzpicture}
    \pgfmathsetmacro{\N}{5};
    \pgfmathsetmacro{\M}{5};
    \pgfmathsetmacro{\P}{\N-1};
    \pgfmathsetmacro{\a}{0.15};
    \pgfmathsetmacro{\h}{0.6};
    \draw[step=1, black!30, thin] (0,0) grid (\N, \M);
    \draw[step=1, black!80, thick] (2,2) grid (3, \M);
    
    \draw[step=1, black!30, thin, shift={(0,5+\h)}] (0,0) grid (\N,2);
    \draw[step=1, black!30, thin, shift={(5+\h,0)}] (0,0) grid (2,\N);
    \draw[step=1, black!30, thin, shift={(5+\h,5+\h)}] (0,0) grid (2,2);
    \draw[step=1, black!80, thick, shift={(2,5+\h)}] (0,0) grid (1,1);

    \draw[thin, black!50, densely dashed] (0,0) -- (\N,\M);
    \draw[thin, black!50, densely dashed] (\N+\h,\M+\h) -- (\N+2+\h,\M+2+\h);

    \draw[-latex, very thick] (0,0) -- (\N+2.1+\h,0) node[right] {$x$};
    \draw[-latex, very thick] (0,0) -- (0,\M+2.1+\h) node[above] {$\phi(x)$};
    
    \draw[red, thick, dashed] (2.5, 0) -- (2.5, 2.5);
    \draw[red, thick, dashed] (0, 2.5) -- (2.5, 2.5);
    \filldraw[red] (2.5, 2.5) circle (2pt);
    \draw (2.5,\a) -- (2.5,0)  node[anchor=north, pos=1.25] {$x$};
    \draw (\a,2.5) -- (-\a,2.5)  node[anchor=east] {$x$};

    \draw[black!70, font=\footnotesize] (0,0) -- (0,-0.25);
    \draw[black!70, font=\footnotesize] (1,0) -- (1,-0.25);
    \draw[black!70, font=\footnotesize] (2,0) -- (2,-0.25)  node[anchor=north] {$x_{c_1}^{min}$};
    \draw[black!70, font=\footnotesize] (3,0) -- (3,-0.25)  node[anchor=north] {$x_{c_1}^{max}$};
    \draw[black!70, font=\footnotesize] (4,0) -- (4,-0.25)  node[anchor=north] {$x_{c_2}^{max}$};
    \draw[black!70, font=\footnotesize] (5,0) -- (5,-0.25)  node[anchor=north] {$x_{c_3}^{max}$};
    \draw[black!70, font=\footnotesize] (5+\h,0) -- (5+\h,-0.25) ;
    \draw[black!70, font=\footnotesize] (6+\h,0) -- (6+\h,-0.25)  node[anchor=north] {$x_{c_m}^{max}$};

    \draw[black!70, font=\footnotesize] (0,0) -- (-0.25,0);
    \draw[black!70, font=\footnotesize] (0,1) -- (-0.25,1);
    \draw[black!70, font=\footnotesize] (0,2) -- (-0.25,2)  node[anchor=east] {$x_{c_1}^{min}$};
    \draw[black!70, font=\footnotesize] (0,3) -- (-0.25,3)  node[anchor=east] {$x_{c_1}^{max}$};
    \draw[black!70, font=\footnotesize] (0,4) -- (-0.25,4)  node[anchor=east] {$x_{c_2}^{max}$};
    \draw[black!70, font=\footnotesize] (0,5) -- (-0.25,5)  node[anchor=east] {$x_{c_3}^{max}$};
    \draw[black!70, font=\footnotesize] (0,5+\h) -- (-0.25,5+\h)  node[anchor=east] {$x_{c_m}^{min}$};
    \draw[black!70, font=\footnotesize] (0,6+\h) -- (-0.25,6+\h)  node[anchor=east] {$x_{c_m}^{max}$};

    \filldraw[black] (2.5, 3) circle (2pt);
    \filldraw[black] (2.5, 4) circle (2pt);
    \filldraw[black] (2.5, 5) circle (2pt);
    \filldraw[black] (2.5, 5+\h) circle (2pt);

    \draw[thick] (3,2.5) -- (3.25,2.5)  node[anchor=west,fill=white] {$c_1 , t_1$};
    \draw[thick] (3,3.5) -- (3.25,3.5)  node[anchor=west,fill=white] {$c_2 , t_2$};
    \draw[thick] (3,4.5) -- (3.25,4.5)  node[anchor=west,fill=white] {$c_3 , t_3$};
    \draw[thick] (3,5.5+\h) -- (3.25,5.5+\h)  node[anchor=west,fill=white] {$c_m , t_m$};
    \foreach \i in {0,...,4}{
        \node at (\i+.5,5.4) {$\vdots$};
        \node at (4.7+\h,\i+0.5) {$\cdots$};
    }
    \node at (4.7+\h,5.5+\h) {$\cdots$};
    \node at (4.7+\h,6.5+\h) {$\cdots$};
    \node at (5.5+\h,5.4) {$\vdots$};
    \node at (6.5+\h,5.4) {$\vdots$};

    \foreach \i in {4,...,6}{
        \draw[-Triangle, shift={(1.9,\i-1.3)}, thick] (0, 0) .. controls(-0.35,0) and (-0.35,0.25) .. (0, 0.5);
    }
    \draw[-Triangle, shift={(1.9,5.5)}, thick] (0, 0) .. controls(-0.35,0) and (-0.35,0.25) .. (0, 0.5);
\end{tikzpicture}
    }
    \caption{Iterative process of integration: starting at the initial cell $c_1$ at integration time $t_1=1$, several cells are crossed, and the process finishes at cell $c_m$ at integration time $t_m$.}
    \label{fig:integration_process_grid}
    \end{center}
\end{figure}

\begin{remark}
    The closed-form term was drawn from \cite{Freifeld2015}, who state that the integration of one-dimensional CPA fields has a closed form solution (Theorem 2). Definitions vary and are ambiguous \cite{borwein2013closed}, but we extend the definition presented in the article (\cref{fn:closedform}) and define a closed form solution as an expression for an exact solution given a finite amount of data.
    An infinite sum would generally not be considered closed-form. Closed form solutions and numerical solutions are similar in that they both can be evaluated with a finite number of standard operations, but differ in that a closed-form solution is exact whereas a numerical solution is only approximate.
    Even though the number of iterations depend on the number of cells, for a given $N_\mathcal{P}$, both the integral solution and its gradient are given explicitly (in terms of function and mathematical operations from a given generally-accepted set) and are also computed exactly, without any approximation given that scaling and squaring (\cref{sec:scaling_squaring}) is not applied. 
\end{remark}

\subsection{Closed-Form Partial Derivatives of $\phi^\theta(x,t)$ w.r.t. $\theta$}\label{sec:closed_form_derivatives}

Let's recall that the trajectory (the solution for the integral equation \ref{eq:integral}) is a composition of a finite number of solutions $\psi$, as given by:
\begin{equation}
\phi^\theta(x,t) = \Big(\psi_{\theta,c_m}^{t_m} \circ \psi_{\theta,c_{m-1}}^{t_{m-1}} \circ \cdots \circ \psi_{\theta,c_2}^{t_2} \circ \psi_{\theta,c_1}^{t_1} \Big)(x)
\end{equation}
During the iterative process of integration, several cells are crossed, starting from $c_1$ at integration time $t_1=1$, and finishing at $c_m$ at integration time $t_m$. The integration time $t_m$ of the last cell $c_m$ can be calculated by subtracting from the initial integration time the accumulated boundary hitting times $t_{hit}$: 
\begin{equation}
t_m = t_1 - \sum_{i=1}^{m-1} t_{hit}^\theta(c_i, x_i)
\end{equation}

The final integration point $x_m$ is the boundary of the penultimate cell $c_{m-1}$: $x_m = x_{c_{m-1}}$. In case only one cell is visited, both time and space remain unchanged: $t_m = 1$ and $x_m = x$. Taking this into consideration, the trajectory can be calculated as follows:
\begin{equation}\label{eq:closed_form_integration}
\phi^\theta(x,t) = \psi^\theta(x=x_m,t=t_m) = 
\bigg(
    x e^{t a_c} + \Big(e^{t a_c}-1\Big) \frac{b_c}{a_c}
\bigg)_{\substack{x = x_m \\ t = t_m}}
\end{equation}

In this section both the $\theta$-dependent $a^\theta$ and $b^\theta$ are expressed as $a$ and $b$ to avoid overloading the notation.
Therefore, the derivative can be calculated by going backwards in the integration direction. We are interested in the derivative of the trajectory w.r.t. the parameters 
$\boldsymbol{\theta} = \begin{bmatrix} \theta_1 & \theta_2 & \cdots & \theta_d \end{bmatrix} \in 
\mathbb{R}^{d}$. To obtain such gradient, we focus on the partial derivative w.r.t. one of the coefficients of $\theta$, i.e., $\theta_k$:
\begin{equation}\label{eq:derivative_chain_rule}
\frac{\partial \phi^\theta(x,t)}{\partial \theta_k} = 
\bigg(
\frac{\partial \psi^\theta(x,t)}{\partial \theta_k} + 
\frac{\partial \psi^\theta(x,t)}{\partial t^\theta} \cdot
\frac{\partial t^\theta}{\partial \theta_k} + 
\frac{\partial \psi^\theta(x,t)}{\partial x} \cdot
\frac{\partial x}{\partial \theta_k}
\bigg)_{\substack{x = x_m \\ t = t_m}}
\end{equation}
Let's derive each of the terms of this derivative:

\begin{center}
    $\color{Magenta}\boxed{\color{black} \cfrac{\partial \psi^\theta(x,t)}{\partial \theta_k}}\;$,
    $\color{Violet}\boxed{\color{black} \cfrac{\partial \psi^\theta(x,t)}{\partial t^\theta\vphantom{t_k}}}\;$,
    $\color{ForestGreen}\boxed{\color{black} \cfrac{\partial t^\theta}{\partial \theta_k}}\;$,
    $\color{Cyan}\boxed{\color{black} \cfrac{\partial \psi^\theta(x,t)}{\partial x\vphantom{x_k}}}\;$,
    $\color{Orange}\boxed{\color{black} \cfrac{\partial x}{\partial \theta_k}}\;$
\end{center}

\clearpage
\begin{quote}
\textbf{Notation.} Recall the expression for $\psi(x,t)$ from \cref{sec:cpa_diffeomorphic_transformations}:
\begin{equation}\label{eq:psi_expression_1}
\psi^\theta(x,t) = x e^{t a_c} + \Big(e^{t a_c}-1\Big) \frac{b_c}{a_c}
\end{equation}
Note that the slope $a_c$ and intercept $b_c$ are a linear combination of the orthogonal basis $B$ of the constraint matrix $L$, with $\theta$ as coefficients.
\begin{equation}\label{eq:vec_A}
vec(\textbf{A}) = \textbf{B} \cdot \boldsymbol{\theta} = \sum_{j=1}^{d} \theta_j \cdot \textbf{B}_j
\end{equation}
If we define one of the components from the orthogonal basis $\textbf{B}_j$ as:
\begin{equation}\label{eq:basis}
\textbf{B}_{j} = \begin{bmatrix} a_1^{(j)} & b_1^{(j)} & \cdots & a_c^{(j)} & b_c^{(j)} & \cdots & a_{N_\mathcal{P}}^{(j)} & b_{N_\mathcal{P}}^{(j)}\end{bmatrix}^T
\end{equation}
Then,
\begin{equation}\label{eq:theta_dot_basis}
\theta_j \cdot \textbf{B}_j = \begin{bmatrix} & \cdots & \theta_j a_c{(j)} & \theta_j b_c{(j)} & \cdots & \end{bmatrix}^T
\end{equation}
And as a result,
\begin{equation}\label{eq:vec_A_complete}
vec(\textbf{A}) = \sum_{j=1}^{d} \theta_j \cdot \textbf{B}_j = \begin{bmatrix} & \cdots & \sum_{j=1}^{d} \theta_j a_c^{(j)} & \sum_{j=1}^{d} \theta_j b_c^{(j)} & \cdots & \end{bmatrix}^T
\end{equation}
Thus, the slope $a_c$ and intercept $b_c$ (parameters of the affine transformation) are denoted as follows:
\begin{equation}\label{eq:ac_bc}
a_c = \sum_{j=1}^{d} \theta_j a_c^{(j)} \quad\quad\quad
b_c = \sum_{j=1}^{d} \theta_j b_c^{(j)}
\end{equation}
\end{quote}
\begin{center}
\rule{\textwidth}{.4pt}
\end{center}

\subsubsection{Expression for $\color{Magenta}\boxed{\color{black} \cfrac{\partial \psi^\theta(x,t)}{\partial \theta_k}}$}
\label{sec:expression_derivative_1}

On the basis of the expression for $\psi(x,t)$,
\begin{equation}\label{eq:psi_expression_2}
\psi^\theta(x,t) = x e^{t a_c} + \Big(e^{t a_c}-1\Big) \frac{b_c}{a_c}
\end{equation}
apply the chain-rule to obtain the derivative w.r.t $\theta_k$:
\begin{equation}\label{eq:psi_derivative}
\frac{\partial \psi^\theta(x,t)}{\partial \theta_k} = 
\frac{\partial \psi^\theta(x,t)}{\partial a_c} \cdot
\frac{\partial a_c}{\partial \theta_k} +
\frac{\partial \psi^\theta(x,t)}{\partial b_c} \cdot
\frac{\partial b_c}{\partial \theta_k}
\end{equation}
Considering \cref{eq:psi_expression_2} it is immediate to obtain the partial derivatives of $a_c$ and $b_c$ w.r.t $\theta_k$:
\begin{equation}\label{eq:ac_derivative}
a_c = \sum_{j=1}^{d} \theta_j a_c^{(j)} \rightarrow \frac{\partial a_c}{\partial \theta_k} = a_c^{(k)}
\end{equation}
\begin{equation}\label{eq:bc_derivative}
b_c = \sum_{j=1}^{d} \theta_j b_c^{(j)} \rightarrow \frac{\partial b_c}{\partial \theta_k} = b_c^{(k)}
\end{equation}
Next, we find the derivative $\psi(x,t)$ w.r.t $a_c$ and $b_c$:
\begin{equation}\label{eq:psi_derivative_ac}
\frac{\partial \psi^\theta(x,t)}{\partial a_c} = 
x \, t \, e^{t a_c} + \Big(t\,e^{t a_c}\Big)\frac{b_c}{a_c} - \Big(e^{t a_c}-1\Big)\frac{b_c}{a_c^2} =
t \, e^{t a_c} \Big(x + \frac{b_c}{a_c} \Big) - \Big(e^{t a_c}-1\Big)\frac{b_c}{a_c^2}
\end{equation}
\begin{equation}\label{eq:psi_derivative_bc}
\frac{\partial \psi^\theta(x,t)}{\partial b_c} = 
\Big(e^{t a_c}-1\Big)\frac{1}{a_c}
\end{equation}
Finally, \cref{eq:ac_derivative,eq:bc_derivative,eq:psi_derivative_ac,eq:psi_derivative_bc} are joined together into \cref{eq:psi_derivative}:
\begin{equation}\label{eq:psi_derivative_developed}
\begin{split}
\frac{\partial \psi^\theta(x,t)}{\partial \theta_k} &= 
\frac{\partial \psi^\theta(x,t)}{\partial a_c} \cdot
\frac{\partial a_c}{\partial \theta_k} +
\frac{\partial \psi^\theta(x,t)}{\partial b_c} \cdot
\frac{\partial b_c}{\partial \theta_k} = \\
&=
\underbrace{
\Bigg(t \, e^{t a_c} \Big(x + \frac{b_c}{a_c} \Big) - \Big(e^{t a_c}-1\Big)\frac{b_c}{a_c^2}\Bigg) 
}_{\cfrac{\partial \psi^\theta(x,t)}{\partial a_c}}
\cdot
\underbrace{\vphantom{\Bigg(}
a_c^{(k)}
}_{\cfrac{\partial a_c}{\partial \theta_k}}
+
\underbrace{\vphantom{\Bigg(}
\Big(e^{t a_c}-1\Big)\frac{1}{a_c} 
}_{\cfrac{\partial \psi^\theta(x,t)}{\partial b_c}}
\cdot
\underbrace{\vphantom{\Bigg(}
b_c^{(k)}
}_{\cfrac{\partial b_c}{\partial \theta_k}}
= \\
&= 
a_c^{(k)} \, t \, e^{t a_c} \Big(x + \frac{b_c}{a_c} \Big) + \Big(e^{t a_c}-1\Big)\frac{b_c^{(k)} \, a_c - a_c^{(k)} \, b_c}{a_c^2}
\end{split}
\end{equation}

\begin{equation}\label{eq:psi_derivative_developed_final}
    \color{magenta}
    \boxed{\color{black} 
    \frac{\partial \psi^\theta(x,t)}{\partial \theta_k} = 
    a_c^{(k)} \, t \, e^{t a_c} \Big(x + \frac{b_c}{a_c} \Big) + \Big(e^{t a_c}-1\Big)\frac{b_c^{(k)} \, a_c - a_c^{(k)} \, b_c}{a_c^2}
    }\color{black}
\end{equation}

\subsubsection{Expression for $\color{Violet}\boxed{\color{black} \cfrac{\partial \psi^\theta(x,t)}{\partial t^\theta}}$}
\label{sec:expression_derivative_2}

Starting from the expression for $\psi(x,t)$,
\begin{equation}\label{eq:psi_expression_3}
\psi^\theta(x,t) = x e^{t a_c} + \Big(e^{t a_c}-1\Big) \frac{b_c}{a_c}
\end{equation}
we explicitly get the derivative w.r.t $t^\theta$:
\begin{equation}\label{eq:psi_derivative_t}
    \color{Violet}
    \boxed{\color{black} 
    \frac{\partial \psi^\theta(x,t)}{\partial t^\theta} = x \, a_c \, e^{t a_c} + a_c \, e^{t a_c} \frac{b_c}{a_c} = 
    e^{t a_c} \Big( a_c x + b_c \Big)
    }\color{black}
\end{equation}

\subsubsection{Expression for $\color{ForestGreen}\boxed{\color{black} \cfrac{\partial t^\theta}{\partial \theta_k}}$}
\label{sec:expression_derivative_3}

After visiting $m$ cells, the integration time $t^\theta$ can be expressed as:
\begin{equation}\label{eq:time}
t^\theta = t_1 - \sum_{i=1}^{m-1} t_{hit}^\theta(c_i, x_i)
\end{equation}
The derivative w.r.t. $\theta_k$ is obtained as follows:
\begin{equation}\label{eq:time_derivative_theta}
\frac{\partial t^\theta}{\partial \theta_k} = 
-\sum_{i=1}^{m-1} \frac{\partial t_{hit}^\theta(c_i, x_i)}{\partial \theta_k}
\end{equation}
where
\begin{equation}\label{eq:thit}
t_{hit}^\theta(c, x) = \frac{1}{a_c} \log \bigg( \frac{a_c x_c + b_c}{a_c x + b_c} \bigg)
\end{equation}
and $x_c$ is the boundary for cell index $c$. Now, apply the chain rule operation to the hitting time $t_{hit}^\theta(c, x)$ expression:
\begin{equation}\label{eq:thit_derivative_theta}
\frac{\partial t_{hit}^\theta(c, x)}{\partial \theta_k} = 
\frac{\partial t_{hit}^\theta(c, x)}{\partial a_c} \cdot 
\frac{\partial a_c}{\partial \theta_k} +
\frac{\partial t_{hit}^\theta(c, x)}{\partial b_c} \cdot 
\frac{\partial b_c}{\partial \theta_k}
\end{equation}
As it was previously derived in \cref{sec:expression_derivative_2}, considering \cref{eq:psi_expression_2} it is immediate to obtain the partial derivatives of $a_c$ and $b_c$ w.r.t $\theta_k$:
\begin{equation}\label{eq:ac_derivative_theta}
a_c = \sum_{j=1}^{d} \theta_j a_c^{(j)} \rightarrow \frac{\partial a_c}{\partial \theta_k} = a_c^{(k)}
\end{equation}
\begin{equation}\label{eq:bc_derivative_theta}
b_c = \sum_{j=1}^{d} \theta_j b_c^{(j)} \rightarrow \frac{\partial b_c}{\partial \theta_k} = b_c^{(k)}
\end{equation}
Next, we develop the derivatives of $t_{hit}^\theta(c, x)$ w.r.t $a_c$ and $b_c$:
\begin{equation}\label{eq:thit_derivative_ac}
\begin{split}
\frac{\partial t_{hit}^\theta(c, x)}{\partial a_c} &= 
-\frac{1}{a_c^2} \log \bigg( \frac{a_c x_c + b_c}{a_c x + b_c} \bigg) + 
\frac{1}{a_c} \frac{x_c (a_c x + b_c) - x (a_c x_c + b_c)}{(a_c x + b_c)(a_c x_c + b_c)} = \\ 
&= -\frac{1}{a_c^2} \log \bigg( \frac{a_c x_c + b_c}{a_c x + b_c} \bigg) + 
\frac{b_c}{a_c} \frac{x_c - x}{(a_c x + b_c)(a_c x_c + b_c)}
\end{split}
\end{equation}
\begin{equation}\label{eq:thit_derivative_bc}
\frac{\partial t_{hit}^\theta(c, x)}{\partial b_c} = 
\frac{1}{a_c} \frac{(a_c x + b_c) - (a_c x_c + b_c)}{(a_c x + b_c)(a_c x_c + b_c)} = 
\frac{x - x_c}{(a_c x + b_c)(a_c x_c + b_c)}
\end{equation}
Finally, these expressions (\cref{eq:ac_derivative_theta,eq:bc_derivative_theta,eq:thit_derivative_ac,eq:thit_derivative_bc}) are joined together to obtain the derivative of $t_{hit}^\theta(c, x)$ w.r.t $\theta_k$ into \cref{eq:thit_derivative_theta}:
\begin{equation}\label{eq:thit_derivative_theta_complete}
\begin{split}
\frac{\partial t_{hit}^\theta(c, x)}{\partial \theta_k} &= 
\frac{\partial t_{hit}^\theta(c, x)}{\partial a_c} \cdot 
\frac{\partial a_c}{\partial \theta_k} +
\frac{\partial t_{hit}^\theta(c, x)}{\partial b_c} \cdot 
\frac{\partial b_c}{\partial \theta_k} = \\
&= 
\underbrace{\vphantom{\Bigg(}
\Bigg(
    -\frac{1}{a_c^2} \log \bigg( \frac{a_c x_c + b_c}{a_c x + b_c} \bigg) + 
    \frac{b_c}{a_c} \frac{x_c - x}{(a_c x + b_c)(a_c x_c + b_c)}
\Bigg) 
}_{\cfrac{\partial t_{hit}^\theta(c, x)}{\partial a_c}}
\cdot
\underbrace{\vphantom{\Bigg(}
a_c^{(k)}
}_{\cfrac{\partial a_c}{\partial \theta_k}} +\\
&+
\underbrace{\vphantom{\Bigg(}
\frac{x - x_c}{(a_c x + b_c)(a_c x_c + b_c)} 
}_{\cfrac{\partial t_{hit}^\theta(c, x)}{\partial b_c}}
\cdot
\underbrace{\vphantom{\Bigg(}
b_c^{(k)} 
}_{\cfrac{\partial b_c}{\partial \theta_k}} =\\
&=
-\frac{a_c^{(k)}}{a_c^2} \log \bigg( \frac{a_c x_c + b_c}{a_c x + b_c} \bigg) +
\frac{(x - x_c)(b_c^{(k)} a_c - a_c^{(k)} b_c)}{a_c(a_c x + b_c)(a_c x_c + b_c)}
\end{split}
\end{equation}
\begin{equation}\label{eq:thit_derivative_theta_final}
    \color{ForestGreen}
    \boxed{\color{black}
    \frac{\partial t_{hit}^\theta(c, x)}{\partial \theta_k} = 
    -\frac{a_c^{(k)}}{a_c^2} \log \bigg( \frac{a_c x_c + b_c}{a_c x + b_c} \bigg) +
    \frac{(x - x_c)(b_c^{(k)} a_c - a_c^{(k)} b_c)}{a_c(a_c x + b_c)(a_c x_c + b_c)}
    }\color{black}
\end{equation}

\subsubsection{Expression for $\color{Cyan}\boxed{\color{black} \cfrac{\partial \psi^\theta(x,t)}{\partial x}}$}
\label{sec:expression_derivative_4}

Starting from the expression for $\psi(x,t)$,
\begin{equation}\label{eq:psi_expression_4}
\psi^\theta(x,t) = x e^{t a_c} + \Big(e^{t a_c}-1\Big) \frac{b_c}{a_c}
\end{equation}
we explicitly get the derivative w.r.t $x$:
\begin{equation}\label{eq:psi_derivative_x}
    \color{Cyan}
    \boxed{\color{black} 
    \frac{\partial \psi^\theta(x,t)}{\partial x} = e^{t a_c}
    }\color{black}
\end{equation}

\vspace{-0.5cm}
\subsubsection{Expression for $\color{Orange}\boxed{\color{black} \cfrac{\partial x}{\partial \theta_k}}$}
\label{sec:expression_derivative_5}

After visiting $m$ cells, the integration point $x$ is the boundary point of the last visited cell. In case only one cell is visited, the integration point remains unchanged.
In either case, $x$ does not depend on the parameters $\theta$, thus 
\begin{equation}\label{eq:x_derivative_theta}
    \color{Orange}
    \boxed{\color{black}
    \frac{\partial x}{\partial \theta_k} = 0
    }\color{black}
\end{equation}

\vspace{-0.5cm}
\subsubsection{Final Expression for $\cfrac{\partial \phi^\theta(x,t)}{\partial \theta_k}$}
\label{sec:expression_derivative_6}

Joining all the terms together and evaluating the derivative at $x = x_m$ and $t = t_m$ yields the expression for the partial derivative w.r.t. one of the coefficients of $\boldsymbol{\theta}$, i.e., $\theta_k$:
\begin{equation}\label{eq:derivative_complete}
\boxed{
\begin{aligned}
\quad \frac{\partial \phi^\theta(x,t)}{\partial \theta_k} &= 
\bigg(
\frac{\partial \psi^\theta(x,t)}{\partial \theta_k} + 
\frac{\partial \psi^\theta(x,t)}{\partial t^\theta} \cdot
\frac{\partial t^\theta}{\partial \theta_k} + 
\frac{\partial \psi^\theta(x,t)}{\partial x} \cdot
\frac{\partial x}{\partial \theta_k}
\bigg)_{\substack{x = x_m \\ t = t_m}} = \\
 &= a_{c_m}^{(k)} \, t_m \, e^{t_m a_{c_m}} \Big(x_m + \frac{b_{c_m}}{a_{c_m}} \Big) + 
\Big(e^{t_m a_{c_m}}-1\Big)\frac{b_{c_m}^{(k)} \, a_{c_m} - a_{c_m}^{(k)} \, b_{c_m}}{a_{c_m}^2} - \\ & \quad 
e^{t_m a_{c_m}} \Big( a_{c_m} x_m + b_{c_m} \Big) 
\sum_{i=1}^{m-1} 
\Bigg(
-\frac{a_{c_i}^{(k)}}{a_{c_i}^2} \log \bigg( \frac{a_{c_i} x_{c_i} + b_{c_i}}{a_{c_i} x_i + b_{c_i}} \bigg) + \\ & \quad
 \frac{(x_{c_i} - x_i)(b_{c_i}^{(k)} a_{c_i} - a_{c_i}^{(k)} b_{c_i})}{a_{c_i}(a_{c_i} x_i + b_{c_i})(a_{c_i} x_{c_i} + b_{c_i})}
\Bigg)
\end{aligned}
}
\end{equation}

\subsection{Closed-Form Transformation $\phi^\theta(x,t)$ when $a_c=0$}\label{sec:closed_form_slope_integration}

The special case where the slope is zero ($a_{c}=0$) leads to some indeterminations in the previous expressions for the integral (\cref{eq:closed_form_integration}) and derivative (\cref{eq:derivative_complete}), consequently in this section we analyze and provide specific solutions.
For this section, keep in mind the infinitesimal equivalents of the exponential and natural logarithm functions: $e^x - 1 \sim x$, if  $x \rightarrow 0$, and $\log x \sim x-1$, if  $x \rightarrow 1$.

Recall the expression for $\psi(x,t)$ from \cref{sec:cpa_diffeomorphic_transformations}:

\begin{equation}\label{eq:psi_init_azero}
\psi^\theta(x,t) = x e^{t a_c} + \Big(e^{t a_c}-1\Big) \frac{b_c}{a_c}
\end{equation}

Estimate  the limit of this expression when $a_c$ tends to zero:
\begin{equation}\label{eq:psi_limit_azero}
\lim_{a_c \to 0} \psi^\theta(x,t) = 
\lim_{a_c \to 0} x e^{t a_c} + \Big(e^{t a_c}-1\Big) \frac{b_c}{a_c} \simeq  
\lim_{a_c \to 0} x e^{0} + \lim_{a_c \to 0} \Big(t \, a_c\Big) \frac{b_c}{a_c} = x + t \, b_c
\end{equation}

\begin{equation}\label{eq:psi_final_azero}
\boxed{\psi^\theta(x,t) \Big|_{a_c \to 0} = x + t \, b_c}
\end{equation}

Also, for the gradient computation it will be convenient to derive the expression for the hitting time of the boundary of $U_c$:

\begin{equation}\label{eq:thit_init_azero}
t_{hit}^\theta = \frac{1}{a_c} \log \bigg( \frac{a_c x_c + b_c}{a_c x + b_c} \bigg)
\end{equation}

\begin{equation}\label{eq:thit_limit_azero}
\lim_{a_c \to 0} t_{hit}^\theta = 
\lim_{a_c \to 0} \frac{1}{a_c} \log \bigg( \frac{a_c x_c + b_c}{a_c x + b_c} \bigg)  \simeq 
\lim_{a_c \to 0} \frac{1}{a_c}  \bigg( \frac{a_c x_c + b_c}{a_c x + b_c} - 1 \bigg) = \frac{x_c - x}{b_c}
\end{equation}

Therefore,

\begin{equation}\label{eq:thit_final_azero}
\boxed{t_{hit}^\theta \Big|_{a_c \to 0} = \frac{x_c - x}{b_c}}
\end{equation}

\clearpage
\subsection{Closed-Form Partial Derivatives of $\phi^\theta(x,t)$ w.r.t. $\theta$ when $a_c=0$}\label{sec:closed_form_slope_derivative}

As the special case $a_{c}=0$ leads to some indeterminations, in this section we also provide a specific solution for the derivative when $a_c=0$. 
Note that the partial derivative w.r.t. one of the coefficients of $\theta$, i.e., $\theta_k$ is expressed as:
\begin{equation}\label{eq:psi_derivative_init_azero}
\frac{\partial \phi^\theta(x,t)}{\partial \theta_k} = 
\bigg(
\frac{\partial \psi^\theta(x,t)}{\partial \theta_k} + 
\frac{\partial \psi^\theta(x,t)}{\partial t^\theta} \cdot
\frac{\partial t^\theta}{\partial \theta_k} + 
\frac{\partial \psi^\theta(x,t)}{\partial x} \cdot
\frac{\partial x}{\partial \theta_k}
\bigg)_{\substack{x = x_m \\ t = t_m}}
\end{equation}

To particularize this derivative to the special case $a_c=0$, we first introduce the expressions obtained in \cref{sec:closed_form_derivatives} and then apply the limit when $a_c$ tends to zero. Note that the last two terms $\cfrac{\partial \psi^\theta(x,t)}{\partial x}$ and $\cfrac{\partial x}{\partial \theta_k}$ are not analyzed since $\cfrac{\partial x}{\partial \theta_k} = 0 \; \forall a_c$.

\subsubsection{Expression for $\cfrac{\partial \psi^\theta(x,t)}{\partial \theta_k}$ when $a_c=0$}
\label{sec:closed_form_slope_2a}

\begin{equation}\label{eq:psi_derivative_theta_init_azero}
\frac{\partial \psi^\theta(x,t)}{\partial \theta_k} = 
a_c^{(k)} \, t \, e^{t a_c} \Big(x + \frac{b_c}{a_c} \Big) + \Big(e^{t a_c}-1\Big)\frac{b_c^{(k)} \, a_c - a_c^{(k)} \, b_c}{a_c^2}
\end{equation}
\begin{equation}\label{eq:psi_derivative_theta_limit_azero}
\begin{split}
\lim_{a_c \to 0} \frac{\partial \psi^\theta(x,t)}{\partial \theta_k} &= 
\lim_{a_c \to 0} \Bigg(a_c^{(k)} \, t \, e^{t a_c} \Big(x + \frac{b_c}{a_c} \Big) + \Big(e^{t a_c}-1\Big)\frac{b_c^{(k)} \, a_c - a_c^{(k)} \, b_c}{a_c^2}
\Bigg) = \\
&=
a_c^{(k)} \, t \, x + 
\lim_{a_c \to 0} \bigg(a_c^{(k)} \, t \, \frac{b_c}{a_c} \bigg) +
\lim_{a_c \to 0} \bigg(t a_c \frac{b_c^{(k)} \, a_c - a_c^{(k)} \, b_c}{a_c^2}\bigg) = \\
&= 
a_c^{(k)} \, t \, x + 
\lim_{a_c \to 0} \bigg(
a_c^{(k)} \, t \, \frac{b_c}{a_c} +
t b_c^{(k)} - 
a_c^{(k)} \, t \, \frac{b_c}{a_c}
\bigg) = \\
&=
t \Big( a_c^{(k)} x + b_c^{(k)} \Big)
\end{split}
\end{equation}

\subsubsection{Expression for $\cfrac{\partial \psi^\theta(x,t)}{\partial t^\theta}$ when $a_c=0$}
\label{sec:closed_form_slope_2b}

If the partial derivative of $\psi^\theta(x,t)$ w.r.t $t^\theta$ is given by:
\begin{equation}\label{eq:psi_derivative_t_init_azero}
\frac{\partial \psi^\theta(x,t)}{\partial t^\theta} = 
e^{t a_c} \Big( a_c x + b_c \Big)
\end{equation}
then,
\begin{equation}\label{eq:psi_derivative_t_limit_azero}
\lim_{a_c \to 0} \frac{\partial \psi^\theta(x,t)}{\partial t^\theta} = 
\lim_{a_c \to 0} e^{t a_c} \Big( a_c x + b_c \Big) = b_c
\end{equation}

\subsubsection{Expression for $\cfrac{\partial t^\theta}{\partial \theta_k}$ when $a_c=0$}
\label{sec:closed_form_slope_2c}

Recall that the partial derivative of the $t^\theta$ w.r.t $\theta_k$ depends on the previously hit boundaries:
\begin{equation}\label{eq:t_derivative_theta_init_azero}
\frac{\partial t^\theta}{\partial \theta_k} = 
-\sum_{i=1}^{m-1} \frac{\partial t_{hit}^\theta(c_i, x_i)}{\partial \theta_k}
\end{equation}
\begin{equation}\label{eq:thit_derivative_theta_init_azero}
\frac{\partial t_{hit}^\theta(c, x)}{\partial \theta_k} = 
-\frac{a_c^{(k)}}{a_c^2} \log \bigg( \frac{a_c x_c + b_c}{a_c x + b_c} \bigg) +
 \frac{(x - x_c)(b_c^{(k)} a_c - a_c^{(k)} b_c)}{a_c(a_c x + b_c)(a_c x_c + b_c)}
\end{equation}
Therefore,
\begin{equation}\label{eq:thit_derivative_theta_limit_azero}
\begin{split}
\lim_{a_c \to 0} \frac{\partial t_{hit}^\theta(c, x)}{\partial \theta_k} &= 
\lim_{a_c \to 0} \Bigg(
    -\frac{a_c^{(k)}}{a_c^2} \log \bigg( \frac{a_c x_c + b_c}{a_c x + b_c} \bigg) +
    \frac{(x - x_c)(b_c^{(k)} a_c - a_c^{(k)} b_c)}{a_c(a_c x + b_c)(a_c x_c + b_c)}
\Bigg) =\\
&=
\lim_{a_c \to 0} \Bigg(
    -\frac{a_c^{(k)}}{a_c^2} \bigg( \frac{a_c x_c + b_c}{a_c x + b_c} - 1 \bigg) +
    \frac{(x - x_c)(b_c^{(k)} a_c - a_c^{(k)} b_c)}{a_c(a_c x + b_c)(a_c x_c + b_c)}
\Bigg) = \\
&=
\lim_{a_c \to 0} \Bigg(
    -\frac{a_c^{(k)}}{a_c} \bigg( \frac{x_c - x}{\underbrace{a_c x + b_c}_{\to b_c}} \bigg) +
    \frac{(x - x_c)(b_c^{(k)} a_c - a_c^{(k)} b_c)}{a_c(\underbrace{a_c x + b_c}_{\to b_c})(\underbrace{a_c x_c + b_c}_{\to b_c})}
\Bigg) = \\
&=
\lim_{a_c \to 0} \Bigg(
    -\frac{a_c^{(k)} (x_c - x) }{a_c \, b_c} +
    \frac{(x - x_c)(b_c^{(k)} a_c - a_c^{(k)} b_c)}{a_c \, b_c^2}
\Bigg) = \\
&=
\lim_{a_c \to 0} \Bigg(
    -\frac{a_c^{(k)} (x_c - x) }{a_c \, b_c} +
    \frac{b_c^{(k)}(x - x_c)  }{b_c^2} -
    \frac{a_c^{(k)}(x - x_c)  }{a_c \, b_c} 
\Bigg) = \\
&=
\frac{b_c^{(k)}(x - x_c)  }{b_c^2}
\end{split}
\end{equation}
\begin{equation}
\frac{\partial t^\theta}{\partial \theta_k} \Bigg|_{a_c \to 0} = 
-\sum_{i=1}^{m-1} \frac{\partial t_{hit}^\theta(c_i, x_i)}{\partial \theta_k} \Bigg|_{a_c \to 0} = 
-\sum_{i=1}^{m-1} \frac{b_{c_i}^{(k)}(x_i - x_{c_i})  }{b_{c_i}^2}
\end{equation}

\subsubsection{Final Expression for $\cfrac{\partial \phi^\theta(x,t)}{\partial \theta_k}$ when $a_c=0$}\label{sec:closed_form_slope_2d}

Joining all the terms together and evaluating the derivative at $x = x_m$ and $t = t_m$ yields:
\begin{equation}\label{eq:phi_derivative_theta_final_azero}
\boxed{
\begin{aligned}
\frac{\partial \phi^\theta(x,t)}{\partial \theta_k} \Bigg|_{a_c \to 0} &= 
\Bigg(
    t \Big( a_c^{(k)} x + b_c^{(k)} \Big) -
    b_c \sum_{i=1}^{m-1} \frac{b_{c_i}^{(k)}(x_i - x_{c_i})  }{b_{c_i}^2}
\Bigg)_{\substack{x = x_m \\ t = t_m}} = \\
&=
t_m \Big( a_{c_m}^{(k)} x_m + b_{c_m}^{(k)} \Big) -
b_{c_m} \sum_{i=1}^{m-1} \frac{b_{c_i}^{(k)}(x_i - x_{c_i})  }{b_{c_i}^2}
\end{aligned}
}
\end{equation}

\subsection{Scaling-and-Squaring}\label{sec:scaling_squaring}

The scaling and squaring algorithm, also known as the binary exponentiation algorithm, is a technique used to efficiently calculate the power of a number. This algorithm is particularly useful in computational settings where it is necessary to perform repeated multiplication of large numbers, such as in cryptography or in certain scientific and engineering applications. The basic idea behind the scaling and squaring algorithm is to decompose the exponentiation operation as a product of smaller powers of two.

The algorithm is based on the well-known identity that $x^n = (x^{(n/2)})^2$ if $n$ is even, and $x^n = x*(x^{((n-1)/2)})^2$ if $n$ is odd. This identity can be used to compute the value of $x^n$ by recursively computing the value of $x^{n/2}$ until it is reduced to a small enough value that can be directly computed.

For instance, to calculate the power 43 of a number ($x^{43}$), we first decompose it into powers of two: $43 = 32 + 8 + 2 + 1$. Then, we calculate the powers of two using a recursive approach: $x^{43} = x^{32} * x^{8} * x^{2} * x$. The recursive approach of the scaling and squaring algorithm allows us to calculate large powers of a number with a time complexity of $\mathcal{O}(\log n)$, where $n$ is the exponent. This is significantly faster than the traditional naive approach, which has a time complexity of $\mathcal{O}(n)$.

One of the key advantages of the scaling-squaring algorithm is that it can be easily implemented in a recursive fashion, which allows for further optimization through the use of memorization. 
However, the scaling and squaring algorithm also has some disadvantages. The algorithm requires some pre-processing to scale the input number to a suitable range, which can be time-consuming for large numbers. 

We use the scaling-and-squaring method \cite{Moler2003,Higham2009} to approximate the numerical or closed-form integration of the velocity field. This method uses the following property of diffeomorphic transformations to accelerate the computation of the integral: $\phi(x,t+s) = \phi(x,t) \,\circ\,\phi(x,s)$. Namely, computing the transformation $\phi$ at time $t+s$ is equivalent to composing the transformations at time $t$ and $s$. The scaling-and-squaring method imposes $t=s$, so it only needs to compute one transformation and self-compose it: 
$\phi(x,2t) = \phi(x,t)\,\circ\,\phi(x,t)$. Repeating this procedure multiple times (N) we can efficiently approximate the integration:
\begin{equation}\label{eq:scaling_squaring}
\phi(x,t^{2N}) = \phi(x,t) \; \underbrace{\circ \; \cdots \; \circ}_{N} \; \phi(x,t)
\end{equation}
This procedure relates to the log-euclidean framework \cite{Arsigny2006,Arsigny2006a} that uses (inverse) scaling and squaring for exponential (logarithm) operation on velocity fields. 

The self-composition operation $\circ$ can be estimated by first approximating $\phi$ as a piecewise linear function and then evaluate itself onto it. We are interested in making this operation differentiable as well, a task that is explored in detail in the next \cref{sec:linear_interpolation_grid}.

\subsection{Linear Interpolation Grid}\label{sec:linear_interpolation_grid}

Let $f$ be a discretized one-dimensional function defined by two arrays, i.e. the input
$x = \{x_{1}, \cdots, x_{n}\}$, and the output $y = \{y_{1}, \cdots, y_{n}\}$, such that $x_{1} < x_{2} < \cdots < x_{n}$. We want to get an estimate of the function value at $x = \hat{x}$ using piecewise linear interpolation $\hat{y} = f(\hat{x})$. 

For a general irregular grid $x$, $\hat{y}$ is defined as follows:
\begin{equation}\label{eq:grid_y}
\hat{y} = \sum_{i=1}^{n} y_i \cdot \phi_{i}(x)
\end{equation}
where $\phi_{i}(x)$ are known as the basis (hat functions), denoted as:
\begin{equation}\label{eq:grid_basis}
\phi_{i}(x) = 
\left\{\begin{matrix}
    \cfrac{\hat{x} - x_{i-1}}{x_{i} - x_{i-1}} & & x_{i-1} < \hat{x} < x_{i}
    \\
    \cfrac{x_{i+1} - \hat{x}}{x_{i+1} - x_{i}}  & & x_{i} < \hat{x} < x_{i+1}
    \\
    0 & &  \text{otherwise}
\end{matrix}\right.
\end{equation}
for $i=2,\cdots,n-1$, with the boundary "half-hat" basis $\phi_{1}(x)$ and $\phi_{n}(x)$.

In a regular grid $x$ every pair of values are separated by the same distance $h=x_{i+1}-x_{i}=x_{i}-x_{i-1}$. In such case, the above expression for $\phi_{i}(x)$ simplifies to: 
\begin{equation}\label{eq:grid_basis_regular}
\phi_{i}(x) = 
\left\{\begin{matrix}
    (\hat{x} - x_{i-1})/h & & x_{i-1} < \hat{x} < x_{i}
    \\
    (x_{i+1} - \hat{x})/h  & & x_{i} < \hat{x} < x_{i+1}
    \\
    0 & &  \text{otherwise}
\end{matrix}\right.
\end{equation}

The piecewise linear interpolation method is differentiable, which is a necessary condition to propagate the loss to the rest of the model. Thus, let's compute the derivative of the piecewise linear interpolation function with respect to $y_{i}$, $\hat{x}$ and $x_{i}$:

\begin{equation}\label{eq:grid_y_derivative_yi}
\cfrac{\partial \hat{y}}{\partial y_{i}}=\phi_{i}(x)
\end{equation}
\begin{equation}\label{eq:grid_y_derivative_x}
\cfrac{\partial \hat{y}}{\partial \hat{x}}=\sum_{i=1}^{n} y_{i}(x)
\left\{\begin{matrix}
    1/(x_{i} - x_{i-1}) & & x_{i-1} < \hat{x} < x_{i}
    \\
    -1/(x_{i+1} - x_{i})  & & x_{i} < \hat{x} < x_{i+1}
    \\
    0 & &  \text{otherwise}
\end{matrix}\right.
\end{equation}
\begin{equation}\label{eq:grid_y_derivative_xi}
\cfrac{\partial \hat{y}}{\partial x_{i}}=\sum_{i=1}^{n}
\left\{\begin{matrix}
    (y_{i-1}(x) - y_{i}(x))\cfrac{\hat{x} - x_{i-1}}{(x_{i} - x_{i-1})^2}  & & x_{i-1} < \hat{x} < x_{i}
    \\
    (y_{i}(x) - y_{i+1}(x))\cfrac{x_{i+1} - \hat{x}}{(x_{i+1} - x_{i})^2}  & & x_{i} < \hat{x} < x_{i+1}
    \\
    0 & & \text{otherwise}
\end{matrix}\right.
\end{equation}

\paragraph{Self-composition} In order to apply the scaling-squaring approximate method we are interested in a special case of the piecewise linear interpolation in which the input $\hat{x}$ is one of the output values $y_k$, i.e. $\hat(x) = y_{k}$. In such case the derivative with respect to $y_k$ is extended as follows:

\begin{equation}\label{eq:grid_y_derivative_yk}
\cfrac{\partial \hat{y}}{\partial y_{k}}=
\cfrac{\partial \hat{y}}{\partial y_{i}}+
\cfrac{\partial \hat{y}}{\partial \hat{x}}=
\phi_{i}(x) + 
\sum_{i=1}^{n} y_{i}(x)
\left\{\begin{matrix}
    1/(x_{i} - x_{i-1}) & & x_{i-1} < \hat{x} < x_{i}
    \\
    -1/(x_{i+1} - x_{i})  & & x_{i} < \hat{x} < x_{i+1}
    \\
    0 & & \text{otherwise}
\end{matrix}\right.
\end{equation}

\clearpage
\section{Experiments and Results}\label{sec:results_2}

\subsection{Null Space of Constraint Matrix $\mathbf{L}$: Performance Analysis}\label{sec:results:null_space}

To obtain the null space of $\mathbf{L}$, four different methods were implemented (\cref{sec:null_space}): SVD decomposition, QR decomposition, Reduced Row Echelon Form (RREF), and Sparse form (SPARSE).
\textit{PyTorch's Benchmark} module was used to measure the amount of time it takes to compute the null space for different tessellation sizes ranging from 10 to 1500. 

The Blocked Autorange feature was used to take multiple measurements that (in total) elapsed at least 2 seconds, ensuring that the overhead for timing the measurements is a small portion of the total measurement time. This is fulfilled by first running the operation multiple times to warm it up and then increasing the number of runs until the total runtime is much larger than the measurement overhead. Once this point is reached, measurements are taken until the target time is accomplished.

\vspace{1cm}
\begin{figure}[!htb]
    \begin{center}
    \centerline{\includegraphics[width=0.8\linewidth]{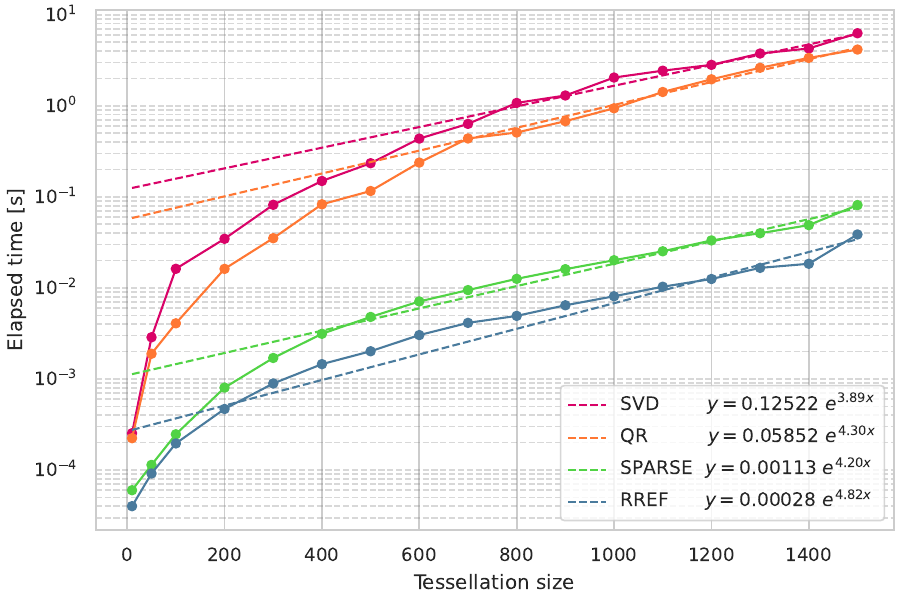}}
    \caption{Computation times (in seconds) for different velocity field basis and tessellation sizes. Elapsed time $y$ is fitted to an exponential function $y(x)=c_{1} e^{c_{2} \cdot x}$, where $x$ is the tessellation size, $c_1$ the intercept and $c_2$ the exponential slope.}
    \label{fig:basis_benchmark_fitted}
    \end{center}
\end{figure}

\clearpage
This procedure is repeated for each basis and tessellation size. Then, the obtained elapsed time $y$ is fitted to an exponential function $y(x)=c_{1} e^{c_{2} \cdot x}$, where $x$ is the tessellation size, $c_1$ the intercept and $c_2$ the exponential slope. \cref{fig:basis_benchmark_fitted} shows the obtained benchmark results and the fitted exponential functions as well. SVD basis is the most time-consuming basis, followed by QR which requires on average half the time ($c_1^{SVD}=0.125$ vs $c_1^{QR}=0.058$). Given that the RREF and the SPARSE bases can be computed efficiently in closed-form, the computation time of these null spaces is several orders of magnitude (between 2 and 3) faster than SVD and QR. Nonetheless, note that the null space only needs to be computed once, and as a result, this operation is not a critical step in the whole process to obtain the diffeomorphic trajectory.

\subsection{Diffeomorphic Properties}\label{sec:results:diffeomorphic_properties}

\begin{figure*}[!htb]
    \begin{center}
    \begin{subfigure}{\linewidth}
        \centering
        \includegraphics[width=0.85\linewidth,trim=0 13 0 10,clip]{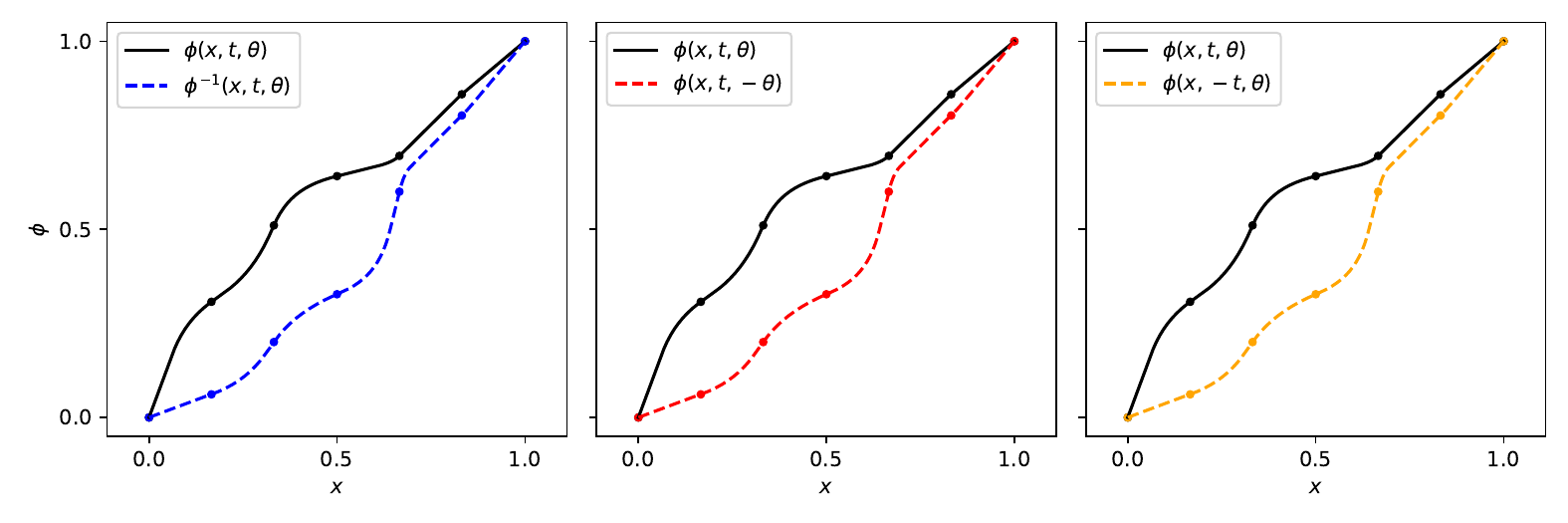}
        \caption{SVD}
        \label{fig:diffeoproperties_svd}
    \end{subfigure}
    \begin{subfigure}{\linewidth}
        \centering
        \includegraphics[width=0.85\linewidth,trim=0 13 0 10,clip]{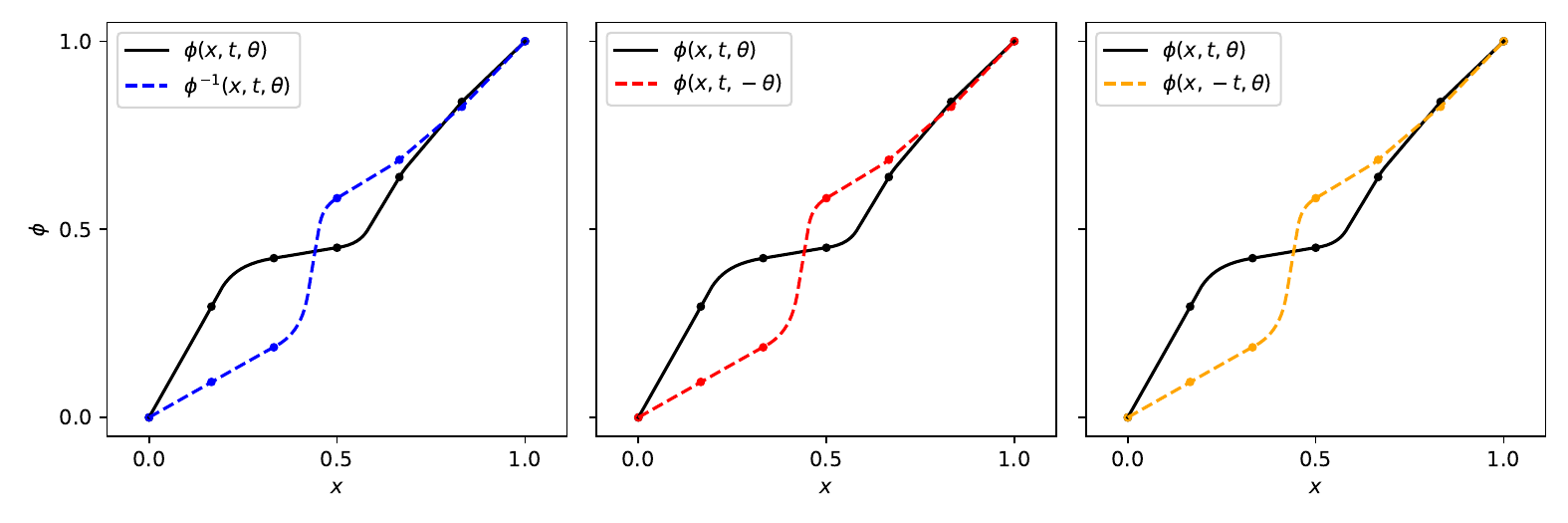}
        \caption{QR}
        \label{fig:diffeoproperties_qr}
    \end{subfigure}
    \begin{subfigure}{\linewidth}
        \centering
        \includegraphics[width=0.85\linewidth,trim=0 13 0 10,clip]{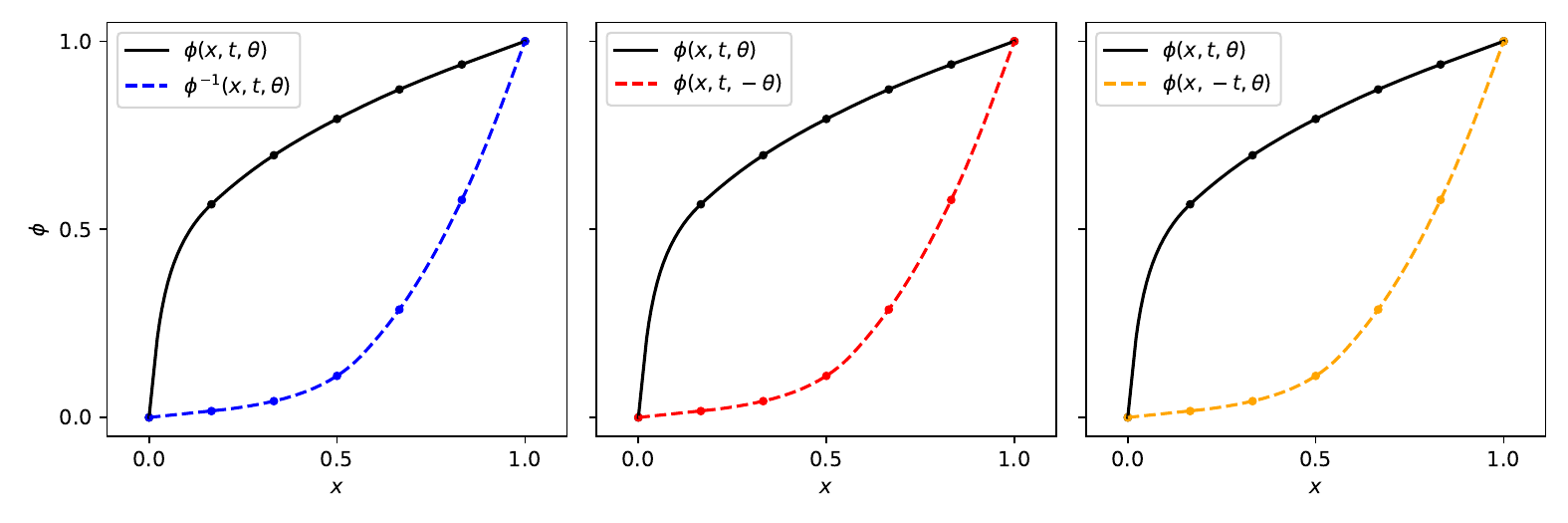}
        \caption{RREF}
        \label{fig:diffeoproperties_rref}
    \end{subfigure}
    \begin{subfigure}{\linewidth}
        \centering
        \includegraphics[width=0.85\linewidth,trim=0 13 0 10,clip]{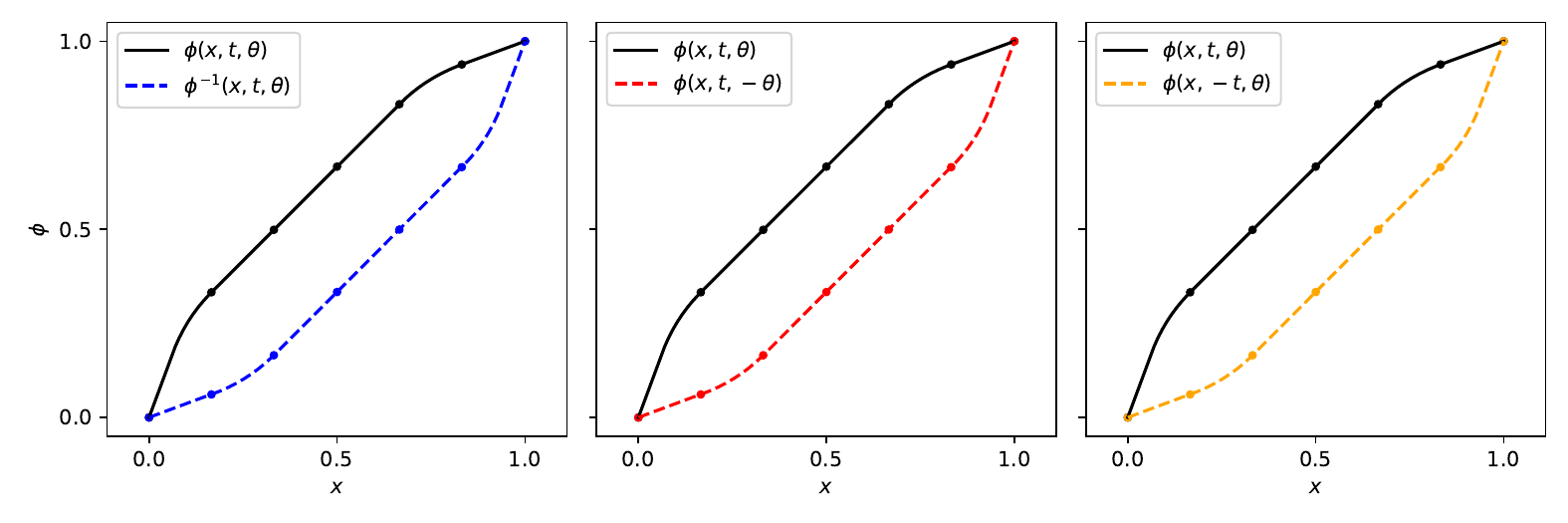}
        \caption{SPARSE}
        \label{fig:diffeoproperties_sparse}
    \end{subfigure}
    \caption{Qualitative validation of the diffeomorphic property $ \phi^{-1}(x,t,\theta) = \phi(x,t,-\theta) = \phi(x,-t,\theta)$ with $N_{\mathcal{P}}=5$ for different null space bases. \textbf{Left}: $\phi^{-1}(x,t,\theta)$ inverse function obtained by interpolation. \textbf{Center}: $\phi(x,t,-\theta)$ closed-form integration reversing the sign of the parameters. \textbf{Right}: $\phi(x,-t,\theta)$ closed-form integration reversing time.}
    \label{fig:diffeoproperties}
    \end{center}
\end{figure*}

Diffeomorphic functions are able to change the shape of an object through a smooth, continuous, and one-to-one mapping $\phi(x,t,\theta)$; where $x$ is the spatial dimension, $t$ the integration time and $\theta$ the transformation parameters. By understanding the properties of diffeomorphic transformations, we can gain a deeper appreciation for the versatility and flexibility of these mathematical objects. In this section, we will delve into one of the main characteristics of diffeomorphic transformations:

\begin{equation}\label{eq:diffeomorphic_properties}
    \phi^{-1}(x,t,\theta) = \phi(x,t,-\theta) = \phi(x,-t,\theta)
\end{equation}

That is, the inverse function can be obtained by reversing the sign of the parameters $\theta$, or alternatively, by reversing the time direction $t$. In this experiment, we integrate the ODE for a certain number of tessellation size $N_{\mathcal{P}}$ and for each null space basis (SVD, QR, RREF, SPARSE) and compute:
\begin{itemize}
    \item $\boxed{\;\;\phi(x,t,\theta)\;\,}\rightarrow$ closed-form integration. 
    \item $\boxed{\phi^{-1}(x,t,\theta)}\rightarrow$ empirical inverse function obtained by interpolation.
    \item $\boxed{\phi(x,t,-\theta)\,}\rightarrow$ closed-form integration reversing the sign of the parameters.
    \item $\boxed{\phi(x,-t,\theta)\,}\rightarrow$ closed-form integration reversing the time direction. 
\end{itemize}

The empirical inverse coincides with the closed-form integration reversing the sign of the parameters or the integration time, thus probing \cref{eq:diffeomorphic_properties}.

\clearpage
\subsection{Computation Time}\label{sec:results:computation_time} 

This section compares the performance of forward and backward operations between the proposed closed-form method and \textit{libcpab} \cite{detlefsen2018LIBCPAB}, which supports CPAB-based diffeomorphic transformations and implements a custom-made ODE solver \cite{Freifeld2015} that alternates between the analytic solution and a generic solver.

\begin{figure}[!htb]
    \begin{center}
    \centerline{\includegraphics[width=0.8\linewidth]{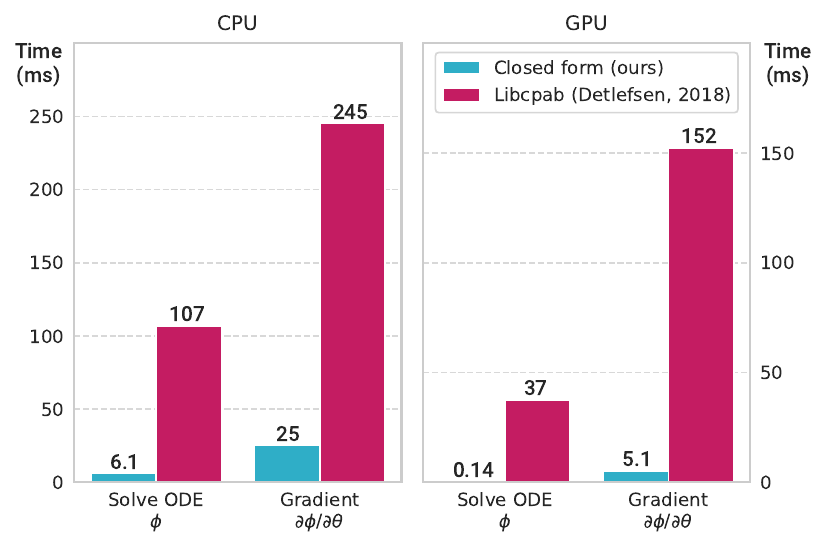}}
    \caption{Computation time ($ms$) for the forward (velocity field integration) and backward (gradient w.r.t. parameters) operations on CPU (\textbf{left}) and GPU (\textbf{right}). Our closed form method is compared to libcpab numeric solution implementation \cite{detlefsen2018LIBCPAB} which is based on \cite{Freifeld2017}. 
    }
    \label{fig:benchmark_times_compiled_linear}
    \end{center}
\end{figure}

The proposed CPA-diffeomorphic transformations were implemented on multiple backends for CPU (\textit{NumPy} and \textit{PyTorch} with \textit{C++}) and GPU (\textit{PyTorch} with \textit{CUDA}) and presented under an open-source library called Diffeomorphic Fast Warping \textit{DIFW}\footnote{\footnotesize\url{https://github.com/imartinezl/difw}} (see Supplementary Material). 
Speed tests show a \textbf{x18} and \textbf{x10} improvement on CPU over \textit{libcpab} for forward and backward operations respectively. On GPU the performance gain of \textit{DIFW} reaches \textbf{x260} and \textbf{x30} (see \cref{fig:benchmark_times_compiled_linear}). (Parameters: $10^{3}$ points in $\Omega$, $N_\mathcal{P}=30$, and batch size $40$).

Furthermore, a fast computation of the exponential function is essential for these operations. The exponential is embedded in the closed-form solution, and must be called several times during the forward and backward processes. \cref{apx:exponential_function} includes a performance analysis of different methods to compute the exponential function with floating-point arithmetic.

Performance comparisons with other methods such as SoftDTW, and ResNet-TW were avoided due to parametrization and complexity differences. SoftDTW and its gradient have quadratic time \& space complexity, but produce non-diffeomorphic warping functions. ResNet-TW requires multiple convolutional layers to incrementally compute the warping function and is limited by the recursive computation of the monotonic temporal constraints.

\subsection{Comparing ODE Solvers for CPA Velocity Functions}\label{sec:results:comparison_numerical}

Recall that we construct diffeomorphic curves by the integration of velocity functions, and specifically we work with continuous piecewise affine (CPA) velocity functions. Such integral equation or the equivalent ODE can be solved numerically using any generic ODE solver. 

\begin{figure*}[!htb]
    \begin{center}
    \begin{subfigure}{0.48\linewidth}
        \centering
        \includegraphics[width=\linewidth]{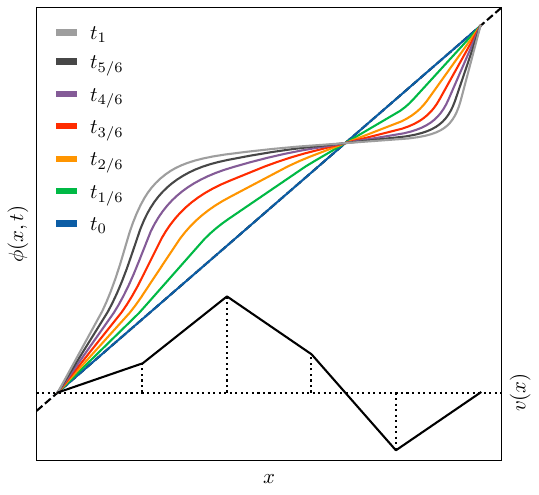}
        \caption{Velocity field $v(x)$ is a continuous piecewise affine function of five intervals. The integration of the flow equation with the initial condition $\phi(x, t_{0}) = x$ is shown for different integration times $t=\{0, 1/6, 2/6,\cdots, 1\}$. The gray curve corresponds to the final diffeomorphism at $t=1$.}
        \label{fig:integration_1}
    \end{subfigure}
    \hfill
    \begin{subfigure}{0.48\linewidth}
        \centering
        \includegraphics[width=\linewidth]{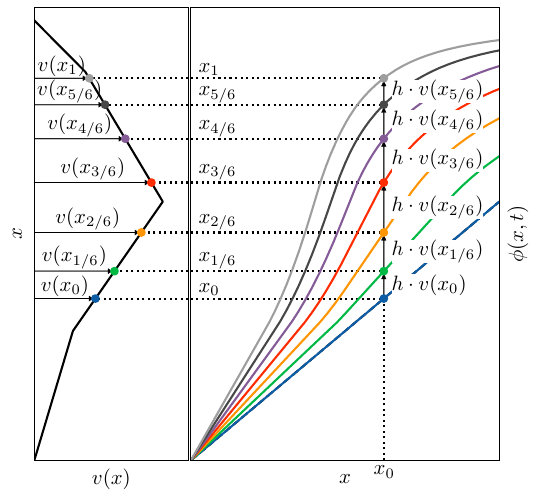}
        \caption{Illustration of the numerical integration of the flow equation: $\phi(x,t+h)=\phi(x,t) + h \cdot v(x)$. The velocity field $v(x)$ is shown along the y-axis. This scheme produces only increasing functions (invertible one-dimensional functions), when the integration step $h$ is chosen small enough.}
        \label{fig:integration_2}
    \end{subfigure}
    \caption{Construction of diffeomorphic curves by integration of velocity functions. In this chapter, we work with continuous piecewise affine (CPA) velocity functions.}
    \label{fig:integration}
    \end{center}
\end{figure*}

\vspace{-1em}
In this sense, \cite{Freifeld2015,Freifeld2017} proposed a specialized solver for integrating CPA fields which alternates between the analytic solution and a generic solver. This specialized solver is faster and more accurate than a generic solver since most of the trajectory is solved exactly, while the generic solver is only called in small portions. The two parameters used in \cite{Freifeld2017}, $N_{steps}$ and $n_{steps}$ imply two-step sizes, one large $\Delta t = t/N_{steps}$ used for exact updates, and one small, $\delta t = \Delta t / n_{steps}$ used for the generic-solver subroutine. By construction, the proposed closed-form solution is more accurate (exact solution) for both the ODE solution and for its derivative with respect to the velocity function parameters. We verified this empirically, comparing the specialized solver presented in \cite{Freifeld2015,Freifeld2017} against the closed-form methods described in \cref{sec:closed_form_integration,sec:closed_form_derivatives}. 

\begin{figure}[!htb]
    \begin{center}
    \begin{subfigure}{0.48\linewidth}
        \centering
        \includegraphics[width=\linewidth]{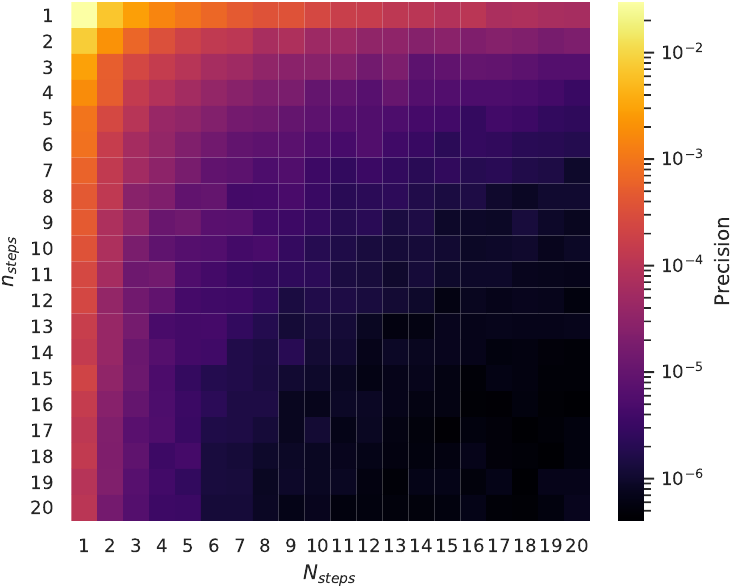}
        \caption{Integration error: $[10^{-6} \leftrightarrow 10^{-2}]$ range}
        \label{fig:error_integration}
    \end{subfigure}
    \hfill
    \begin{subfigure}{0.48\linewidth}
        \centering
        \includegraphics[width=\linewidth]{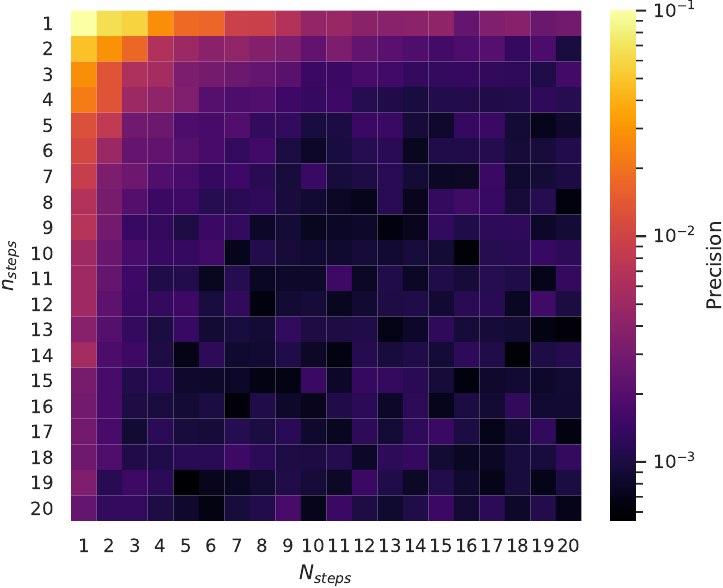}
        \caption{Gradient error: $[10^{-3} \leftrightarrow 10^{-1}]$ range}
        \label{fig:error_gradient}
    \end{subfigure}
    \caption{Comparing ODE numerical and closed-form methods for random CPA Velocity Functions: Precision obtained for different values of $N_{steps} \in [1,20]$ and $n_{steps} \in [1,20]$}
    \label{fig:error_integration_numerical}
    \end{center}
\end{figure}

\vspace{-1em}
We compared the integration (and gradient) error averaged over random CPA velocity fields; i.e. we drew 100 CPA velocity functions from the prior (normal distribution of zero mean and unit standard deviation), $\{v^{\theta_i}\}_{i=1}^{100}$, sampled uniformly 1000 points in $\Omega$, $\{x^{\theta_i}\}_{i=1}^{1000}$, and then computed the error:
\vspace{-0.75em}
\begin{equation}\label{eq:error_comparison}
    \epsilon = \frac{1}{100}\sum_{i=1}^{100} \; ||\text{abs}(\phi_{exact}^{\theta_i}(x_j) - \phi_{solver}^{\theta_i}(x_j))||_{\infty} \quad \forall j \in [1,1000]
\end{equation}
The intention with this analysis was to study the number of numerical steps necessary to guarantee good enough integration and gradient estimations. The warping function is a bijective map from $(0,1)$ to $(0,1)$.
\cref{fig:error_integration_numerical} shows the precision obtained for different values of $N_{steps} \in [1,20]$ and $n_{steps} \in [1,20]$. For instance, using the recommended values for $N_{steps}$ and $n_{steps}$ by the authors in \cite{Freifeld2017} ($N_{steps}=10$ and $n_{steps}=10$) we get a 5 decimals precision for the integration result but only 3 decimals for the gradient computation\footnote{Note, for comparative purposes, that transformations are computed for the interval $[0,1]$, i.e. precision $10^0$.}. Recall that \cref{eq:error_comparison} computes the maximum absolute difference, and that the total error can be much higher over the entire gradient function. Lack of precision in the computation of transformation gradient translates to inefficient search in the parameter space, which leads to worse training processes and non-optimal solutions.

\subsection{Scaling-and-Squaring Integration Method}\label{sec:results:scaling_squaring}

The scaling-and-squaring method can be applied to approximate the flow and speed up the forward and backward operations. 
This method needs to balance two factors to be competitive: the faster integration computation (scaling step) versus the extra time to self-compose the trajectory multiple times (squaring step). 
The scaled-down integral solution is computed using the closed-form expression (\cref{sec:closed_form_integration}), while adjoint equations are directly applied to compute derivatives. 

\begin{figure}[!htb]
    \begin{center}
    \begin{subfigure}[t]{0.46\linewidth}
        \centering
        \includegraphics[width=\linewidth]{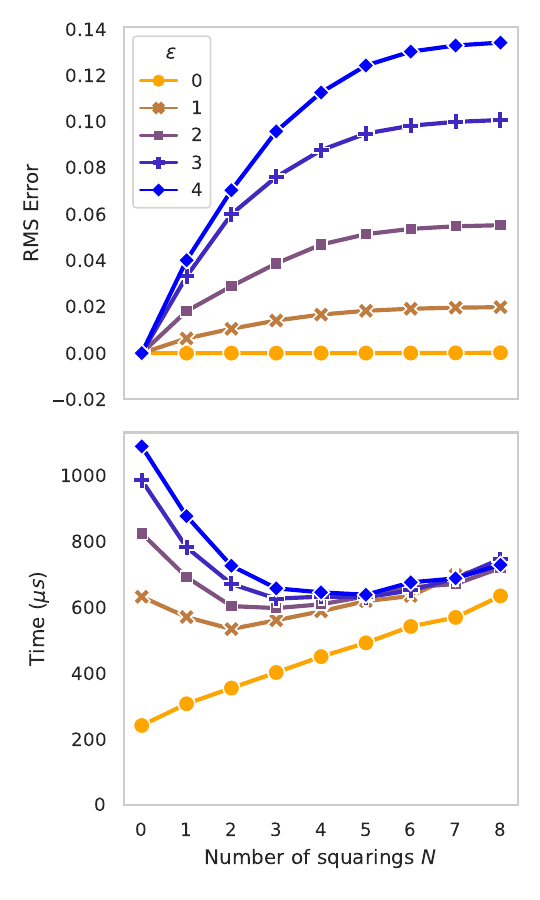}
        \caption{Forward operation (ODE integration)}
        \label{fig:scaling_squaring_integration}
    \end{subfigure}
    \hspace{1cm}
    \begin{subfigure}[t]{0.46\linewidth}
        \centering
        \includegraphics[width=\linewidth]{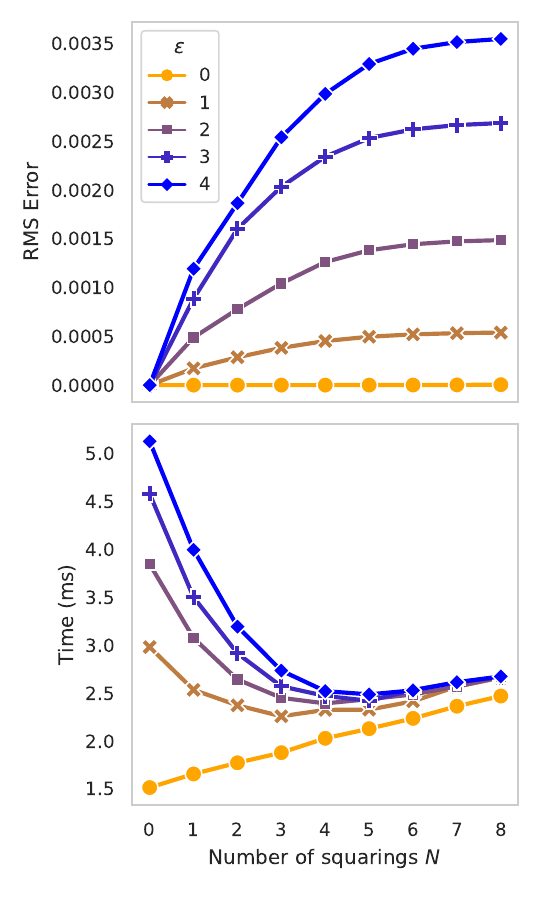}
        \caption{Backward operation (ODE solution gradient)}
        \label{fig:scaling_squaring_gradient}
    \end{subfigure}
    \caption{Scaling-and-squaring method: \textbf{Left}: Forward operation (ODE integration).
    \textbf{Right}: Backward operation (derivative of ODE solution). Top row illustrates the RMS error, while the bottom row show the computation time for each calculation.  
    }
    \label{fig:scaling_squaring}
    \end{center}
\end{figure}

\cref{fig:scaling_squaring_integration} shows (\textbf{top}) the integration error (RMS) and (\textbf{bottom}) the computation time for different squaring iterations $N \in [1,8]$. 
Similarly, the same \cref{fig:scaling_squaring_gradient} shows the error and the time for computing the derivative of the ODE solution for different scaling values $N \in [0,8]$. Under this setting, $\epsilon$ represents the parameter vector $\boldsymbol{\theta}=\textbf{1}_{d} \cdot \epsilon$. Thus, a large $\epsilon$ value generates large deformations. 

Executing more squaring iterations (large $N$) yields larger errors (both in the forward and backward operations). 
Even though the computation time increases for small deformations (positive slope for $\epsilon=0$) as the number of squaring iterations increases, experiments show that for large deformations, computation speed can be gained at the expense of small integration errors. For example, in the backward operation with $\epsilon=4$, takes around $5$ ms to compute if no scaling-and-squaring method is used. On the contrary, applying the scaling-and-squaring method 8 times (a reduction factor of $2^8=256$) reduces the computation time to around 2.5 ms (50\% improvement), at the expense of 0.0035 RMS error in the gradient value. The reduced number of time steps benefits the memory footprint. Therefore, this study shows that the scaling-and-squaring method can boost the speed performance for large deformations.

\section{Conclusions}
\label{sec:conclusions_2}

In this chapter a novel closed-form expression is presented for the gradient of CPA-based one-dimensional diffeomorphic transformations, providing efficiency, speed and precision.  
We present Diffeomorphic Fast Warping (\textit{DIFW}), an open-source library and highly optimized implementation of 1D diffeomorphic transformations on multiple backends for CPU (\textit{NumPy} and \textit{PyTorch} with \textit{C++}) and GPU (\textit{PyTorch} with \textit{CUDA}). 
Speed tests show a \textbf{x18} and \textbf{x10} improvement on CPU over \textit{libcpab} for forward and backward operations respectively. On GPU the performance gain of \textit{DIFW} reaches \textbf{x260} and \textbf{x30}.

The applications of efficient one-dimensional closed-form diffeomorphic transformations are multiple and can be a useful tool for modeling and analyzing real-world problems: 
For example, they can be used as transfer functions for a process called \textbf{image tone mapping}, which adjusts the brightness and contrast levels in a perceptually uniform manner. The transfer function monotonically maps the original and the modified intensity values, preserving  the image's contrast while adjusting the brightness levels. 
Another application is in \textbf{probability theory}, as the cumulative distribution function of a random variable can be modelled as a monotonically increasing function.
Related to this, they can be used as coupling functions for \textbf{normalizing flows} (see \cref{chapter:6}), that require a bijective one-dimensional function and the derivative of the function with respect the input variable.
Finally, they are an ideal fit for \textbf{time series alignment} methods, that call for highly expressive, differentiable and invertible warping functions which preserve temporal topology. Next \cref{chapter:3} focuses on this specific application.

\graphicspath{{content/chapter3/}}

\chapter{Time Series Elastic Averaging based on Diffeomorphic Warping Functions}\label{chapter:3}
\begingroup
\hypertarget{chapter3}{}
\hypersetup{linkcolor=black}
\setstretch{1.0}
\minitoc
\endgroup

\clearpage
\section{Introduction}\label{sec:introduction_3}

Time series data recurrently appears misaligned or warped in time despite exhibiting amplitude and shape similarity. 
As illustrated in \cref{sec:shape_lock_step}, rigid similarity metrics such as the Euclidean distance do not capture existing temporal variabilities in time series data, and as a result are not the optimal choice for time series averaging as well.

Temporal misalignment, often caused by differences in execution, sampling rates, or the number of measurements, confounds statistical analysis to the point where even the sample mean of semantically-close time series can obfuscate actual peaks and valleys, generate non-existent features and be rendered almost meaningless. 
Indeed, time warping (linear or non-linear) poses serious threats for time series analysis methods and ignoring such temporal variability can greatly decrease recognition and classification performance \cite{veeraraghavan2009rate}. 

\begin{figure}[!htb]
    \begin{center}
        \begin{tikzpicture}[scale=1]
            \node[draw=none,fill=none] (original) at (0,0){\includegraphics[width=10em]{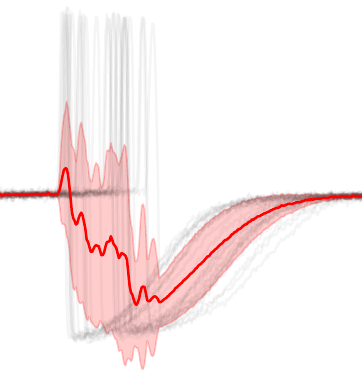}};
            \node[draw=none,fill=none] (aligned) at (10,0){\includegraphics[width=10em]{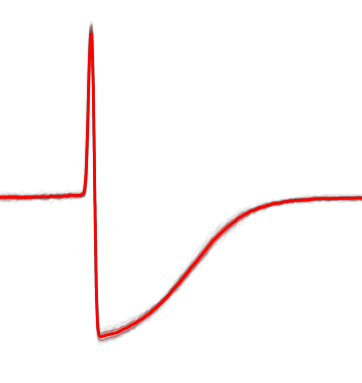}};
            
            \node[text width=10em, text centered] (A1) at (original.north) {Original Data};
            \node[text width=10em, text centered] (A2) at (original.south) {$\{\mathbf{y}_{i}\}_{i=1}^{N}$};
            \node[text width=10em, text centered] (B1) at (aligned.north) {Aligned Data};
            \node[text width=10em, text centered] (B2) at (aligned.south) {$\{\mathbf{y}_i \circ \boldsymbol{\phi}_i \}_{i=1}^{N}$};
    
            \draw[draw=black!50, fill=none, style=densely dotted, rounded corners] (A2.south west) rectangle (A1.north east);
            \draw[draw=black!50, fill=none, style=densely dotted, rounded corners] (B2.south west) rectangle (B1.north east);
            
            \draw[-latex, color=black, double] (original.east) -- node[anchor=south] {Time Series Alignment} (aligned.west);
            \draw[-latex, color=black, double] (original.east) -- node[anchor=north] {$\min \sum_{i=1}^{N} \mathcal{D}(\bar{\mathbf{y}}, \mathbf{y}_i \circ \boldsymbol{\phi}_i)$} (aligned.west);
        \end{tikzpicture}
    \caption{Time series alignment: the goal is to find a set of optimal warping functions $\{\phi_{i}^{*}\}_{i=1}^N$ such that the average sequence, $\bar{\mathbf{y}}=\argmin_{\mathbf{y}} \sum_{i=1}^{N} \mathcal{D}(\mathbf{y},\mathbf{y}_i)$ minimizes the sum of distance to all elements in the set.}
    \label{fig:alignment}
    \end{center}
  \end{figure}
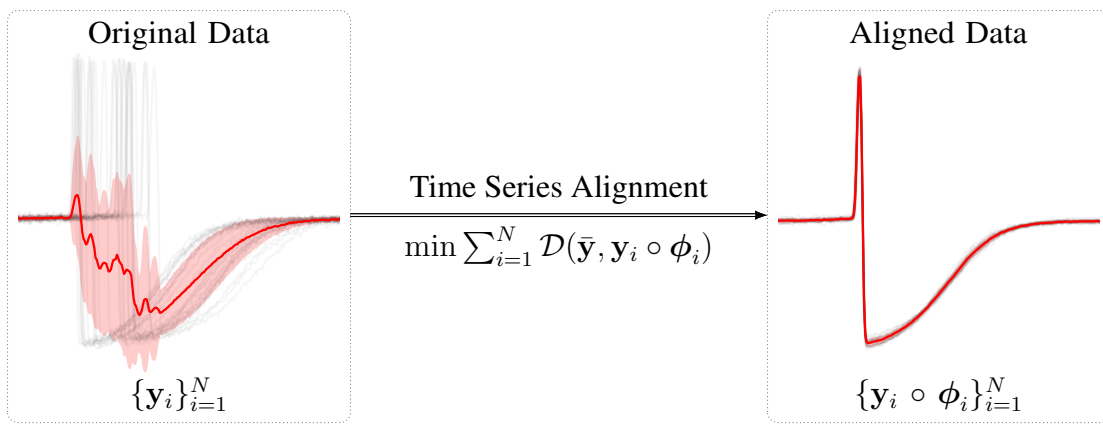

Therefore, elastic averaging a set of time series requires a simultaneous alignment that gets rid of undesired warping. Time series alignment can be solved by finding a set of optimal and plausible warping functions $\boldsymbol{\phi}$ between observations of the same process so that their temporal variability is minimized. 
Alignment methods have become critical for many applications involving time series data, such as bioinformatics, computer vision, industrial data, speech recognition \& synthesis, or human action recognition \cite{srivastava2016functional}.
Temporal warping functions extend from low dimensional global transformations described by a few parameters (e.g. affine transformations) to high dimensional non-rigid transformations specified at each point of the domain (e.g. diffeomorphisms).
In fact, temporal alignment problems have been widely stated as the estimation of diffeomorphic transformations between input and output data \cite{Huang2021}. 
Finding the optimal warping function for pairwise or joint alignment requires solving an optimization problem. 
However, aligning a batch of time series data is insufficient because new optimization problems arise as new data batches arrive. 
Inspired by Spatial Transformer Networks (STN), Temporal Transformer Networks (TTN) such as \cite{Weber2019,Lohit2019,Nunez2020} and recently \cite{Huang2021}, generalize inferred alignments from the original batch to the new data without having to solve a new optimization problem each time. 
Temporal transformer's neural network training proceeds via gradient descent on model parameters. 

In this chapter, we incorporate closed-form diffeomorphic transformations presented in \cref{chapter:2} into a temporal transformer network resembling \cite{Weber2019} for time series alignment and averaging.
We conduct extensive experiments on 84 datasets from the UCR archive \cite{dau2019ucr} to validate the generalization ability of our model to unseen data for time series joint alignment. Results show significant improvements both in terms of efficiency and accuracy (see \cref{fig:ucr:accuracy}).

\section{Related Work}
\label{sec:related_work_3}

\subsection{Pairwise Alignment}\label{sec:pairwise_alignment}

Given two time series observations $\mathbf{y}$ and $\mathbf{z}$: $\Omega \rightarrow \mathbb{R}^d$, where $\Omega \subseteq \mathbb{R}$ is the temporal domain, the goal of pairwise alignment is to compute an optimal transformation $\boldsymbol{\phi}^{*}: \Omega \rightarrow \Omega$, such than the data fit between the query $\mathbf{y}$ and transformed target $\mathbf{z} \circ \boldsymbol{\phi}$ is high, that is, their difference according to a distance measure $\mathcal{D}$ is minimum\footnote{The $\circ$ operator refers to function composition\vspace{-14pt}}.
This problem is typically solved as an optimization problem. The “goodness” of the transformation is measured by a cost of the form:
\begin{equation}
\boldsymbol{\phi}^{*} = \argmin_{\boldsymbol{\phi} \in \Phi} \mathcal{D}(\mathbf{y}, \mathbf{z} \circ \boldsymbol{\phi}) + \mathcal{R}(\boldsymbol{\phi})    
\end{equation}
The set $\Phi$ is the set of all monotonically-increasing functions from $\Omega$ to itself. 
The first term $\mathcal{D}(\mathbf{y}, \mathbf{z} \circ \boldsymbol{\phi})$ evaluates the similarity between $\mathbf{y}$ and $\mathbf{z} \circ \boldsymbol{\phi}$, whereas the second term $\mathcal{R}(\boldsymbol{\phi})$ imposes constraints on the warping function $\boldsymbol{\phi}$, as smoothness, monotonicity preserving and boundary conditions.

Elastic similarity measures such as the ones introduced in \cref{sec:similarity_metrics} can be used to align two time series. Dynamic Time Warping (DTW) \cite{sakoe1978dynamic} is a popular method that finds the optimal alignment by first computing a pairwise distance matrix and then solving a dynamic program using Bellman's recursion with a quadratic cost. However, DTW is not differentiable everywhere, is sensitive to noise and leads to bad local optima when used as a loss \cite{blondel2021differentiable}. SoftDTW \cite{cuturi2017soft} is a differentiable loss function that replaces the minimum over alignments in DTW with a soft minimum, which induces a probability distribution over alignments.

\subsection{Joint Alignment \& Averaging}\label{sec:joint_alignment}

Given a set of $N$ time series observations $\mathcal{Y}=\{\mathbf{y}_i\}_{i=1}^N$ the goal of joint alignment is to find a set of optimal warping functions $\{\phi_{i}^{*}\}_{i=1}^N$ such that the average sequence, 
\begin{equation}
\bar{\mathbf{y}}=\argmin_{\mathbf{y}} \sum_{i=1}^{N} \mathcal{D}(\mathbf{y},\mathbf{y}_i)
\end{equation}
minimizes the sum of distance to all elements in the set $\mathcal{Y}$: 
\begin{equation}
\{\boldsymbol{\phi}_{i}^{*}\}_{i=1}^{N} = \argmin_{\boldsymbol{\phi}_i \in \Phi} \sum_{i=1}^{N} \mathcal{D}(\bar{\mathbf{y}}, \mathbf{y}_i \circ \boldsymbol{\phi}_i) + \mathcal{R}(\boldsymbol{\phi}_i)
\end{equation}
The presented formulation poses time series averaging and the equivalent joint alignment problem as an optimization problem.
The computation of the average sequence or, the centroid, is a difficult task and it critically depends on the distance measure $\mathcal{D}$ used to compare time series. For instance, with the Euclidean distance the arithmetic mean is used to compute an average sequence.
When distance $\mathcal{D}$ is reduced to DTW, the sample mean in DTW space is any time series that minimizes the so called Fréchet function $F(\mathbf{y}) = \frac{1}{N} \sum_{i=1}^{N} \mathcal{D}(\mathbf{y},\mathbf{y}_i)$. 

Unfortunately, computing the sample mean of a set of time series under elastic measures like dynamic time warping is \textit{NP}-hard and a polynomial-time algorithm for finding a global minimum of the non-differentiable, non-convex Fréchet function is unknown. Therefore, an ongoing research problem is to devise improved approximation algorithms and heuristics that efficiently find acceptable sub-optimal solutions of the Fréchet function.
Time series averaging methods can be classified by two dimensions: symmetry and batch size.

\paragraph{Asymmetric vs. Symmetric Averages}
The asymmetric-symmetric dimension defines the form of an average. 
An asymmetric average of $\mathbf{y}$ and $\mathbf{z}$ is computed as follows: 
First, select a reference time series, say $\mathbf{y}$, and compute an optimal warping function $\phi$ between $\mathbf{y}$ and $\mathbf{z}$. Second, average $\mathbf{y}$ and $\mathbf{z}$ along the warping function $\phi$ with respect to the time axis of reference $\mathbf{y}$.
The form and length of an asymmetric average depends on the choice of the reference time series.

A symmetric average has no reference time series, and consists of averaging the matching elements between $\mathbf{y}$ and $\mathbf{z}$ after aligning them. However, the optimal warping function (or path) has generally more points and a finer sampling rate than the original time series $\mathbf{y}$ and $\mathbf{z}$. Therefore, a symmetric average requires methods for adjusting the sampling rate of the average to that of $\mathbf{y}$ and $\mathbf{z}$ if one wants to avoid increasingly long time series averages.

\paragraph{Batch vs. Incremental Averaging}
The batch-incremental dimension describes the strategy for combining more than two time series to an average.
Batch averaging first warps all sample time series into an appropriate form and then averages the warped time series. Incremental methods synthesize several sample time series to an average time series by pairwise averaging and repeatedly apply the following steps: First, select a pair of time series $\mathbf{y}$ and $\mathbf{z}$ from the batch; Second, compute their average and include it into the sample.

Asymmetric-batch methods first chooses an initial time series $\mathbf{y}$ as reference. Then the following steps are repeated until termination: warp all time series onto the time axis of the reference $\mathbf{y}$; and assign the average of the warped time series as new reference $\mathbf{y}$. By construction, the length of the reference z is identical in every iteration. Consequently, the final solution (reference) of the asymmetric-batch method depends on the choice of the initial solution.

Symmetric-batch methods require to find an optimal warping in an $N$-dimensional hypercube, where $N$ is the sample size. Since finding an optimal common warping path for $N$ sample time series is computationally intractable, symmetric-batch methods have not been further explored.

\subsubsection{Current Strategies for Time Series Averaging}
In this section we review available solutions to compute time series averaging.

NonLinear Alignment and Averaging Filters (\textbf{NLAAF}) \cite{gupta1996nonlinear} uses a simple pairwise method where each coordinate of the average sequence is calculated as the center of the mapping produced by DTW. This method is applied sequentially to pairs of sequences until only one pair is left.
In each iteration of NLAAF, time series are randomly grouped into a set of pairs and then each pair is fused using the DTW averaging. This procedure is repeated until the centroid time series is achieved. Although NLAAF is practical for averaging multiple sequences, it has two problems: the centroid is sensitive to the order of pairing time series, and there is no feedback in the algorithm to reduce the errors that occur at early steps.
Prioritized Shape Averaging (\textbf{PSA}) \cite{niennattrakul2009shape} uses hierarchical clustering to automatically create a reasonable order for pairing time series, and in this way is robust to outlier sequences.
The coordinates of an average sequence are computed as the weighted center of the coordinates of two time series sequences that were coupled by DTW. Initially, all sequences have weight one, and each average sequence produced in the nodes of the tree has a weight that corresponds to the number of sequences it averages.
Several drawbacks of these two methods have led to the creation of a more robust techniques.

DTW Barycenter Averaging (\textbf{DBA}) \cite{Petitjean2011-DBA} minimizes the sum of the DTW discrepancies to the input time series and iteratively refines an initial average sequence. Each coordinate of the average sequence is updated with the use of barycenter of one or more coordinates of the other sequences that were associated with the use of DTW.
However, DBA is sensitive to the quality of the initial average signal. When the initial average is considerably different from the best solution, DBA is highly likely to converge to a weak local minimum.
Generalized Time Warping (\textbf{GTW}) \cite{Zhou2012} approximates the optimal temporal warping by a linear combination of monotonic basis functions and a Gauss-Newton-based method is used to learn the weights of the basis functions.
However, GTW requires a large number of complex basis functions to be effective and defining these basis functions is very difficult. 

Time Elastic Kernel Averaging (\textbf{TEKA}) \cite{marteau2019times} expresses the averaging process in terms of stochastic alignment automata. It uses an iterative agglomerative heuristic method for averaging the aligned samples, while also averaging the times of their occurrence. This algorithm exhibits quite interesting denoising capabilities. 
Neural Time Warping (\textbf{NTW}) \cite{kawano2020neural} relaxes the DTW-based discrete formulation of the joint alignment problem to a continuous optimization in which a neural network learns the optimal warping functions.
Also related is the square-root velocity function (\textbf{SRVF}) representation \cite{Srivastava2011} for analyzing shapes of curves in euclidean spaces under an elastic metric that is invariant to reparametrization. 
Note that these solutions are cast as an optimization problem, and as a result, lack a generalization mechanism and must compute alignments from scratch for new data. 

With regard to other relevant methods, 
\cite{Cohen2020} proposes a differentiable distance between time series on potentially incomparable spaces,
\cite{Abid2018} defines a flexible and differentiable family of warping metrics such as DTW, Euclidean, and edit distance; and uses sequence autoencoders to optimize for a member of this warping distance family. 
Also related, \cite{Grabocka2018} models a warping function as the upper level neural network between deeply-encoded time series values and
\cite{Koneripalli2020} explicitly disentangles temporal parameters in the form of order-preserving diffeomorphisms with respect to a learnable template.
Finally, in \cite{Kaufman2021} CPA based diffeomorphic transformations are applied to cyclic, closed shapes.

\subsection{Warping Functions}

In the context of time series alignment, a warping function is a transformation that aligns two series by mapping the time indices of one series to the time indices of another series. It should be noted that the warping function must be strictly monotone with a strictly positive first derivative. This section discusses the two main types of warping functions: non-parametric and parametric.

Non-parametric warping functions do not have a predetermined mathematical form and depend on the data dimensionality, such as the paths derived from DTW, Soft-DTW and related elastic metrics. These functions are typically more flexible than parametric warping functions, can capture complex relationships between the time indices, but are more difficult to work with due to the variable number of parameters and the computational resources required to compute them.

Parametric warping functions, on the other hand, have a fixed number of parameters that can be learned from the data. Examples of parametric warping functions include linear, polynomial, and spline functions, which are typically easier to work with due to their well-defined form and fixed number of parameters. However, they may not be as flexible as non-parametric functions in capturing complex relationships between the time indices.

Parametric warping functions may be preferred in situations where computational efficiency is a concern, while non-parametric warping functions may be chosen when greater flexibility is needed.
One approach to compare the performance of parametric and non-parametric warping functions is through the use of simulated data. By generating time series with known alignments, one can evaluate the accuracy and computational efficiency of different warping functions in recovering the true alignment. However, this approach is not optimal since it is biased to the type of warping functions applied in first instance to the data.

Furthermore, different similarity measures can result in divergent warping functions and average results, making it difficult to compare the performance of different algorithms or to evaluate the quality of the warping function. 
Given this lack of ground truth for the latent warps in real data, in this chapter we use a simple classification model like nearest centroid as a proxy metric for the quality of the joint alignment and the average signal. This can provide insights into the practical effectiveness of different warping functions in different contexts.

\paragraph{Differentiability}

When using gradient-based optimization techniques, a critical requirement for temporal transformations is differentiability. 
Gradient-based optimization is a calculus-based point-by-point technique that relies on the gradient (derivative) information of the objective function with respect to a number of independent variables. 
ML and DL algorithms that use gradient-based optimization are modeled as convex optimization problems over the loss function, even though the functions themselves may be non-convex. Hence, the optimization method can only work when all functions are differentiable, and thus differentiability is an important property. 

Regarding warping functions, it is essential that the parametrized transformation $T^{\theta}(x): \mathbb{R} \rightarrow \mathbb{R}$ is differentiable with respect to both the temporal dimension $x$ and the parameters $\theta$. Differentiability also implies that drastic kinks or corners in the transformed time series are avoided.
Being differentiable does not imply that the computation of such gradients is feasible and easy to compute. In the case of diffeomorphic functions obtained from the integration of continuous piecewise-affine velocity fields a ODE solver is required.
Among the available methods to compute derivatives to the ODE's solution (presented in \cref{sec:related_work_2}), we follow the first principle of automatic differentiation and use the closed-form derivative provided in \cref{sec:closed_form_derivatives}.

\subsection{Temporal Transformer Networks}

Spatial Transformer\footnote{It should be noted that the Spatial Transformer does not share any similarity with the Transformer model architecture \cite{vaswani2017attention}, which is at the core of most state of the art ML research, and is composed of multiple \textit{attention} layers which learn which parts of the input data are the most important for a given task. Transformers started in language modeling, then expanded into computer vision, audio, and other modalities.} Networks are a type of deep neural network architecture that allows the spatial manipulation of data within the network, and consist of three components: a localization network, a grid generator, and a sampler. The localization network takes in an input image and produces a set of transformation parameters, which are then used by the grid generator to produce a grid of sampling points. The sampler uses this grid to sample the input image and produce the output image.
One of the main benefits of STNs is that they allow a neural network to learn to align or transform its input data, rather than having to rely on pre-processing or hand-designed image transformations. This makes them useful for tasks such as image translation, rotation, and scaling. They have also been applied to a variety of other tasks, including object detection and segmentation.

Inspired by Spatial Transformer Networks (STN), \textbf{Temporal Transformer Networks} (TTN) such as \cite{Weber2019,Lohit2019,Nunez2020} and recently \cite{Huang2021}, generalize inferred alignments from the original batch to the new data without having to solve a new optimization problem each time. 
TTNs are designed to be differentiable, meaning that they can be trained using gradient descent and backpropagation on model parameters.

Diffeomorphisms have been actively studied in the field of image registration and have been successfully applied in a variety of methods \cite{Beg2005,vercauteren2009diffeomorphic,dalca2018unsupervised,fu2020deep}, 
such as the log-Euclidean polyaffine method \cite{Arsigny2006, Arsigny2006a}.
Hereof \cite{Detlefsen2018} first implemented flexible CPA-based diffeomorphic image transformations within STN, and were later extended to variational autoencoders \cite{Detlefsen2019}. 
On this subject, \cite{Ouderaa2021} presented an STN with diffeomorphic transformations and used the scaling-and-squaring method for solving 2D stationary velocity fields and the Baker-Campbell-Hausdoorff formula for solving time-dependent velocity fields. 

The counterpart STN model for time series alignment is \textbf{Diffeomorphic Temporal Alignment Nets} (DTAN) \cite{Weber2019} and its recurrent variant R-DTAN. DTAN closely resembles the TTN model presented in this chapter, even though the core diffeomorphic transformations are computed differently, as presented in \cref{sec:method_2}.
In this regard, \cite{Lohit2019} generated input-dependent warping functions that lead to rate-robust representations, reduce intra-class variability and increase inter-class separation.

Also related is \cite{Rousseau2019}, a residual network for the numerical approximation of exponential diffeomorphic operators. The velocity field in this model is a linear combination of basis functions which are parametrized with convolutional and \textit{ReLu} layers.
Recently, \cite{Huang2021} proposed a deep residual network for time warping (\textbf{ResNet-TW}) that echoes an Eulerian discretization of the flow equation (ODE) for CPA time-dependent vector fields to build diffeomorphic transformations. Inspired by the elegant Large Deformation Diffeomorphic Metric Mapping (LDDMM) framework \cite{Beg2005}, ResNet-TW tackles the alignment problem by compositing a flow of incremental diffeomorphic mappings.
As an alternative approach, \cite{Nunez2020} proposed a deep neural network for learning the warping functions directly from DTW matches, and used it to predict optimal diffeomorphic warping functions.
Regarding the SRVF framework, \cite{Nunez2021} presented an unsupervised generative encoder-decoder architecture (\textbf{SRVF-Net}) to produce a distribution space of SRVF warping functions for the joint alignment of functional data.
Beyond diffeomorphic transformations, \textbf{Trainable Time Warping} (TTW) \cite{Khorram2019} performs alignment in the continuous-time domain using a \textit{sinc} convolutional kernel and a gradient-based optimization technique. 
TTW first aligns the input sequences and then takes an average over the synchronized sequences to calculate the centroid signal. Its complexity is linear in both the number and the length of time series.

\section{Temporal Transformer Network (TTN) Model}
\label{sec:method_3}

\begin{figure}[p]
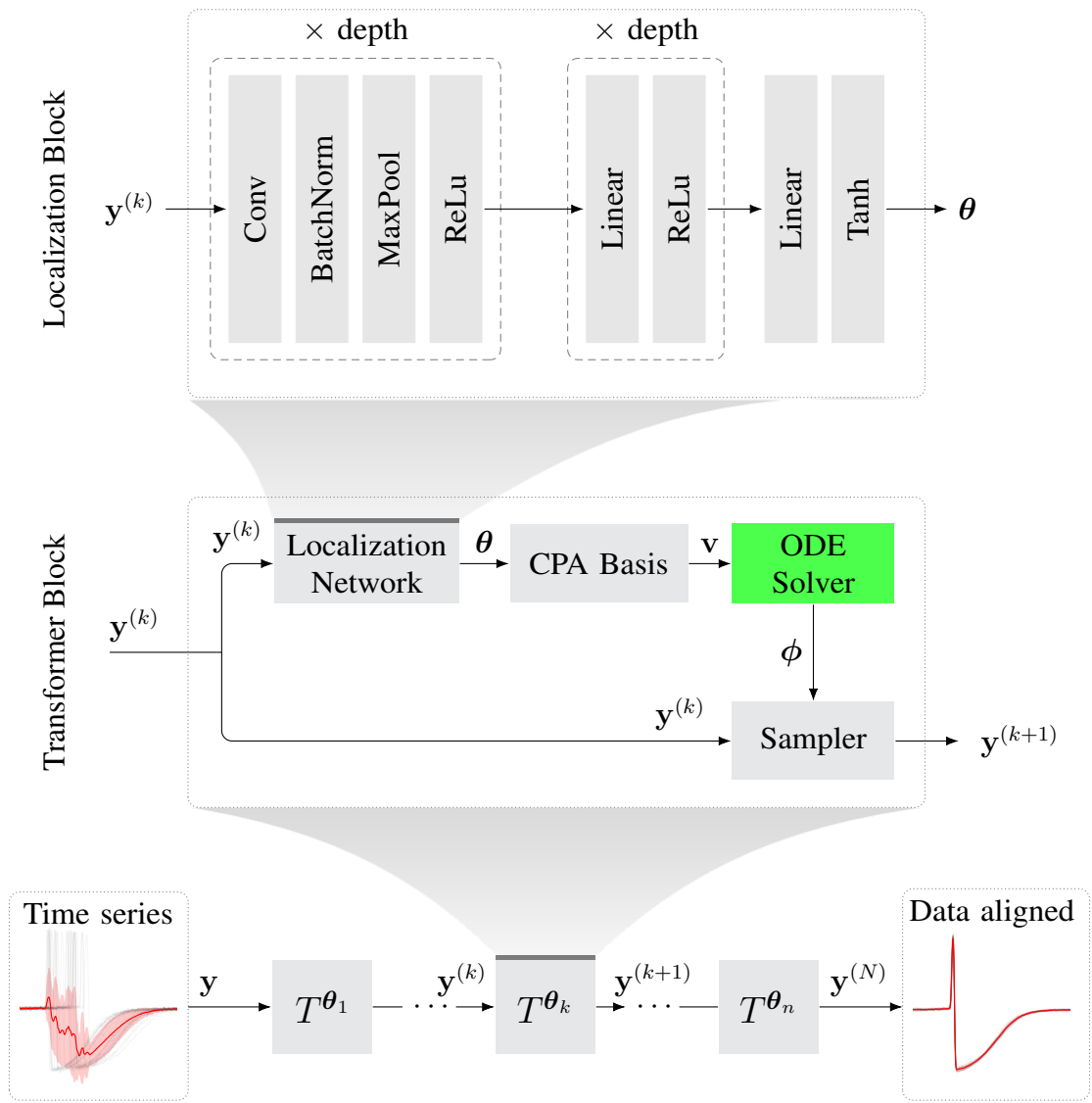

    \begin{center}
        \begin{tikzpicture}[scale=1.2]
            \draw[draw=white, fill, top color=black!15, bottom color=white] (5,0) to[bend right=20] (1, 2.25) -- (9.25, 2.25) to[bend right=20] cycle;
            \draw[draw=white, fill, top color=black!15, bottom color=white] (3,4.75) -- (1.95,5.5) to[bend right=20] (1, 6.85) -- (9.25, 6.85) to[bend right=18] (4.05,5.5) -- cycle;

            \node[draw=none, text centered, rotate=90] at (-0.5,4) {Transformer Block};
            \node[draw=none, text centered, rotate=90] at (-0.5, 9) {Localization Block};
        
            \begin{scope}[shift={(0,0)}]
                \tikzstyle{r}=[draw=none, fill=Black!10, text width=2.5em, minimum height=3em, text centered, thick, rounded corners=0]
                
                \node[draw=none,fill=none] (original) at (0,0){\includegraphics[width=5em]{figures/original.pdf}};
                \node[draw=none,fill=none] (aligned) at (10,0){\includegraphics[width=5em]{figures/aligned.pdf}};
                
                \node[text width=5em, text centered] (A) at (original.north) {Time series};
                \node[text width=5.5em, text centered] (B) at (aligned.north) {Data aligned};
        
                \draw[draw=black!50, fill=none, style=densely dotted, rounded corners] (original.south west) rectangle (A.north east);
                \draw[draw=black!50, fill=none, style=densely dotted, rounded corners] (aligned.south west) rectangle (B.north east);
        
                \node[r] (C) at (2.5,0) {\large $T^{\boldsymbol{\theta}_1}$};
                \node[r, drop shadow={fill=black, shadow xshift=0pt, shadow yshift=2pt}] (D) at (5,0) {\large $T^{\boldsymbol{\theta}_k}$};
                \node[r] (E) at (7.5,0) {\large $T^{\boldsymbol{\theta}_n}$};
                
                \draw[-Latex, color=black] (original.east) -- node[anchor=south east] {$\mathbf{y}$} (C.west);
                \draw[color=black] (C.east) -- (3.4,0);
                \node[] at (3.75,0) {$\cdots$};
                \draw[-Latex, color=black] (4,0) node[anchor=south west, xshift=-0.4cm] {$\mathbf{y}^{(k)}$} -- (D.west);
        
                \draw[-Latex, color=black] (D.east) -- node[anchor=south west, xshift=-0.1cm] {$\mathbf{y}^{(k+1)}$} (5.9,0);
                \draw[color=black] (6.5, 0) -- (E.west);
                \node[] at (6.25,0) {$\cdots$};
                \draw[-Latex, color=black] (E.east) node[anchor=south west] {$\mathbf{y}^{(N)}$} -- (aligned.west);
            \end{scope}
        
            \begin{scope}[shift={(3,5)}]
                \tikzstyle{r}=[text width=4em, minimum height=2.5em, text centered, thick]
                \node[r, draw=none, fill=Black!10, text=black, text width=5.2em, drop shadow={fill=black, shadow xshift=0pt, shadow yshift=2pt}] (A) at (0,0) {Localization Network};
                \node[r, draw=none, fill=Black!10, text=black, text width=5em] (B) at (2.6,0) {CPA Basis};
                \node[r, draw=none, fill=green!70, text=black, text width=4.5em] (C) at (5,0) {ODE Solver};
                \node[r, draw=none, fill=Black!10, text=black, text width=4.5em] (D) at (5,-2) {Sampler};
                \node[] (E) at (-1.5, -1) {};
                \node[] (F) at (-3, -1) {};
                \node[] (G) at (6.75, -2) {};
        
                \draw[] (F) node[anchor=south west] {$\mathbf{y}^{(k)}$} -- (E);
                \draw[-Latex, color=black, rounded corners] (E.west) |- (A.west) node[anchor=south east] {$\mathbf{y}^{(k)}$};
                \draw[-Latex, color=black] (A.east) -- (B.west) node[midway, anchor=south] {$\boldsymbol{\theta}$};
                \draw[-Latex, color=black] (B.east) -- (C.west) node[midway, anchor=south] {$\mathbf{v}$};
                \draw[-Latex, color=black] (C.south) -- (D.north) node[midway, anchor=east] {$\boldsymbol{\phi}$};
                \draw[-Latex, color=black, rounded corners] (E.west) |- (D.west) node[anchor=south east, xshift=-0.2cm] {$\mathbf{y}^{(k)}$};
                \draw[-Latex, color=black] (D.east) -- (G) node[anchor=west] {$\mathbf{y}^{(k+1)}$};
                \draw[draw=black!50, fill=none, style=densely dotted, rounded corners] (-2, -2.75) rectangle (6.25,0.75);
            \end{scope}

            \begin{scope}[scale=0.5, shift={(17,18)}, rotate=90, nodes={rotate=90}]
                \begin{scope}[shift={(0,9)}, nodes={draw, text width=8em, align=center, thick}]
                    \node[draw=RoyalBlue!0,   fill=black!10, text=black, minimum height=1.75em] (D3) at (0,4.5) {Conv};
                    \node[draw=BurntOrange!0, fill=black!10, text=black, minimum height=1.75em] (C3) at (0,3) {BatchNorm};
                    \node[draw=ForestGreen!0, fill=black!10, text=black, minimum height=1.75em] (B3) at (0,1.5) {MaxPool};
                    \node[draw=Violet!0,      fill=black!10, text=black, minimum height=1.75em] (A3) at (0,0) {ReLu};
                    \draw[draw=black!50, fill=none, style=densely dashed, rounded corners] (-3.4,-1) rectangle (3.4,5.5);
                    \node[draw=none, rotate=-90] at (4,2.25) {$\times$ depth};
                \end{scope}
        
                \begin{scope}[shift={(0,4)}, nodes={draw, text width=8em, align=center, thick}]
                    \node[draw=Red!0,    fill=black!10, text=black, minimum height=1.75em] (B2) at (0,1.5) {Linear};
                    \node[draw=Violet!0, fill=black!10, text=black, minimum height=1.75em] (A2) at (0,0) {ReLu};
                    \draw[draw=black!50, fill=none, style=densely dashed, rounded corners] (-3.4,-1) rectangle (3.4,2.5);
                    \node[draw=none, rotate=-90] at (4,0.75) {$\times$ depth};
                \end{scope}
        
                \begin{scope}[shift={(0,0)}, nodes={draw, text width=8em, align=center, thick}]
                    \node[draw=Red!0, fill=black!10, text=black, minimum height=1.75em] (B1) at (0,1.5) {Linear};
                    \node[draw=RubineRed!0, fill=black!10, text=black, minimum height=1.75em] (A1) at (0,0) {Tanh};
                \end{scope}
        
                \draw[-Latex, color=black] (A3) -- (B2);
                \draw[-Latex, color=black] (A2) -- (B1);
                
                \draw[draw=black!50, fill=none, style=densely dotted, rounded corners] (-4.25, -1.5) rectangle (4.5,15);
                
                \draw[-Latex, color=black]  (0,15.5) node[anchor=east, rotate=-90] {$\mathbf{y}^{(k)}$}  -- (D3);
                \draw[-Latex, color=black] (A1) -- (0,-2) node[anchor=west, rotate=-90] {$\boldsymbol{\theta}$};
        
            \end{scope}
        
        \end{tikzpicture}    
    \caption{Proposed temporal transformer architecture. \textbf{Bottom}: time series $\mathbf{y}$ is aligned by applying a sequence of transformers $T^{\boldsymbol{\theta}}$ that minimize the empirical variance of the warped signals. \textbf{Middle}: each transformer block resembles the STN proposed by \cite{Jaderberg2015}. 
    Transformed data is sampled based on a diffeomorphic flow $\boldsymbol{\phi}$ obtained from the integration of a velocity function $\mathbf{v}$ from a first order ordinary differential equation (ODE). \textbf{Top}: The parameters $\boldsymbol{\theta}$ of the velocity function $\mathbf{v}$ are computed by the localization network based on each signal $\mathbf{y}^{(k)}$.
    }
    \label{fig:architecture}
    \end{center}
\end{figure}

The Temporal Transformer Network (TTN) is a end-to-end trainable module designed to capture temporal warping and magnitude invariances, and it can be easily added at the beginning of a time series classifier. Its function is to warp the input sequences so as to maximize the classification performance. \cite{Lohit2019}
The proposed TTN model resembles the Spatial Transformer Network proposed by \cite{Jaderberg2015}, later adapted for time series alignment by \cite{oh2018learning,Lohit2019,Weber2019} (see \cref{fig:architecture}), and can recurrently apply nonlinear time warps to the input signal. The model consists of four main components: a localization network, a CPA basis generator, ODE solver and a sampler. The localization network is a neural network that takes the input data $\mathbf{y}$ and produces a set of transformation parameters $\boldsymbol{\theta}$. The CPA basis generator then creates a set of velocity coordinates $\mathbf{v}$ that define a regular grid over the input data. Then, the ODE solver (which is also a differentiable module) integrates the velocity function and yields the diffeormorphic warping function $\boldsymbol{\phi}$. Finally, the sampler uses the warping function $\boldsymbol{\phi}$ to sample the input data  $\mathbf{y}$ at the transformed coordinates.

Composing warps increases the expressiveness without refining $\Omega$ as it implies non-stationary velocity functions which are CPA in $\Omega$ and piecewise constant in time. Note that unlike \cite{Weber2019}, in the proposed TTN model the localization network parameters are not shared across layers. 
The differentiable sampler uses piecewise linear interpolation to estimate the warped signals. We refer the reader to \cref{sec:linear_interpolation_grid} for details about the sampler estimation and its derivatives.

\subsection{Loss Function}
Let $\mathbf{y}_i$ denote an input signal among $N$ time series samples, and $\boldsymbol{\theta}_i = F_{loc}(\mathbf{w}, \mathbf{y}_i)$ denote the corresponding output of the localization network $F_{loc}(\mathbf{w},\cdot)$, and let 
$\mathbf{z}_i = \mathbf{y}_i \circ \boldsymbol{\phi}_{\boldsymbol{\theta}_i}$ 
denote the result of warping $\mathbf{y}_i$ by $ \boldsymbol{\phi}_{\boldsymbol{\theta}_i}$, where $\boldsymbol{\theta}_i$ depends on $\mathbf{w}$ and $\mathbf{y}_i$. The variance of the observed $(\mathbf{y}_i)_{i=1}^{N}$ is partially explained by the latent warps $(\mathbf{y}_i)_{i=1}^{N}$, so we seek to minimize the empirical variance of the warped signals:

\begin{equation}\label{eq:loss_data_single}
\mathcal{L}_{data}^{1} (\mathbf{y}_i|_{i=1}^{N}) =
\frac{1}{N} \sum_{i=1}^{N} 
\Big\Vert 
\mathbf{y}_i \circ \boldsymbol{\phi}_i - \frac{1}{N} \sum_{j=1}^{N} \mathbf{y}_j \circ \boldsymbol{\phi}_j
\Big\Vert_{2}^2
\end{equation}
where $\Vert\cdot \Vert_{2}$ is the $l_2$ norm. For multi-class problems, the expression is the sum of within-class variances:
\begin{equation}\label{eq:loss_data_multi}
\mathcal{L}_{data}^{K} (\mathbf{y}_i|_{i=1}^{N})= 
\sum_{k=1}^{K}
\frac{1}{N_k} 
\mathcal{L}_{data}^{1} (\mathbf{y}_{r_{i}=k})
\end{equation}
where $K$ is the number of classes, $r_i$ takes values in $\{1,\cdots,K\}$ and is the class label associated with $\mathbf{y}_i$ and $N_k$ is the number of examples in class $k$.
Hence, we formulate the joint alignment problem as to simultaneously compute the centroid and align all sequential data within a class, under a semi-supervised schema: A single net learns how to perform joint alignment within each class without knowing the class labels at test time. Thus, while the within-class alignment remains unsupervised (as in the single-class case), for multi-class problems labels are used during training (not on testing) to reduce the variance within each class separately.
Both single- and multi-class cases use a regularization term for the warps:
\begin{equation}\label{eq:loss_reg}
\mathcal{L}_{reg}(\mathbf{y}_i|_{i=1}^{N}) = 
\frac{1}{N} \sum_{i=1}^{N}
\boldsymbol{\theta}_i^T \boldsymbol{\Sigma}_{CPA}^{-1} \boldsymbol{\theta}_i
\end{equation}
In the context of joint-alignment using regularization is critical, partly since it is too easy to minimize $\mathcal{L}_{data}$ by unrealistically-large deformations that would cause most of the inter-signal variability to concentrate on a small region of the domain \cite{Weber2019}; the regularization term prevents that. 
Therefore, our loss function, to be minimized over $\mathbf{w}$, is: 
\begin{equation}
    \mathcal{L}=\mathcal{L}_{data} + \mathcal{L}_{reg}
\end{equation}

\paragraph{Variable length \& multi-channel data} 
As in \cite{Weber2019}, generalization to multichannel signal is trivial. Variable-length signals can be managed with the presented loss function as long as a fixed number of points is set on the linear interpolation sampler. Yet, a neural network that can handle variable-length (a Recurrent Neural Network, for instance) is required for the localization network $F_{loc}$.

\paragraph{Implementation} 
The Temporal Transformer Network model was implemented in \textit{PyTorch} \cite{paszke2019pytorch} and can be integrated with a few lines of code with other deep learning architectures for time series classification or prediction. 

\section{Experiments and Results}\label{sec:results_3}

Extensive experiments are conducted to determine whether the proposed Temporal Transformer Network model can effectively joint aligning a collection of univariate and multivariate time series and generalize inferred alignments from the original batch to new data without solving a new optimization problem.

\subsection{Nearest Centroid Classification (NCC)} 

NCC first computes the centroid (average) of each class in the training set by minimizing the loss function for each class. During prediction, NCC assigns the class of the nearest centroid. 
NCC has a lower variance than nearest neighbors (NN), which can represent much more complex decision boundaries. This greater power comes at the cost of greater capacity to overfit the data. 
In the lack of ground truth for the latent warps in real data, NCC accuracy rates provide an indicative metric for the quality of the joint alignment and the average signal. 
The nearest centroid classifier does not involve any parameter, and thus the accuracy depends only on the similarity measure, which ensure fair comparisons between different similarity measures.
Thus, we perform NCC on the UCR archive, comparing our model to: (1) the sample mean of the misaligned sets (Euclidean); (2) DBA \cite{Petitjean2011-DBA}; (3) SoftDTW \cite{cuturi2017soft}, (4) DTAN \cite{Weber2019} and (5) ResNet-TW \cite{Huang2021}.

\paragraph{Experiment details}
For each of the UCR datasets, we train our TTN for joint alignment as in \cite{Weber2019}, where 
$N_{\mathcal{P}} \in \{16,32,64\}$,
$\lambda_{\sigma} \in \{10^{-3},10^{-2}\}$,
$\lambda_{s} \in \{0.1,0.5\}$,
the number of transformer layers $\in \{1,5\}$,
scaling-and-squaring iterations $\in \{0,8\}$
and the option to apply the zero-boundary constraint.
Summarized results of hyperparameters grid-search have been included on \cref{sec:hyperparameter}, and the full table of results is available in the Supplementary Material. 
The network was initialized with Xavier initialization \cite{glorot2010understanding} using a normal distribution and was trained for 500 epochs with $10^{-5}$ learning rate, a batch size of $32$ and Adam \cite{kingma2014adam} optimizer with $\beta_{1}=0.9$, $\beta_{2}=0.98$ and $\epsilon=10^{-8}$.
Regarding the tessellation size $N_{\mathcal{P}}$, an ablation study was conducted to investigate how partition fineness in CPA velocity functions controls the trade-off between expressiveness and computational complexity (see \cref{sec:ablation}).

\paragraph{Computing Infrastructure}
We used the following computing infrastructure in our experiments: 
Intel® Core™ i7-6560U CPU @2.20GHz, 4 cores, 16GB RAM with an Nvidia Tesla P100 graphic card.

\subsection{UCR Time Series Classification Archive}\label{sec:results:ucr}

The UCR \cite{dau2019ucr} time series classification archive currently contains 128 real-world datasets and we use a subset containing 84 univariate datasets, as in DTAN \cite{Weber2019} and ResNet-TW \cite{Huang2021}. Details about these datasets can be found on \cref{tab:ucr_dataset_table}. 
Experiments were conducted with the provided train and test split. 

\small
\renewcommand{\arraystretch}{0.8}
\begin{longtable}{lrrrrl}
\caption{UCR univariate datasets \cite{dau2019ucr} information}\label{tab:ucr_dataset_table}\\
\toprule
                       Dataset &  Train &  Test &  Length &  Classes &      Type \\
\midrule\endfirsthead
\caption{(Cont.) UCR univariate datasets \cite{dau2019ucr} information}\\
\toprule
                       Dataset &  Train &  Test &  Length &  Classes &      Type \\
\midrule \endhead 
                         Adiac &    390 &   391 &     176 &       37 &     Image \\
                     ArrowHead &     36 &   175 &     251 &        3 &     Image \\
                          Beef &     30 &    30 &     470 &        5 &   Spectro \\
                     BeetleFly &     20 &    20 &     512 &        2 &     Image \\
                   BirdChicken &     20 &    20 &     512 &        2 &     Image \\
                           Car &     60 &    60 &     577 &        4 &    Sensor \\
                           CBF &     30 &   900 &     128 &        3 & Simulated \\
         ChlorineConcentration &    467 &  3840 &     166 &        3 & Simulated \\
                  CinCECGTorso &     40 &  1380 &    1639 &        4 &       ECG \\
                        Coffee &     28 &    28 &     286 &        2 &   Spectro \\
                     Computers &    250 &   250 &     720 &        2 &    Device \\
                      CricketX &    390 &   390 &     300 &       12 &    Motion \\
                      CricketY &    390 &   390 &     300 &       12 &    Motion \\
                      CricketZ &    390 &   390 &     300 &       12 &    Motion \\
           DiatomSizeReduction &     16 &   306 &     345 &        4 &     Image \\
  DistalPhalanxOutlineAgeGroup &    400 &   139 &      80 &        3 &     Image \\
   DistalPhalanxOutlineCorrect &    600 &   276 &      80 &        2 &     Image \\
               DistalPhalanxTW &    400 &   139 &      80 &        6 &     Image \\
                   Earthquakes &    322 &   139 &     512 &        2 &    Sensor \\
                        ECG200 &    100 &   100 &      96 &        2 &       ECG \\
                       ECG5000 &    500 &  4500 &     140 &        5 &       ECG \\
                   ECGFiveDays &     23 &   861 &     136 &        2 &       ECG \\
               ElectricDevices &   8926 &  7711 &      96 &        7 &    Device \\
                       FaceAll &    560 &  1690 &     131 &       14 &     Image \\
                      FaceFour &     24 &    88 &     350 &        4 &     Image \\
                      FacesUCR &    200 &  2050 &     131 &       14 &     Image \\
                    FiftyWords &    450 &   455 &     270 &       50 &     Image \\
                          Fish &    175 &   175 &     463 &        7 &     Image \\
                         FordA &   3601 &  1320 &     500 &        2 &    Sensor \\
                         FordB &   3636 &   810 &     500 &        2 &    Sensor \\
                      GunPoint &     50 &   150 &     150 &        2 &    Motion \\
                           Ham &    109 &   105 &     431 &        2 &   Spectro \\
                  HandOutlines &   1000 &   370 &    2709 &        2 &     Image \\
                       Haptics &    155 &   308 &    1092 &        5 &    Motion \\
                       Herring &     64 &    64 &     512 &        2 &     Image \\
                   InlineSkate &    100 &   550 &    1882 &        7 &    Motion \\
           InsectWingbeatSound &    220 &  1980 &     256 &       11 &    Sensor \\
              ItalyPowerDemand &     67 &  1029 &      24 &        2 &    Sensor \\
        LargeKitchenAppliances &    375 &   375 &     720 &        3 &    Device \\
                    Lightning2 &     60 &    61 &     637 &        2 &    Sensor \\
                    Lightning7 &     70 &    73 &     319 &        7 &    Sensor \\
                        Mallat &     55 &  2345 &    1024 &        8 & Simulated \\
                          Meat &     60 &    60 &     448 &        3 &   Spectro \\
                 MedicalImages &    381 &   760 &      99 &       10 &     Image \\
  MiddlePhalanxOutlineAgeGroup &    400 &   154 &      80 &        3 &     Image \\
   MiddlePhalanxOutlineCorrect &    600 &   291 &      80 &        2 &     Image \\
               MiddlePhalanxTW &    399 &   154 &      80 &        6 &     Image \\
                    MoteStrain &     20 &  1252 &      84 &        2 &    Sensor \\
    NonInvasiveFetalECGThorax1 &   1800 &  1965 &     750 &       42 &       ECG \\
    NonInvasiveFetalECGThorax2 &   1800 &  1965 &     750 &       42 &       ECG \\
                      OliveOil &     30 &    30 &     570 &        4 &   Spectro \\
                       OSULeaf &    200 &   242 &     427 &        6 &     Image \\
      PhalangesOutlinesCorrect &   1800 &   858 &      80 &        2 &     Image \\
                       Phoneme &    214 &  1896 &    1024 &       39 &    Sensor \\
                         Plane &    105 &   105 &     144 &        7 &    Sensor \\
ProximalPhalanxOutlineAgeGroup &    400 &   205 &      80 &        3 &     Image \\
 ProximalPhalanxOutlineCorrect &    600 &   291 &      80 &        2 &     Image \\
             ProximalPhalanxTW &    400 &   205 &      80 &        6 &     Image \\
          RefrigerationDevices &    375 &   375 &     720 &        3 &    Device \\
                    ScreenType &    375 &   375 &     720 &        3 &    Device \\
                   ShapeletSim &     20 &   180 &     500 &        2 & Simulated \\
                     ShapesAll &    600 &   600 &     512 &       60 &     Image \\
        SmallKitchenAppliances &    375 &   375 &     720 &        3 &    Device \\
         SonyAIBORobotSurface1 &     20 &   601 &      70 &        2 &    Sensor \\
         SonyAIBORobotSurface2 &     27 &   953 &      65 &        2 &    Sensor \\
                    Strawberry &    613 &   370 &     235 &        2 &   Spectro \\
                   SwedishLeaf &    500 &   625 &     128 &       15 &     Image \\
                       Symbols &     25 &   995 &     398 &        6 &     Image \\
              SyntheticControl &    300 &   300 &      60 &        6 & Simulated \\
              ToeSegmentation1 &     40 &   228 &     277 &        2 &    Motion \\
              ToeSegmentation2 &     36 &   130 &     343 &        2 &    Motion \\
                         Trace &    100 &   100 &     275 &        4 &    Sensor \\
                    TwoLeadECG &     23 &  1139 &      82 &        2 &       ECG \\
                   TwoPatterns &   1000 &  4000 &     128 &        4 & Simulated \\
        UWaveGestureLibraryAll &    896 &  3582 &     945 &        8 &    Motion \\
          UWaveGestureLibraryX &    896 &  3582 &     315 &        8 &    Motion \\
          UWaveGestureLibraryY &    896 &  3582 &     315 &        8 &    Motion \\
          UWaveGestureLibraryZ &    896 &  3582 &     315 &        8 &    Motion \\
                         Wafer &   1000 &  6164 &     152 &        2 &    Sensor \\
                          Wine &     57 &    54 &     234 &        2 &   Spectro \\
                  WordSynonyms &    267 &   638 &     270 &       25 &     Image \\
                         Worms &    181 &    77 &     900 &        5 &    Motion \\
                 WormsTwoClass &    181 &    77 &     900 &        2 &    Motion \\
                          Yoga &    300 &  3000 &     426 &        2 &     Image \\
\bottomrule
\end{longtable}
\renewcommand{\arraystretch}{1}
\normalsize

\subsection{NCC Accuracy Results}

\begin{figure}[!htb]
    \begin{center}
    \includegraphics[width=0.75\linewidth]{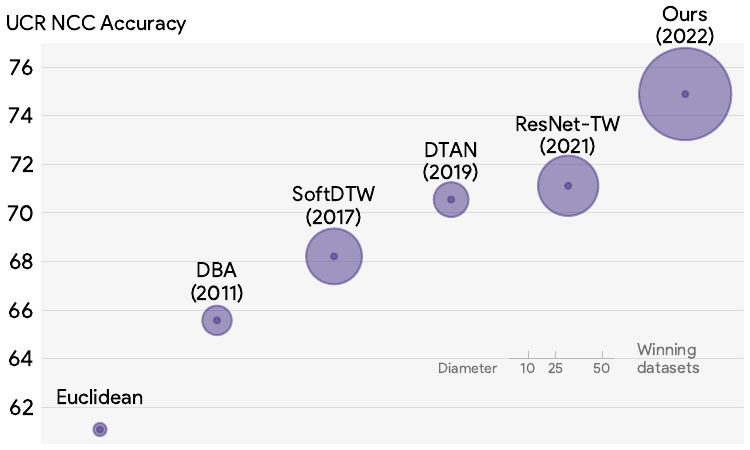}
    \caption{
    Correct classification rates using NCC on UCR archive \cite{dau2019ucr}. Radius denotes the number of datasets in which each method achieved top accuracy. Our method was better or no worse than 94\% of the datasets compared with Euclidean, DBA (77\%), SoftDTW (69\%),  DTAN (76\%),  ResNet-TW (70\%).
    }
    \label{fig:ucr:accuracy}
    \end{center}
\end{figure}

\begin{figure}[!htb]
    \begin{center}
    \includegraphics[width=\linewidth]{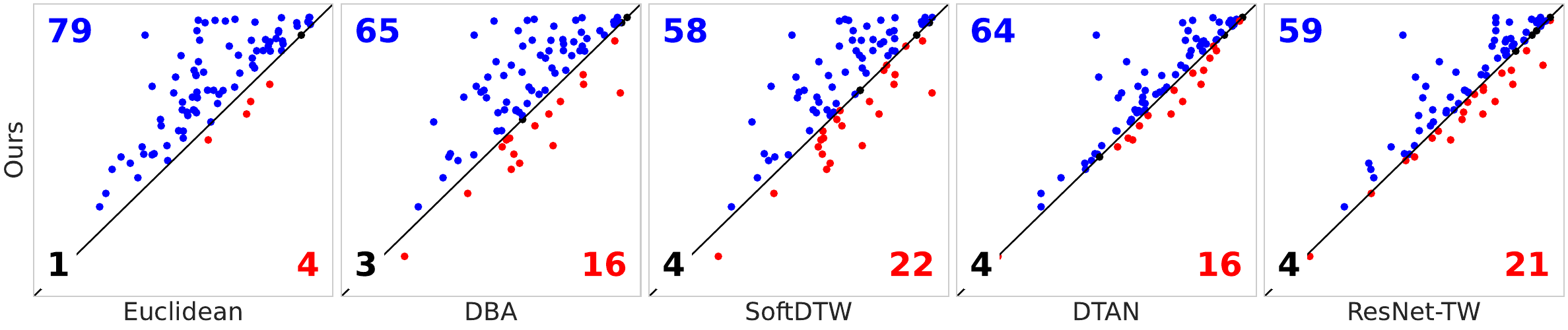}
    \caption{Correct classification rates using NCC. Each point above (below) the diagonal indicates an entire UCR archive dataset where our model achieves better (worse) accuracy. 
    }
    \label{fig:ucr_dataset_rates}
    \end{center}
\end{figure}

\begin{figure}[!htb]
    \begin{center}
        \includegraphics[width=0.48\linewidth]{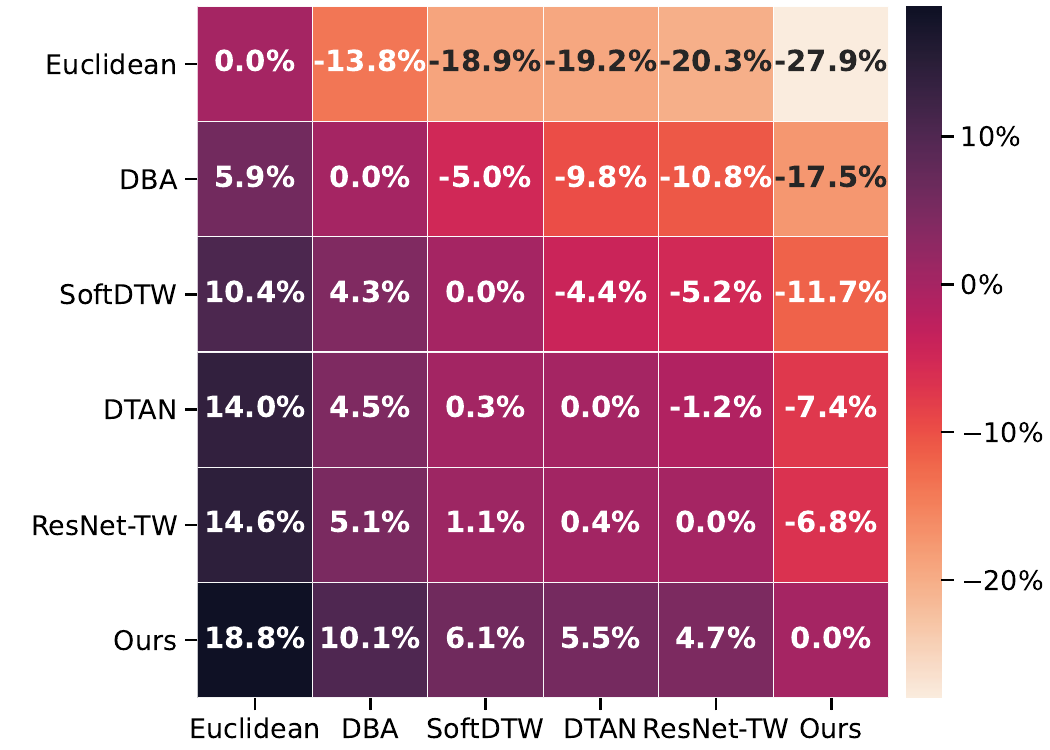}
        \hfill
        \includegraphics[width=0.48\linewidth]{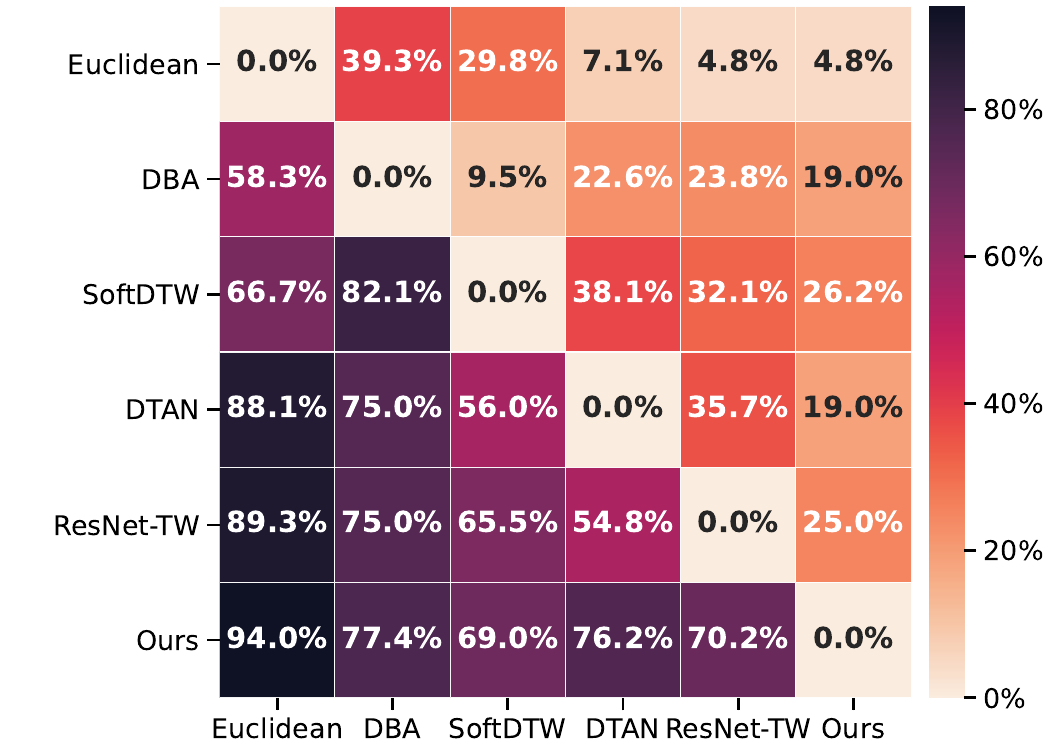}
        \caption{Percentage of wins (left) (row against column) and average percentage of change in accuracy (right) over UCR archive.}
        \label{fig:ucr_dataset_percentages}
    \end{center}
\end{figure}

In this section, we show detailed quantitative results of the NCC experiment on UCR archive \cite{dau2019ucr} in \cref{fig:ucr:accuracy,fig:ucr_dataset_cd,fig:ucr_dataset_rates,fig:ucr_dataset_percentages,tab:ucr_dataset_by_type,tab:ucr_dataset_accuracy}.
We compare our Temporal Transformer Network with Euclidean average (as a baseline), DTW Barycenter Averaging (DBA) \cite{Petitjean2011-DBA}, SoftDTW \cite{cuturi2017soft}, DTAN \cite{Weber2019} and ResNet-TW \cite{Huang2021}.

Results show that temporal misalignment strongly affects the Euclidean mean, and DBA usually reaches a local minimum. SoftDTW, DTAN and ResNet-TW show similar quantitative results. NCC test accuracy found that our method was better or no worse in 94\% of the datasets compared with Euclidean,  DBA (77\%), SoftDTW (69\%), DTAN (76\%) and ResNet-TW (70\%).
Overall, our proposed TTN using \textit{DIFW} beats all 5 comparing methods. The precise computation of the gradient of the transformation translates to an efficient search in the parameter space, which leads to faster and better solutions at convergence.

\clearpage
\small
\renewcommand{\arraystretch}{0.8}
\begin{longtable}{lcccccc}
\caption{Accuracy of nearest centroid classification on UCR dataset \cite{dau2019ucr}. Highest values per dataset (row) are shown in bold.}\label{tab:ucr_dataset_accuracy}\\
\toprule
Dataset &   Euclidean &        DBA &    SoftDTW &       DTAN &  ResNet-TW &       Ours \\ 
\midrule 
\endfirsthead 
\caption{Accuracy of nearest centroid classification on UCR dataset \cite{dau2019ucr}. Highest values per dataset (row) are shown in bold.}\\
\toprule
Dataset &   Euclidean &        DBA &    SoftDTW &       DTAN &  ResNet-TW &       Ours \\ 
\midrule \endhead 
                            Adiac &   0.5498 &   0.4629 &   0.5012 &   0.6956 &   0.6982 &   \textbf{0.7186} \\ 
                        ArrowHead &   0.6114 &   0.4742 &   0.5200 &   0.7485 &   \textbf{0.7542} &   0.7257 \\ 
                            Beef &   0.5333 &   0.4000 &   0.5666 &   0.6333 &   0.6333 &   \textbf{0.7000} \\ 
                        BeetleFly &   0.8500 &   0.9000 &   0.8500 &   0.8000 &   0.8000 &   \textbf{0.9500} \\ 
                    BirdChicken &   0.5500 &   0.6000 &   0.7000 &   0.8000 &   \textbf{0.9500} &   \textbf{0.9500} \\ 
                            Car &   0.6166 &   0.6333 &   0.6833 &   0.8166 &   \textbf{1.0000} &   0.9889 \\ 
                            CBF &   0.7633 &   0.9655 &   0.9711 &   0.9144 &   0.8500 &   \textbf{0.9822} \\ 
            ChlorineConcentration &   0.3330 &   0.3236 &   0.3481 &   0.3330 &   0.3518 &   \textbf{0.3966} \\ 
                    CinCECGTorso &   0.3855 &   0.4456 &   0.3985 &   0.6159 &   0.5427 &   \textbf{0.7405} \\ 
                        Coffee &   0.9642 &   0.9642 &   0.9642 &   \textbf{1.0000} &   0.9642 &   \textbf{1.0000} \\ 
                        Computers &   0.4160 &   0.6160 &   0.6400 &   0.5920 &   \textbf{0.6760} &   0.6160 \\ 
                        CricketX &   0.2384 &   0.5743 &   \textbf{0.6025} &   0.4230 &   0.3410 &   0.4282 \\ 
                        CricketY &   0.3487 &   0.5410 &   \textbf{0.5717} &   0.5410 &   0.4153 &   0.5128 \\ 
                        CricketZ &   0.3051 &   0.6051 &   \textbf{0.6153} &   0.4205 &   0.3333 &   0.4512 \\ 
            DiatomSizeReduction &   0.9575 &   0.9509 &   0.9509 &   0.9705 &   0.9738 &   \textbf{0.9836} \\ 
    DistalPhalanxOutlineAgeGroup &   0.8175 &   0.8400 &   0.8500 &   0.8475 &   \textbf{0.8625} &   0.7482 \\ 
    DistalPhalanxOutlineCorrect &   0.4716 &   0.4883 &   0.4900 &   0.4716 &   0.5050 &   \textbf{0.7753} \\ 
                DistalPhalanxTW &   0.7475 &   0.7550 &   0.7600 &   0.7800 &   \textbf{0.7975} &   0.6834 \\ 
                    Earthquakes &   0.7546 &   0.5745 &   0.8229 &   0.7732 &   \textbf{0.9732} &   0.8200 \\ 
                        ECG200 &   0.7500 &   0.7200 &   0.7300 &   0.7900 &   0.7950 &   \textbf{0.9137} \\ 
                        ECG5000 &   0.8604 &   0.8346 &   0.8537 &   0.8913 &   0.8000 &   \textbf{0.9988} \\ 
                    ECGFiveDays &   0.6898 &   0.6585 &   0.6701 &   0.9779 &   0.9315 &   \textbf{0.9930} \\ 
                ElectricDevices &   0.4826 &   0.5389 &   0.5397 &   0.5348 &   0.5188 &   \textbf{0.5737} \\ 
                        FaceAll &   0.4917 &   0.7964 &   0.8278 &   0.8047 &   0.8409 &   \textbf{0.8562} \\ 
                        FaceFour &   0.8409 &   0.8522 &   0.8522 &   0.8295 &   0.8551 &   \textbf{0.9204} \\ 
                        FacesUCR &   0.5395 &   0.7746 &   0.8126 &   0.8570 &   \textbf{0.8571} &   0.8009 \\ 
                    FiftyWords &   0.5164 &   0.6153 &   0.6153 &   \textbf{0.6527} &   0.5164 &   0.6307 \\ 
                            Fish &   0.5600 &   0.6514 &   0.6971 &   0.9028 &   0.9028 &   \textbf{0.9142} \\ 
                            FordA &   0.4959 &   0.5495 &   0.5529 &   0.6048 &   0.5681 &   \textbf{0.6522} \\ 
                            FordB &   0.4997 &   0.5684 &   \textbf{0.5913} &   0.5797 &   0.5662 &   0.5456 \\ 
                        GunPoint &   0.7533 &   0.7000 &   0.7333 &   \textbf{0.8800} &   0.8066 &   0.8466 \\ 
                            Ham &   0.7619 &   0.7238 &   0.7333 &   0.7904 &   0.7619 &   \textbf{0.8095} \\ 
                    HandOutlines &   0.8180 &   0.8040 &   0.8120 &   0.8500 &   0.8350 &   \textbf{0.9081} \\ 
                        Haptics &   0.3928 &   0.3506 &   0.3733 &   0.4577 &   0.4642 &   \textbf{0.4870} \\ 
                        Herring &   0.5468 &   0.5468 &   0.6093 &   0.7031 &   0.7656 &   \textbf{0.7812} \\ 
                    InlineSkate &   0.1927 &   0.2327 &   0.2527 &   0.2600 &   0.2436 &   \textbf{0.2872} \\ 
            InsectWingbeatSound &   0.6010 &   0.2893 &   0.3282 &   0.5873 &   0.5707 &   \textbf{0.6065} \\ 
                ItalyPowerDemand &   0.9183 &   0.7308 &   0.7502 &   0.9620 &   0.9650 &   \textbf{0.9669} \\ 
        LargeKitchenAppliances &   0.4400 &   0.7280 &   \textbf{0.7333} &   0.4826 &   0.5013 &   0.5173 \\ 
                    Lightning2 &   0.6885 &   0.6393 &   0.6229 &   0.7213 &   \textbf{0.7540} &   0.7377 \\ 
                    Lightning7 &   0.5890 &   0.6986 &   \textbf{0.7260} &   0.7123 &   0.6849 &   \textbf{0.7260} \\
                        Mallat &   0.9667 &   0.9526 &   0.9539 &   0.9688 &   0.9667 &   \textbf{0.9735} \\ 
                            Meat &   \textbf{0.9333} &   0.9166 &   \textbf{0.9333} &   \textbf{0.9333} &   \textbf{0.9333} &   \textbf{0.9333} \\ 
                    MedicalImages &   0.3855 &   0.4368 &   0.4618 &   0.4684 &   0.4736 &   \textbf{0.4828} \\ 
    MiddlePhalanxOutlineAgeGroup &   0.7325 &   0.7125 &   \textbf{0.7950} &   0.7375 &   0.7525 &   0.6363 \\ 
    MiddlePhalanxOutlineCorrect &   0.5516 &   0.4833 &   0.4950 &   0.5433 &   0.5316 &   \textbf{0.6975} \\ 
                MiddlePhalanxTW &   0.5914 &   0.5563 &   0.5814 &   0.5964 &   \textbf{0.6340} &   0.5389 \\ 
                    MoteStrain &   0.8610 &   0.8266 &   0.8434 &   0.9041 &   \textbf{0.9129} &   0.8746 \\ 
    NonInvasiveFetalECGThorax1 &   0.7694 &   0.7129 &   0.7109 &   0.8534 &   0.8386 &   \textbf{0.8743} \\ 
    NonInvasiveFetalECGThorax2 &   0.8020 &   0.7638 &   0.7730 &   0.9053 &   0.8386 &   \textbf{0.9165} \\ 
                        OliveOil &   0.8666 &   0.7666 &   0.8000 &   0.8666 &   0.8666 &   \textbf{0.9000} \\ 
                        OSULeaf &   0.3595 &   0.4380 &   0.4752 &   0.4628 &   0.4586 &   \textbf{0.9333} \\ 
        PhalangesOutlinesCorrect &   0.6258 &   0.6328 &   0.6375 &   0.6421 &   0.6631 &   \textbf{0.6759} \\ 
                        Phoneme &   0.0785 &   0.1824 &   \textbf{0.2046} &   0.1017 &   0.1165 &   0.1007 \\ 
                            Plane &   0.9619 &   \textbf{1.0000} &   0.9904 &   \textbf{1.0000} &   \textbf{1.0000} &   \textbf{1.0000} \\ 
ProximalPhalanxOutlineAgeGroup &   0.8195 &   0.8439 &   0.8536 &   0.8536 &   \textbf{0.8731} &   \textbf{0.8731} \\ 
    ProximalPhalanxOutlineCorrect &   0.6460 &   0.6494 &   \textbf{0.7250} &   0.6426 &   0.6872 &   \textbf{0.7250} \\
                ProximalPhalanxTW &   0.7075 &   0.7350 &   0.7475 &   0.8175 &   \textbf{0.8225} &   0.7902 \\ 
            RefrigerationDevices &   0.3546 &   0.5840 &   \textbf{0.5866} &   0.4667 &   0.48267 &   0.4853 \\ 
                    ScreenType &   0.4426 &   0.3786 &   0.3893 &   0.4453 &   \textbf{0.4693} &   0.4613 \\ 
                    ShapeletSim &   0.5000 &   0.5222 &   \textbf{0.5888} &   0.5389 &   \textbf{0.5888} &   0.5722 \\ 
                        ShapesAll &   0.5133 &   0.6033 &   0.6283 &   0.6283 &   \textbf{0.6817} & 0.6433 \\
        SmallKitchenAppliances &   0.4186 &   \textbf{0.6613} &   0.6586 &   0.6213 &   0.5600 &   0.5920 \\ 
            SonyAIBORobotSurface1 &   0.8119 &   0.8352 &   \textbf{0.8935} &   \textbf{0.8935} &   0.8602 &   0.8918 \\
            SonyAIBORobotSurface2 &   0.7932 &   0.7660 &   0.7722 &   0.8111 &   0.8300 &   \textbf{0.8751} \\ 
                    Strawberry &   0.6688 &   0.6166 &   0.6492 &   0.8433 &   0.7862 &   \textbf{0.8918} \\ 
                    SwedishLeaf &   0.7024 &   0.6816 &   0.7232 &   0.8064 &   0.8368 &   \textbf{0.8576} \\ 
                        Symbols &   0.8643 &   \textbf{0.9547} &   \textbf{0.9547} &   0.8572 &   0.9065 &   0.9115 \\
                SyntheticControl &   0.9166 &   \textbf{0.9800} &   \textbf{0.9800} &   0.9500 &   0.9500 &   \textbf{0.9800} \\ 
                ToeSegmentation1 &   0.5745 &   0.6140 &   0.6710 &   0.6403 &   0.6535 &   \textbf{0.7938} \\ 
                ToeSegmentation2 &   0.5461 &   0.8384 &   \textbf{0.8538} &   0.7538 &   0.7461 &   0.7846 \\
                            Trace &   0.5800 &   0.9700 &   0.9700 &   0.7800 &   0.8000 &   \textbf{0.9800} \\ 
                    TwoLeadECG &   0.5548 &   0.8112 &   0.8015 &   0.9561 &   0.9552 &   \textbf{0.9894} \\ 
                    TwoPatterns &   0.4647 &   0.9750 &   \textbf{0.9897} &   0.5557 &   0.7005 &   0.7157 \\
        UWaveGestureLibraryAll &   0.8495 &   0.8319 &   0.8336 &   0.9207 &   0.9115 &   \textbf{0.9438} \\ 
            UWaveGestureLibraryX &   0.6312 &   0.6764 &   0.7068 &   0.6811 &   \textbf{0.7219} &   0.7104 \\ 
            UWaveGestureLibraryY &   0.5482 &   0.5254 &   0.5647 &   0.6116 &   0.6172 &   \textbf{0.6412} \\ 
            UWaveGestureLibraryZ &   0.5374 &   0.5924 &   0.6041 &   0.6420 &   0.6462 &   \textbf{0.6521} \\ 
                            Wafer &   0.6544 &   0.5110 &   0.6494 &   \textbf{0.9889} &   0.9828 &   0.9862 \\ 
                            Wine &   0.5555 &   0.5185 &   0.5740 &   0.5740 &   0.5925 &   \textbf{0.8333} \\ 
                    WordSynonyms &   0.2711 &   0.3448 &   0.4122 &   0.4749 &   \textbf{0.5015} &   0.4749 \\ 
                            Worms &   0.2154 &   \textbf{0.4143} &   0.4088 &   0.2596 &   0.3425 &   0.3376 \\
                    WormsTwoClass &   0.5414 &   0.5911 &   \textbf{0.6519} &   0.6187 &   0.6187 &   0.6493 \\
                            Yoga &   0.4970 &   0.5570 &   0.5740 &   0.6316 &   \textbf{0.6967} &   0.6810 \\
\midrule
Average accuracy   &   0.6108 &   0.6557 &   0.6821 &   0.7054 &   0.7111 &   \textbf{0.7489} \\
Average arithmetic ranking &   4.964  &  4.488 &   3.297 &   2.988 &   2.666 &   \textbf{1.940} \\
Average geometric ranking  &  4.760 &   4.077 &   2.816 &   2.730 &   2.299 &   \textbf{1.589} \\
Winning times      &   1 &   5 &  18 &   7 &  21 &  \textbf{49} \\
\bottomrule
\bottomrule
\end{longtable}
\renewcommand{\arraystretch}{1}
\normalsize
\clearpage

\begin{table}[!htb]
    \captionsetup{skip=0pt}
    \caption{Accuracy of nearest centroid classification on UCR dataset \cite{dau2019ucr}, grouped by time series type. Highest values per dataset (row) are shown in bold.}
    \label{tab:ucr_dataset_by_type}
    \begin{center}
    \begin{small}
    \begin{tabular}{lcccccc}
        \toprule
            Type (count) &   Euclidean &        DBA &    SoftDTW &       DTAN &  ResNet-TW &       Ours \\
        \midrule
        Device (6) &   0.4257 &   0.5844 &   \textbf{0.5912} &   0.5238 &   0.5347 &   0.5409 \\
            ECG (7) &   0.6874 &   0.7067 &   0.7054 &   0.8557 &   0.8145 &   \textbf{0.9180} \\
        Image (29) &   0.6254 &   0.6600 &   0.6866 &   0.7199 &   0.7391 &   \textbf{0.7679} \\
        Motion (14) &   0.4768 &   0.5777 &   0.6031 &   0.5793 &   0.5615 &   \textbf{0.6090} \\
        Sensor (15) &   0.6603 &   0.6517 &   0.6934 &   0.7491 &   0.7723 &   \textbf{0.7835} \\
    Simulated (6) &   0.6574 &   0.7865 &   \textbf{0.8053} &   0.7101 &   0.7346 &   0.7700 \\
        Spectro (7) &   0.7548 &   0.7009 &   0.7458 &   0.8058 &   0.7912 &   \textbf{0.8668} \\
            \midrule
        Total (84) &   0.6108 &   0.6557 &   0.6821 &   0.7054 &   0.7111 &   \textbf{0.7489} \\
        \bottomrule
    \end{tabular}
    \end{small}
    \end{center}
\end{table}

\subsubsection{Critical difference} 
\begin{figure}[!htb]
    \begin{center}
    \includegraphics[width=0.85\linewidth]{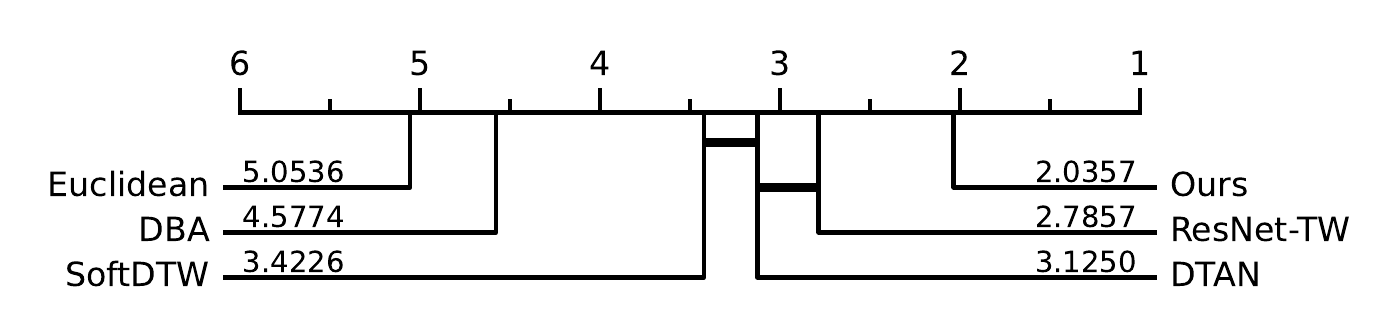}
    \caption{Critical difference diagram for the proposed method and five competing methods (Euclidean, DBA, SoftDTW, DTAN, ResNet-TW). 
    Solid bars indicate cliques, within which there is no significant difference in rank. Tests are performed with the Wilcoxon sign rank test using the Holm correction. }
    \label{fig:ucr_dataset_cd}
    \end{center}
\end{figure}

A critical difference diagram (see \cref{fig:ucr_dataset_cd}) is a graphical tool used to compare the performance of multiple algorithms or techniques. It is particularly useful when comparing experiments that have been run on different datasets. The diagram provides a visual representation of the relative performance of the methods being compared, allowing for easy identification of which methods are significantly different from one another.
The diagram is constructed by plotting the performance of each method on the x-axis. 
The methods are then connected by solid bars to indicate cliques, within which there is no significant difference in rank. The critical difference is performed with the Wilcoxon sign rank test using the Holm correction. 

\subsubsection{The Texas Sharpshooter Fallacy}\label{sec:texas}

The Texas Sharpshooter Fallacy is a common logic error that occurs when comparing methods across multiple datasets. In fact, it is not sufficient to have a method that can be more accurate on some datasets unless one can predict on which datasets it will be more accurate.
Therefore, in this section we test whether the \textbf{expected} accuracy gain over another competing method coincides with the \textbf{actual} accuracy gain. The expected accuracy gain and the actual accuracy gain are defined by \cref{eq:texas_expected} and \cref{eq:texas_actual}, respectively.
An expected accuracy gain larger than one indicates that we predict our method will perform better, while an actual accuracy gain larger than one indicates that our method indeed performs better.

\begin{equation}\label{eq:texas_expected}
\text{expected gain} = \frac{\text{accuracy}_{train}(\text{our method})}{\text{accuracy}_{train}(\text{other method})}
\end{equation}
\begin{equation}\label{eq:texas_actual}
\text{actual gain} = \frac{\text{accuracy}_{test}(\text{our method})}{\text{accuracy}_{test}(\text{other method})}
\end{equation}

\begin{figure}[!htb]
    \begin{center}
    \centerline{\includegraphics[width=\linewidth]{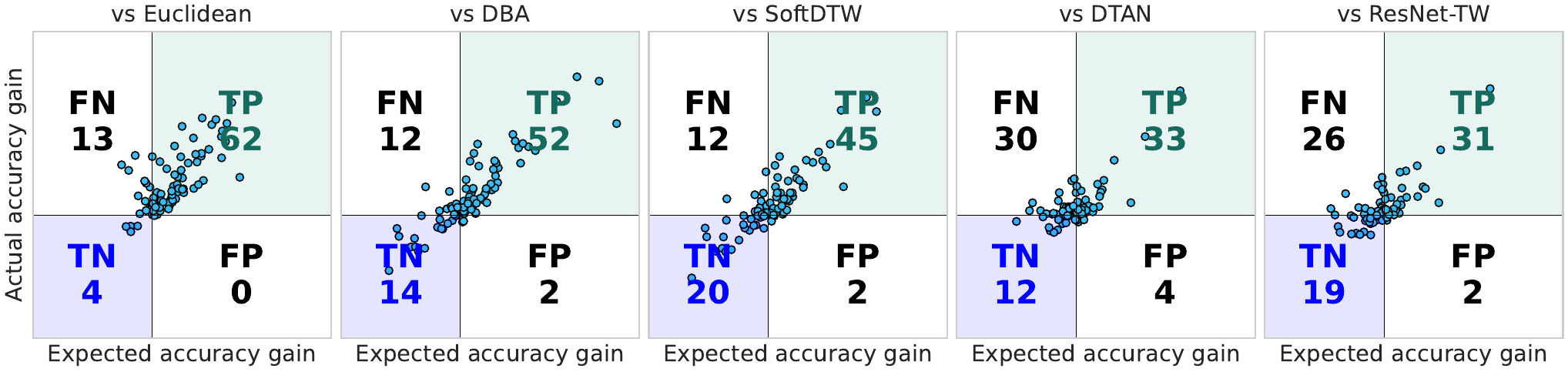}}
    \caption{Texas sharpshooter plot of the proposed method against the other five competing methods respectively. Each point represents an entire dataset.}
    \label{fig:ucr_dataset_texas}
    \end{center}
\end{figure}

As shown in \cref{fig:ucr_dataset_texas}, Texas sharpshooter plot is a convenient tool to visualize the comparison between the expected accuracy gain and the actual accuracy gain on multiple datasets. Each point represents a dataset which falls into one of the following four possibilities:

\begin{table}[!htb]
    \centering
    \caption{Number of datasets in each Texas sharpshooter quadrant.}
    \label{tab:ucr_dataset_texas}
    \begin{tabular}{lcccc}
    \toprule
    \textbf{Method} & \textbf{TP}     & \textbf{TN}     & \textbf{FN}     & \textbf{FP}    \\ \midrule
    vs Euclidean   & 62              & 4               & 13              & 0              \\
    vs DBA         & 52              & 14               & 12              & 2              \\
    vs SoftDTW     & 45              & 20              & 12              & 2              \\
    vs DTAN        & 33              & 12              & 30              & 4              \\
    vs ResNet-TW   & 31              & 19              & 26              & 2              \\\midrule
    Total (395) & 223             & 69              & 93              & 10             \\
    & 56.5\% & 17.5\% & 23.5\% & 2.5\% \\ \bottomrule
    \end{tabular}%
\end{table}

\begin{itemize}
    \item \textbf{TP} (True Positive, 223/395, 56.5\%): we predicted that our method would increase accuracy, and it did. This is the most beneficial situation and the majority of points fall into this region.
    \item \textbf{TN} (True Negative, 69/395, 17.5\%): we predicted that our method would decrease accuracy, and it did. This is not truly a bad region since knowing ahead of time that a method will do worse allows choosing another method to avoid the loss of accuracy.
    \item \textbf{FN} (False Negative, 93/395, 23.5\%): we predicted that our method would decrease accuracy but it actually increased. This is also not a bad case even though we might miss the opportunity to improve.
    \item \textbf{FP} (False Positive, 10/395, 2.5\%): we predicted our method would increase accuracy and it did not. Truly the painful region to be, but fortunately not many points fall into this region.
\end{itemize}

\subsubsection{Correlation With Dataset Size}\label{sec:correlation}

\begin{figure}[!htb]
    \begin{center}
    \centerline{\includegraphics[width=\linewidth]{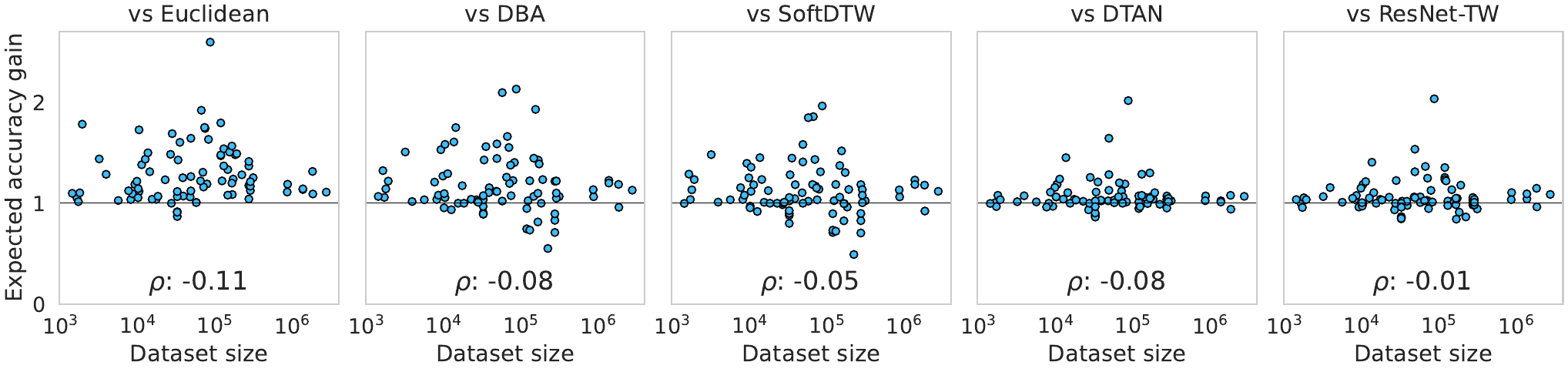}}
    \caption{Dataset size (number of training signals $\times$ time series length) vs expected accuracy gain against the other five competing methods. Each point represents an entire dataset. The Pearson \cite{pearson1895vii} coefficient $\rho$ is included as well.}
    \label{fig:ucr_dataset_correlation}
    \end{center}
\end{figure}

Finally, we analyze the correlation between the expected accuracy gain and the dataset size. For each competing method, we analyze whether more data availability translates into larger accuracy gains. As visualized in \cref{fig:ucr_dataset_correlation}, there is a lack of correlation between data availability and accuracy improvement across the five competing methods. The Pearson correlation coefficients are less than 0.15, indicating a very weak bond.

\clearpage

\subsection{Hyperparameter Grid Search}\label{sec:hyperparameter}

For each of the UCR datasets, we trained our TTN for joint alignment as in \cite{Weber2019}, where 
$N_{\mathcal{P}} \in \{16,32,64\}$,
$\lambda_{\sigma} \in \{10^{-3},10^{-2}\}$,
$\lambda_{s} \in \{0.1,0.5\}$,
the number of transformer layers $\in \{1,5\}$,
scaling-and-squaring iterations $\in \{0,8\}$
and the option to apply the zero-boundary constraint.
The full table of results is available in the Supplementary Materials, and was not included in the manuscript due to the large number of experiments for the 84 UCR datasets. 

\begin{figure}[!htb]
    \begin{center}
    \centerline{\includegraphics[width=0.9\linewidth]{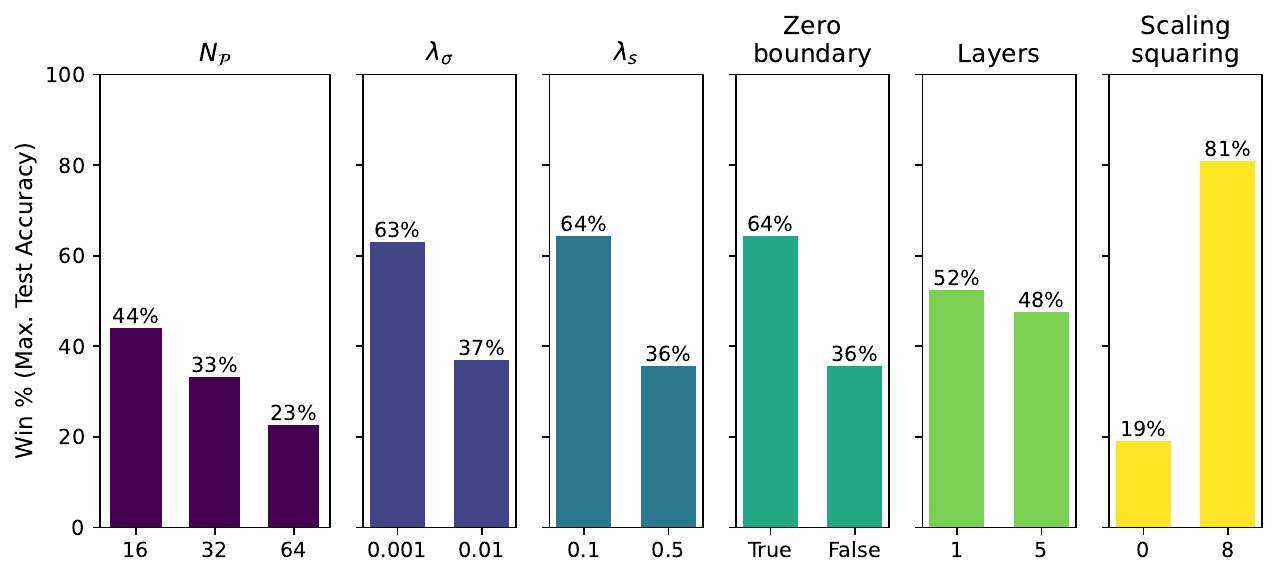}}
    \caption{Winning percentage over 84 UCR datasets for each hyperparameter: 
    (a) tessellation size $N_{\mathcal{P}} \in \{16,32,64\}$, 
    (b) regularization kernel's overall variance $\lambda_{\sigma} \in \{10^{-3},10^{-2}\}$,
    (c) regularization kernel's length-scale $\lambda_{s} \in \{0.1,0.5\}$,
    (d) the number of transformer layers $\in \{1,5\}$,
    (e) scaling-and-squaring iterations $\in \{0,8\}$
    and (f) the option to apply the zero-boundary constraint. A hyperparameter is considered to win in a dataset if it yields maximum NCC test accuracy.
}
    \label{fig:hyperparameter}
    \end{center}
\end{figure}

In total, $84\times3\times2^{5}=8064$ different hyperparameter configurations were tested. To provide a summary of these experiments, we first gathered the hyperparameters that yield optimal (maximum) NCC test accuracy for each UCR dataset. Then we counted how many times each parameter value appears in the optimal NCC accuracy configuration for each of the 84 UCR datasets. \cref{fig:hyperparameter} shows the winning percentage of each hyperparameter value. 
For example, a tessellation size $N_{\mathcal{P}}=16$ appears in $44\%$ of the winning parameter combinations among 84 datasets, while $N_{\mathcal{P}}=32$ and $N_{\mathcal{P}}=64$ get reduced to $33\%$ in $23\%$ respectively. The scaling-and-squaring method appears to be preferable based on the results, with $80\%$ of the datasets using 8 iterations of the method.

\clearpage

\subsection{Ablation Study on Model Expressivity}\label{sec:ablation}

Regarding the expressivity of the proposed model, it is worth noting that the fineness of the partition in CPA velocity functions controls the trade-off between expressiveness and computational complexity. An ablation study is presented in this section to investigate how partition fineness control affects such balance.
We computed the training time and the model accuracy for different values of the tessellation size $N_{\mathcal{P}} \in \{4,16,32,64,128,256\}$.

\begin{figure}[!htb]
    \begin{center}
    \centerline{\includegraphics[width=0.45\linewidth]{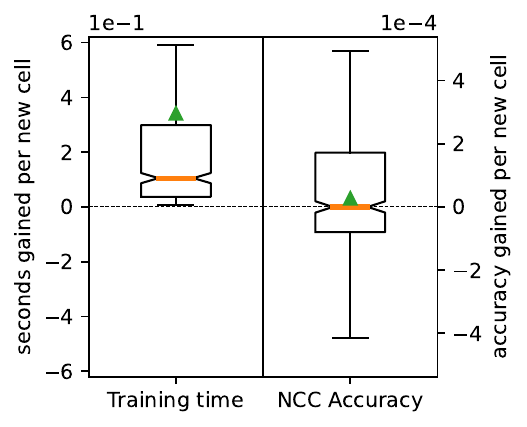}}
    \caption{Ablation study on transformation expressiveness. Distribution of training time (\textbf{left}) and NCC accuracy (\textbf{right}) gained per new cell added to the piecewise transformation.}
    \label{fig:ablation}
    \end{center}
\end{figure}

\vspace{-1cm}
\cref{fig:ablation} and \cref{tab:ablation} show the results from the ablation study in terms of model accuracy and compute time for all 84 UCR datasets and for different values of $N_\mathcal{P}$. Results show that increasing $N_\mathcal{P}$ lead to longer training times, and large $N_\mathcal{P}$ values can hinder accuracy performance. For instance, on average, going from 4 to 16 cells increments the training time by 1.5\% and increments the NCC accuracy by 0.6962\%. However, going from 16 to 32 cells increments the training time by 1.8\% but reduces the NCC accuracy by 0.0905\%.

\begin{table}[!htb]
    \captionsetup{width=0.6\linewidth, skip=0pt}
    \caption{Impact of higher expressive CPA functions (more cells) on NCC model training time and accuracy.}
    \label{tab:ablation}
    \begin{center}
    \begin{tabular}{lcc}
        \toprule
        $N_\mathcal{P}$ & \begin{tabular}[c]{@{}l@{}}Training\\Time\end{tabular} & \begin{tabular}[c]{@{}l@{}}NCC\\Accuracy\end{tabular} \\
        \midrule
        4\hfill$\rightarrow$\hfill16 & \trianbox1{cgreen} 1.5\% & \trianbox1{cgreen} 0.69\% \\
        16\hfill$\rightarrow$\hfill32 & \trianbox1{cgreen} 1.8\% & \uptrianbox1{cred} 0.09\% \\
        32\hfill$\rightarrow$\hfill64 & \trianbox1{cgreen} 5.5\% & \uptrianbox1{cred} 0.20\% \\
        64\hfill$\rightarrow$\hfill128 & \trianbox1{cgreen} 16.3\% & \uptrianbox1{cred} 0.24\% \\
        128\hfill$\rightarrow$\hfill256 & \trianbox1{cgreen} 25.0\% & \uptrianbox1{cred} 0.48\% \\ \bottomrule
    \end{tabular}
    \end{center}
\end{table}

\clearpage
\subsection{Qualitative Alignment Results}\label{sec:additional_results}

In this section, we provide qualitative results of joint alignment of test data in different datasets from the UCR archive \cite{dau2019ucr}.

\subsubsection{Timeline \& Heatmap Visualization}

To begin, we depict the visual differences before and after time series alignment. The layout presented in \cref{fig:alignment_example_1a} is repeated multiple times throughout the rest of the chapter, therefore here we discuss it in further depth: The figure is divided into two parts, the left side represents the original data before alignment, while the right side displays the aligned data.
Note that the data in these figures always corresponds to the test set; the model learns to generalize alignments from the training set, so it does not need to solve a new optimization problem every time. The bottom graphic consists of a timeline of every time series instance (lines in \textcolor{gray}{gray}), and the Euclidean average (line in \textcolor{red}{red}) with one standard deviation shadow area (colored in translucent \textcolor{red}{red}). The heatmap on top represents the time in the x-axis, each time series instance in the y-axis, and the color z-scale corresponds to the series value. This plot makes it easy to compare each time series, which may be overlapping at the bottom plot.

\begin{figure}[!htb]
    \begin{center}
        \begin{subfigure}{0.49\linewidth}
            \centering
            \includegraphics[width=\linewidth]{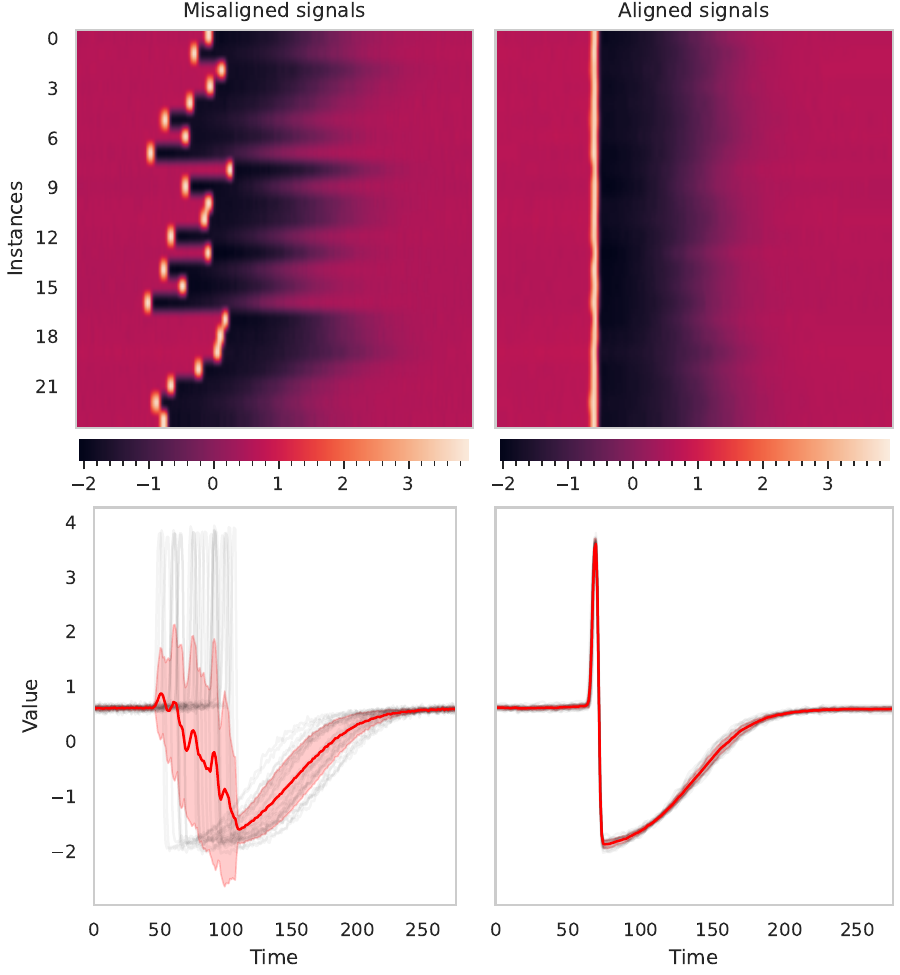}
            \caption{Trace test set, class 0}
        \end{subfigure}
        \begin{subfigure}{0.49\linewidth}
            \centering
            \includegraphics[width=\linewidth]{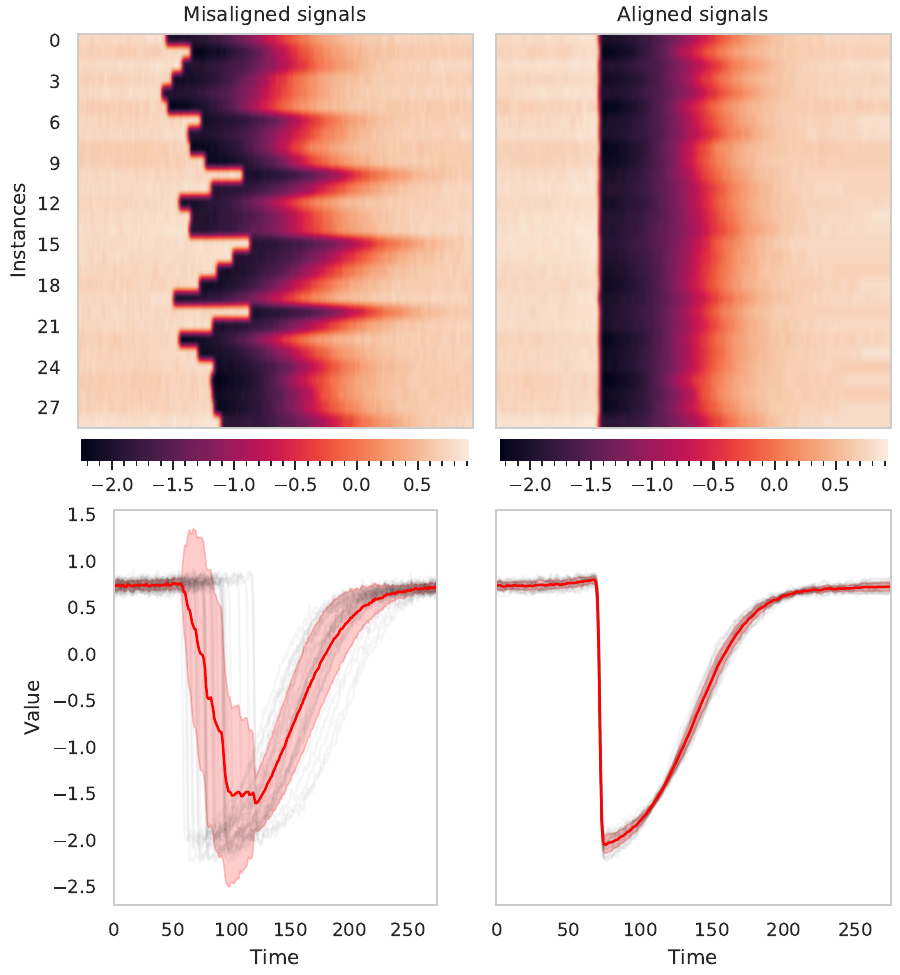}
            \caption{Trace test set, class 1}
        \end{subfigure}
        \caption{Multi-class time series alignment on multiple datasets. \textbf{Top}: heatmap of each time series sample (row). \textbf{Bottom}: overlapping time series, red line represents Euclidean average. \textbf{Left}: original signals. \textbf{Right}: signals after alignment. (cont.)}
        \label{fig:alignment_example_1a}
    \end{center}
\end{figure}

\begin{figure}[!htb]\ContinuedFloat
    \begin{center}
        \begin{subfigure}{0.49\linewidth}
            \centering
            \includegraphics[width=\linewidth, trim=32 262 0 12, clip]{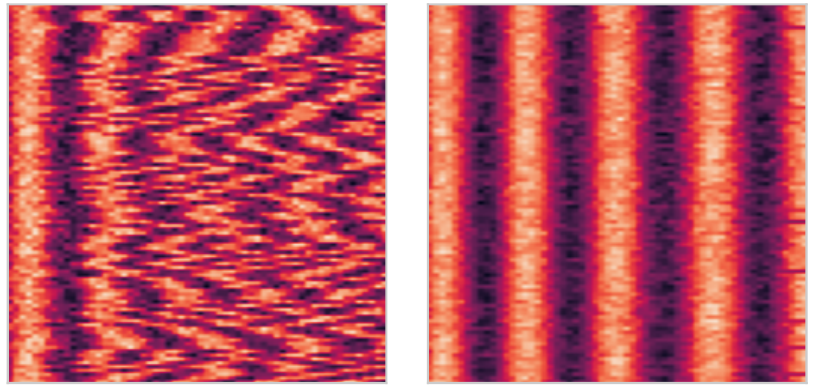}\\
            \includegraphics[width=\linewidth, trim=32 25 0 230, clip]{figures/ucr_dataset/alignment/SyntheticControl1_example_crop.pdf}
            \caption{SyntheticControl test set, class 1}
        \end{subfigure}
        \begin{subfigure}{0.49\linewidth}
            \centering
            \includegraphics[width=\linewidth, trim=32 262 0 12, clip]{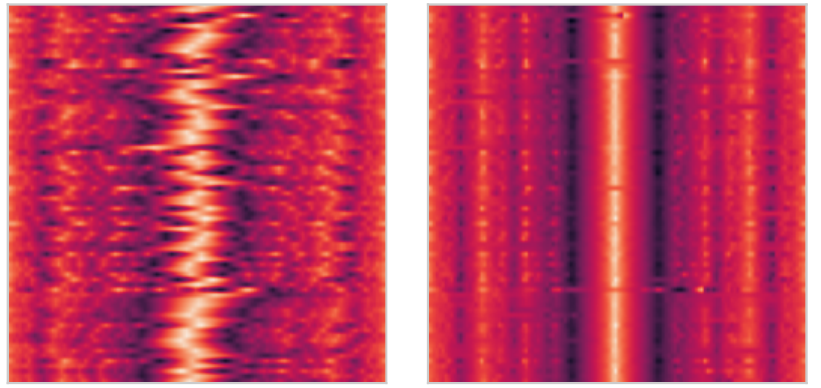}\\
            \includegraphics[width=\linewidth, trim=32 25 0 230, clip]{figures/ucr_dataset/alignment/SwedishLeaf1_example_crop.pdf}
            \caption{SwedishLeaf test set, class 1}
        \end{subfigure}
        \\
        \begin{subfigure}{0.49\linewidth}
            \centering
            \includegraphics[width=\linewidth, trim=34 260 0 12, clip]{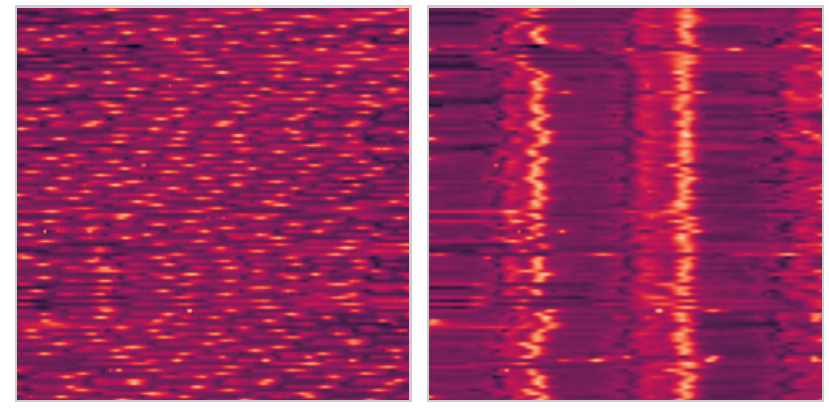}
            \includegraphics[width=\linewidth, trim=34 27 0 240, clip]{figures/ucr_dataset/alignment/ToeSegmentation1_aligned_heatmap_test_0.pdf}
            \caption{ToeSegmentation1 test set, class 0}
        \end{subfigure}
        \begin{subfigure}{0.49\linewidth}
            \centering
            \includegraphics[width=\linewidth, trim=34 262 0 12, clip]{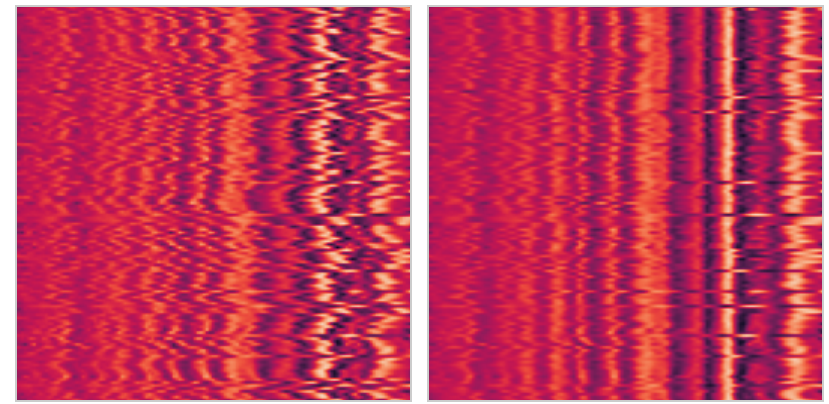}\\
            \includegraphics[width=\linewidth, trim=34 25 0 240, clip]{figures/ucr_dataset/alignment/FacesUCR_aligned_heatmap_test_6.pdf}
            \caption{FacesUCR test set, class 6}
        \end{subfigure}
        \caption{Multi-class time series alignment on multiple datasets. \textbf{Top}: heatmap of each time series sample (row). \textbf{Bottom}: overlapping time series, red line represents Euclidean average. \textbf{Left}: original signals. \textbf{Right}: signals after alignment.}
        \label{fig:alignment_example_1b}
    \end{center}
    \end{figure}

\clearpage
\subsubsection{t-SNE Visualization}

\begin{figure}[!htb]
    \begin{center}
        \begin{subfigure}{0.48\linewidth}
            \centering
            \includegraphics[width=0.95\linewidth, trim=28 17 0 14, clip]{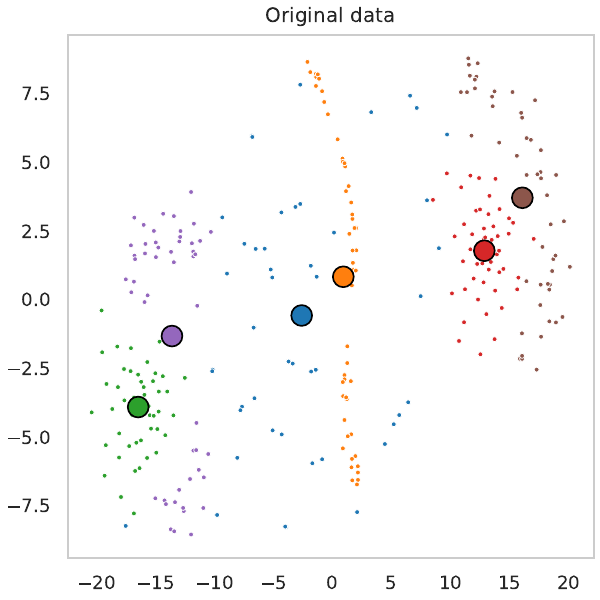}
            \caption{Original data}
        \end{subfigure}
        \hfill
        \begin{subfigure}{0.48\linewidth}
            \centering
            \includegraphics[width=0.95\linewidth, trim=28 17 0 14, clip]{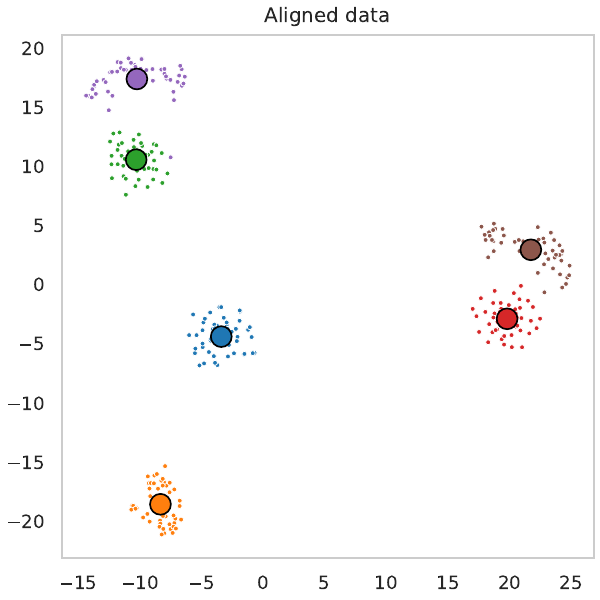}
            \caption{Aligned data}
        \end{subfigure}
    \caption{t-SNE visualization before and after alignment: SyntheticControl dataset.}
    \label{fig:alignment_tsne_SyntheticControl}
    \end{center}
\end{figure}

\begin{figure}[!htb]
    \begin{center}
        \begin{subfigure}{0.48\linewidth}
            \centering
            \includegraphics[width=0.95\linewidth, trim=28 17 0 14, clip]{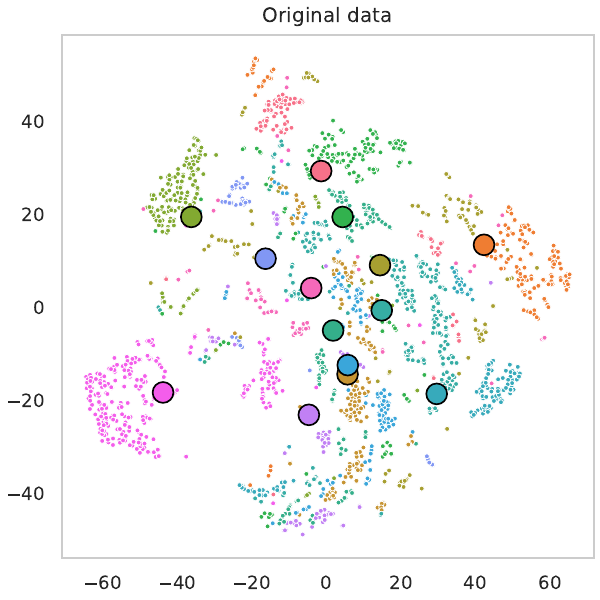}
            \caption{Original data}
        \end{subfigure}
        \hfill
        \begin{subfigure}{0.48\linewidth}
            \centering
            \includegraphics[width=0.95\linewidth, trim=28 17 0 14, clip]{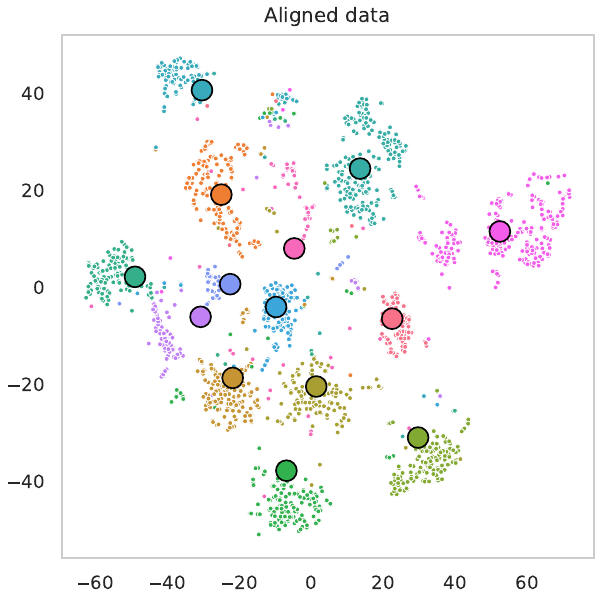}
            \caption{Aligned data}
        \end{subfigure}
    \caption{t-SNE visualization before and after alignment: FacesUCR dataset.}
    \label{fig:alignment_tsne_FacesUCR}
    \end{center}
\end{figure}

\cref{fig:alignment_tsne_SyntheticControl,fig:alignment_tsne_FacesUCR} show a t-SNE \cite{van2008visualizing} visualization of the original and aligned data of the 6-class Synthetic and the 11-class FacesUCR dataset respectively. This illustrates how the proposed TTN indeed decreases intra-class variance while increasing inter-class one, thus improving the performance of classification.

\vspace{-1em}

\subsubsection{Within-class Variance Reduction}

\begin{figure}[!htb]
    \begin{center}
    \begin{subfigure}[t]{\linewidth}
        \centering
        \includegraphics[width=\linewidth]{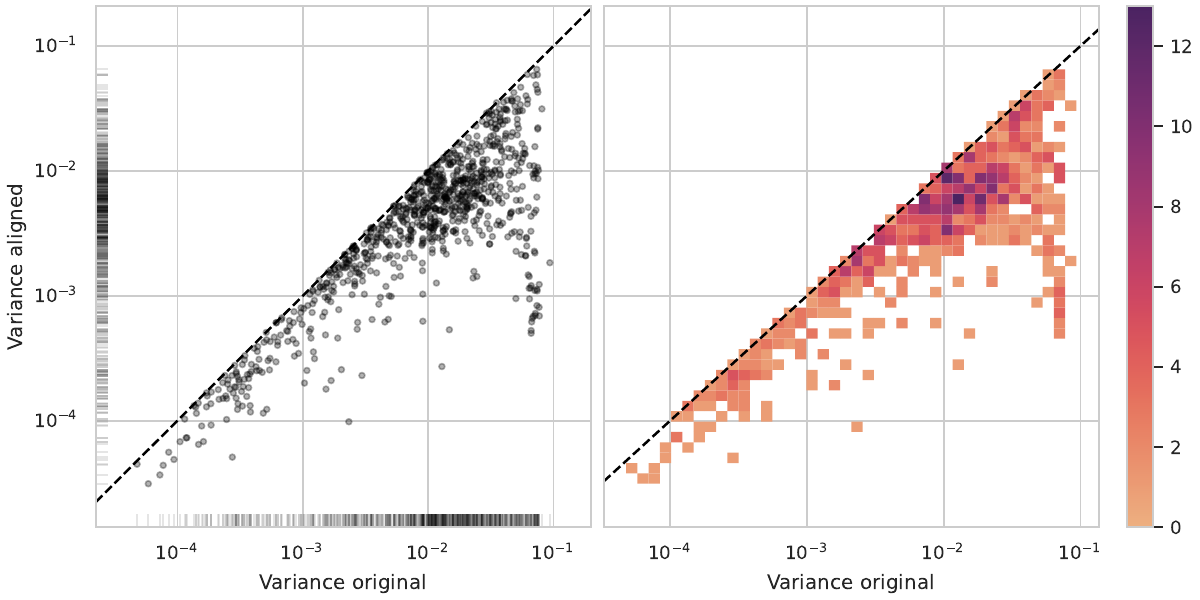}
        \caption{Time series variance before and after alignment. \textbf{Left}: scatterplot, each point represents a (dataset, class) tuple. \textbf{Right}: 2D Histogram, counts the number of (dataset, class) tuples in each bin.}
    \end{subfigure}
    \\
    \begin{subfigure}[t]{0.48\linewidth}
        \centering
        \includegraphics[width=\linewidth]{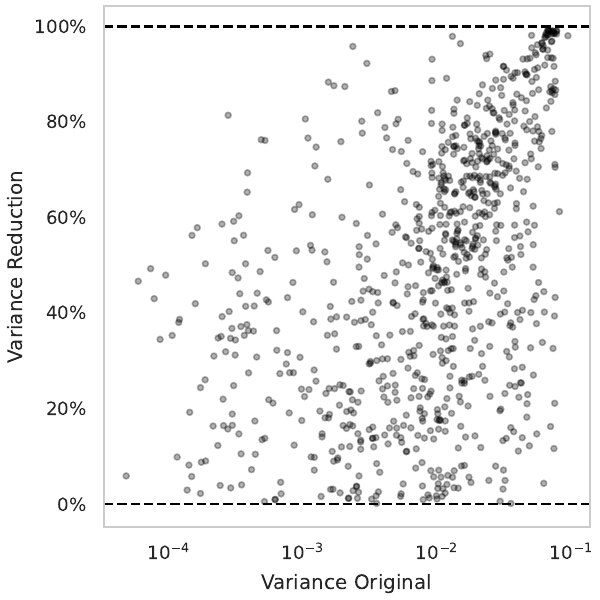}
        \caption{Reduction relative to original variance.}
    \end{subfigure}
    \hfill
    \begin{subfigure}[t]{0.49\linewidth}
        \centering
        \includegraphics[width=0.95\linewidth, trim=0 0 300 0, clip]{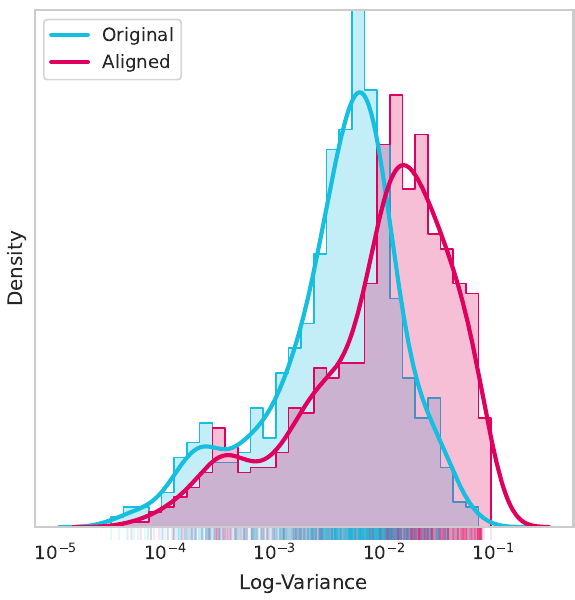}
        \caption{Density of variance before and after alignment.}
    \end{subfigure}
    \caption{Study of within-class variance reduction before \& after time series alignment. Univariate UCR datasets \cite{dau2019ucr}.}
    \label{fig:variance_univariate}
\end{center}
\end{figure}

This section explores the reduction of within-class variance after alignment. This statistic provides an idea of how much temporal misalignment has been repaired. We report results for all UCR (dataset, class) tuples in \cref{fig:variance_univariate}, and more detailed stats in \cref{fig:alignment_variance_1,fig:alignment_variance_2,fig:alignment_variance_3}.

The distribution of variance reduction varies depending on the dataset and the extent of temporal misalignment. 
In \cref{fig:alignment_variance_1}, it is important to note that the decrease achieved in the train set is carried over to the test set, indicating that the model has been able to learn and generalize the warping misalignment in the data to unseen samples.

\begin{figure}[!htb]
    \begin{center}
    \begin{subfigure}[t]{0.37\linewidth}
        \centering
        \includegraphics[height=1.3\linewidth]{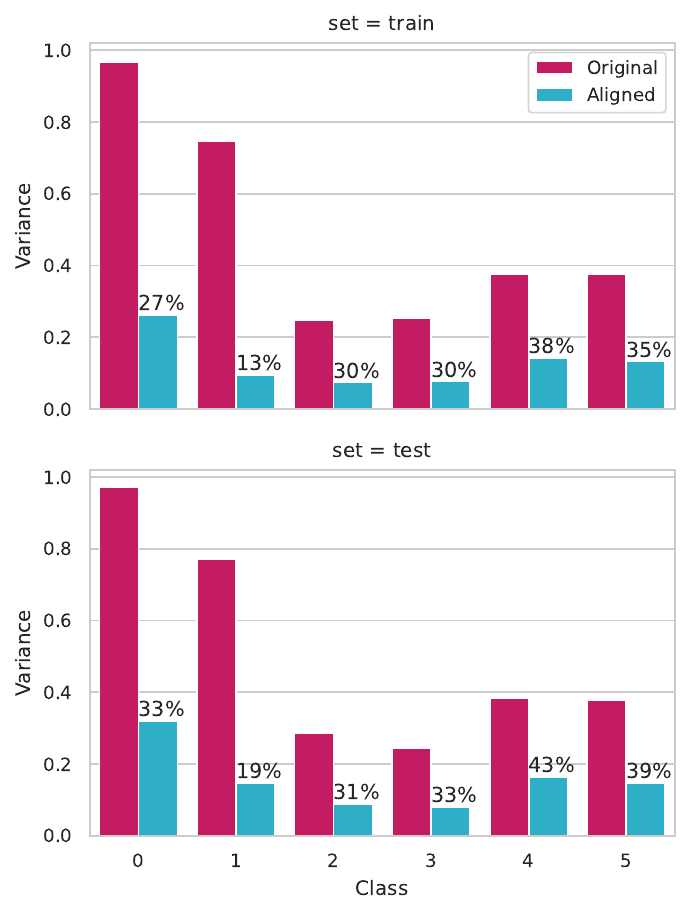}
        \caption{SyntheticControl dataset}
    \end{subfigure}
    \hfill
    \begin{subfigure}[t]{0.24\linewidth}
        \centering
        \includegraphics[height=2\linewidth]{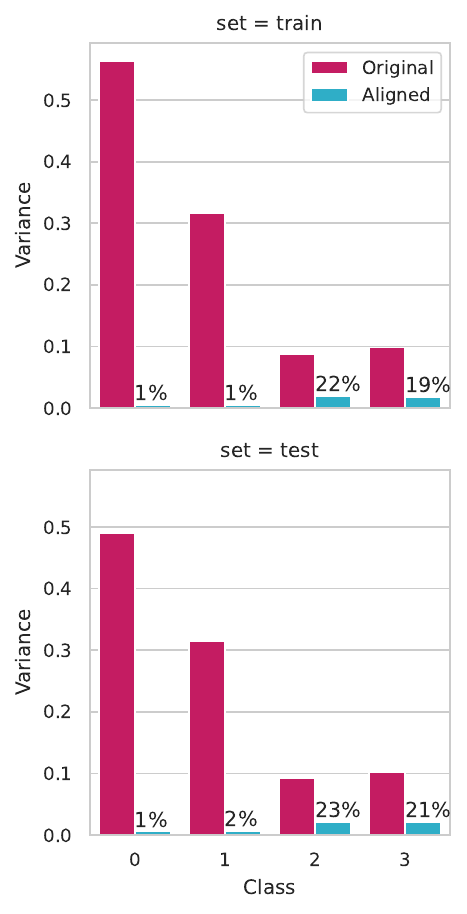}
        \caption{Trace dataset}
    \end{subfigure}
    \hfill
    \begin{subfigure}[t]{0.37\linewidth}
        \centering
        \includegraphics[height=1.3\linewidth]{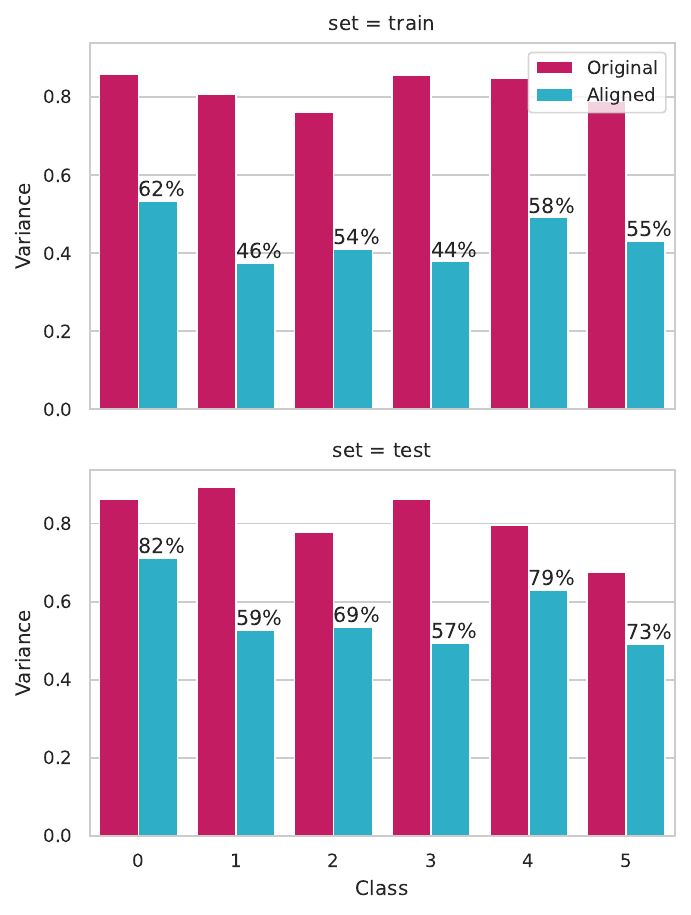}
        \caption{OSULeaf dataset}
    \end{subfigure}
    \\
    \begin{subfigure}[t]{0.48\linewidth}
        \centering
        \includegraphics[height=1.1\linewidth]{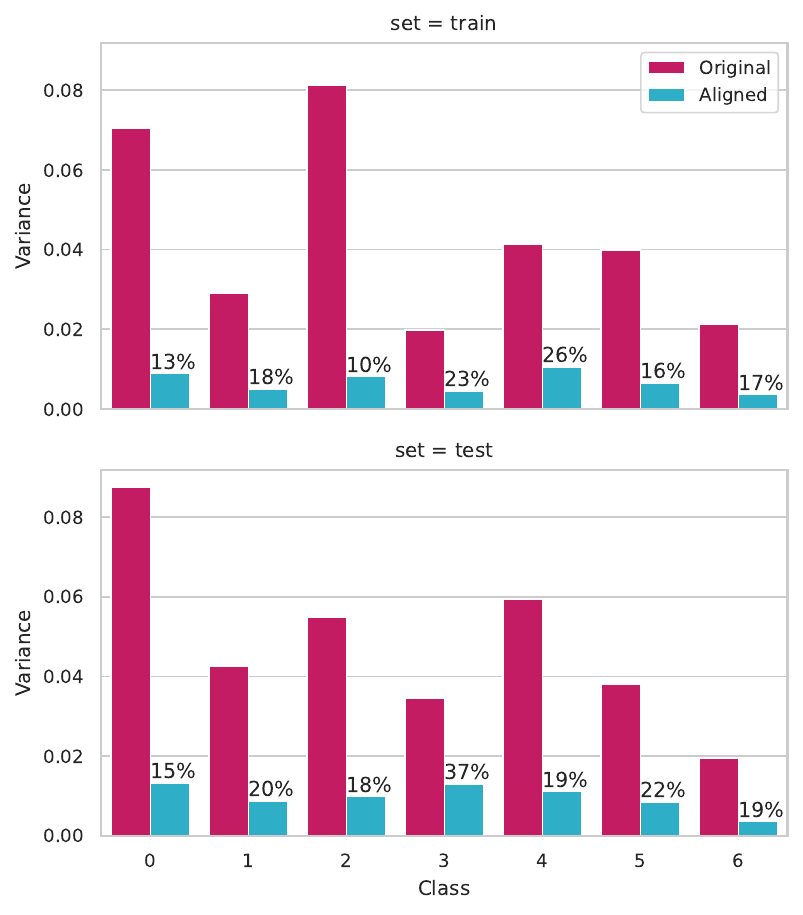}
        \caption{Fish dataset}
    \end{subfigure}
    \hfill
    \begin{subfigure}[t]{0.48\linewidth}
        \centering
        \includegraphics[height=1.1\linewidth]{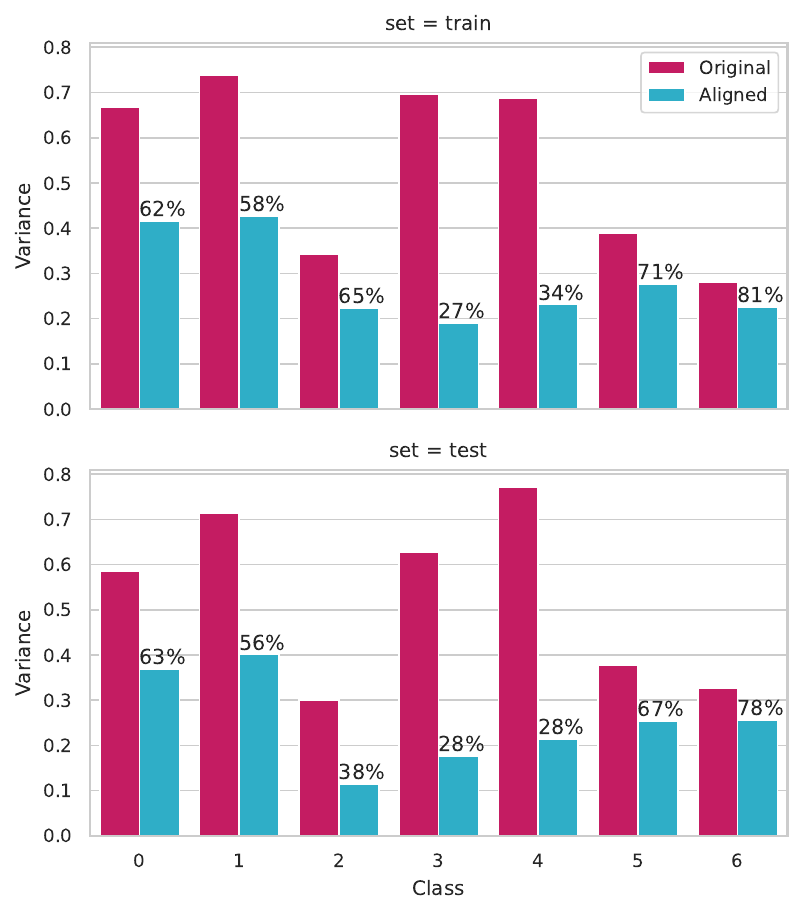}
        \caption{Lightning7 dataset}
    \end{subfigure}
    \caption{Within-class variances on several UCR archive datasets before \& after alignment.}
    \label{fig:alignment_variance_1}
\end{center}
\end{figure}

\renewcommand\x{0.32}
\begin{figure}[!htb]\ContinuedFloat
    \begin{center}
        \begin{subfigure}[t]{\x\linewidth}
            \centering
            \includegraphics[height=1.6\linewidth]{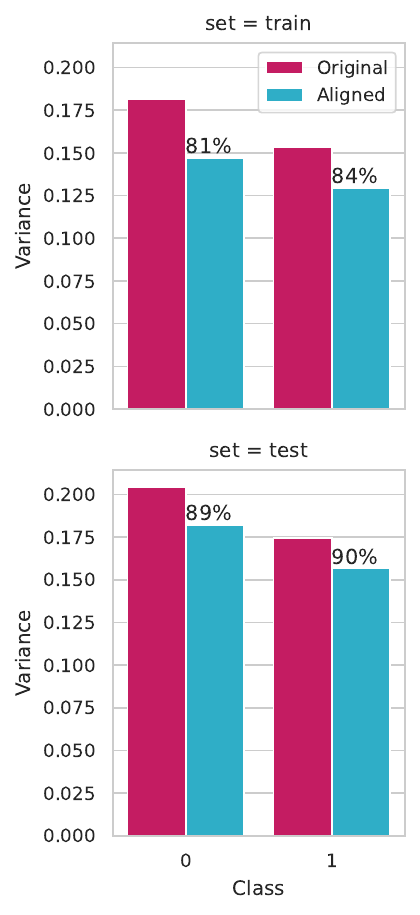}
            \caption{Ham dataset}
        \end{subfigure}
        \hfill
        \begin{subfigure}[t]{\x\linewidth}
            \centering
            \includegraphics[height=1.6\linewidth]{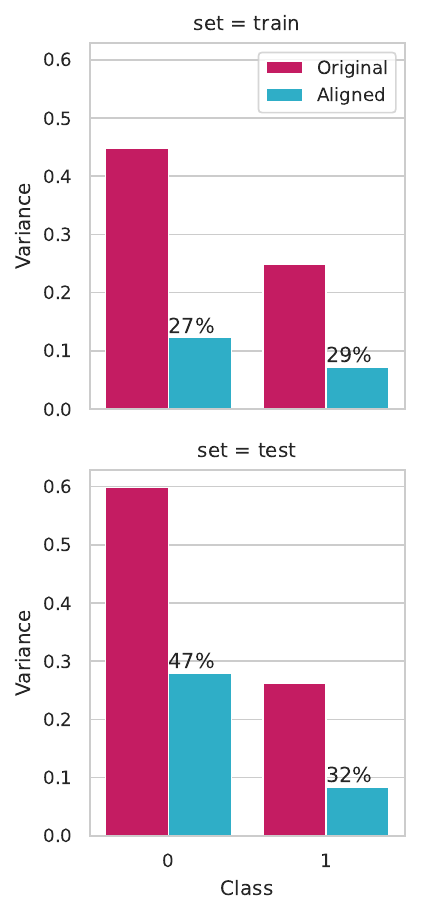}
            \caption{ECG200 dataset}
        \end{subfigure}
        \hfill
        \begin{subfigure}[t]{\x\linewidth}
            \centering
            \includegraphics[height=1.6\linewidth]{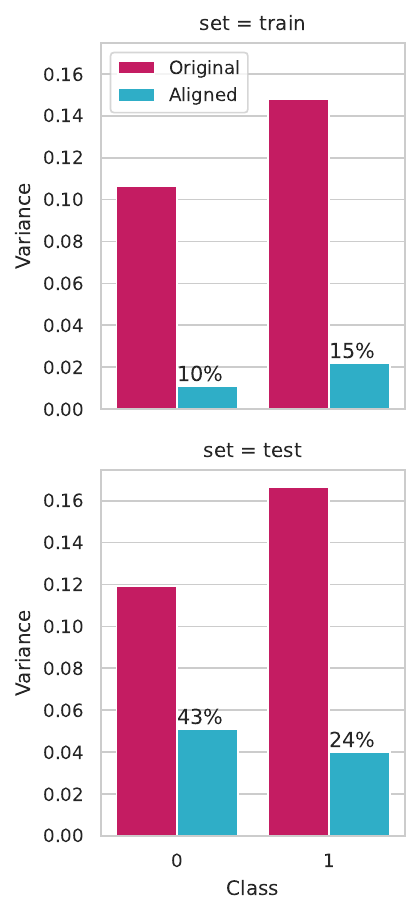}
            \caption{TwoLeadECG dataset}
        \end{subfigure}
        \\
        \begin{subfigure}[t]{\x\linewidth}
            \centering
            \includegraphics[height=1.6\linewidth]{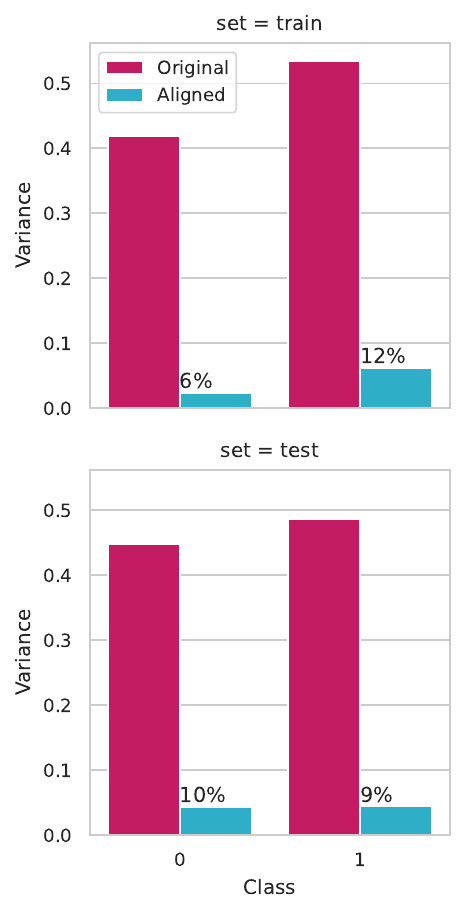}
            \caption{ECGFiveDays dataset}
        \end{subfigure}
        \hfill
        \begin{subfigure}[t]{\x\linewidth}
            \centering
            \includegraphics[height=1.6\linewidth]{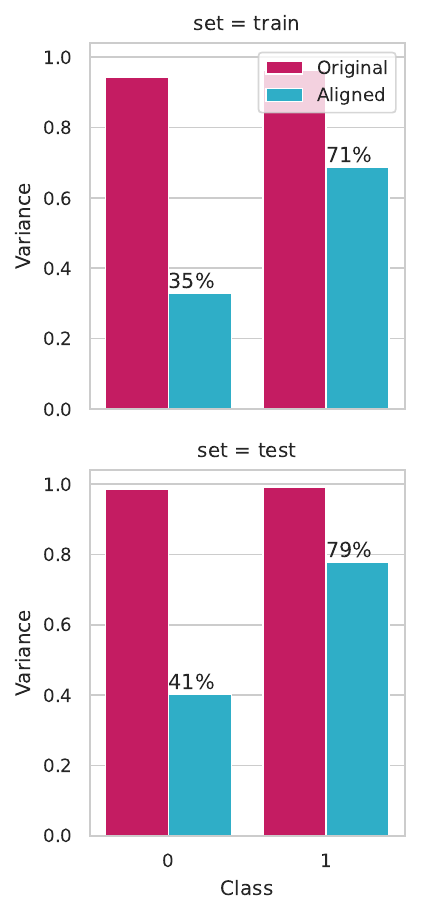}
            \caption{ToeSegmentation1 dataset}
        \end{subfigure}
        \hfill
        \begin{subfigure}[t]{\x\linewidth}
            \centering
            \includegraphics[height=1.6\linewidth]{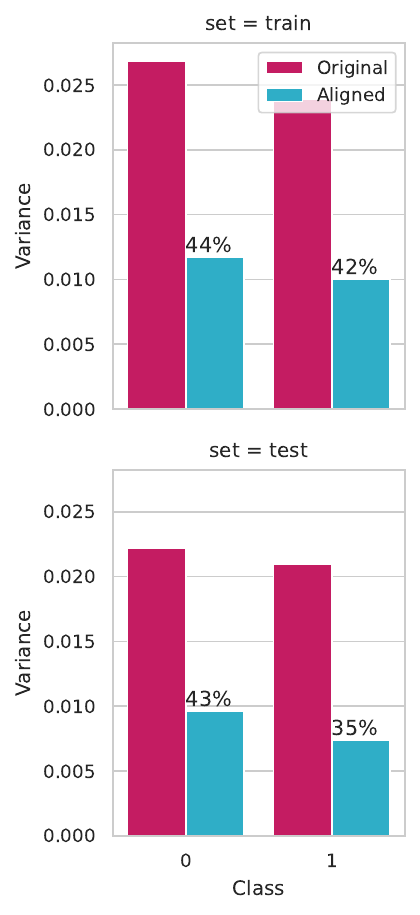}
            \caption{MiddlePhalanxOutlineCorrect dataset}
        \end{subfigure}
        \caption{Within-class variances on several UCR archive datasets before \& after alignment. (cont.)}
        \label{fig:alignment_variance_2}
    \end{center}
    \end{figure}

\begin{figure}[!htb]\ContinuedFloat
\begin{center}
    \begin{subfigure}{\linewidth}
        \centering
        \includegraphics[width=\linewidth]{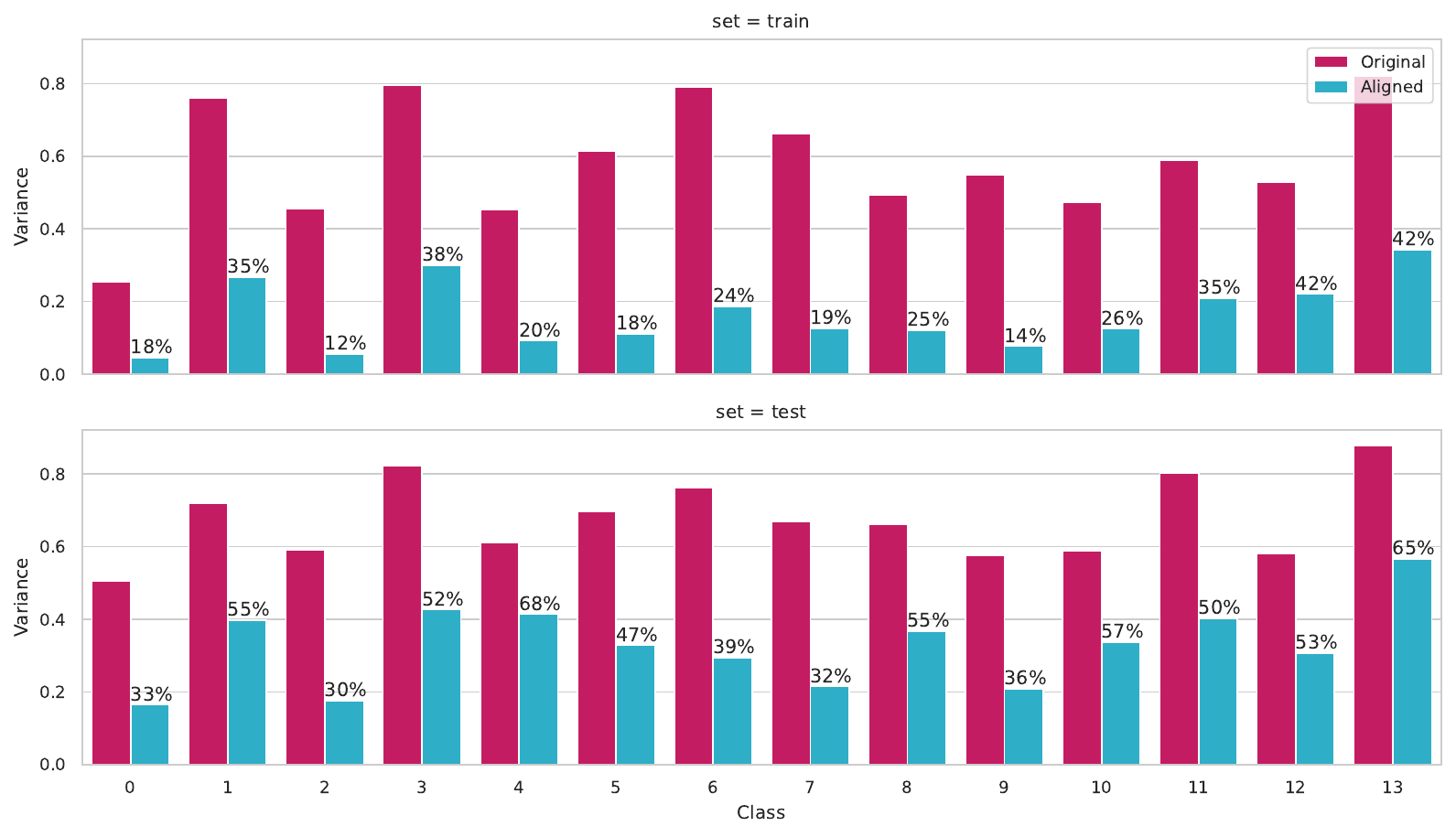}
        \caption{FacesUCR dataset}
    \end{subfigure}
    \\
    \begin{subfigure}{\linewidth}
        \centering
        \includegraphics[width=\linewidth]{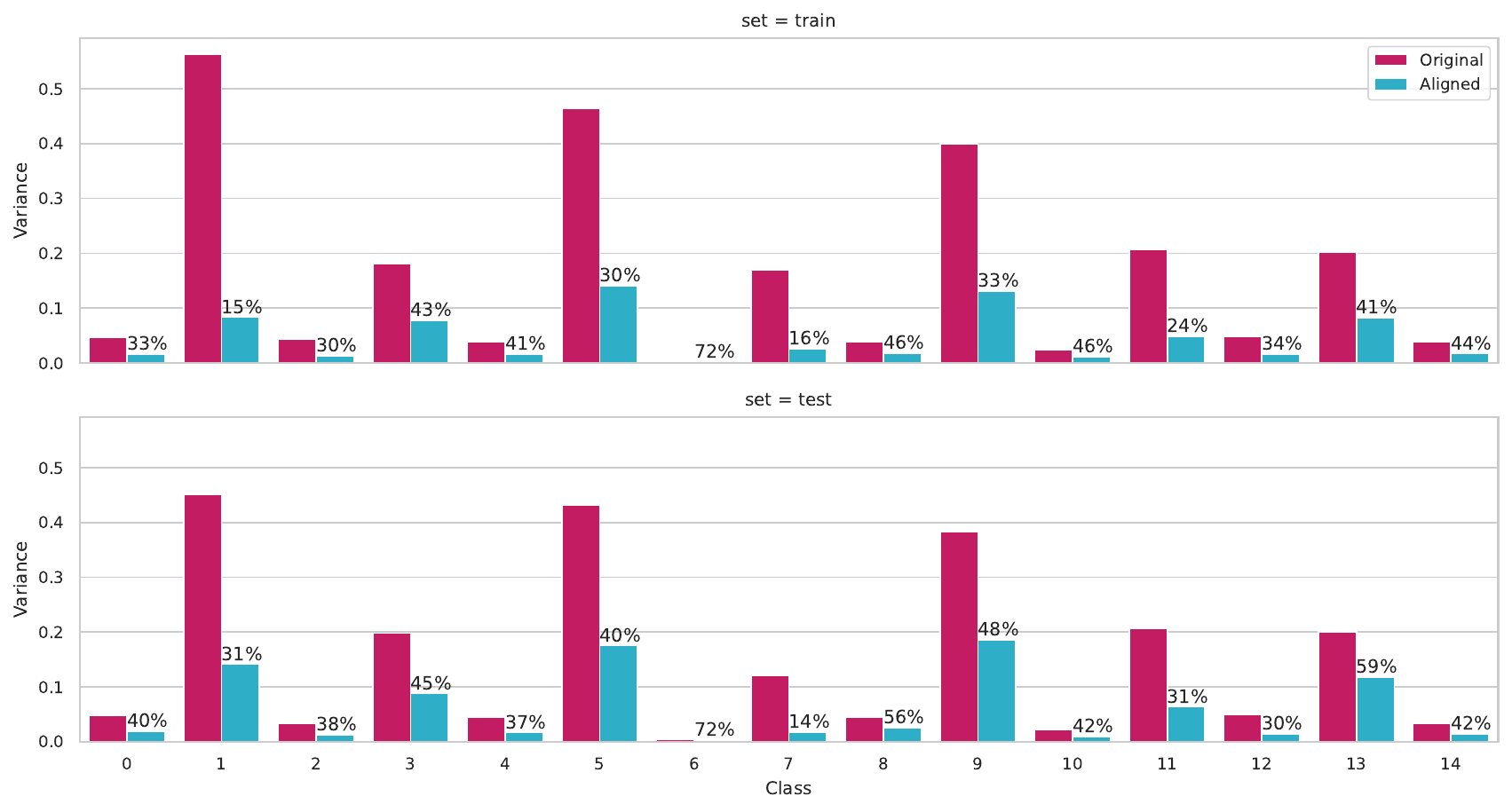}
        \caption{SwedishLeaf dataset}
    \end{subfigure}
    \caption{Within-class variances on several UCR archive datasets before \& after alignment. (cont.)}
    \label{fig:alignment_variance_3}
\end{center}
\end{figure}

\clearpage
\subsection{Extension to Multivariate Datasets}

This final section covers the generalization of the presented method to multivariate (also called multichannel) time series datasets: UCR Multivariate Dataset and MSRAction3D Human Activity Recognition. In both datasets, we apply the proposed Temporal Transformer Network (TTN) model (see \cref{sec:method_3}) and visualize the resulting average signal.

\subsubsection{UCR Multivariate Dataset}

It consists of 30 multivariate datasets of different time series lengths, channels, and number of classes.

\small
\renewcommand{\arraystretch}{0.8}
\begin{longtable}{lrrrrl}
\caption{UCR multivariate datasets \cite{dau2019ucr} information}\label{tab:ucr_dataset_table_multivariate}\\
\toprule
        Dataset &  Train &  Test &  Length &  Classes & Channels \\
\midrule\endfirsthead
\caption{UCR multivariate datasets \cite{dau2019ucr} information}\\
\toprule
        Dataset &  Train &  Test &  Length &  Classes & Channels \\
\midrule \endhead 
        ArticularyWordRecognition & 275   & 300   & 144   & 25 & 9    \\
        AtrialFibrillation        & 15    & 15    & 640   & 3  & 2    \\
        BasicMotions              & 40    & 40    & 100   & 4  & 6    \\
        CharacterTrajectories     & 1422  & 1436  & 182   & 20 & 3    \\
        Cricket                   & 108   & 72    & 1197  & 12 & 6    \\
        DuckDuckGeese             & 50    & 50    & 270   & 5  & 1345 \\
        EigenWorms                & 128   & 131   & 17984 & 5  & 6    \\
        Epilepsy                  & 137   & 138   & 206   & 4  & 3    \\
        EthanolConcentration      & 261   & 263   & 1751  & 4  & 3    \\
        ERing                     & 30    & 270   & 65    & 6  & 4    \\
        FaceDetection             & 5890  & 3524  & 62    & 2  & 144  \\
        FingerMovements           & 316   & 100   & 50    & 2  & 28   \\
        HandMovementDirection     & 160   & 74    & 400   & 4  & 10   \\
        Handwriting               & 150   & 850   & 152   & 26 & 3    \\
        Heartbeat                 & 204   & 205   & 405   & 2  & 61   \\
        InsectWingbeat            & 30000 & 20000 & 30    & 10 & 200  \\
        JapaneseVowels            & 270   & 370   & 29    & 9  & 12   \\
        Libras                    & 180   & 180   & 45    & 15 & 2    \\
        LSST                      & 2459  & 2466  & 36    & 14 & 6    \\
        MotorImagery              & 278   & 100   & 3000  & 2  & 64   \\
        NATOPS                    & 180   & 180   & 51    & 6  & 24   \\
        PenDigits                 & 7494  & 3498  & 8     & 10 & 2    \\
        PEMS-SF                   & 267   & 173   & 144   & 7  & 963  \\
        Phoneme                   & 3315  & 3353  & 217   & 39 & 11   \\
        RacketSports              & 151   & 152   & 30    & 4  & 6    \\
        SelfRegulationSCP1        & 268   & 293   & 896   & 2  & 6    \\
        SelfRegulationSCP2        & 200   & 180   & 1152  & 2  & 7    \\
        SpokenArabicDigits        & 6599  & 2199  & 93    & 10 & 13   \\
        StandWalkJump             & 12    & 15    & 2500  & 3  & 4    \\
        UWaveGestureLibrary       & 120   & 320   & 315   & 8  & 3    \\
\bottomrule
\end{longtable}
\renewcommand{\arraystretch}{1}
\normalsize

\paragraph{Within-class Variance Reduction}

This statistic provides an idea of how much temporal misalignment has been repaired. There is a clear shift in the within-class variance distribution.

\begin{figure}[!htb]
    \begin{center}
    \begin{subfigure}[t]{\linewidth}
        \centering
        \includegraphics[width=\linewidth]{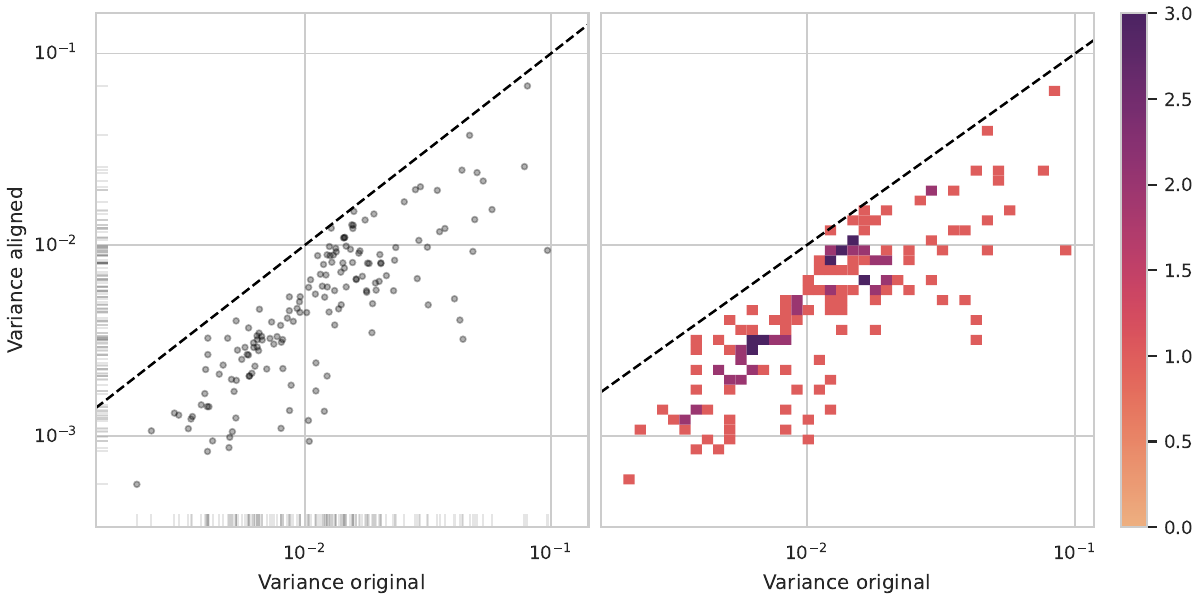}
        \caption{Time series variance before and after alignment. \textbf{Left}: scatterplot, each point represents a (dataset, class) tuple. \textbf{Right}: 2D Histogram, counts the number of (dataset, class) tuples in each bin.}
    \end{subfigure}
    \\
    \begin{subfigure}[t]{0.48\linewidth}
        \centering
        \includegraphics[width=\linewidth]{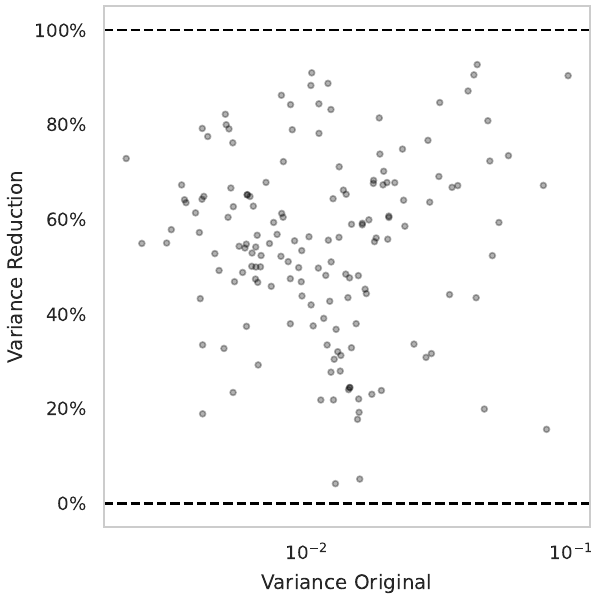}
        \caption{Reduction relative to original variance.}
    \end{subfigure}
    \hfill
    \begin{subfigure}[t]{0.49\linewidth}
        \centering
        \includegraphics[width=0.95\linewidth, trim=0 0 300 0, clip]{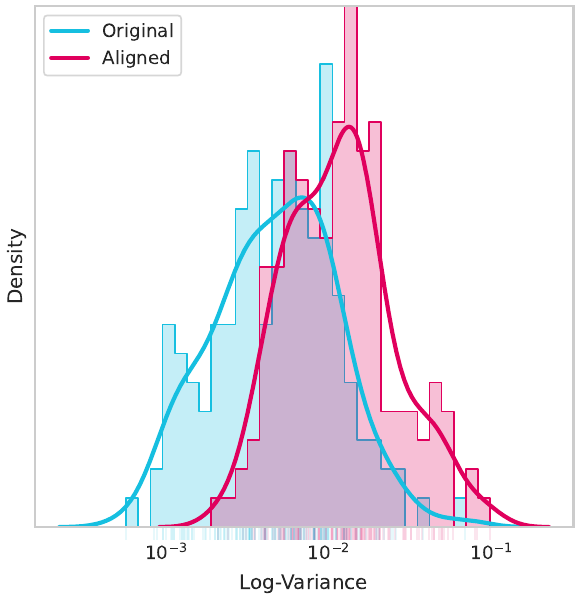}
        \caption{Density of variance before and after alignment.}
    \end{subfigure}
    \caption{Study of variance reduction before and after time series alignment. Multivariate UCR datasets \cite{dau2019ucr}.}
    \label{fig:variance_multivariate}
\end{center}
\end{figure}

\paragraph{Qualitative Alignment Results}

We report qualitative results in \cref{fig:alignment_example_2,fig:alignment_example_3,fig:alignment_example_4,fig:alignment_example_5}. Each figure corresponds to a different dataset, as the number of channels varies between them. 

\renewcommand\x{0.32}
\begin{figure}[!htb]
    \begin{center}
        \begin{subfigure}{\x\linewidth}
            \centering
            \includegraphics[width=\linewidth, trim=32 262 0 20, clip]{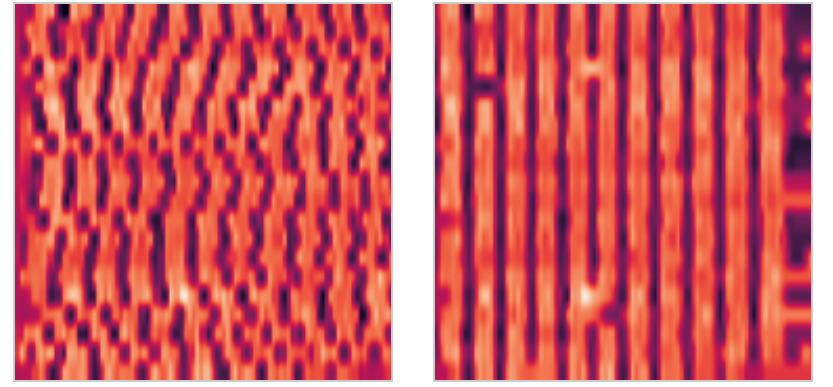}\\
            \includegraphics[width=\linewidth, trim=32 35 0 240, clip]{figures/ucr_dataset_multivariate/alignment/BasicMotions_1_0.pdf}
            \caption{Channel 0}
        \end{subfigure}
        \begin{subfigure}{\x\linewidth}
            \centering
            \includegraphics[width=\linewidth, trim=32 262 0 20, clip]{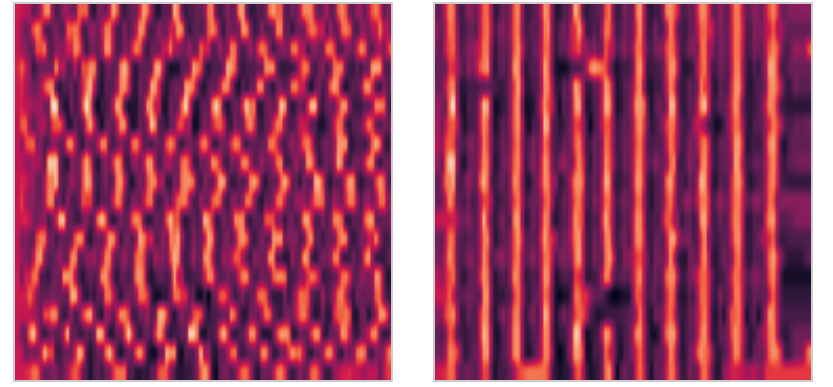}\\
            \includegraphics[width=\linewidth, trim=32 35 0 240, clip]{figures/ucr_dataset_multivariate/alignment/BasicMotions_1_1.pdf}
            \caption{Channel 1}
        \end{subfigure}
        \begin{subfigure}{\x\linewidth}
            \centering
            \includegraphics[width=\linewidth, trim=32 262 0 20, clip]{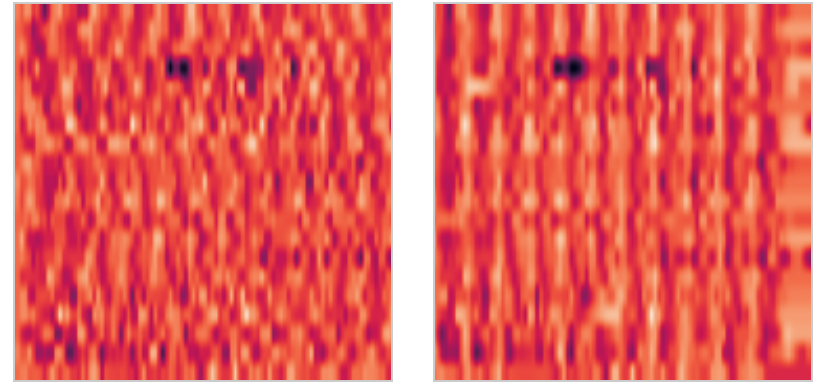}\\
            \includegraphics[width=\linewidth, trim=32 35 0 240, clip]{figures/ucr_dataset_multivariate/alignment/BasicMotions_1_2.pdf}
            \caption{Channel 2}
        \end{subfigure}
        \\
        \begin{subfigure}{\x\linewidth}
            \centering
            \includegraphics[width=\linewidth, trim=32 262 0 20, clip]{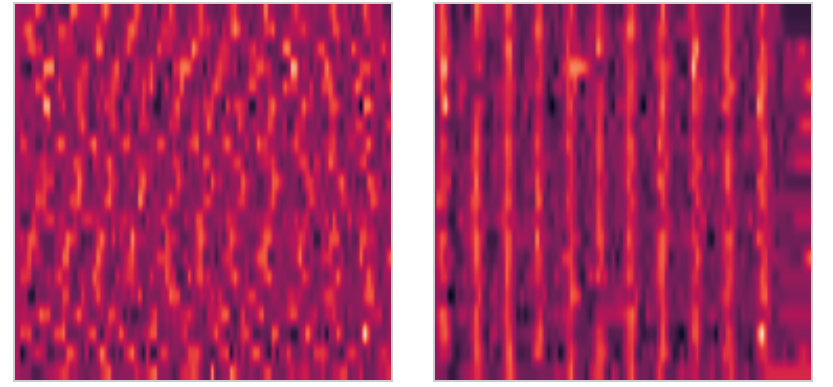}\\
            \includegraphics[width=\linewidth, trim=32 35 0 240, clip]{figures/ucr_dataset_multivariate/alignment/BasicMotions_1_3.pdf}
            \caption{Channel 3}
        \end{subfigure}
        \begin{subfigure}{\x\linewidth}
            \centering
            \includegraphics[width=\linewidth, trim=32 262 0 20, clip]{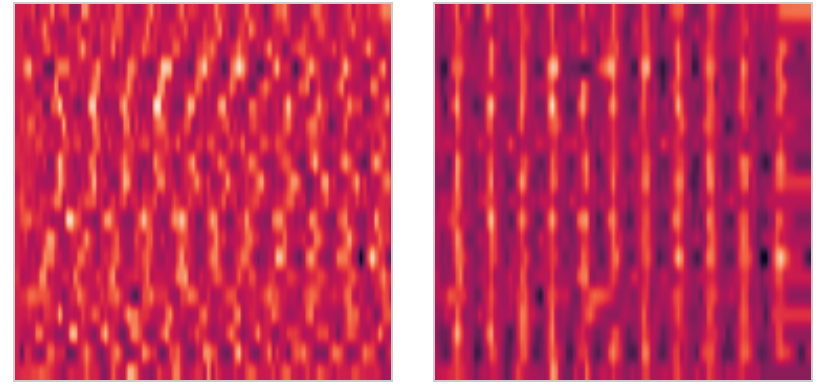}\\
            \includegraphics[width=\linewidth, trim=32 35 0 240, clip]{figures/ucr_dataset_multivariate/alignment/BasicMotions_1_4.pdf}
            \caption{Channel 4}
        \end{subfigure}
        \begin{subfigure}{\x\linewidth}
            \centering
            \includegraphics[width=\linewidth, trim=32 262 0 20, clip]{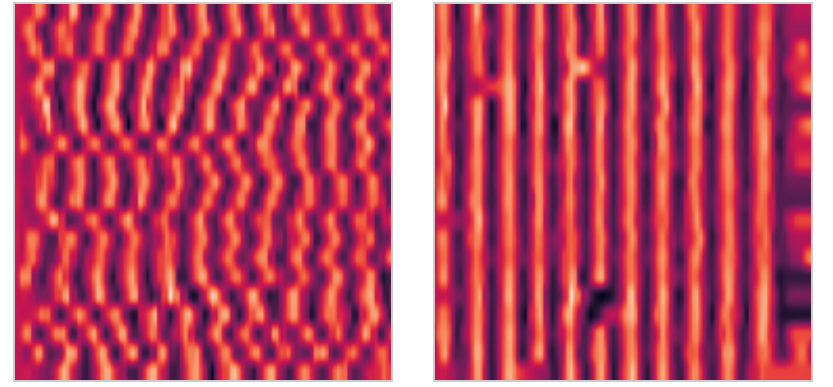}\\
            \includegraphics[width=\linewidth, trim=32 35 0 240, clip]{figures/ucr_dataset_multivariate/alignment/BasicMotions_1_5.pdf}
            \caption{Channel 5}
        \end{subfigure}
        \caption{Multi-class time series alignment on BasicMotions test set, class 1. \textbf{Top}: heatmap of each time series sample (row). \textbf{Bottom}: overlapping time series, red line represents Euclidean average. \textbf{Left}: original signals. \textbf{Right}: signals after alignment.}
        \label{fig:alignment_example_2}
    \end{center}
\end{figure}

\renewcommand\x{0.32}
\begin{figure}[!htb]
    \begin{center}
        \begin{subfigure}{\x\linewidth}
            \centering
            \includegraphics[width=\linewidth, trim=35 262 0 22, clip]{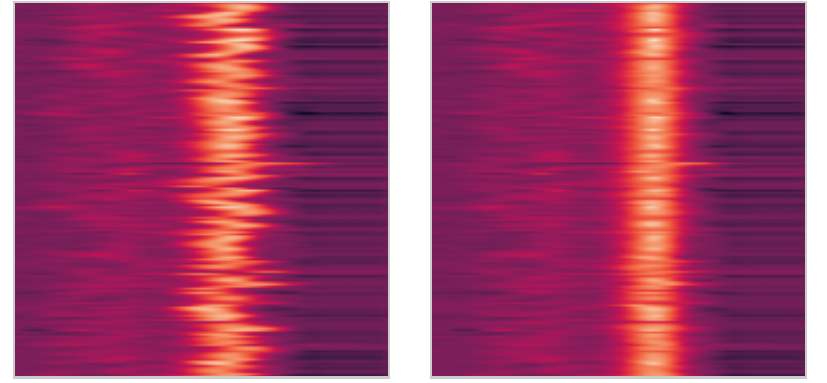}\\
            \includegraphics[width=\linewidth, trim=35 35 0 240, clip]{figures/ucr_dataset_multivariate/alignment/CharacterTrajectories_8_0.pdf}
            \caption{Channel 0}
        \end{subfigure}
        \begin{subfigure}{\x\linewidth}
            \centering
            \includegraphics[width=\linewidth, trim=35 262 0 22, clip]{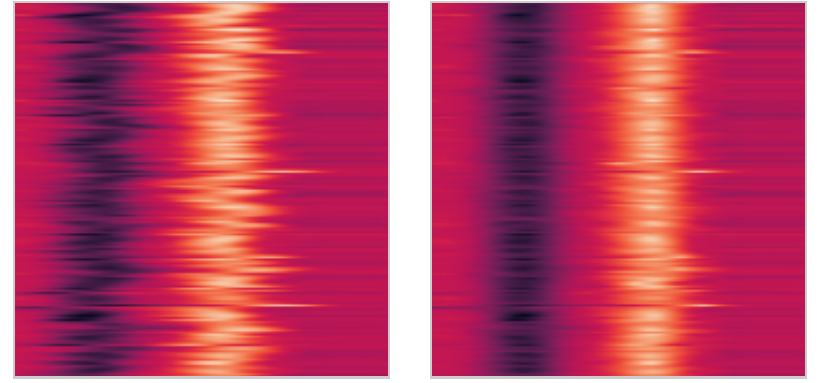}\\
            \includegraphics[width=\linewidth, trim=35 35 0 240, clip]{figures/ucr_dataset_multivariate/alignment/CharacterTrajectories_8_1.pdf}
            \caption{Channel 1}
        \end{subfigure}
        \begin{subfigure}{\x\linewidth}
            \centering
            \includegraphics[width=\linewidth, trim=35 262 0 22, clip]{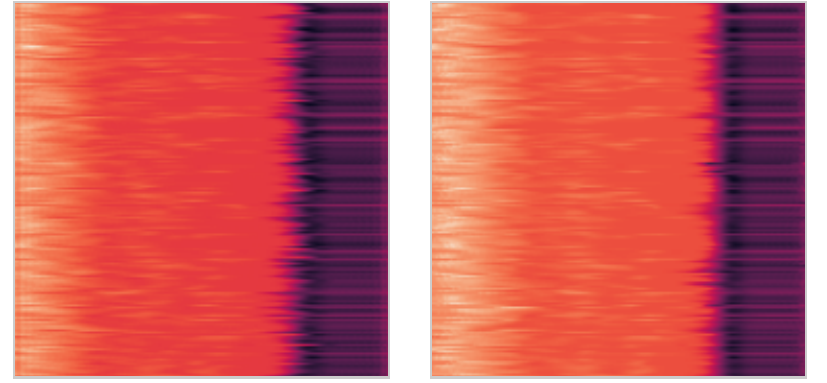}\\
            \includegraphics[width=\linewidth, trim=35 35 0 240, clip]{figures/ucr_dataset_multivariate/alignment/CharacterTrajectories_8_2.pdf}
            \caption{Channel 2}
        \end{subfigure}
        \caption{Multi-class time series alignment on CharacterTrajectories test set, class 8.}
        \label{fig:alignment_example_4}
    \end{center}
\end{figure}

\renewcommand\x{0.32}
\begin{figure}[!htb]
    \begin{center}
        \begin{subfigure}{\x\linewidth}
            \centering
            \includegraphics[width=\linewidth, trim=35 262 0 22, clip]{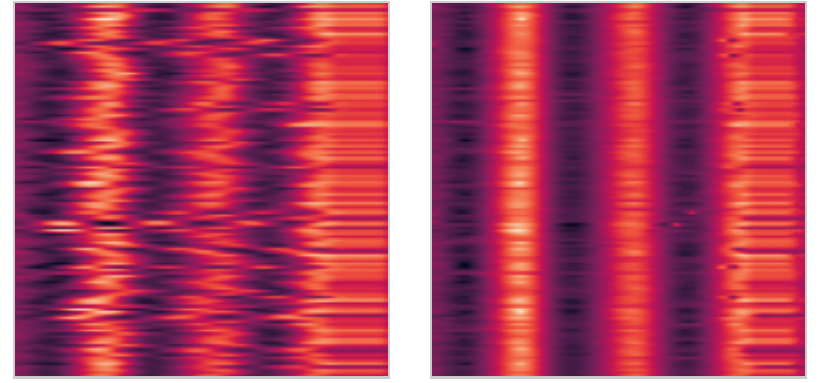}\\
            \includegraphics[width=\linewidth, trim=35 35 0 240, clip]{figures/ucr_dataset_multivariate/alignment/CharacterTrajectories_19_0.pdf}
            \caption{Channel 0}
        \end{subfigure}
        \begin{subfigure}{\x\linewidth}
            \centering
            \includegraphics[width=\linewidth, trim=35 262 0 22, clip]{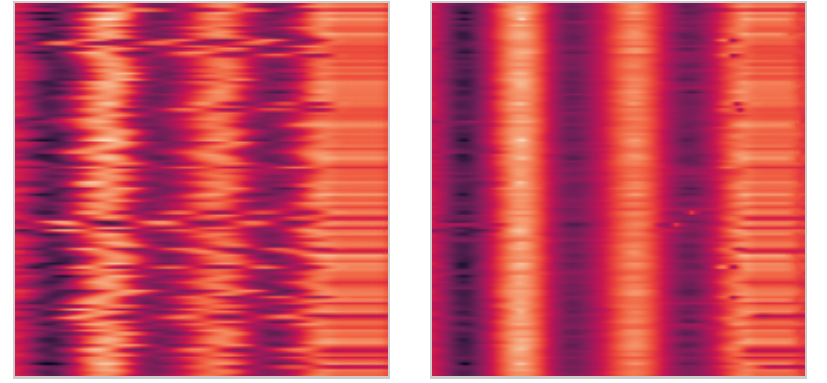}\\
            \includegraphics[width=\linewidth, trim=35 35 0 240, clip]{figures/ucr_dataset_multivariate/alignment/CharacterTrajectories_19_1.pdf}
            \caption{Channel 1}
        \end{subfigure}
        \begin{subfigure}{\x\linewidth}
            \centering
            \includegraphics[width=\linewidth, trim=35 262 0 22, clip]{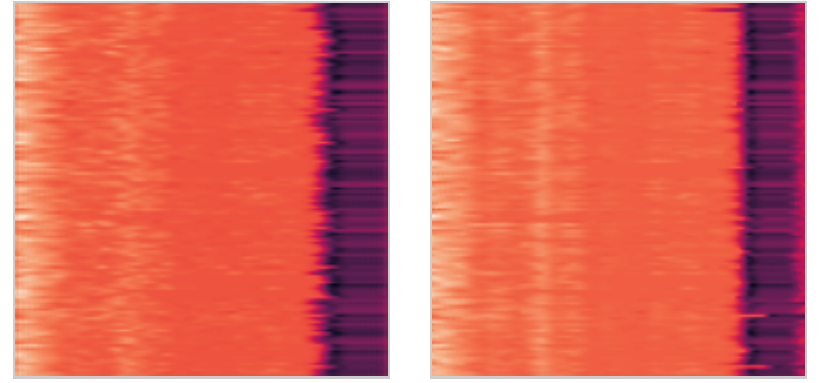}\\
            \includegraphics[width=\linewidth, trim=35 35 0 240, clip]{figures/ucr_dataset_multivariate/alignment/CharacterTrajectories_19_2.pdf}
            \caption{Channel 2}
        \end{subfigure}
        \caption{Multi-class time series alignment on CharacterTrajectories test set, class 19. \textbf{Top}: heatmap of each time series sample (row). \textbf{Bottom}: overlapping time series, red line represents Euclidean average. \textbf{Left}: original signals. \textbf{Right}: signals after alignment.}
        \label{fig:alignment_example_5}
    \end{center}
\end{figure}

\renewcommand\x{0.32}
\begin{figure}[!htb]
    \begin{center}
        \begin{subfigure}{\x\linewidth}
            \centering
            \includegraphics[width=\linewidth, trim=32 262 0 20, clip]{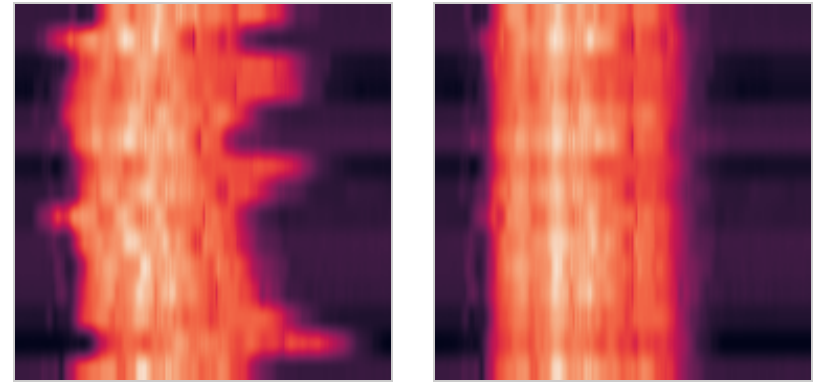}\\
            \includegraphics[width=\linewidth, trim=32 35 0 240, clip]{figures/ucr_dataset_multivariate/alignment/Cricket_2_0.pdf}
            \caption{Channel 0}
        \end{subfigure}
        \begin{subfigure}{\x\linewidth}
            \centering
            \includegraphics[width=\linewidth, trim=32 262 0 20, clip]{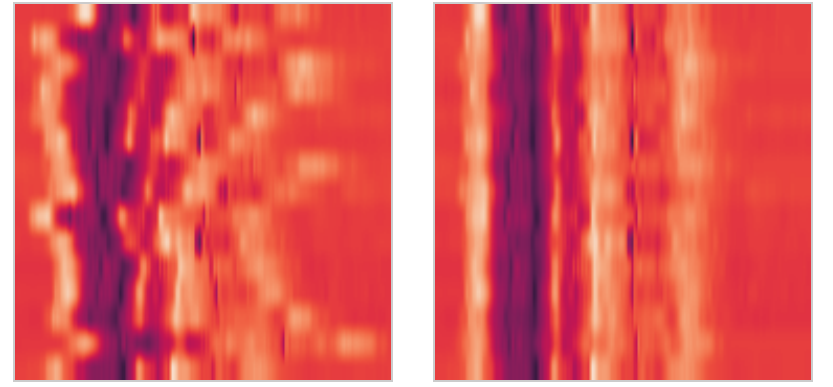}\\
            \includegraphics[width=\linewidth, trim=32 35 0 240, clip]{figures/ucr_dataset_multivariate/alignment/Cricket_2_1.pdf}
            \caption{Channel 1}
        \end{subfigure}
        \begin{subfigure}{\x\linewidth}
            \centering
            \includegraphics[width=\linewidth, trim=32 262 0 20, clip]{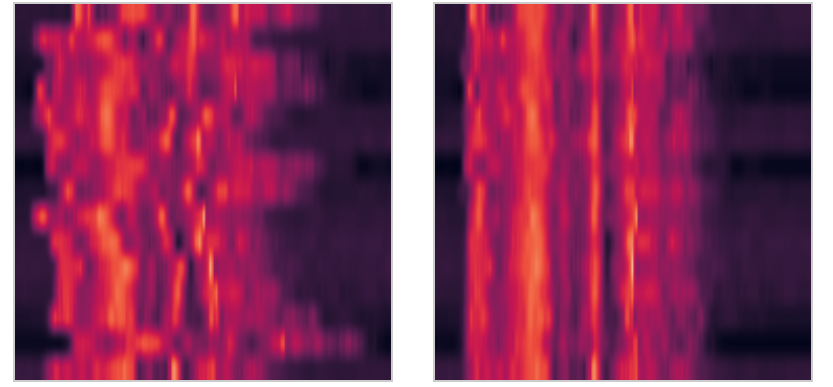}\\
            \includegraphics[width=\linewidth, trim=32 35 0 240, clip]{figures/ucr_dataset_multivariate/alignment/Cricket_2_2.pdf}
            \caption{Channel 2}
        \end{subfigure}
        \\
        \begin{subfigure}{\x\linewidth}
            \centering
            \includegraphics[width=\linewidth, trim=32 262 0 20, clip]{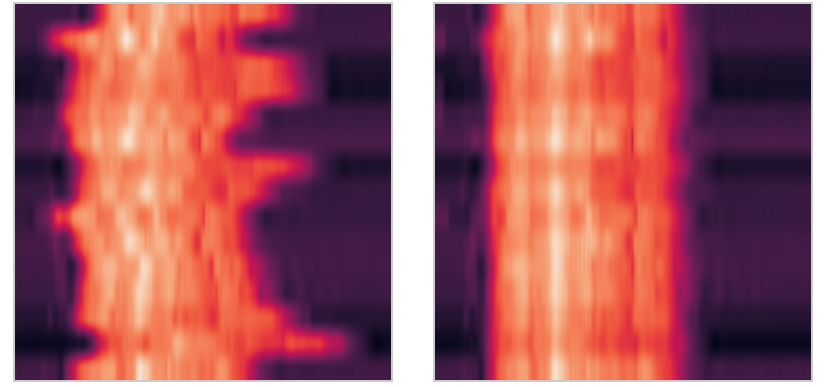}\\
            \includegraphics[width=\linewidth, trim=32 35 0 240, clip]{figures/ucr_dataset_multivariate/alignment/Cricket_2_3.pdf}
            \caption{Channel 3}
        \end{subfigure}
        \begin{subfigure}{\x\linewidth}
            \centering
            \includegraphics[width=\linewidth, trim=32 262 0 20, clip]{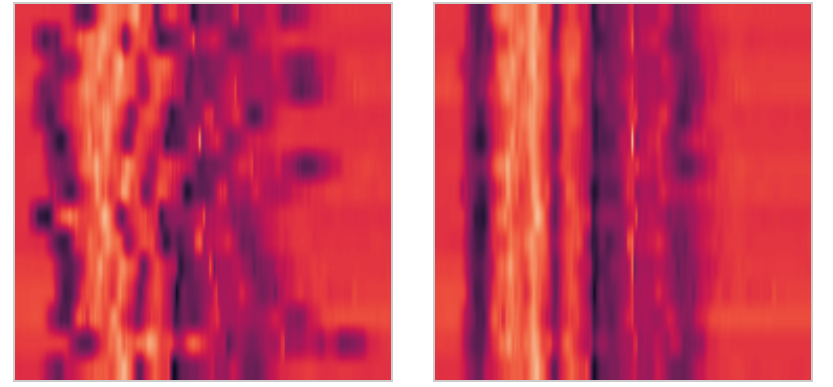}\\
            \includegraphics[width=\linewidth, trim=32 35 0 240, clip]{figures/ucr_dataset_multivariate/alignment/Cricket_2_4.pdf}
            \caption{Channel 4}
        \end{subfigure}
        \begin{subfigure}{\x\linewidth}
            \centering
            \includegraphics[width=\linewidth, trim=32 262 0 20, clip]{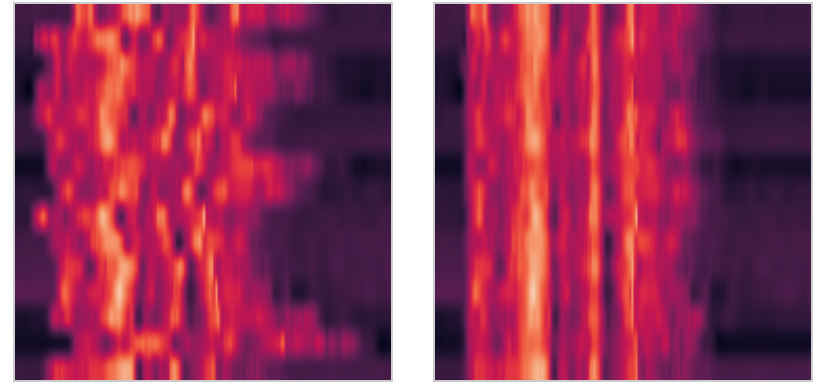}\\
            \includegraphics[width=\linewidth, trim=32 35 0 240, clip]{figures/ucr_dataset_multivariate/alignment/Cricket_2_5.pdf}
            \caption{Channel 5}
        \end{subfigure}
        \caption{Multi-class time series alignment on Cricket test set, class 2. \textbf{Top}: heatmap of each time series sample (row). \textbf{Bottom}: overlapping time series, red line represents Euclidean average. \textbf{Left}: original signals. \textbf{Right}: signals after alignment.}
        \label{fig:alignment_example_3}
    \end{center}
\end{figure}

\clearpage
\subsubsection{Human Action 3D Dataset}


This dataset studies the recognition of human actions from sequences of depth maps. The human body is an articulated system of rigid segments connected by joints and human motion is often considered as a continuous evolution of the spatial configuration of the segments or body posture. Given a sequence of depth maps, if the body joints can be reliably extracted and tracked, action recognition can be achieved by using the tracked joint positions. 

The MSRAction3D dataset \cite{li2010action} consists of 10 subjects, 20 3D joints (\textit{x,y,z}), and 20 different actions (see \cref{tab:action}). 
Each action was performed three times, and the subjects were facing the camera during the performance. Four real numbers are stored for each joint: \textit{x}, \textit{y}, \textit{z}, \textit{c}; where (\textit{x},\textit{y}) are screen coordinates, \textit{z} is the depth value, and \textit{c} is the confidence score. The depth maps were captured at 15 frames per second by a depth camera that acquires the depth through structure infra-red light. The size of the depth map is $640\times480$. Altogether, the dataset has 23797 frames of depth maps for the 4020 action samples.

\begin{figure}[!htb]
    \begin{center}
        \begin{subfigure}[b]{0.45\linewidth}
            \centering
            \includegraphics[width=0.9\linewidth]{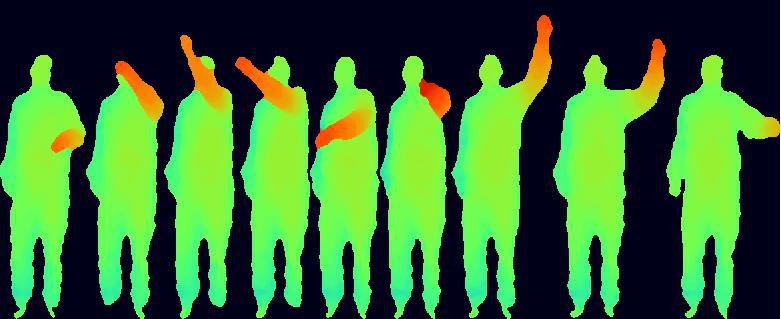}
            \includegraphics[width=0.9\linewidth]{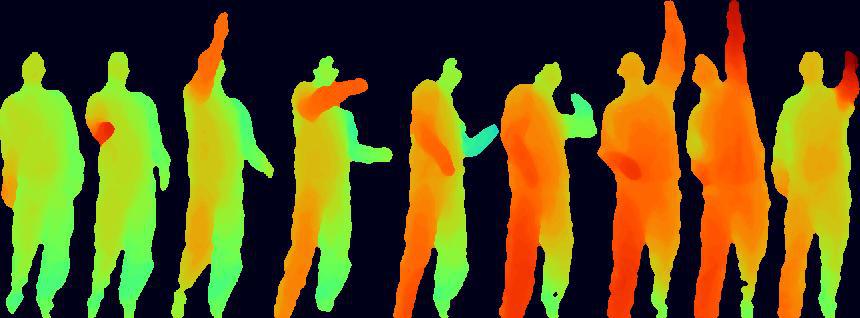}
            \caption{Example of the sequences of depth maps for "\textit{Draw Tick}" and "\textit{Tennis Serve}" actions. Adapted with permission from \cite{li2010action}}
        \end{subfigure}
        \hfill
        \begin{subfigure}[b]{0.4\linewidth}
            \centering
            \includegraphics[width=\linewidth]{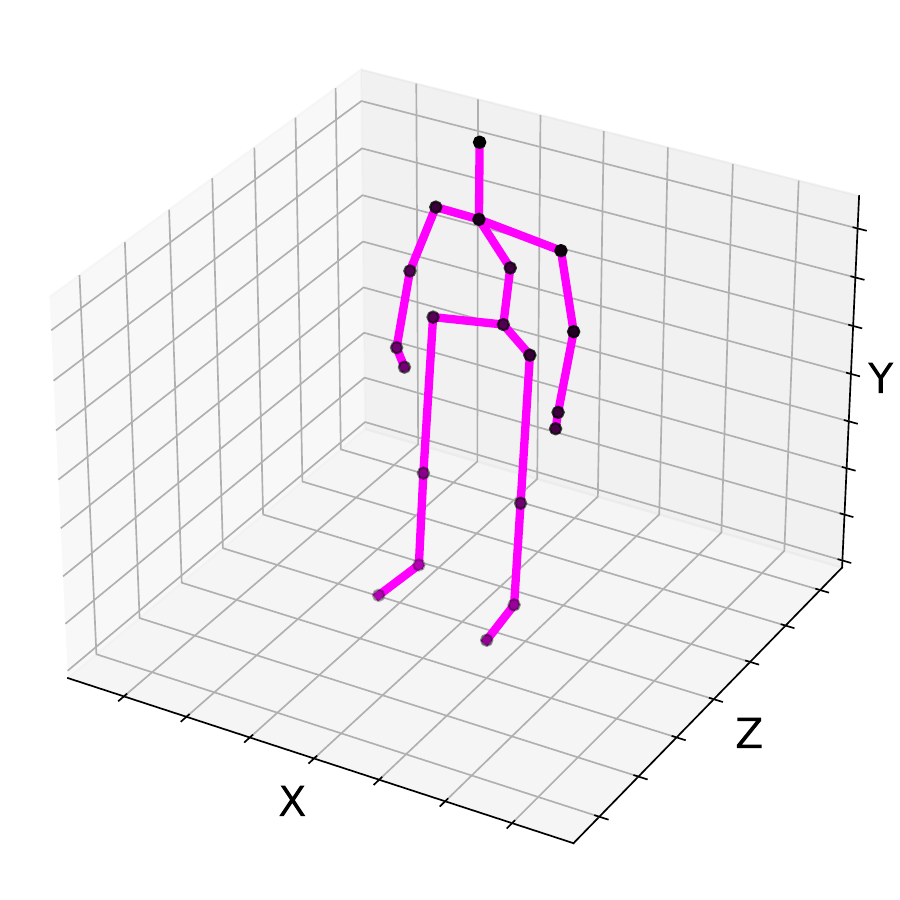}
            \caption{3D visualization of skeleton's joints and links.}
        \end{subfigure}
        \end{center}
        \caption{MSRAction3D dataset.}
    \label{fig:action}
\end{figure}

\cref{tab:action} shows the variance reduction after applying the proposed Temporal Transformer Network (TTN) to this dataset. We obtain reduction rates ranging from 12.0\% to 77.8 \%, and a 33.7\% variance reduction on average.
Furthermore, \cref{fig:human_action_1} shows a 3D representation of the skeleton joints before and after alignment, for the "\textit{Draw X}" action.

\begin{equation}
    \text{Variance Reduction} = 100 \times \frac{\text{Var}_{\text{aligned}} - \text{Var}_{\text{original}}}{\text{Var}_{\text{original}}}
\end{equation}

\clearpage

\begin{table}[!htb]
    \centering
    \caption{Within-class variance reduction in human action MSR 3D dataset.}
    \label{tab:action}
    \resizebox{0.8\textwidth}{!}{%
    \begin{tabular}{@{}llllll@{}}
        \toprule
        \textbf{Action}              & \textbf{Samples} & \textbf{Length} & & \textbf{Variance} & \\ 
        \cline{4-6}
        & & & Original & Aligned & Reduction \\
        \midrule
        Bend                & 30      & 255    & 1413.6   & 507.8    & 64.0\%   \\
        Draw Circle         & 30      & 58     & 86.5     & 73.2     & 15.4\%   \\
        Draw Tick           & 30      & 45     & 70.3     & 60.8     & 13.6\%   \\
        Draw X              & 28      & 67     & 200.4    & 89.6     & 55.2\%   \\
        Forward Kick        & 30      & 58     & 129.7    & 85.9     & 33.7\%   \\
        Forward Punch       & 26      & 76     & 114.9    & 85.5     & 25.5\%   \\
        Golf swing          & 30      & 71     & 159.7    & 118.0    & 26.0\%   \\
        Hammer              & 27      & 68     & 109.5    & 84.0     & 23.2\%   \\
        Hand Catch          & 26      & 100    & 126.6    & 94.5     & 25.2\%   \\
        High Arm Wave       & 27      & 67     & 108.1    & 90.2     & 16.4\%   \\
        High Throw          & 26      & 56     & 4166.2   & 1769.6   & 57.5\%   \\
        Horizontal Arm Wave & 27      & 66     & 95.1     & 83.6     & 12.0\%   \\
        Hand Clap           & 30      & 49     & 89.1     & 70.6     & 20.7\%   \\
        Jogging             & 30      & 58     & 94.1     & 78.4     & 16.7\%   \\
        Pick up \& Throw    & 30      & 71     & 609.2    & 230.6    & 62.1\%   \\
        Side-Boxing         & 30      & 61     & 117.3    & 92.6     & 21.0\%   \\
        Side Kick           & 20      & 56     & 76.9     & 61.8     & 19.6\%   \\
        Tennis Swing        & 30      & 57     & 136.2    & 95.2     & 30.1\%   \\
        Tennis Serve        & 30      & 69     & 316.3    & 132.3    & 58.1\%   \\
        Two Hand Wave       & 30      & 72     & 1400.3   & 309.8    & 77.8\%   \\ \bottomrule
        \end{tabular}%
    }
\end{table}

\renewcommand\x{0.035}
\begin{figure}[!htb]
  \begin{center}
    \begin{subfigure}{\linewidth}
      \includegraphics[width=\x\linewidth, trim=130 110 480 50, clip]{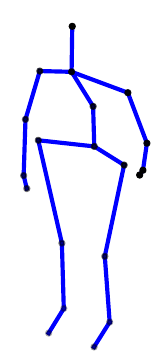}\hspace{-0.18em}
      \includegraphics[width=\x\linewidth, trim=130 110 480 50, clip]{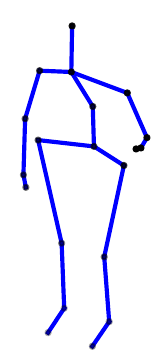}\hspace{-0.18em}
      \includegraphics[width=\x\linewidth, trim=130 110 480 50, clip]{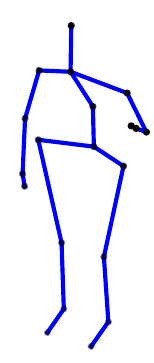}\hspace{-0.18em}
      \includegraphics[width=\x\linewidth, trim=130 110 480 50, clip]{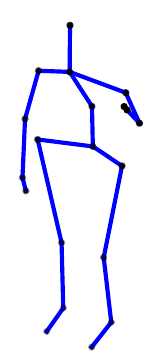}\hspace{-0.18em}
      \includegraphics[width=\x\linewidth, trim=130 110 480 50, clip]{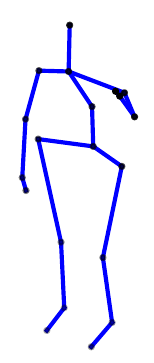}\hspace{-0.18em}
      \includegraphics[width=\x\linewidth, trim=130 110 480 50, clip]{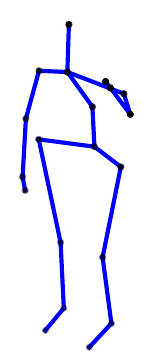}\hspace{-0.18em}
      \includegraphics[width=\x\linewidth, trim=130 110 480 50, clip]{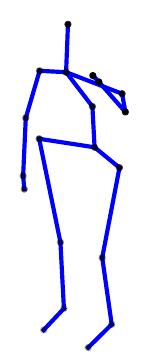}\hspace{-0.18em}
      \includegraphics[width=\x\linewidth, trim=130 110 480 50, clip]{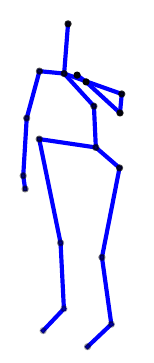}\hspace{-0.18em}
      \includegraphics[width=\x\linewidth, trim=130 110 480 50, clip]{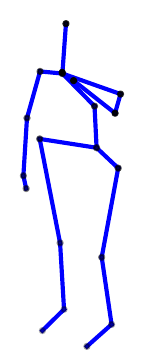}\hspace{-0.18em}
      \includegraphics[width=\x\linewidth, trim=130 110 480 50, clip]{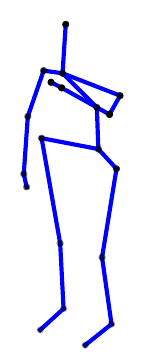}\hspace{-0.18em}
      \includegraphics[width=\x\linewidth, trim=130 110 480 50, clip]{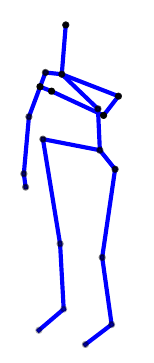}\hspace{-0.18em}
      \includegraphics[width=\x\linewidth, trim=130 110 480 50, clip]{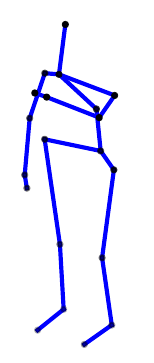}\hspace{-0.18em}
      \includegraphics[width=\x\linewidth, trim=130 110 480 50, clip]{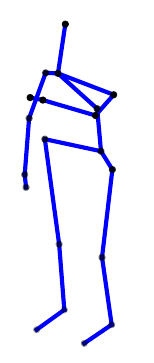}\hspace{-0.18em}
      \includegraphics[width=\x\linewidth, trim=130 110 480 50, clip]{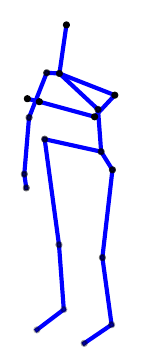}\hspace{-0.18em}
      \includegraphics[width=\x\linewidth, trim=130 110 480 50, clip]{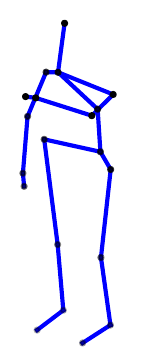}\hspace{-0.18em}
      \includegraphics[width=\x\linewidth, trim=130 110 480 50, clip]{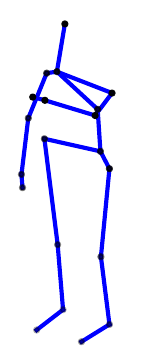}\hspace{-0.18em}
      \includegraphics[width=\x\linewidth, trim=130 110 480 50, clip]{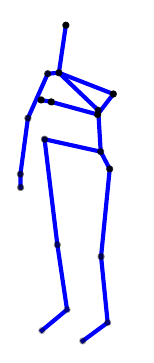}\hspace{-0.18em}
      \includegraphics[width=\x\linewidth, trim=130 110 480 50, clip]{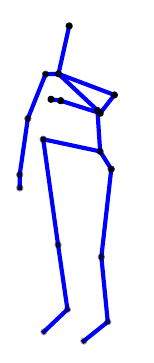}\hspace{-0.18em}
      \includegraphics[width=\x\linewidth, trim=130 110 480 50, clip]{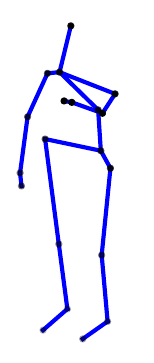}\hspace{-0.18em}
      \includegraphics[width=\x\linewidth, trim=130 110 480 50, clip]{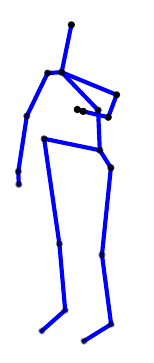}\hspace{-0.18em}
      \includegraphics[width=\x\linewidth, trim=130 110 480 50, clip]{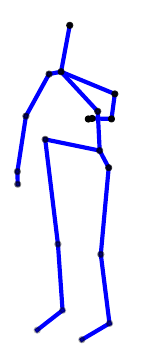}\hspace{-0.18em}
      \includegraphics[width=\x\linewidth, trim=130 110 480 50, clip]{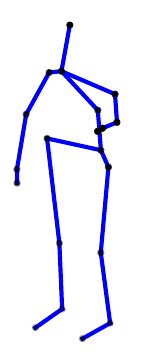}\hspace{-0.18em}
      \includegraphics[width=\x\linewidth, trim=130 110 480 50, clip]{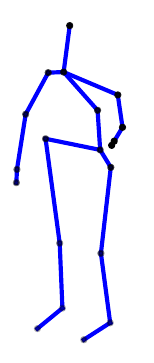}\hspace{-0.18em}
      \includegraphics[width=\x\linewidth, trim=130 110 480 50, clip]{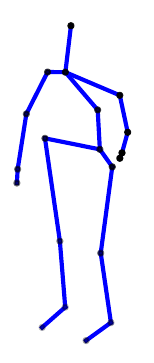}\hspace{-0.18em}
      \includegraphics[width=\x\linewidth, trim=130 110 480 50, clip]{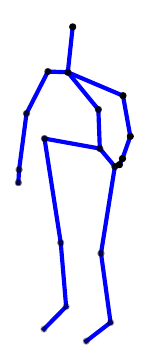}\hspace{-0.18em}
      \includegraphics[width=\x\linewidth, trim=130 110 480 50, clip]{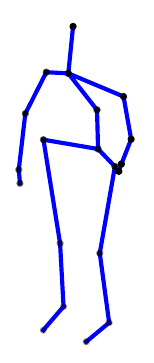}\hspace{-0.18em}
      \includegraphics[width=\x\linewidth, trim=130 110 480 50, clip]{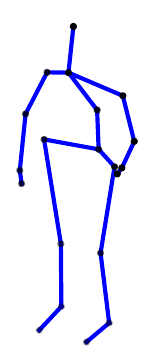}\hspace{-0.18em}
      \includegraphics[width=\x\linewidth, trim=130 110 480 50, clip]{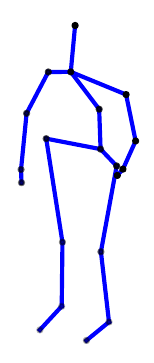}\hspace{-0.18em}
      \includegraphics[width=\x\linewidth, trim=130 110 480 50, clip]{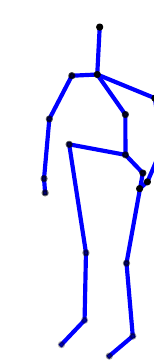}\hspace{-0.18em}
      \includegraphics[width=\x\linewidth, trim=130 110 480 50, clip]{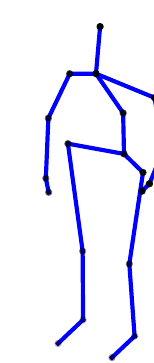}\hspace{-0.18em}
      \includegraphics[width=\x\linewidth, trim=130 110 480 50, clip]{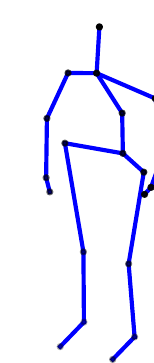}\hspace{-0.18em}
      \includegraphics[width=\x\linewidth, trim=130 110 480 50, clip]{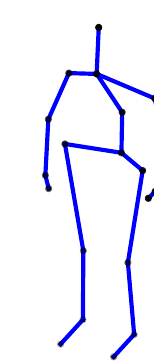}\hspace{-0.18em}
      \includegraphics[width=\x\linewidth, trim=130 110 480 50, clip]{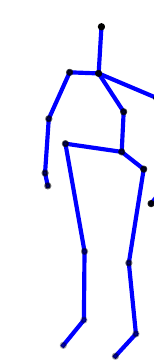}\hspace{-0.18em}
      \includegraphics[width=\x\linewidth, trim=130 110 480 50, clip]{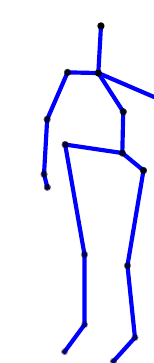}\hspace{-0.18em}
      \includegraphics[width=\x\linewidth, trim=130 110 480 50, clip]{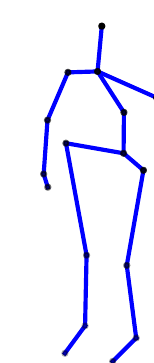}\hspace{-0.18em}
      \includegraphics[width=\x\linewidth, trim=130 110 480 50, clip]{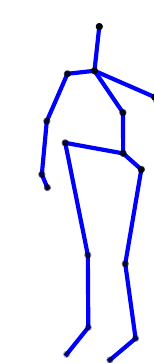}\hspace{-0.18em}
      \includegraphics[width=\x\linewidth, trim=130 110 480 50, clip]{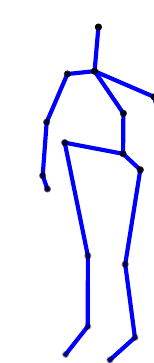}\hspace{-0.18em}
      \includegraphics[width=\x\linewidth, trim=130 110 480 50, clip]{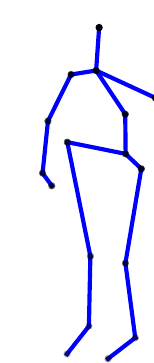}\hspace{-0.18em}
      \includegraphics[width=\x\linewidth, trim=130 110 480 50, clip]{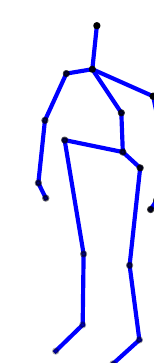}\hspace{-0.18em}
      \includegraphics[width=\x\linewidth, trim=130 110 480 50, clip]{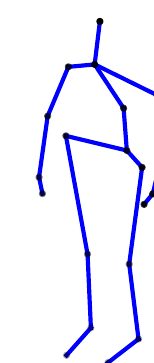}\hspace{-0.18em}
      \includegraphics[width=\x\linewidth, trim=130 110 480 50, clip]{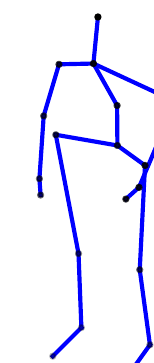}\hspace{-0.18em}
      \includegraphics[width=\x\linewidth, trim=130 110 480 50, clip]{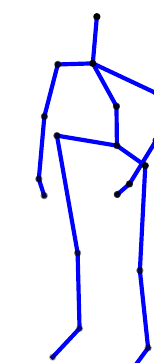}\hspace{-0.18em}
      \includegraphics[width=\x\linewidth, trim=130 110 480 50, clip]{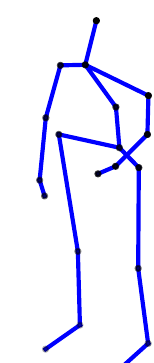}\hspace{-0.18em}
      \includegraphics[width=\x\linewidth, trim=130 110 480 50, clip]{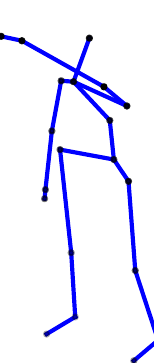}\hspace{-0.18em}
      \includegraphics[width=\x\linewidth, trim=130 110 480 50, clip]{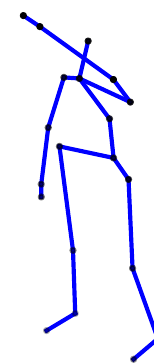}\hspace{-0.18em}
      \includegraphics[width=\x\linewidth, trim=130 110 480 50, clip]{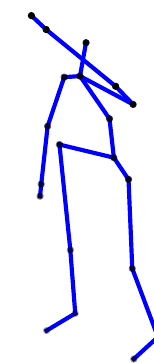}\hspace{-0.18em}
      \includegraphics[width=\x\linewidth, trim=130 110 480 50, clip]{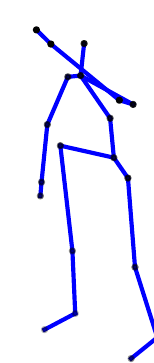}\hspace{-0.18em}
      \includegraphics[width=\x\linewidth, trim=130 110 480 50, clip]{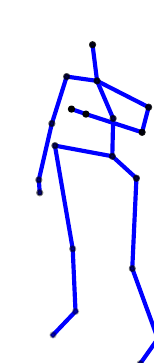}\hspace{-0.18em}
      \includegraphics[width=\x\linewidth, trim=130 110 480 50, clip]{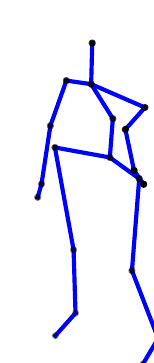}\hspace{-0.18em}
      \includegraphics[width=\x\linewidth, trim=130 110 480 50, clip]{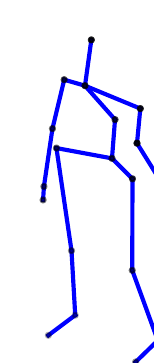}\hspace{-0.18em}
      \includegraphics[width=\x\linewidth, trim=130 110 480 50, clip]{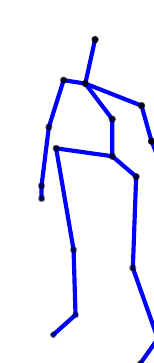}\hspace{-0.18em}
      \includegraphics[width=\x\linewidth, trim=130 110 480 50, clip]{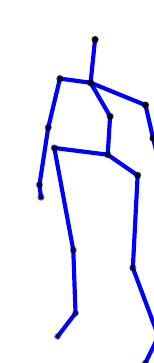}\hspace{-0.18em}
      \includegraphics[width=\x\linewidth, trim=130 110 480 50, clip]{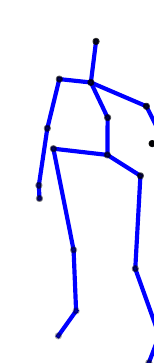}\hspace{-0.18em}
      \includegraphics[width=\x\linewidth, trim=130 110 480 50, clip]{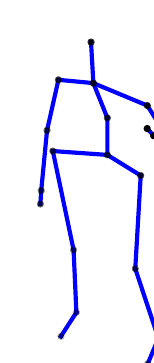}\hspace{-0.18em}
    \caption{Euclidean average on original data.}
    \end{subfigure}
    \begin{subfigure}{\linewidth}
      \includegraphics[width=\x\linewidth, trim=500 110 110  50, clip]{figures/action_draw_x/action_8.pdf}\hspace{-0.18em}
      \includegraphics[width=\x\linewidth, trim=500 110 110  50, clip]{figures/action_draw_x/action_9.pdf}\hspace{-0.18em}
      \includegraphics[width=\x\linewidth, trim=500 110 110  50, clip]{figures/action_draw_x/action_10.pdf}\hspace{-0.18em}
      \includegraphics[width=\x\linewidth, trim=500 110 110  50, clip]{figures/action_draw_x/action_11.pdf}\hspace{-0.18em}
      \includegraphics[width=\x\linewidth, trim=500 110 110  50, clip]{figures/action_draw_x/action_12.pdf}\hspace{-0.18em}
      \includegraphics[width=\x\linewidth, trim=500 110 110  50, clip]{figures/action_draw_x/action_13.pdf}\hspace{-0.18em}
      \includegraphics[width=\x\linewidth, trim=500 110 110  50, clip]{figures/action_draw_x/action_14.pdf}\hspace{-0.18em}
      \includegraphics[width=\x\linewidth, trim=500 110 110  50, clip]{figures/action_draw_x/action_15.pdf}\hspace{-0.18em}
      \includegraphics[width=\x\linewidth, trim=500 110 110  50, clip]{figures/action_draw_x/action_16.pdf}\hspace{-0.18em}
      \includegraphics[width=\x\linewidth, trim=500 110 110  50, clip]{figures/action_draw_x/action_17.pdf}\hspace{-0.18em}
      \includegraphics[width=\x\linewidth, trim=500 110 110  50, clip]{figures/action_draw_x/action_18.pdf}\hspace{-0.18em}
      \includegraphics[width=\x\linewidth, trim=500 110 110  50, clip]{figures/action_draw_x/action_19.pdf}\hspace{-0.18em}
      \includegraphics[width=\x\linewidth, trim=500 110 110  50, clip]{figures/action_draw_x/action_20.pdf}\hspace{-0.18em}
      \includegraphics[width=\x\linewidth, trim=500 110 110  50, clip]{figures/action_draw_x/action_21.pdf}\hspace{-0.18em}
      \includegraphics[width=\x\linewidth, trim=500 110 110  50, clip]{figures/action_draw_x/action_22.pdf}\hspace{-0.18em}
      \includegraphics[width=\x\linewidth, trim=500 110 110  50, clip]{figures/action_draw_x/action_23.pdf}\hspace{-0.18em}
      \includegraphics[width=\x\linewidth, trim=500 110 110  50, clip]{figures/action_draw_x/action_24.pdf}\hspace{-0.18em}
      \includegraphics[width=\x\linewidth, trim=500 110 110  50, clip]{figures/action_draw_x/action_25.pdf}\hspace{-0.18em}
      \includegraphics[width=\x\linewidth, trim=500 110 110  50, clip]{figures/action_draw_x/action_26.pdf}\hspace{-0.18em}
      \includegraphics[width=\x\linewidth, trim=500 110 110  50, clip]{figures/action_draw_x/action_27.pdf}\hspace{-0.18em}
      \includegraphics[width=\x\linewidth, trim=500 110 110  50, clip]{figures/action_draw_x/action_28.pdf}\hspace{-0.18em}
      \includegraphics[width=\x\linewidth, trim=500 110 110  50, clip]{figures/action_draw_x/action_29.pdf}\hspace{-0.18em}
      \includegraphics[width=\x\linewidth, trim=500 110 110  50, clip]{figures/action_draw_x/action_30.pdf}\hspace{-0.18em}
      \includegraphics[width=\x\linewidth, trim=500 110 110  50, clip]{figures/action_draw_x/action_31.pdf}\hspace{-0.18em}
      \includegraphics[width=\x\linewidth, trim=500 110 110  50, clip]{figures/action_draw_x/action_32.pdf}\hspace{-0.18em}
      \includegraphics[width=\x\linewidth, trim=500 110 110  50, clip]{figures/action_draw_x/action_33.pdf}\hspace{-0.18em}
      \includegraphics[width=\x\linewidth, trim=500 110 110  50, clip]{figures/action_draw_x/action_34.pdf}\hspace{-0.18em}
      \includegraphics[width=\x\linewidth, trim=500 110 110  50, clip]{figures/action_draw_x/action_35.pdf}\hspace{-0.18em}
      \includegraphics[width=\x\linewidth, trim=500 110 110  50, clip]{figures/action_draw_x/action_36.pdf}\hspace{-0.18em}
      \includegraphics[width=\x\linewidth, trim=500 110 110  50, clip]{figures/action_draw_x/action_37.pdf}\hspace{-0.18em}
      \includegraphics[width=\x\linewidth, trim=500 110 110  50, clip]{figures/action_draw_x/action_38.pdf}\hspace{-0.18em}
      \includegraphics[width=\x\linewidth, trim=500 110 110  50, clip]{figures/action_draw_x/action_39.pdf}\hspace{-0.18em}
      \includegraphics[width=\x\linewidth, trim=500 110 110  50, clip]{figures/action_draw_x/action_40.pdf}\hspace{-0.18em}
      \includegraphics[width=\x\linewidth, trim=500 110 110  50, clip]{figures/action_draw_x/action_41.pdf}\hspace{-0.18em}
      \includegraphics[width=\x\linewidth, trim=500 110 110  50, clip]{figures/action_draw_x/action_42.pdf}\hspace{-0.18em}
      \includegraphics[width=\x\linewidth, trim=500 110 110  50, clip]{figures/action_draw_x/action_43.pdf}\hspace{-0.18em}
      \includegraphics[width=\x\linewidth, trim=500 110 110  50, clip]{figures/action_draw_x/action_44.pdf}\hspace{-0.18em}
      \includegraphics[width=\x\linewidth, trim=500 110 110  50, clip]{figures/action_draw_x/action_45.pdf}\hspace{-0.18em}
      \includegraphics[width=\x\linewidth, trim=500 110 110  50, clip]{figures/action_draw_x/action_46.pdf}\hspace{-0.18em}
      \includegraphics[width=\x\linewidth, trim=500 110 110  50, clip]{figures/action_draw_x/action_47.pdf}\hspace{-0.18em}
      \includegraphics[width=\x\linewidth, trim=500 110 110  50, clip]{figures/action_draw_x/action_48.pdf}\hspace{-0.18em}
      \includegraphics[width=\x\linewidth, trim=500 110 110  50, clip]{figures/action_draw_x/action_49.pdf}\hspace{-0.18em}
      \includegraphics[width=\x\linewidth, trim=500 110 110  50, clip]{figures/action_draw_x/action_50.pdf}\hspace{-0.18em}
      \includegraphics[width=\x\linewidth, trim=500 110 110  50, clip]{figures/action_draw_x/action_51.pdf}\hspace{-0.18em}
      \includegraphics[width=\x\linewidth, trim=500 110 110  50, clip]{figures/action_draw_x/action_52.pdf}\hspace{-0.18em}
      \includegraphics[width=\x\linewidth, trim=500 110 110  50, clip]{figures/action_draw_x/action_53.pdf}\hspace{-0.18em}
      \includegraphics[width=\x\linewidth, trim=500 110 110  50, clip]{figures/action_draw_x/action_54.pdf}\hspace{-0.18em}
      \includegraphics[width=\x\linewidth, trim=500 110 110  50, clip]{figures/action_draw_x/action_55.pdf}\hspace{-0.18em}
      \includegraphics[width=\x\linewidth, trim=500 110 110  50, clip]{figures/action_draw_x/action_56.pdf}\hspace{-0.18em}
      \includegraphics[width=\x\linewidth, trim=500 110 110  50, clip]{figures/action_draw_x/action_57.pdf}\hspace{-0.18em}
      \includegraphics[width=\x\linewidth, trim=500 110 110  50, clip]{figures/action_draw_x/action_58.pdf}\hspace{-0.18em}
      \includegraphics[width=\x\linewidth, trim=500 110 110  50, clip]{figures/action_draw_x/action_59.pdf}\hspace{-0.18em}
      \includegraphics[width=\x\linewidth, trim=500 110 110  50, clip]{figures/action_draw_x/action_60.pdf}\hspace{-0.18em}
      \includegraphics[width=\x\linewidth, trim=500 110 110  50, clip]{figures/action_draw_x/action_61.pdf}
    \caption{Elastic average after alignment.}
    \end{subfigure}
    \caption{Human action dataset, activity "\textit{Draw X}".}
    \label{fig:human_action_1}
  \end{center}
\end{figure}

\section{Conclusions}
\label{sec:conclusions_3}

In this chapter, we successfully incorporated closed-form diffeomorphic transformations (presented in \cref{chapter:2}) into a temporal transformer network (TTN) resembling \cite{Weber2019} that can both align pairwise sequential data and learn representative average sequences for multiclass joint alignment.
We add regularization to the alignment loss function to address the warping function's sensitivity to noise and outliers. 
Furthermore, given the lack of ground truth for the latent warps in real data, we use a simple classification model (nearest centroid) as a proxy metric for the quality of the joint alignment and the average signal. 
We conducted extensive experiments on 84 univariate and 30 multivariate datasets from the UCR archive \cite{dau2019ucr}, to validate the generalization ability of our model to unseen data for time series joint alignment. Results show significant improvements both in terms of efficiency and accuracy (see \cref{fig:ucr:accuracy}): our method was better or no worse in 94\% of the datasets compared with Euclidean,  DBA (77\%), SoftDTW (69\%), DTAN (76\%) and ResNet-TW (70\%).
Overall, our proposed TTN using \textit{DIFW} beats all 5 comparing methods. The precise computation of the gradient of the transformation translates to an efficient search in the parameter space, which leads to faster and better solutions at convergence.

\graphicspath{{content/chapter4/}}

\chapter[Time Series Classification with Diffeomorphic Elastic Averaging: A case study]{Time Series Classification with Diffeomorphic Elastic Averaging: \\ A case study}\label{chapter:4}
\begingroup
\hypertarget{chapter4}{}
\hypersetup{linkcolor=black}
\vspace{-0.7cm}
\setstretch{0.93}
\minitoc
\endgroup

\section{Introduction}\label{sec:introduction_4}

Time series alignment methods, such as those presented in \cref{chapter:3} can be used as a preprocessing step before performing other machine learning tasks, such as clustering, classification, averaging, and so on. In this chapter one of these applications is discussed: \textbf{time series classification (TSC)}.
Applications of time series classification are diverse and include areas such as finance \cite{majumdar2020clustering}, power systems \cite{susto2018time}, healthcare \cite{wang2022multihead}, manufacturing \cite{hsu2021multiple} and environmental monitoring \cite{gundersen2020binary}. On the whole, time series classification is a powerful tool that provides valuable insights into patterns and trends in sequential data, and has the potential to significantly improve decision-making and outcomes in many fields.

The mining industry, for instance, has been exploring the \textbf{benefits of automation and remote operation}, which can enhance safety by minimizing human presence near machines and hazardous work areas. One promising application of such automation is the use of TSC methods for failure detection in hydraulic rock drills through the analysis of pressure sensor data. These machines operate under severe performance demands in harsh environments, which generate vibrations that affect the behavior of the pressure signals.
The oscillations resulting from pressure propagation at high frequencies can lead to external changes that are not faults, such as changes in hose lengths, significantly altering the pressure signals' dynamics. Such \textbf{complexities produce time series data that are not directly comparable}, rendering current time series classification methods incapable of accurately classifying the data. Therefore, to be effective, consistent alignment of the time series data is crucial. 

This chapter focuses on the already mentioned industrial application, i.e. the \textbf{fault classification of hydraulic rock drill pressure data} under different configurations and scenarios, which was put forward by the Prognostics and Health Management Society \cite{phm} for the 2022 PHM Data Challenge. 
The PHM Data Challenge\footnote{\url{https://data.phmsociety.org/2022-phm-conference-data-challenge/}} is an international open data competition specialized in industrial data analytics and covers a wide spectrum of real-world industrial problems. 
This chapter summarizes our participation in the 2022 data challenge, in which we presented a model that integrates the time series alignment methods proposed in \cref{chapter:3} and the latest advances in deep learning to simultaneously align and classify faulty signals. 

This chapter is structured as follows.
Work related to time series classification is discussed in \cref{sec:related_work_4}. Then, the PHM challenge is described in \cref{sec:phm_challenge} and present the proposed model in \cref{sec:method_4}. Experimental results are included in \cref{sec:results_4} and final remarks are included in \cref{sec:conclusions_4}.

\section{Related Work}\label{sec:related_work_4}

\textbf{Time series classification} is defined as the process of assigning a category to an unlabeled time series observation by exploiting patterns of the training data.
Let 
$D=\{(X_{1}, y_{1}), \ldots, (X_{n}, y_{n})\}$ 
be a dataset containing a collection of pairs, where $X_{i}$ could either be a univariate or multivariate time series with its corresponding label denoted by $Y_{i}$.
Given such a set of $n$ labelled examples consisting of $d$ features measured at $T$ time steps ($X \in \mathbb{R}^{n \times T \times d}$), and the outcome labels $y \in \mathbb{R}^{n}$, the goal is to learn a mapping from $\{x_{t}^{(i)}\}_{t=1}^{T}$ to $y_{i}$ where $x_{t}^{(i)} \in \mathbb{R}^{d}$ and $i \in \{1,\cdots,n\}$ is an index into the $i^{th}$ sample. Each feature is represented as a set of $T$ measurements. 
Note that, in general, each time series can have a different number of observations $T$.

One of the key challenges in time series classification is to capture and model the \textbf{inherent temporal dependencies} and patterns in the data. Traditional machine learning algorithms, such as decision trees and support vector machines, are designed for cross-sectional data and are not suitable for TSC due to their inability to handle and harness the temporal information. 
For \textbf{traditional classification models} the order of the attributes is irrelevant and the interaction between variables is considered independent of their relative positions. 
For instance, the Naive Bayes algorithm assumes conditional independence between each feature given the label.
In contrast, for time series data, the ordering of the variables is crucial for identifying the best discriminating features, as the order of the values is an essential part of the time series, and consecutive time points are likely to be highly correlated. 
To address this, \textbf{specialized time series algorithms} have been developed, such as dynamic time warping and recurrent neural networks, that are capable of capturing the temporal dependencies in the data. 
This section presents a review of current approaches for time series classification.

\subsection{$k$ Nearest-Neighbor Methods}

Nearest-neighbor methods are based on the notion of similarity, and the prediction of a new sample is made based on the target value of $k$ similar samples.
Consequently, their key element is the metric that defines the \textbf{similarity between any pair of samples}.
The choice of distance measure is dependent on the domain and, more specifically, the invariances required by the domain  (see \cref{sec:invariances}). 
A classic benchmark for TSC is the nearest neighbor algorithm coupled with the Dynamic Time Warping (NN-DTW) \cite{bagnall2017great}. Elastic measures, such as DTW, are intended to compensate for local distortions, misalignments, or warpings in time series caused by stretched or shrunken sections. It should be noted, however, that NN-DTW is highly sensitive to noise in the training set, which is a strong trait of time series datasets \cite{abanda2019review}.
Much research has focused on finding alternative efficient similarity measures, many of those were reviewed in \cref{sec:shape_elastic}. Commonly used similarity measures include variations of DTW such as TWED \cite{marteau2008time}, SoftDTW \cite{cuturi2017soft}, DDTW \cite{keogh2001derivative} and others SBD \cite{paparrizos2017fast}, LCSS \cite{vlachos2002discovering}, or GAK \cite{cuturi2011fast}. A comprehensive review of distance-based TSC methods can be found in \cite{abanda2019review}.

\subsection{Tree-based Algorithms}

Tree-based machine learning algorithms, such as random forest \cite{breiman2001random}, build a tree-like structure in which each internal \textbf{node represents a decision based on certain features} of the data, and each leaf node represents a final prediction. In TSC, these algorithms are applied to the time series data by segmenting it and extracting features from each segment to feed the decision tree.
Time Series Forest \cite{deng2013time} considers subsequences generated from random start and end indices, applies three time domain transformations (mean, standard deviation and slope), and then trains a decision tree using these features. The operation is repeated to learn an ensemble of decision trees on different randomly chosen intervals.
Time Series Bag-Of-Features \cite{baydogan2013bag} extracts more features than Time Series Forest. Each randomly sampled interval is divided into non-overlapping sub-intervals, and four features are extracted: mean, standard deviation, as well as the start and end indices of each sub-interval.
The Proximity Forest algorithm \cite{lucas2019proximity} works directly with raw time series to  build an ensemble of classification trees in which data at each node is split based on similarity to a representative time series from each class. This differs from the typical attribute-value splitting methods used in decision trees.
In a similar vein, TS-CHIEF \cite{shifaz2020ts} is a stochastic tree-based ensemble designed for speed and high accuracy, where at each node the algorithm selects from a random selection of TSC methods the one that best classifies the data reaching that node.

\subsection{Dictionary-based Algorithms}

Dictionary-based algorithms efficiently transform time series data into a bag of words representation. These techniques are particularly adept at addressing noisy data and recognizing repeating patterns. To achieve this, the series is first shortened using an approximation method, and then a quantization method is used to discretize the values, thereby forming words. Each time series is represented by a histogram that counts words frequency. Notable dictionary based algorithms are BOP \cite{lin2012rotation}, SAX \cite{lin2003symbolic}, SAX-VSM \cite{senin2013sax}, SFA \cite{schafer2012sfa} and BOSS \cite{schafer2015boss}. We refer the reader to \cref{sec:bag_words} for a detailed review of these methods.

\subsection{Deep Learning}

Over the past decade, deep learning has led to many breakthroughs in several fields such as computer vision \cite{fawaz2019deep} and natural language processing \cite{floridi2020gpt}. Deep learning has also been recently investigated for time series classification.
\textbf{Convolutional Neural Networks} (CNNs) have showed promising results for TSC \cite{wang2017time}. 
In this sense, Fully Convolutional Neural Networks (FCNs) \cite{wang2017time} were shown to achieve great performance without the need to add pooling layers to reduce the input data's dimensionality. 
\textbf{InceptionTime} \cite{fawaz2019deep} is an ensemble of 60 neural network models to perform time series classification. 
CNNs have a large number of trainable parameters in comparison to more classic algorithms such as logistic regression or support vector machines, thus usually requiring a large sample size to find good values for the trainable parameters. Based on this observation, the \textbf{Random Convolutional Kernel Transform} (ROCKET) \cite{dempster2020rocket} extracts features from time series using a large number of random convolutional kernels, meaning that all the parameters of all the kernels (length, weights, bias, dilation, and padding) are randomly generated from fixed distributions. 
%

\begin{remark}
    Applied to time series data, 1D convolutions inherently capture phase invariance and noise invariance, to a degree.
    Max pooling coupled with multiple layers allows the model to smooth the inputs and learn higher-level abstractions.
    However, temporal invariances such as warping must also be considered. Data may be collected from different sources, at different times, and using different sampling frequencies, which may lead to misleading results.
\end{remark}

\subsection{Ensemble Models}

Ensemble-based methods combine different classifiers together to achieve a higher accuracy. For example, Collective of Transformation-Based Ensembles (COTE) \cite{bagnall2015time} uses an ensemble of different classifiers over different time series representations.
Hierarchical Vote Collective of Transformation-Based Ensembles (HIVE-COTE) \cite{lines2018time} is an extension of COTE with a new type of spectral classifier called Random Interval Spectral Ensemble, two more classifiers (BOSS \cite{schafer2015boss} and Time Series Forest \cite{deng2013time}) and a hierarchical voting scheme, which further improves the decision taken by the ensemble.
In terms of classification accuracy, HIVE-COTE is the current state of the art, but it is often infeasible to run on even modest amounts of data.

\section{Rock Drill Fault Detection: PHM Challenge}\label{sec:phm_challenge}

The 2022 PHM challenge \cite{phm} addresses the problem of \textbf{fault classification for a rock drill} application under different individual configurations of the rock drill. The task is to develop a fault diagnosis model using the provided pressure sensor data as input. 
The application of TSC methods for fault classification holds significant potential for various industries, including the hydraulics sector within the mining industry. The drive towards automation and remote operation in the mining industry is motivated by the desire to create a safer working environment by reducing human presence near machines and hazardous work areas \cite{jakobsson2022time}.

\subsection{Industrial Application}

\begin{figure}[!b]
    \begin{center}
        \begin{subfigure}[t]{0.38\linewidth}
        \includegraphics[width=\linewidth, trim=80 80 80 20, clip]{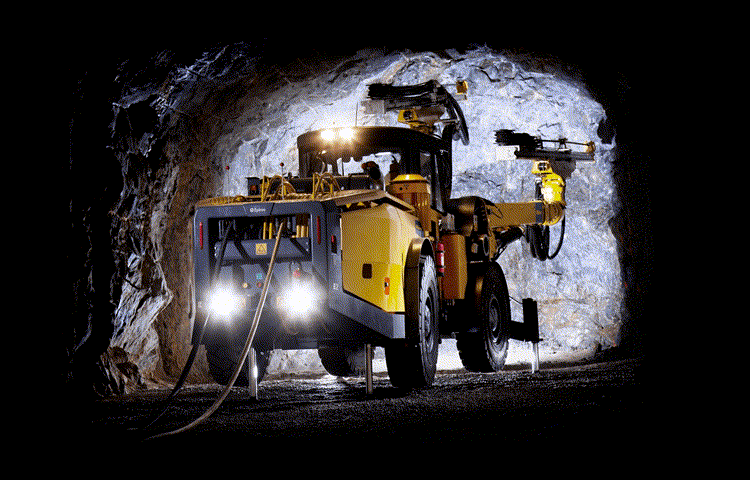}
        \caption{Rock drills on a drill rig in their natural underground habitat. 
        The rock drill is composed of four primary systems: percussion, rotation, damper and flushing. This work primarily targets the percussion and damping system.
        }
        \label{fig:phm_1}
    \end{subfigure}
    \hfill
    \begin{subfigure}[t]{0.6\linewidth}
        \includegraphics[width=\linewidth]{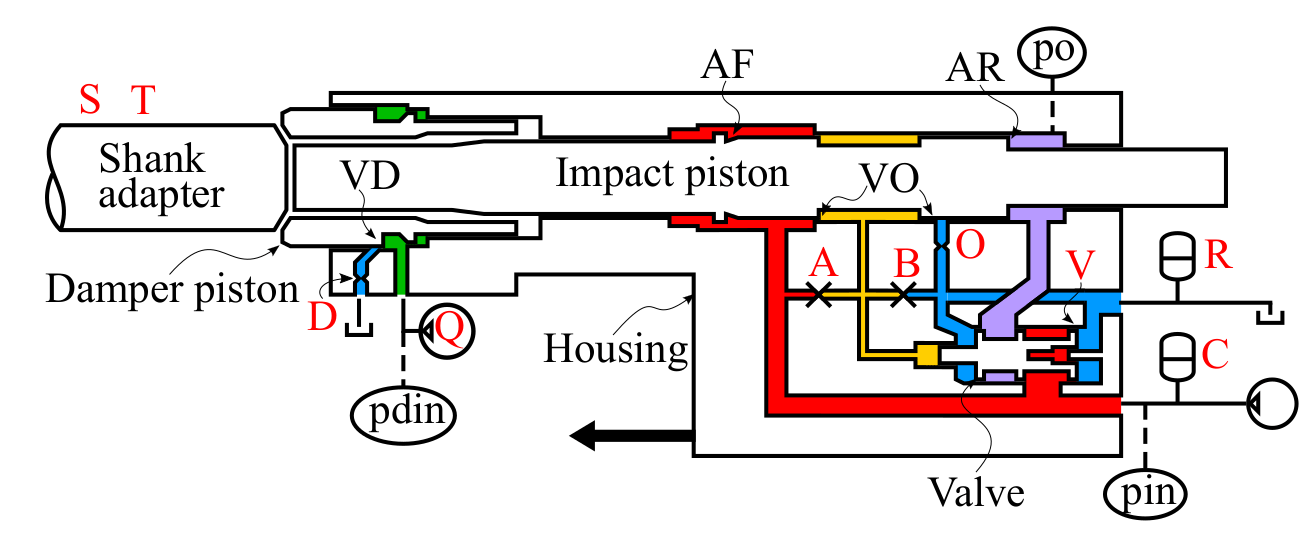}
        \caption{Diagram of the percussion and damper system, with 10 induced fault modes in red capital letters, and sensor locations (\textit{pin, po, pdin}) in ovals. Colored areas represent various hydraulic lines such as \textcolor{red}{high pressure supply} (red), \textcolor{blue}{low pressure return} (blue), \textcolor{violet}{alternating pressure} (purple), \textcolor{ForestGreen}{damper pressure} (green) and \textcolor{orange}{control pressure for the valve} (yellow).}
        \label{fig:phm_3}
    \end{subfigure}
    \caption{Rock drill information. Adapted with permission from \cite{jakobsson2022dataset}.}
    \label{fig:phm_1_3}
  \end{center}
\end{figure}

Hydraulic rock drills are used in a wide range of applications where holes are needed in hard rock materials. Such systems, as seen in \cref{fig:phm_1}, often operate under high performance demands in harsh environments, with vibrations and moisture. 
For this work, data is measured using a single pressure sensor located on the inlet pressure line where many effects of internal conditions are believed to manifest.

A hydraulic rock drill is a hydromechanical device using for generating stress waves in a drill steel (see \cref{fig:phm_3}). It operates at frequencies where oscillations from \textbf{pressure propagation} is a phenomenon that needs to be taken into account. 
This causes small external changes that are not faults, such as changes in hose lengths, to significantly alter the behavior of the pressure signals. To capture such differences, different individuals are tested and included in the data. 
Furthermore, the measured signal is periodic, and governed by some cyclic phenomenon such as the repeated opening of a valve. It has strong non-linearities, produced by impacts and sudden valve openings. The fundamental machine frequency is influenced by various disturbances, causing different events to occur at different times during a cycle depending on faults and individual variation such as unit configuration and manufacturing tolerances.

Such complex temporal dynamics produce time series data that is not directly comparable, making it difficult for TSC methods to accurately classify the data. In order for these techniques to be effective, time series data must be consistently aligned. The model proposed in \cref{sec:method_4} automatically handles these temporal dependencies in the data using an extension of the Temporal Transformer Network presented in \cref{chapter:3}.

\subsection{Nature of Data}

The data set provides \textbf{normalized pressure measurements} from a hydraulic rock drill, operating in a close-to-reality type of test cell, while different faults and other  pre-defined variations are introduced to the system in a controlled manner. 

There are 11 different \textbf{fault classification categories} (see \cref{tab:faults}), in which 10 are different failure modes and one class is from the healthy/no fault condition. 
The approximate location in the rock drill where faults are introduced are shown in red capital letters in \cref{fig:phm_3}. 
The differences between some fault classes are small compared to the individual differences, making this a challenging classification task.

\begin{table}[!htb]
    \small
    \caption{Fault classification categories}
    \label{tab:faults}
    \vspace{-1em}
    \begin{center}    
    \begin{tabular}{lll}   
        \toprule 
        Label & Letter & Description \\
        \midrule
        1 & NF & No-Fault \\
        2 & T & Thicker drill steel. \\
        3 & A & A-seal missing. Leakage from high pressure channel to control channel. \\
        4 & B & B-seal missing. Leakage from control channel to return channel. \\
        5 & R & Return accumulator, damaged. \\
        6 & S & Longer drill steel. \\
        7 & D & Damper orifice is larger than usual. \\
        8 & Q & Low flow to the damper circuit. \\
        9 & V & Valve damage. A small wear-flat on one of the valve lands. \\
        10 & O & Orifice on control line outlet larger than usual. \\
        11 & C & Charge level in high pressure accumulator is low. \\
        \bottomrule
    \end{tabular}
    \end{center}
\end{table}

\paragraph{Reference Data}
Furthermore, data from multiple individuals is scarce and does not cover the expected variability upon deployment. 
As a result, organizers provide reference data from each individual upon deployment. 
The challenge is to create models able to generalize over individuals, while having access to some reference data from the specific individual.

\paragraph{Data Collection}

\begin{figure}[!htb]
    \begin{center}
        \begin{subfigure}[t]{0.46\linewidth}
        \includegraphics[width=\linewidth]{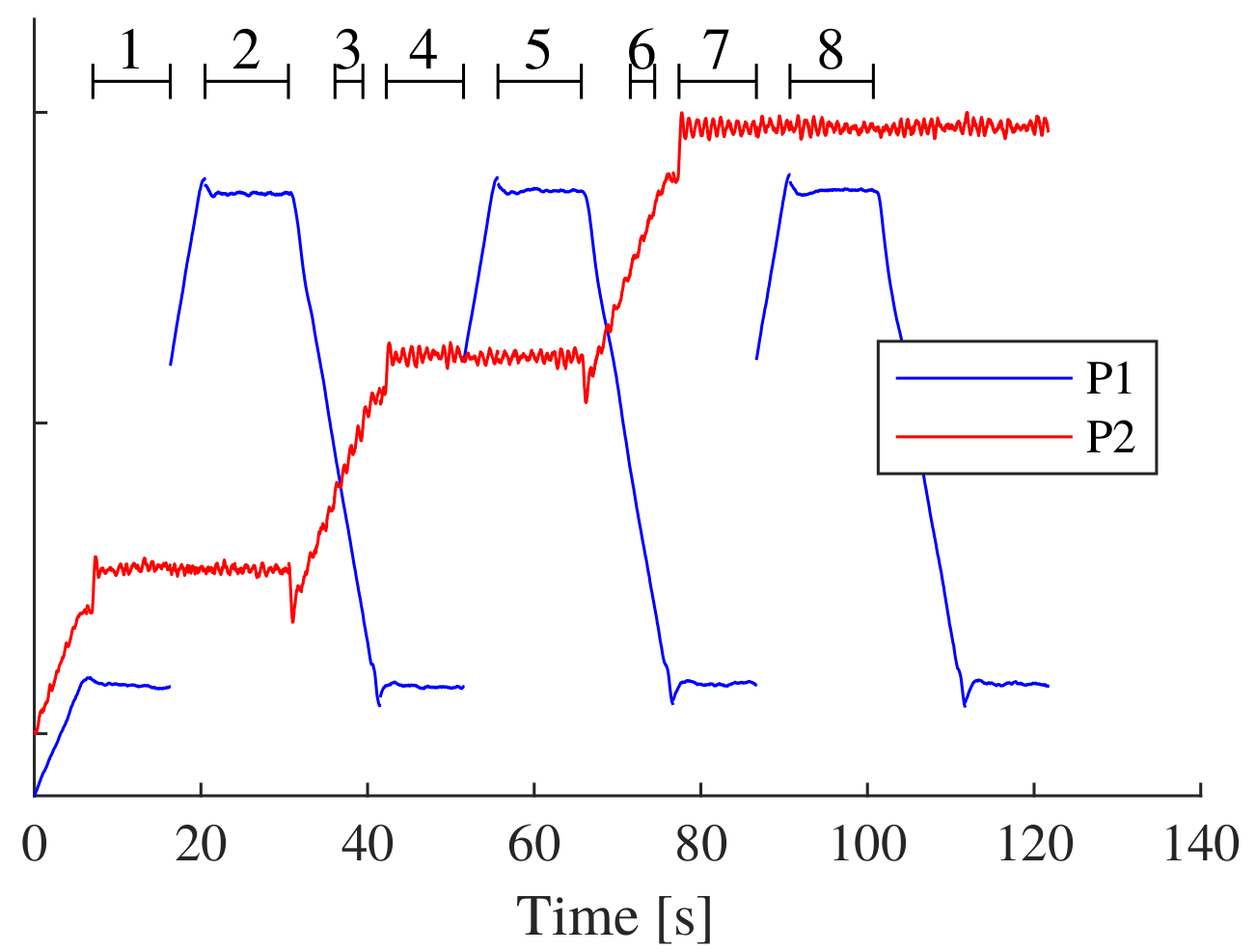}
        \caption{To simulate different individual configurations, the percussion pressure (P1) and feed force (P2) are changed. Markers in black show where individuals 1 through 8 are collected.}
        \label{fig:phm_5}
    \end{subfigure}
    \hfill
    \begin{subfigure}[t]{0.51\linewidth}
        \includegraphics[width=\linewidth]{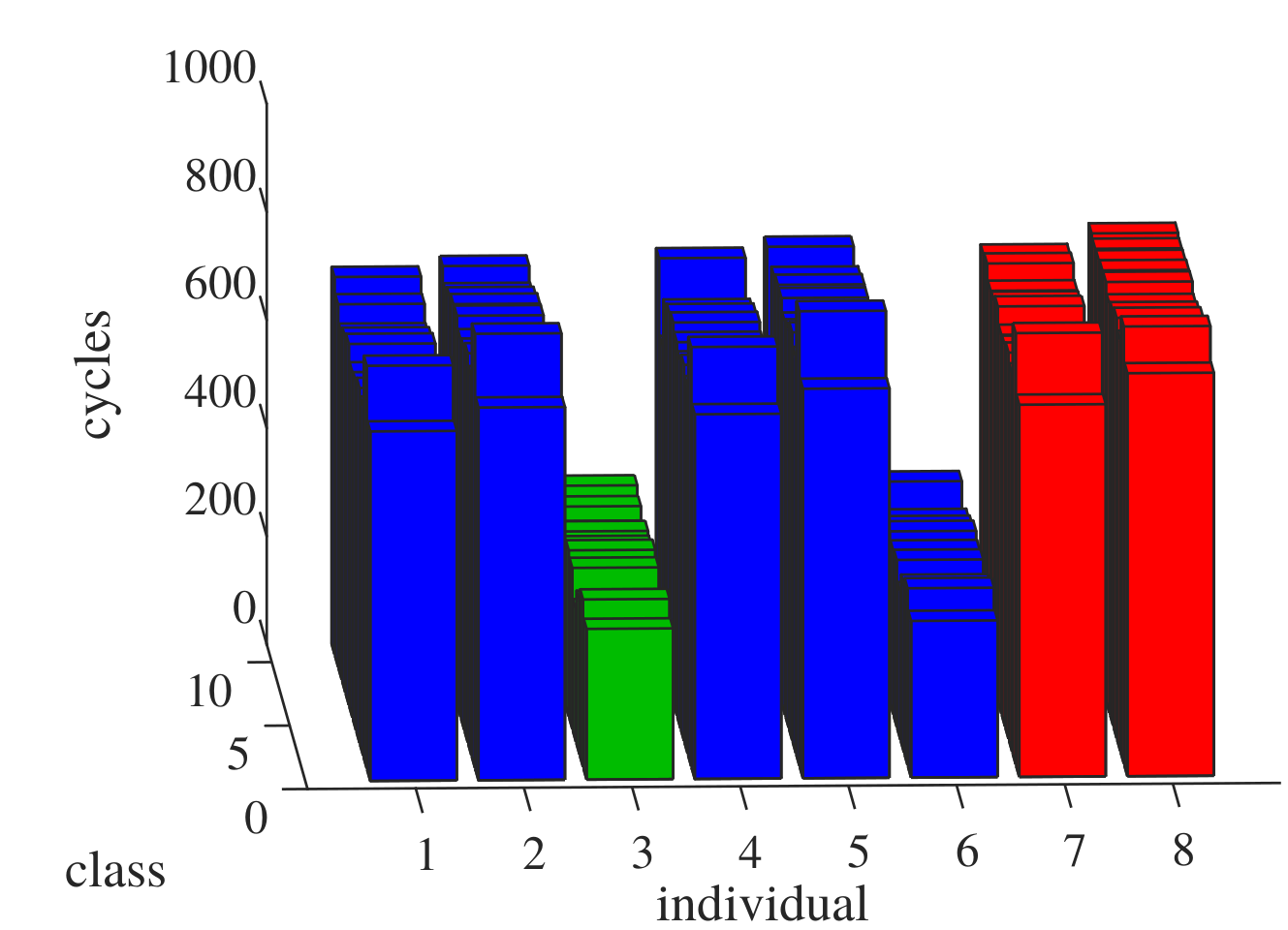}
        \caption{Number of cycles for each class and individual. Blue represents training data, green is data used for online scoring (test set) and red a holdout test set used for official competition scoring (validation).}
        \label{fig:phm_6}
    \end{subfigure}
    \caption{Data collection stats. Adapted with permission from \cite{jakobsson2022dataset}.}
    \label{fig:phm_5_6}
  \end{center}
  \end{figure}

For each case, ten seconds of data is collected, corresponding to approximately 800 impact cycles sampled at 50kHz. The ten-second long time series are divided into shorter segments, containing one impact cycle each. 
The automated collection sequence is shown in \cref{fig:phm_5}, where two configuration variables percussion pressure (P1) and feed force (P2) are changed according to the graph. 
Note that P2 has a larger influence on the pressure signature than P1, so it is easier to generalize between certain classes.
The structure of the data is shown in \cref{fig:phm_6}. In total 8 individuals $\times$ 11 classes $\times$ 300 to 700 cycles each give approximately 54000 cycles in total, each with data from 3 sensors. Each pressure time series range from 556 to 748 samples. The size of the dataset in $.csv$ format is 976 MB.

\paragraph{Train / Test / Validation Split}
The training set consists of data from various faults from five individual configurations, while the test data is from one individual setup of the rock drill. The validation set for the final scoring for the competition is from two individual configurations from the rock drill and the labels were blind to the contest participants. For both the test and validation sets a reference from a No-Fault health condition is provided.

\paragraph{Individual Differences}

\begin{figure}[!htb]
    \begin{center}
    \centerline{\includegraphics[width=0.9\linewidth]{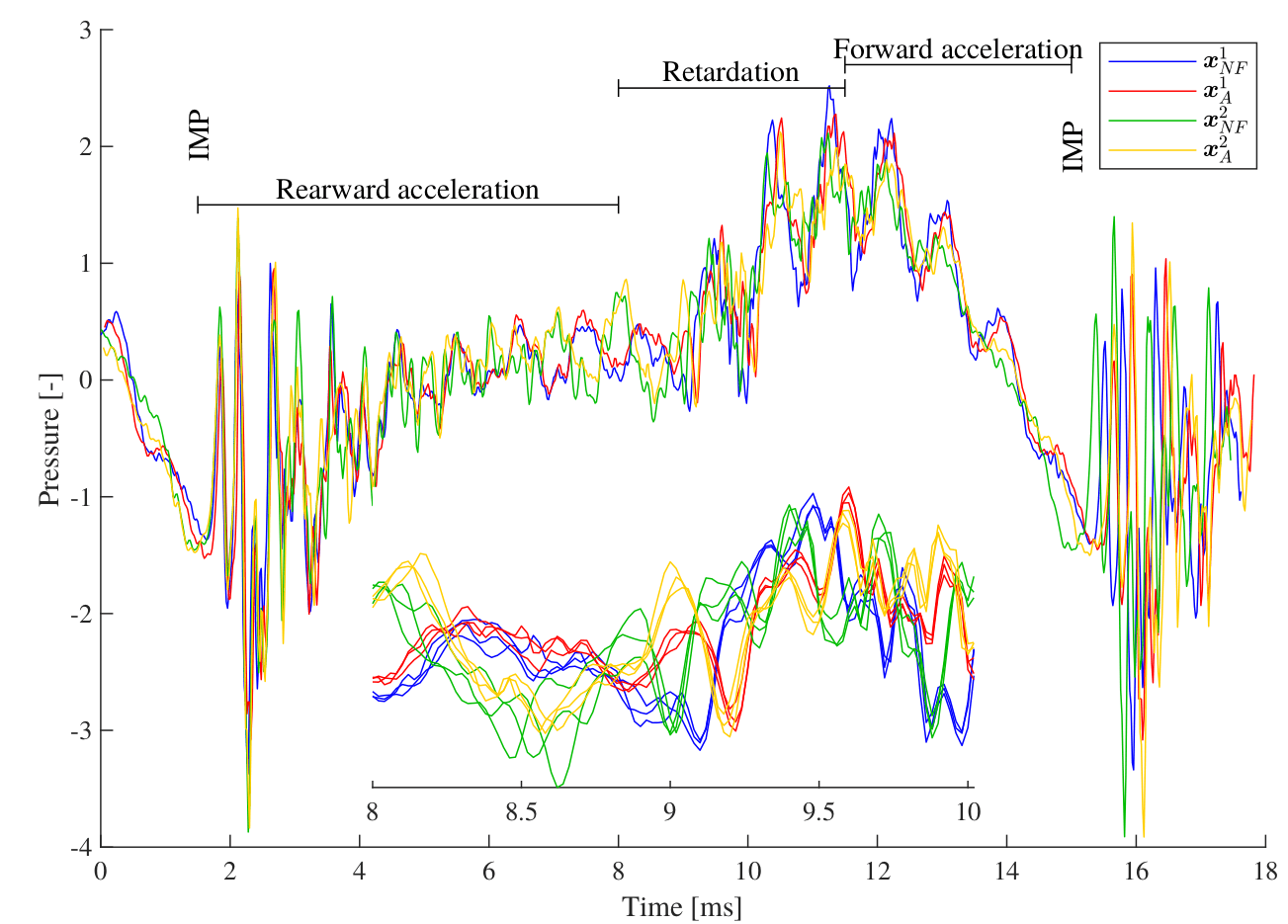}}
    \caption{Pressure from two classes (NF and A), and two individuals (1 and 2). 
    At $t = 9$, a distinction between the classes becomes apparent as the defining valley is delayed for both individuals. This difference can be challenging to observe without the reference NF class.
    Adapted with permission from \cite{jakobsson2022dataset}.
    }
    \label{fig:phm_2}
    \end{center}
\end{figure}

Two individuals can have quite different pressure signatures due to a difference in supply hose length. \cref{fig:phm_2} shows an example of how individual differences can confound the pressure trace in a fault scenario compared with a No-Fault (NF) case. 
In fact, the oscillation seen around $t = 9$ arrives later as the fault is introduced, but this disparity can be masked by individual differences. A known reference from the current individual might be required for calibration \cite{jakobsson2022time}. 

To address this issue various strategies can be employed. The machine learning approach is to collect enough fault data from multiple individuals to train a model that can generalize well for unseen variations. Another approach is to build a physics-based model rooted on fundamental principles and refine it using data obtained from specific individuals. This method, although effective, can be both costly and time-consuming, particularly for complex systems. 
A third strategy is to gather limited data from few individuals and then personalize the model to suit specific individuals. The model in \cref{sec:method_4} adopts this approach.

\section{Classification Model}\label{sec:method_4}

This section presents the classification model that was submitted to the 2022 PHM Data Challenge. 
The submitted approach combines the latest advances in deep learning with the innovative alignment methods presented in \cref{chapter:3} to create a state-of-the-art model that can \textbf{simultaneously align and classify faulty signals}. 
\cref{tab:phm_notation} shows the notation used throughout this section. Note that $X_{dat}$ corresponds to the multivariate time series pressure data of one impact cycle, while $X_{ref}$ is the reference pressure data (also multivariate) of the No-Fault class of a given individual. 
\begin{table}[!htb]
    \caption{Notation}
    \label{tab:phm_notation}
    \begin{center}
    \begin{tabular}{ll}
    \toprule
    $X_{dat}$ & data multivariate time series \\
    $X_{ref}$ & reference multivariate time series \\
    $Y$ & time series label \\
    $\theta$ & alignment parameters \\
    $v$ & velocity function \\
    $\phi$ & warping function \\ \bottomrule
    \end{tabular}
\end{center}
\end{table}

\subsection{Model Overview}
\begin{figure}[!htb]
    \begin{center}
    \centerline{\includegraphics[width=0.9\linewidth,trim=0 50 0 0,clip]{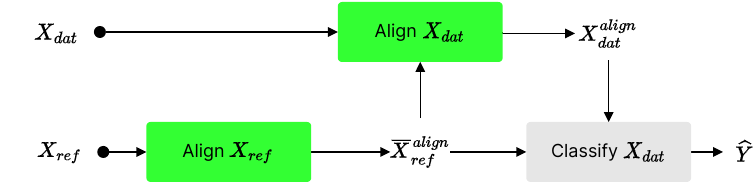}}
    \caption{Overall strategy of the proposed classification model. It simultaneously 1) aligns reference data $X_{ref}$, 2) aligns pressure data $X_{dat}$ conditioned on the elastic average of the reference data and 3) classifies it to a given label $\hat{Y}$. 
    The final classification module is fed with aligned pressure data $X_{dat}$ and with the elastic average of reference data $X_{ref}$ as well.
    The loss function is defined as the weighted average of each task loss.}
    \label{fig:phm_method2}
    \end{center}
\end{figure}

An overview of the proposed model is displayed in \cref{fig:phm_method2}:
\begin{enumerate}
    \item For each individual, multiple cycles are provided for the reference data $X_{ref}$, so the first task is to align and average this data, yielding $\bar{X}_{ref}^{align}$. 
    \item Then, raw data $X_{dat}$ is aligned conditioned on the reference average, obtaining $X_{dat}^{align}$. 
    \item Finally, the reference average $\bar{X}_{ref}^{align}$ and the aligned data $X_{data}^{align}$ are fed into the classification module that provides the predicted label $\hat{Y}$.
\end{enumerate}

The \textbf{loss function} minimizes a weighted average of three terms: the variance of the aligned reference data  $X_{ref}$, the variance of the raw data $X_{dat}$ and the classification cross-entropy. The corresponding weights are denoted as $C_{ref}$, $C_{align}$, and $C_{class}$.
\begin{equation}
    \mathcal{L} = C_{ref} \cdot \text{Var}(X_{ref}^{align}) + C_{align} \cdot \text{Var}(X_{dat}^{align}) + C_{class} \cdot \text{CrossEntropy}(Y, \hat{Y})
\end{equation}

\begin{figure}[!htb]
    \begin{center}
    \includegraphics[width=0.9\linewidth]{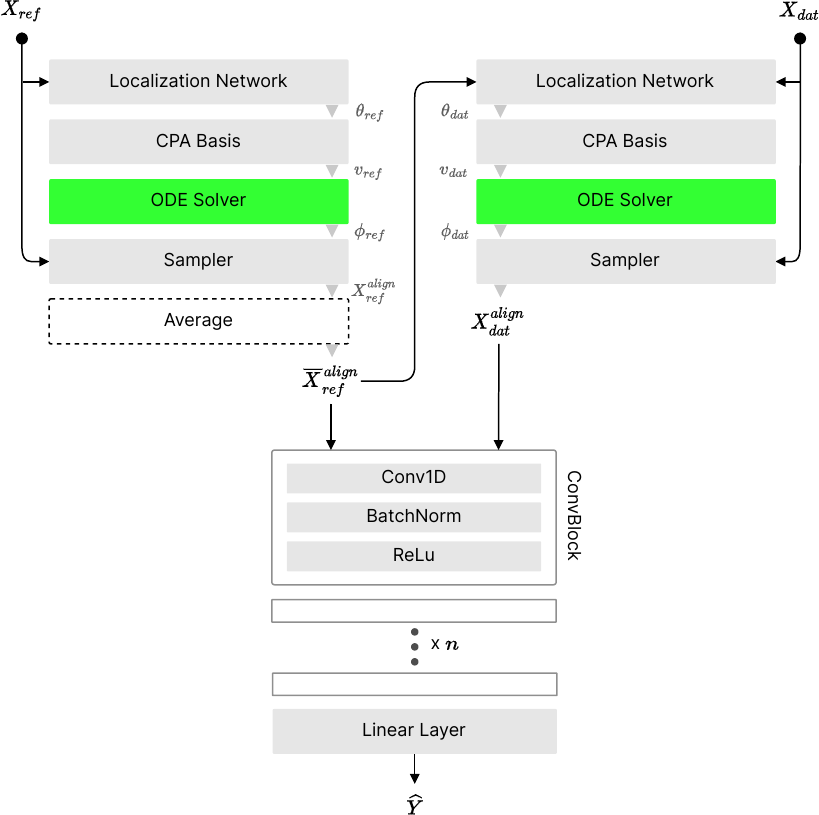}
    \caption{Proposed model for time series classification. Reference data $X_{ref}$ is first aligned using the alignment module, which is composed of a localization network, a CPA basis generator, ODE solver and a sampler. An identical architecture is used to align pressure data $X_{dat}$. Note that in both cases, the localization network computes the alignment parameters $\theta_{ref}$ and $\theta_{dat}$ conditioned on the respective raw data $X_{ref}$ and $X_{dat}$. The classification module is composed of multiple convolutional blocks that estimate the fault class $\hat{Y}$ based on aligned data $X_{dat}^{align}$ and the elastic average of reference data $\bar{X}_{ref}^{align}$.}
    \label{fig:phm_method1}
    \end{center}
\end{figure}

The model is designed to address the temporal variability and the individual differences present in the data. It uses deep learning techniques to capture the underlying patterns in the time series signals, while simultaneously aligning the signals to reduce the effect of variability. The alignment methods presented in \cref{chapter:3} provide the foundation for this model and are integrated into the deep learning architecture, allowing the model to learn both the temporal dependencies and the alignment parameters simultaneously.
A more detailed diagram of the model is shown in \cref{fig:phm_method1}.

\subsection{Alignment Module}
The alignment module is composed of a localization network, a CPA basis generator, ODE solver and a sampler. As mentioned in \cref{sec:method_3}, the localization network is a neural network that takes the input data $X$ and produces a set of transformation parameters $\theta$. The CPA basis generator then creates a set of velocity coordinates $v$ that define a regular grid over the input data. Then, the ODE solver (which is also a differentiable module) integrates the velocity function and yields the diffeormorphic warping function $\phi$. Finally, the sampler uses the warping function $\boldsymbol{\phi}$ to sample the input data $X$ at the transformed coordinates.
The only add-inn of the reference alignment module is that this data is averaged after alignment.

\clearpage
\subsection{Classification Module}
The classification module consists of $n$ convolutional blocks and a fully-connected layer. The convolutional block is made of a 1D convolutional layer, a batch norm layer and a ReLU activation layer. 

\begin{figure}[!htb]
    \begin{center}
        \begin{tikzpicture}[arr/.style={matrix of nodes, nodes={draw}}]
            
            \matrix(A) [arr]
            {
              0 & 1 & 1 & |[fill=orange!30]| 2 & |[fill=orange!30]| 4 & |[fill=orange!30]| 0 & 1 & 1 & 2 \\
            };
            \node[right=0.5 em of A] {Input $I$};

            \matrix (B) [arr, below=0.5em of A, nodes={draw, fill=red!30}]
            {
              0 & 2 & 1 \\
            };
            \node[right=4 em of B] {Filter $F$};

            \matrix (B1) [arr, below=0.5em of B]
            {
              0 & 8 & 0 \\
            };
            \node[right=4 em of B1] {$I*F$};

            \matrix (C) [arr, below=0.5em of B1]
            {
              3 & 4 & 8 & |[fill=blue!80!black!30]| 8 & 1 & 3 & 4 \\
            };
            
            \node[right=1.5 em of C] {Output $O$};

            \draw[dashed, thick, orange] (A-1-4.south west) -- (B-1-1.north west);
            \draw[dashed, thick, orange] (A-1-6.south east) -- (B-1-3.north east);

            \draw[dashed, thick, red] (B-1-1.south west) -- (B1-1-1.north west);
            \draw[dashed, thick, red] (B-1-3.south east) -- (B1-1-3.north east);

            \draw[dashed, thick, blue] (B1-1-1.south west) -- (C-1-4.north west);
            \draw[dashed, thick, blue] (B1-1-3.south east) -- (C-1-4.north east);

            \node[] at (0, -0.6){$\times$};
            \node[] at (0, -1.65){$=$};
            \node[] at (0, -2.7){$\oplus$};
        \end{tikzpicture}
\caption{1D convolution operator slides the filter across the input and records element-wise products. Input size $L_{in}=9$, kernel size $k=3$, padding $p=0$, stride $s=1$.}
\label{fig:phm_convolution}
\end{center}
\end{figure}
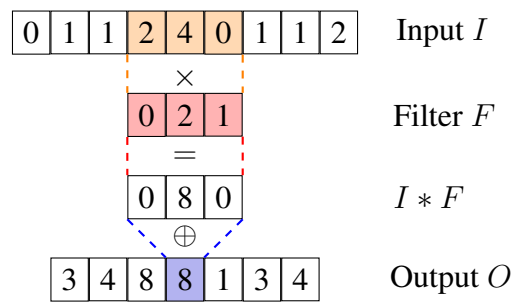

In a \textbf{1D convolution}, a kernel or filter slides across the input signal, computing the dot product between the entries of the filter and the input, producing a new feature map. The goal of a 1D convolution is to extract local features and patterns from the input signal while reducing its length, preserving spatial information and relations between values (see \cref{fig:phm_convolution}). 1D convolutions can inherently capture some shift invariance in the data, but note that these are not shift-invariant but shift-equivariant: a shift of the input to a convolutional layer produces a shift in the output feature maps by the same amount. Therefore, these convolutional layers serve to additionally eliminate any remaining warping present in the aligned data.

The kernel size, padding and stride hyperparameters determine how the filter interacts with the input signal. Padding refers to adding zero-valued elements to the edges of the input signal, allowing the filter to operate on the border elements of the input without reducing its spatial size. Stride refers to the number of positions the filter moves in each step along the input. A larger stride value results in a down-sampled output, while a smaller stride value allows the filter to capture more local information.
The output dimension for an input vector of size $L_{in}$, padding $p$, stride $s$ and kernel size $k$ can be computed as $1+(L_{in}+2p-s(k-1)-1)/s$.

A \textbf{batch norm layer} normalizes the activations of neurons for each channel across a mini-batch of data, helps to stabilize the learning process, reduce the covariate shift, and increase the training speed. 

\begin{figure}[!htb]
   \begin{center}
       \scalebox{0.9}{
       \begin{tikzpicture}
           \begin{axis}[
               legend pos=north west,
               legend cell align={left},
               legend style={draw=none},
               axis x line=middle,
               axis y line=middle,
               xtick={-2, -1, 1, 2},
               ytick={-1, 1, 2},
               grid = major,
               width=14cm,
               height=7cm,
               grid style={dashed, gray!30},
               xmin=-2,
               xmax= 2,
               ymin=-1,
               ymax= 2,
               xlabel=$x$,
               ylabel=$\varphi$,
               tick align=outside,
               enlargelimits=false]
           \addplot[domain=-2:2, orange, ultra thick,samples=100] {tanh(x)};
           \addplot[domain=-2:2, violet, ultra thick,samples=100] {ln(exp(x) + 1)};
           \addplot[domain=-2:2, red, ultra thick,samples=100] {max(x, exp(x) - 1)};
           \addplot[domain=-2:2, ForestGreen, ultra thick,samples=100] {1/(1+exp(-x))};
           \addplot[domain=-2:2, blue, ultra thick,samples=100] {max(0, x)};
           \addlegendentry{$\varphi_1(x)=\max(0, x)$}
           \addlegendentry{$\varphi_2(x)=\tanh(x)$}
           \addlegendentry{$\varphi_3(x)=\log(e^x + 1)$}
           \addlegendentry{$\varphi_4(x)=\max(x, e^x - 1)$}
           \addlegendentry{$\varphi_5(x)=(1+e^{-x})^{-1}$}
           \end{axis}
       \end{tikzpicture}
       }
\caption{Activation functions for neural networks.}
\label{fig:phm_activation}
\end{center}
\end{figure}
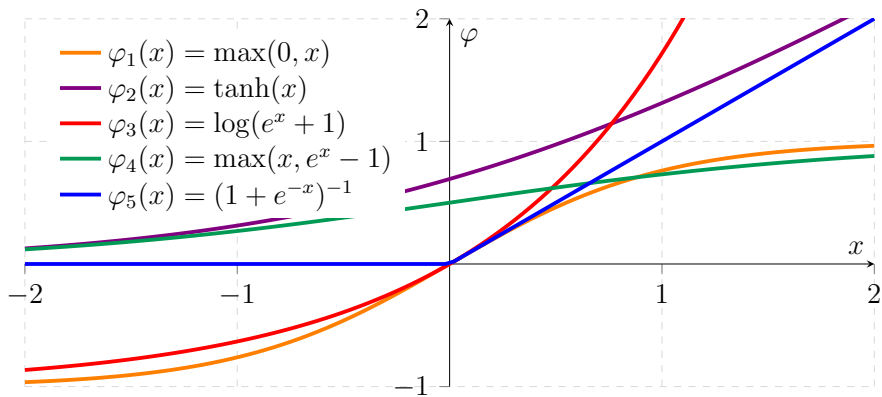

\textbf{ReLU} (Rectified Linear Unit) is defined as $\varphi(x) = \max(0, x)$, where $x$ is the input to the activation function. Other activation functions can be used, as visualized in \cref{fig:phm_activation}. Major benefits of ReLUs are sparsity and a reduced likelihood of vanishing gradient. The constant gradient of ReLUs results in faster learning, whereas the gradient of sigmoids becomes increasingly small as the absolute value of $x$ increases. In addition, sigmoids are more likely to generate some non-zero value resulting in dense representations.

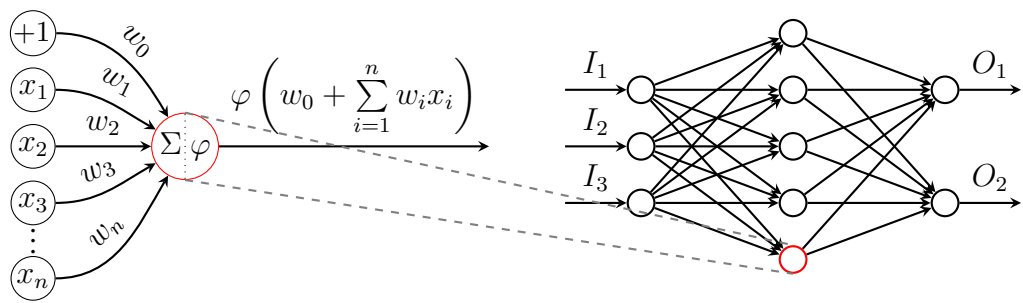
\begin{figure}[!htb]
    \begin{center}
    \scalebox{1.0}{
    \begin{tikzpicture}
        \tikzstyle{inputNode}=[draw,circle,minimum size=10pt,inner sep=0pt]
        \tikzstyle{stateTransition}=[-stealth, thick]
    \node[draw,circle,red,minimum size=25pt,inner sep=0pt] (x) at (0,0) {\color{black}$\Sigma$ $\varphi$};

    \node[inputNode,text width=15pt,align=center] (x0) at (-2, 1.5) {\small $+1$};
    \node[inputNode,text width=15pt,align=center] (x1) at (-2, 0.75) {$x_1$};
    \node[inputNode,text width=15pt,align=center] (x2) at (-2, 0) {$x_2$};
    \node[inputNode,text width=15pt,align=center] (x3) at (-2, -0.75) {$x_3$};
    \node[inputNode,text width=15pt,align=center] (xn) at (-2, -1.75) {$x_n$};

    \draw[stateTransition] (x0) to[out=0,in=120] node [midway, sloped, above] {$w_0$} (x);
    \draw[stateTransition] (x1) to[out=0,in=150] node [midway, sloped, above] {$w_1$} (x);
    \draw[stateTransition] (x2) to[out=0,in=180] node [midway, sloped, above] {$w_2$} (x);
    \draw[stateTransition] (x3) to[out=0,in=210] node [midway, sloped, above] {$w_3$} (x);
    \draw[stateTransition] (xn) to[out=0,in=240] node [midway, sloped, above] {$w_n$} (x);
    \draw[stateTransition] (x) -- (4,0.) node [midway,above] {$\varphi\left(w_0 + \sum\limits_{i=1}^{n}{w_ix_i}\right)$};
    \draw[dotted] (0,-0.43) -- (0,0.43);
    \node (dots) at (-2, -1.15) {$\vdots$};
    \node[inputNode, thick] (i1) at (6, 0.75) {};
    \node[inputNode, thick] (i2) at (6, 0) {};
    \node[inputNode, thick] (i3) at (6, -0.75) {};

    \node[inputNode, thick] (h1) at (8, 1.5) {};
    \node[inputNode, thick] (h2) at (8, 0.75) {};
    \node[inputNode, thick] (h3) at (8, 0) {};
    \node[inputNode, thick] (h4) at (8, -0.75) {};
    \node[inputNode, thick, red] (h5) at (8, -1.5) {};

    \node[inputNode, thick] (o1) at (10, 0.75) {};
    \node[inputNode, thick] (o2) at (10, -0.75) {};

    \draw[stateTransition] (5, 0.75) -- node[above] {$I_1$} (i1);
    \draw[stateTransition] (5, 0) -- node[above] {$I_2$} (i2);
    \draw[stateTransition] (5, -0.75) -- node[above] {$I_3$} (i3);

    \draw[stateTransition] (i1) -- (h1);
    \draw[stateTransition] (i1) -- (h2);
    \draw[stateTransition] (i1) -- (h3);
    \draw[stateTransition] (i1) -- (h4);
    \draw[stateTransition] (i1) -- (h5);
    \draw[stateTransition] (i2) -- (h1);
    \draw[stateTransition] (i2) -- (h2);
    \draw[stateTransition] (i2) -- (h3);
    \draw[stateTransition] (i2) -- (h4);
    \draw[stateTransition] (i2) -- (h5);
    \draw[stateTransition] (i3) -- (h1);
    \draw[stateTransition] (i3) -- (h2);
    \draw[stateTransition] (i3) -- (h3);
    \draw[stateTransition] (i3) -- (h4);
    \draw[stateTransition] (i3) -- (h5);

    \draw[stateTransition] (h1) -- (o1);
    \draw[stateTransition] (h1) -- (o2);
    \draw[stateTransition] (h2) -- (o1);
    \draw[stateTransition] (h2) -- (o2);
    \draw[stateTransition] (h3) -- (o1);
    \draw[stateTransition] (h3) -- (o2);
    \draw[stateTransition] (h4) -- (o1);
    \draw[stateTransition] (h4) -- (o2);
    \draw[stateTransition] (h5) -- (o1);
    \draw[stateTransition] (h5) -- (o2);

    \draw[stateTransition] (o1) -- node[above] {$O_1$} (11, 0.75);
    \draw[stateTransition] (o2) -- node[above] {$O_2$} (11, -0.75);

    \path[dashed, double, thick, gray] (x.north) edge[bend left=0] (h5.north);
    \path[dashed, double, thick, gray] (x.south) edge[bend right=0] (h5.south);
    \end{tikzpicture}
    }
    \caption{Illustration of a fully-connected neural network, also known as a linear layer.}
    \label{fig:phm_neural_network}
    \end{center}
\end{figure}

The final \textbf{linear layer} is a fully-connected neural network like the one shown in \cref{fig:phm_neural_network}.

\clearpage
\section{Experiments and Results}\label{sec:results_4}

\subsection{Experimental Setup}

The proposed classification model is trained with the training set made available by the organizers of the PHM Data Challenge.
Multivariate time series of pressure sensor readings $X_{dat}$, and their corresponding class $Y$ from a fleet of eight individuals are provided. A label of 1 represents the No-Fault condition, and a label of 2-11 corresponds to different fault types. 
The reference measurements $X_{ref}$ for each individual are included as well. 
The objective of the classification model is to predict the label $Y$ for each cycle.
Evaluation of the model performance is carried out using the independent test and validation sets, using the accuracy metric:
\begin{equation}\label{eq:phm_accuracy}
\text{Accuracy} = \cfrac{\text{Correctly classified cycles}}{\text{Total number of cycles}}
\end{equation}
Regarding the data structure, a three-dimensional tensor is used, where each row contains one impact cycle, the column dimension represents time and the channel dimension corresponds to the three sensor locations (\textit{pin, po, pdin}). The length of the data varies between different cycles.

Both localization networks are composed of 3 identical convolutional blocks, where each block contains a 1D convolutional layer followed by a BatchNorm and a ReLu activation layer. The 1D convolution operation has 10 filters, kernel size 8 and unit stride. The padding is chosen to match the output and input dimensions.
Regarding the CPA basis, the number of cells of the piecewise velocity function is set to 100 ($N_{\mathcal{P}}=100$) and the zero-boundary additional constraint is applied. The ODE solver uses 4 scaling-and-squaring iterations of the closed-form method presented in \cref{chapter:2}.

The classification module is composed of 3 identical convolutional blocks, where each block contains a 1D convolutional layer followed by a BatchNorm and a ReLu activation layer, as illustrated in \cref{fig:phm_method1}. The 1D convolution operation has 128 filters, kernel size 8 and unit stride. The padding is chosen to match the output and input dimensions. Aligned data $\bar{X}_{ref}^{align}$ and $X_{dat}^{align}$ are concatenated across the last dimension, so the classification module receives a three-dimensional tensor input with 6 channels (3 + 3), each corresponding to the three sensor locations (\textit{pin, po, pdin}). The output dimension is set to the number of classes (11).
The network was initialized with Xavier initialization \cite{glorot2010understanding} using a normal distribution and was trained for 100 epochs with $8e^{-5}$ learning rate, a batch size of $64$ and Adam \cite{kingma2014adam} optimizer with $\beta_{1}=0.9$, $\beta_{2}=0.999$ and $\epsilon=1e^{-8}$.

\subsection{Accuracy Results}
The hyperparameters of the loss function were optimized: $C_{ref}$ weights the variance of the aligned reference data $X_{ref}$, $C_{align}$ the variance of the raw data $X_{dat}$, and $C_{class}$ penalizes the classification cross-entropy. A grid-search was conducted from predefined sets of values: $C_{ref}=\{0,1,10,50,100\}$, $C_{align}=\{0,1,10,50,100\}$ and $C_{class}=\{1,10\}$.
\cref{tab:phm_hyperparameters} shows the train and test loss and accuracy values. Note how the baseline classification model without alignment ($C_{align}=0$, $C_{ref}=0$, $C_{class}=1$) scored significantly lower accuracy ($0.9715$) than the other cases where the alignment contributes in less or more measure ($C_{align}>0$ and $C_{ref}>0$). 
The grid-search analysis allowed to select the optimal weighting schemes, ensuring maximum accuracy. 
These experiments highlight the effectiveness of aligning time series data to reduce the risk of errors in classification by ensuring that the data is properly synchronized and accurately reflect the underlying patterns.

\small
\begingroup
\renewcommand\arraystretch{0.85}
\begin{longtable}{llr|rrrrrrrrrrrr}
    \caption{Grid-search analysis for the loss function hyperparameters}\label{tab:phm_hyperparameters}\\
    \toprule
    $C_{ref}$ &  $C_{align}$ &  $C_{class}$ &  $\mathcal{L}_{train}$ &  $\mathcal{L}_ {test}$ &  $\text{Accuracy}_{train}$ &  $\text{Accuracy}_{test}$ \\
    \midrule\endfirsthead
    \caption{(Cont.) Grid-search analysis for the loss function hyperparameters}\\
    \toprule
    $C_{ref}$ &  $C_{align}$ &  $C_{class}$ &  $\mathcal{L}_{train}$ &  $\mathcal{L}_{test}$ &  $\text{Accuracy}_{train}$ &  $\text{Accuracy}_{test}$ \\
    \midrule \endhead 
    0 &             0 &                   1 &     1.6047 &    1.6072 &     0.9695 &    0.9715 \\
    1 &               1 &                    1 &      1.5433 &     1.5434 &     1.0000 &    1.0000 \\
    1 &              10 &                    1 &      1.5439 &     1.5440 &     1.0000 &    0.9999 \\
    1 &              50 &                    1 &      1.5510 &     1.5514 &     0.9997 &    0.9993 \\
    1 &             100 &                    1 &      1.5485 &     1.5486 &     1.0000 &    1.0000 \\
    10 &               1 &                    1 &      1.5432 &     1.5433 &     1.0000 &    1.0000 \\
    10 &              10 &                    1 &      1.5440 &     1.5440 &     1.0000 &    1.0000 \\
    10 &              50 &                    1 &      1.5523 &     1.5531 &     0.9995 &    0.9990 \\
    10 &             100 &                    1 &      1.5486 &     1.5489 &     1.0000 &    0.9999 \\
    50 &               1 &                    1 &      1.5464 &     1.5476 &     0.9996 &    0.9989 \\
    50 &              10 &                    1 &      1.5499 &     1.5517 &     0.9991 &    0.9975 \\
    50 &              50 &                    1 &      1.5509 &     1.5522 &     0.9999 &    0.9997 \\
    50 &             100 &                    1 &      1.5525 &     1.5532 &     1.0000 &    0.9996 \\
    100 &               1 &                    1 &      1.5432 &     1.5434 &     1.0000 &    0.9999 \\
    100 &              10 &                    1 &      1.5439 &     1.5440 &     1.0000 &    1.0000 \\
    100 &              50 &                    1 &      1.5513 &     1.5522 &     0.9997 &    0.9996 \\
    100 &             100 &                    1 &      1.5536 &     1.5542 &     0.9997 &    0.9999 \\
    1 &               1 &                   10 &     15.4643 &    15.4763 &     0.9998 &    0.9989 \\
    1 &              10 &                   10 &     15.4842 &    15.4912 &     0.9992 &    0.9989 \\
    1 &              50 &                   10 &     15.4825 &    15.4859 &     0.9978 &    0.9980 \\
    1 &             100 &                   10 &     15.4838 &    15.4921 &     0.9987 &    0.9988 \\
    10 &               1 &                   10 &     15.4618 &    15.4683 &     0.9992 &    0.9989 \\
    10 &              10 &                   10 &     15.5214 &    15.5309 &     0.9961 &    0.9964 \\
    10 &              50 &                   10 &     15.4708 &    15.4798 &     0.9996 &    0.9990 \\
    10 &             100 &                   10 &     15.4588 &    15.4696 &     0.9998 &    0.9990 \\
    50 &               1 &                   10 &     15.4554 &    15.4659 &     0.9994 &    0.9984 \\
    50 &              10 &                   10 &     15.4611 &    15.4665 &     0.9999 &    0.9997 \\
    50 &              50 &                   10 &     15.4883 &    15.5000 &     0.9996 &    0.9986 \\
    50 &             100 &                   10 &     15.4850 &    15.4956 &     0.9998 &    0.9994 \\
    100 &               1 &                   10 &     15.4645 &    15.4740 &     0.9995 &    0.9989 \\
    100 &              10 &                   10 &     15.4529 &    15.4656 &     0.9999 &    0.9990 \\
    100 &              50 &                   10 &     15.4674 &    15.4814 &     0.9992 &    0.9982 \\
    100 &             100 &                   10 &     15.5046 &    15.5177 &     0.9994 &    0.9985 \\
\bottomrule
\end{longtable}
\endgroup
\normalsize

\begin{figure}[!htb]
    \begin{center}
    \centerline{\includegraphics[width=\linewidth]{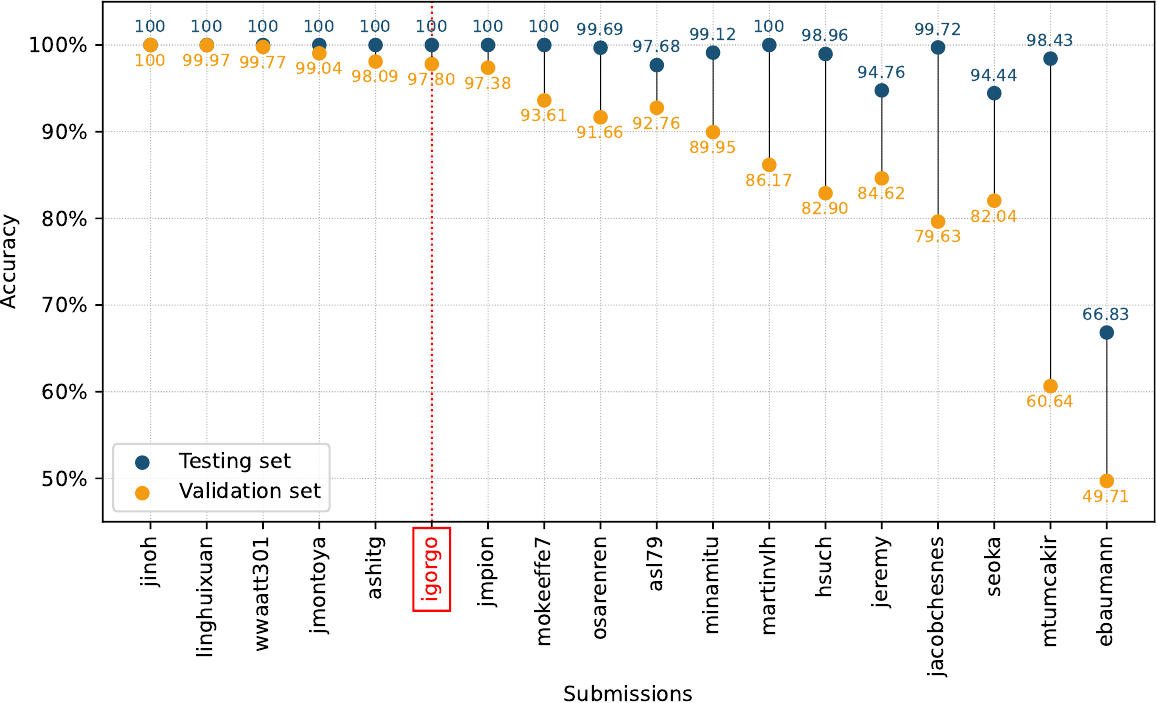}}
    \caption{Competition results in terms of accuracy. Test and validation sets. Extracted from \url{https://data.phmsociety.org/2022-phm-conference-data-challenge/}}
    \label{fig:phm_results}
    \end{center}
\end{figure}

In the competition, the proposed model ranked sixth out of 18 competitors, with an accuracy of $100\%$ and $97.80\%$ for the test and validation sets respectively (see \cref{fig:phm_results}, our submission was named \textit{igorgo}). Overall, the model has shown high accuracy in classifying faulty signals and has demonstrated its robustness to variability and complexity in real-world time series data.

\subsection{Within-class Variance Reduction}

\begin{figure}[!htb]
    \begin{center}
    \centerline{\includegraphics[width=\linewidth]{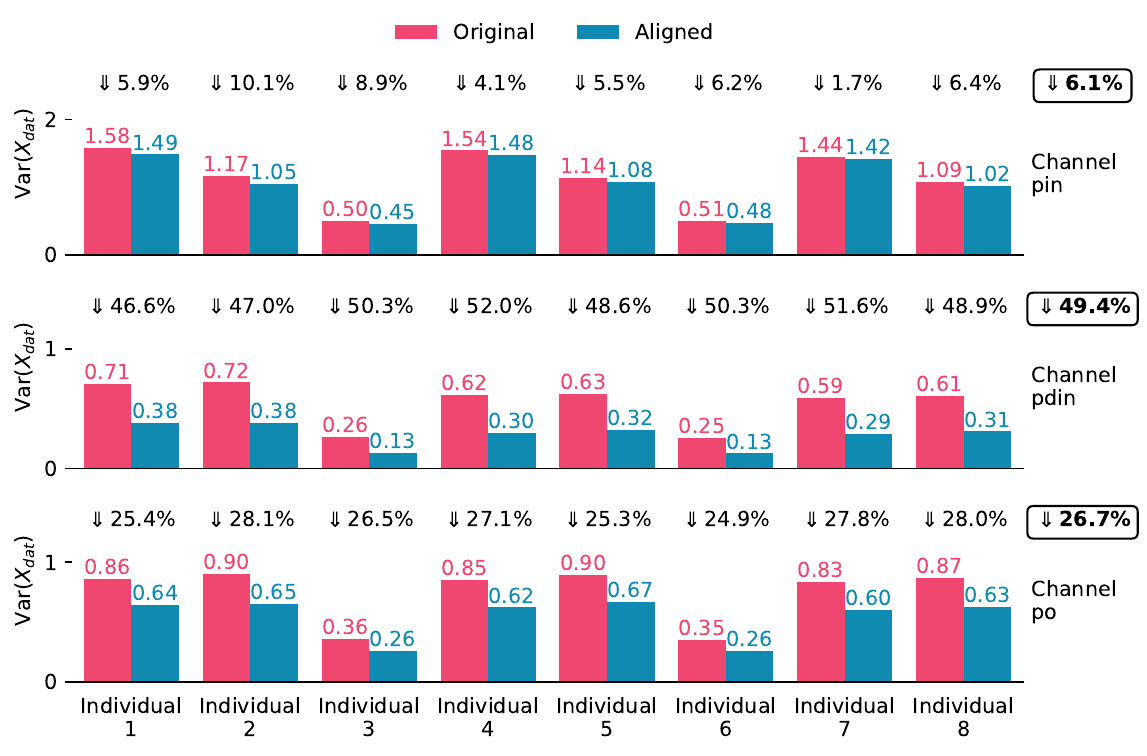}}
    \vspace{-0.3cm}
    \caption{Reduction of variance per individual and channel (test \& validation sets).}
    \label{fig:phm_variance}
    \end{center}
    \vspace{-1cm}
\end{figure}

This section explores the reduction of within-class variance after alignment. This statistic provides an idea of how much temporal misalignment has been repaired. Results for all  (individual, channel) tuples are reported in \cref{fig:phm_variance}. 
The distribution of variance reduction varies depending on the channel and the extent of temporal misalignment. It is important to note that the decrease achieved in the train set is carried over to the test and validation sets, indicating that the model has been able to learn and generalize the warping misalignment in the data to unseen samples.

\subsection{Qualitative Alignment Results}

\begin{figure}[!htb]
    \begin{center}
        \begin{subfigure}[t]{\linewidth}
            \centering
            \includegraphics[width=0.32\linewidth]{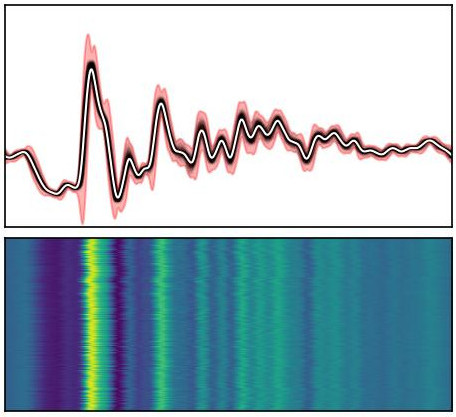}
            \includegraphics[width=0.32\linewidth]{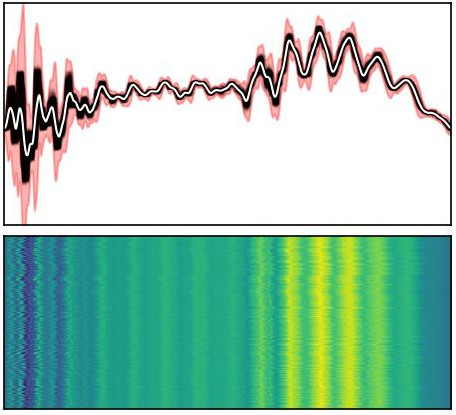}
            \includegraphics[width=0.32\linewidth]{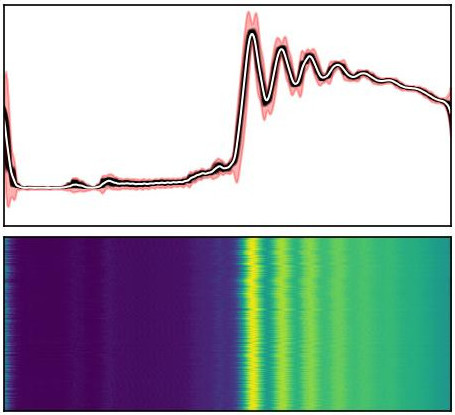}
            \caption{Original data (training set)}
            \label{fig:phm_alignment_1a}
        \end{subfigure}
        \begin{subfigure}[t]{\linewidth}
            \centering
            \includegraphics[width=0.32\linewidth]{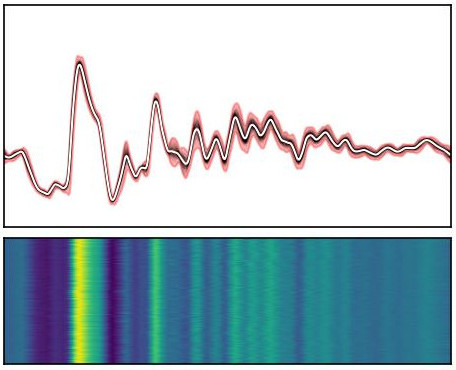}
            \includegraphics[width=0.32\linewidth]{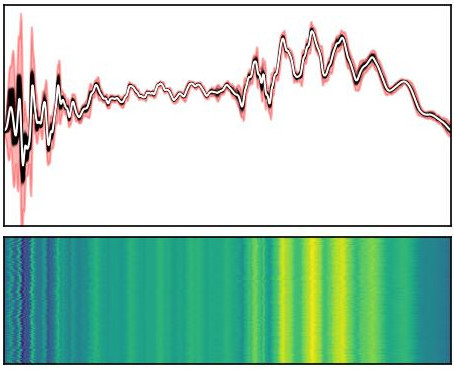}
            \includegraphics[width=0.32\linewidth]{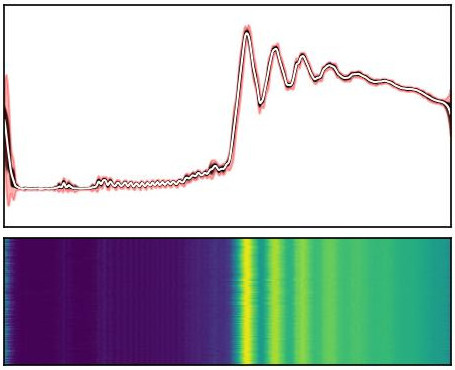}
            \caption{Aligned data (test set)}
            \label{fig:phm_alignment_1b}
        \end{subfigure}
        \caption{Pressure sensor data before and after alignment: detail of one individual (number 5), and 3 channels (left: \textit{pin}, center: \textit{pdin}, right: \textit{po}). Note the reduction in the standard deviation after alignment, represented by the shadow area in translucent red color.}
    \label{fig:phm_alignment_1}
    \end{center}
\end{figure}

\begin{figure}[!htb]
    \begin{center}
    \begin{subfigure}{\linewidth}
        \includegraphics[width=\linewidth]{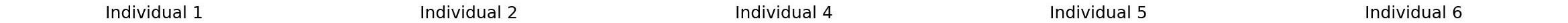}
        \includegraphics[width=\linewidth]{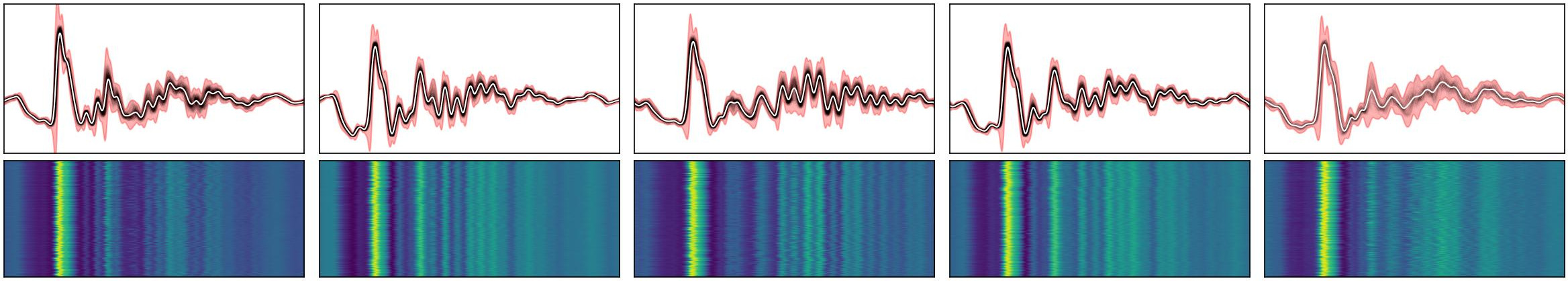}
        \includegraphics[height=0.161\linewidth,trim=0 0 900 0,clip]{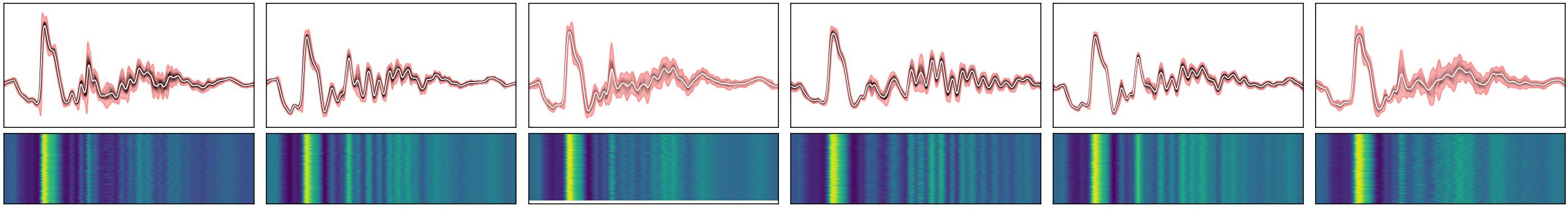}
        \includegraphics[height=0.161\linewidth,trim=675 0 0 0,clip]{figures/aligned/plot_heatmap_class_1_2.jpg}
        \caption{\textit{pin} sensor, with original data (training set) on top, and aligned data (test set) on the bottom.}
    \end{subfigure}
    \begin{subfigure}{\linewidth}
        \includegraphics[width=\linewidth]{figures/original/plot_heatmap_class_1_1.jpg}
        \includegraphics[width=\linewidth]{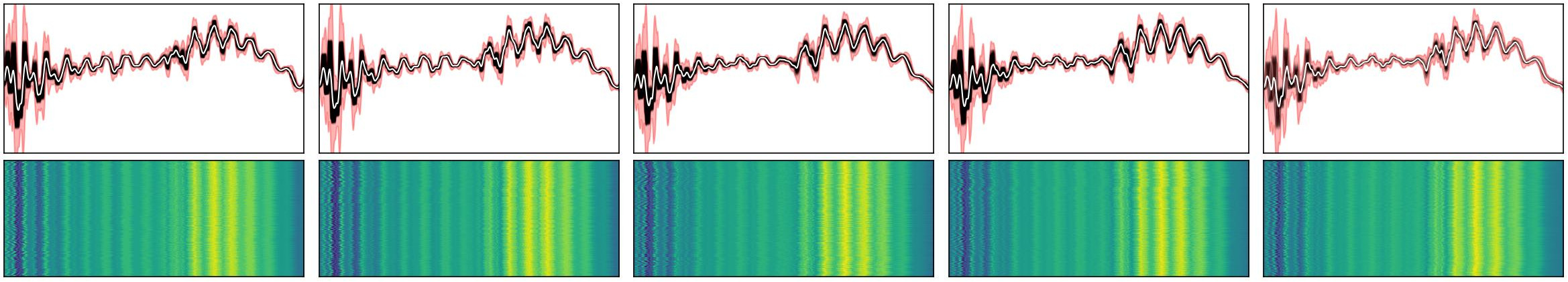}
        \includegraphics[height=0.159\linewidth,trim=0 0 900 0,clip]{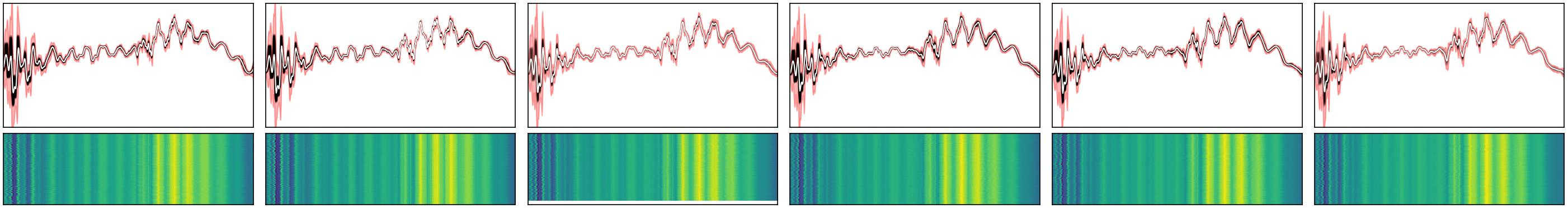}
        \includegraphics[height=0.159\linewidth,trim=675 0 0 0,clip]{figures/aligned/plot_heatmap_class_1_3.jpg}
        \caption{\textit{pdin} sensor, with original data (training set) on top, and aligned data (test set) on the bottom.}
    \end{subfigure}
    \begin{subfigure}{\linewidth}
        \includegraphics[width=\linewidth]{figures/original/plot_heatmap_class_1_1.jpg}
        \includegraphics[width=\linewidth]{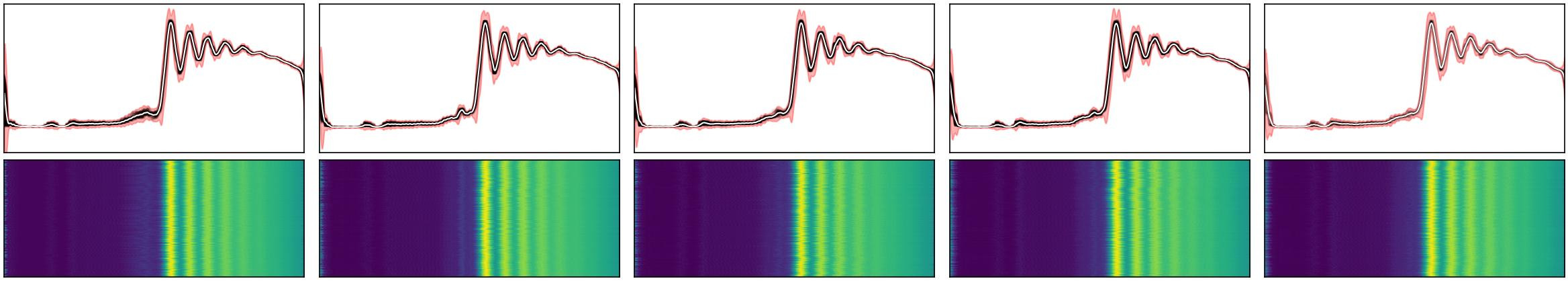}
        \includegraphics[height=0.159\linewidth,trim=0 0 900 0,clip]{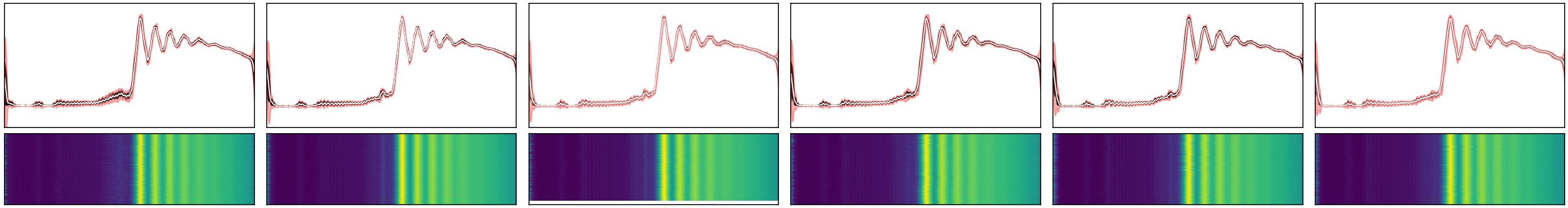}
        \includegraphics[height=0.159\linewidth,trim=675 0 0 0,clip]{figures/aligned/plot_heatmap_class_1_4.jpg}
        \caption{\textit{po} sensor, with original data (training set) on top, and aligned data (test set) on the bottom.}
    \end{subfigure}
    \caption{Pressure sensor data before and after alignment: multiple individuals (1,2,4,5,6), and 3 channels (top: \textit{pin}, middle: \textit{pdin}, bottom: \textit{po})}
    \label{fig:phm_alignment_2}
    \end{center}
\end{figure}

This section illustrates the visual differences before and after time series alignment. \cref{fig:phm_alignment_1a} represents the original training data before alignment, while \cref{fig:phm_alignment_1b} displays the aligned data from the test set.
The model learns to generalize alignments from the training set, so it does not need to solve a new optimization problem every time. The top graphic consists of a timeline of every time series instance (lines in \textcolor{gray}{gray}), and the Euclidean average (line in white) with one standard deviation shadow area (colored in translucent \textcolor{red}{red}). The heatmap on the bottom represents the time in the x-axis, each time series instance in the y-axis, and the color z-scale corresponds to the series value. This plot makes it easy to compare each time series, which may be overlapping at the top plot. Note the variance reduction between the left and right side. Results on more individuals were included in \cref{fig:phm_alignment_2}.

\clearpage
\section{Conclusions}\label{sec:conclusions_4}

This chapter presents an application of time series warping for time series classification using the methodological contributions discussed in \cref{chapter:2,chapter:3}.
The case study focuses on the fault diagnosis of a hydraulic rock drill using pressure data, and its performance is evaluated in the 2022 PHM Data Challenge. 
The results show that our deep learning-based solution achieved a noteworthy accuracy of 97.80\%.
The success of this approach can be attributed to the importance and effectiveness of aligning time series data for classification tasks. Accurate alignment reduces the risk of misinterpretation and misclassification, which results in more meaningful comparisons. Furthermore, the reduction in variance seen in the training set also exists in the test and validation sets, indicating that the model can learn and generalize the warping functions present in the data to new samples.

Overall, the results of this study demonstrate that time series classification has a wide range of applications and is a valuable tool for analyzing sequential data. 
By leveraging time series warping and deep learning techniques, a solution was developed that outperformed many competitors and demonstrated the potential of these approaches for solving complex problems. We believe that the combination of these two methodologies has a promising future in the field of time series classification and beyond.

\graphicspath{{content/chapter5/}}

\chapter{Incremental Time Series Clustering with Diffeomorphic Elastic Methods}\label{chapter:5}
\begingroup
\hypertarget{chapter5}{}
\hypersetup{linkcolor=black}
\setstretch{1.0}
\minitoc
\endgroup

\clearpage
\section{Introduction}\label{sec:introduction_5}

The preceding chapter focused on the supervised classification of time series using deep learning, wherein a set of time series instances is assigned to a fixed number of classes. Supervised learning techniques based on deep learning methods require large amounts of annotated data in order to be trained, and acquiring annotations can be challenging, preventing the training of these models. 
Clustering can be seen as an alternative to partition time series into homogeneous groups allowing a better analysis of the structure of the data. 
This chapter addresses the problem of whole-time series clustering, that aggregates and summarizes time series into a set of small, understandable, meaningful, and manageable representatives. 

\begin{wrapfigure}{r}{0.34\textwidth}
    \vspace{-0.8cm}
    \begin{center}
    \includegraphics[width=\linewidth]{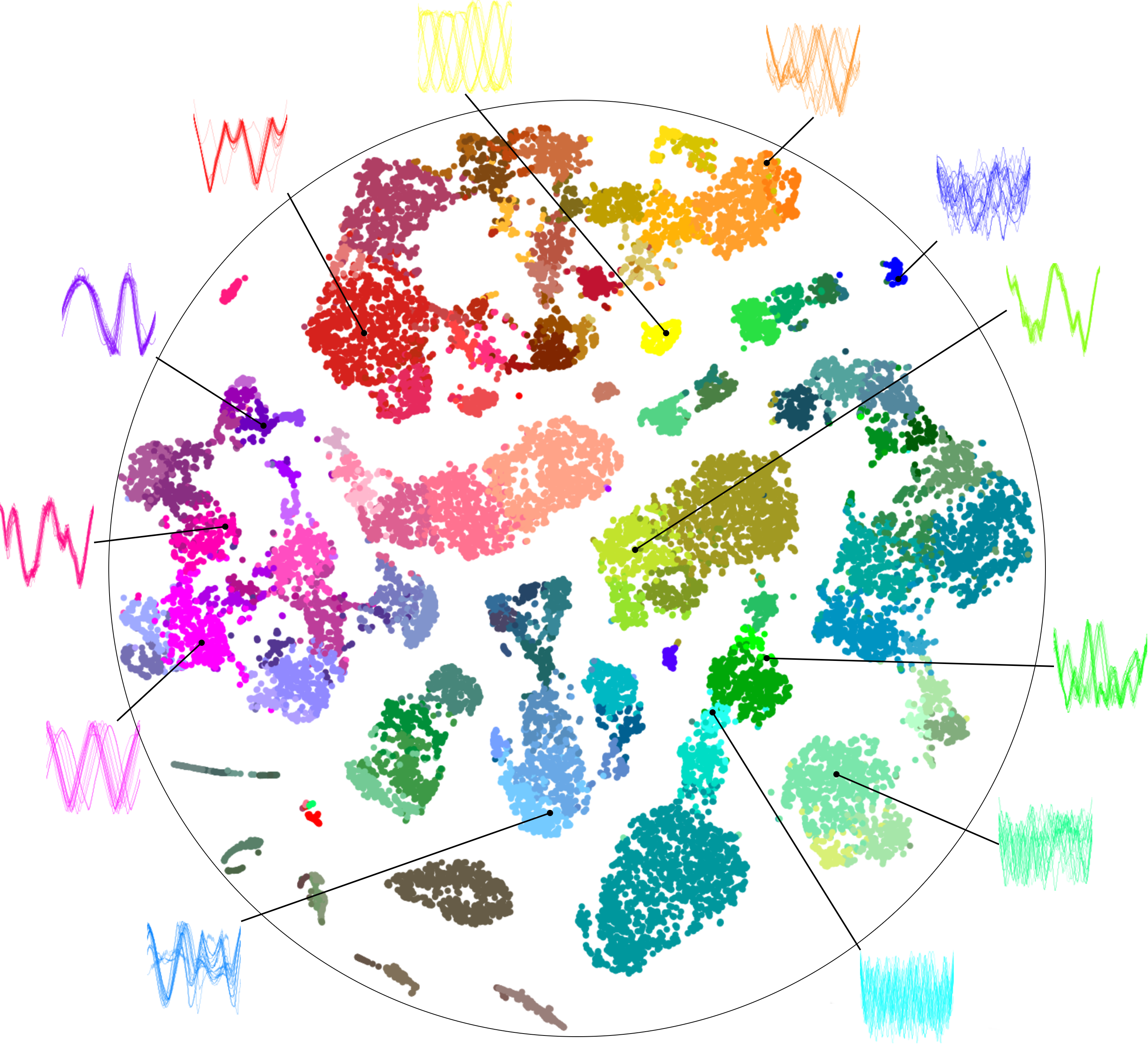}
    \caption{Time series clustering, t-SNE \cite{van2008visualizing} illustration. Each point represents a time series, colors different classes. 
    }
    \label{fig:time_series_clustering}
    \vspace{-2cm}
    \end{center}
\end{wrapfigure}

In recent years, there has been an increasing demand for clustering algorithms that can handle large datasets and high-dimensional data, especially in the context of time series analysis \cite{ding2015yading,aghabozorgi2014clustering,lin2004iterative,he2019fast}. 
Clustering algorithms typically require the calculation of pairwise distances between data points to determine their similarity. 
For time series data, the specificity of the time dimension makes the use of traditional clustering methods challenging. Indeed, each time step cannot be seen as an independent feature. Two time series can represent similar objects, but as we have seen throughout this dissertation, the time signal can be delayed, stretched, or subject to noise. This may result in high differences in the Euclidean space, even though time series denote a similar signal \cite{maharaj2019time} (see \cref{fig:time_series_clustering}).
In addition, the large number of time series instances and the high dimensionality of each time series instance makes clustering on large-scale time series more computationally expensive and memory-intensive. 

Another important challenge lies in the context of non-stationary distributions, also called concept drift \cite{Zliobaite2010, Webb2016}. Most clustering models live under the assumption that the systems they predict are static and stationary \cite{Gama2014}, neglecting that in reality, data generating processes are often dynamic and non-stationary. Unexpected changes in the environment or perturbed, incorrect and missing data can statistically detach the measured variable from the monitored feature, inducing to wrong decisions \cite{Hoens2012, Ditzler2015}.

This chapter proposes DIFW-IC, a novel incremental clustering algorithm for time series data that assigns each incoming time series into the nearest cluster using an elastic distance and is able to handle high-dimensional \& large datasets. The proposed method is based on a combination of elastic alignment and incremental clustering techniques. The algorithm starts by clustering a small subset of the data, known as initial buffer, and incrementally adds new time series points to the existing clusters. The assignment decision is based on the elastic distance of the new point to the existing clusters. Each cluster has a prototype time series (average) and a maximum boundary radius (standard deviation) that is analogous to the variance of all data points contained in the cluster. 
When the query time series does not match with any of the existing clusters, it is allocated to a new temporary group that can be updated with more incoming data. This helps to ensure that outliers do not skew the clustering results and that the algorithm can accurately identify clusters of similar time series.

We evaluate the performance of the proposed algorithm on several benchmark datasets and compare it with state-of-the-art clustering methods for time series data. The results show that the proposed algorithm outperforms existing ones in terms of clustering quality and scalability, and moreover, is also able to handle high-dimensional data and can identify the most relevant features for clustering the data.

This chapter is structured as follows: 
\cref{sec:time_series_clustering} introduces the reader to the problem of time series clustering; Work related to incremental time series clustering is discussed in \cref{sec:related_5}. Then, our proposed method is presented in \cref{sec:method_5}.  Experimental results are included in \cref{sec:results_5} and final remarks are included in \cref{sec:conclusions_5}.

\section{Time Series Clustering}\label{sec:time_series_clustering}

Unsupervised learning involves discovering structure in data without any prior labeling. 
\textbf{Cluster analysis} is a subset of unsupervised learning that aims to identify groups or clusters with high internal homogeneity and high external heterogeneity. Clustering complex objects such as time series is particularly advantageous because it leads to the discovery of interesting patterns, either frequent or rare ones, in the time series datasets.
Mathematically speaking, clustering is the general problem of partitioning a set of $N$ observations $\mathbf{X} = \{\mathbf{x}_{1}, \mathbf{x}_{2}, \cdots, \mathbf{x}_{N} \}$, where $\mathbf{x}_{i} \in \mathbb{R}^{d}$, into $K$ clusters, where a cluster is characterized with the notions of \textbf{homogeneity}—the similarity of observations within a cluster—and \textbf{separation}—the dissimilarity of observations from different clusters. The higher the similarity within a group and the dissimilarity between groups, the higher the clustering quality. Finding a global optimum for such problem would require trying all possible clusterings, which is infeasible even for small sets. In fact, this optimization problem is $\mathcal{NP}$-hard in Euclidean space even for two clusters \cite{garey1982complexity}, so several heuristics for finding local optima have been developed.

\clearpage

Cluster analysis can be divided into three stages: (1) determining a measure to quantify the similarity between observations, (2) choosing which method to use for obtaining the clustering and (3) selecting the desired number of clusters. The order in which these stages are executed depends on the clustering algorithm, for example, partitional methods require the number of clusters as input, while hierarchical ones decide on this number after the clustering is generated.

Applied to time series data, a fundamental point is to find a \textbf{suitable similarity measure} to quantify the distance between two observations. In this regard, \cref{chapter:2} presented a thorough literature review of similarity metrics for time series data.
For instance, the canonical similarity measure for points is the Euclidean distance.
However, Euclidean distance can be an extremely brittle distance measure, is very sensitive to small distortions in the time axis and may fail to produce an intuitively correct measure of similarity between two sequences. 

Hence, the high dimensionality of time series data presents a significant challenge for conventional clustering algorithms.
Large datasets make this task even more challenging because these algorithms are not designed to deal with data that is too large to fit in memory or is being continuously generated in real time. In this sense, incremental clustering methods allows new data to be added to an existing cluster over time, rather than starting with a completely new set of data. This allows the algorithm to adapt and evolve over time, making it useful for applications where the data is constantly changing or when it is not possible to cluster all the data at once due to its size or complexity. 
With incremental clustering, the existing clusters can be updated as new data becomes available, allowing for more accurate and up-to-date clusters.

\section{Related Work}\label{sec:related_5}

\textbf{$K$-means} \cite{steinhaus1956division} was first proposed in 1956, and it is still one of the most commonly used clustering methods. In general, the $K$-means aims to partition a set $\mathcal{X}=\{\mathbf{x}_1,\cdots,\mathbf{x}_N\}$ of $N$  vectors $\mathbf{x}_n\in \mathbb{R}^{d}$ into a predefined number $K$ of clusters $\mathcal{C}=\{\mathbf{c}_1, \cdots, \mathbf{c}_K\} \in \mathbb{R}^d$ by minimizing the distance between each $\mathbf{x}_i$ to the center $\mathbf{c}_k\in \mathbb{R}^d$ of the cluster $k$ it belongs to ---the optimal center $\mathbf{c}_k$ is the mean vector of the points assigned to cluster $k$. A cluster is identified by its center or centroid and consists of all points for which this centroid is closest. 
Formally, one wishes to minimize the within-cluster sum of squares (WCSS) objective function: 
$\{\mathbf{c}_1, \dots, \mathbf{c}_K\} \mapsfrom  \min \sum_{k=1}^K \sum_{i \in a_{k}} \mathcal{D}(\mathbf{x}_i, \mathbf{c}_k)$, 
where $a_{k}$ if the set of points that belong to cluster $k$ and $\mathcal{D}(x,y)=(x - y)^2$. 

\clearpage
A straightforward generalization of the above minimization problem is to choose a different metric $\mathcal{D}$. 
When Euclidean distance is used, the centroid can be computed with the arithmetic mean property. In many cases where alignment of observations is required, this problem is referred to as the multiple sequence alignment problem, which is known to be NP-complete \cite{srivastava2011registration}. The $K$-means clustering algorithm can be applied to time series with \textbf{elastic similarity metrics} by replacing the $\mathcal{D}(x,y)$ function and computing the cluster centroids, or barycenters, with respect to such metric \cite{sangalli2010k}. 
Recall that here barycenter is known as the average sequence from a group of time series in elastic space. 
With this approach, the centroids have an average shape that mimics the shape of the members of the cluster, regardless of where temporal shifts occur among the members.

\begin{algorithm}
    \caption{$K$-means, Lloyd-Forgy algorithm}
    \label{algo:kmeans}
    \renewcommand{\algorithmicrequire}{\textbf{Input:}}
    \renewcommand{\algorithmicensure}{\textbf{Return:}}
    \begin{algorithmic}[1]
        \Require Data $\mathcal{X}=\{\mathbf{x}_1,\cdots,\mathbf{x}_N\} \in \mathbb{R}^{N \times d}$, $K$ clusters, Centroids $\mathcal{C}=\{\mathbf{c}_k \in \mathbb{R}^d\}_{k=1}^{K}$
        \State $\tau \gets 0$ \Comment{Initialize number of steps}
        \Repeat
        \State $\tau \gets \tau + 1$ \Comment{Update number of steps}
        \State $a_{k} \gets \{i \;|\quad \mathcal{D}(\mathbf{x}_i,\mathbf{c}_k) \leq \mathcal{D}(\mathbf{x}_i,\mathbf{c}_j) \quad \forall j \in [K]\}$ \Comment{Assign points to nearest center}\label{line:kmeans:assignment}
        \ForAll {$k \in K$} 
        \State $\mathbf{c}_k \gets \frac{1}{|a_{k}|} \sum_{i \in a_{k}} {\mathbf{x}_i}$ \Comment{Update each cluster's center}\label{line:kmeans:compute_means}
        \EndFor
        \Until{stop criterion} \Comment{Repeat until stop criterion} 
        \Ensure $\mathcal{C}=\{\mathbf{c}_k\}_{k=1}^{K}$ the $K$-means of $N$ $d$-dimensional samples
    \end{algorithmic}
\end{algorithm}

The most popular procedure to approximately solve the $K$-means problem is \textbf{Lloyd-Forgy's algorithm}, which starts with an initialized set of $K$ center-points $\{\mathbf{c}_k \in \mathbb{R}^d\}_{k=1}^{K}$. Each update step $\tau$ is divided into two parts: (i) assignment step: all observations $\mathbf{x}_i$ are assigned to their nearest cluster based on the center-points $\mathbf{c}_k^{(\tau-1)}$s at this step. (ii) reestimation step: the new center-points $\mathbf{c}_k^{(\tau)}$s are computed as the means of the assigned $\mathbf{x}_i$.
Since the objective function decreases in every step, this algorithm is guaranteed to converge to a local optimum.

The cost of the assignment step is $\mathcal{O}(NdK)$, while that of the centers update is $\mathcal{O}(Nd)$, where $N$ is the number of $d$-dimensional observations and $K$ is the number of
clusters. Hence, the computational bottleneck lies in the assignment step at \cref{line:kmeans:assignment} of \cref{algo:kmeans}. This operation is repeated for a number of iterations $\tau$, giving the whole algorithm a time complexity of $\mathcal{O}(\tau NdK)$, which is sometimes considered to be "linear", due to $\tau, K, d \ll N$.

Lloyd's algorithm requires multiple passes over the data and thus isn't applicable in a streaming setting.
\textbf{Minibatch $K$-means} \cite{sculley2010web} modifies the $K$-means procedure to update the cluster centroid one example at a time, rather than all at once. This is particularly attractive when examples are acquired over a period of time, and one wants to start clustering before we have seen all the examples. This algorithm starts by finding the cluster that is closest to the current observation, and then it moves the cluster's central position towards the new observation based on the halflife parameter.
\cref{algo:minikmeans} shows how the minibatch $K$-means operates sequentially.
Also related, \textbf{Stream $K$-means} \cite{shindler2011fast} is an alternative version of Stream $L$-search by \cite{o2002streaming} that  uses an incremental $K$-means each time the temporary chunk of data points is full.

\begin{algorithm}
    \caption{Mini-batch $K$-means}
    \label{algo:minikmeans}
    \renewcommand{\algorithmicrequire}{\textbf{Input:}}
    \renewcommand{\algorithmicensure}{\textbf{Return:}}
    \begin{algorithmic}[1]
        \Require Data $\mathcal{X}=\{\mathbf{x}_1,\cdots,\mathbf{x}_N\} \in \mathbb{R}^{N \times d}$, $K$ clusters, \\ Mini-batch size $b$, Centroids $\mathcal{C}=\{\mathbf{c}_k \in \mathbb{R}^d\}_{k=1}^{K}$
        \State $v \gets 0$ \Comment{Initialize per-cluster counts}
        \Repeat
        \State $M \gets b$ examples picked randomly from $\mathcal{X}$
        \State $a_{k} \gets \{\mathbf{x}_i \;|\; \mathcal{D}(\mathbf{x}_i,\mathbf{c}_k) \leq \mathcal{D}(\mathbf{x}_i,\mathbf{c}_j) \quad \forall j \in [K]\}$ \Comment{Assign points to nearest center}\label{line:minikmeans:assignment}
        \ForAll {$k \in K$} 
        \State $v_k \gets v_k + |a_{k}|$ \Comment{Update per-cluster counts}
        \State $\eta \gets \frac{1}{v_k}$ \Comment{Get per-cluster learning rate}
        \State $\mathbf{c}_k \gets (1-\eta)\mathbf{c}_k  + \eta\sum_{i \in a_{k}} {\mathbf{x}_i}$ \Comment{Update each cluster's center}\label{line:minikmeans:compute_means}
        \EndFor
        \Until{stop criterion} \Comment{Repeat until stop criterion} 
        \Ensure $\mathcal{C}=\{\mathbf{c}_k\}_{k=1}^{K}$ the mini-batch $K$-means of $N$ $d$-dimensional samples
    \end{algorithmic}
\end{algorithm}

\textbf{CluStream} \cite{aggarwal2003framework} proposes to divide the clustering process into two components: 
The online component computes and stores summary statistics about the data stream using microclusters. The offline component answers queries using the stored summary statistics and a pyramidal time frame model.
The microclusters are updated incrementally as new data points arrive and represent a summary of the data stream (number, linear sum and squared sum), thus reducing the clustering's computational complexity. The algorithm can also handle evolving clusters, as it can detect changes in the data stream and update the microcluster accordingly. 
Clustream has good scalability as the online phase satisfies the one-pass constraint: access information about one data point only once. 

\textbf{DenStream} \cite{cao2006density} is based on the DBSCAN \cite{ester1996density} algorithm, and combines the concepts of microclusters \& density to discover clusters with arbitrary shape. 
In the first phase, incoming data points are assigned to microclusters that have properties such as center point, weight and variance. Only microclusters whose weight exceeds a certain threshold enter the second phase and are clustered by the DBSCAN algorithm. Thus, it is not necessary to retain the information about each data point, thus reducing memory and computing requirements. 
It is also possible to apply a fading function to the weights of the microclusters, that allows the algorithm to capture changes such as the movement of clusters or their disappearance/appearance over time.

\textbf{DBStream} \cite{hahsler2016clustering} is a microcluster-based online clustering component that explicitly captures the density between microclusters via a shared density graph. The density information in the graph is then exploited for re-clustering based on actual density between adjacent microclusters.
The algorithm includes a cleanup that removes weak microclusters and weak entries in the shared density graph to recover memory and improve the clustering algorithm's processing speed.

\textbf{Birch} \cite{zhang1996birch} constructs a hierarchical clustering structure in memory without needing to store all the data points in memory at once. Instead, it uses a tree-based data structure called the Clustering Feature Tree (CFT) that holds the necessary information for clustering: number of samples, linear sum, squared sum, centroids and squared norm of the centroids.
Birch can be viewed as an instance reduction method, since it reduces the input data to a set of subclusters which are obtained directly from the leaves of the CFT. This reduced data can be further processed by feeding it into a global clustering method. 
\noindent\rule{4cm}{0.2pt}

Overall, time series data are complex and high-dimensional, with intricate temporal dependencies that traditional clustering approaches may struggle to capture effectively. For instance, specific patterns may repeat at particular times of day or week, or exhibit long-term trends or cycles. Current clustering algorithms do not always consider these temporal dependencies, leading to suboptimal cluster assignments. 

The presented analysis indicates that current incremental clustering algorithms have shown some promise in addressing the challenges posed by time series data. However, there is still room for improvement in explicitly capturing the temporal dependencies. Elastic similarity metrics can help in using the temporal structure of the data, but their implementation is not straightforward, as it requires computing cluster centroids in an efficient and scalable manner. 
Other clustering algorithms rely on fixed-length representations of time series, such as feature vectors, which may also not be able to capture the full temporal structure of the data.
Therefore, incremental clustering algorithms still have significant potential for improvement in handling temporal dependencies effectively when applied to time series data.

\clearpage
\section{Incremental Time Series Clustering}\label{sec:method_5}

In this section, a novel incremental clustering algorithm for time series data is presented, called DIFW-IC (Diffeomorphic Fast Warping for Incremental Clustering). This algorithm assigns each incoming time series to the nearest cluster based on an elastic similarity distance and is capable of handling high-dimensional \& large datasets. 
Unlike traditional clustering algorithms that require the entire dataset to be loaded into memory, this algorithm is scalable and can process incoming high-rate data, whether univariate or multivariate, making it suitable for use in a variety of applications that require continuous monitoring and analysis.

The proposed method is a combination of elastic alignment and incremental clustering techniques, which together provide a robust and flexible approach. 
The algorithm is warmed-up through an offline process: using the beginning of data stream, an initial buffer of data is clustered using the elastic $K$-means method (see \cref{algo:myclustering_warmup}). Then, in the online phase, new time series points are added incrementally to existing clusters. \cref{algo:myclustering} shows the pseudocode for the proposed incremental clustering algorithm.

\begin{algorithm}
    \setstretch{0.95}
    \caption{DIFW-IC Warm-up: Elastic $K$-Means}
    \label{algo:myclustering_warmup}
    \renewcommand{\algorithmicrequire}{\textbf{Input:}}
    \renewcommand{\algorithmicensure}{\textbf{Return:}}
    \begin{algorithmic}[1]
    \Require{Data $\mathcal{X}_{B}=\{\mathbf{x}_1,\cdots,\mathbf{x}_{N_{B}}\} \in \mathbb{R}^{N_{B} \times d}$, $K$ clusters, $n_{init}$ initializations, $n_{iter}$ iterations}
    \State $\mathcal{C} \gets \{\,\}$, $\mathcal{A} \gets \{\,\}$ \Comment{Initialize centroids $\mathcal{C}$ and assignments $\mathcal{A}$}
    \State $\text{WCSS} \gets \infty$ \Comment{Initialize within cluster sum-of-squares}
    \For{$\tau \in [n_{init}]$}
        \ForAll {$k \in [K]$}
            \State $\mathbf{c}_{k}^{\tau} \gets \{\mathbf{x}_i \sim \mathcal{X}_{B}\}$ \Comment{Initial  $\mathcal{C}$ by random selection}
        \EndFor
        \ForAll {$j \in [n_{iter}]$}
            \For{$i \in [N_{B}]$}
                \State $k \gets \argmin_{k \in [K]} \mathcal{D}(\mathbf{x}_i,\mathbf{c}_{k}^{\tau})$ \Comment{Select closest cluster}
                \State $a_{k}^{\tau} \gets \{a_{k}^{\tau}, i\}$ \Comment{Assign time series $i$ to $a_{k}$}
            \EndFor
            \ForAll {$k \in [K]$}
                \State $\mathbf{c}_{k}^{\tau} \gets \frac{1}{|a_{k}^{\tau}|} \sum_{i \in a_{k}^{\tau}} {\mathbf{x}_i}$ \Comment{Update each cluster's centroids}
            \EndFor
        \EndFor
        \State $\text{WCSS}^{\tau} \gets \sum_{k=1}^K \sum_{i \in a_{k}^{\tau}} \mathcal{D}(\mathbf{x}_i, \mathbf{c}_k)$ \Comment{Compute within cluster sum-of-squares}
        \If{$\text{WCSS}^{\tau} < \text{WCSS}$}
        \State $\text{WCSS} \gets \text{WCSS}^{\tau}$\Comment{Update the best SSE}
        \State $\mathcal{C} \gets \{\mathbf{c}_{k}^{\tau}, \quad \forall k \in [K]\}$ \Comment{Update the best cluster centroids}
        \State $\mathcal{A} \gets \{a_{k}^{\tau}, \quad \forall k \in [K]\}$ \Comment{Update the best assignments}
        \EndIf
    \EndFor
    \Ensure{$\mathcal{C}$ cluster centroids and $\mathcal{A}$ cluster assignments}
    \end{algorithmic}
\end{algorithm}

Given a new incoming time series $\mathbf{x}$, the assignment decision is based on the elastic distance $\mathcal{D}$ of the new point to the existing clusters. This elastic distance measure $\mathcal{D}$ is based on the closed-form diffeomorphic transformations presented in \cref{chapter:2}, and takes into account the dynamic nature of time series data, which can often exhibit variations in both shape and amplitude over time. To obtain the aligned time series, the differentiable sampler from \cref{sec:linear_interpolation_grid} is used.

Each cluster stores relevant information: a prototype time series (elastic average) $\{\mathbf{c}_k\}_{k=1}^{K}$ and a maximum boundary radius (elastic standard deviation), that that is analogous to the scaled variance $\{\mathbf{v}_k\}_{k=1}^{K}$ of all time series contained in the cluster. This maximum boundary radius is controlled by an $R$ factor that multiplies the cluster's scaled variance.
At each iteration, the algorithm uses an elastic metric to compute the distance from the query time series to each existing cluster, then selects the nearest one, and finally updates the mean and standard deviation of the selected cluster. 

\begin{algorithm}
    \caption{DIFW-IC Incremental Clustering}
    \label{algo:myclustering}
    \renewcommand{\algorithmicrequire}{\textbf{Input:}}
    \renewcommand{\algorithmicensure}{\textbf{Return:}}
    \begin{algorithmic}[1]
        \Require Cluster mean $\mathcal{C}=\{\mathbf{c}_k\}_{k=1}^{K}$, scaled variance $\mathcal{V}=\{\mathbf{v}_k\}_{k=1}^{K}$, size $\mathcal{S}=\{s_k\}_{k=1}^{K}$, time series $\mathbf{x}$, radius threshold $R$, elastic distance $\mathcal{D}$, alignment function $\mathcal{F}$.
        \State $D \gets \{\,\}$, $Y \gets \{\,\}$ \Comment{Initalize candidate distances and aligned time series}
        \ForAll {$k \in [K]$}
        \State $d_{k} \gets \mathcal{D}(\mathbf{x},\mathbf{c}_{k})$ \Comment{Compute elastic distance}
        \If {$d_{k} \leq R \cdot \mathbf{v}_k \cdot s_k $} \Comment{Filter by radius threshold}
            \State $D \gets \{D, d_{k}\}$ \Comment{Save elastic distance}
            \State $Y \gets \{Y, \mathcal{F}(\mathbf{x},\mathbf{c}_{k})\}$ \Comment{Save aligned time series}
        \EndIf
        \EndFor
        \If{$Y \in \varnothing$} \Comment{If no valid clusters are found}
            \State $k \gets K +1$ \Comment{Create new cluster}
            \State $\mathcal{C} \gets \{\mathcal{C}, \mathbf{0}\}$ \Comment{Assign zero mean}
            \State $\mathcal{V} \gets \{\mathcal{V}, \mathbf{0}\}$ \Comment{Assign zero scaled variance}
        \Else
            \State $k \gets \argmin_{k} D$ \Comment{Otherwise, assign to the closest cluster}
        \EndIf 
        \State $\mathbf{y}_{k} \gets Y[k]$ \Comment{Get aligned time series}
        \State $\mathbf{\hat{c}}_k \gets \mathbf{c}_k$ \Comment{Save current mean}
        \State $\mathbf{c}_k \gets \mathbf{c}_k + (\mathbf{y}_{k} - \mathbf{c}_k)/s_{k}$ \Comment{Update cluster mean}\label{alg:cluster_mean}
        \State $\mathbf{v}_k \gets \mathbf{v}_k + (\mathbf{y}_{k} - \mathbf{\hat{c}}_k)(\mathbf{y}_{k} - \mathbf{c}_k)$ \Comment{Update cluster scaled variance}\label{alg:cluster_std}

        \Ensure{Updated cluster mean $\mathcal{C}$, scaled variance $\mathcal{V}$ and cluster size $\mathcal{S}$}
    \end{algorithmic}
\end{algorithm}

\clearpage
When the query time series does not lie within the boundary of its nearest cluster, it is allocated to a new cluster that can be updated with more incoming data. This may be the result of the new time series being an outlier or the emergence of a new natural cluster in the stream. The creation of new clusters ensures that outliers do not skew the clustering results and that the algorithm is able to accurately identify clusters of similar time series. These clusters can be updated with more incoming data, allowing the algorithm to adapt to changing data patterns over time.
To handle concept drift, the algorithm employs this adaptive online learning approach. This allows the algorithm to continuously update its clustering model as new data becomes available, adapting to changes in the underlying data distribution over time. By doing so, the algorithm is able to maintain its accuracy and effectiveness even in the face of evolving data patterns. 

Hence, at any moment a total of $K$ clusters are maintained. The value of $K$ is significantly larger than the natural number of clusters in the data set and also significantly smaller than the number of data points in the stream. This allows us to maintain a high level of temporal and spatial granularity while reducing the storage requirement. 
The maximum number of clusters during the execution of the algorithm depends on the distribution of the input data, but also on the setting of the radius parameter $R$. There are fewer clusters at higher values of $R$ as this results in coarser clusters that can contain time series from a wider range. In general, the number of outlier clusters increases at the beginning of the analysis, but decreases again when outlier clusters transform into clusters. If the patterns in the data stream do not change over time, the number of clusters settles after an initial phase. The more clusters there are, the higher the computing requirements: note that the elastic distance from the query time series to all existing clusters is computed at each iteration. A compromise must therefore be reached between the computing speed and the coarseness of the clusters. Other strategies could be adopted, for example, limiting the maximum number of clusters \cite{aggarwal2003framework}. 

In case the time series instance finds a cluster, that cluster representative (prototype or average) is updated with that information. It is very important that the averaging process is also incremental, so that we only need to have the current representative and the new time series member to update it, and not all the past-members of that cluster. Of course, there is a trade-off between the quality of the cluster average/representative and the computational and memory requirements to calculate the exact average.

The proposed algorithm requires an incremental update of both the cluster mean (\cref{alg:cluster_mean}) and the cluster variance (\cref{alg:cluster_std}). Note that both expressions make use of the aligned time series, and as a consequence, the arithmetic mean can be applied to obtain these statistics. 

\paragraph{Incremental Mean}
The Euclidean average or mean is often computed using the formula:
$\mu_{n} = \frac{1}{n} \sum_{i=1}^{n} x_{i}$.
When the number of items in the collection is small, this formula is not a big deal. However, as the number of items in the collection $n$ increases, a potential problem occurs: the sum of all the items gets large. If it gets too large, it might cause an overflow issue. More subtly, loss of precision may occur if there is a large difference in the magnitude of the sum compared to the number of items in the collection (especially when dealing with floating point numbers).

An incremental approach to the average computation can solve these issues. Recalling that the formula for the mean is the sum divided by the number of items, one can pull off the most recently added sample out of the sum with no change.
\begin{equation}
\mu_{n} = \frac{1}{n} \sum_{i=1}^{n} x_{i} = \frac{1}{n} \left(x_{n} + \sum_{i=1}^{n-1} x_{i}\right) = \frac{1}{n} \left(x_{n} + (n-1) \mu_{n-1}\right)
\end{equation}
Rearranging this formula leads to:
\begin{equation}\label{eq:incremental_mean}
\mu_{n} = \mu_{n-1} + \frac{x_{n} - \mu_{n-1}}{n}
\end{equation}
This expression appears intuitive at first glance; the new average is derived by adjusting the old average by a normalized component representing the degree to which the current sample deviates from the historical norm. This equation is more stable because it avoids the accumulation of, potentially, large sums.

\paragraph{Incremental Variance}
The classic formula for the variance is: $\sigma_{n}^{2} = \frac{1}{n} \sum_{i=1}^{n} \left( x_{i} - \mu_{n}\right)^{2}$
Similarly to the Euclidean average, the variance formula is not stable and can lead to overflow for a large collection of numbers. One can solve these issues by deriving an incremental variance formula. Let's define the scaled variance as $v_{n} = \sigma_{n}^{2} n$. Then:
\begin{equation}
    \begin{aligned}  
        v_{n} &= \sum_{i=1}^{n} \left( x_{i} - \mu_{n}\right)^{2} = \\ 
        &= \sum_{i=1}^{n} \left( x_{i}^2 - 2x_{i}\mu_{n}x_{i} + \mu_{n}^{2}\right) = \\
        &= \left(\sum_{i=1}^{n} x_{i}^2\right) - 2\mu_{n}\left(\sum_{i=1}^{n} x_{i}\right) + \mu_{n}^{2} \left(\sum_{i=1}^{n} 1\right) = \\
        &= \left(\sum_{i=1}^{n} x_{i}^2\right) - 2n\mu_{n}^{2} + \mu_{n}^2n = 
    \left(\sum_{i=1}^{n} x_{i}^2\right) - \mu_{n}^2n
\end{aligned}
\end{equation}

Then, one can use this result to generate a similar result from $n-1$ samples instead of $n$ points, and subtract the two of them to find the difference:
\begin{equation}
\begin{aligned} 
    v_{n} - v_{n-1} &= 
    \left(\sum_{i=1}^{n} x_{i}^2\right) - n\mu_{n}^{2} - 
    \left(\sum_{i=1}^{n-1} x_{i}^2\right) - n\mu_{n-1}^{2} = \\
    &= \left(\sum_{i=1}^{n} x_{i}^2 - \sum_{i=1}^{n-1} x_{i}^2\right) - 
    n\mu_{n}^{2} + (n-1)\mu_{n-1}^2 = \\
    &= x_{n}^{2} - n\mu_{n}^{2} + (n-1)\mu_{n-1}^2 = \\
    &= x_{n}^{2} - \mu_{n-1}^2 + n(\mu_{n-1}^{2} -\mu_{n}^2) = \\
    &= x_{n}^{2} - \mu_{n-1}^2 + n(\mu_{n-1} + \mu_{n})(\mu_{n-1} - \mu_{n})
\end{aligned}
\end{equation}

Recall back to \cref{eq:incremental_mean} for the incremental mean $\mu_{n} = \mu_{n-1} + (x_{n} - \mu_{n-1})/{n}$, which one can use apply to rearrange the previous formula.
\begin{equation}
\begin{aligned} 
v_{n} - v_{n-1} &= 
x_{n}^{2} - \mu_{n-1}^2 + (\mu_{n-1} + \mu_{n})(\mu_{n-1} - x_{n})= \\
&= x_{n}^{2} - \mu_{n-1}^2 + \mu_{n-1}^{2} - x_{n}\mu_{n-1} + \mu_{n}\mu_{n-1} - x_{n}\mu_{n} = \\ 
&= x_{n}^{2} + \mu_{n}\mu_{n-1} - x_{n}\mu_{n} - x_{n}\mu_{n-1} = \\
&= (x_{n} - \mu_{n-1})(x_{n} - \mu_{n})
\end{aligned}
\end{equation}
We now have an equation $v_{n}$ in terms of just the previous value, the new data point, and the previous means. From there, it's easy to divide by the number of samples and square root. 
\begin{equation}
    v_{n} = v_{n-1} + (x_{n} - \mu_{n-1})(x_{n} - \mu_{n})
\end{equation}
\begin{equation}
    \sigma_{n} = \sqrt{v_{n}/n}
\end{equation}

\section{Experiments and Results}\label{sec:results_5}

In this section, we assess the performance of DIFW-IC on several benchmark datasets and compare it with state-of-the-art clustering methods for time series data.
To evaluate the effectiveness of incremental time series clustering algorithms, it is essential to use appropriate experimental datasets. The dataset should be sufficiently large and complex to test the scalability and efficiency of the algorithm, and the data should be representative of real-world time series data. Furthermore, the dataset should include a sufficient and potentially unbounded number of clusters with varying sizes and shapes, to test the algorithm's ability to identify diverse patterns within the data. 

\subsection{Experimental Setup}

The proposed experimental setup comprises a data augmentation procedure and a non-parametric cluster generation process. Starting from a dataset with finite number of classes and time series instances, the data augmentation procedure increases the number of classes and creates data variability. Then, a stochastic non-parametric data generation process is used to generate new clusters.

\paragraph{Data Augmentation}
UCR univariate and multivariate \cite{dau2019ucr} time series classification archive is considered as the initial dataset. 
Data variability can be generated from a single dataset of $c$ classes by augmenting and combining existing samples. The proposed procedure goes as follows: To generate a new class, we first extract $n$ segments of different classes. Second, a coin is tossed to decide whether to flip horizontally and/or vertically each segment. Finally, all the segments are joined together across the time dimension to form the new sample. A sub-sampling algorithm is finally applied to reduce the data dimensionality to original size.

With this proposed procedure, an initial dataset of $c$ classes will yield $2^{n}\cdot 2^{n} \cdot n\cdot c$ new classes, where $n$ is the number of segments (hyperparameter). For instance, using a dataset of 4 classes and just 1 segment produces 16 classes (4 of which are the original ones), whereas increasing to 2 segments will yield 256 potential new classes.

\paragraph{Cluster generation}
A non-parametric data generation process is required to create a potential unbound number of clusters. For this purpose, the Chinese Restaurant Process (CRP) is used, a method for modeling random partitions of data, where each observation is assigned to one of an infinite number of possible clusters. 

The process works by imagining a restaurant with an infinite number of tables, each representing a potential cluster. The first customer enters the restaurant and sits at a randomly chosen table. The next customer then has the option to either join an existing table or start a new one, with the probability of selecting a table being proportional to the number of customers already sitting there. This process continues for an arbitrary number of customers, resulting in a random partition of the data into clusters. 

\begin{figure}[!htb]
    \begin{center}
    \includegraphics[width=0.9\linewidth]{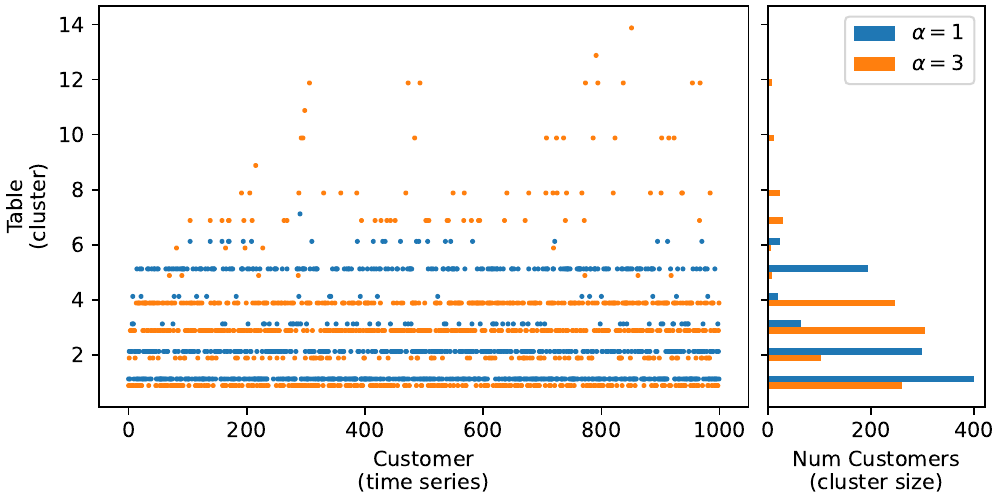}
    \caption{Chinese Restaurant Process (CRP). Effect of the concentration parameter $\alpha$.}
    \label{fig:crp}
    \end{center}
\end{figure}

In the CRP, the concentration parameter $\alpha$ controls the degree of clustering in the data. A higher value of $\alpha$ leads to more groups (see \cref{fig:crp}). 
The CRP is a powerful tool for modeling complex data structures that do not fit neatly into traditional statistical models. Its ability to generate random partitions with an infinite number of clusters makes it particularly useful for this data generation application, where the number of clusters is not known in advance.

\paragraph{Validation Methods}\label{sec:clustering_validation}

Any evaluation metric should not consider the absolute values of the cluster labels, but rather whether the clustering defines separations of the data similar to the ground truth set of classes or satisfies some assumption, for example: members of the same class are more similar than members of different classes according to a similarity metric.

External validation methods can be computed given the knowledge of the ground truth class $y$ and the assignments of the clustering algorithm for the same samples $\hat{y}$.  See \cref{tab:external_validation_methods} for a summary of the most relevant external validation methods: \textit{Fowlkes-Mallows Index, Adjusted Rand-Index (ARI), Adjusted Mutual Information, Homogeneity, Completeness and V-measure}.

All these metrics require knowledge of the ground truth classes which is almost never available in practice or requires manual assignment by human annotators. 
However, given the synthetic nature of the datasets used in the experiments, the ground truth class $y$ is available at run time, so external validation methods will be used.
It is important to note that some metrics do not yield zero-scores for random label assignments, for example, the unadjusted Rand-Index, the unadjusted Mutual Information or the V-measure. On the contrary, the Fowlkes-Mallows Index, the Adjusted Mutual Information and the Adjusted Rand-Index are normalized against random chance for any number of clusters and samples.

\begin{landscape}
    \begin{table}[!ht]
    \caption{External validation methods for $N$ points of dimensionality $d$ and $K$ clusters.}
    \label{tab:external_validation_methods}
        \begin{center}
        \resizebox*{\linewidth}{!}{%
        \begin{tabular}{lp{7cm}lp{7cm}}
        \toprule
        \textbf{Method} &
        \textbf{Description} &
        \textbf{Formula} &
        \textbf{Comments}
        \\ 

        \midrule 

        Pair Confusion Matrix & A $2 \times 2$ similarity matrix between two clusterings assignments $y$ and $\hat{y}$
        & 
        $M = \begin{bmatrix}
            \text{TN} & \text{FP} \\  \text{FN} & \text{TP}
        \end{bmatrix}$
        & 
        For example, FN (False Negatives) is the number of pairs with $y$ having the samples clustered together but  $\hat{y}$ not.
        \\

        Fowlkes-Mallows Index & Geometric mean of the pairwise precision and recall
        & 
        $\text{FMI}=\cfrac{\text{TP}}{\sqrt{(\text{TP}+\text{FP})(\text{TP}+\text{FN})}} \in [0,1]$
        & 
        Random label assignments have an FMI score close to zero for any number of clusters and samples.
        \\

        Adjusted Rand-Index & Measures the similarity of the ground truth $y$ and clustering assignments $\hat{y}$, ignoring permutations
        & 
        \begin{tabular}[t]{l}
        $\text{ARI} = \cfrac{\text{RI} - \mathbb{E}[\text{RI}]}{\max(\text{RI}) - \mathbb{E}[\text{RI}]} \in [-1,1]$ \\
        $\text{RI} = \cfrac{A + B}{W_N} \in [0,1]$ \\
        $W_N = \cfrac{N(N-1)}{2}$
        \end{tabular}
        & $A$ ($B$) is the number of pairs that are in the same (different) set in $y$ and in the same (different) set in $\hat{y}$, and $W_{N}$ is the total number of possible pairs in the dataset of size $N$.
        \\

        Adjusted Mutual Information & Measures the similarity of the ground truth $y$ and clustering assignments $\hat{y}$, ignoring permutations
        & 
        \begin{tabular}[t]{l}
        $\text{AMI} = \cfrac{\text{MI} - \mathbb{E}[\text{MI}]}{\mathrm{mean}(H(y), H(\hat{y}))- \mathbb{E}[\text{MI}]}$ \\
        $H(y)=-\sum_{i=1}^{|y|} P_i \log(P_i)$ \\
        $H(\hat{y})=-\sum_{j=1}^{|\hat{y}|} \hat{P}_j \log(\hat{P}_j)$ \\
        $\text{MI} = \sum_{i=1}^{|y|} \sum_{j=1}^{|\hat{y}|} P_{ij} \log\left(\cfrac{P_{ij}}{P_i\hat{P}_j}\right)$ \\
        $P_{ij} = |\hat{y}_{i} \cap y_{j}|/n$ \\
        $P_i = |y_{i}|/n$
        \end{tabular}
        & Entropy $H(y)$ is the amount of uncertainty for a partition set. $P_i$ is the probability that an object picked at random from $y$ falls into class $y_i$, whereas $P(i,j)$ is the probability that an object picked at random falls into both classes $y_{i}$ and $\hat{y}_{j}$.
        \\

        Homogeneity & Measures whether each cluster contains only members of a single class
        & 
        $h = 1 - \cfrac{H(y|\hat{y})}{H(y)} \in [0,1]$ 
        & 
        \\

        Completeness & Measures whether all members of a given class are assigned to the same cluster
        & 
        $c = 1 - \cfrac{H(\hat{y}|y)}{H(\hat{y})} \in [0,1]$
        & 
        \\

        V-measure & Harmonic mean of homogeneity $h$ and completeness $c$
        & 
        \begin{tabular}[t]{l}
        $v = \cfrac{(1+\beta) \cdot h \cdot c}{\beta \cdot h + c}  \in [0,1]$
        \end{tabular}
        &The parameter $\beta$ controls the balance between homogeneity and completeness.
        \\

        \bottomrule
        \end{tabular}%
        }
        \end{center}
    \end{table}
\end{landscape}

\paragraph{Experiments Hyperparameters}
As stated above, UCR time series classification archive \cite{dau2019ucr} is considered as the initial dataset. For the data augmentation process, the number of segments is $n=\{1,2\}$. The concentration parameter for the Chinese Restaurant Process is chosen among $\alpha=\{0, 1, 5, 10\}$, and the algorithm is tested with $10^4$ time series samples. For a given data generation process, the total number of clusters can be known in advance. This information is provided to some models such as $K$-means, that require to set the number of clusters prior to the algorithm execution.
Finally, the following external validation metrics are used: Adjusted Rand-Index, Adjusted Mutual Information, Completeness, Homogeneity, and V-measure.
The following models were implemented:
\begin{itemize}
    \item \textbf{Random}: predicts a random label from the cardinality of existing clusters.
    \item \textbf{Oracle}: predicts the correct cluster with 100\% accuracy.
    \item \textbf{DIFW-IC}: for the warm-up, buffer size $N_{B}=60$, buffer clusters $K=8$, $n_{init}=3$, $n_{iter}=5$; for the incremental clustering, radius threshold $R=2.0$ (equivalent to 99\% Gaussian quantile); and for the diffeomorphic elastic alignment, 
    $N_{\mathcal{P}} \in \{10, 20, 50, 100\}$,
    learning rate $\mu \in \{10^{-2}, 10^{-3}, 10^{-4}\}$,
    scaling-and-squaring iterations $\in \{2,4,6,8\}$,
    and the option to apply the zero-boundary constraint. 
    \item \textbf{$K$-means Euclidean}: the \textit{River} library \cite{montiel2021river} implementation is used with default hyperparameters $\textit{halflife}=0.5$, $\textit{mu}=0$, $\textit{sigma}=0.1$, and $\textit{p}=2$. The number of clusters matches those of the data generation process.
    \item \textbf{$K$-means elastic}: same diffeomorphic alignment hyperparameters as in DIFW-IC.
    \item \textbf{Stream $K$-means}: \textit{River} library \cite{montiel2021river} default implementation.
    \item \textbf{MiniBatch $K$-means}: \textit{Scikit-learn} library \cite{scikit-learn} implementation with hyperparameters $\textit{init}=\textit{'k-means++'}$, $\textit{max\_iter}=100$, $\textit{batch\_size}=1024$, $\textit{tol}=0$, $\textit{max\_no\_improvement}=10$, $\textit{n\_init}=3$, $\textit{reassignmentratio}=0.01$.
    \item \textbf{CluStream}: \textit{River} library \cite{montiel2021river} implementation with hyperparameters $\textit{time\_window}=100$, $\textit{max\_micro\_clusters}=50$, $\textit{micro\_cluster\_r\_factor}=2$, $\textit{halflife}=0.5$, $\textit{mu}=0$, $\textit{sigma}=0.05$, $\textit{p}=2$. Also, $\textit{n\_macro\_clusters}$ matches the number of clusters of the data generation process.
    \item \textbf{Denstream}: \textit{River} library \cite{montiel2021river} implementation with $\textit{decaying\_factor}=0.25$, $\textit{beta}=1$, $\textit{mu}=2$, $\textit{epsilon}=0.25$, $\textit{n\_samples\_init}=60$, $\textit{stream\_speed}=1$.
    \item \textbf{DBstream}: \textit{River} library \cite{montiel2021river} implementation with hyperparameters  $\textit{clustering\_threshold}=2.0$, $\textit{fading\_factor}=0.75$, $\textit{cleanup\_interval}=2$,\\ $\textit{intersection\_factor}=0.02$, $\textit{minimum\_weight}=20$.
    \item \textbf{Birch}: \textit{Scikit-learn} \cite{scikit-learn} implementation with hyperparameters $\textit{threshold}=0.9$, $\textit{branching\_factor}=50$.
\end{itemize}

\subsection{Results}

First, let's analyze the clustering results for CRP concentration values $\alpha=0$ and $\alpha=1$, and applied to one example dataset from the UCR univariate archive (Trace dataset).

\begin{figure}[!htb]
    \begin{center}
    \begin{subfigure}[b]{0.48\linewidth}
        \includegraphics[width=\linewidth]{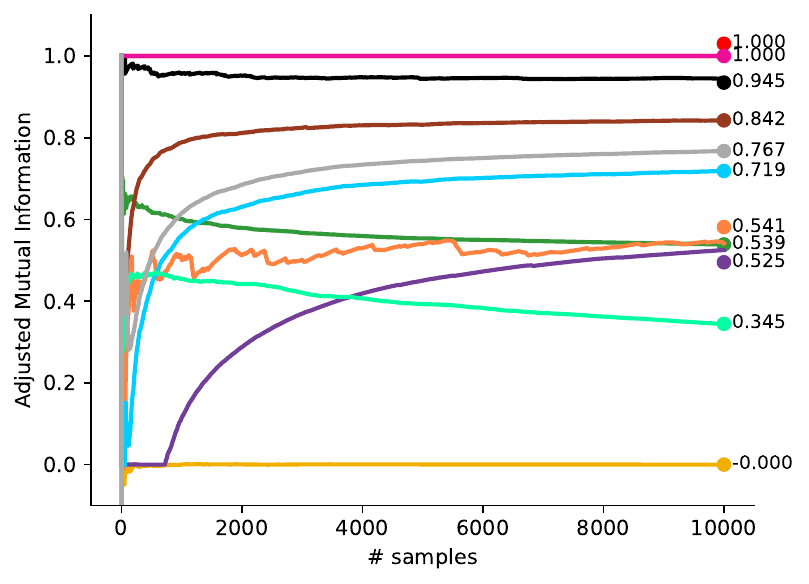}
        \caption{Adjusted Mutual Information }
    \end{subfigure}
    \begin{subfigure}[b]{0.48\linewidth}
        \includegraphics[width=\linewidth]{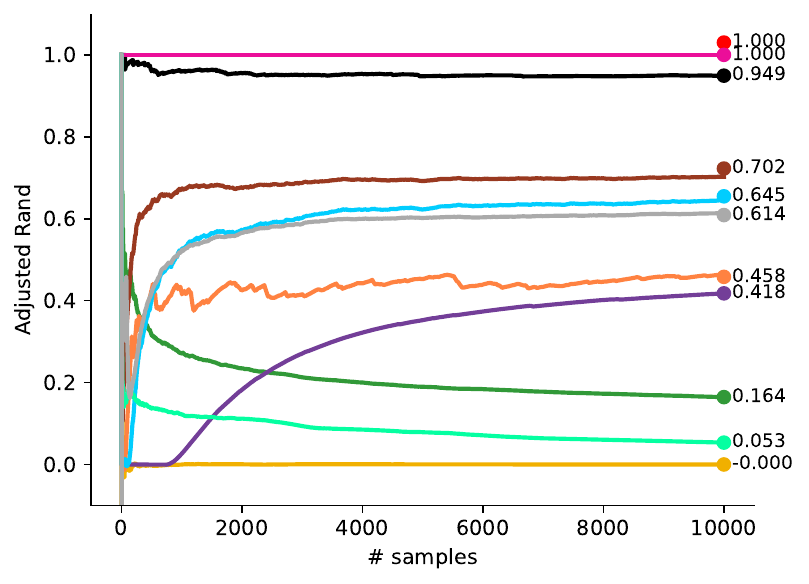}
        \caption{Adjusted Rand-index }
    \end{subfigure}
    \begin{subfigure}[b]{0.48\linewidth}
       \includegraphics[width=\linewidth]{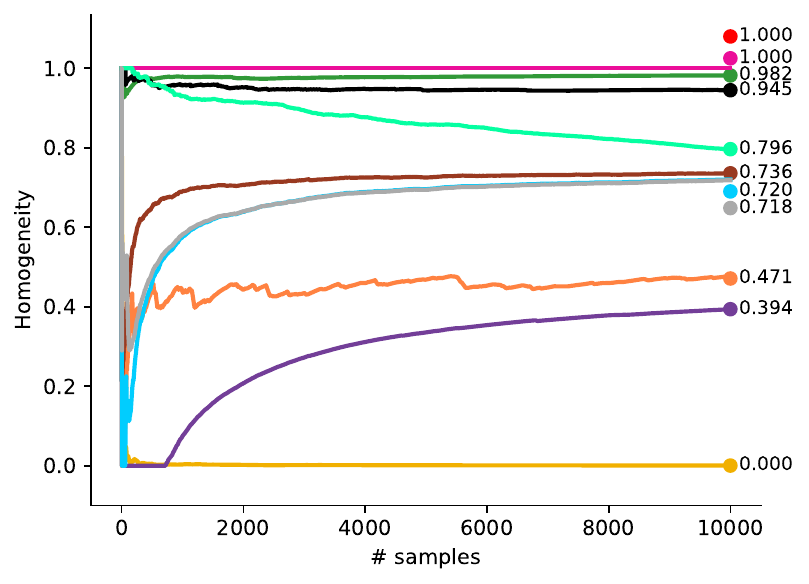}
        \caption{Homogeneity}
    \end{subfigure}
    \begin{subfigure}[b]{0.48\linewidth}
        \includegraphics[width=\linewidth]{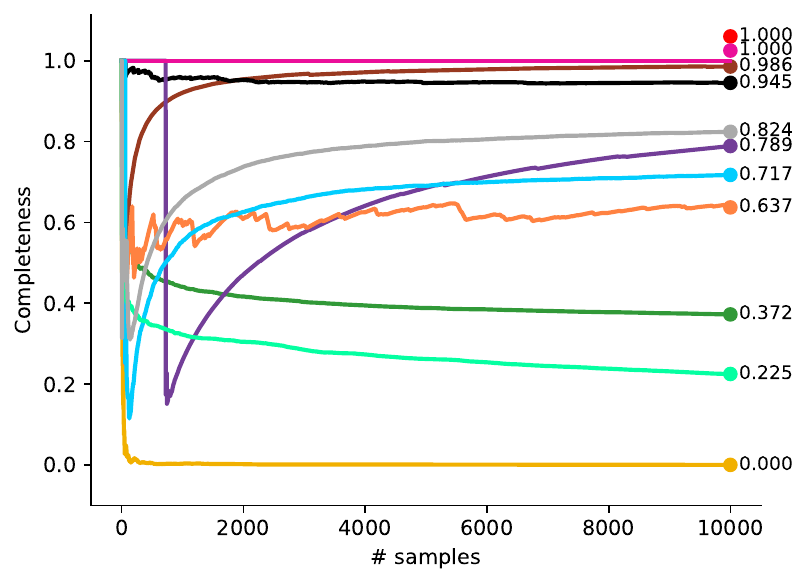}
        \caption{Completeness }
    \end{subfigure}
    \includegraphics[width=0.95\linewidth, trim=0 0 0 230, clip]{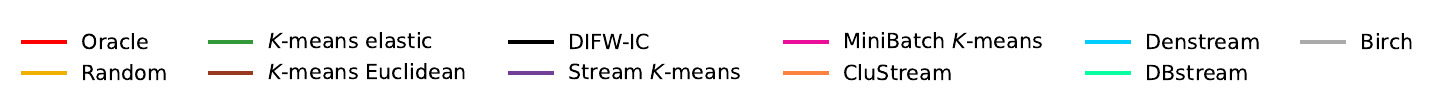}
    \caption{External validation metrics for CRP concentration value $\alpha=0$}
    \label{fig:plot_validation_0}
    \end{center}
\end{figure}

\paragraph{Case with $\boldsymbol{\alpha}\boldsymbol{=}\mathbf{0}$, dataset=Trace}
In this case there is no growth in the number of clusters regardless of how many time series samples are collected. 
In \cref{fig:plot_validation_0}, the evolution of external validation metrics is depicted in detail. 
In addition to the benchmark algorithms, the random and oracle models are included to be used as a reference.
The results show that including a priori information about the number of clusters helps static clustering algorithms to produce high-quality clusters that accurately capture the underlying structure of the data. Overall, DIFW-IC obtains very good quality clusterings, with metrics above 0.945.

In addition, it is worth noting the trade-off between homogeneity and completeness that is shown by some  models, for instance, Stream $K$ means, Elastic $K$-means, or DBstream.
Let's recall that homogeneity measures how well each cluster contains only data points that belong to the same class, while completeness measures how well all data points belonging to the same class are assigned to the same cluster.
The \textbf{trade-off between homogeneity and completeness} arises because maximizing one metric often leads to a decrease in the other. 

Specifically, a clustering algorithm that is optimized for homogeneity will tend to create clusters that are highly pure, meaning that all data points within a given cluster belong to the same class or category. However, such algorithm may also produce many small clusters that do not capture the overall structure of the data well. This is the case of the elastic version of $K$-means, that goes from a homogeneity of 0.982 to a completeness value of 0.382. 

In contrast, a clustering algorithm that is optimized for completeness, will tend to produce larger clusters that capture the overall structure of the data well. However, these clusters may contain data points from multiple classes, which can reduce the homogeneity of the clusters. An example of such case is the Stream $K$ means algorithm, that goes from a completeness of 0.789 to a homogeneity value of 0.394. 

Hence, it is important to have \textbf{consolidated metrics} such as the V-measure, and adjusted Rand index for evaluating this trade-off.
The benefit of homogeneity and completeness metrics is that they aid in understanding and  troubleshooting the behavior of clustering methods when applied to datasets of different nature and in a variety of hyperparameter settings.

\paragraph{Case with $\boldsymbol{\alpha}\boldsymbol{=}\mathbf{1}$, dataset=Trace}

\begin{figure}[!htb]
    \begin{center}
    \begin{subfigure}[b]{0.48\linewidth}
        \includegraphics[width=\linewidth]{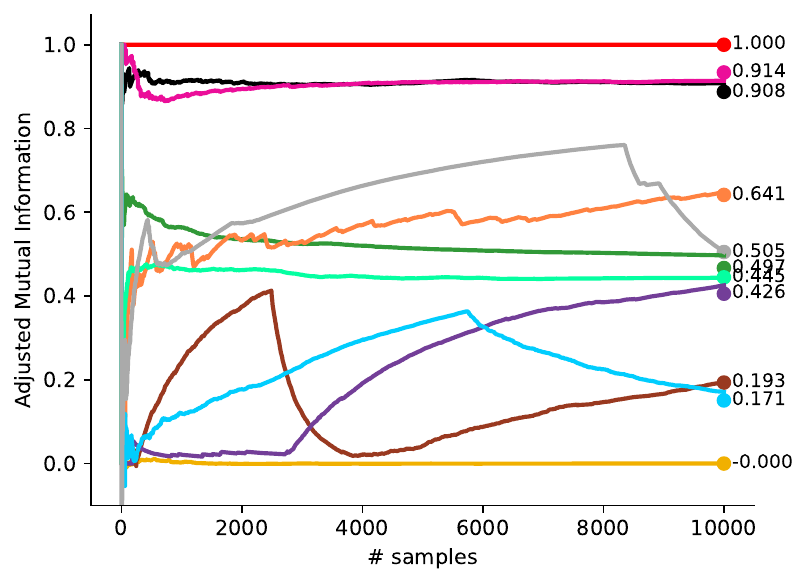}
        \caption{Adjusted Mutual Information }
    \end{subfigure}
    \begin{subfigure}[b]{0.48\linewidth}
        \includegraphics[width=\linewidth]{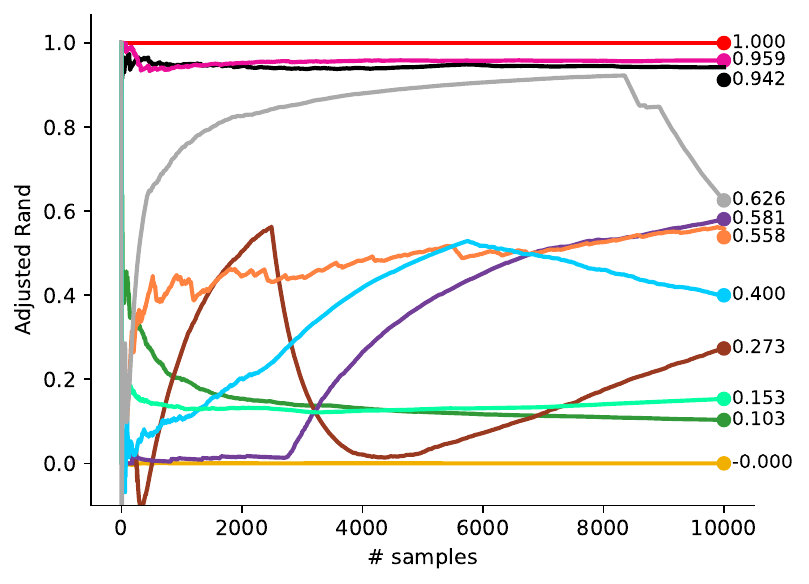}
        \caption{Adjusted Rand-index }
    \end{subfigure}
    \begin{subfigure}[b]{0.48\linewidth}
       \includegraphics[width=\linewidth]{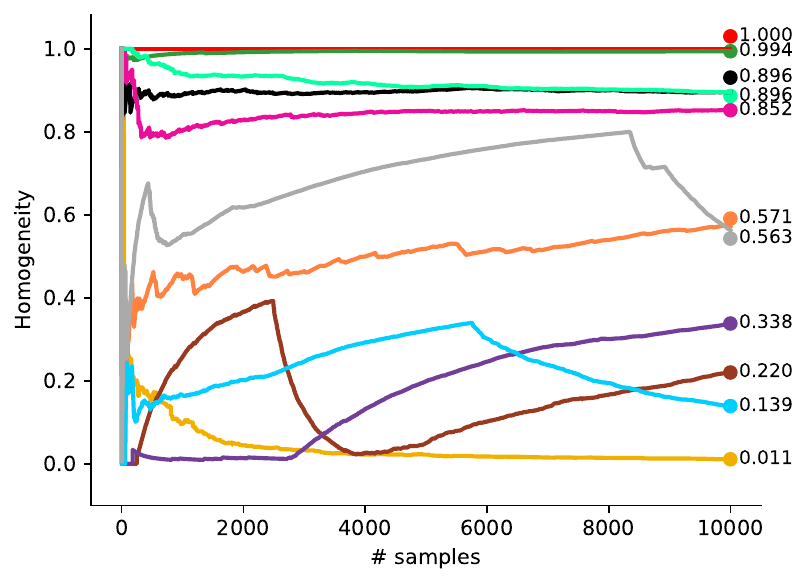}
        \caption{Homogeneity}
    \end{subfigure}
    \begin{subfigure}[b]{0.48\linewidth}
        \includegraphics[width=\linewidth]{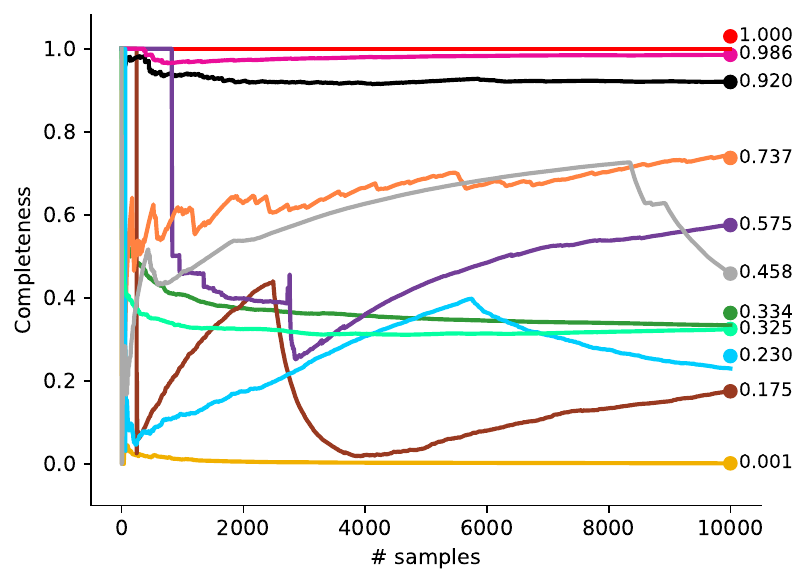}
        \caption{Completeness }
    \end{subfigure}
    \includegraphics[width=0.95\linewidth, trim=0 0 0 230, clip]{figures/plot_legend.pdf}
    \caption{External validation metrics for CRP concentration value $\alpha=1$}
    \label{fig:plot_validation_1}
    \end{center}
\end{figure}

When the CRP concentration value is set to $\alpha=1$, the number of clusters increases with more time series samples. In fact, the probability of the \textit{n}-th time series to be assigned to a new cluster is $\alpha / (\alpha + i - 1)$, and the expected number of clusters for $n$ time series grows logarithmically according to $\mathcal{O}(\alpha \log n$). 
In \cref{fig:plot_validation_1}, the mentioned external validation metrics are illustrated, as well as the random and oracle models to be use as a reference.

In terms of clustering performance, the \textbf{efficiency of the incremental algorithms is generally better} than the non-incremental, traditional ones. This is because they are designed to update the clustering model using small batches of data, which reduces the computational cost compared to processing the entire dataset at once. See \cref{tab:clustering_speed} for a comparison on the clustering speed of the analyzed algorithms.
Among the implemented algorithms, some are specifically designed for incremental clustering. These include DIFW-IC, Stream $K$-means, MiniBatch $K$-means, CluStream, DenStream, and DBStream. Other algorithms, such as $K$-means with Euclidean distance or Elastic distance, can also be adapted to incremental clustering, although they may not be as efficient as the dedicated incremental clustering algorithms.

DIFW-IC (with Adjusted Rand-Index ARI: 0.942) and MiniBatch $K$-means (ARI: 0.959) are both well-suited for large-scale datasets and have been shown to provide good results with reduced computational cost. While it is true that DIFW-IC is not as fast as MiniBatch $K$-means, its memory consumption is kept low. Other specialized algorithms for data streams do not obtain such good results for this specific dataset, for instance, Birch (ARI: 0.626), CluStream (ARI: 0.558) and DenStream (ARI: 0.400). Indeed, Birch shows a latter drop (at around 9000 time series samples) in all clustering metrics, which can be related to tree branching limitations.
\begin{table}[!htb]
    \caption{Incremental clustering algorithm speed in number of samples per second. The following computing infrastructure was used in these experiments: Intel® Core™ i7-6560U CPU @2.20GHz, 4 cores, 16GB RAM.}
    \label{tab:clustering_speed}
    \vspace{-1em}
    \begin{center}
    \resizebox{0.45\linewidth}{!}{%
    \begin{tabular}{lr}
        \toprule
        Algorithm &  Speed [samples/s] \\
        \midrule
        Oracle &              333.3 \\
        Random &              333.3 \\
        MiniBatch $K$-means &              166.7 \\
        $K$-means Euclidean &               95.2 \\
        Stream $K$-means &               95.2 \\
        Denstream &               83.3 \\
        Birch &               44.4 \\
        DIFW-IC &               33.3 \\
        $K$-means elastic &               28.6 \\
        CluStream &               18.5 \\
        DBstream &                5.6 \\
        \bottomrule
    \end{tabular}
    }
    \end{center}
    \vspace{-1em}
\end{table}

\vspace{-1em}
\paragraph{V-Measure on UCR archive, $\boldsymbol{\alpha}\boldsymbol{=}\mathbf{1}$}
\cref{tab:clustering_vmeasure} shows the V-measure on univariate and multivariate time series datasets from the UCR archive \cite{dau2019ucr} when the CRP concentration value is set to $\alpha = 1$. 
Results show that the proposed algorithm outperforms existing ones in terms of clustering quality and scalability.
The V-measure validation measure found that our method was better or no worse in 60\% of the datasets compared with MiniBatch $K$-means, and obtaining maximum V-measure at 59 out of the 158 datasets tested.

\footnotesize
\renewcommand{\arraystretch}{0.8}
\addtolength{\tabcolsep}{-2pt} 
\begin{longtable}{lcrrrrrrrr}
    \caption{V-measure benchmark on univariate and multivariate time series datasets from the UCR archive \cite{dau2019ucr}, for CRP concentration value $\alpha=1$.}\label{tab:clustering_vmeasure}\\
    \toprule
    Dataset & 
    \rotatebox{90}{\begin{tabular}[c]{@{}l@{}}DIFW-IC\end{tabular}} & 
    \rotatebox{90}{\begin{tabular}[c]{@{}l@{}}$K$-means\\Euclidean\end{tabular}} & 
    \rotatebox{90}{\begin{tabular}[c]{@{}l@{}}$K$-means\\elastic\end{tabular}} & 
    \rotatebox{90}{\begin{tabular}[c]{@{}l@{}}Stream\\$K$-means\end{tabular}} & 
    \rotatebox{90}{\begin{tabular}[c]{@{}l@{}}MiniBatch\\$K$-means\end{tabular}} & 
    \rotatebox{90}{\begin{tabular}[c]{@{}l@{}}CluStream\\\end{tabular}} & 
    \rotatebox{90}{\begin{tabular}[c]{@{}l@{}}Denstream\\\end{tabular}} & 
    \rotatebox{90}{\begin{tabular}[c]{@{}l@{}}DBstream\end{tabular}} & 
    \rotatebox{90}{\begin{tabular}[c]{@{}l@{}}Birch\end{tabular}} \\
    \midrule
    \endfirsthead
    \caption{(Cont.) V-measure benchmark on univariate and multivariate time series datasets from the UCR archive \cite{dau2019ucr}, for CRP concentration value $\alpha=1$.}\label{tab:clustering_vmeasure}\\
    \toprule
    Dataset & 
    \rotatebox{90}{\begin{tabular}[c]{@{}l@{}}DIFW-IC\\\end{tabular}} & 
    \rotatebox{90}{\begin{tabular}[c]{@{}l@{}}$K$-means\\Euclidean\end{tabular}} & 
    \rotatebox{90}{\begin{tabular}[c]{@{}l@{}}$K$-means\\elastic\end{tabular}} & 
    \rotatebox{90}{\begin{tabular}[c]{@{}l@{}}Stream\\$K$-means\end{tabular}} & 
    \rotatebox{90}{\begin{tabular}[c]{@{}l@{}}MiniBatch\\$K$-means\end{tabular}} & 
    \rotatebox{90}{\begin{tabular}[c]{@{}l@{}}CluStream\\\end{tabular}} & 
    \rotatebox{90}{\begin{tabular}[c]{@{}l@{}}Denstream\\\end{tabular}} & 
    \rotatebox{90}{\begin{tabular}[c]{@{}l@{}}DBstream\\\end{tabular}} & 
    \rotatebox{90}{\begin{tabular}[c]{@{}l@{}}Birch\\\end{tabular}} \\
    \midrule
    \endhead 
    Adiac & \textbf{0.730} & 0.396 & 0.566 & 0.522 & 0.696 & 0.584 & 0.408 & 0.559 & 0.560 \\
ArrowHead & \textbf{0.484} & 0.464 & 0.399 & 0.376 & 0.461 & 0.407 & 0.322 & 0.365 & 0.389 \\
Beef & 0.571 & 0.329 & 0.424 & 0.420 & 0.529 & 0.463 & 0.338 & 0.416 & \textbf{0.592} \\
BeetleFly & 0.520 & 0.319 & 0.427 & 0.417 & 0.513 & 0.458 & 0.319 & \textbf{0.573} & 0.422 \\
BirdChicken & 0.539 & 0.329 & 0.418 & 0.389 & \textbf{0.547} & 0.427 & 0.307 & 0.403 & 0.414 \\
Car & 0.611 & 0.353 & 0.469 & \textbf{0.636} & 0.579 & 0.510 & 0.518 & 0.492 & 0.480 \\
CBF & 0.644 & 0.364 & 0.490 & 0.478 & \textbf{0.682} & 0.488 & 0.504 & 0.502 & 0.488 \\
ChlorineConcentration & 0.509 & 0.312 & 0.399 & 0.391 & 0.503 & 0.398 & 0.316 & 0.396 & \textbf{0.716} \\
CinCECGTorso & 0.528 & 0.328 & 0.439 & 0.422 & 0.507 & 0.443 & 0.312 & 0.418 & \textbf{0.591} \\
Coffee & 0.579 & 0.332 & 0.466 & \textbf{0.583} & 0.563 & 0.481 & 0.341 & 0.419 & 0.471 \\
Computers & 0.512 & 0.317 & 0.424 & 0.383 & 0.506 & 0.423 & 0.311 & 0.390 & \textbf{0.566} \\
CricketX & 0.515 & 0.322 & 0.393 & 0.365 & 0.503 & \textbf{0.550} & 0.322 & 0.395 & 0.381 \\
CricketY & \textbf{0.655} & 0.522 & 0.504 & 0.454 & 0.622 & 0.521 & 0.363 & 0.516 & 0.499 \\
CricketZ & 0.590 & 0.359 & 0.480 & 0.466 & \textbf{0.607} & 0.492 & 0.508 & 0.458 & 0.486 \\
DiatomSizeReduction & \textbf{0.544} & 0.320 & 0.441 & 0.419 & 0.538 & 0.468 & 0.333 & 0.424 & 0.457 \\
DistalPhalanxOutlineAgeGroup & 0.569 & 0.363 & 0.501 & 0.447 & \textbf{0.581} & 0.474 & 0.349 & 0.430 & 0.470 \\
DistalPhalanxOutlineCorrect & 0.562 & 0.335 & 0.592 & 0.450 & 0.576 & \textbf{0.627} & 0.352 & 0.459 & 0.470 \\
DistalPhalanxTW & 0.616 & 0.352 & 0.471 & 0.459 & 0.603 & \textbf{0.637} & 0.355 & 0.442 & 0.628 \\
Earthquakes & 0.632 & 0.376 & 0.522 & 0.457 & \textbf{0.633} & 0.550 & 0.372 & 0.511 & 0.535 \\
ECG200 & \textbf{0.674} & 0.386 & 0.537 & 0.506 & 0.661 & 0.597 & 0.411 & 0.537 & 0.512 \\
ECG5000 & \textbf{0.570} & 0.351 & 0.475 & 0.421 & 0.563 & 0.512 & 0.342 & 0.435 & 0.471 \\
ECGFiveDays & 0.656 & 0.348 & 0.509 & 0.480 & 0.645 & 0.554 & 0.388 & 0.474 & \textbf{0.693} \\
ElectricDevices & 0.604 & 0.354 & 0.490 & 0.475 & \textbf{0.643} & 0.536 & 0.368 & 0.508 & 0.490 \\
FaceAll & \textbf{0.739} & 0.376 & 0.585 & 0.567 & 0.734 & 0.585 & 0.396 & 0.545 & 0.593 \\
FaceFour & 0.570 & \textbf{0.645} & 0.482 & 0.436 & 0.565 & 0.499 & 0.372 & 0.445 & 0.486 \\
FacesUCR & \textbf{0.681} & 0.389 & 0.529 & 0.535 & 0.646 & 0.582 & 0.372 & 0.510 & 0.536 \\
FiftyWords & 0.676 & 0.381 & 0.504 & 0.506 & \textbf{0.703} & 0.552 & 0.405 & 0.687 & 0.535 \\
Fish & \textbf{0.568} & 0.339 & 0.426 & 0.465 & 0.554 & 0.488 & 0.503 & 0.431 & 0.430 \\
FordA & \textbf{0.658} & 0.366 & 0.490 & 0.481 & 0.635 & 0.540 & 0.373 & 0.487 & 0.535 \\
FordB & 0.631 & 0.378 & 0.493 & \textbf{0.634} & 0.605 & 0.558 & 0.402 & 0.471 & 0.519 \\
GunPoint & \textbf{0.737} & 0.715 & 0.713 & 0.561 & 0.711 & 0.582 & 0.419 & 0.607 & 0.681 \\
Ham & \textbf{0.606} & 0.378 & 0.515 & 0.455 & 0.590 & 0.558 & 0.384 & 0.491 & 0.496 \\
HandOutlines & \textbf{0.587} & 0.341 & 0.432 & 0.466 & 0.560 & 0.477 & 0.357 & 0.447 & 0.463 \\
Haptics & 0.713 & 0.389 & 0.569 & 0.672 & 0.681 & \textbf{0.726} & 0.393 & 0.566 & 0.564 \\
Herring & 0.764 & 0.552 & 0.572 & 0.669 & \textbf{0.769} & 0.750 & 0.382 & 0.576 & 0.598 \\
InlineSkate & \textbf{0.492} & 0.319 & 0.432 & 0.407 & 0.482 & 0.428 & 0.315 & 0.372 & 0.426 \\
InsectWingbeatSound & \textbf{0.609} & 0.351 & 0.467 & 0.453 & 0.591 & 0.512 & 0.355 & 0.465 & 0.484 \\
ItalyPowerDemand & \textbf{0.675} & 0.373 & 0.545 & 0.492 & 0.666 & 0.558 & 0.385 & 0.497 & 0.532 \\
LargeKitchenAppliances & 0.648 & 0.525 & 0.643 & 0.480 & 0.644 & 0.520 & 0.389 & 0.521 & \textbf{0.666} \\
Lightning2 & 0.558 & 0.353 & 0.459 & 0.445 & \textbf{0.577} & 0.461 & 0.343 & 0.401 & 0.450 \\
Lightning7 & \textbf{0.578} & 0.333 & 0.438 & 0.448 & 0.561 & 0.504 & 0.347 & 0.466 & 0.474 \\
Mallat & 0.589 & 0.328 & 0.455 & 0.415 & 0.594 & 0.490 & 0.349 & \textbf{0.596} & 0.460 \\
Meat & \textbf{0.621} & 0.367 & 0.477 & 0.453 & 0.584 & 0.467 & 0.357 & 0.464 & 0.504 \\
MedicalImages & 0.574 & 0.489 & 0.437 & 0.451 & \textbf{0.590} & 0.474 & 0.345 & 0.457 & 0.477 \\
MiddlePhalanxOutlineAgeGroup & 0.704 & 0.405 & 0.566 & 0.542 & \textbf{0.728} & 0.597 & 0.425 & 0.548 & 0.582 \\
MiddlePhalanxOutlineCorrect & \textbf{0.528} & 0.307 & 0.434 & 0.394 & 0.490 & 0.443 & 0.323 & 0.395 & 0.421 \\
MiddlePhalanxTW & \textbf{0.703} & 0.398 & 0.544 & 0.490 & 0.699 & 0.550 & 0.407 & 0.499 & 0.568 \\
MoteStrain & 0.487 & 0.332 & 0.421 & 0.426 & \textbf{0.493} & 0.449 & 0.305 & 0.406 & 0.451 \\
NonInvasiveFetalECGThorax1 & 0.578 & 0.323 & 0.477 & 0.437 & \textbf{0.619} & 0.504 & 0.359 & 0.454 & 0.484 \\
NonInvasiveFetalECGThorax2 & 0.595 & 0.364 & 0.468 & 0.484 & 0.607 & 0.531 & 0.356 & \textbf{0.621} & 0.483 \\
OliveOil & 0.510 & 0.313 & 0.404 & 0.433 & 0.489 & \textbf{0.574} & 0.321 & 0.400 & 0.417 \\
OSULeaf & 0.692 & 0.545 & 0.537 & 0.531 & \textbf{0.698} & 0.612 & 0.389 & 0.526 & 0.595 \\
PhalangesOutlinesCorrect & \textbf{0.619} & 0.353 & 0.470 & 0.459 & 0.602 & 0.549 & 0.375 & 0.508 & 0.483 \\
Phoneme & 0.482 & 0.329 & 0.431 & 0.385 & \textbf{0.488} & 0.419 & 0.305 & 0.391 & 0.406 \\
Plane & 0.549 & 0.318 & 0.414 & 0.401 & 0.509 & 0.457 & 0.314 & \textbf{0.584} & 0.426 \\
ProximalPhalanxOutlineAgeGroup & 0.728 & 0.405 & 0.585 & 0.707 & \textbf{0.748} & 0.615 & 0.566 & 0.564 & 0.610 \\
ProximalPhalanxOutlineCorrect & \textbf{0.571} & 0.504 & 0.420 & 0.400 & 0.537 & 0.474 & 0.327 & 0.445 & 0.443 \\
ProximalPhalanxTW & 0.615 & 0.345 & 0.486 & 0.459 & 0.577 & \textbf{0.650} & 0.365 & 0.447 & 0.630 \\
RefrigerationDevices & 0.561 & 0.346 & 0.427 & 0.429 & 0.575 & 0.472 & 0.344 & 0.435 & \textbf{0.613} \\
ScreenType & \textbf{0.536} & 0.298 & 0.420 & 0.403 & 0.517 & 0.441 & 0.318 & 0.402 & 0.450 \\
ShapeletSim & 0.530 & 0.328 & 0.440 & \textbf{0.584} & 0.510 & 0.440 & 0.497 & 0.430 & 0.449 \\
ShapesAll & 0.527 & 0.322 & 0.406 & 0.407 & 0.514 & 0.461 & 0.322 & \textbf{0.569} & 0.437 \\
SmallKitchenAppliances & \textbf{0.607} & 0.495 & 0.455 & 0.416 & 0.559 & 0.488 & 0.345 & 0.438 & 0.470 \\
SonyAIBORobotSurface1 & 0.541 & 0.338 & 0.425 & 0.436 & \textbf{0.564} & 0.459 & 0.343 & 0.387 & 0.457 \\
SonyAIBORobotSurface2 & \textbf{0.696} & 0.392 & 0.503 & 0.508 & 0.664 & 0.578 & 0.384 & 0.658 & 0.490 \\
StarLightCurves & \textbf{0.745} & 0.412 & 0.552 & 0.531 & 0.675 & 0.568 & 0.367 & 0.509 & 0.509 \\
Strawberry & 0.735 & 0.388 & 0.541 & 0.533 & \textbf{0.741} & 0.606 & 0.415 & 0.549 & 0.560 \\
SwedishLeaf & 0.591 & 0.358 & \textbf{0.630} & 0.458 & 0.563 & 0.524 & 0.353 & 0.442 & 0.472 \\
Symbols & \textbf{0.796} & 0.411 & 0.717 & 0.549 & 0.703 & 0.773 & 0.416 & 0.571 & 0.596 \\
SyntheticControl & 0.667 & 0.366 & 0.495 & 0.493 & 0.635 & \textbf{0.726} & 0.368 & 0.475 & 0.501 \\
ToeSegmentation1 & \textbf{0.593} & 0.341 & 0.446 & 0.431 & 0.570 & 0.496 & 0.356 & 0.468 & 0.455 \\
ToeSegmentation2 & 0.683 & 0.375 & 0.520 & 0.628 & \textbf{0.701} & 0.554 & 0.531 & 0.668 & 0.559 \\
Trace & 0.908 & 0.195 & 0.500 & 0.426 & \textbf{0.914} & 0.642 & 0.173 & 0.451 & 0.505 \\
TwoLeadECG & \textbf{0.743} & 0.396 & 0.560 & 0.519 & 0.734 & 0.592 & 0.400 & 0.552 & 0.572 \\
TwoPatterns & 0.599 & 0.350 & 0.492 & 0.466 & 0.646 & 0.526 & \textbf{0.651} & 0.490 & 0.491 \\
UWaveGestureLibraryAll & 0.643 & 0.380 & 0.520 & 0.510 & 0.617 & \textbf{0.682} & 0.371 & 0.444 & 0.494 \\
UWaveGestureLibraryX & 0.502 & 0.316 & 0.416 & 0.403 & 0.514 & 0.434 & 0.344 & 0.437 & \textbf{0.577} \\
UWaveGestureLibraryY & 0.729 & 0.384 & 0.549 & 0.539 & 0.741 & \textbf{0.752} & 0.411 & 0.553 & 0.733 \\
UWaveGestureLibraryZ & 0.643 & 0.366 & 0.493 & 0.491 & \textbf{0.659} & 0.525 & 0.374 & 0.460 & 0.494 \\
Wafer & 0.533 & 0.318 & 0.443 & 0.453 & \textbf{0.552} & 0.445 & 0.369 & 0.440 & 0.482 \\
Wine & 0.685 & 0.524 & 0.522 & 0.532 & 0.707 & 0.541 & 0.384 & 0.530 & \textbf{0.882} \\
WordSynonyms & \textbf{0.699} & 0.372 & 0.694 & 0.495 & 0.676 & 0.590 & 0.420 & 0.532 & 0.564 \\
Worms & \textbf{0.672} & 0.370 & 0.522 & 0.532 & 0.669 & 0.587 & 0.378 & 0.520 & 0.507 \\
WormsTwoClass & \textbf{0.623} & 0.335 & 0.615 & 0.471 & 0.578 & 0.473 & 0.521 & 0.438 & 0.469 \\
Yoga & \textbf{0.517} & 0.316 & 0.389 & 0.407 & 0.481 & 0.444 & 0.311 & 0.389 & 0.377 \\
ACSF1 & 0.514 & 0.314 & 0.374 & 0.404 & 0.534 & \textbf{0.571} & 0.315 & 0.544 & 0.396 \\
AllGestureWiimoteX & 0.676 & 0.376 & 0.501 & 0.603 & 0.589 & 0.515 & 0.385 & \textbf{0.765} & 0.496 \\
AllGestureWiimoteY & \textbf{0.518} & 0.310 & 0.420 & 0.392 & 0.503 & 0.435 & 0.314 & 0.398 & 0.434 \\
AllGestureWiimoteZ & \textbf{0.732} & 0.391 & 0.566 & 0.538 & 0.696 & 0.595 & 0.568 & 0.690 & 0.713 \\
BME & \textbf{0.567} & 0.333 & 0.470 & 0.446 & 0.559 & 0.452 & 0.500 & 0.433 & 0.460 \\
EthanolLevel & 0.595 & 0.356 & 0.493 & 0.443 & 0.589 & 0.507 & 0.513 & \textbf{0.654} & 0.464 \\
FreezerRegularTrain & \textbf{0.516} & 0.320 & 0.403 & 0.409 & 0.506 & 0.419 & 0.313 & 0.398 & 0.445 \\
FreezerSmallTrain & 0.585 & 0.356 & \textbf{0.633} & 0.478 & 0.582 & 0.539 & 0.361 & 0.463 & 0.489 \\
GunPointAgeSpan & 0.660 & 0.365 & \textbf{0.666} & 0.514 & 0.662 & 0.545 & 0.395 & 0.534 & 0.540 \\
GunPointMaleVersusFemale & 0.567 & 0.338 & 0.451 & 0.434 & 0.569 & 0.500 & 0.507 & 0.449 & \textbf{0.612} \\
GunPointOldVersusYoung & \textbf{0.661} & 0.365 & 0.510 & 0.511 & 0.643 & 0.551 & 0.374 & 0.505 & 0.547 \\
InsectEPGRegularTrain & 0.553 & 0.483 & 0.459 & 0.423 & 0.583 & \textbf{0.638} & 0.501 & 0.467 & 0.476 \\
InsectEPGSmallTrain & \textbf{0.602} & 0.346 & 0.462 & 0.454 & 0.598 & 0.491 & 0.347 & 0.455 & 0.491 \\
PickupGestureWiimoteZ & \textbf{0.582} & 0.358 & 0.458 & 0.446 & 0.567 & 0.498 & 0.513 & 0.457 & 0.480 \\
PigAirwayPressure & \textbf{0.702} & 0.396 & 0.539 & 0.509 & 0.694 & 0.572 & 0.374 & 0.691 & 0.578 \\
PigArtPressure & \textbf{0.752} & 0.382 & 0.582 & 0.557 & 0.674 & 0.594 & 0.415 & 0.556 & 0.604 \\
PigCVP & 0.518 & 0.332 & 0.453 & 0.457 & 0.515 & 0.492 & 0.327 & \textbf{0.591} & 0.450 \\
PLAID & 0.743 & 0.430 & 0.588 & 0.547 & \textbf{0.786} & 0.595 & 0.410 & 0.574 & 0.596 \\
PowerCons & 0.596 & 0.365 & 0.487 & 0.467 & \textbf{0.625} & 0.493 & 0.354 & 0.468 & 0.479 \\
ShakeGestureWiimoteZ & 0.494 & 0.446 & 0.395 & 0.372 & \textbf{0.507} & 0.421 & 0.328 & 0.426 & 0.414 \\
SmoothSubspace & 0.750 & 0.406 & \textbf{0.751} & 0.526 & 0.749 & 0.631 & 0.424 & 0.578 & 0.560 \\
UMD & 0.673 & 0.403 & 0.549 & 0.728 & 0.718 & 0.600 & 0.560 & 0.543 & \textbf{0.732} \\
Fungi & \textbf{0.658} & 0.387 & 0.505 & 0.539 & 0.613 & 0.553 & 0.542 & 0.537 & 0.523 \\
GesturePebbleZ1 & 0.535 & 0.353 & 0.414 & 0.426 & \textbf{0.553} & 0.463 & 0.327 & 0.442 & 0.444 \\
GesturePebbleZ2 & 0.571 & 0.344 & 0.442 & 0.439 & 0.565 & \textbf{0.637} & 0.354 & 0.438 & 0.477 \\
HouseTwenty & 0.660 & 0.360 & 0.525 & 0.469 & \textbf{0.675} & 0.587 & 0.536 & 0.660 & 0.527 \\
DodgerLoopDay & 0.648 & 0.394 & 0.539 & 0.655 & 0.646 & 0.607 & 0.376 & 0.684 & \textbf{0.690} \\
DodgerLoopWeekend & \textbf{0.621} & 0.364 & 0.486 & 0.427 & 0.581 & 0.510 & 0.364 & 0.474 & 0.501 \\
DodgerLoopGame & 0.540 & 0.338 & 0.443 & 0.444 & \textbf{0.562} & 0.484 & 0.332 & 0.424 & 0.450 \\
SemgHandGenderCh2 & \textbf{0.474} & 0.317 & 0.385 & 0.388 & 0.454 & 0.416 & 0.315 & 0.391 & 0.397 \\
SemgHandMovementCh2 & \textbf{0.748} & 0.374 & 0.554 & 0.656 & 0.707 & 0.584 & 0.399 & 0.521 & 0.551 \\
SemgHandSubjectCh2 & 0.605 & 0.369 & 0.459 & 0.621 & 0.628 & 0.500 & 0.361 & 0.465 & \textbf{0.659} \\
MixedShapesRegularTrain & 0.656 & 0.372 & 0.466 & 0.487 & 0.602 & 0.486 & 0.370 & 0.499 & \textbf{0.670} \\
MixedShapesSmallTrain & \textbf{0.558} & 0.328 & 0.442 & 0.413 & 0.524 & 0.473 & 0.323 & 0.431 & 0.447 \\
EOGHorizontalSignal & 0.526 & 0.343 & 0.478 & 0.416 & \textbf{0.549} & 0.466 & 0.346 & 0.456 & 0.446 \\
EOGVerticalSignal & 0.723 & 0.418 & 0.565 & 0.696 & \textbf{0.735} & 0.577 & 0.567 & 0.531 & 0.595 \\
GestureMidAirD1 & 0.763 & 0.412 & 0.565 & 0.536 & \textbf{0.772} & 0.593 & 0.415 & 0.725 & 0.575 \\
GestureMidAirD2 & 0.610 & 0.335 & 0.509 & 0.478 & 0.588 & 0.505 & 0.352 & 0.478 & \textbf{0.653} \\
GestureMidAirD3 & \textbf{0.708} & 0.548 & 0.495 & 0.583 & 0.693 & 0.586 & 0.396 & 0.692 & 0.524 \\
Rock & 0.722 & 0.407 & 0.558 & 0.520 & \textbf{0.795} & 0.581 & 0.434 & 0.572 & 0.565 \\
Crop & \textbf{0.682} & 0.558 & 0.537 & 0.519 & 0.663 & 0.576 & 0.404 & 0.540 & 0.554 \\
Chinatown & 0.610 & 0.351 & 0.511 & 0.488 & \textbf{0.652} & 0.543 & 0.363 & 0.488 & 0.491 \\
MelbournePedestrian & \textbf{0.728} & 0.414 & 0.607 & 0.542 & 0.705 & 0.556 & 0.419 & 0.590 & 0.590 \\
ArticularyWordRecognition & \textbf{0.745} & 0.386 & 0.688 & 0.651 & 0.717 & 0.586 & 0.397 & 0.524 & 0.574 \\
AtrialFibrillation & 0.567 & 0.340 & \textbf{0.609} & 0.589 & 0.595 & 0.468 & 0.334 & 0.448 & 0.600 \\
BasicMotions & 0.475 & 0.306 & 0.411 & 0.384 & 0.464 & \textbf{0.572} & 0.307 & 0.395 & 0.406 \\
CharacterTrajectories & \textbf{0.709} & 0.382 & 0.692 & 0.533 & 0.689 & 0.580 & 0.394 & 0.518 & 0.547 \\
Cricket & 0.438 & 0.301 & 0.385 & \textbf{0.560} & 0.489 & 0.384 & 0.316 & 0.386 & 0.397 \\
DuckDuckGeese & 0.506 & 0.332 & 0.428 & 0.599 & 0.554 & \textbf{0.790} & 0.339 & 0.440 & 0.453 \\
EigenWorms & 0.635 & 0.372 & 0.522 & 0.544 & \textbf{0.697} & 0.563 & 0.378 & 0.541 & 0.518 \\
Epilepsy & 0.635 & 0.345 & 0.628 & 0.476 & \textbf{0.642} & 0.533 & 0.354 & 0.489 & 0.527 \\
EthanolConcentration & 0.727 & 0.393 & 0.568 & 0.558 & \textbf{0.747} & 0.626 & 0.398 & 0.565 & 0.598 \\
ERing & 0.669 & 0.549 & 0.473 & 0.456 & 0.612 & \textbf{0.700} & 0.391 & 0.480 & 0.520 \\
FaceDetection & \textbf{0.649} & 0.405 & 0.509 & 0.529 & 0.610 & 0.573 & 0.410 & 0.530 & 0.561 \\
FingerMovements & 0.537 & 0.315 & 0.578 & \textbf{0.585} & 0.537 & 0.483 & 0.337 & 0.445 & 0.423 \\
HandMovementDirection & 0.682 & 0.382 & 0.487 & 0.616 & 0.679 & \textbf{0.701} & 0.386 & 0.520 & 0.584 \\
Handwriting & 0.545 & 0.324 & \textbf{0.583} & 0.553 & 0.484 & 0.493 & 0.327 & 0.434 & 0.425 \\
Heartbeat & 0.605 & 0.339 & 0.462 & 0.455 & \textbf{0.618} & 0.518 & 0.334 & 0.471 & 0.488 \\
InsectWingbeat & 0.517 & 0.455 & 0.395 & 0.383 & 0.474 & 0.410 & 0.327 & \textbf{0.539} & 0.413 \\
JapaneseVowels & 0.649 & 0.387 & 0.549 & 0.520 & \textbf{0.665} & 0.602 & 0.409 & 0.524 & 0.589 \\
Libras & \textbf{0.613} & 0.348 & 0.475 & 0.512 & 0.569 & 0.531 & 0.356 & 0.461 & 0.490 \\
LSST & 0.526 & 0.332 & 0.437 & \textbf{0.581} & 0.568 & 0.482 & 0.496 & 0.431 & 0.446 \\
MotorImagery & 0.558 & 0.333 & 0.578 & 0.421 & 0.571 & \textbf{0.616} & 0.338 & 0.442 & 0.449 \\
NATOPS & \textbf{0.724} & 0.528 & 0.521 & 0.454 & 0.715 & 0.587 & 0.379 & 0.517 & 0.546 \\
PenDigits & 0.626 & 0.332 & \textbf{0.644} & 0.446 & 0.577 & 0.642 & 0.373 & 0.480 & 0.473 \\
PEMS-SF & 0.521 & 0.481 & 0.431 & \textbf{0.712} & 0.535 & 0.450 & 0.341 & 0.560 & 0.463 \\
Phoneme  & \textbf{0.741} & 0.540 & 0.555 & 0.528 & 0.727 & 0.572 & 0.578 & 0.528 & 0.605 \\
RacketSports & 0.621 & 0.376 & 0.503 & 0.635 & 0.659 & 0.533 & 0.380 & \textbf{0.793} & 0.552 \\
SelfRegulationSCP1 & 0.677 & 0.380 & 0.557 & 0.660 & \textbf{0.705} & 0.535 & 0.385 & 0.483 & 0.549 \\
SelfRegulationSCP2 & \textbf{0.718} & 0.434 & 0.566 & 0.585 & 0.711 & 0.598 & 0.553 & 0.527 & 0.602 \\
SpokenArabicDigits & 0.536 & 0.486 & 0.449 & 0.431 & \textbf{0.572} & 0.463 & 0.338 & 0.476 & 0.436 \\
StandWalkJump & \textbf{0.769} & 0.410 & 0.564 & 0.563 & 0.740 & 0.602 & 0.414 & 0.565 & 0.617 \\
UWaveGestureLibrary & 0.646 & 0.365 & 0.504 & 0.483 & \textbf{0.669} & 0.542 & 0.375 & 0.531 & 0.536 \\
\midrule
Average V-measure & \textbf{0.618} & 0.381 & 0.502 & 0.494 & 0.611 & 0.536 & 0.386 & 0.499 & 0.519 \\
Average arithmetic ranking & \textbf{1.886} & 8.111 & 5.282 & 5.902 & 2.142 & 3.703 & 7.848 & 5.601 & 4.525 \\
Average geometric ranking & \textbf{1.699} & 7.785 & 4.878 & 5.419 & 1.953 & 3.364 & 7.588 & 5.110 & 4.079 \\
Winning times & \textbf{59} & 1 & 7 & 8 & 40 & 17 & 1 & 10 & 15 \\
\bottomrule
    \end{longtable}
\renewcommand{\arraystretch}{1}
\addtolength{\tabcolsep}{-3pt} 
\normalsize

\section{Conclusions}\label{sec:conclusions_5}

The demand for clustering algorithms that can handle large datasets and high-dimensional data has been increasing in recent years, particularly in the context of time series analysis. 
In this sense, traditional clustering methods may not be effective for time series data due to the specificity of the time dimension, which can result in high differences in the Euclidean space even for similar signals. In this section DIFW-IC has been proposed, a novel incremental clustering algorithm for time series that outperforms existing methods in terms of clustering quality and scalability. The proposed method is based on a combination of elastic alignment and incremental clustering techniques, and it is robust to outliers and concept drift. Contributions from \cref{chapter:2} where incorporated into the clustering model: the decision of assigning new points to the existing clusters is based on a diffeomorphic elastic distance, and aligned time series are obtained using the proposed differentiable sampler. We evaluated the performance of DIFW-IC on several benchmark datasets and compared it with state-of-the-art clustering methods for time series data. Our results demonstrate that overall DIFW-IC outperforms existing methods in terms of clustering quality and scalability. 

\graphicspath{{content/chapter6/}}

\chapter{Normalizing Flows based on Diffeomorphic Coupling Functions}\label{chapter:6}
\begingroup
\hypertarget{chapter6}{}
\hypersetup{linkcolor=black}
\minitoc
\endgroup

\section{Introduction}\label{sec:introduction_6}

Generative models have become increasingly popular in recent years due to their ability to capture complex data distributions and generate high-quality, realistic samples.
Discriminative ML models only focus on making predictions based on the input data and do not have the ability to generate new samples.
One can think of any kind of observed data $\mathcal{X} = \{\mathbf{x}_1, \mathbf{x}_2 \cdots \mathbf{x}_N\}$ as a finite set of $N$ samples from an underlying distribution $p_{\mathcal{X}}$. At its very core, the goal of any generative model is to approximate such data distribution given access to the dataset $\mathcal{X}$. The hope is that if we are able to \textit{learn} a good generative model, we can use the learned model for downstream \textit{inference}.
Among the most prevalent generative models one can find Generative adversarial networks (GANs) \cite{goodfellow2020generative}, Variational Autoencoders (VAEs) \cite{kingma2013auto}, Energy-based models \cite{lecun2005loss} and Normalizing flows (NFs) \cite{rezende2015variational}, which are studied in this chapter. 

\textbf{Normalizing flows} transform a simple probability distribution into a more complex one through a series of invertible transformations $f$.
The key defining property of flow-based generative models is that transformations $f$ must be \textbf{invertible} and both $f$ and $f^{-1}$ must be \textbf{differentiable}.
Normalizing flows based on coupling functions require a bijective one-dimensional function $h(z)$ and the derivative of the function with respect the input variable $z$ (also called the spatial dimension), i.e. $\frac{\partial h}{\partial z}$.
Related flows based on coupling layers such as NICE \cite{dinh2014nice} and RealNVP \cite{dinh2016density}, have an analytic one-pass inverse, but are often less flexible than their autoregressive counterparts. 

Based on these limitations, this work proposes to implement the \textbf{coupling function $h$ using the integration of continuous piecewise-affine (CPA) velocity functions} as a building block. A fully-differentiable module based on the integration of CPA velocity functions is presented, which yield diffeomorphic curves. Computing the inverse of such curves is equivalent to computing the forward curve backward in time or with opposite sign of the parameters. These diffeomorphic curves are used as the coupling function $h(z)$. 

The module acts as a \textbf{drop-in replacement for the affine or additive transformations} commonly found in coupling and autoregressive transforms. 
When combined with alternating invertible linear transformations, the resulting class of normalizing flows is referred to as \textbf{closed-form diffeomorphic spline flows} (DIFW-NF), which may feature coupling layers, DIFW-NF (C), or autoregressive layers, DIFW-NF (AR).
Overall, DIFW-NF resembles a traditional feed-forward neural network architecture, alternating between linear transformations and element-wise non-linearities, while retaining an exact, analytic inverse.
Experiments demonstrate that this module significantly enhances the flexibility of both classes of flows, obtaining competitive results in a variety of high-dimensional datasets.

Given that all previous chapters were devoted to time series and this one stands on its own, the author considers that an appropriate introduction to the matter is required. As a consequence, in \cref{sec:normalizing_flows} normalizing flows are formally introduced and in \cref{sec:normalizing_flows_models} an extensive review of the state-of-the-art is presented. The proposed model is detailed in \cref{sec:method_6} and the experiments applied to 1D, 2D and ND data are included in \cref{sec:results_6}.

\section{Normalizing Flows}\label{sec:normalizing_flows}

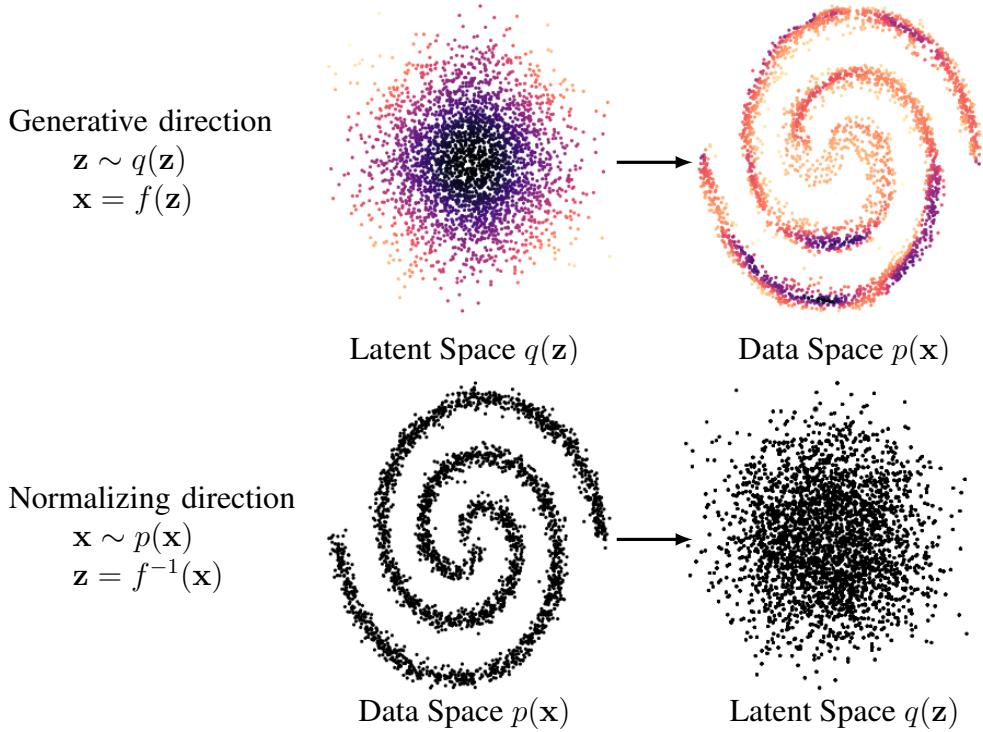
\begin{figure}[!htb]
  \begin{center}
    \begin{tikzpicture}
      \begin{scope}[shift={(0,0)}]
          \node[text width=4cm] at (0,0) {Generative direction \\ $\quad\quad \mathbf{z} \sim q(\mathbf{z})$ \\ $\quad\quad \mathbf{x} = f(\mathbf{z})$};
          \node[] at (6.5,0) {\includegraphics[width=0.6\linewidth,trim=0 0 0 25cm, clip]{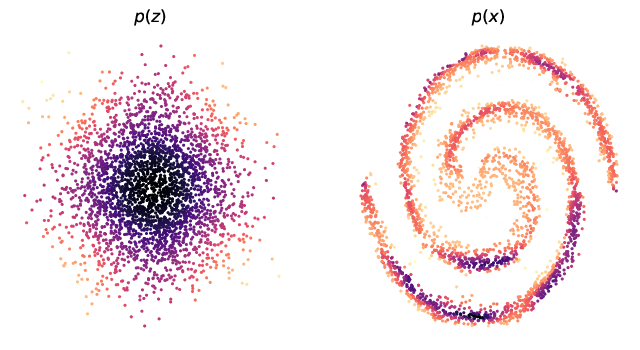}};
          \path[draw, -latex, very thick] (6,0) -- (7,0);
          \node[] at (4,-2.5) {Latent Space $q(\mathbf{z})$};
          \node[] at (9,-2.5) {Data Space $p(\mathbf{x})$};
      \end{scope}
      \begin{scope}[shift={(0,-5)}]
          \node[text width=4cm] at (0,0) {Normalizing direction \\ $\quad\quad \mathbf{x} \sim p(\mathbf{x})$ \\ $\quad\quad \mathbf{z} = f^{-1}(\mathbf{x})$};
          \node[] at (6.5,0) {\includegraphics[width=0.6\linewidth,trim=0 0 0 25cm, clip]{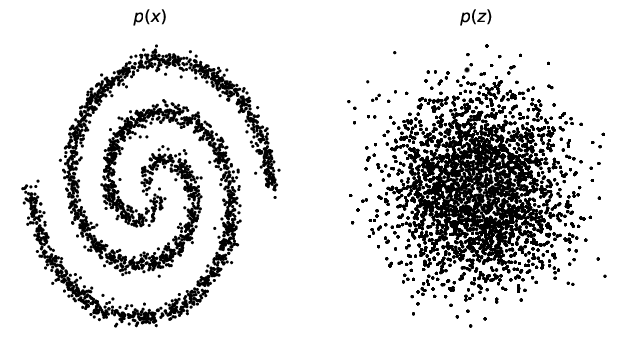}};
          \path[draw, -latex, very thick] (6,0) -- (7,0);
          \node[] at (4,-2.3) {Data Space $p(\mathbf{x})$};
          \node[] at (9,-2.3) {Latent Space $q(\mathbf{z})$};
      \end{scope}
    \end{tikzpicture}
  \caption{Normalizing flow on a toy two-dimensional dataset. 
  The function $f(\mathbf{z})$ maps samples $\mathbf{z}$ from the latent distribution in the upper left into approximate samples $\mathbf{x}$ from the data distribution in the upper right. This corresponds to exact generation of samples from the model.
  The inverse function $f^{-1}(\mathbf{x})$ maps samples $\mathbf{x}$ from the data distribution in the lower left into approximate samples $\mathbf{z}$ from the latent distribution in the lower right. This corresponds to the exact inference of the latent state given the data. 
  }
  \label{fig:normalizing_flow_example_0}
  \end{center}
\end{figure}

A normalizing flow is a transformation that converts a simple probability distribution $q(\mathbf{z})$  into a more complex one $p(\mathbf{x})$ by applying a sequence of invertible and differentiable mappings \cite{Papamakarios2021} (see \cref{fig:normalizing_flow_example_0}). 
According to the change of variables theorem the original variable is substituted in a repetitive manner for a new one as it "flows" through a series of transformations to ultimately obtain the probability distribution of the final target variable.
Thus, normalizing flows offer a general method for constructing flexible probability distributions for continuous random variables.
By repeatedly applying simple transformations to an unimodal initial density, complex models can be created.

There has been a growing interest in normalizing flows in the deep learning community, driven by successful applications and structural advantages they have over alternatives: \textbf{model flexibility} and \textbf{generation speed}. Normalizing flows can represent a richer family of distributions without requiring approximations. Flows have been explored both to increase the flexibility of the variational posterior in the context of VAEs, and directly as a generative model. 

Let $\mathbf{x}$ be a $d$-dimensional real vector and suppose one would like to define a joint distribution over $\mathbf{x}$. The main idea of flow-based modeling is to express $\mathbf{x}$ as a transformation $f$ of a real vector $\mathbf{z}$ from $q(\mathbf{z})$, that is, $\mathbf{x} = f(\mathbf{z})$ where $\mathbf{z} \sim q(\mathbf{z})$.
$q(\mathbf{z})$ is referred to as the base distribution of the flow-based model. The transformation $f$ and the base distribution $q(\mathbf{z})$ can have parameters on their own. 
Depending on the flow's intended use cases, there are practical constraints in addition to formal invertibility:
\begin{itemize}
\item To train a \textbf{density estimator}, one needs to evaluate the Jacobian determinant and the inverse function $f^{-1}$ quickly. Note that $f$ is not evaluated, so the flow is usually defined by specifying $f^{-1}$.
\item To \textbf{draw samples}, one would like $f$ to be available analytically, rather than having to invert $f^{-1}$ with iterative or approximate methods.
\end{itemize}
In the ideal case, one would like both $f$ and $f^{-1}$ to require only a single pass of a neural network to compute, so that both density estimation and sampling can be performed quickly.
The key defining property of flow-based models is that the transformation must be \textbf{invertible} and both $f$ and $f^{-1}$ must be \textbf{differentiable}. Such transformations are known as diffeomorphisms and require that $\mathbf{z}$ be $d$-dimensional as well. Under these conditions, the density of $\mathbf{x}$ is well-defined and can be obtained by a change of variables:
\begin{equation}
  p(\mathbf{x}) = q(\mathbf{z}) \left\vert \det \dfrac{d \mathbf{z}}{d \mathbf{x}} \right\vert = q(f^{-1}(\mathbf{x})) \left\vert \det \dfrac{d f^{-1}(\mathbf{x})}{d \mathbf{x}} \right\vert
\end{equation}

Regarding the \textbf{expressive power} of flow-based models, it has been shown that for any pair of well-behaved distributions $p(\mathbf{x})$ (the target) and $q(\mathbf{z})$ (the base), there exists a diffeomorphism that can turn $p(\mathbf{x})$ into $q(\mathbf{z})$.
In practice, a flow-based model is often constructed by implementing $f$ or $f^{-1}$ with a neural network and taking $q(\mathbf{z})$ to be a simple density such as a multivariate normal. 

\begin{figure}[!htb]
  \begin{center}
  \includegraphics[width=0.65\linewidth]{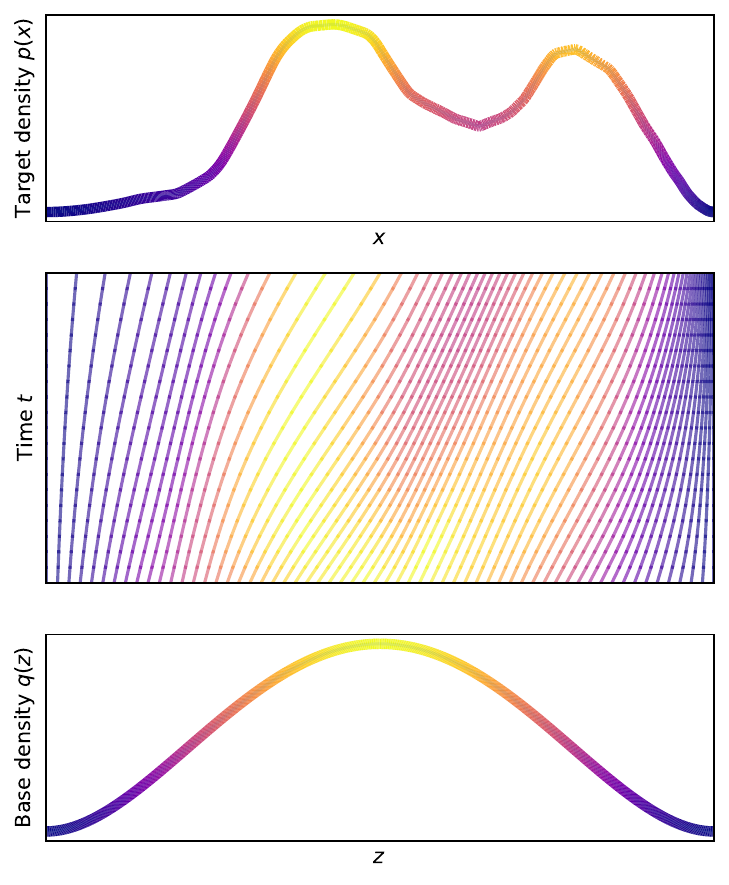}
  \caption{A continuous normalizing flow continuously deforms one distribution into another distribution. The flow lines show how particles from the base distribution Beta($\alpha=3,\beta=3$) are perturbed until they approximate the target multi-modal distribution.}
  \label{fig:normalizing_flow_example_1}
  \end{center}
\end{figure}

An important property of invertible and differentiable transformations is that they are composable. Given two such transformations $f_1$ and $f_2$, their composition $f_{2} \circ f_{1}$ is also invertible and differentiable. 
In practice, it is common to chain together multiple transformations $f_{1}, \cdots, f_{k}$ to obtain $f=f_{1} \circ \dots \circ f_{k}$, where each $f_{i}$ transforms $\mathbf{z}_{i-1}$ into $\mathbf{z}_{i}$, assuming $\mathbf{z}_{0}=\mathbf{z}$ and $\mathbf{z}_{k}=\mathbf{x}$. 
\cref{fig:normalizing_flow_example_1} shows a one-dimensional bijective transformation.
The name “normalizing flow” can be interpreted as the following:
Flow refers to the trajectory that a collection of samples from $q(\mathbf{z})$ follow as they are gradually transformed by the sequence of invertible transformations $f_{1}, \cdots, f_{k}$ which are composed with each other to create more complex invertible transformations.
Normalizing refers to the fact that the inverse flow through $f_{k}^{-1}, \cdots, f_{1}^{-1}$ takes a collection of samples from $p(\mathbf{x})$ and transform them into a collection of samples from a prescribed density $q(\mathbf{z})$, in a sense, normalizes them.

\subsection{Sampling}

In the sampling case, the bijective functions $f$ (with parameters $\theta$) and the initial distribution $q(z)$ (with parameters $\phi$) are given. 
Let $f$ be a normalizing flow, $f: \mathbf{z} \rightarrow \mathbf{x}$ where $\mathbf{x} = (x_1, x_2 \dots x_{d})$ and $\mathbf{z} = (z_1, z_2 \dots z_{d})$. More generally, one can decompose $f$ into $f_1, f_2 \dots f_{d}$ such that $x_{i} = f_{i}(z_1, z_2, \dots, z_{d})$. 
Sampling points from the output distribution requires calculating the forward pass, and hence an efficient calculation of the transformation. Thus sampling performance is determined by the cost of the generative direction.

Let $\mathbf{z}_0$ be a continuous random variable belonging to a simple probability distribution $p_{0}(\mathbf{z}_{0})$ such as a multivariate Gaussian with parameters $(\mu, \sigma)$: $\mathbf{z}_{0} \sim p_{0}(\mathbf{z}_{0}) = N(\mathbf{z}_{0}; \mu, \sigma)$.
Normalizing flows transform a simple distribution into the desired output probability distribution with random variable $\mathbf{x}$, with a sequence of invertible transformations $f_{i}$:
\begin{equation}
  \mathbf{x} = \mathbf{z}_k = f_k \circ f_{k-1} \circ \dots \circ f_1 (\mathbf{z}_0)
\end{equation}


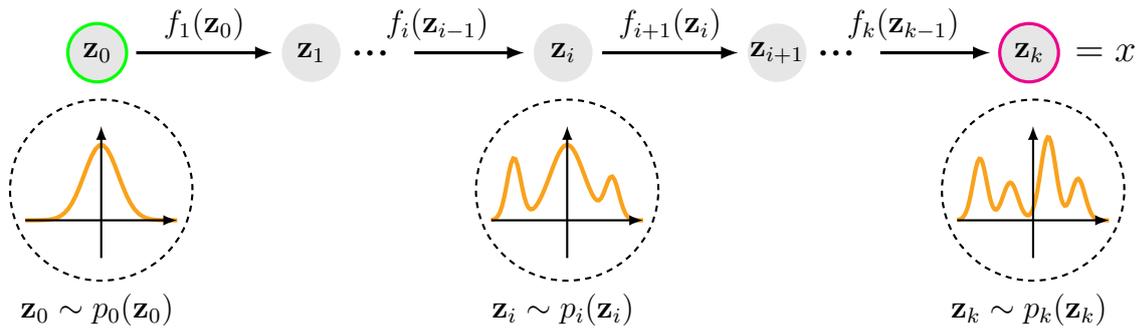
\begin{figure}[!htb]
  \begin{center}
\begin{tikzpicture}[
  node distance=2, very thick,
  flow/.style={shorten >=3, shorten <=3, -latex},
  znode/.style={circle, fill=black!10, minimum size=22, inner sep=0},
  ]
  \newcommand{\distro}[4][40]{
    \begin{tikzpicture}[thick]
      \draw[dashed, dash pattern={on 2.3 off 2}] (0, .4) circle (12mm);
      \draw[YellowOrange, ultra thick] plot[variable=\t, domain=-1:1, samples=#1] ({\t}, {#2 * exp(-10*(\t)^2) + #3 * exp(-60*(\t-0.6)^2 - \t) + #3 * exp(-60*(\t+0.7)^2 - 0.2) + #4 * 0.5 * exp(-50*(\t+0.3)^2) + #4 * exp(-50*(\t-0.2)^2 + 0.1)});
      \draw[solid, -latex] (-1, 0)--(1, 0);
      \draw[solid, -latex] (0, -0.5)--(0, 1.25);
    \end{tikzpicture}
  }

\node[znode, draw=green] (z0) {$\mathbf{z}_0$};
\node[znode, right=of z0] (z1) {$\mathbf{z}_1$};
\draw[flow] (z0) -- node[above, midway] {$f_1(\mathbf{z}_0)$} (z1);

\node[znode, right=2.5 of z1] (zi) {$\mathbf{z}_i$};
\node[znode, right=of zi] (zip1) {$\mathbf{z}_{i+1}$};
\draw[flow] (zi) -- node[above, midway] {$f_{i+1}(\mathbf{z}_i)$} (zip1);
\draw[flow, shorten <=5ex] (z1) -- node[pos=0.16, inner sep=1] {\textbf\dots} node[above, midway] {$f_i(\mathbf{z}_{i-1})$} (zi);

\node[znode, draw=Magenta, right=2.5 of zip1] (zk) {$\mathbf{z}_k$};
\draw[flow, shorten <=5ex] (zip1) -- node[pos=0.16, inner sep=1] {\textbf\dots} node[above, midway] {$f_k(\mathbf{z}_{k-1})$} (zk);
\node[right=0 of zk, scale=1.2] {$= x$};
\node[outer sep=0, inner sep=0, below=0.2 of z0, label={below:$\mathbf{z}_0 \sim p_0(\mathbf{z}_0)$}] (f0) {\distro{1}{0}{0}};
\node[outer sep=0, inner sep=0, below=0.2 of zi, label={below:$\mathbf{z}_i \sim p_i(\mathbf{z}_i)$}] (fi) {\distro[70]{1}{1}{0}};
\node[outer sep=0, inner sep=0, below=0.2 of zk, label={below:$\mathbf{z}_k \sim p_k(\mathbf{z}_k)$}] (fk) {\distro[90]{0}{1}{1}};

\end{tikzpicture}
\caption{A chain of bijections $f = f_k \circ \dots \circ f_1$ constituting a normalizing flow which step by step transforms a simple distribution $p_0(\mathbf{z}_0)$ into a complex one $p_k(\mathbf{z}_k)$. The bijections are trained to fit $p_k(\mathbf{z}_k)$ to some target distribution $p_x(\mathbf{x})$.}
\label{fig:normalizing_chain}
\end{center}
\end{figure}

\subsection{Density Estimation}

When applied as density estimators, NFs provide a general way of constructing flexible probability distributions over continuous random variables starting from a simple probability distribution. By constraining the transformations to be invertible, flow-based models provide a tractable method to calculate the exact likelihood for a wide variety of generative modeling problems.

\begin{equation}
\begin{aligned}
  \mathbf{z}_{i-1} &\sim p_{i-1}(\mathbf{z}_{i-1}) \\ 
  \mathbf{z}_i &= f_i(\mathbf{z}_{i-1})\text{, thus }\mathbf{z}_{i-1} = f_i^{-1}(\mathbf{z}_i) \\ 
  p_i(\mathbf{z}_i) &= p_{i-1}(f_i^{-1}(\mathbf{z}_i)) \left\vert \det\dfrac{d f_i^{-1}(\mathbf{z}_i)}{d \mathbf{z}_i} \right\vert 
\end{aligned}
\end{equation}

Then, in order to infer with the base distribution, the equation is converted to a function of $\mathbf{z}_i$:
\begin{equation}
\begin{aligned}
  p_i(\mathbf{z}_i) &= p_{i-1}(f_i^{-1}(\mathbf{z}_i)) \left\vert \det\dfrac{d f_i^{-1}}{d \mathbf{z}_i} \right\vert \\
  &\underset{\footnotemark}{=} p_{i-1}(\mathbf{z}_{i-1}) \left\vert \det \color{blue}{\Big(\dfrac{d f_i}{d\mathbf{z}_{i-1}}\Big)^{-1}} \right\vert \\ 
  &\underset{\footnotemark}{=} p_{i-1}(\mathbf{z}_{i-1}) \color{blue}{\left\vert \det \dfrac{d f_i}{d\mathbf{z}_{i-1}} \right\vert^{-1}} \\
\end{aligned}
\end{equation}
\addtocounter{footnote}{-2}
\stepcounter{footnote}\footnotetext{Inverse function theorem: If $y=f(x)$ and $x=f^{-1}(y)$ then $\dfrac{df^{-1}(y)}{dy} = \dfrac{dx}{dy} = \big(\dfrac{dy}{dx}\big)^{-1} = \big(\dfrac{df(x)}{dx}\big)^{-1} $}
\stepcounter{footnote}\footnotetext{Property of Jacobians of invertible functions: The determinant of the inverse of an invertible matrix is the inverse of the determinant: $\det(M^{-1}) = (\det(M))^{-1}$, because $\det(M)\det(M^{-1}) = \det(M \cdot M^{-1}) = \det(I) = 1$}

NFs usually work with log-probabilities, as they allow for stable numerical computations:
\begin{equation}
\log p_i(\mathbf{z}_i) = \log p_{i-1}(\mathbf{z}_{i-1}) - \log \left\vert \det \dfrac{d f_i}{d\mathbf{z}_{i-1}} \right\vert 
\end{equation}
Given such a chain of probability density functions, one knows the relationship between each pair of consecutive variables. The equation of the output $\mathbf{x}$ can be expanded step by step until tracing back to the initial distribution $\mathbf{z}_0$.
\begin{equation}
\begin{aligned} 
  \mathbf{x} = \mathbf{z}_k &= f_k \circ f_{k-1} \circ \dots \circ f_1 (\mathbf{z}_0) \\ 
  \log p(\mathbf{x}) = \log q_k(\mathbf{z}_k) &= \log q_{k-1}(\mathbf{z}_{k-1}) - \log\left\vert\det\dfrac{d f_k}{d \mathbf{z}_{k-1}}\right\vert \\
  &= \log q_{k-2}(\mathbf{z}_{k-2}) - \log\left\vert\det\dfrac{d f_{k-1}}{d\mathbf{z}_{k-2}}\right\vert - \log\left\vert\det\dfrac{d f_k}{d\mathbf{z}_{k-1}}\right\vert \\ 
  &= \dots \\ 
  &= \log q_0(\mathbf{z}_0) - \sum_{i=1}^k \log\left\vert\det\dfrac{d f_i}{d\mathbf{z}_{i-1}}\right\vert 
\end{aligned}
\end{equation}

For density estimation and training, the calculation of both the inverse and determinant of the Jacobian are needed.
The path traversed by the random variables $\mathbf{z}_i = f_i(\mathbf{z}_{i-1})$ is the flow and the full chain formed by the successive distributions $q_i$ is called a normalizing flow. Required by the change of variables equation, a transformation function $f_i$ should satisfy two properties: (a) It is easily invertible. (b) Its Jacobian determinant is easy to compute.
However, calculating the inverse and the determinant of the Jacobian of a sequence of high dimensional transformations can be very time-consuming (for dimensionality $d$ matrix, both are of complexity $\mathcal{O}(d^3)$). There are various tricks that are used to reduce the complexity of these two operations, one of the popular ones being the use of \textbf{triangular maps}.

\begin{equation}\label{eq:jacobian}
\cfrac{df}{d\mathbf{z}} = \nabla_{\mathbf{z}} f = \begin{bmatrix}
  \cfrac{\partial f_1}{\partial z_1} & 0 & \cdots & 0 \\
  \cfrac{\partial f_2}{\partial z_1} & \cfrac{\partial f_2}{\partial z_2} & \cdots & 0 \\
  \vdots & \vdots & \ddots & \vdots \\
  \cfrac{\partial f_d}{\partial z_1} & \cfrac{\partial f_d}{\partial z_2} & \cdots &  \cfrac{\partial f_d}{\partial z_d} \\
\end{bmatrix}
\end{equation}

For $f$ to be a triangular map, each $f_{i}$ should be a function of $(z_1, z_2 \dots z_{i})$ i.e. the first $i$ elements and not all the $d$ elements. For triangular maps/matrices, both the inverse and the determinant of the Jacobian are easy to compute. The Jacobian for a triangular map is shown in \cref{eq:jacobian}. The determinant is the product of the diagonals and has a complexity of $\mathcal{O}(d)$ instead of $\mathcal{O}(d^3)$, whereas the complexity for the inverse is $\mathcal{O}(d^2)$ instead of $\mathcal{O}(d^3)$.

\subsection{Training}

With normalizing flows in our toolbox, the exact log-likelihood of input data $\log p(\mathbf{x})$ becomes tractable. 
As a result, the training criterion of flow-based generative model is simply the negative log-likelihood (NLL) over the training dataset $\mathcal{X}$ and one can recover the optimal parameters via maximizing likelihood estimation: 
$\max_{\theta \in \mathcal{M}} \mathbb{E}_{\mathbf{x} \sim p_{\mathcal{X}}} [\log p_{\theta}(\mathbf{x})]$. 
Here, $\log p_{\theta}(\mathbf{x})$ is referred to as the log-likelihood of the data point $\mathbf{x}$ with respect to the model distribution $p_{\theta}$. To approximate the expectation over the unknown $p_{\mathcal{X}}$, one makes an assumption: points in the dataset $\mathcal{X}$ are sampled i.i.d. from $p_{\mathcal{X}}$. This allows us to obtain an unbiased Monte Carlo estimate of the objective as
\begin{equation}
\max_{\theta \in \mathcal{M}} \frac{1}{\vert\mathcal{X}\vert}\sum_{\mathbf{x} \in \mathcal{X}} \log p_{\theta}(\mathbf{x}) = \mathcal{L}(\theta | \mathcal{X})
\end{equation}
The maximum likelihood estimation (MLE) objective has an intuitive interpretation: pick the model parameters $\theta \in \mathcal{M}$ that maximize the log-probability of the observed data points in $\mathcal{X}$.
In practice, the MLE objective is optimized using mini-batch gradient ascent. The algorithm operates in iterations. At every iteration $i$, a mini-batch $\beta_i$ of data points is sampled randomly from the dataset ($|\beta_i|<|\mathcal{X}|$) and gradients of the objective are evaluated for the mini-batch. These parameters at iteration $i+1$ are then given via the following update rule:
$\theta^{(i+1)} = \theta^{(i)} + \lambda_{i} \nabla_{\theta} \mathcal{L}(\theta^{(i)} | \mathcal{X})$, where $\theta^{(i+1)}$ and $\theta^{(i)}$ are the parameters at iterations $i+1$ and $i$ respectively, and $\lambda_{i}$ is the learning rate at iteration $i$. Typically, one only specifies the initial learning rate $\lambda_{1}$ and update the rate based on a schedule. Variants of stochastic gradient ascent, such as RMSprop and Adam, employ modified update rules that work slightly better in practice.

\section{Related Work}\label{sec:normalizing_flows_models}

\textbf{Autoregressive models} such as MAF \cite{papamakarios2017masked} achieve state-of-the-art density estimation performance on many challenging real-world datasets, but generally suffer from slow sampling time due to their autoregressive structure. These flows are $d$ times slower to invert than to evaluate, where $d$ is the dimensionality of $\mathbf{x}$. 
\textbf{Inverse autoregressive models} such as IAF \cite{kingma2016improved} can sample quickly and potentially have strong modeling capacity, but they cannot be trained efficiently by maximum likelihood.
Subsequent work which enhances the flexibility of autoregressive flows has resulted in models which do not have an analytic inverse, and require numerical optimization to invert. For instance, inverting the non-affine transformations used by NAF \cite{huang2018neural} and block-NAF \cite{de2020block} would require numerical optimization.
Transformations that are equally fast to invert and evaluate do exist. 
\textbf{Continuous flows} such as Neural ODEs \cite{chen2019neural} and FFJORD \cite{Grathwohl2019} are equally fast in both directions, but they require numerically integrating a differential equation in each direction, which can be slower than a single neural-network pass.

\textbf{Coupling flows} and autoregressive flows have a similar functional form and both have coupling functions as building blocks. A coupling function is a bijective differentiable function $\mathbf{h}(\cdot, \theta): \mathcal{R}^{d} \rightarrow \mathcal{R}^{d} $, parametrized by $\theta$. In coupling flows, these functions are typically constructed by applying a scalar coupling function $h(\cdot, \theta): \mathcal{R} \rightarrow \mathcal{R}$ element-wise. In autoregressive flows, $d = 1$ and hence they are also scalar valued. 
\textbf{Additive and affine} coupling functions are used for coupling flows in NICE \cite{dinh2014nice}, RealNVP \cite{dinh2016density}, Glow \cite{kingma2018glow} and for autoregressive architectures in IAF and MAF. Both the affine and additive transformations are easy to invert, but they lack flexibility.
\cite{ziegler2019latent} proposed an invertible \textbf{non-linear squared} transformation that adds an inverse-quadratic perturbation to an affine transformation in an autoregressive flow. 
The \textbf{Flow++ model} \cite{ho2019flow} uses the cumulative distribution function (CDF) of a mixture of logistic distributions as a monotonic transformation. In this case the computation of the inverse is done numerically with the bisection algorithm since a closed form is not available. The derivative of the transformation with respect to $z$ is expressed in terms of probability density function of logistic mixture.
Also related, \cite{Jaini2019} parametrize the monotonic transformation as a strictly increasing \textbf{sum of squares polynomial} (SOS).
For low-degree polynomials, an analytic inverse may be available, but the method would require an iterative solution in general. SOS is easier to train than NAF, because there are no restrictions on the parameters (like positivity of weights).

Regarding \textbf{linear splines}, 
\cite{muller2019neural} divided the domain into $k$ equal bins and modeled $h$ as the integral of a positive piecewise-constant function. 
\cite{muller2019neural} also used a \textbf{monotone quadratic spline} on the unit interval and modeled it as the integral of a positive piecewise-linear function. Such spline is invertible but finding its inverse requires solving a quadratic equation.
\cite{durkan2019cubic} proposed using \textbf{monotone cubic splines} defined only on the interval $[0,1]$. To ensure that the input is always between 0 and 1, a sigmoid transformation was placed before each coupling layer, and a logit transformation after each coupling layer.
A monotone cubic polynomial has only one real root and for inversion, one can find this either analytically or numerically. 
The flow can be trained by gradient descent by differentiating through the numerical root finding method. However, the procedure is numerically unstable if not treated carefully, as noted by \cite{durkan2019neural} when the sigmoid saturates for values far from zero, 
Also related, \cite{durkan2019neural} model a coupling function as a \textbf{monotone rational-quadratic spline} on an interval and as the identity in the rest of the domain.
The derivative is a quotient derivative and the inverse is obtained by solving a quadratic equation. The RQ-NSF (C) coupling architecture and RQ-NSF(AR) autoregressive architectures used these coupling functions.

In the hope of creating an ideal likelihood-based generative model that simultaneously has fast sampling, fast inference, and strong density estimation performance, this article proposes replacing the conditional affine transformation \cite{kingma2016improved} with a more rich family of transformations, and note the requirements for doing so. In particular, we propose a novel normalizing-flow model based on a coupling function $h$ using the integration of continuous piecewise-affine velocity functions as a building block. Approaches like SOS \cite{Jaini2019}, cubic splines \cite{durkan2019cubic} and Flow++  \cite{ho2019flow} present couplings that are similar in spirit to our approach. 

\section{Diffeomorphic Coupling Functions}\label{sec:method_6}

\begin{figure}[!htb]
  \begin{center}
    \includegraphics[width=\linewidth]{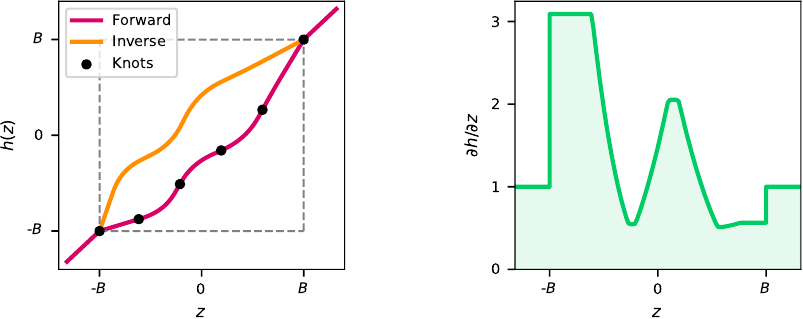}
    \caption{The proposed diffeomorphic transforms are drop-in replacements for additive or affine transformations in coupling or autoregressive layers, greatly enhancing their flexibility while retaining exact invertibility. 
    \textbf{Left}: Forward and inverse transformer of a random diffeomorphic monotonic function with tessellation size of 5 bins and linear tails, which is parametrized by a series of 6 points in the plane, and the 5 derivatives at the internal knots. \textbf{Right}: Transformer derivative with respect to $z$. }
    \label{fig:nf_simple3}
  \end{center}
  \vspace{-0.5cm}
\end{figure}

Normalizing flows based on coupling functions require a \textbf{bijective one-dimensional function} $h(z)$ and the derivative of the function with respect the input variable $z$ (also called the spatial dimension), i.e. $\frac{\partial h}{\partial z}$.
This work proposes a fully-differentiable module based on the \textbf{integration of continuous piecewise-affine (CPA) velocity functions}, which yield diffeomorphic curves (differentiable, invertible and with differentiable inverse). As a matter of fact, computing the inverse is exactly the same as computing the forward curve backward in time or by flipping the sign of the parameters (check the diffeomorphic properties in \cref{sec:results:diffeomorphic_properties}). These diffeomorphic curves will be used as the coupling function, so $h(z) = \phi^{\theta}(x,t)$. 
The diffeomorphic function itself maps an interval $[-B, B]$ to $[-B, B]$, as illustrated on \cref{fig:nf_simple3}. The transformation outside this range is defined as the identity, resulting in linear tails, so that the overall transformation can take unconstrained inputs.

The module acts as a \textbf{drop-in replacement for the affine or additive transformations} commonly found in coupling and autoregressive transforms. 
Unlike the additive and affine transformations, which have limited flexibility, the proposed differentiable monotonic function with sufficiently many intervals can approximate any differentiable monotonic function on the specified interval $[-B, B]$, yet has a closed-form, tractable Jacobian determinant, and can be inverted analytically. The proposed parameterization is fully-differentiable, which allows for training by gradient methods.

The proposed formulation can also easily be adapted for autoregressive transforms; each $\theta_{k}$ can be computed as a function of $\mathbf{z}_{1:i}$ using an autoregressive neural network, and then all elements of $\mathbf{z}$ can be transformed at once. 
When combined with alternating invertible linear transformations, the resulting class of normalizing flows is referred to as \textbf{closed-form diffeomorphic spline flows (DIFW-NF)}, which may feature coupling layers, DIFW-NF (C), or autoregressive layers, DIFW-NF (AR).
Experiments demonstrate that this module significantly \textbf{enhances the flexibility} of both classes of flows, and in some cases brings the performance of coupling transforms on par with the best-known autoregressive flows.
DIFW-NF only requires a single neural-network pass in either the forward or the inverse direction, but in practice is as flexible as state-of-the-art autoregressive flows.
Overall, DIFW-NF resembles a traditional feed-forward neural network architecture, alternating between linear transformations and element-wise non-linearities, while retaining an exact, analytic inverse.
Like Real NVP or Glow, DIFW-NF flows can represent either a transformation from data to noise, or from noise to data. In both cases, the transformation requires only a single pass of the neural network defining the flow.
In order to keep the same notation used in \cref{chapter:2}, for the rest of the section the spatial dimension of the transformation will be referred to as $x$ instead of $z$.

\begin{figure}[!htb]
  \begin{center}
  \includegraphics[width=0.9\linewidth]{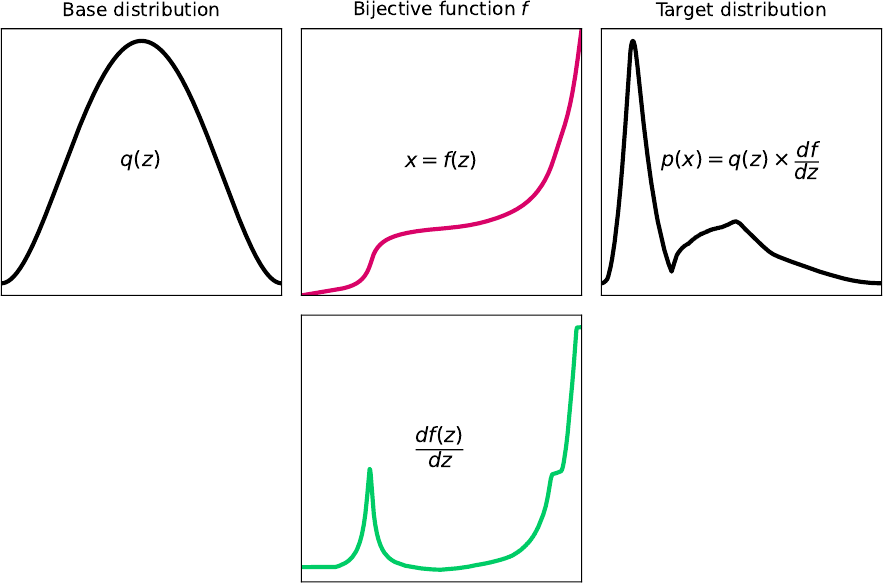}
  \caption{An example of a transformation obtained by a proposed normalizing flow DIFW-NF, which transforms a base Beta($\alpha=3,\beta=3$) distribution into a target multi-modal distribution with the help of the proposed diffeomorphic bijective function.}
  \label{fig:normalizing_flow_example_2}
  \end{center}
\end{figure}

\clearpage

\subsection{CPA-based Diffeomorphic Transformations}

Let's recall the CPA-based diffeomorphic transformations introduced in \cref{chapter:2}. A diffeomorphism can be obtained, via integration, from uniformly continuous stationary velocity fields $T^{\theta}(x) = \phi^{\theta}(x,1)$ 
where $\phi^{\theta}(x,t) = x + \int_0^t v^{\theta}(\phi^{\theta}(x,\tau)) d\tau$ for uniformly continuous $v: \Omega \rightarrow \mathbb{R}$ and integration time $t$. 
Here $x$ refers to the spatial dimension and $t$ the integration time.
The solution for the integral equation is a composition of a finite number of solutions $\psi$, given by: $\phi^\theta(x,t) = \big(\psi_{\theta,c_m}^{t_m} \circ \psi_{\theta,c_{m-1}}^{t_{m-1}} \circ \cdots \circ \psi_{\theta,c_2}^{t_2} \circ \psi_{\theta,c_1}^{t_1} \big)(x)$, 
where $m$ is the number of cells visited and $\psi_{\theta,c}^{t}$ is the solution of a basic ODE $\frac{d\psi}{dt}=v^\theta(\psi)$
with an $\mathbb{R} \rightarrow \mathbb{R}$ affine velocity field: 
$v^\theta(\psi) = a^\theta \psi + b^\theta$ and an initial condition $\psi(x,0) = x$.

During the iterative process of integration, several cells are crossed, starting from $c_1$ at integration time $t_1=1$, and finishing at $c_m$ at integration time $t_m$. The integration time $t_m$ of the last cell $c_m$ can be calculated by subtracting from the initial integration time the accumulated boundary hitting times $t_{hit}$: $t_m = t_1 - \sum_{i=1}^{m-1} t_{hit}^\theta(c_i, x_i)$. The final integration point $x_m$ is the boundary of the penultimate cell $c_{m-1}$: $x_m = x_{c_{m-1}}$. In case only one cell is visited, both time and space remain unchanged: $t_m = 1$ and $x_m = x$. Taking this into consideration, the trajectory can be calculated as follows:
\begin{equation}\label{eq:closed_form_integration:repeat}
\phi^\theta(x,t) = \psi^\theta(x=x_m,t=t_m) = 
\bigg(
    x e^{t a_c} + \Big(e^{t a_c}-1\Big) \frac{b_c}{a_c}
\bigg)_{\substack{x = x_m \\ t = t_m}}
\end{equation}
Gradient-based optimization algorithms require the derivatives of the transformation with respect to the model parameters $\theta$ and $x$. That is, we need to obtain: 
\begin{itemize}
  \item $\color{Magenta}\boxed{\color{black}\cfrac{\partial \phi^\theta(x,t)}{\partial \theta}}$: partial derivative of transformation $\phi^\theta(x,t)$ w.r.t. $\theta$. Presented at \ref{sec:closed_form_derivatives}.
  \item $\color{Violet}\boxed{\color{black}\cfrac{\partial \phi^\theta(x,t)}{\partial x}}$: partial derivative of transformation $\phi^\theta(x,t)$ w.r.t. $x$.
\end{itemize}

Note that in \cref{chapter:2}, only the derivative ${\partial \phi^\theta(x,t)}/{\partial \theta}$ was obtained as the spatial dimension $x$ was not a variable of the transformation and remained constant. 
In this case, however, the spatial variable is in the model, so the derivative ${\partial \phi^\theta(x,t)}/{\partial x}$ is also required.
In addition, a normalizing flow model involves computing the Jacobian determinant, so under gradient-based optimization techniques, second order derivatives are also required:
\begin{itemize}
  \item  $\color{ForestGreen}\boxed{\color{black}\cfrac{\partial^2 \phi^\theta(x,t)}{\partial x^2}}$: partial second derivative of transformation $\phi^\theta(x,t)$ w.r.t. $x$.
  \item  $\color{Orange}\boxed{\color{black}\cfrac{\partial^2 \phi^\theta(x,t)}{\partial x \partial \theta}}$: partial second derivative of transformation $\phi^\theta(x,t)$ w.r.t. $x$ and $\theta$.
\end{itemize}
\clearpage

\subsection{Closed-Form Derivatives of $\phi^\theta(x,t)$ w.r.t. $x$}\label{sec:closed_form_derivative_x}

The derivative can be calculated by going backwards in the integration direction. We focus on the partial derivative w.r.t. the spatial dimension $x$, and derive each of the terms of this derivative:
\begin{equation}
\color{Violet}\boxed{\color{black}\frac{\partial \phi^\theta(x,t)}{\partial x}}\color{black} = 
\bigg(
\frac{\partial \psi^\theta(x,t)}{\partial x} + 
\frac{\partial \psi^\theta(x,t)}{\partial t^\theta} \cdot
\frac{\partial t^\theta}{\partial x}
\bigg)_{\substack{x = x_m \\ t = t_m}}
\end{equation}

\subsubsection{Expression for $\cfrac{\partial \psi^\theta(x,t)}{\partial x}$}

Based on the expression for $\psi(x,t)$ from \cref{eq:closed_form_integration:repeat}, we can obtain the derivative w.r.t. $x$:
\begin{equation}
\phi^\theta(x,t) = \psi^\theta(x=x_m,t=t_m) = 
  \bigg(
      x e^{t a_c} + \Big(e^{t a_c}-1\Big) \frac{b_c}{a_c}
  \bigg)_{\substack{x = x_m \\ t = t_m}}
\end{equation}
It is important to note that the variable $x$ only affects this expression when $m=1$, otherwise is zero. That is, when the integration process does not traverse to adjacent cells the value of $x$ is present on $\phi^\theta(x,t)$, otherwise is $x_m$. Here $\indicator{C}$ is the indicator function that takes value 1 when the condition $C$ is true.
\begin{equation}
  \frac{\partial \psi^\theta(x,t)}{\partial x} = 
  e^{t a_c} \cdot \indicator{m=1} 
\end{equation}

\subsubsection{Expression for $\cfrac{\partial \psi^\theta(x,t)}{\partial t^\theta}$}

Similarly, from the same expression for $\psi(x,t)$, we explicitly get the derivative w.r.t $t^\theta$:
\begin{equation}\label{eq:psi_derivative_t:repeated}
    \frac{\partial \psi^\theta(x,t)}{\partial t^\theta} = x \, a_c \, e^{t a_c} + a_c \, e^{t a_c} \frac{b_c}{a_c} = 
    e^{t a_c} \big( a_c x + b_c \big)
\end{equation}

\subsubsection{Expression for $\cfrac{\partial t^\theta}{\partial x}$}

After visiting $m$ cells, the integration time $t^\theta$ can be expressed as:
\begin{equation}\label{eq:time:repeated}
t^\theta = t_1 - \sum_{i=1}^{m-1} t_{hit}^\theta(c_i, x_i)
\end{equation}
In this case, the variable $x$ only affects \cref{eq:time:repeated} if we move away from the first cell, i.e. $m>1$, and on the first cell $c=1$, which results the following expresion:
\begin{equation}\label{eq:time_derivative_x}
\frac{\partial t^\theta}{\partial x} = 
-\sum_{i=1}^{m-1} \frac{\partial t_{hit}^\theta(c_i, x_i)}{\partial x} = 
  -\cfrac{\partial t_{hit}^\theta(c, x)}{\partial x} \cdot \indicator{m>1} 
\end{equation}
where
\begin{equation}\label{eq:thit:repeat}
t_{hit}^\theta(c, x) = \frac{1}{a_c} \log \bigg( \frac{a_c x_c + b_c}{a_c x + b_c} \bigg)
\end{equation}
and $x_c$ is the boundary for cell index $c$. Now, apply the chain rule operation to the hitting time $t_{hit}^\theta(c, x)$ expression:
\begin{equation}\label{eq:thit_derivative_x}
\frac{\partial t_{hit}^\theta(c, x)}{\partial x} = 
\frac{1}{a_c} \frac{-a_{c}\cfrac{a_c x_c + b_c}{(a_c x + b_c)^2}}{\cfrac{a_c x_c + b_c}{a_c x + b_c}} = 
\frac{-1}{a_c x + b_c}
\end{equation}
Then, based on \cref{eq:time_derivative_x}, the derivative of the integration time $t^\theta$ can be expressed as:
\begin{equation}\label{eq:time_derivative_x_final}
\frac{\partial t^\theta}{\partial x} = 
-\cfrac{\partial t_{hit}^\theta(c, x)}{\partial x} \cdot \indicator{m>1} = 
\cfrac{1}{a_c x + b_c} \cdot \indicator{m>1} 
\end{equation}

\subsubsection{Final Expression for $\color{Violet}\boxed{\color{black}\cfrac{\partial \phi^\theta(x,t)}{\partial x}}$}

Joining all the terms together and evaluating the derivative at $x = x_m$ and $t = t_m$ yields the expression for the partial derivative w.r.t. $x$:

\begin{equation}\label{eq:derivative_x_complete}
\begin{aligned}
  \frac{\partial \phi^\theta(x,t)}{\partial x} &= 
  \bigg(
  \frac{\partial \psi^\theta(x,t)}{\partial x} + 
  \frac{\partial \psi^\theta(x,t)}{\partial t^\theta} \cdot
  \frac{\partial t^\theta}{\partial x}
  \bigg)_{\substack{x = x_m \\ t = t_m}} = \\ &= 
  e^{t_m a_{c_m}} \cdot \indicator{m=1}  + e^{t_m a_{c_m}} \cfrac{a_{c_m} x_m + b_{c_m}}{a_c x + b_c}  \cdot \indicator{m>1} = \\ &=
  \left\{\begin{array}{ll}
    e^{t_m a_{c_m}} \cfrac{a_{c_m} x_m + b_{c_m}}{a_c x + b_c} & \text{if} \quad m>1\\ 
    e^{t_m a_{c_m}} & \text{otherwise}
    \end{array}\right.
\end{aligned}
\end{equation}

\subsection{Closed-Form Derivatives of $\cfrac{\partial \phi^\theta(x,t)}{\partial x}$ w.r.t. $x$}\label{sec:closed_form_derivative_x_x}

In this case, we start from the previous \cref{eq:derivative_x_complete} and derive again w.r.t. $x$:

\begin{equation}
  \color{ForestGreen}\boxed{\color{black}\frac{\partial^2 \phi^\theta(x,t)}{\partial x^2}}\color{black} = 
  \bigg(
  \frac{\partial^2 \psi^\theta(x,t)}{\partial x^2} + 
  \frac{\partial^2 \psi^\theta(x,t)}{\partial x \, \partial t^\theta} \cdot
  \frac{\partial t^\theta}{\partial x} + 
  \frac{\partial \psi^\theta(x,t)}{\partial t^\theta} \cdot
  \frac{\partial^2 t^\theta}{\partial x^2}
  \bigg)_{\substack{x = x_m \\ t = t_m}}
\end{equation}

\subsubsection{Expression for $\cfrac{\partial^2 \psi^\theta(x,t)}{\partial x^2}$}

\begin{equation}
\frac{\partial^2 \psi^\theta(x,t)}{\partial x^2} = 
\frac{\partial}{\partial x}\Big(\frac{\partial \psi^\theta(x,t)}{\partial x}\Big) = 
\frac{\partial}{\partial x}\Big(e^{t a_c}\cdot \indicator{m>1}\Big) = 0
\end{equation}

\subsubsection{Expression for $\cfrac{\partial^2 \psi^\theta(x,t)}{\partial t^\theta \partial x}$}

\begin{equation}
  \frac{\partial^2 \psi^\theta(x,t)}{\partial t^\theta \partial x} = 
  \frac{\partial}{\partial x}\Big(\frac{\partial \psi^\theta(x,t)}{\partial t^\theta}\Big) = 
  \frac{\partial}{\partial x}\Big(e^{t a_c} \big( a_c x + b_c \big)\Big) = e^{t a_c} a_c
\end{equation}

\subsubsection{Expression for $\cfrac{\partial^2 t^\theta}{\partial x^2}$}

Again, in this case the variable $x$ only affects \cref{eq:time:repeated} if we move away from the first cell, i.e. $m>1$, and on the first cell $c=1$:

\begin{equation}
  \frac{\partial^2 t^\theta}{\partial x^2} = 
  \frac{\partial}{\partial x}\Big(\frac{\partial  t^\theta}{\partial x}\Big) = 
  \frac{\partial}{\partial x} \cfrac{1}{a_c x + b_c} \cdot \indicator{m>1} = 
  - \frac{a_{c}}{(a_{c} x + b_{c})^2} \cdot \indicator{m>1} 
\end{equation}

\subsubsection{Final Expression for $\color{ForestGreen}\boxed{\color{black}\cfrac{\partial^2 \phi^\theta(x,t)}{\partial x^2}}$}

Joining all the terms together and evaluating the derivative at $x = x_m$ and $t = t_m$ yields the expression for the partial derivative w.r.t. $x$:

\begin{equation}\label{eq:derivative_x2_complete}
\begin{aligned}
  \frac{\partial^2 \phi^\theta(x,t)}{\partial x^2} &= 
  \bigg(
  \frac{\partial^2 \psi^\theta(x,t)}{\partial x^2} + 
  \frac{\partial^2 \psi^\theta(x,t)}{\partial t^\theta \partial x} \cdot
  \frac{\partial t^\theta}{\partial x} + 
  \frac{\partial \psi^\theta(x,t)}{\partial t^\theta} \cdot
  \frac{\partial^2 t^\theta}{\partial x^2}
  \bigg)_{\substack{x = x_m \\ t = t_m}} = \\ &=
  e^{t_m a_{c_m}} a_{c_m} \cfrac{1}{a_c x + b_c} \cdot \indicator{m>1} - 
  e^{t_m a_{c_m}} \big( a_{c_m} x_m + b_{c_m} \big) \frac{a_{c}}{(a_{c} x + b_{c})^2} \cdot \indicator{m>1} 
\end{aligned}
\end{equation}

\subsection{Closed-Form Derivatives of $\cfrac{\partial \phi^\theta(x,t)}{\partial x}$ w.r.t. $\theta$}\label{sec:closed_form_derivative_x_theta}

We start from \cref{eq:derivative_x_complete} and derive w.r.t. one of the coefficients of $\theta$, i.e., $\theta_k$:
\begin{equation}
  \color{Orange}\boxed{\color{black}\frac{\partial^2 \phi^\theta(x,t)}{\partial \theta_k \, \partial x}}\color{black} = 
  \bigg(
  \frac{\partial^2 \psi^\theta(x,t)}{\partial \theta_k \, \partial x} + 
  \frac{\partial^2 \psi^\theta(x,t)}{\partial \theta_k \partial t^\theta} \cdot
  \frac{\partial t^\theta}{\partial x} + 
  \frac{\partial \psi^\theta(x,t)}{\partial t^\theta} \cdot
  \frac{\partial^2 t^\theta}{\partial \theta_k \partial x}
  \bigg)_{\substack{x = x_m \\ t = t_m}}
\end{equation}
Note that the slope $a_c$ and intercept $b_c$ are a linear combination of the orthogonal basis $B$ of the constraint matrix $L$, with $\theta$ as coefficients. We recommend the reader to review \cref{sec:velocity_continuity_constraints,sec:null_space} for more information about these matrices.
\begin{equation}\label{eq:vec_A:repeat}
vec(\textbf{A}) = \textbf{B} \cdot \boldsymbol{\theta} = \sum_{j=1}^{d} \theta_j \cdot \textbf{B}_j
\end{equation}
If we define one of the components from the orthogonal basis $\textbf{B}_j$ as:
\begin{equation}\label{eq:basis:repeat}
\textbf{B}_{j} = \begin{bmatrix} a_1^{(j)} & b_1^{(j)} & \cdots & a_c^{(j)} & b_c^{(j)} & \cdots & a_{N_\mathcal{P}}^{(j)} & b_{N_\mathcal{P}}^{(j)}\end{bmatrix}^T
\end{equation}
Then,
\begin{equation}\label{eq:vec_A_complete:repeat}
vec(\textbf{A}) = \sum_{j=1}^{d} \theta_j \cdot \textbf{B}_j = \begin{bmatrix} & \cdots & \sum_{j=1}^{d} \theta_j a_c^{(j)} & \sum_{j=1}^{d} \theta_j b_c^{(j)} & \cdots & \end{bmatrix}^T
\end{equation}
Thus, the slope $a_c$ and intercept $b_c$ (parameters of the affine transformation) and their derivatives w.r.t. one of the coefficients of $\theta$, i.e., $\theta_k$, are denoted as follows:
\begin{equation}\label{eq:ac_bc:repeat}
a_c = \sum_{j=1}^{d} \theta_j a_c^{(j)} \quad\quad\quad
b_c = \sum_{j=1}^{d} \theta_j b_c^{(j)}
\end{equation}
\begin{equation}\label{eq:ac_bc_partial}
  \frac{\partial a_c}{\partial \theta_k} = a_c^{(k)} \quad\quad\quad
  \frac{\partial b_c}{\partial \theta_k} = b_c^{(k)}
  \end{equation}

\subsubsection{Expression for $\cfrac{\partial^2 \psi^\theta(x,t)}{\partial \theta_k \, \partial x}$}

Given that $\cfrac{\partial \psi^\theta(x,t)}{\partial x} = e^{t a_c} \cdot \indicator{m=1}$, we derive w.r.t. one of the coefficients of $\theta$, i.e., $\theta_k$:

\begin{equation}
\frac{\partial^2 \psi^\theta(x,t)}{\partial \theta_k \, \partial x} =
\frac{\partial^2 \psi^\theta(x,t)}{\partial a_c \, \partial x} \cdot \frac{\partial a_c}{\partial \theta_k} = 
t \cdot e^{t a_c} \cdot \indicator{m=1} \cdot a_c^{(k)} 
\end{equation}

\subsubsection{Expression for $\cfrac{\partial^2 \psi^\theta(x,t)}{\partial \theta_k \partial t^\theta}$}

Given that $\cfrac{\partial \psi^\theta(x,t)}{\partial t^\theta} = e^{t a_c} \big( a_c x + b_c \big)$, we derive w.r.t. one of the coefficients of $\theta$, i.e., $\theta_k$:
\begin{equation}\label{eq:derivative_phi_x_thetak}
  \frac{\partial^2 \psi^\theta(x,t)}{\partial \theta_k \, \partial t^\theta} =
  \frac{\partial^2 \psi^\theta(x,t)}{\partial {t^\theta} \, \partial t^\theta} \cdot \frac{\partial t^\theta}{\partial \theta_k} +
  \frac{\partial^2 \psi^\theta(x,t)}{\partial a_c \, \partial t^\theta} \cdot \frac{\partial a_c}{\partial \theta_k} +
  \frac{\partial^2 \psi^\theta(x,t)}{\partial b_c \, \partial t^\theta} \cdot \frac{\partial b_c}{\partial \theta_k}
\end{equation}

We reutilize $\cfrac{\partial t^\theta}{\partial \theta_k}$ from \cref{sec:expression_derivative_3}, and derive each remaining term.

\begin{equation}\label{eq:derivative_t_t}
\frac{\partial^2 \psi^\theta(x,t)}{\partial {t^\theta} \, \partial t^\theta} = 
e^{t a_c} ( a_c x + b_c) \cdot a_c
\end{equation}

\begin{equation}\label{eq:derivative_t_a}
\frac{\partial^2 \psi^\theta(x,t)}{\partial a_c \, \partial t^\theta} = 
\frac{\partial}{\partial a_c}\bigg( e^{t a_c} ( a_c x + b_c) \bigg) = 
t e^{t a_c} (a_c x + b_c) + e^{t a_c} x = 
e^{t a_c} (t(a_c x + b_c) + x)
\end{equation}

\begin{equation}\label{eq:derivative_t_b}
\frac{\partial^2 \psi^\theta(x,t)}{\partial b_c \, \partial t^\theta} = 
\frac{\partial}{\partial b_c}\bigg( e^{t a_c} ( a_c x + b_c) \bigg) = 
e^{t a_c} 
\end{equation}

Therefore, joining the expressions from \cref{eq:derivative_t_t,eq:derivative_t_a,eq:derivative_t_b} into \cref{eq:derivative_phi_x_thetak}:

\begin{equation}
\begin{aligned}
\frac{\partial^2 \psi^\theta(x,t)}{\partial \theta_k \, \partial t^\theta} =
e^{t a_c} ( a_c x + b_c) a_c \cdot \frac{\partial t^\theta}{\partial \theta_k} + 
e^{t a_c} (t(a_c x + b_c) + x) \cdot a_c^{(k)} + 
e^{t a_c}  \cdot b_c^{(k)}
\end{aligned}
\end{equation}

\subsubsection{Expression for $\cfrac{\partial^2 t^\theta}{\partial \theta_k \partial x}$}

Given that $\cfrac{\partial t^\theta}{\partial x} = \cfrac{1}{a_c x + b_c}  \cdot \indicator{m>1}$, we derive w.r.t. one of the coefficients of $\theta$, i.e., $\theta_k$:

\begin{equation}
  \begin{aligned}  
    \frac{\partial^2 t^\theta}{\partial \theta_k \, \partial x} &=
    \frac{\partial^2 t^\theta}{\partial a_c \, \partial x} \cdot \frac{\partial a_c}{\partial \theta_k} +
    \frac{\partial^2 t^\theta}{\partial b_c \, \partial x} \cdot \frac{\partial b_c}{\partial \theta_k} = \\ &=
    \frac{-x}{(a_c x + b_c)^2}\cdot a_c^{(k)} + 
    \frac{-1}{(a_c x + b_c)^2}\cdot b_c^{(k)}
\end{aligned}
\end{equation}

\subsubsection{Final Expression for $\color{Orange}\boxed{\color{black}\cfrac{\partial^2 \phi^\theta(x,t)}{\partial \theta_k \, \partial x}}$}

Joining all the terms together and evaluating the derivative at $x = x_m$ and $t = t_m$ yields the expression for the partial derivative w.r.t. $x$:
\begin{equation}
\begin{aligned}
  \frac{\partial^2 \phi^\theta(x,t)}{\partial \theta_k \, \partial x} &= 
  \bigg(
  \frac{\partial^2 \psi^\theta(x,t)}{\partial \theta_k \, \partial x} + 
  \frac{\partial^2 \psi^\theta(x,t)}{\partial \theta_k \partial t^\theta} \cdot
  \frac{\partial t^\theta}{\partial x} + 
  \frac{\partial \psi^\theta(x,t)}{\partial t^\theta} \cdot
  \frac{\partial^2 t^\theta}{\partial \theta_k \partial x}
  \bigg)_{\substack{x = x_m \\ t = t_m}} = \\ &= 
  t \cdot e^{t a_c} \cdot \indicator{m=1} \cdot a_c^{(k)} + \\
  &\quad\, \Big (
  e^{t a_c} ( a_c x + b_c) a_c \cdot \frac{\partial t^\theta}{\partial \theta_k} + 
  e^{t a_c} (t(a_c x + b_c) + x) \cdot a_c^{(k)} + 
  e^{t a_c}  \cdot b_c^{(k)} 
  \Big) \cdot
  \cfrac{1}{a_c x + b_c}  \cdot \indicator{m>1} + \\
  &\quad\, e^{t a_c} ( a_c x + b_c ) \cdot
  \Big (
  \frac{-x}{(a_c x + b_c)^2}\cdot a_c^{(k)} + 
  \frac{-1}{(a_c x + b_c)^2}\cdot b_c^{(k)}
  \Big )
\end{aligned}
\end{equation}

\clearpage

\section{Experiments and Results}\label{sec:results_6}

The performance of DIFW-NF flows were evaluated on both synthetic and real-world datasets for density estimation, and compare it to several alternative autoregressive models and flow based methods.
For these experiments, we define the forward and backward operators of the proposed normalizing flow model. The forward (generative) direction samples data $\mathbf{z}$ from a known distribution  $p(\mathbf{z})$ and applies a transformation $f$ to generate data samples $\mathbf{x} = f(\mathbf{z})$.
Similarly, the backward (normalizing) direction takes real data samples  $\mathbf{x} \sim p(\mathbf{x})$ and applies the inverse transformation $f^{-1}$. Model's loss function is simply the negative log-likelihood. 

\subsection{1-dimensional Data}

First, we performed a host of experiments on one-dimensional simulated data to gain in-depth understanding of DIFW-NF. Five datasets are tested: 
\textsc{blobs} generates isotropic one-dimensional Gaussian blobs,
\textsc{gaussian} creates an unimodal Gaussian distribution,
\textsc{gaussianmix} generates multimodal data from a mixture of Gaussians,
\textsc{power} applies the power transformation $5^x$
and \textsc{uniform} samples from a uniform 0-1 distribution. 

A grid-hyperparameter search is conducted for each dataset. 
Flows can be composed after $1$,$2$ or $3$ steps and the transformation tessellation size is chosen among $\{4,8,16,32\}$.
The Adam optimizer \cite{kingma2014adam} is used with default hyperparameters and an initial learning rate of $\{1e^{-4}, 1e^{-3}, 5e^{-3}\}$ over $500$ training epochs with batch size $256$. For training and test, $5000$ and $2000$ data points are used respectively. 
The final hyperparameters are shown in \cref{tab:nf_1d}, along with the log-likelihood of the generated data.

\begin{table}[!htb]
  \small
  \caption{Hyperparameters and log-probability for density-estimation results in 1D-datasets}
  \label{tab:nf_1d}
  \begin{center}
  \begin{tabular}{llllll}
    \toprule
    Dataset & \textsc{blobs} & \textsc{gaussian} & \textsc{gaussianmix} & \textsc{power} & \textsc{uniform} \\
    \midrule
    Train Size           &              5000 &              5000 &              5000 &             5000 &             5000 \\
    Test Size            &              2000 &              2000 &              2000 &             2000 &             2000 \\
    Batch Size           &               256 &               256 &               256 &              256 &              256 \\
    Tessellation Size    &                32 &                32 &                32 &                8 &               32 \\
    Flow Steps           &                 2 &                 2 &                 2 &                3 &                2 \\
    Epochs               &               500 &               500 &               500 &              500 &              500 \\
    Learning Rate        &             0.005 &             0.005 &             0.001 &            0.005 &            0.005 \\ \midrule
    $\log p(\mathbf{x})$ &  -3.70 $\pm$ 0.22 &  -1.49 $\pm$ 0.08 &  -1.22 $\pm$ 0.14 &  0.80 $\pm$ 0.02 &  0.01 $\pm$ 0.01 \\
    \bottomrule
    \end{tabular}    
  \end{center}
\end{table}

\begin{figure}[!htb]
  \begin{center}
    \begin{subfigure}[b]{0.32\linewidth}
      \includegraphics[width=\linewidth, trim=32 35 445 0, clip]{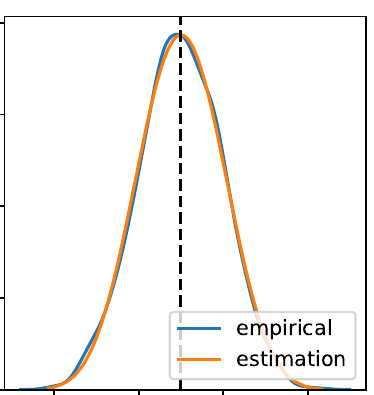}
      \caption{Base distribution $q(\mathbf{z})$}
    \end{subfigure}
    \begin{subfigure}[b]{0.32\linewidth}
      \includegraphics[width=\linewidth, trim=463 35 15 0, clip]{figures/1D/BLOBS/plot_1D.pdf}
      \caption{Transformation $f(\mathbf{z})$}
    \end{subfigure} 
    \begin{subfigure}[b]{0.32\linewidth}
      \includegraphics[width=\linewidth, trim=245 35 232 0, clip]{figures/1D/BLOBS/plot_1D.pdf}
      \caption{Target distribution $p(\mathbf{x})$}
    \end{subfigure} 
    \\
    \begin{subfigure}[b]{0.48\linewidth}
      \includegraphics[width=\linewidth, trim=0 0 0 20, clip]{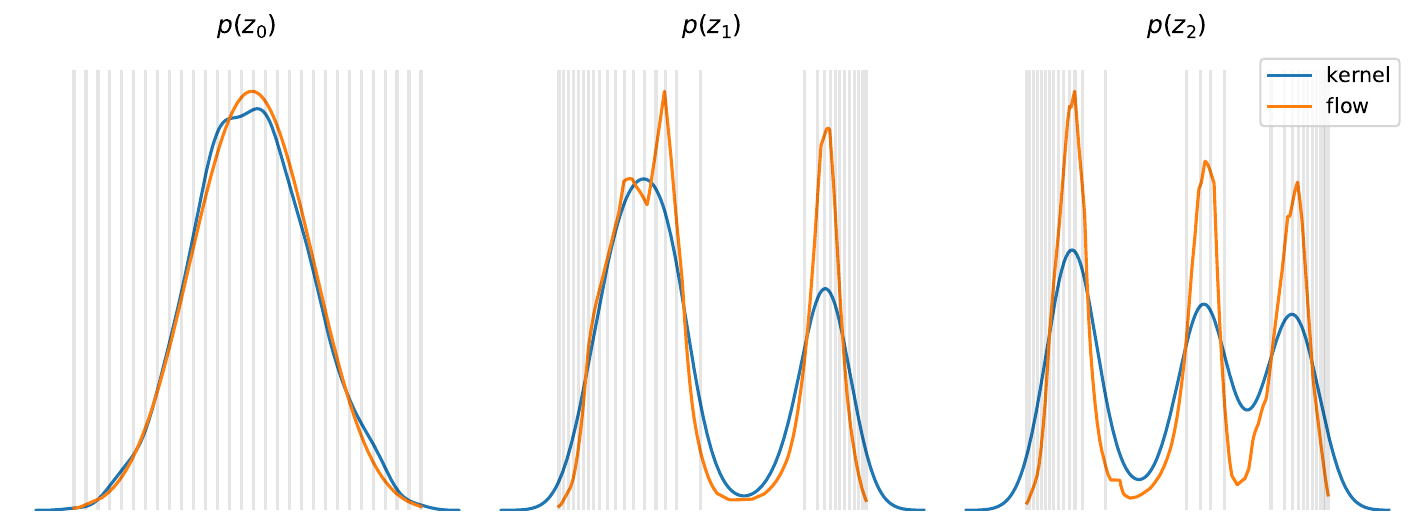}
      \caption{Generative flow: noise $\Rightarrow$ data}
      \label{fig:nf_1d_example:generative}
    \end{subfigure}
    \hfill{\color{lightgray}\vrule}\hfill
    \begin{subfigure}[b]{0.48\linewidth}
      \includegraphics[width=\linewidth, trim=0 0 0 26, clip]{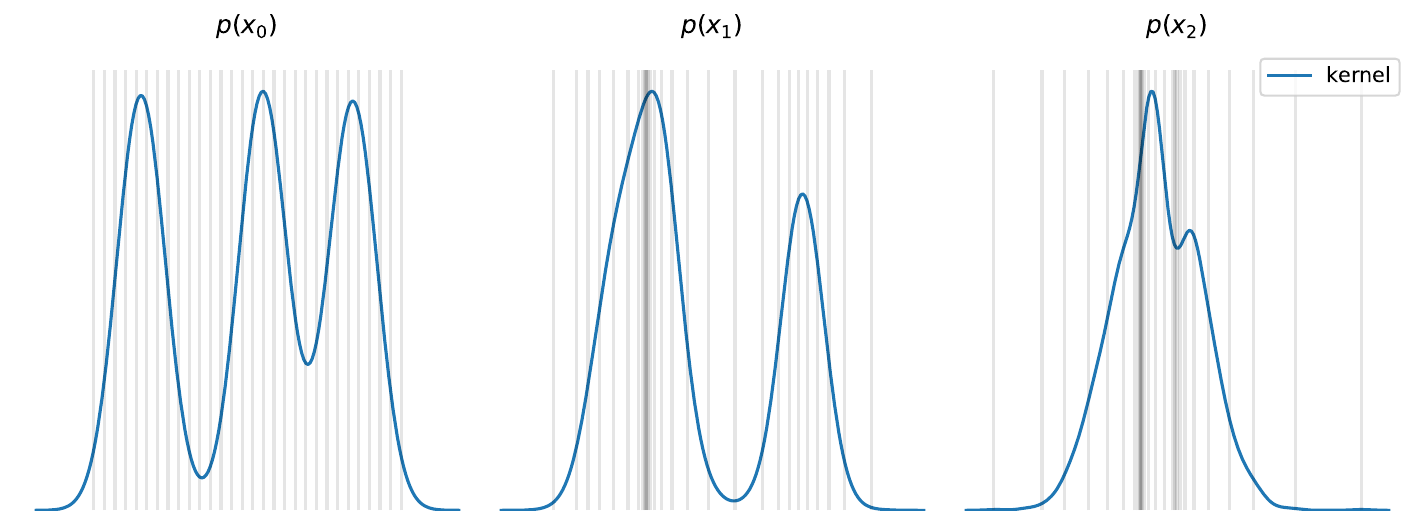}
      \caption{Normalizing flow: data $\Rightarrow$ noise}
      \label{fig:nf_1d_example:normalizing}
    \end{subfigure}
  \caption{DIFW-NF model applied to 1D \textsc{blobs} dataset: kernel probability density estimate in \textcolor{blue}{blue} and inferred probability using the change-of-variable formula in \textcolor{orange}{orange}.}
  \label{fig:nf_1d_example}
  \end{center}
\end{figure}

\cref{fig:nf_1d_example} shows a detailed example of a one-dimensional flow using the \textsc{blobs} dataset. The base $q(\mathbf{z})$ and target $p(\mathbf{x})$ distributions are shown, as well as the applied monotonic function $f(\mathbf{z})$. In this case, the flow is applied in two steps. 
\cref{fig:nf_1d_example:generative} shows the generative direction, in which the function $f(\mathbf{z})$ maps samples $\mathbf{z}$ from the latent distribution into approximate samples $\mathbf{x}$ from the data distribution. This corresponds to exact generation of samples from the model.
\cref{fig:nf_1d_example:normalizing} shows the normalizing direction, in which the inverse function $f^{-1}(\mathbf{x})$ maps samples $\mathbf{x}$ from the data distribution into approximate samples $\mathbf{z}$ from the latent distribution. This corresponds to the exact inference of the latent state given the data. 

\begin{figure}[!htb]
  \begin{center}
    \begin{subfigure}[b]{0.48\linewidth}
      \centering
      \includegraphics[width=\linewidth, trim=0 0 0 20, clip]{figures/1D/BLOBS/plot_generative_flow_evolution.pdf}
      \includegraphics[width=\linewidth, trim=0 0 0 26, clip]{figures/1D/BLOBS/plot_normalizing_flow_evolution.pdf}
      \caption{\textsc{blobs} dataset}
      \label{fig:NF_1D_BLOBS}
    \end{subfigure}
    \hfill{\color{lightgray}\vrule}\hfill
    \begin{subfigure}[b]{0.48\linewidth}
      \centering
      \includegraphics[width=\linewidth, trim=0 0 0 20, clip]{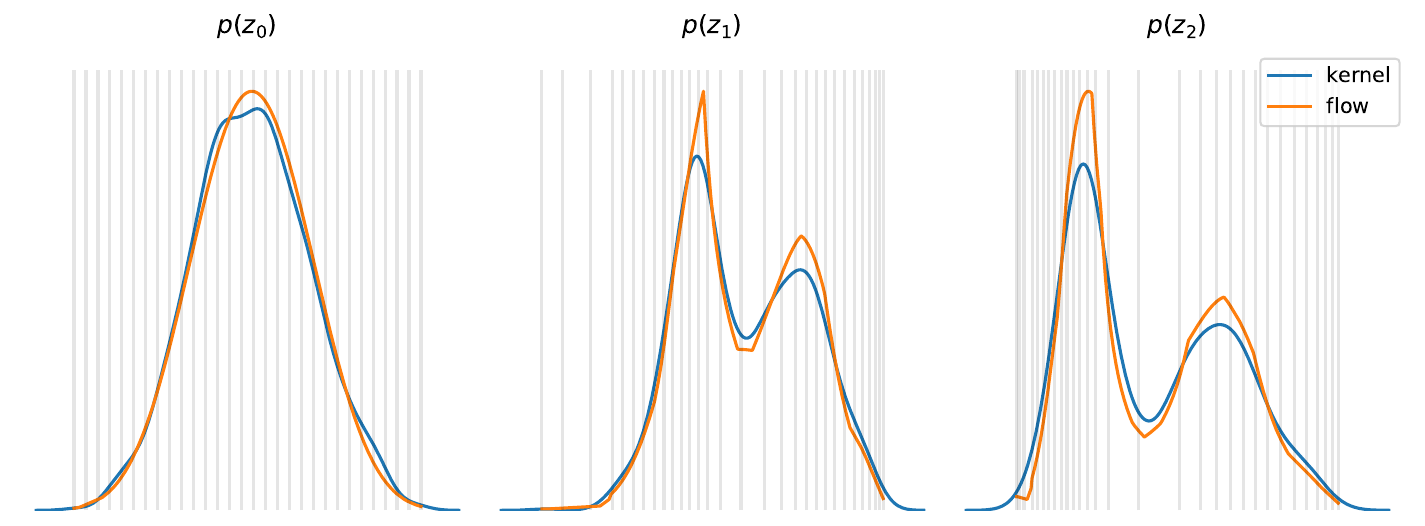}
      \includegraphics[width=\linewidth, trim=0 0 0 26, clip]{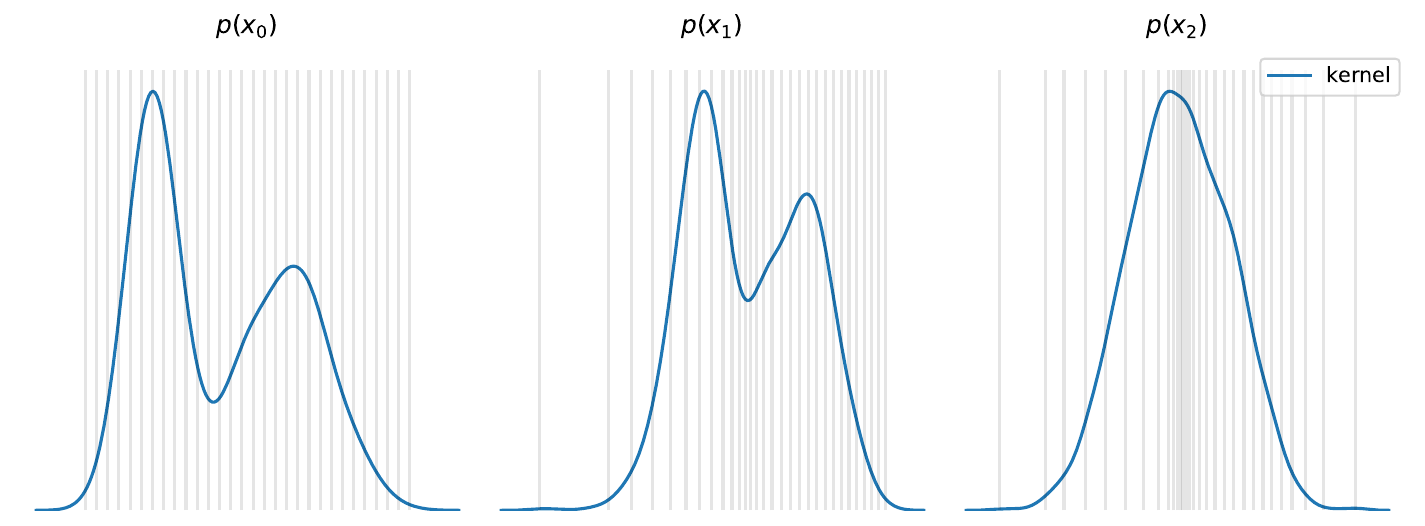}
      \caption{\textsc{gaussianmix} dataset}
      \label{fig:NF_1D_GAUSSIANMIXTURE}
    \end{subfigure}
    \vspace{1em}
    \begin{subfigure}[b]{0.48\linewidth}
      \centering
      \includegraphics[width=\linewidth, trim=0 0 0 20, clip]{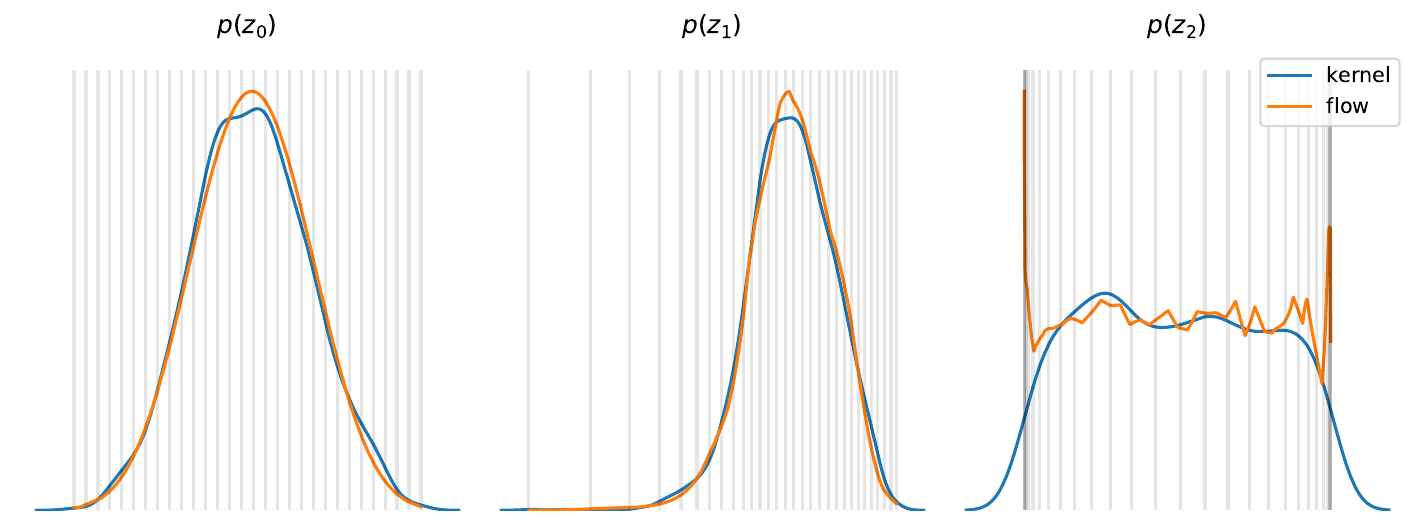}
      \includegraphics[width=\linewidth, trim=0 0 0 26, clip]{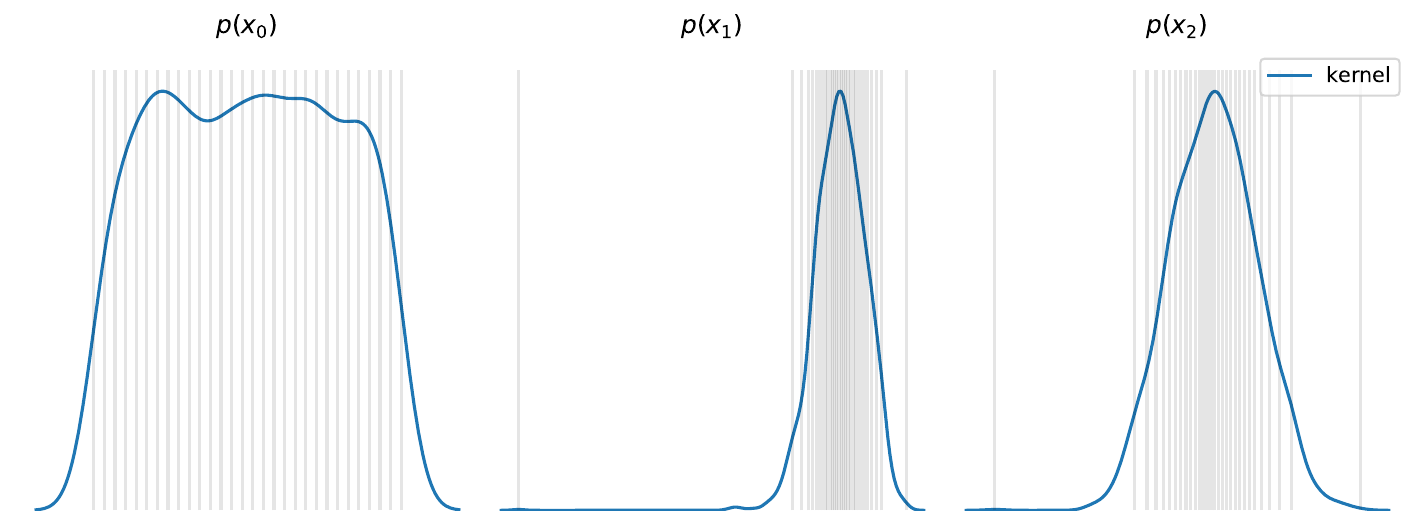}
      \caption{\textsc{uniform} dataset}
      \label{fig:NF_1D_UNIFORM}
    \end{subfigure}
    \hfill{\color{lightgray}\vrule}\hfill
    \begin{subfigure}[b]{0.48\linewidth}
      \centering
      \includegraphics[width=\linewidth, trim=0 0 0 20, clip]{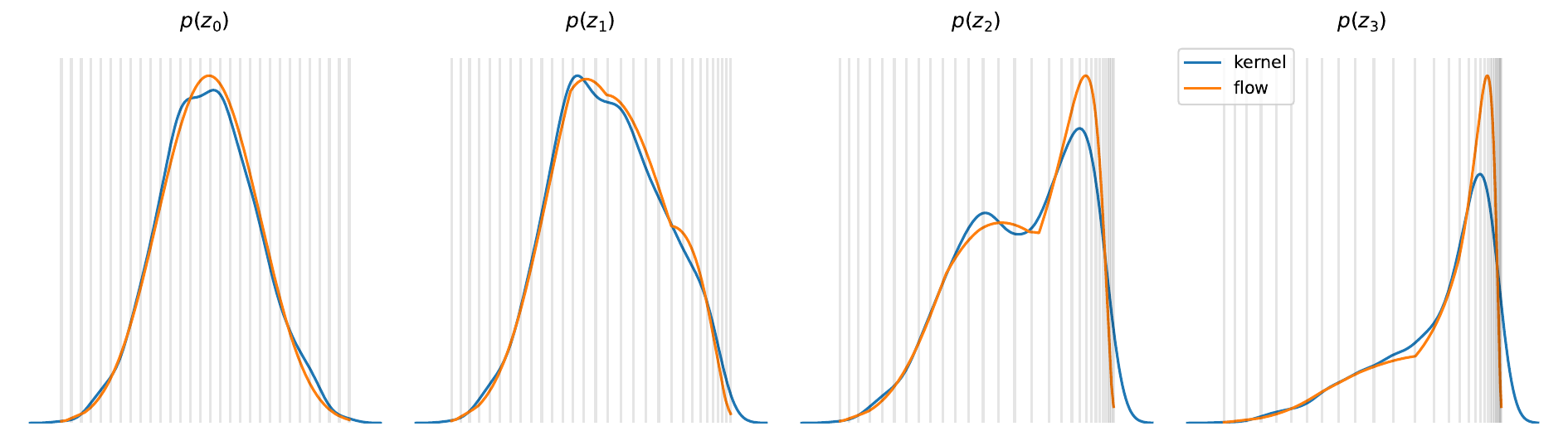}
      \includegraphics[width=\linewidth, trim=0 0 0 26, clip]{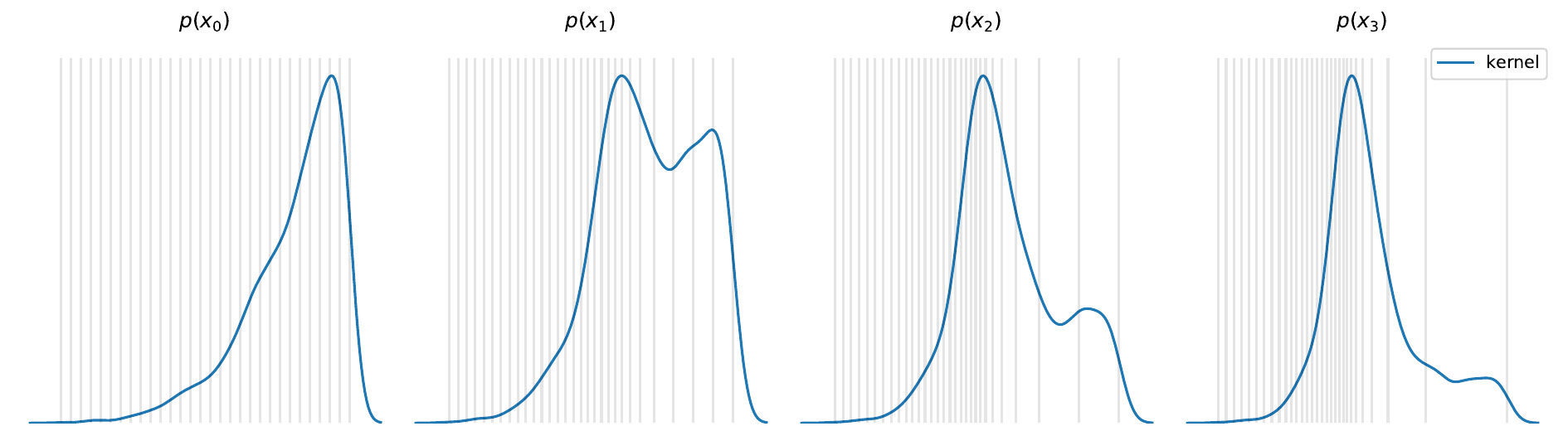}
      \caption{\textsc{power} dataset}
      \label{fig:NF_1D_POWER}
    \end{subfigure}
    \caption{1D normalizing flows on multiple datasets. \textbf{Top}: Generative flow from Gaussian distribution to data. Kernel probability density estimate in \textcolor{blue}{blue} and inferred probability using the change-of-variable formula in \textcolor{orange}{orange}. \textbf{Bottom}: Normalizing flow from data back to the Gaussian distribution.}
    \label{fig:nf_1d_datasets}
  \end{center}
\end{figure}

More results on other one-dimensional datasets are presented in \cref{fig:nf_1d_datasets}. The visual comparison between the empirical kernel density function (\textcolor{blue}{blue}) and the  probability density value inferred using the change-of-variable formula \textcolor{orange}{orange} indicates that the proposed model is flexible enough to appropriately capture the target data distribution. Based on these results we can proceed to 2D and high-dimensional data.

\subsection{2-dimensional Data}

In this section the flexibility of DIFW-NF is demonstrated on synthetic two-dimensional datasets. A fully-connected neural network with ReLu activation function computes the parameters of the element-wise transformations.
A grid-hyperparameter search is conducted for each dataset. Flows can be composed after $1$,$2$,$3$,$4$ or $8$ steps and the transformation tessellation size is chosen among $\{4,8,16,32\}$.
In this case, a grid-search over network architectures was also performed; we searched over models with $1$,$2$ or $4$ layers with $8$ or $16$ hidden layers per flow.
The Adam optimizer \cite{kingma2014adam} is used with default hyperparameters and an initial learning rate of $1e^{-4}$ over $500$ training epochs with batch size $256$. For training and test, $5000$ and $2000$ data points are used respectively. 
The final hyperparameters are shown in \cref{tab:nf_2d}, along with the log-likelihood of the generated data.

\begin{table}[!htb]
  \small
  \caption{Hyperparameters and log-probability for density-estimation results in 2D-datasets}
  \label{tab:nf_2d}
  \vspace{-1em}
  \setstretch{0.87}
  \begin{center}
  \begin{tabular}{lccccl}
    \toprule
          Dataset & \begin{tabular}[c]{l}\# Neurons\\ per Layer\end{tabular} & \begin{tabular}[c]{l}\# Hidden\\Layers\end{tabular} & \begin{tabular}[c]{l}Tessellation\\Size $N_\mathcal{P}$\end{tabular} & \begin{tabular}[c]{l}Flow\\Steps\end{tabular} & $\log p(\mathbf{x})$ \\
    \midrule
               \textsc{abs} &                   16 &                1 &                32 &          2 &     -1.15 $\pm$ 0.06 \\
      \textsc{checkerboard} &                   16 &                4 &                32 &          1 &     -3.71 $\pm$ 0.06 \\
           \textsc{circles} &                   16 &                2 &                32 &          1 &     -1.24 $\pm$ 0.06 \\
          \textsc{crescent} &                   16 &                2 &                 4 &          1 &     -1.79 $\pm$ 0.06 \\
     \textsc{crescentcubed} &                    8 &                4 &                 4 &          1 &     -1.75 $\pm$ 0.09 \\
           \textsc{diamond} &                   16 &                4 &                32 &          1 &     -3.40 $\pm$ 0.06 \\
       \textsc{fourcircles} &                   16 &                4 &                 8 &          2 &     -2.90 $\pm$ 0.06 \\
             \textsc{moons} &                   16 &                4 &                32 &          2 &     -1.10 $\pm$ 0.06 \\
              \textsc{sign} &                   16 &                2 &                32 &          3 &     -1.43 $\pm$ 0.07 \\
          \textsc{sinewave} &                   16 &                2 &                32 &          3 &     -1.96 $\pm$ 0.05 \\
        \textsc{twospirals} &                   16 &                2 &                 4 &          8 &     -2.81 $\pm$ 0.05 \\
    \bottomrule
    \end{tabular}
  \end{center}
\end{table}
\begin{figure}[!htb]
  \begin{center}
    \begin{subfigure}{\linewidth}
        \centering
        \includegraphics[width=\linewidth,trim=0cm 0cm 0cm 1.6cm, clip]{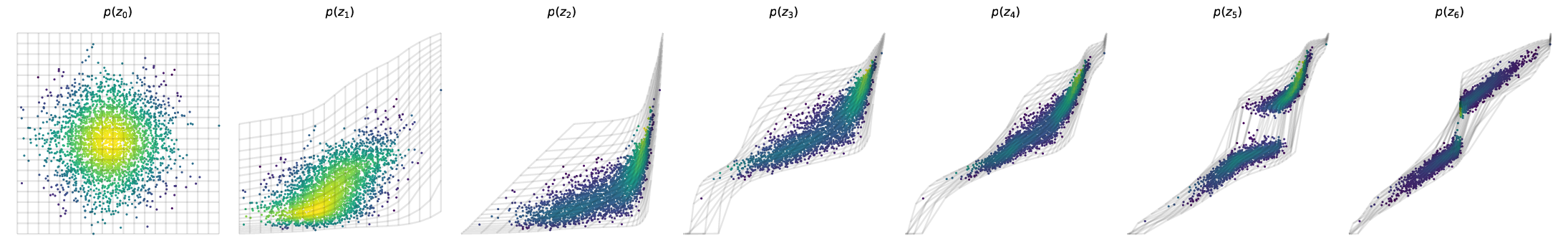}
        \includegraphics[width=\linewidth,trim=0cm 0cm 0cm 1.6cm, clip]{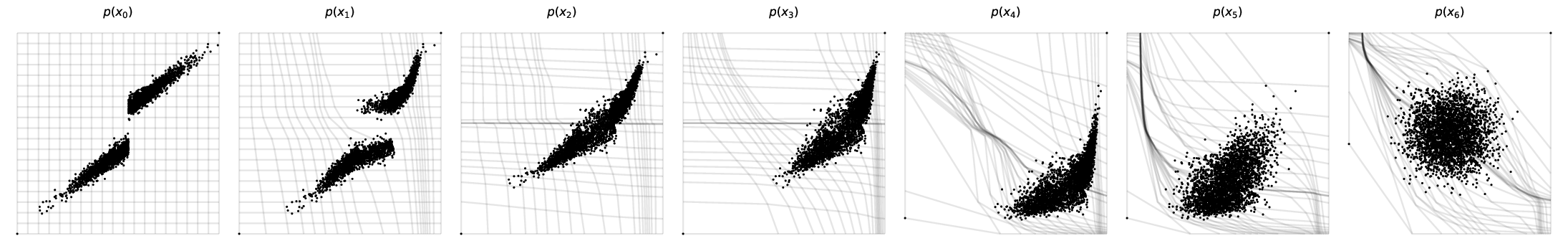}
        \caption{\textsc{sign} dataset}
        \label{fig:NF_2D_SIGN}
    \end{subfigure}
    \vspace{1em}
    \begin{subfigure}{\linewidth}
      \centering
      \includegraphics[width=\linewidth,trim=0cm 0cm 0cm 1.6cm, clip]{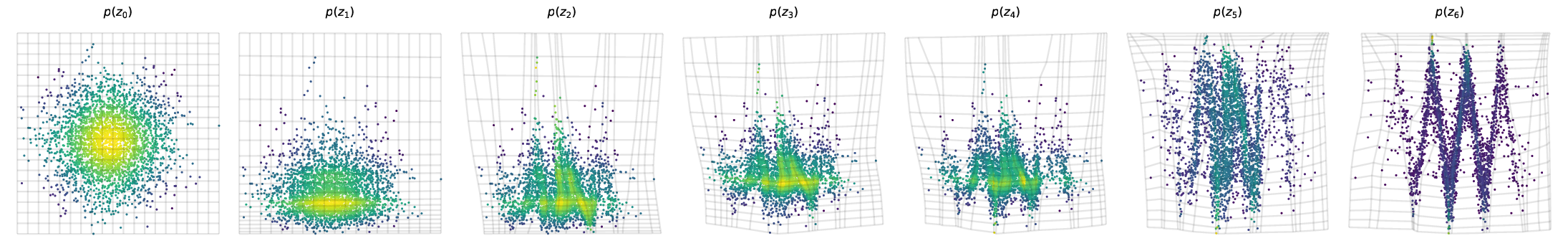}
      \includegraphics[width=\linewidth,trim=0cm 0cm 0cm 1.6cm, clip]{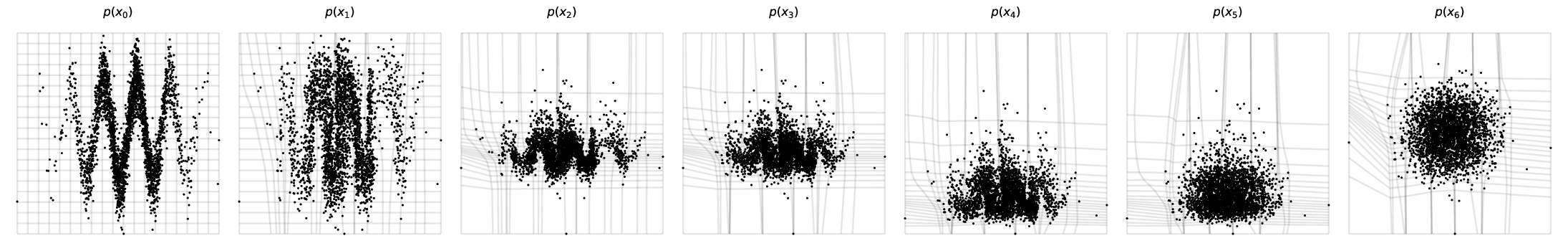}
      \caption{\textsc{sinewave} dataset}
      \label{fig:NF_2D_SINEWAVE}
    \end{subfigure}
  \vspace{-2em}
  \caption{Density estimation for two-dimensional synthetic datasets, including multi-modal and discontinuous densities. Flow transforms samples from a standard-normal base density to the target density. The top row shows the generative direction, visualizing the transformation from noise to data: $\mathbf{z} \sim p(\mathbf{z}) \rightarrow \mathbf{x} = f(\mathbf{z})$. 
  The color scale represents the probability density function. 
  Normalizing flows are reversible, so one can train on a density estimation task and still be able to sample from the learned density efficiently. 
  The bottom row shows the normalizing direction from data to noise: $\mathbf{x} \sim p(\mathbf{x}) \rightarrow \mathbf{z} = f^{-1}(\mathbf{x})$.}
  \label{fig:nf_2d_examples:1}
\end{center}
\end{figure}

\clearpage

\begin{figure}[!htb]\ContinuedFloat
  \begin{center}
    \begin{subfigure}{\linewidth}
        \centering
        \includegraphics[width=\linewidth,trim=0cm 0cm 0cm 1.6cm, clip]{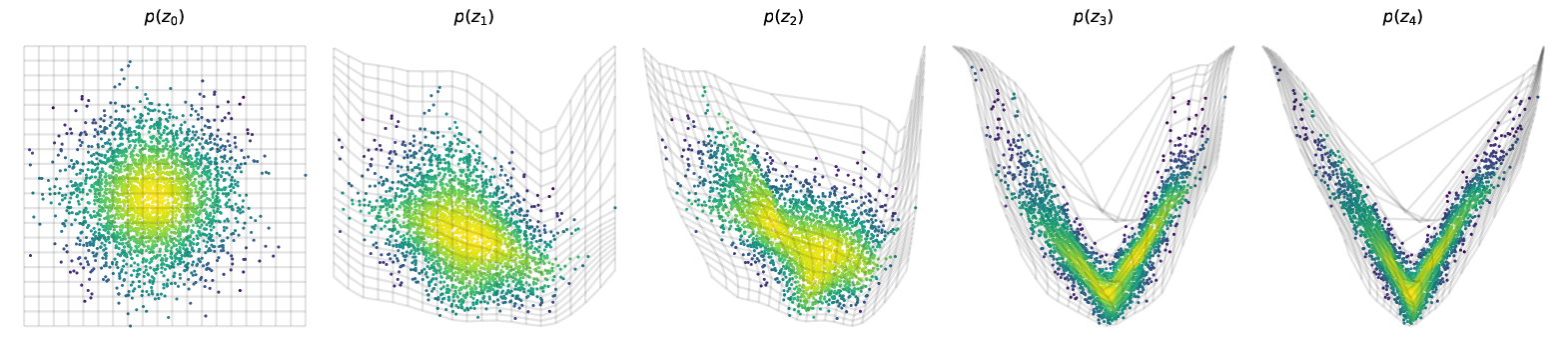}
        \includegraphics[width=\linewidth,trim=0cm 0cm 0cm 1.6cm, clip]{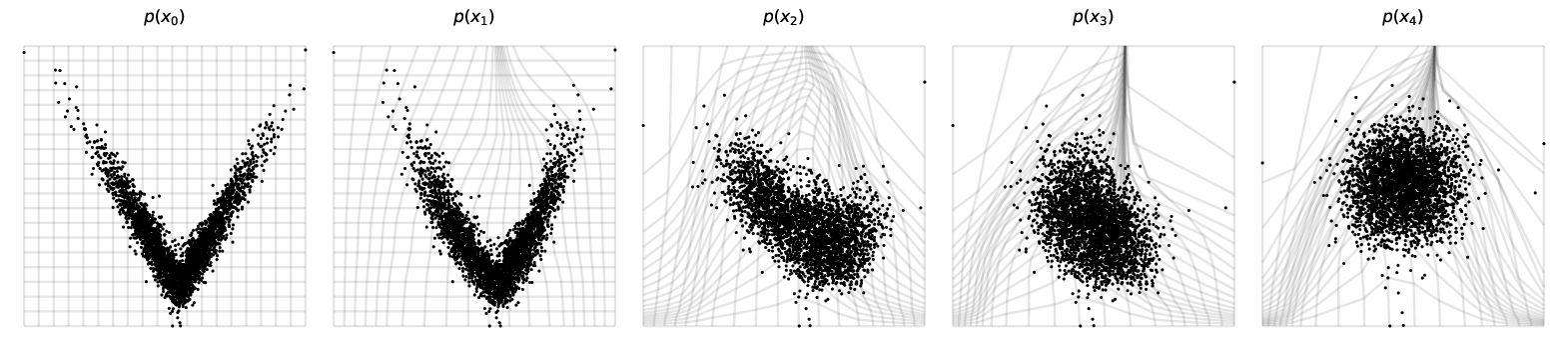}
        \caption{\textsc{abs} dataset}
        \label{fig:NF_2D_ABS}
    \end{subfigure}
    \vspace{1em}
    \begin{subfigure}{\linewidth}
      \centering
      \includegraphics[width=\linewidth,trim=0cm 0cm 0cm 1.6cm, clip]{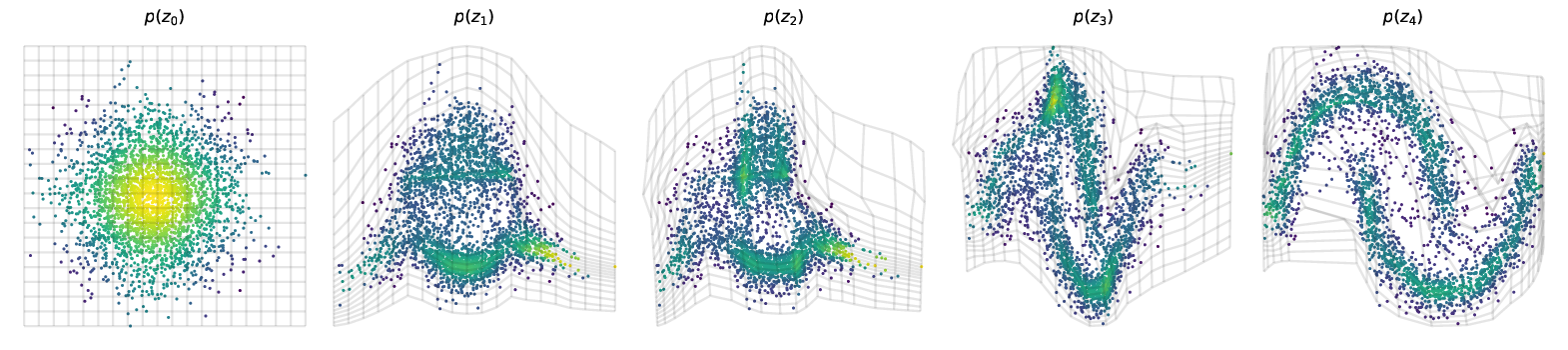}
      \includegraphics[width=\linewidth,trim=0cm 0cm 0cm 1.6cm, clip]{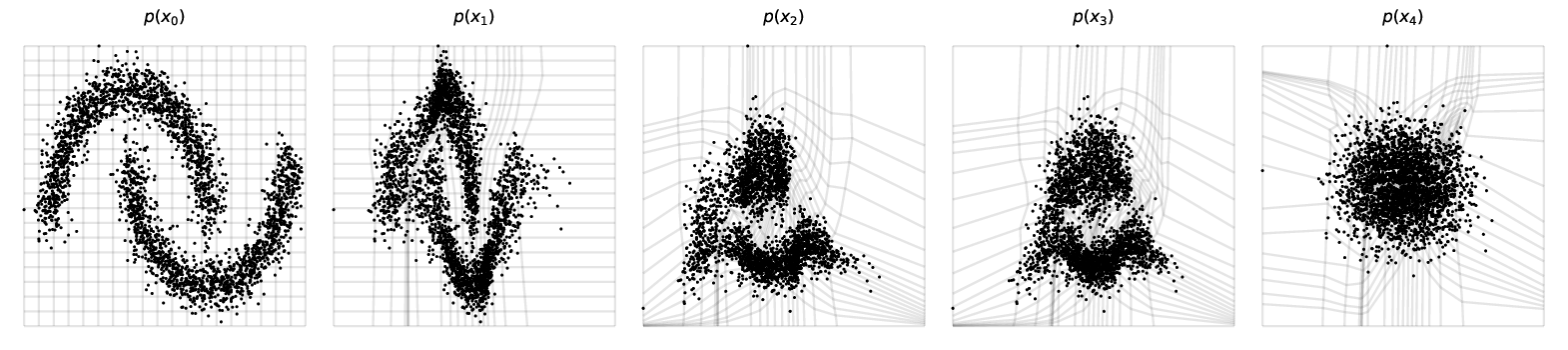}
      \caption{\textsc{moons} dataset}
      \label{fig:NF_2D_MOONS}
    \end{subfigure}
    \vspace{1em}
    \begin{subfigure}{\linewidth}
      \centering
      \includegraphics[width=\linewidth,trim=0cm 0cm 0cm 1.6cm, clip]{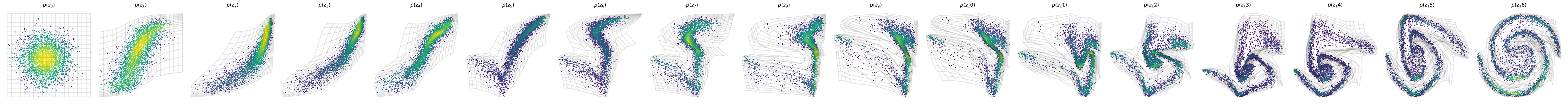}
      \includegraphics[width=\linewidth,trim=0cm 0cm 0cm 1.6cm, clip]{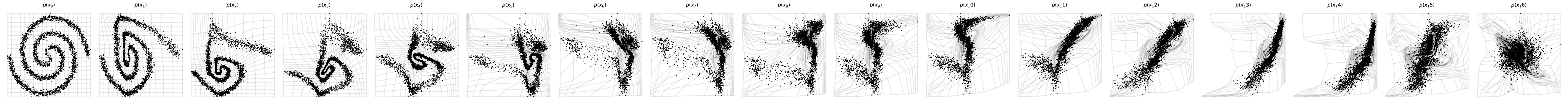}
      \caption{\textsc{twospirals} dataset}
      \label{fig:NF_2D_TWOSPIRALS}
    \end{subfigure}
  
  \caption{(Cont.) Density estimation for two-dimensional synthetic datasets, including multi-modal and discontinuous densities. Flow transforms samples from a standard-normal base density to the target density. The top row shows the generative direction, visualizing the transformation from noise to data: $\mathbf{z} \sim p(\mathbf{z}) \rightarrow \mathbf{x} = f(\mathbf{z})$. The color scale represents the probability density function. Normalizing flows are reversible, so one can train on a density estimation task and still be able to sample from the learned density efficiently. The bottom row shows the normalizing direction from data to noise: $\mathbf{x} \sim p(\mathbf{x}) \rightarrow \mathbf{z} = f^{-1}(\mathbf{x})$}
  \label{fig:nf_2d_examples:2}
\end{center}
\end{figure}

\begin{figure}[!htb]\ContinuedFloat
  \begin{center}
    \begin{subfigure}{0.48\linewidth}
      \centering
      \includegraphics[width=\linewidth,trim=0cm 0cm 0cm 1.6cm, clip]{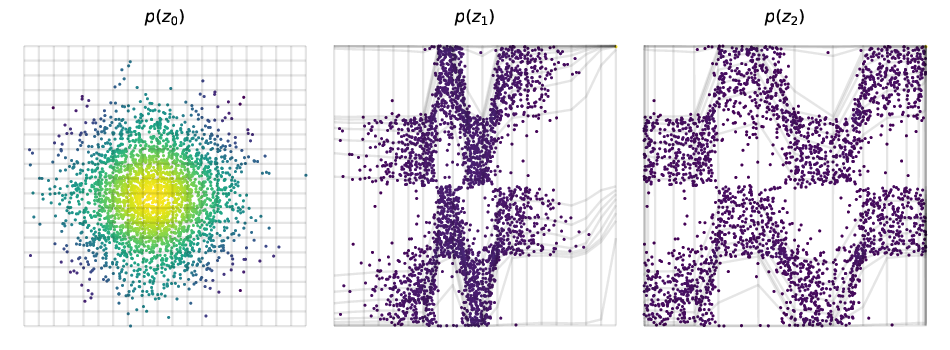}
      \includegraphics[width=\linewidth,trim=0cm 0cm 0cm 1.6cm, clip]{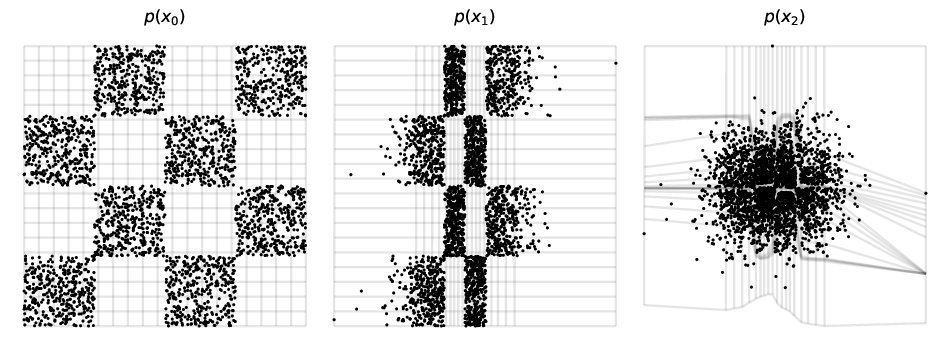}
      \caption{\textsc{checkerboard} dataset}
      \label{fig:NF_2D_CHECKERBOARD}
    \end{subfigure}
    \hfill{\color{lightgray}\vrule}\hfill
    \begin{subfigure}{0.48\linewidth}
      \centering
      \includegraphics[width=\linewidth,trim=0cm 0cm 0cm 1.6cm, clip]{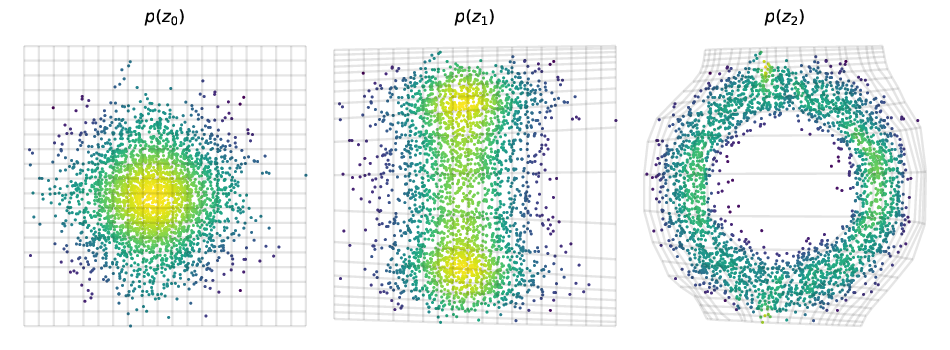}
      \includegraphics[width=\linewidth,trim=0cm 0cm 0cm 1.6cm, clip]{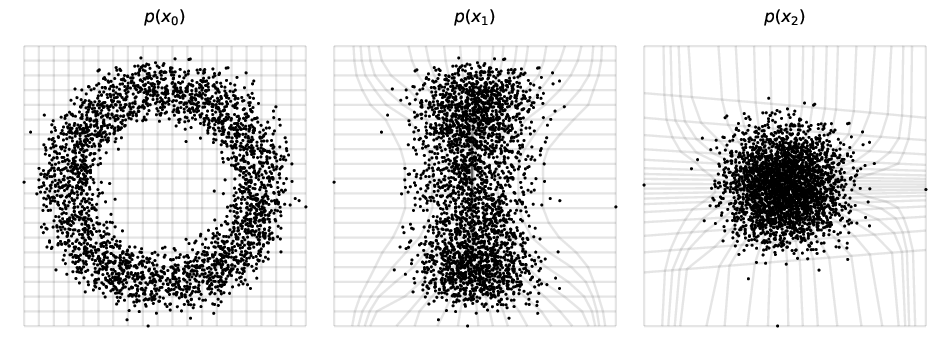}
      \caption{\textsc{circles} dataset}
      \label{fig:NF_2D_CIRCLES}
    \end{subfigure}
    \vspace{1em}
    \begin{subfigure}{0.48\linewidth}
      \centering
      \includegraphics[width=\linewidth,trim=0cm 0cm 0cm 1.6cm, clip]{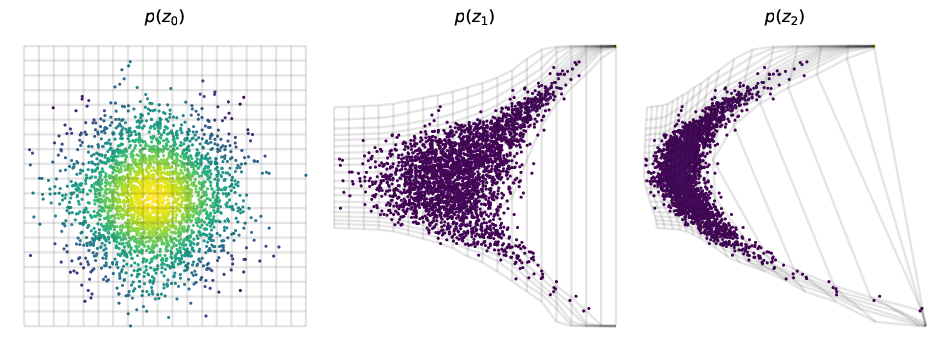}
      \includegraphics[width=\linewidth,trim=0cm 0cm 0cm 1.6cm, clip]{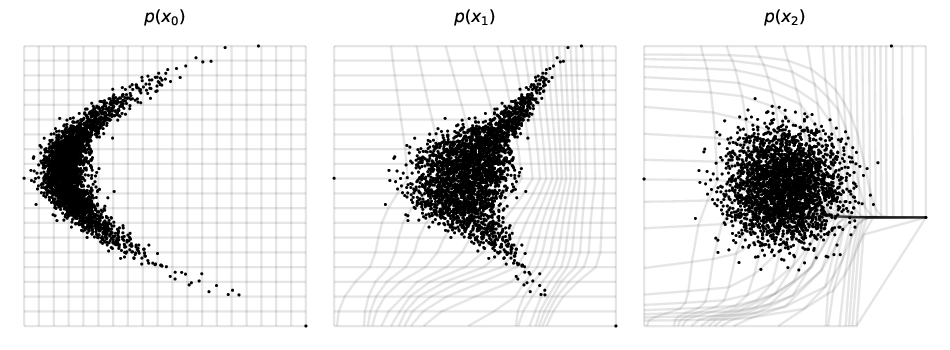}
      \caption{\textsc{crescent} dataset}
      \label{fig:NF_2D_CRESCENT}
    \end{subfigure}
    \hfill{\color{lightgray}\vrule}\hfill
    \begin{subfigure}{0.48\linewidth}
      \centering
      \includegraphics[width=\linewidth,trim=0cm 0cm 0cm 1.6cm, clip]{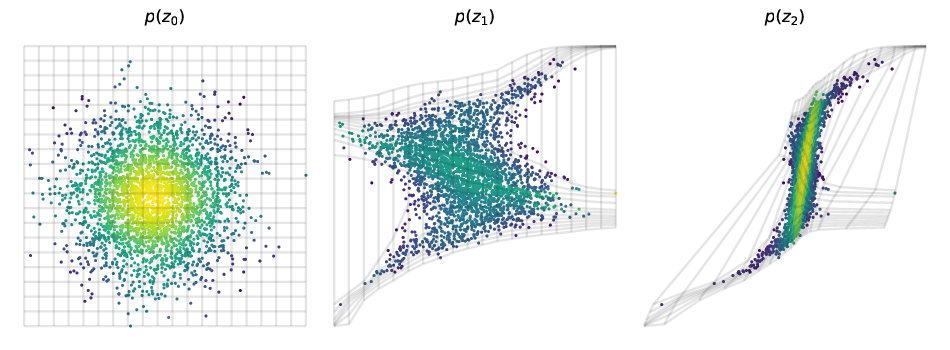}
      \includegraphics[width=\linewidth,trim=0cm 0cm 0cm 1.6cm, clip]{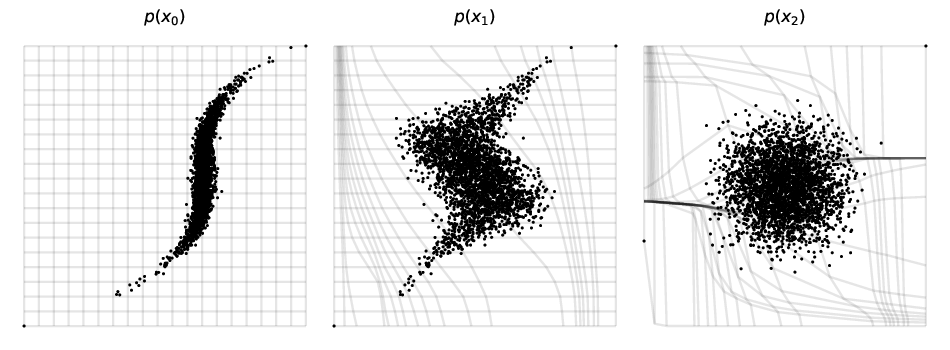}
      \caption{\textsc{crescentcubed} dataset}
      \label{fig:NF_2D_CRESCENTCUBED}
    \end{subfigure}
    \caption{(Cont.) Density estimation for two-dimensional synthetic datasets, including multi-modal and discontinuous densities. Flow transforms samples from a standard-normal base density to the target density. The top row shows the generative direction, visualizing the transformation from noise to data: $\mathbf{z} \sim p(\mathbf{z}) \rightarrow \mathbf{x} = f(\mathbf{z})$. The color scale represents the probability density function. Normalizing flows are reversible, so one can train on a density estimation task and still be able to sample from the learned density efficiently. The bottom row shows the normalizing direction from data to noise: $\mathbf{x} \sim p(\mathbf{x}) \rightarrow \mathbf{z} = f^{-1}(\mathbf{x})$}
    \label{fig:nf_2d_examples:3}
  \end{center}
\end{figure}

\cref{fig:nf_2d_examples:1,fig:nf_2d_examples:2,fig:nf_2d_examples:3} shows qualitative results for two-dimensional synthetic datasets. The comparison of the generated points with the original data shows that the normalizing flows can accurately replicate multi-modal and discontinuous distributions. DIFW-NF is able to achieve this with a limited number of flow layers and small neural network architecture, i.e., small number of hidden layers and neurons per layer. 
It is worth noting that the \textsc{twospirals} dataset, which has a complex and non-linear distribution, requires more than three flow steps to generate accurate results. Considering other models, comparisons are avoided in this instance because these datasets are small, simple and not representative of real-world data. Models that perform well on toy datasets may not necessarily perform well on larger, more complex datasets. Hence, the next section focuses on experiments on N-dimensional real-world data. 

\subsection{N-dimensional Data}

We evaluate our proposed flows using a selection of datasets from the UCI machine-learning
repository \cite{asuncion2007uci} and BSDS300 collection of natural images \cite{martin2001database}. This is the standard suite of benchmarks for density estimation of tabular data. The experimental setup and pre-processing of \cite{papamakarios2017masked} is followed, who make their data available online.

Model selection is performed using the standard validation splits for these datasets. The norm of gradients is clipped to the range $[-5, 5]$, and find this helps stabilize training. 
In this case a fully-connected neural network of $4$ layers with $64$ neurons per layer with ReLu activation function was chosen to compute the parameters of the element-wise transformations. A grid-search is not carried out to optimize the neural network architecture.

On the contrary, a grid-hyperparameter search is conducted for the transformation of each dataset. Flows can be composed after $3$,$5$,$8$,$10$,$15$ or $20$ steps and the transformation tessellation size is chosen among $\{5,10,20,50\}$.
In addition, a grid-search over network architectures was also performed; we searched over models with $1$,$2$ or $4$ layers with $8$ or $16$ hidden layers per flow.
The Adam optimizer \cite{kingma2014adam} was used with default hyperparameters and an initial learning rate of $3e^{-4}$ or $5e^{-4}$ over $128$, $256$ or $512$ training epochs with batch size $512$. 

Hyperparameter settings are shown for coupling flows in \cref{tab:nf_nd}. The dimensionality and number of training data points are included in each table for reference. 

\begin{table}[!htb]
  \small
  \caption{Hyperparameters and log-probability for density-estimation results in ND-datasets}
  \label{tab:nf_nd}
  \begin{center}
  \begin{tabular}{llllllll}
    \toprule
    Dataset &  \textsc{bsds}\footnotesize300 & \textsc{gas} & \textsc{hepmass} & \textsc{miniboone} & \textsc{power} \\
    \midrule
    Train Points         &  1000000 &  852174 &  315123 &     29556 &  1615917  \\
    Dimension            &       63 &       8 &      21 &        43 &        6  \\
    \midrule
    Batch Size           &      512 &     512 &     512 &       512 &      512  \\
    \# Neurons per Layer &       64 &      64 &      64 &        64 &       64  \\
    \# Hidden Layers     &        4 &       4 &       4 &         4 &        4  \\
    Tessellation Size    &       10 &      10 &      10 &        10 &       50  \\
    Flow Steps           &       15 &      15 &      10 &        10 &        3  \\
    Epochs               &      256 &     256 &     256 &       512 &      128  \\
    Learning Rate        &   0.0005 &  0.0005 &  0.0005 &    0.0003 &   0.0005  \\
    
    \bottomrule
  \end{tabular}
\end{center}
\end{table}

\clearpage

\begin{table}[!htb]
  \small
  \caption{Test log-likelihood (in nats) for UCI datasets and BSDS300, with error bars corresponding to two standard deviations. Higher is better.}
  \label{tab:nf_nd_results}
  \begin{center}
  \begin{tabular}{llllll}
    \toprule
    Dataset &  \textsc{bsds300} & \textsc{gas} & \textsc{hepmass} & \textsc{miniboone} & \textsc{power} \\
    \midrule
    DIFW (AR)    &  155.95 $\pm$ 0.39 &  11.60 $\pm$ 0.02 &  -14.18 $\pm$ 0.04 &   -9.06 $\pm$ 0.07 &  0.45 $\pm$ 0.01 \\
    DIFW (CL)    &  \textbf{160.25} $\pm$ 0.40 &  10.79 $\pm$ 0.03 &  -15.32 $\pm$ 0.03 &   \textbf{-8.02} $\pm$ 0.06 &  0.30 $\pm$ 0.01 \\
    BLOCK-NAF    &  157.36 $\pm$ 0.03 &  12.06 $\pm$ 0.09 &  -14.71 $\pm$ 0.38 &   -8.95 $\pm$ 0.07 &  0.61 $\pm$ 0.01 \\
    FFJORD       &  157.40 $\pm$ 0.19 &   8.59 $\pm$ 0.12 &  -14.92 $\pm$ 0.08 &  -10.43 $\pm$ 0.04 &  0.46 $\pm$ 0.01 \\
    GLOW         &  156.95 $\pm$ 0.28 &  12.24 $\pm$ 0.03 &  -16.99 $\pm$ 0.02 &  -10.55 $\pm$ 0.45 &  0.52 $\pm$ 0.01 \\
    MAF          &  156.95 $\pm$ 0.28 &  12.35 $\pm$ 0.02 &  -17.03 $\pm$ 0.02 &  -10.92 $\pm$ 0.46 &  0.45 $\pm$ 0.01 \\
    NAF          &  157.73 $\pm$ 0.04 &  11.96 $\pm$ 0.33 &  -15.09 $\pm$ 0.40 &   -8.86 $\pm$ 0.15 &  0.62 $\pm$ 0.01 \\
    Q-NSF (AR)   &  157.42 $\pm$ 0.28 &  12.91 $\pm$ 0.02 &  -14.67 $\pm$ 0.03 &   -9.72 $\pm$ 0.47 &  \textbf{0.66} $\pm$ 0.01 \\
    Q-NSF (C)    &  157.65 $\pm$ 0.28 &  12.80 $\pm$ 0.02 &  -15.35 $\pm$ 0.02 &   -9.35 $\pm$ 0.44 &  0.64 $\pm$ 0.01 \\
    RQ-NSF (AR)  &  157.31 $\pm$ 0.28 &  \textbf{13.09} $\pm$ 0.02 &  \textbf{-14.01} $\pm$ 0.03 &   -9.22 $\pm$ 0.48 &  \textbf{0.66} $\pm$ 0.01 \\
    RQ-NSF (C)   &  157.54 $\pm$ 0.28 &  \textbf{13.09} $\pm$ 0.02 &  -14.75 $\pm$ 0.03 &   -9.67 $\pm$ 0.47 &  0.64 $\pm$ 0.01 \\
    SOS          &  157.48 $\pm$ 0.41 &  11.99 $\pm$ 0.41 &  -15.15 $\pm$ 0.10 &   -8.90 $\pm$ 0.11 &  0.60 $\pm$ 0.01 \\\bottomrule
  \end{tabular}
\end{center}
\end{table}
\footnotetext{The average log-likelihood (relative entropy) is usually reported in units of nats. When one uses the natural logarithm to compute entropy, it takes on the "natural units of information", or nats.}

\begin{figure}[!htb]
  \begin{center}
  \includegraphics[width=\linewidth]{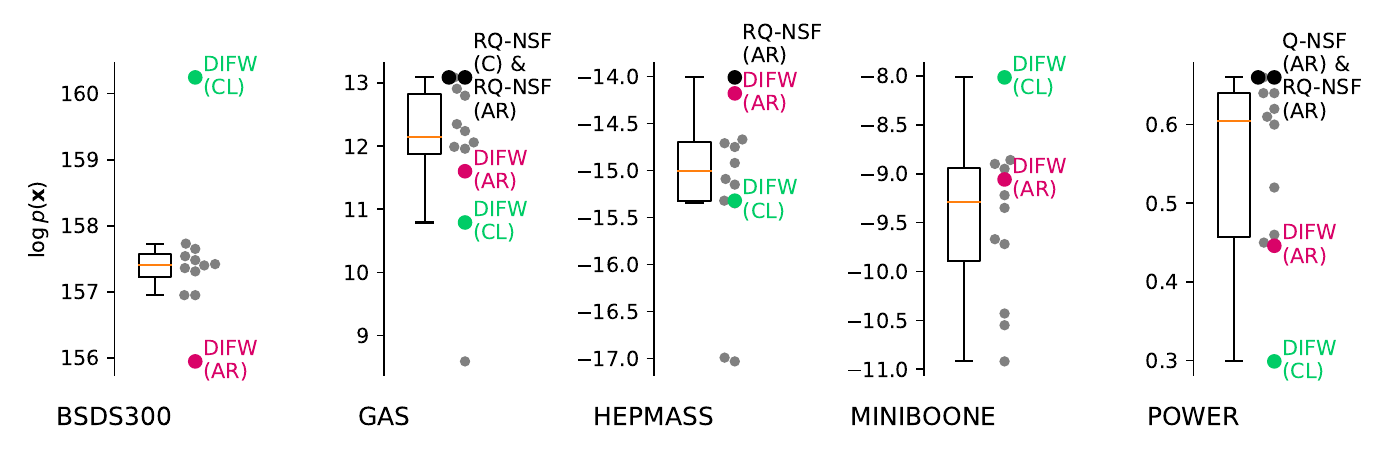}
  \caption{Boxplot visualization. Each point represents the test log-likelihood (in nats) of each model for UCI datasets and BSDS300, higher is better.}
  \label{fig:nf_nd_results}
  \end{center}
\end{figure}

Our results are shown in \cref{tab:nf_nd_results,fig:nf_nd_results}.
DIFW (CL) achieve state-of-the-art results for \textsc{bsds300} and \textsc{miniboone}, while Both RQ-NSF (C) and RQ-NSF (AR) do so on the \textsc{power}, \textsc{gas}, and \textsc{hepmass} datasets, tied with Q-NSF (AR) on the \textsc{power} dataset. Moreover, DIFW (CL) and DIFW (AR) achieve competitive scores with the best autoregressive models (Glow and FFJORD). These results close the gap between autoregressive flows and flows based on coupling layers, and demonstrate that, in some cases, it may not be necessary to sacrifice one-pass sampling for density-estimation performance.

\section{Conclusions}

On the whole, DIFW-NF shows that upgrading the commonly-used affine transformations in coupling and autoregressive layers can significantly improve performance without compromising analytic invertibility. Through the use of monotonic transforms based on the integration of continuous piecewise-affine velocity functions, coupling-layer-based models can achieve density-estimation performance on par with the best autoregressive flows while maintaining exact one-pass sampling. These models find a unique compromise between versatility and flexibility, serving as a powerful off-the-shelf tool for enhancing architectures like the variational autoencoder and boosting parameter efficiency in generative modeling.

The proposed transforms scale to high-dimensional problems, as demonstrated empirically in \cref{sec:results_6}.
Moreover, due to the increased flexibility of the proposed transformations, it requires fewer steps to build flexible flows, reducing the computational cost. 
A potential drawback of the proposed method is a more involved implementation. This is alleviated by providing an extensive \cref{sec:method_6} with technical details, and a reference implementation in PyTorch.

The proposed transformations are also a useful differentiable and invertible module in their own right, which could be included in many models that can be trained end-to-end. For instance, monotonic warping functions with a tractable Jacobian determinant are useful for supervised learning. More generally, invertibility can be useful for training very large networks, since activations can be recomputed on-the-fly for backpropagation, meaning gradient computation requires memory which is constant instead of linear in the depth of the network.
Monotonic transforms based on the integration of continuous piecewise-affine velocity functions are one way of constructing invertible element-wise transformations, but there may be others. The benefits of research in this direction are clear, and so we look forward to future work in this area.

\graphicspath{{content/chapter7/}}

\hypertarget{chapter7}{
\chapter{Conclusions and Future Research}
}\label{chapter:7}
\begingroup
\hypersetup{linkcolor=black}
\endgroup

The purpose of this chapter is to provide a condensed overview of the presented results and deliberate on the principal conclusions derived from this thesis.

\section{Summary of Contributions}
The recent prevalence of time-based data across many disciplines has called for effective and efficient methods to analyze large and complex time series datasets. Addressing this demand, this thesis has harnessed the power of fast, efficient, parametric \& diffeomorphic warping techniques to enhance time-series tasks such as alignment, averaging, classification and clustering.

\subsubsection*{Efficient diffeomorphic transformations suitable for deep learning}

One-dimensional diffeomorphic transformations, which are smooth invertible functions with  and a differentiable inverse, have a variety of applications across different fields: shape and texture mapping for computer graphics, transfer functions for image tone mapping, coupling functions for normalizing flows, and warping functions for time series alignment are just a few examples.
In the context of time series alignment, for instance, warping functions that are continuous, differentiable and invertible are preferred, as they must be strictly monotone with a strictly positive first derivative.

With this in mind, \textbf{fast and efficient one-dimensional diffeomorphic transformations} were proposed in \cref{chapter:2}. Such differentiable, invertible and with differentiable inverse functions were obtained via the integration of an ordinary differential equation (ODE) defined by a continuous piecewise affine (CPA) velocity function. CPA velocity functions yield well-behaved parametrized diffeomorphic transformations, which are efficient to compute, accurate and highly expressive. In addition, these finite-dimensional parametric transformations can handle optional constraints and support convenient modeling choices such as smoothing priors and coarse-to-fine analysis. 
As proved by \cite{Freifeld2015}, this ODE integration can be obtained in closed-form for the one-dimensional case. However, the gradient of CPA-based transformations is only available via the solution of a system of coupled integral equations. In this thesis, we  derived a novel \textbf{closed-form solution for the gradient} of CPA-based diffeomorphic transformations, providing efficiency, speed and precision.  

In this regard, an open-source library called \textbf{Diffeomorphic Fast Warping (DIFW)}, was released with a highly optimized implementation of 1D diffeomorphic transformations on multiple backends for CPU (\textit{NumPy} and \textit{PyTorch} with \textit{C++}) and GPU (\textit{PyTorch} with \textit{CUDA}). 
The experimental tests resulted in \textit{x18} and \textit{x10} speed improvement on CPU over \textit{libcpab} for forward and backward operations respectively. On GPU the performance gain of \textit{DIFW} reached \textit{x260} and \textit{x30} respectively.

\subsubsection*{Differentiable and efficient similarity measure under time warping}

When working with time series data, a key challenge is determining how similar they are to one another. 
In fact, as was illustrated throughout the dissertation, time series data often appears misaligned due to differences in execution, sampling rates, or the number of measurements, despite having amplitude and shape similarities. These and other factors have a major effect on the design of similarity metrics.
With time warping, insignificant differences in the time axis may appear as very significant differences in the ordinate axis. As a result, rigid similarity metrics such as the Euclidean distance are not the optimal choice.

Elastic metrics such as dynamic time warping (DTW), which were thoroughly reviewed in \cref{chapter:1}, offer a promising approach, but are limited by their computational complexity, non-differentiability and sensitivity to noise and outliers. 
Considering this, in \cref{chapter:3} we proposed \textbf{novel elastic alignment methods} that use parametric \& diffeomorphic warping transformations as a means of overcoming the shortcomings of DTW-based metrics.
The proposed method is differentiable \& invertible, well-suited for deep learning architectures, robust to noise and outliers, computationally efficient, and is expressive and flexible enough to capture complex patterns.

\subsubsection*{Warping invariant and generalizable time series averaging}

Obtaining the elastic average of a set of time series requires a simultaneous alignment that gets rid of undesired warping, thereby minimizing set's temporal variability. However, finding optimal and plausible warping functions requires solving a complex optimization problem. 
In this thesis, we deployed the power of \textbf{gradient descent methods and backpropagation} to minimize elastic averaging's loss function.
Closed-form diffeomorphic transformations were successfully incorporated into a \textbf{temporal transformer network (TTN)} resembling \cite{Weber2019} that can both align pairwise sequential data and learn representative average sequences for multiclass joint alignment.
In addition, we added regularization to the alignment loss function to address the warping function's sensitivity to noise and outliers.  

Given the lack of ground truth for the latent warps, the nearest centroid classification model was used as a proxy metric for the quality of the joint alignment and the average signal. 
Extensive experiments were conducted on 84 univariate and 30 multivariate datasets from the UCR archive \cite{dau2019ucr}, to validate the generalization ability of our model to unseen data for time series joint alignment. 
Results showed significant \textbf{improvements both in terms of efficiency and accuracy}, and the proposed TTN model using \textit{DIFW} beat all 5 comparing methods (Euclidean, DBA, SoftDTW, DTAN, ResNet-TW). 
One key factor for such improvement was the closed-form expression for the gradient of the warping transformation; its precise computation translated to an efficient search in the parameter space, which led to faster and better solutions at convergence.
Furthermore, the model was able to generalize \textbf{inferred alignments from the original batch} to the new data without having to solve a new optimization problem each time.

\subsubsection*{Application of elastic time series classification to rock drill pressure data}

\cref{chapter:4} presented an application of the methodological contributions discussed in \cref{chapter:2,chapter:3} to time series classification tasks.
The case study focused on the \textbf{fault diagnosis of a hydraulic rock drill} using pressure data, and its performance was evaluated in the 2022 PHM Data Challenge. 
Experimental results showed that the proposed deep learning-based solution achieved a noteworthy accuracy of 97.80\%. The success of this approach can be attributed to the \textbf{importance of time series alignment for classification} tasks of such nature. 
As a matter of fact, hydraulic rock drill machines operate under severe performance demands in harsh environments, which generate vibrations that affect the behavior of the pressure signals: the oscillations resulting from pressure propagation at high frequencies can lead to significant alterations of the pressure signals' dynamics.
Furthermore, it is worth noting that the model was able to \textbf{learn and generalize the warping functions} present in the data to new samples, as evidenced by the fact that the variance reduction observed in the training set also exists in the test and validation sets.

\subsubsection*{Scalable shape-based time series clustering}

\cref{chapter:5} focused on warping-invariant time series clustering techniques that can operate under limited computational and time resources. Clustering methods can summarize data into a set of small, meaningful, and manageable representatives.
However, traditional clustering methods must contend with several challenges when applied to time series, such as the specificity of the time dimension, its high dimensionality or the underlying dynamic and non-stationary processes.

In response to these challenges, an \textbf{incremental time series clustering algorithm} called DIFW-IC was presented in \cref{chapter:5}. The proposed model assigns each incoming time series into the nearest cluster using a combination of elastic alignment and incremental clustering techniques, and it is robust to outliers and concept drift.  
The algorithm is warmed-up through an offline process, and in the online phase, new time series instances are added incrementally to existing clusters. The assignment decision is based on the \textbf{diffeomorphic elastic distance} of the new point to the existing clusters. 
When the query time series does not match with any of the existing clusters, it is allocated to a new temporary group that can be updated with more incoming data. 
The performance of DIFW-IC was evaluated on several benchmark datasets and compared it with state-of-the-art clustering methods for time series data. Results showed that overall the proposed algorithm outperformed existing ones in terms of \textbf{clustering quality and scalability}.

\subsubsection*{Efficient and flexible coupling functions for normalizing flows}

\cref{chapter:6} dealt with normalizing flows, a type of generative model that converts a simple probability distribution into a more complex one by applying a sequence of invertible and differentiable mappings. 
Related flows based on coupling layers such as NICE and RealNVP have an analytic one-pass inverse, but are less flexible than their autoregressive counterparts. 
Based on these limitations, we proposed to implement the coupling function using the integration of continuous piecewise-affine (CPA) velocity functions as a building block.

The closed-form diffeomorphic transformations presented in \cref{chapter:2} were exploited to act as a \textbf{drop-in replacement for the affine or additive transformations} commonly found in coupling and autoregressive flows, significantly enhancing their flexibility. 
When combined with alternating invertible linear transformations, the resulting class of normalizing flows was referred to as \textbf{closed-form diffeomorphic spline flows} (DIFW-NF), which may feature coupling or autoregressive layers.
DIFW-NF resembles a traditional feed-forward neural network architecture, alternating between linear transformations and element-wise non-linearities, while retaining an exact, analytic inverse.

Unlike the additive and affine transformations, the proposed differentiable monotonic transformation with sufficiently many intervals can approximate any differentiable monotonic function, yet has a closed-form, tractable Jacobian determinant, and can be inverted analytically. Our parameterization is fully-differentiable, which allows for training by gradient methods.
Experiments demonstrated that this module significantly \textbf{enhances the flexibility of coupling \& autoregressive flows}, obtaining competitive results in a variety of high-dimensional datasets.

\section{Future Research}

Building on the insights gained in this work, several promising avenues for future research could be pursued. 

\vspace{-1.0em}
\paragraph{Joint Alignment using Batches}
Due to the training data size, deep learning optimization is typically done batch by batch, where each batch consists of a subset (selected at random) of the time series from the entire set. A single epoch then represents a full pass over the entire data, and the time series are reshuffled between epochs. This typically-necessary batch-by-batch processing creates an optimization difficulty which might appear to be minor but is, in fact, far more critical than it may seem. 

The issue is that the average time series is a function of the entire set, not just the samples in the current batch. A seemingly-obvious solution is to hold the average fixed during each epoch, so it does not affect the computation of the loss' gradient, and then, at the end of each epoch, recompute the average. However, a problem that arises with this approach is that the difference between the alignment targets (that is, the previous average and the recomputed one) in each pair of consecutive epochs might be large, making the optimization difficult since the optimal transformations for one target might be quite far from those that are optimal for the next target. Hence, additional research is necessary to mitigate this jumping-target problem that complicates the joint-alignment optimization problem.

\vspace{-1.0em}
\paragraph{Regularization-free Strategy for Joint Alignment}
Reaching a global optima when attempting to minimize a loss is challenging, if not impossible; if this is accomplished, it is considered to be the pinnacle of success. Despite how (relatively) simple it is to reach, a global minimum of joint alignment actually represents a catastrophic failure; e.g., all the time series are shrunk to an infinitesimally-small point. 

A popular solution in such cases is adding some type of regularization over the transformation parameters $\theta$. However, while various forms of regularization have been suggested, each of them imposes a certain bias.
Likewise, penalizing some norm of the transformation parameters $\theta$ is problematic when the accumulative warping is large, while regularization favoring temporal smoothness is not always compatible with other domains. Another issue is the need of hyperparameter tuning for the weight of the regularization term. Therefore, there remains a significant gap to find a combination of a regularization type and a weight that works well for a sufficiently-large variety of datasets.

\vspace{-1.0em}
\paragraph{Extension to Cyclical Data}
The proposed Temporal Transformer Network (TTN) for time series alignment does not take into account the cyclical nature of some time series data. To address this limitation, an extension could be developed to handle cyclical data, such as electrocardiogram (ECG) data. This would involve modifying the TTN to allow for cyclic behavior in the latent trace. Specifically, an observed trace generated by the model would be allowed to cycle back to the start of the latent trace, and the smoothness constraints would apply to the beginning and end of the traces, coercing them to be similar.

\vspace{-1.0em}
\paragraph{Learning a Notion of Similarity}
Learning from time series data, as illustrated in \cref{chapter:1}, can take many forms depending on the invariants present in the data. 
In case these invariants are known, dedicated time series classification methods can be used, but if there is limited expert knowledge, or it is difficult to choose a suitable learning approach, ensemble techniques that cover various similarity notions such as \cite{bagnall2015time, lines2018time} are typically used. This, however, comes at the expense of increased complexity. In this thesis, we advocated for the explicit removal of unwanted nonlinear warping, but more principled approaches could be designed to learn the notion of similarity from the data.

\vspace{-1.0em}
\paragraph{Hybrid Distance Measures}
Combining the proposed elastic alignment methods with traditional distance measures such as Euclidean distance or cosine similarity could provide more robust similarity metrics. 
More work is needed to investigate the optimal ways to integrate these techniques effectively. Future research could explore different weighting schemes to determine the best way to balance the contribution of each measure.

\vspace{-1.0em}
\paragraph{Handling missing data \& uncertainty}
Time series data often contains missing values, which can make alignment and clustering challenging. Future research could explore methods for handling missing data in the proposed framework, such as imputation techniques or incorporating probabilistic models to handle uncertainty.
In this sense, while the proposed framework in this thesis is robust to noise and outliers, it does not explicitly model uncertainty. Investigating ways to incorporate uncertainty into time series analysis could lead to more accurate predictions and improve the reliability of these time series models.



\begin{appendices} 
    
\graphicspath{{content/appendix1/}}

\chapter{Fast Approximate Exponential Function}\label{apx:exponential_function}
\begingroup
\hypersetup{linkcolor=black}
\setstretch{1.0}
\minitoc
\endgroup

\section{Introduction}\label{sec:introduction_apx}

\begin{figure}[!htb]
  \begin{center}
  \includegraphics[width=0.7\linewidth]{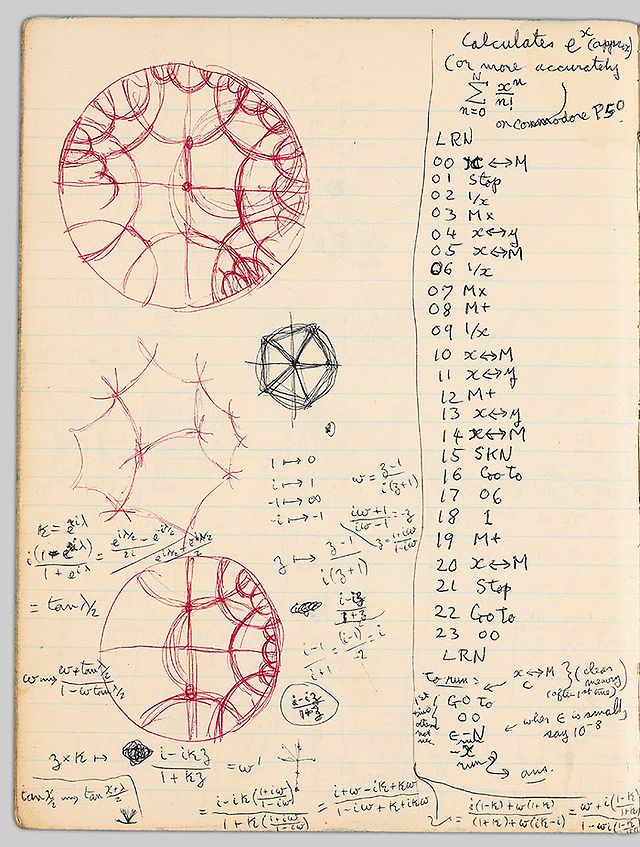}
  \caption{Roger Penrose's journal: how to approximate $e^x$ on a Commodore P50 calculator.}
  \label{fig:penrose}
  \end{center}
\end{figure}

The exponential function, denoted by the symbol 
\color{Blue} $e^{x}$
\color{Black} , is the function where the 
\color{Red} rate of change
\color{Green} is always 
\color{Black} the
\color{Blue} current value\color{Black}, and is one of the most fundamental and widely used functions in mathematics. One of its key properties is that it grows at an exponential rate, which makes it particularly useful for modeling processes that involve exponential growth or decay. 
\begin{equation}
  \color{Red} \cfrac{d \color{Blue} e^{x}}{\color{Red} dx}
  \color{Green} =
  \color{Blue} e^{x}\color{Black}
\end{equation}
In neural computation, for instance, exponentiation is required to compute most of the activation functions and probability distributions used in neural network models. Consequently, much of the time in neural simulations is actually spent on exponentiation. Furthermore, a fast computation of the exponential function is essential for CPA-based diffeomorphic transformations presented in \cref{chapter:2}. The exponential is embedded in the closed-form solution, and must be called several times during the forward and backward processes.
The \textit{exp} functions provided by popular computer math libraries are highly accurate but rather slow. An approximation is adequate for most computation purposes and can save considerable time\footnote{Even Sir Roger Penrose derived a method to compute $e^x$ in a Commodore P50 calculator (see \cref{fig:penrose})}.
Several methods exist in the literature to compute the exponential function exactly or approximately, and these techniques are reviewed and compared in this appendix. The solution that yields the best trade-off between computation time and error was selected to be implemented in the package Diffeomorphic Fast Warping (DIFW)\footnote{\url{https://github.com/imartinezl/difw}}.

\section{Background}\label{sec:background_apx}

\subsection{Taylor Series}
Let $f$ be a smooth function (infinitely differentiable) at some real value $a$. For every $k^{th}$ derivative $f^{(k)}$, we can differentiate $f^{(k)}$ again to obtain the $(k+1)^th$ derivative $f^{(k+1)}$. Any function with such property is called analytic and can be uniquely represented as a Taylor series expansion centered at $a$. The Taylor series of $f$ centered at $a$ and evaluated at $x$ is expressed as:
\begin{equation}
f(x) = \frac{f(a)}{0!}(x-a)^{0} + \frac{f'(a)}{1!}(x-a)^{1} + \cdots + \frac{f^{(k)}(a)}{k!}(x-a)^{k} + \cdots
\end{equation}
A common center of expansion is $a=0$, also called the MacLaurin series. Truncating the Taylor expansion of a function $f$ at any term $k$ gives a finite approximation of $f$ using the $k$ degree Taylor polynomial. A Taylor polynomial of $f$ centered at $a$ produces very accurate approximations of $f(x)$ when is relatively close to $a$. As the absolute value of $x$ increases away from $a$, the accuracy of the Taylor polynomial rapidly decreases, which means it requires more terms of the Taylor series (i.e. a higher degree polynomial) for accurate approximation.

\subsection{Number $e$ and Function $e^{x}$}
The exponential function can be defined as the function which is its own derivative and maps the additive identity 0 to the multiplicative identity 1. 
$f(x)=e^x$ uniquely satisfies this property. We can show this, and define $e$ directly in the process, by starting from the Taylor series representation of an arbitrary function $f$ infinitely differentiable at $a=0$. Suppose $a_0,\,a_1,\,\cdots$ are the coefficients of the Taylor series of $f$ centered at $a$. Then, the Taylor series and its derivative can be expressed as:
\begin{equation}
f(x) = a_{0} + a_{1}x + a_{2}x^2 + a_{3}x^3 + \cdots
\end{equation}
\begin{equation}
f'(x) = a_{1} + 2a_{2}x + 3a_{3}x^2 + 4a_{4}x^3 + \cdots
\end{equation}
To determine a function which is its own derivative, we solve for the coefficients $a_0,\,a_1,\,\cdots$, which satisfy $f=f'$:
\begin{equation}
a_{0} + a_{1}x + a_{2}x^2 + a_{3}x^3 + \cdots = a_{1} + 2a_{2}x + 3a_{3}x^2 + 4a_{4}x^3
\end{equation}
From here we can see the pattern $a_{k} = \cfrac{a_{k-1}}{k}$. 
Given $a_{0}=1$, we find that the Taylor series of a function which is its own derivative is
\begin{equation}
  f(x) = 1 + x + \frac{x^2}{2!} + \frac{x^3}{3!} + \cdots = \sum_{k=0}^{\infty} \frac{x^{k}}{k!}=e^x
\end{equation}
We denote this function with $e^x$, and $e$ to the value of this function at $x=1$.
\begin{equation}
e = f(1) = \sum_{k=0}^{\infty} \frac{1}{k!}
\end{equation}
An exponential function with a base less than $e$ grows more quickly than its derivative, but when the base is greater than $e$, it grows less quickly than its derivative.

\subsection{Floating-point Arithmetic}
The IEEE 754 floating point standard discretizes real intervals into a computable form by mapping all nearby real values in given neighborhoods to a single rounded value. A number $x$ in IEEE 754 binary floating point format is represented in the form $x = (-1)^{S}(1 + M) \cdot 2^{E}$ where $S$ is the sign bit, $M$ is the mantissa (a binary fraction in the range $[0,1)$ and $E$ is the exponent. 

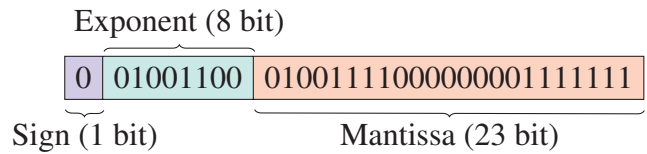
\begin{figure}[!htb]
  \begin{center}
\begin{tikzpicture}[array/.style={rectangle split,rectangle split horizontal, rectangle split parts=#1,draw, anchor=center, rectangle split part fill={Violet!20, Emerald!20, Red!20}}]
  \node[array=3] (a) {
  \nodepart{one}0
  \nodepart{two}01001100
  \nodepart{three}01001111000000001111111
  };
  
  \draw [decoration={brace,raise=4pt},decorate] ($(a.two) + (-0.13,+0.35)$) --  node[above=5pt]{Exponent (8 bit)}($(a.three) + (-0.13,+0.35)$);
  \draw [decoration={brace,mirror,raise=4pt},decorate] ($(a.three) + (-0.13,-0.14)$) --  node[below=5pt]{Mantissa (23 bit)}(a.south east);
  \draw [decoration={brace,mirror,raise=4pt},decorate] ($(a.one) + (-0.13,-0.14)$) --  node[below=5pt]{Sign (1 bit)}($(a.two) + (-0.13,-0.15)$);
\end{tikzpicture}
\caption{IEEE 754 floating point standard, single precision.}
\label{fig:ieee754}
\end{center}
\end{figure}

IEEE 754 \textbf{single precision} floating point numbers have a total size of 32 bits, with 23 bit mantissa (24 accounting for the normalized bit) and an 8 bit exponent $E \in [-126, 127]$. Thus, one can represent $2^{32}$ different values in single precision floating point, with underflow and overflow limits of $2^{-126} \simeq 1.2\times10^{-38}$ and $2^{127} \simeq 3.4\times10^{38}$ respectively.

Likewise, IEEE 754 \textbf{double precision} floating point numbers have a total size of 64 bits, with 52 bit mantissa (53 accounting for the normalized bit) and an 11 bit exponent $E \in [-1022, 1024]$. Thus, one can represent $2^{64}$ different values in double precision floating point, with underflow and overflow limits of $2^{-1022} \simeq 2.2\times10^{-308}$ and $2^{1024} \simeq 1.8\times10^{308}$ respectively. See \cref{tab:floating} for a comparison between single and double precision floating point numbers.

\begin{table}[!htb]
\caption{Single vs double precision floating point numbers.}
\label{tab:floating}
\begin{center}
\resizebox{0.7\linewidth}{!}{%
\begin{tabular}{lcc}
\toprule
\textbf{Characteristics} & \textbf{Single Precision} & \textbf{Double Precision} \\
\midrule
Total size & 32 bits & 64 bits \\
Sign size & 1 bit & 1 bit \\
Mantissa size  & 23 bits & 52 bits \\
Exponent size & 8 bits & 11 bits \\
Exponent range & $[-126, 127]$ & $[-1022, 1024]$ \\
Machine epsilon & $2^{-23}$ & $2^{-52}$ \\
Decimal precision & 7 bits & 16 bits \\
Underflow limit & $2^{-126} \simeq 1.2\times10^{-38}$ & $2^{-1022} \simeq 2.2\times10^{-308}$ \\
Overflow limit & $2^{127} \simeq 3.4\times10^{38}$ & $2^{1024} \simeq 1.8\times10^{308}$ \\
\bottomrule
\end{tabular}%
}
\end{center}
\end{table}

For any binary floating point system, we can derive the maximum precision available in that system using the number of bits available in the mantissa. Given the uneven distribution of floating point values, it follows that the available precision decreases as we move away from the origin. For this reason we define machine epsilon $\epsilon_M$, which is the difference between $1$ and the least representable floating point value greater than $1$.
The smallest distinguishable value larger than $1$ in single precision floating point is $2^{-23}$ and in double precision floating point is $2^{-52}$. The maximum decimal precision of these systems can be obtained by converting $\epsilon_M$ to a decimal value.

Floating point values are not evenly distributed along the real numbers: values are relatively densely clustered near 0 and increasingly sparse the further one moves away from the origin. More generally, in each interval $[2^{n}, 2^{n+1}]$, the available floating point values are distributed with a spacing of $2^{n-p}$ between them, where $p$ is the number of mantissa bits for the precision under consideration. As one moves farther away from the origin, it becomes more likely that your calculations will bump into real values in between the available floating point values, which will be rounded to the nearest available value instead.

A key idea behind the fast exponentiation macro is that any integer written directly to the \textit{exponent component} will be exponentiated when the same memory location is read back in floating-point format.

\subsection{Numerical Errors}

Numerical errors refer to inaccuracies that occur when performing mathematical operations on a computer.
Absolute and relative error provide a way to rigorously quantify the accuracy of an approximation. If we can show an upper bound on one of these error metrics, we can assert the worst-case accuracy of the approximation. 
Given a function $f$ and its approximation $F$, the absolute error at some $x$ is given by $\epsilon(x) = |f(x) - F(x)|$ and the relative error at $x$ is given by
$\mu(x) = {\epsilon(x)}/{|f(x)|}$. 
Relative error depends on, and normalizes, absolute error.

Numerical analysis traditionally considers four main categories of error: fundamental, convergence, truncation and floating-point errors.

\paragraph{1. Fundamental Error} This arises when a model intrinsically deviates from reality in a nontrivial way. It can happen when the model's assumptions or simplifications do not accurately capture the complexity or behavior of the system.

\begin{wrapfigure}{r}{0.3\textwidth}
  \vspace{-0.5cm}
  \footnotesize
  \setstretch{0.8}
  \begin{verbatim}
  x = 10 / 9
  for k in range(20):
    x -= 1
    x *= 10
    print(x)
  >
  1.1111111111111116
  1.1111111111111160
  1.1111111111111605
  1.1111111111116045
  1.1111111111160454
  1.1111111111604544
  1.1111111116045436
  1.1111111160454357
  1.1111111604543567
  1.1111116045435665
  1.1111160454356650
  1.1111604543566500
  1.1116045435665000
  1.1160454356650007
  1.1604543566500070
  1.6045435665000696
  6.0454356650006960
  50.454356650006960
  494.54356650006960
  4935.4356650006960
  \end{verbatim}
  \captionsetup{width=0.7\linewidth}
  \caption{Convergence error}\label{fig:convergence_error}
  \vspace{-1cm}
  \end{wrapfigure}

\paragraph{2. Convergence Error} This refers to the gradual loss of significance, as repeated iterations compound the initial error until it passes a precision threshold. Convergence error is particularly dangerous. Consider a very simple iterative calculation in \textit{Python} (\cref{fig:convergence_error}).

Each result has 16 decimal digits total, but the precision of the result decreases over time. Note that with a starting value of $x=10/9$, after one iteration we will have $f(x) = (x-1)\cdot10 = (\frac{10}{9} - 1)\cdot10 = \frac{1}{9} = \frac{10}{9}$. Therefore, the value of should remain the same after every iteration. But instead we have a value which is rapidly diverging away from the true value with each iteration.

\paragraph{3. Discretization/Truncation Error} It occurs whenever one takes a continuous function and approximates it by finitely many values. It also occurs when approximating functions using their Taylor series. In this case, the infinitely many remaining terms of the Taylor series sliced off after truncation comprise the discretization error in your calculation.

\paragraph{4. Floating Point Error} One cannot achieve more accuracy than is provided for by the precision of the floating point system. For instance, any real number with more than 16 decimal digits simply cannot be expressed to perfect accuracy in double precision. Floating point error also occurs if, in the course of your calculations, an intermediate result is outside the underflow/overflow limits, even if the completion of the calculation would bring it back into the range of available values. We include two examples of floating point errors: one in which the order of the operations matter (\cref{eq:floating_error_1}), and another in which floating point numbers are not associative under addition (\cref{eq:floating_error_2}).

\begin{equation}\label{eq:floating_error_1}
1 + \frac{1}{2}\epsilon_M - 1 = 0 \neq 1.11022 \times 10^{-16} = 1 - 1 + \frac{1}{2}\epsilon_M
\end{equation}

\begin{equation}\label{eq:floating_error_2}
0.1 + (0.2 + 0.3) = 0.600...001 \neq 0.6 = (0.1 + 0.2) + 0.3  
\end{equation}

\section{Methods to Compute $e^{x}$}\label{sec:method_apx}

Computing the exponential function in floating-point arithmetic can present difficulties in terms of both accuracy and speed. This section provides an overview of different techniques for computing the exponential function in floating-point arithmetic, including Taylor series expansion, logarithmic conversion, lookup tables, etc. The advantages and disadvantages of each method will be analyzed and implementation details will be addressed to provide practical guidance.

\subsection{Truncated Product Approximation}\label{sec:method:truncated}

An approximation of the exponential function can be obtained by evaluating the product for a fixed value $m= 2^{D}$.
\begin{equation}
\exp(x) = \lim_{n \rightarrow \infty} \Big(1 + \cfrac{x}{n}\Big)^n \simeq \Big(1 + \cfrac{x}{m}\Big)^m
\end{equation}

This approximation is best for small $x$ but deviates considerably for large $x$.
For positive $x$ the approximation is bounded above by the true exponential function. For negative $x$ the same holds up to a value of $-m$ after which the absolute value of the approximation function increases again, leading to even larger errors. We test two values for $D$, 8 and 10, which yield $m=256$ and $m=1024$.

\subsection{Taylor Series Approximation}\label{sec:method:taylor}

We can use the Taylor series representation to calculate $e^{x}$ to arbitrary precision by evaluating finitely many terms of the Taylor series.
\begin{equation}
e^{x} = \sum_{k=0} \frac{x^k}{k!} = 1 + \frac{x}{1!}+ \frac{x^2}{2!}+ \frac{x^3}{3!} + \cdots
\end{equation}
This method has some limitations: First, it requires exponentiation and we do not have a general power function to evaluate arbitrary exponents of arbitrary bases, and one must take into account that exponentiation is much more expensive than multiplication. Second, there are factorials in the denominators. Dividing by a factorial is highly prone to error and also expensive. This method is numerically unstable, so an alternative way to represent the Taylor series is needed. For that purpose, the Horner's method can be used.

\paragraph{Horner's Method}
For any polynomial $a_{0} + a_{1}x + a_{2}x^2 + \cdots$ we can iteratively factor out the variable $x$ to obtain the equivalent polynomial $a_{0} + x(a_{1} + x(a_{2} + \cdots$. Factoring out terms in order to nest multiplications in this way is known as Horner’s method of polynomial evaluation. When we apply Horner’s method to the Taylor series representation of, we obtain the new form which is much more efficient and stable for implementation:
\begin{equation}
  e^{x} = \sum_{k=0} \frac{x^k}{k!} = 1 + x \Big(1 + \frac{x}{2} \Big( 1 + \frac{x}{3}+ \cdots
\end{equation}

\paragraph{Error Bound}
We can prove an upper bound on the relative error of the Taylor series truncated at the $n^{th}$ term of the series. Let $T_{n}(x)$ be the Taylor series of $e^x$ truncated at the $n^{th}$ term: $e^{x} = T_{n}(x) + \epsilon_{n}(x)$.
The absolute error can be represented using the Lagrange form:
\begin{equation}
  \epsilon_{n}(x) = \frac{e^{(n+1)c}}{(n+1)!}(x-a)^{(n+1)}
\end{equation}
where $e^{(n+1)c}$ denotes the $(n+1)^{th}$ derivative of $e^x$, and $c$ is a real value satisfying $a \le c \le x$.
Since $e^x$ is an increasing function and also is its own derivative, the absolute error is bounded
\begin{equation}
\epsilon_{n}(x) \le e^{x} \frac{x^{n+1}}{(n+1)!}
\end{equation}
and the relative error
\begin{equation}
\mu_{n} = \frac{\epsilon_{n}(x)}{e^{x}} \le \frac{x^{n+1}}{(n+1)!}
\end{equation}
which gives us an upper bound on the relative error of the Taylor series of $e^x$ truncated at the $n^{th}$ term. The main advantage of this strategy is that the accuracy of the exponential function can be controlled by varying $n$. In the limit ($n\rightarrow\infty$) the sum converges to the exact value of the exponential function. However, this approach has a very slow convergence rate for increasing values of $n$, unless $x$ is close to zero.

\paragraph{Precision Bound}
The upper bound on the relative error can be used to determine a lower bound on the number of terms. Suppose one would like to achieve an approximation with a relative error of at most $\mu$. Then we are interested in determining the number $n$ of Taylor terms which will be required to achieve a relative error less than or equal to $\mu$. One can start by finding a $\mu$ to satisfy ${x^n}/{n!} \le \mu$.
By applying logarithms and using the Stirling\footnote{Stirling's approximation: $n! \sim \sqrt{w \pi n}\Big(\cfrac{n}{e}\Big)^n$} approximation, we get to
${xe/}{\mu^{1/n}} \le n$.
As $n$ increases, the denominator approaches $1$, so this becomes asymptotically closer to $xe \le n$.

\paragraph{Implementation Details}

Taylor series approximations provide arbitrary precision at the cost of inefficiency because they converge quite slowly. The size of the relative error at $x$ is proportionate to the distance of $x$ from 0. This is because Taylor series are comparatively accurate near the center of approximation, which in this case is the origin). As $x$ moves away from the origin, the Taylor polynomial requires more terms to achieve the same level of accuracy. 

At the same time, the inherent sparsity of floating point values farther away from the origin works against the approximation, which compounds the overall loss of precision. Thus the approximation becomes increasingly less efficient and less precise as the absolute value of $x$ increases. This algorithm exhibits linear time complexity: as the absolute value of $x$ increases, the number of operations increases by a factor of approximately $36e \approx 100$.

\subsection{Taylor Series Approximation with Range Reduction}\label{sec:method:taylor_range}

We can reduce the area of Taylor approximation to a small neighborhood which is bounded around the origin regardless of the value of $x$. This will trade off some precision in return for significant performance improvements. The basic idea is to decompose $e^x$ by factorizing it as a product of a very small power of $e$ and an integer power of 2, since both of these components are highly efficient to compute: $e^x = e^r \cdot 2^k$, 
where $r$ is a real value satisfying $|r| \le \frac{1}{2} \log 2$ and $k$ is a positive integer. This implies that $x = r + k \log 2$, and thus $k = \lceil \frac{x}{\log 2} - \frac{1}{2} \rceil$ and $r = x - k \log 2$.

With this procedure, we only need 14 Taylor terms to evaluate $e^r$ to 16 digits of precision for any $|r| \le \frac{1}{2} \log 2$, because $r$ is very close to the center $a = 0$. 
The range reduction step achieves a significant increase in efficiency by trading off a modest (single digit) decrease in precision attributable to the left shift step. Whatever error we have in the initial approximation of $e^r$ will be compounded when we multiply $e^r$ by $2^k$: $e^x = 2^k \cdot (e^r - \epsilon)$.

This implementation is significantly faster. This algorithm reduces the number of worst-case operations by three orders of magnitude, from approximately 70000 iterations in the first Taylor algorithm to just 51 in this one. In addition, and as anticipated, we have lost an average of one digit of precision. If the loss of a single digit of precision can be tolerated, this is an excellent performance improvement. The relative error is more predictable and easier to reason about, which is another benefit of range reduction.

\subsection{Lagrange Interpolation}\label{sec:method:lagrange}

Lagrange interpolation is a technique for constructing an approximate function that passes through $n$ given data points (where $n$ also refers to the order of the Lagrange polynomial), and is based on the concept of a Lagrange polynomial
$p_{n}(x) = \sum_{j=0}^{n} y_j \cdot l_j(x)$, where
\begin{equation}
l_j(x) = \frac{(x-x_0)\cdots(x-x_{j-1})(x-x_{j+1})\cdots(x-x_n)}{(x_j-x_0)\cdots(x_j-x_{j-1})(x_j-x_{j+1})\cdots(x_j-x_n)}=\frac{\prod_{k=0, k \neq j}^{n} (x - x_k)}{\prod_{k=0, k \neq j}^{n} (x_j - x_k)}
\end{equation}
  
Lagrange interpolation can be used in conjunction with the range reduction algorithm for exponential computation by first determining a set of discrete data points for the exponential function of interest. In particular, this method first performs a range reduction step by applying the identity $e^x = e^r \cdot 2^k$ for $|r| \le \frac{1}{2} \log 2$ and integer $k$. The Lagrange interpolant approximates the value of $e^r$, then the result is multiplied by $2^k$ to obtain the final approximation for $e^x$.

When using a polynomial of a higher order, the approximate result quickly converges to the exact solution; in fact, if we reduce the distance between the interpolation points by n times, the error decreases by $2^{n+1}$. However, the number of points that need to be evaluated grows in tandem with the order of the polynomial, making the process increasingly inefficient as more points are added.

\subsection{Lagrange Barycentric Form}\label{sec:method:lagrange_barycentric}

Using the barycentric form of the Lagrange interpolation can provide benefits in terms of computation cost and accuracy. Barycentric interpolation is a variation of Lagrange interpolation that is based on barycentric coordinates, and it is more numerically stable, faster, and less sensitive to the distribution of data points.

\begin{equation}
  p_{n}(x) = \cfrac{\prod_{j=0}^{n} \cfrac{w_{j}}{x - x_j}\cdot y_{j}}{\prod_{j=0}^{n} \cfrac{w_{j}}{x - x_j}}
\end{equation}

The weights $w_{j}$ are the same for all values of $x$, so this is calculated once then saved, which takes $\mathcal{O}(n^2)$ time for $n + 1$ of distinct interpolation points. The interpolated member function then uses the weights and $y_j$ values to approximate the function. With the weights pre-calculated, each evaluation of $x$ takes only $\mathcal{O}(n)$ time.



\subsection{Chebyshev Interpolation}\label{sec:method:chebyshev}

Chebyshev interpolation is a method for constructing a polynomial that passes through a given set of points, similar to Lagrange interpolation, but has several advantages over the latter: First, the interpolation points are the roots of Chebyshev polynomials defined as $p_{n}(x) = \cos(n \cdots \text{arccos}(x))$, which are specially chosen to minimize the approximation error. The Chebyshev nodes $x_{k}$ are the roots of these polynomials and are the points where the polynomial interpolant is guaranteed to have the least error:
\begin{equation}
  x_{k} = \cos \Big( \cfrac{2k-1}{n} \cdot \cfrac{\pi}{2} \Big) \quad \forall k=1,\ldots,n
\end{equation}
In addition, Chebyshev polynomials are less sensitive to perturbations in the data and can be faster than Lagrange interpolation, especially for large sets of data. We use the Chebyshev basis with degree 13 and the monomial basis also with degree 13.

\subsection{Lookup Tables}\label{sec:method:lookup}
The exponential function can be analytically converted in a base-two expression and subsequently decomposed in its integer $x_i$ and fractional $x_f$ parts: $e^{x} = 2^{x \log_2(e)} = 2^{x_{i}+x_{f}}$

Different algorithms based on one or more look-up tables can be used to evaluate $2^{x_{i}}$ and $2^{x_{f}}$. Look-up tables are in general faster than Taylor expansions, as they are based on linear interpolation, however they do not fully exploit floating-point arithmetic and cannot be implemented using only SIMD\footnote{SIMD (Single Instruction, Multiple Data) instructions are a type of computer instruction that allows a single operation to be performed on multiple pieces of data at the same time. SIMD instructions are supported by many modern processors, including Intel's SSE, AVX and ARM's NEON.} instructions, due to the presence of conditional statements.

\subsection{Hart Method}\label{sec:method:hart}

To compute exponentials, this method multiplies $x$ by $log_{2}(e)$ to compute $2^{log_{2}(e)x}$ rather than $e^x$.
\cite{hart1978computer} lists polynomials for 46 variants of $2^x$, with precisions ranging from 4 to 25 decimal digits. 
One of them assumes the input argument is between 0 and 0.5, and uses the following ratio of two polynomials: 
\begin{equation}
  2^{x} = \cfrac{Q(x) + x \cdot P(x)}{Q(x) - x \cdot P(x)}
\end{equation}
where $P(x)$ and $Q(x)$ are different polynomials. When different coefficients are used, the precision of the resulting decimal digits varies.

\subsection{Schraudolph Method}\label{sec:method:schraudolph}

This method \cite{schraudolph1999fast} exploits the definition of the IEEE 754 floating point format. Schraudolph developed a technique to take advantage of the $2^x$ power implied in the standard floating-point representation. To apply this method to compute exponentials, we multiply $x$ by $log_{2}(e)$ to compute $2^{log_{2}(e)x}$ rather than $e^x$.

Using $x_i$ and $x_f$ to denote the integer and fractional part of $x=x_i + x_f$, one can write: $2^{x} = 2^{x_i} \cdot 2^{x_f}$. Given a binary representation of an integer $x_i$, the 2-exponential $y=2^x$ can be computed simply by copying the bits representing $x_i$ into the exponent bits of $y$. 
Thus, we shift the exponent by the number of bits required to obtain the integer part of the exponential (i.e., $2^{x_{i}}$), and then we approximate the fractional part  $2^{x_{f}}$ with $(1 + m)$, where $m$ is the mantissa in the IEEE-745 standard. This approach can be summarized in three steps: First, store the manipulated input exponent $x$ in a 32-bit integer variable $\text{int } i = A \cdot x + B - C$, with $A = S / \log{2}$, $B = S \cdot 1023$, $C=60801$, being $S = 2^{20}$ the shift factor. Second, concatenate the 32-bit integer $i$ with another 32-bit integer $j$ to form a 64 bit line. Finally, interpret the 64 bit line as a double-precision float-point number, which coincides with the $e^x$ approximation.

The approximation developed by Schraudolph relies on simple arithmetics, which consists of a single floating-point multiply-add, where A, B, and C are pre-computed constants. However, the accuracy of the approximation is very low, i.e., only a single-digit, even though the C constant has been computed to minimize the RMS relative error.

We also extend the algorithm proposed by \textbf{Schraudolph with a lookup-table (LUT)} to fast compute $2^{x_i}$. In this regard, we only need a look-up table with the constant values of successive square-roots of 2, then operate consecutive multiplying of a selection of those numbers while scanning the mantissa of $x$.

\subsection{Malossi Method}\label{sec:method:malossi}

An extension of the \cite{schraudolph1999fast} method by \cite{malossi2015fast}. Here we also use $x_i$ and $x_f$ to denote the integer and fractional part of $x=x_i + x_f$. Based on this, one can write $2^{x} = 2^{x_i} \cdot 2^{x_f} = (1.0 + \mathcal{K}(x_{f})) \cdot 2^{x_i}$.
Thus, we shift the exponent by the number of bits required to obtain the integer part of the exponential (i.e., $2^{x_{i}}$), and then we approximate the fractional part  $2^{x_{f}}$ with $(1 + \mathcal{K}(x_{f}))$. 

The $\mathcal{K}(x_{f})$ is a correction function that interpolates between the integer exponentials $2^{x_i}$ and $2^{x_{i} + 1}$. The exact form of the correction form is given by $\mathcal{K}(x_{f}) = 2^{x_f} - 1.0$. If there was an efficient way of computing $\mathcal{K}(x_{f})$ for values $x_f \in [0,1]$, it would be possible to compute the exact value of the exponential function. However, since this is not the case, $\mathcal{K}$ is approximated using a polynomial fit. The choice of the degree of the polynomial fit determines how well the exponential function is approximated and how many operations are necessary to approximate the exponential.

To apply this method to compute exponentials, we multiply $x$ by $log_{2}(e)$ to compute $2^{log_{2}(e)x}$ rather than $e^x$. The author uses the Remez algorithm to compute polynomials: the process begins with an initial approximation, and then iteratively improves the approximation by exchanging the roots of the polynomial until a satisfactory approximation is achieved. 
This approximation method provides much smaller errors over the complete range of values. For instance, with a polynomial fit of degree five, the approximation has almost recovered machine precision for single precision arithmetic.

\subsection{Open-source Implementations}

In this section, we will explore some open-source libraries that provide implementation for computing the exponential function. These libraries offer a wide range of precision and speed trade-off, making them suitable for different use cases.

\subsubsection{\textit{fdlimb} library}\label{sec:method:fdlibm}

We test the exponential implementation of \textit{fdlimb} \cite{fdlibm}, a \textit{C} math library that supports IEEE 754 floating-point arithmetic. The exponential is estimated as follows:
\begin{enumerate}
  \item Argument reduction: Reduce $x$ to an $r$ so that $|r| \leq 0.5 \cdot \ln 2 \sim 0.34658$.	That is, given $x$, find $r$ and integer $k$ such that $x = k \cdot \ln 2 + r$, subject to $|r| \leq 0.5 \cdot \ln 2 \sim 0.34658$.
  \item Approximation of $e^r$ by a special rational function on the interval $[0,0.34658]$. Given $R(r^2) = r(e^r+1)/(e^r-1) = 2 + r^2/6 - r^4/360 + \ldots$. They use a special Remes algorithm on $[0,0.34658]$ to generate a polynomial of degree 5 to approximate $R$.
  \item Scale back to obtain $e^x$, i.e., $e^x = 2^k \cdot e^r$.
\end{enumerate}

\clearpage
\subsubsection{\textit{simdgen} library}\label{sec:method:simdgen}

To compute exponentials, the \textit{simdgen} library \cite{simdgen} method multiplies $x$ by $log_{2}(e)$ to compute $2^{log_{2}(e)x}$ rather than $e^x$. Let $y = x \log_2(e)$ and split $y$ into an integer $n$ and a fraction $a$, where $|a| \leq 0.5$. One can get $e^x = 2^n 2^a$, where $2^n$ can be easily computed so we consider $2^a$. Let $2^a = e^{a \log 2} = e^b$ where $b = a \log 2$. If $|a| \leq 0.5$ and $\log 2 = 0.693\ldots$ then $|b| = |a \log 2| \leq 0.346$.

Then, they use a Maclaurin series of $e^x = 1 + x + x^2/2 + x^3/6 + \ldots$. The term of degree 6 is about $0.346^6/6! = 2.4\cdot10^{-6}$, which is close to single floating precision. As a result, $e^b = 1 + b + b^2/2! + b^3/3! + b^4/4! + b^5/5!$. In our implementation we test this series with 5, 6, and 7 degrees.

\subsubsection{\textit{fmath} library}\label{sec:method:fmath}

While the \textit{fmath} library uses a process similar to that of other libraries, there are a few key distinctions:
First, it splits $x$ into two values $s$ and $t$, such that $x=s+t$, and thus $e^x=e^s \cdot e^t$
If $t$ is very small, its exponential can be computed using the Maclaurin series with very few terms and high precision. In fact, \textit{fmath} computes this $e^t$ with just three terms $n=3$: $e^t = 1 + t + \frac{t^2}{2!} + \frac{t^3}{3!}$. 
The discretization or truncation absolute error is bounded by:
\begin{equation}
\epsilon_{n}(t) \le e^{t} \frac{t^{n+1}}{(n+1)!} = e^{t} \frac{t^{4}}{24}  
\end{equation}
Given that $t$ is very near to zero, we can approximate this expression as $\epsilon_{n}(t) \le \frac{t^{4}}{24}$, because $e^t \approx 1$ if $t \approx 0$.
To not exceed the machine epsilon of double float precision ($2^{-52} \approx 2.2 \cdot 10^{-16}$), we would need $t$ to be smaller than:
\begin{equation}
t < (24 \cdot 2.2 \cdot 10^{-16})^{1/4} \approx 2.70 \cdot 10^{-4} \approx \frac{1}{3701}
\end{equation}
Thus, we can choose a more appropriate value that guarantees $t < 1/3701$, such as $t = 1/2^{12} = 1/4096$.

Second, let $y = K \cdot x$ and split it as previously $y = K(s+t) = K \cdot s + K \cdot t = a + b$, where $a = round(y)$ is an integer and $b$ is a fraction $|b| \le 1/2$. Then, $x = s + t = a/K + b/K$.

$$e^x=e^s \cdot e^t = e^{a/K} \cdot e^{b/K}$$

If $K = 2048/\log(2) \approx 2954.6$ then $|b/K| = |b|/K \le 1/6000$. Thus, we have a $t = b/K$ value whose exponential function can be computed with just 3 terms of the Maclaurin series, because $|t| \le 1/4096$.
Finally, we need to compute the exponential value of $s=a/K$:

$$e^s = e^{a/K} = e^{a \cdot \log{2} / 2048} = 2^{a / 2048}$$

To compute this power of two, we split $n/2048$ as follows:  $2^{a / 2048} = 2^{q} \cdot 2^{r}$, where $q = \text{int}(a / 2048)$ and $r = (a/2048) - q$. Given that $q$ is an integer, $2^q$ can be computed by bit shift (taking advantage of the IEEE floating number architecture). Given that there are only 2048 possible values for $r$ (from 0 to 2047) we apply a lookup table to compute $2^r$. $r$ is in the semi-open interval $[0, 1)$ and can take values in $[0, 2048) / 2048$ so we need to pre-compute 2048 values for $2^r \in [1,2)$. To create the lookup table, we use the power function included in the \textit{C++} standard math library. Hence, using this method, the error is only due to the truncation error of the Maclaurin series.

\subsubsection{\textit{std:math} library}\label{sec:method:stdmath}

The implementation uses a combination of mathematical algorithms and approximations to calculate the value of $e^x$ with a high degree of accuracy.
For single precision it is basically an order 3 polynomial approximation. The input value $x$ is written as
\begin{equation}
  x = n \cdot \ln 2 + \cfrac{t}{512} + \delta[t] + y
\end{equation}

where $n$ is an integer $-150 \leq n \leq 127$, $t$ is an integer $-177 \leq t \leq 177$, $\delta$ is based on a table entry ($\delta[t] < 2^{-28}$) and $y$ is whatever is left, $|y| < 2^{-10}$.
Then $e^x$ is approximated as 

\begin{equation}
  e^x = 2^n \cdot e^{t/512 + \delta[t]} \cdot ( 1 + p(x + \delta[t] + n * \ln 2) - \delta[t] )
\end{equation}
where $p(x)$ is a polynomial approximating $e^x-1$, and $e^{t/512 + \delta[t]}$ is obtained from a table. The table used is the same one as for the double precision. The implementation also takes into account the floating-point precision of the input and output values and applies appropriate error handling techniques such as checking for overflow and underflow.

\section{Benchmark}

In this section, we present a benchmark analysis of various algorithms implemented in \textit{C++} for computing the exponential function. The goal of this study is to evaluate the efficiency and accuracy of different implementations of the exponential function and to identify the best-performing algorithm. The benchmark experiments were conducted on a set of test cases with varying input values and the results were analyzed using performance metrics such as execution time and relative error. 

\subsection{Setup}

The reviewed methods in \cref{sec:method_apx} for the exponential function $e^x$ were implemented in \textit{C++} for $x$ in the domain \textcolor{red}{$[-87,87]$} and \textcolor{blue}{$[-708,708]$} to avoid the \textcolor{red}{single} and \textcolor{blue}{double} precision underflow and overflow limits. In particular, we generate $10^6$ points in the corresponding domain, and repeat the computation 100 times. 

We only use functions available in \textit{C++} standard library and avoid any specific, microprocessor dependent instruction set. See \cref{tab:exponential} for more information about each implemented function and its reference to the corresponding section. The native \textit{math:exp} function provided by the \textit{C++} math library is used as a reference to compare against other methods, both for computation time and exact value.
The Chrono\footnote{\url{https://en.cppreference.com/w/cpp/chrono}} library is used for measuring the time benchmark, and the Fstream\footnote{\url{https://en.cppreference.com/w/cpp/io/basic_ofstream}} library to save the obtained results to csv format. 

\begin{table}[!htb]
  \centering
  \small
  \setstretch{1.3}
  \begin{tabular}{llll}
  \toprule
  Function Name        & Method                                                                                                         & Degree & Reference                             \\ \midrule
  Truncated f256       & Truncated product approximation                                                                                & 8      & \ref{sec:method:truncated}            \\
  Truncated f1024      & Truncated product approximation                                                                                & 10     & \ref{sec:method:truncated}            \\
  Taylor               & Taylor range reduction                                                                                         & 3-14   & \ref{sec:method:taylor_range}         \\
  Lagrange Barycentric & \setstretch{0.8}\begin{tabular}[t]{@{}l@{}}Taylor range reduction +\\ Lagrange barycentric interpolation\end{tabular}          & 15     & \ref{sec:method:lagrange_barycentric} \\
  Chebyshev            & \setstretch{0.8}\begin{tabular}[t]{@{}l@{}}Taylor range reduction +\\ Chebyshev interpolation\end{tabular}                     & 13     & \ref{sec:method:chebyshev}            \\
  Chebyshev Monomial   & \setstretch{0.8}\begin{tabular}[t]{@{}l@{}}Taylor range reduction +\\ Chebyshev monomial interpolation\end{tabular} & 13     & \ref{sec:method:chebyshev}            \\
  Hart                 & Hart method                                                                                                    & -      & \ref{sec:method:hart}                 \\
  Schraudolph          & Schraudolph method                                                                                             & 3      & \ref{sec:method:schraudolph}          \\
  Schraudolph + LUT    & Schraudolph method +  Lookup table                                                                             & -      & \ref{sec:method:schraudolph}          \\
  Malossi              & Malossi method                                                                                                 & 3-11   & \ref{sec:method:malossi}              \\
  \textit{fdlibm}      & \textit{fdlibm} library implementation                                                                                  & 5      & \ref{sec:method:fdlibm}               \\
  \textit{simdgen}     & \textit{simdgen} library implementation                                                                                 & 5,6,7  & \ref{sec:method:simdgen}              \\
  \textit{fmath}       & \textit{fmath} library implementation                                                                                   & 3      & \ref{sec:method:fmath}                \\ \bottomrule
  \end{tabular}%
  \caption{Exponential functions implemented in \textit{C++} for benchmark}
  \label{tab:exponential}
  \end{table}

\subsection{Results}

\cref{fig:exponential:single,fig:exponential:double} show the computation time and median relative error for each method, in comparison with the native \textit{math:exp} function. The results of the study show that the best methods in terms of \textbf{execution time} were: truncated product approximation (\ref{sec:method:truncated}), Schraudolph methods (\ref{sec:method:schraudolph}) and fmath library (\ref{sec:method:fmath}). On the other hand, the best methods in terms of \textbf{relative error} were the Taylor series (\ref{sec:method:taylor_range}) of degree 12,13,14, and fblibm (\ref{sec:method:fdlibm}) and fmath (\ref{sec:method:fmath}) open source implementations. Overall, the \textbf{fmath} library obtained the best trade-off between execution time and relative error for both single and double precision, and as a result was selected to be implemented in the package Diffeomorphic Fast Warping (DIFW)\footnote{\url{https://github.com/imartinezl/difw}}.

\begin{landscape}

  \subsubsection{Single precision}
  \vspace{-0.75cm}
  \begin{figure}[!htb]
    \begin{center}
    \begin{subfigure}[t]{0.43\columnwidth}
        \centering
        \includegraphics[width=\linewidth]{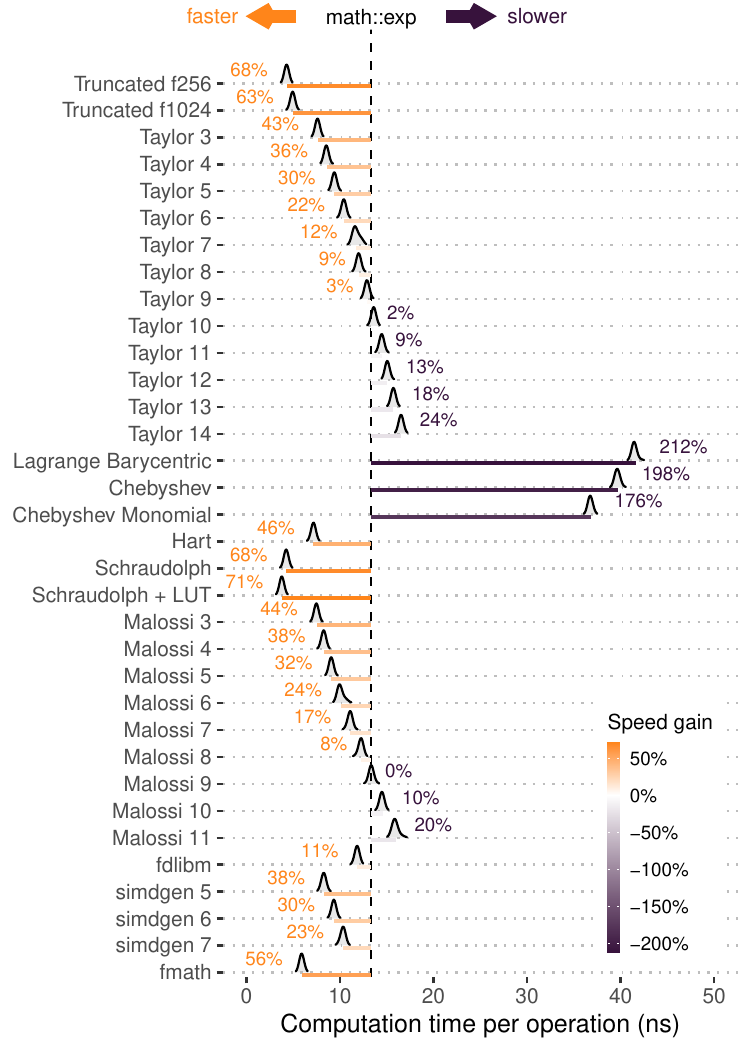}
        \caption{Computation time.}
        \label{fig:exp_float_time_dist}
    \end{subfigure}
    \hfill
    \begin{subfigure}[t]{0.43\columnwidth}
      \centering
      \includegraphics[width=\linewidth]{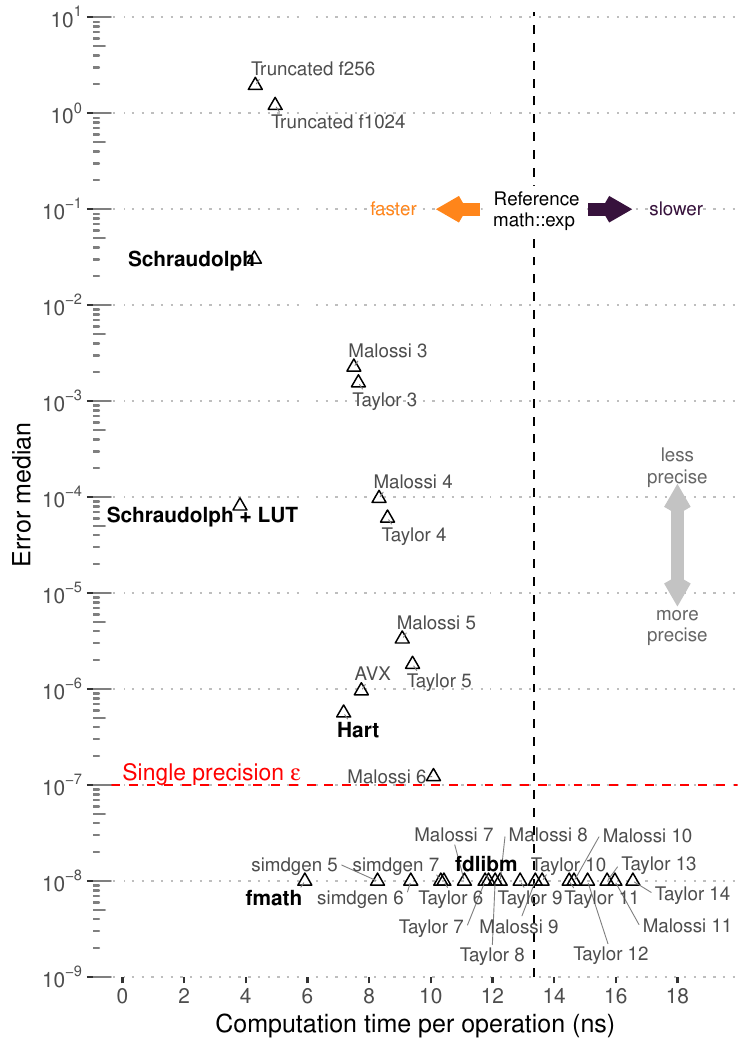}
      \caption{Median relative error vs computation time.}
      \label{fig:exp_float_time_vs_error}
    \end{subfigure}
    \end{center}
    \vspace{-0.75cm}
    \caption{Benchmark results: single precision}
    \label{fig:exponential:single}
  \end{figure}

  \subsubsection{Double precision}
  \vspace{-0.75cm}
  \begin{figure}[!htb]
    \begin{center}
    \begin{subfigure}[t]{0.43\columnwidth}
        \centering
        \includegraphics[width=\linewidth]{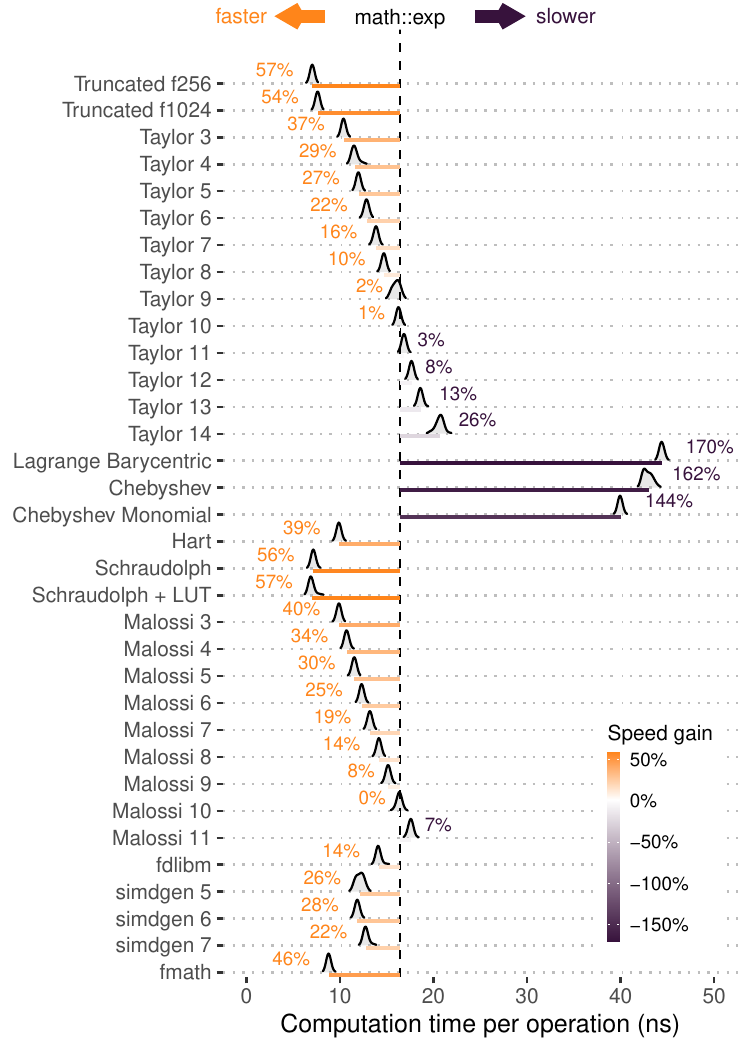}
        \caption{Computation time.}
        \label{fig:exp_double_time_dist}
    \end{subfigure}
    \hfill
    \begin{subfigure}[t]{0.43\columnwidth}
      \centering
      \includegraphics[width=\linewidth]{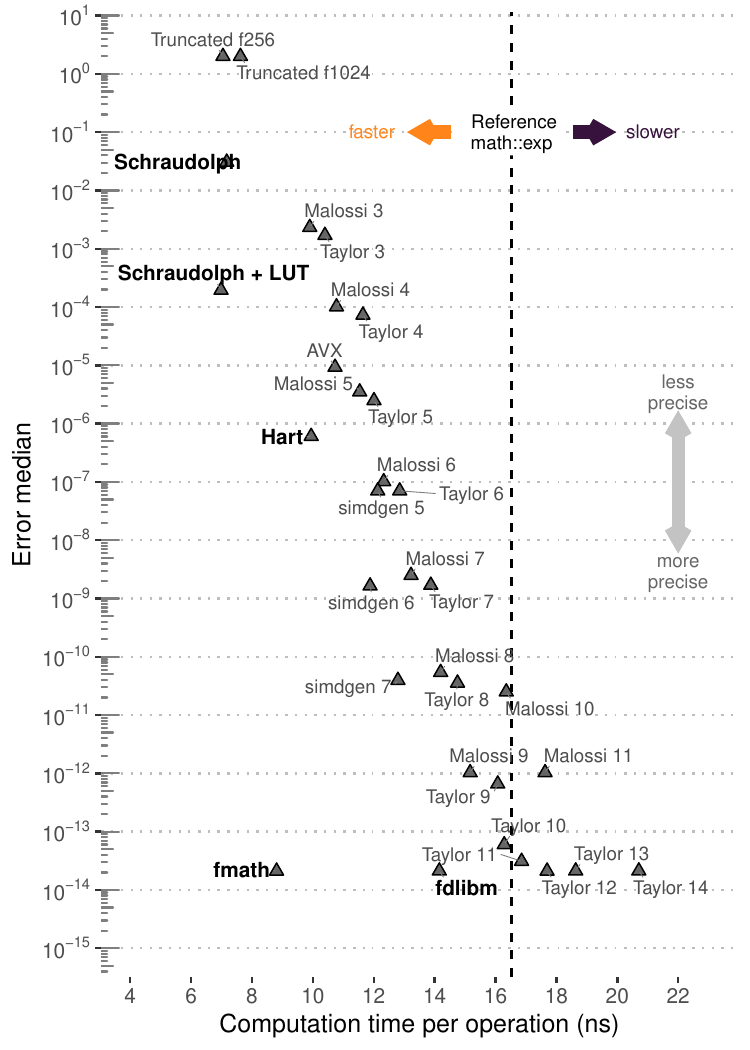}
      \caption{Median relative error vs computation time.}
      \label{fig:exp_double_time_vs_error}
    \end{subfigure}
    \end{center}
    \vspace{-0.75cm}
    \caption{Benchmark results: double precision}
    \label{fig:exponential:double}
  \end{figure}

\end{landscape}

\end{appendices}

\begin{spacing}{0.9}


\cleardoublepage
\bibliography{
    references/introduction,
    references/data_science_methodologies,
    references/diffeomorphisms,
    references/time_series_averaging,
    references/time_series_classification,
    references/time_series_clustering,
    references/registration,
    references/stream_clustering,
    references/concept_drift,
    references/normalizing_flows,
    references/appendix
    } 
\bibliographystyle{icml2022}



\end{spacing}




\end{document}